\documentclass[10pt,twocolumn,letterpaper]{article}
\usepackage[T1]{fontenc}
\usepackage{textcomp}

\usepackage{listings,lipsum,xcolor}
\definecolor{background}{HTML}{EEEEEE}
\definecolor{delim}{RGB}{20,105,176}
\colorlet{numb}{magenta!60!black}

\usepackage{etoolbox}
\usepackage[pdftex]{graphicx}
\usepackage{cuted}
\usepackage{dblfloatfix} 
\usepackage{afterpage} 
\usepackage{caption}
\usepackage{float} 
\usepackage{algorithm2e}
\RestyleAlgo{ruled}
\usepackage{comment}
\usepackage{placeins}

\usepackage{multirow,bigdelim}
\usepackage{booktabs} %
\usepackage{enumitem} %
\usepackage[skip=0.033\baselineskip]{subfig}
\usepackage{makecell}
\usepackage{xcolor}
\usepackage{tikz}
\usepackage{pifont}
\usepackage{svg}
\newif\ifcomments
\commentstrue
\ifcomments
    \providecommand\yuval[1]{[\textcolor{blue}{Yuval: {#1}}]}
    \providecommand\shelly[1]{[\textcolor{purple}{Shelly: {#1}}]}
    \providecommand\adam[1]{[\textcolor{magenta}{Adam: {#1}}]}
    \providecommand\amit[1]{[\textcolor{violet}{Amit: {#1}}]}
    \providecommand\uriel[1]{[\textcolor{green}{Uriel: {#1}}]}
    \providecommand\yaniv[1]{[\textcolor{red}{Yaniv: {#1}}]}
    
    \providecommand\devi[1]{[\textcolor{purple}{Devi: {#1}}]}
\else
    \providecommand{\yuval}[1]{}
    \providecommand{\shelly}[1]{}
    \providecommand{\adam}[1]{}
    \providecommand{\amit}[1]{}
    \providecommand{\uriel}[1]{}
    \providecommand{\devi}[1]{}
    \providecommand{\yaniv}[1]{}\providecommand{\reviewer}[1]{}
\fi

\newcommand{\model}{Emu Edit\xspace}

\newcommand{\ntasks}{sixteen\xspace}
\newcommand{\ntaskstest}{seven\xspace}
\newcommand{\myparagraph}[1]{\noindent\textbf{#1}}

\newcommand{\equalcontribution}{\textsuperscript{*}} %

\usepackage[pagenumbers]{cvpr} %

\newif\ifreview

\definecolor{cvprblue}{rgb}{0.21,0.49,0.74}
\usepackage[pagebackref,breaklinks,colorlinks,citecolor=cvprblue]{hyperref}

\def\paperID{*****} %
\def\confName{CVPR}
\def\confYear{2024}

\begin{document}
\title{\model: Precise Image Editing via Recognition and Generation Tasks}

\author{Shelly Sheynin\equalcontribution, Adam Polyak\equalcontribution, Uriel Singer\equalcontribution, Yuval Kirstain\equalcontribution, Amit Zohar\equalcontribution, Oron Ashual,\\
Devi Parikh and Yaniv Taigman\\
\\
GenAI, Meta\\
}

\lstset{
    language=R,
    basicstyle=\footnotesize\ttfamily,
    numbers=left,           
    numberstyle=\tiny,       
    numbersep=5pt,        
    tabsize=4,                 
    extendedchars=true,         
    breaklines=true,
    keywordstyle=\textbf,           
    stringstyle=\color{white}\ttfamily,
    showspaces=false,       
    showtabs=false,           
    xleftmargin=17pt,
    framexleftmargin=17pt,
    framexrightmargin=5pt,
    framexbottommargin=4pt,
    showstringspaces=false, 
    morestring=[b]"
}

\twocolumn[{%
\renewcommand\twocolumn[1][]{#1}%
\maketitle

\begin{center}
    \vspace{-1.0em}
    \centering
    \captionsetup{type=figure}
\includegraphics[width=0.163\linewidth]{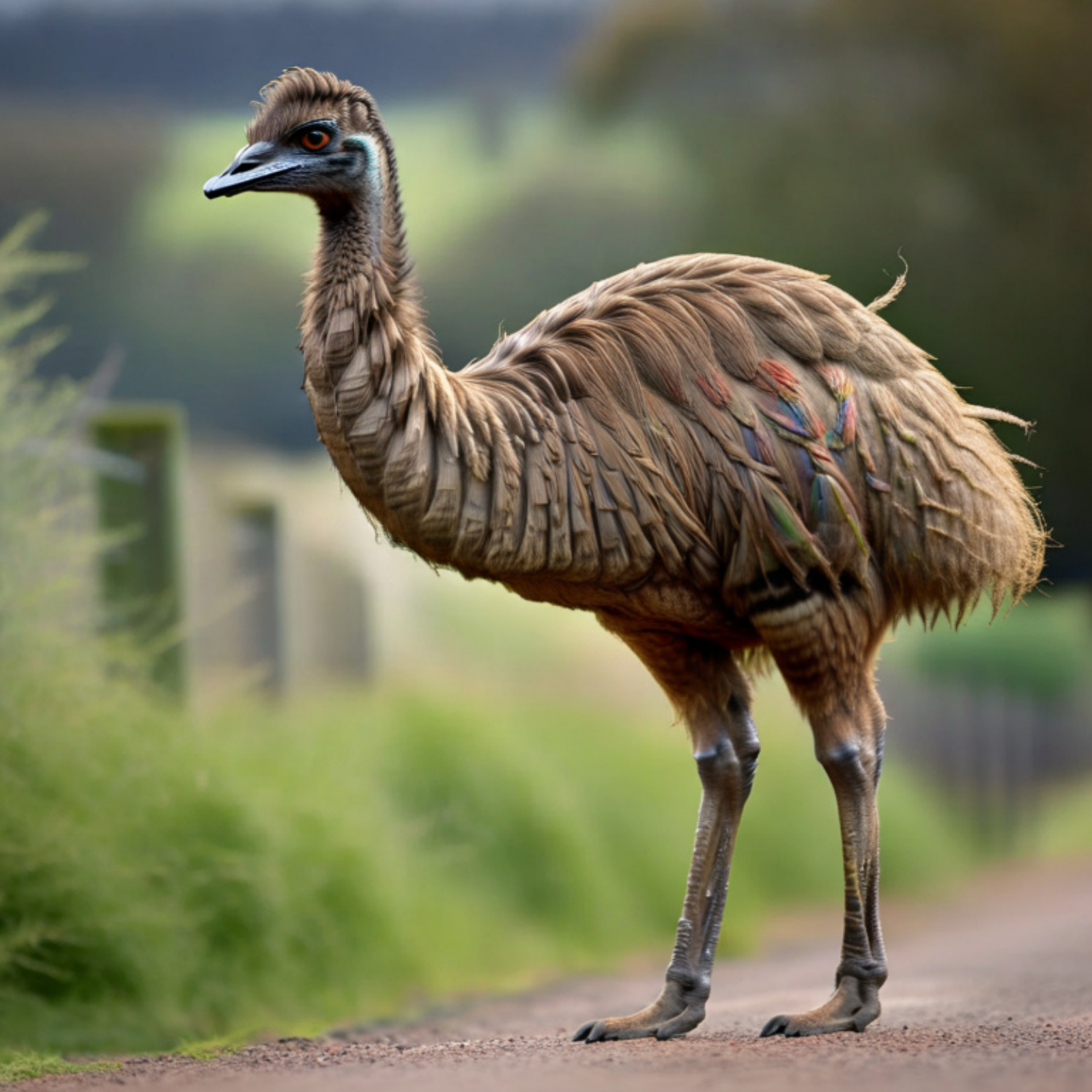}
\hspace{-0.08cm}\includegraphics[width=0.163\linewidth]{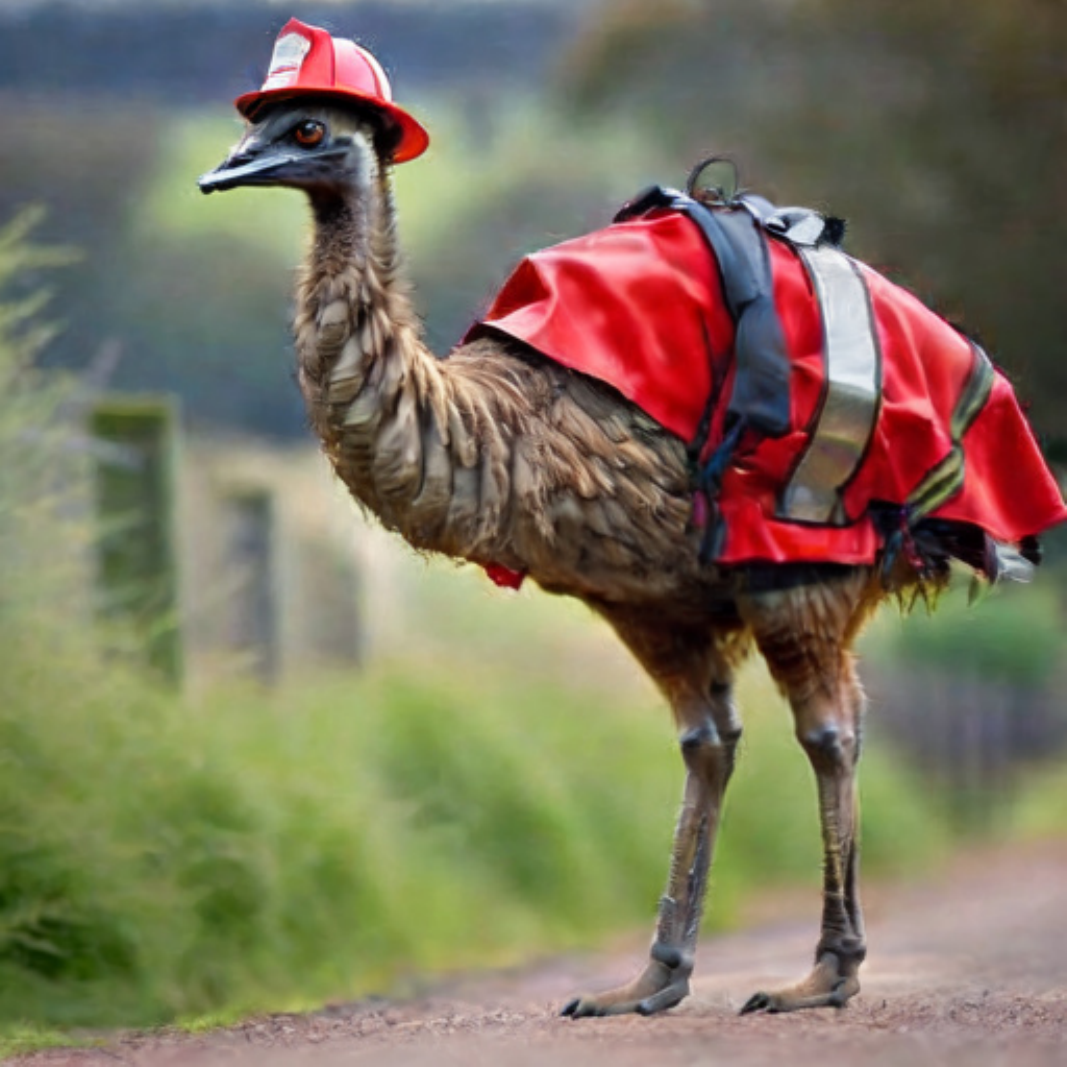}
\includegraphics[width=0.163\linewidth]{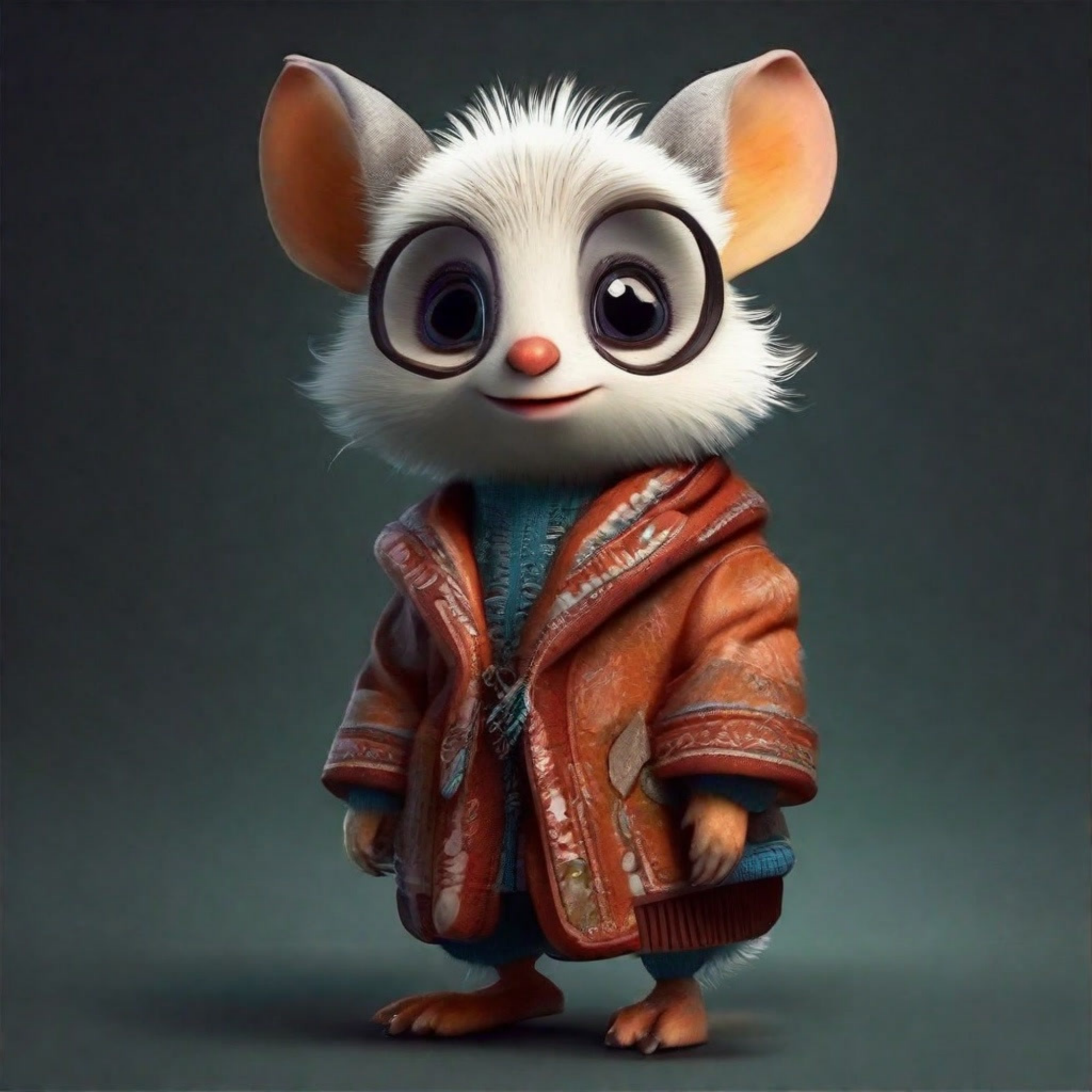}
\hspace{-0.085cm}\includegraphics[width=0.163\linewidth]{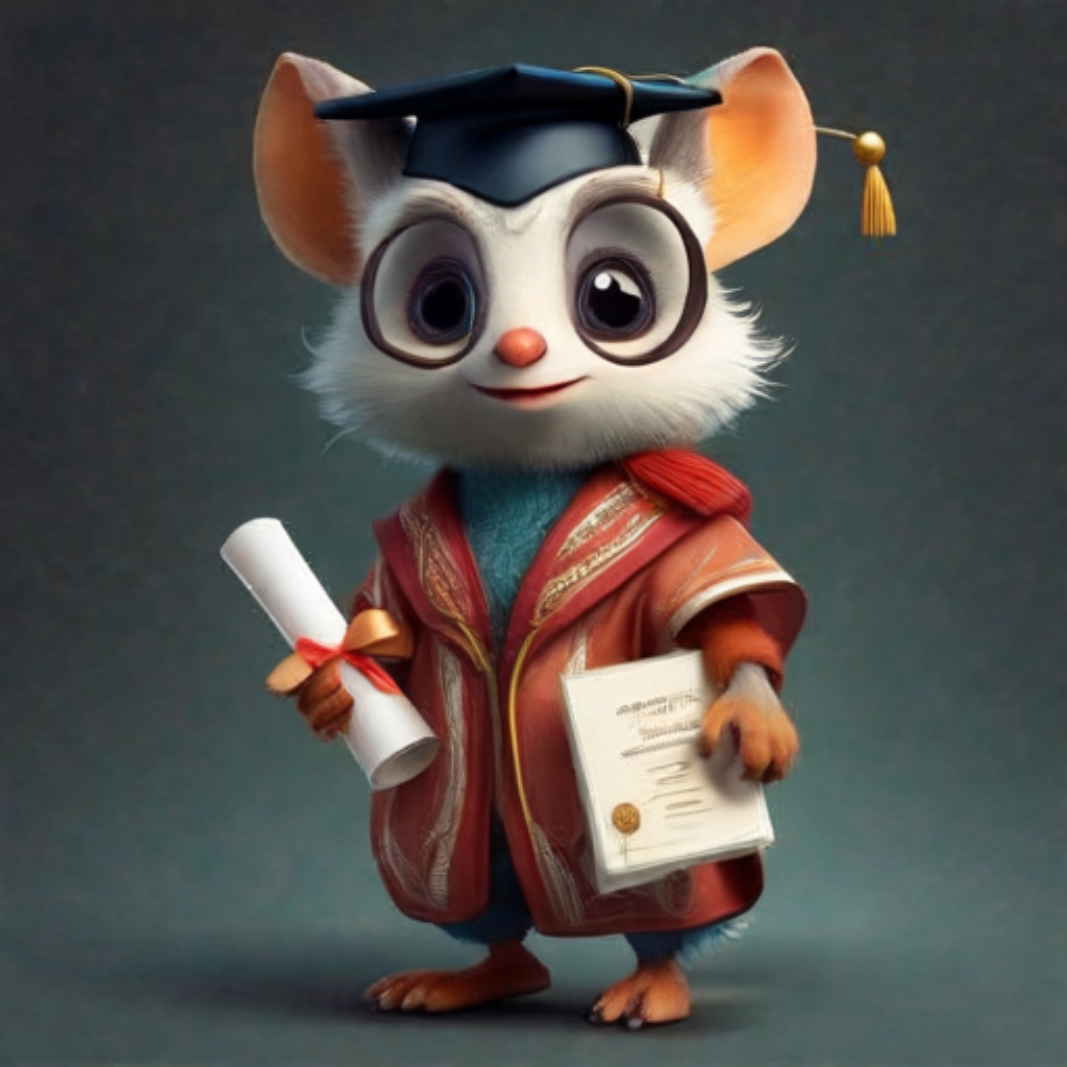}
\includegraphics[width=0.163\linewidth]{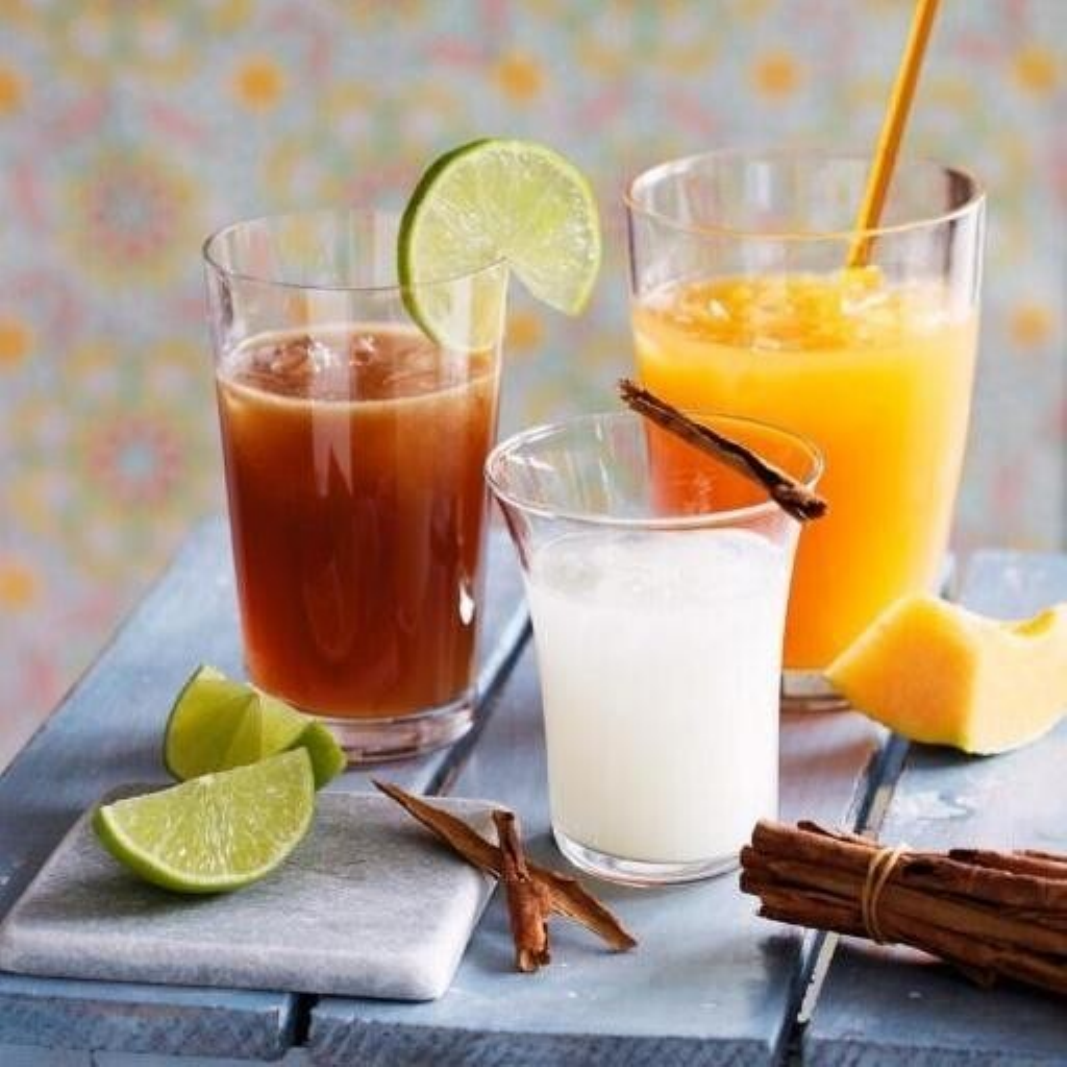}
\hspace{-0.08cm}\includegraphics[width=0.163\linewidth]{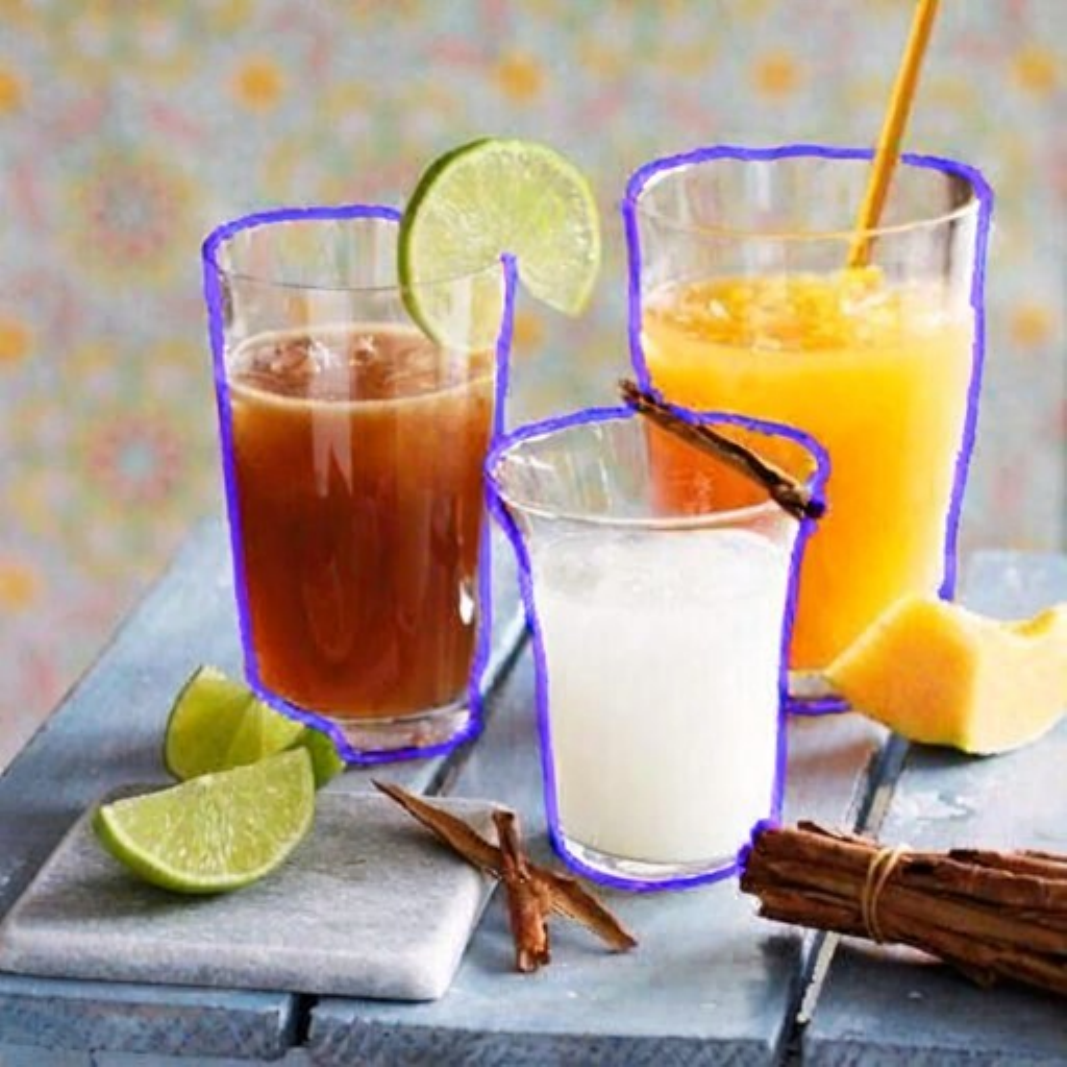}
\scriptsize{\hspace{1.0cm}\textit{Dress the emu with a fireman outfit} \hspace{2.9cm} \textit{Let's see it graduating} \hspace{3.5cm} \textit{Mark the drinks}\hspace{0.7cm}}
\captionof{figure}{\model is a multi-tasking model that combines various editing (left, middle) and vision (right) tasks for precise image editing.}
    \label{fig:tcandidate}
\end{center}%

}]

\begin{abstract}
\vspace{-0.3cm}
Instruction-based image editing holds immense potential for a variety of applications, as it enables users to perform any editing operation using a natural language instruction.
However, current models in this domain often struggle with accurately executing user instructions. 
We present \model, a multi-task image editing model which sets state-of-the-art results in instruction-based image editing.
To develop \model we train it to multi-task across an unprecedented range of tasks, such as region-based editing, free-form editing, and Computer Vision tasks, all of which are formulated as generative tasks.
Additionally, to enhance \model's multi-task learning abilities, we provide it with learned task embeddings which guide the generation process towards the correct edit type.
Both these elements are essential for \model's outstanding performance.
Furthermore, we show that \model can generalize to new tasks, such as image inpainting, super-resolution, and compositions of editing tasks, with just a few labeled examples.
This capability offers a significant advantage in scenarios where high-quality samples are scarce.
Lastly, to facilitate a more rigorous and informed assessment of instructable image editing models, we release a new challenging and versatile benchmark that includes \ntaskstest different image editing tasks.\footnote{Project Page: \url{https://emu-edit.metademolab.com/}}
\end{abstract}

\ifreview
\else
\renewcommand{\thefootnote}{\fnsymbol{footnote}}  %
\footnotetext[1]{Equal contribution.}
\renewcommand{\thefootnote}{\arabic{footnote}}  
\fi

\begin{figure*}[hbt!]
    \centering
      \subfloat[\textit{A cat}]%
  {\includegraphics[width=0.199\linewidth]{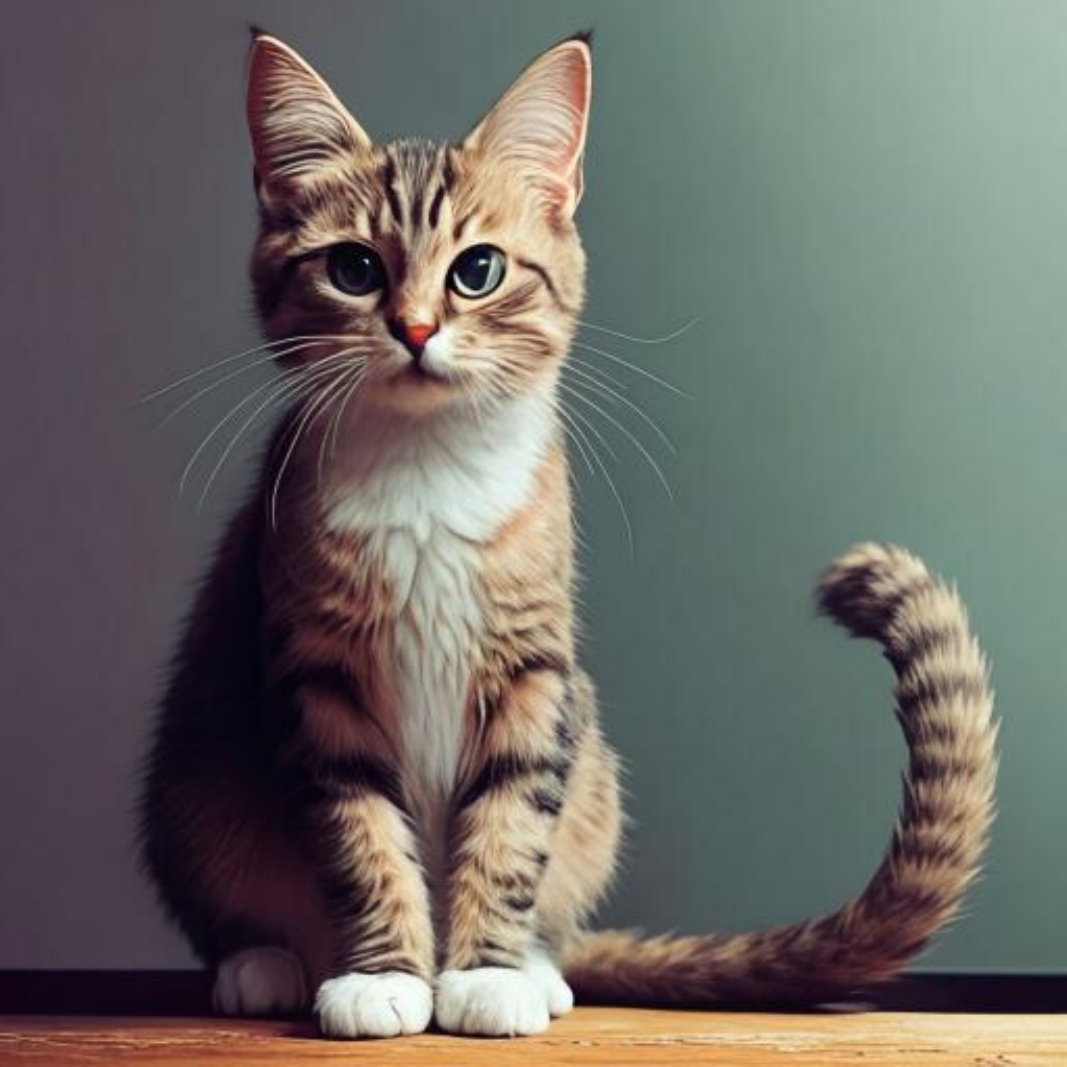}}%
          \hspace{0.2mm}%
        \subfloat[\textit{Remove the tail}]%
  {\includegraphics[width=0.199\linewidth]{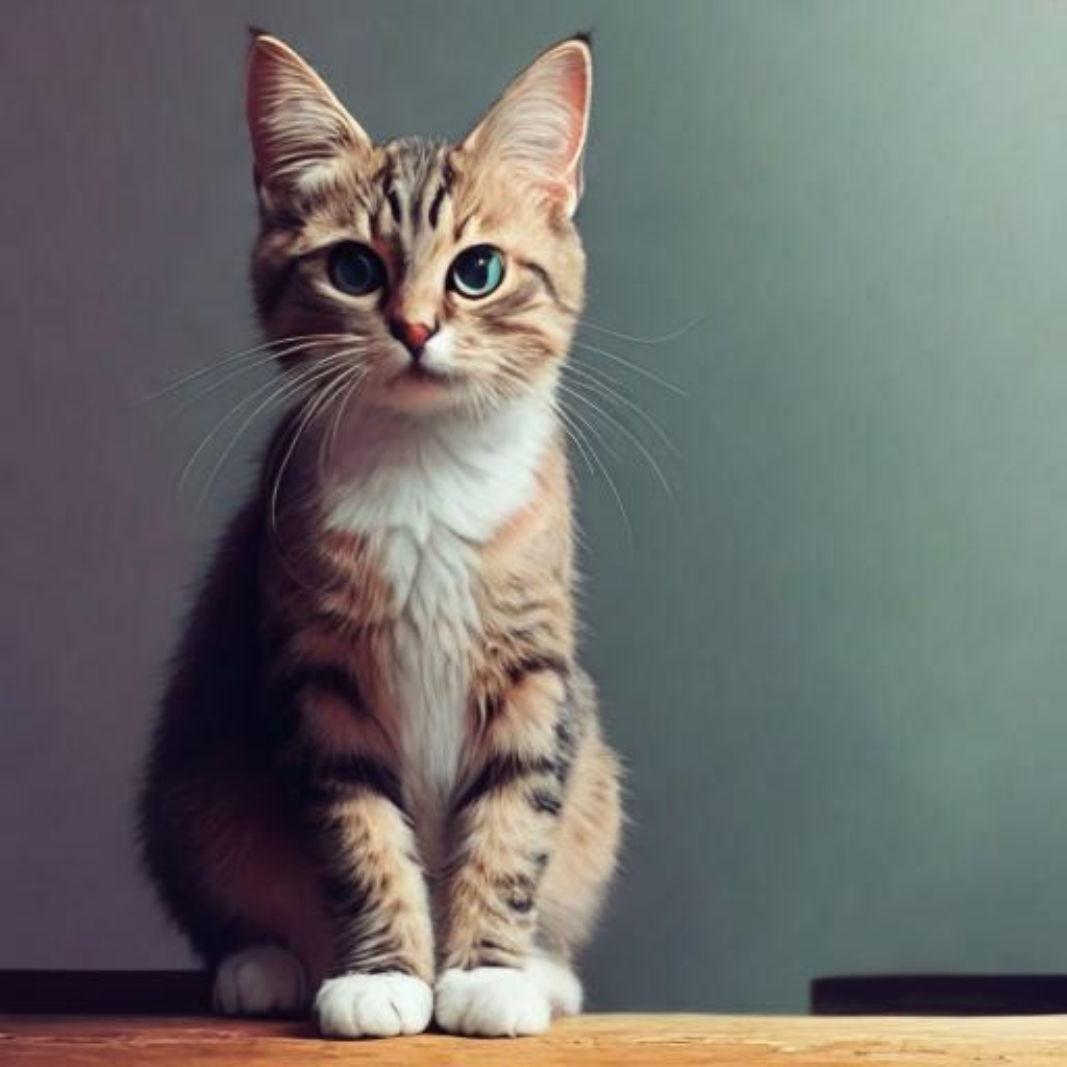}}%
          \hspace{0.2mm}%
        \subfloat[\textit{Add a pink jacket}]%
  {\includegraphics[width=0.199\linewidth]{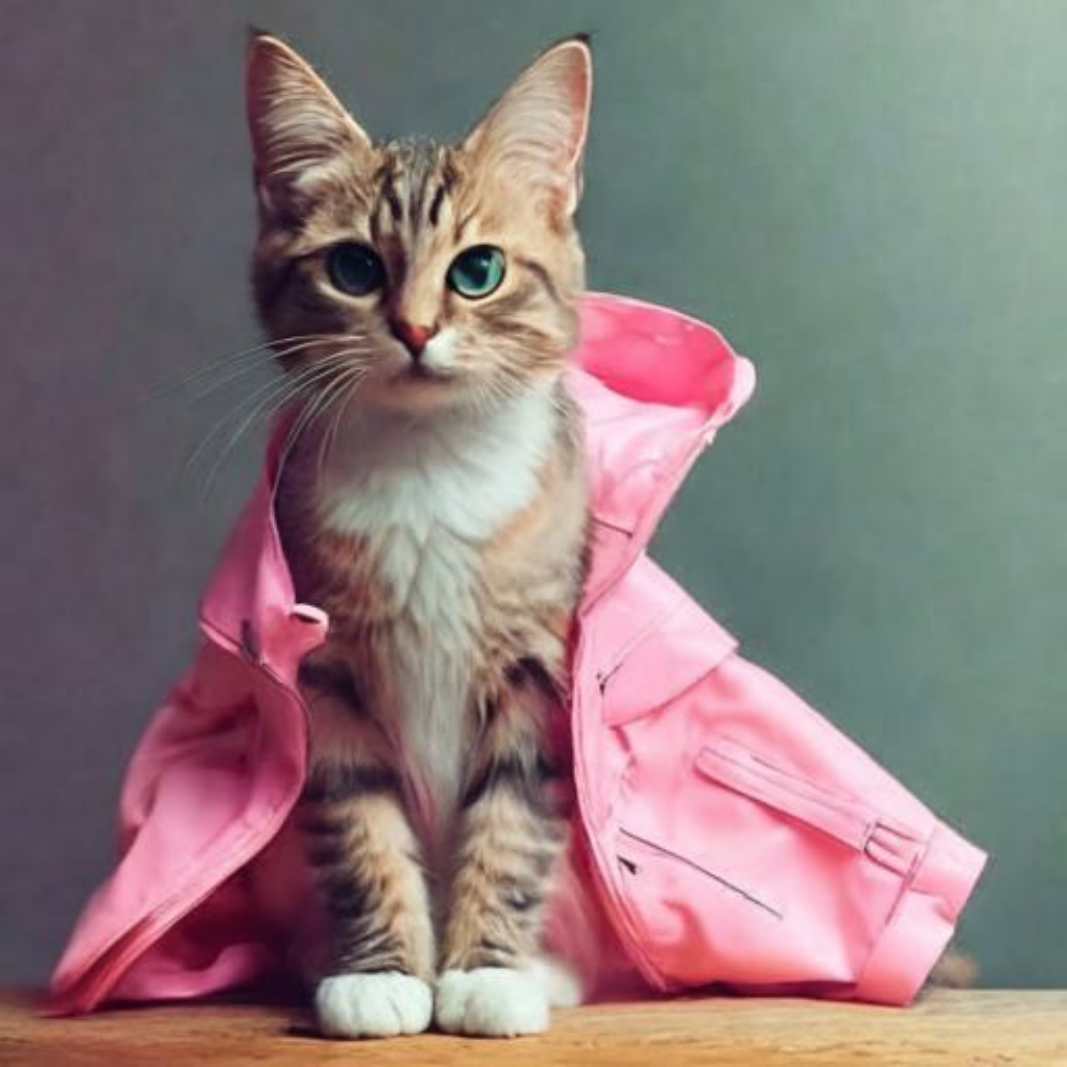}}%
          \hspace{0.2mm}%
        \subfloat[\textit{Make it rainy}]%
  {\includegraphics[width=0.199\linewidth]{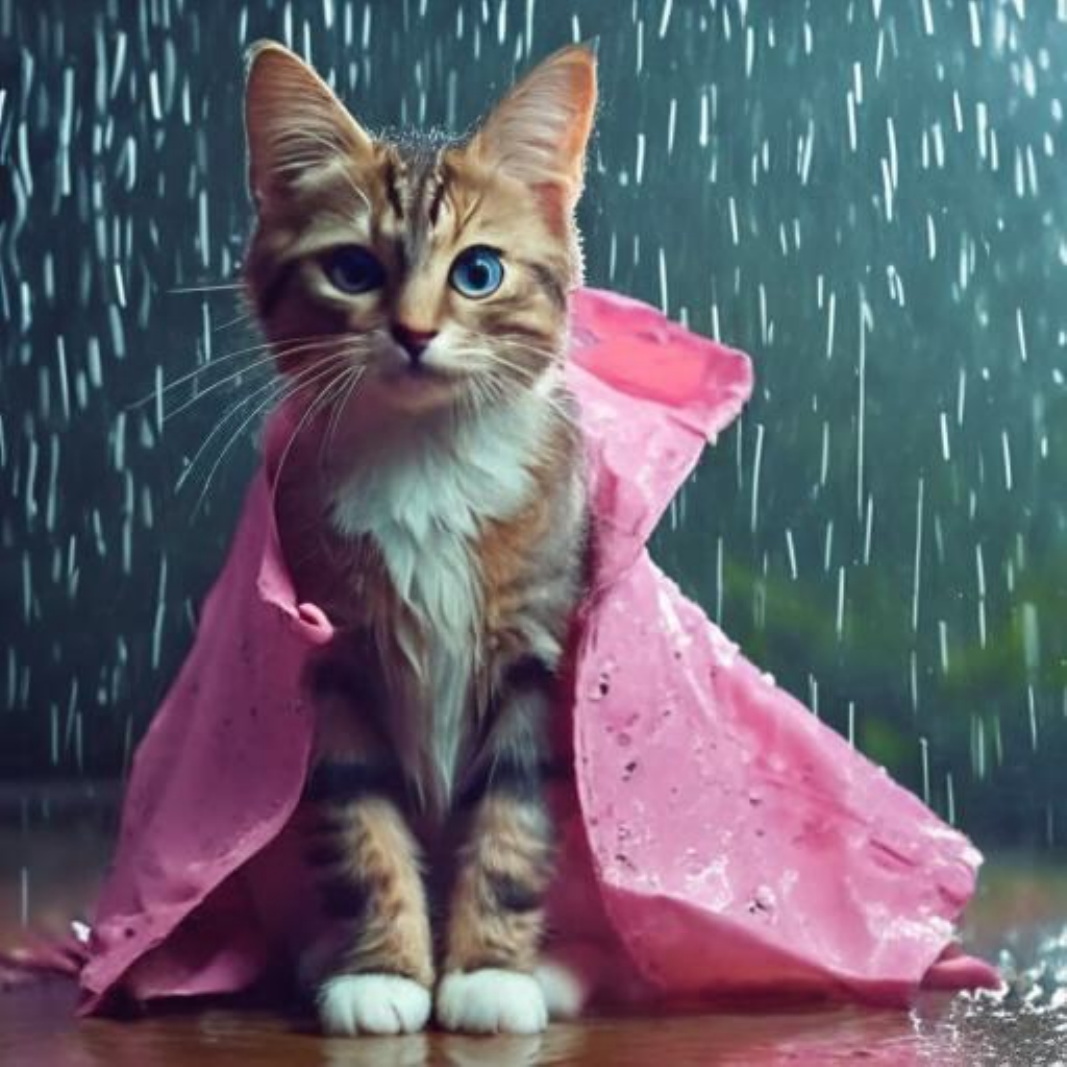}}%
            \hspace{0.2mm}%
        \subfloat[\textit{Have the cat look shocked}]%
  {\includegraphics[width=0.199\linewidth]{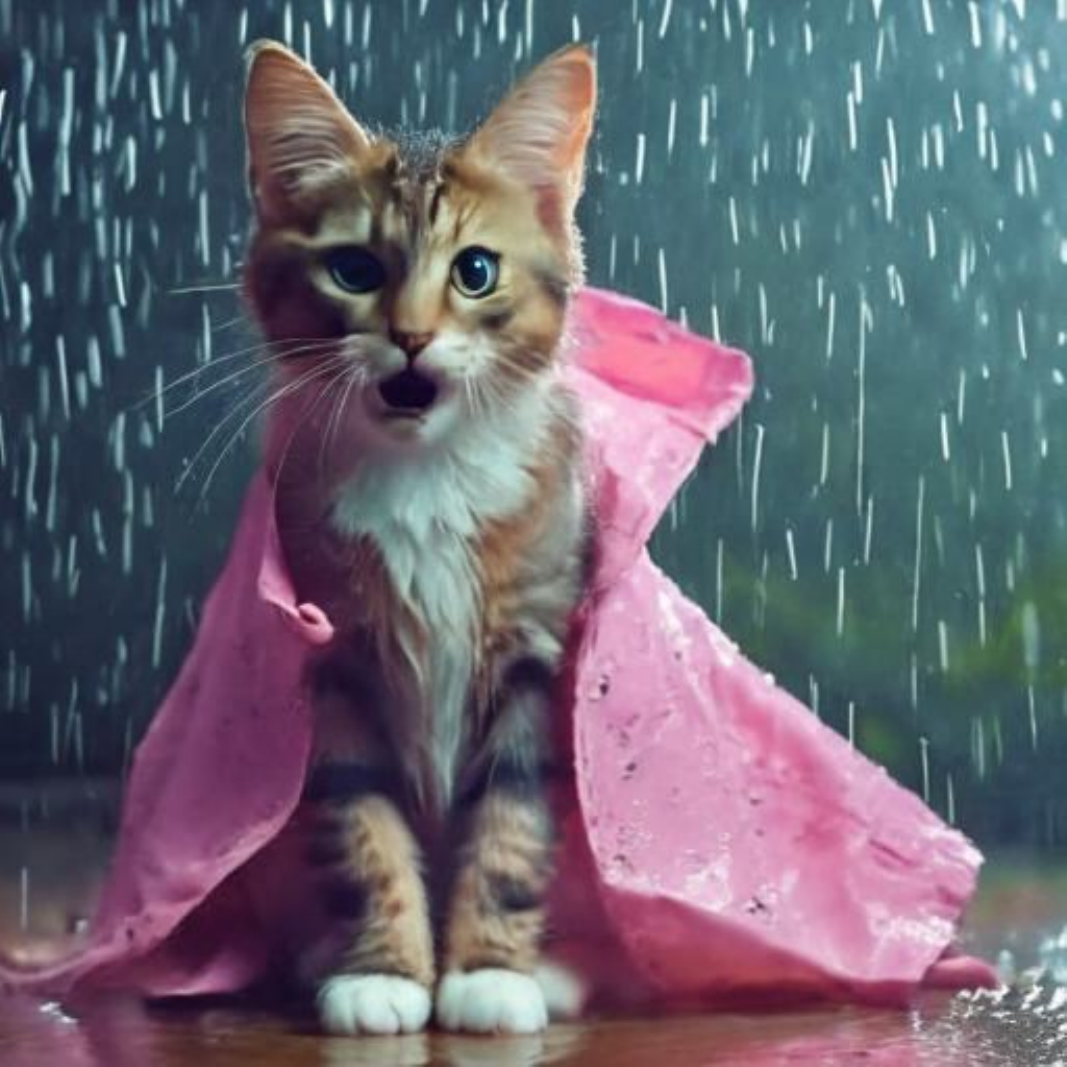}}%
        \vspace{-0.35cm} %
            \subfloat[\textit{Extract the depth map}]%
  {\includegraphics[width=0.199\linewidth]{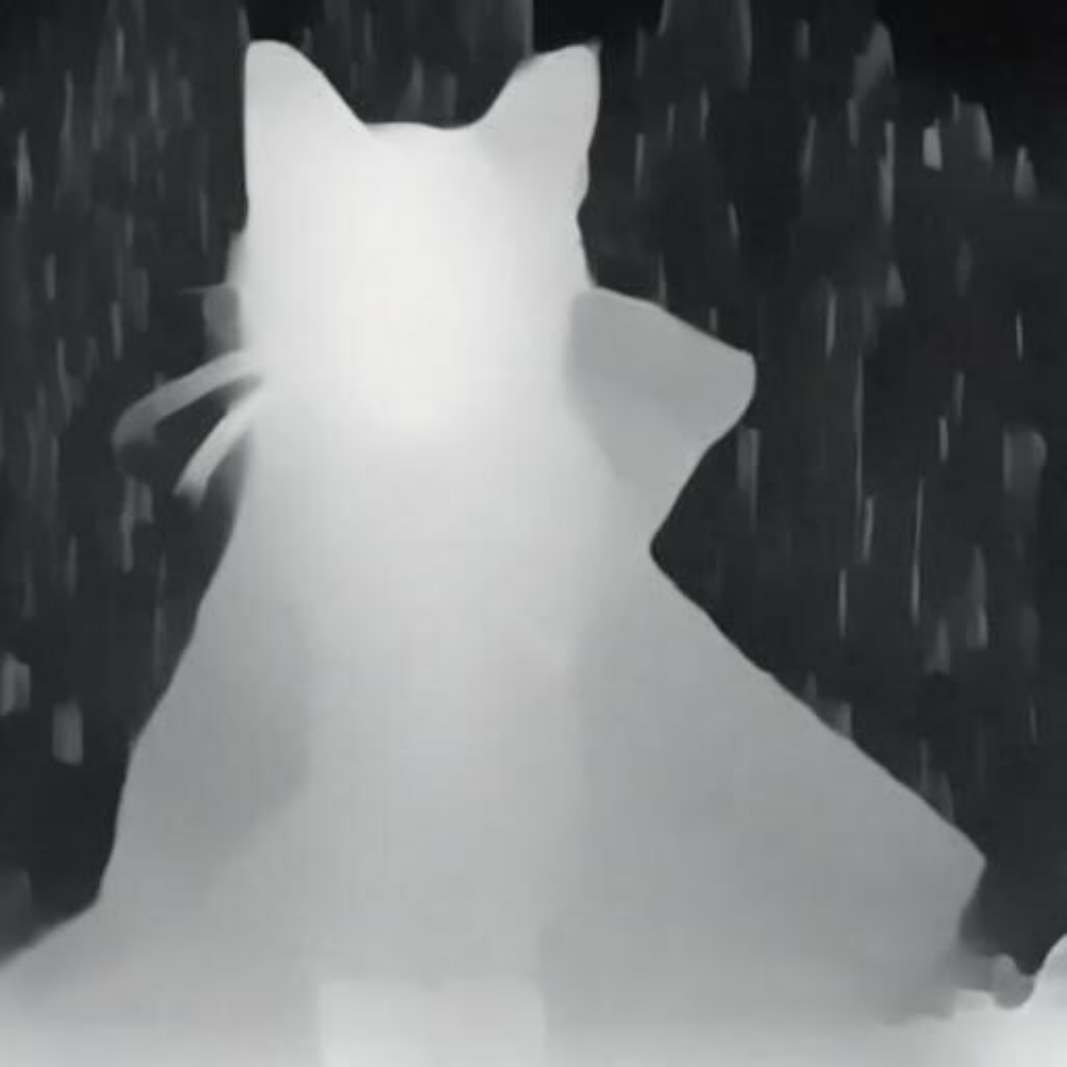}}%
          \hspace{0.2mm}%
       \subfloat[\centering \textit{ Generate a rainy day image of a hedgehog in a dress using the depth map}]%
  {\includegraphics[width=0.199\linewidth]{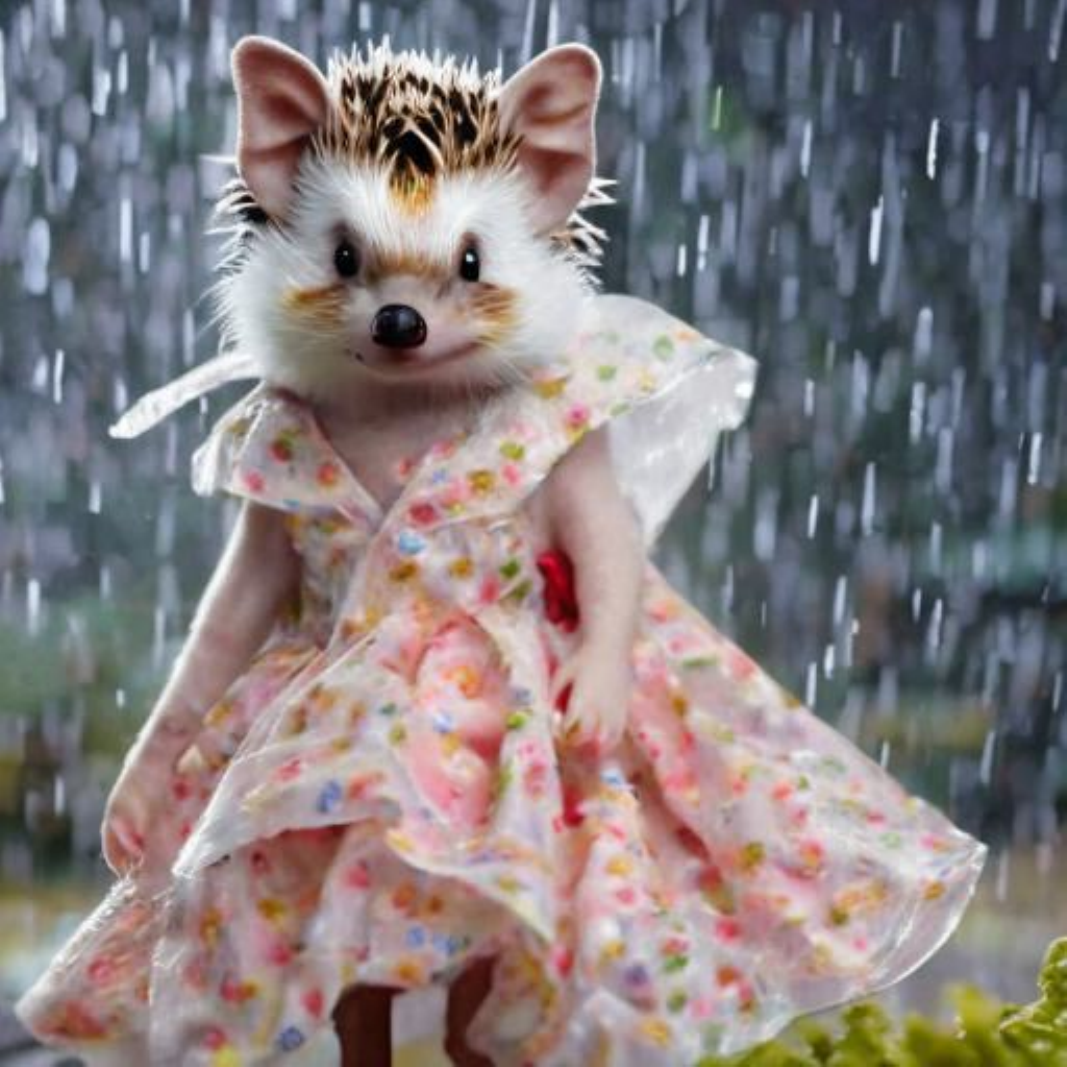}}%
          \hspace{0.2mm}%
        \subfloat[\centering \textit{Replace the dress with an astronaut outfit}]%
  {\includegraphics[width=0.199\linewidth]{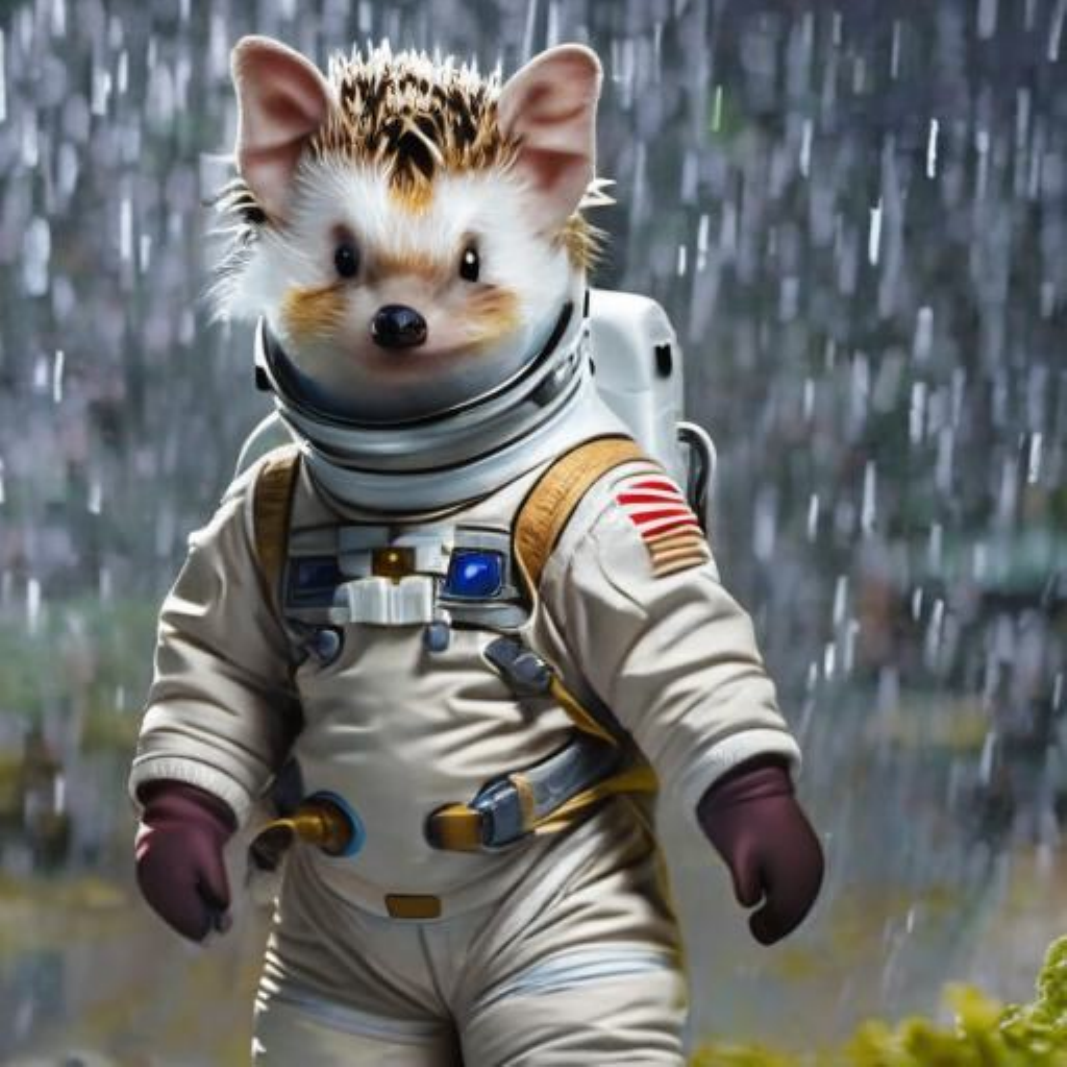}}%
          \hspace{0.2mm}%
        \subfloat[\centering \textit{ Segment the spacesuit, and detect the hands}]%
    {\includegraphics[width=0.199\linewidth]{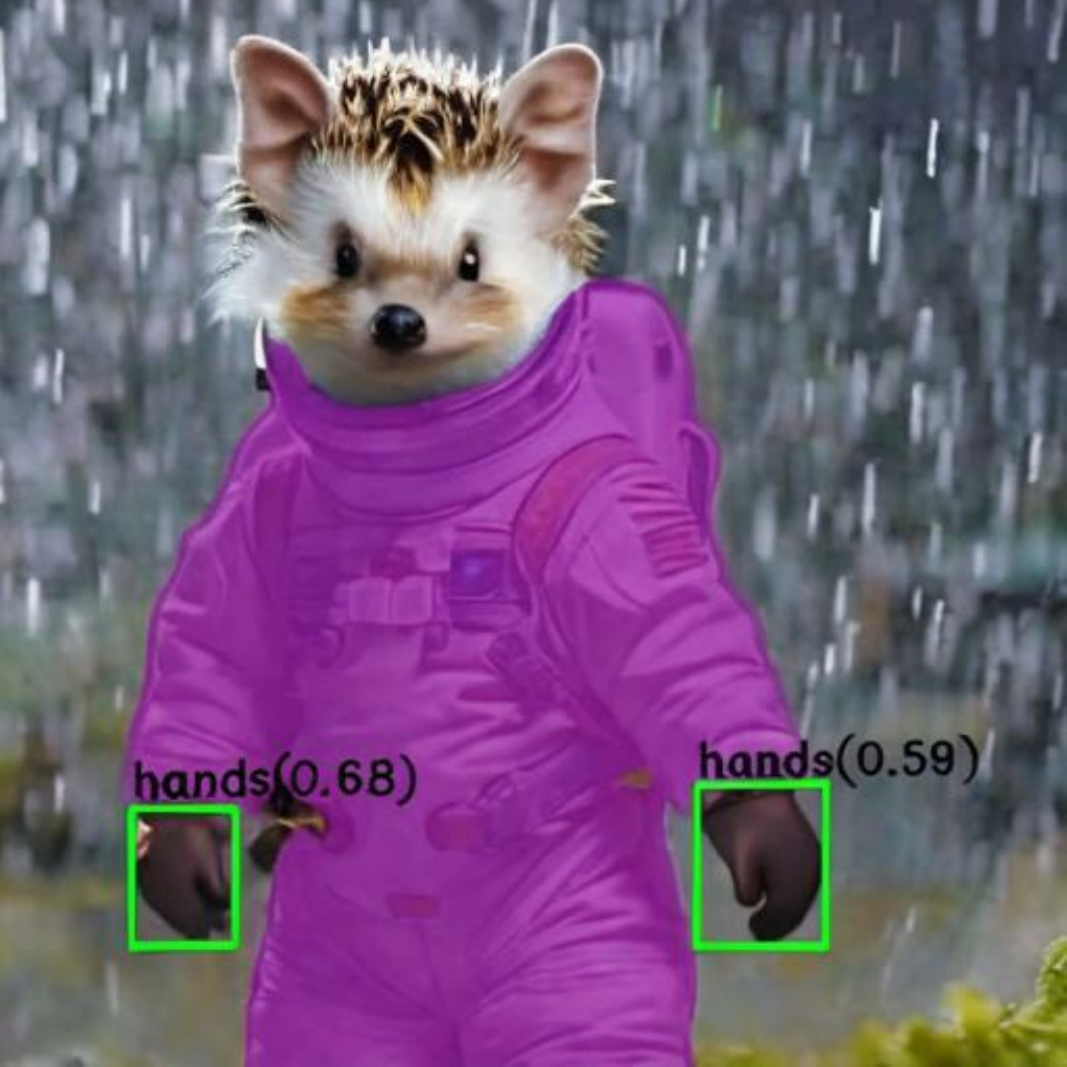}}%
            \hspace{0.2mm}%
                \subfloat[\centering \textit{ Add the text "purple cat" using a purple font}]%
    {\includegraphics[width=0.199\linewidth]{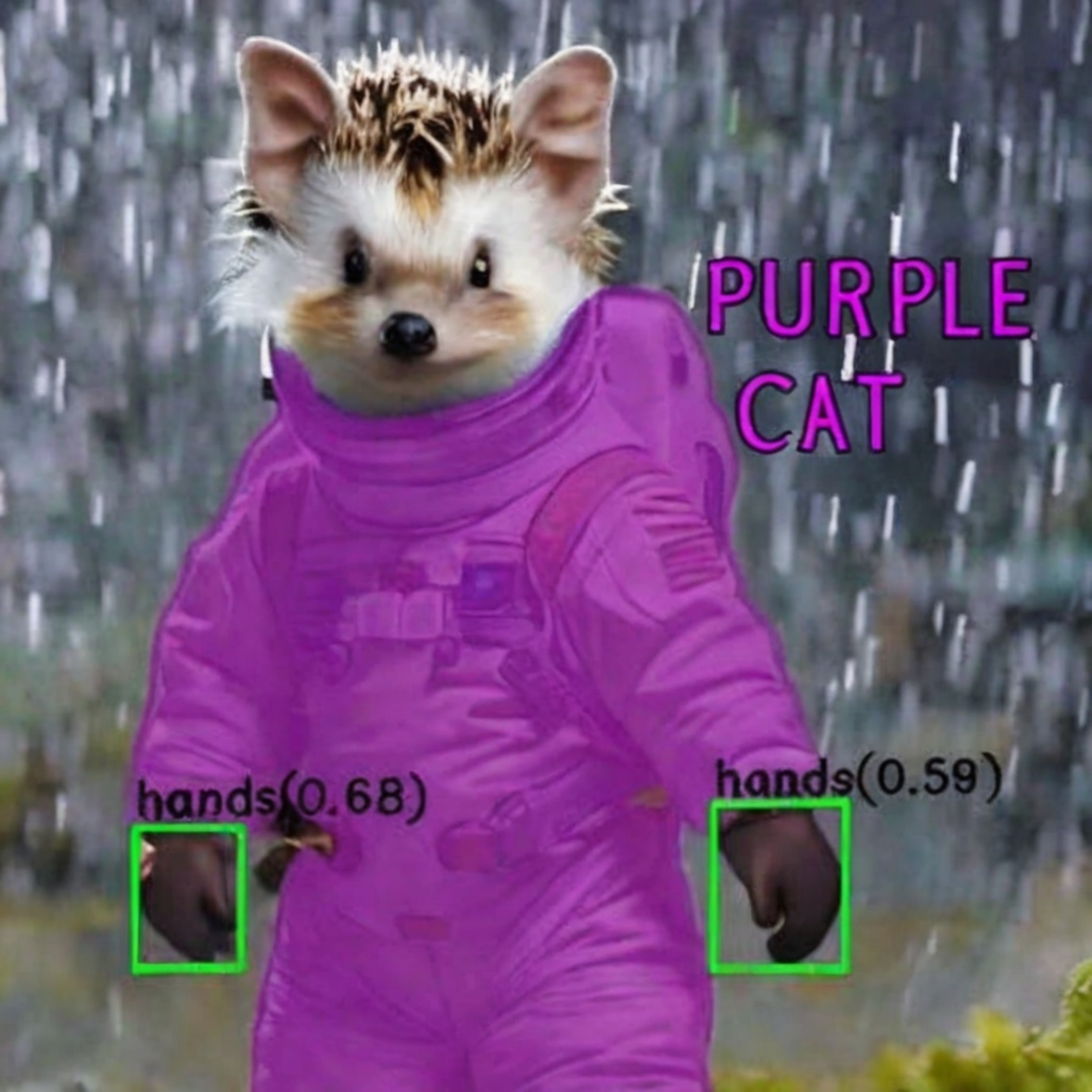}}%
            \hspace{0.2mm}%
    \caption{\textbf{Multi-turn image editing.} Each subsequent image is derived from the prior one, using its associated caption. The initial image is based on a zeroed reference. }
    \label{fig:cat}
    \vspace{-0.6cm} %
\end{figure*}

\section{Introduction}
\label{sec:intro}
Image editing is a widely-used application that millions engage with every day. 
Popular image editing tools, however, either demand considerable expertise and are time-consuming to use, or are quite limited, providing only a predefined set of editing operations, such as specific filters.
Instruction-based image editing~\cite{brooks2023instructpix2pix,zhang2023magicbrush} attempts to resolve these limitations by allowing users to effortlessly describe their editing goals using natural language instructions. %
For instance, a user can provide a model with an image and instruct it to ``Dress the emu with a fireman outfit'' or ``Let's see it graduating" (see Fig.~\ref{fig:tcandidate}).

Nevertheless, while instruction-based image editing models like InstructPix2Pix~\cite{brooks2023instructpix2pix} are \textit{designed} to process any given instruction, they often struggle to \textit{accurately} interpret and execute such instructions.
Moreover, their generalization is limited, often falling short on tasks that deviate slightly from those they were trained on (see Fig. \ref{fig:comp_our_images}).
To address these gaps, we introduce \model, the first image editing model trained on an extensive and diverse set of tasks, including both image editing and computer vision tasks.
\model provides a substantial improvement in both compliance with the edit instruction and preservation of the visual fidelity of the original image.
As we show through both automatic metrics~\cite{radford2021learning} and human judgments on two benchmarks~\cite{zhang2023magicbrush}, \model achieves state-of-the-art results in instruction-based image editing.

The success of \model stems from two key contributions. 
First, we train our model to multi-task across \ntasks distinct image editing tasks.
These tasks span region-based editing tasks, free-form editing tasks and computer vision tasks, all formulated as generative tasks.
Unlike previous work, we develop a distinct data curation pipeline for each task, allowing us to gather a training set that is not only more diverse but also more precise in its examples.
We find that training a single model on all tasks yields better results than training expert models on each task independently.
Additionally, we show that as the number of training tasks increases, so does the performance of \model.
Furthermore, we discover that surprisingly, computer vision tasks such as detection, segmentation, and others, significantly enhance editing performance, as validated both by human raters as well as quantitative measures. 

Second, to process this wide array of tasks effectively, we introduce the concept of \textit{learned task embeddings}, which are used to steer the generation process toward the correct generative task.
Concretely, for each task, we learn a unique task embedding vector, and integrate it into the model through cross-attention interactions, and by adding it to the timestep embeddings.
We demonstrate that learned task embeddings significantly enhance the ability of our model to accurately infer the appropriate edit type from the free-form instruction and execute the correct edit.

Equipped with a robust model trained across a broad spectrum of tasks and guided by learned task embeddings, we explore few-shot adaptation to unseen tasks via \textit{task inversion}.
In this process, we maintain the model weights untouched, and solely update a task embedding to fit the new task.
Our experiments demonstrate that \model can swiftly adapt to new tasks, such as super-resolution, contour detection, and others. 
Notably, for some tasks, fine-tuning the model on just a \textit{handful} of examples yields results that nearly match those of an expert model trained on \textit{one hundred thousand} examples.
This makes task inversion with \model particularly advantageous in scenarios where labeled examples are limited, or when the compute budget is low.
Finally, to support future research for instruction-based image editing, we publicly release a diverse and challenging benchmark that includes \ntaskstest different image editing operations, as well as our model's generations on this dataset.

In summary, this work addresses the limitations of instruction-based image editing models in accurately following user instructions. 
We demonstrate that by employing multi-task training across a diverse array of tasks, including recognition, generation, and editing, we can enhance our model's performance. 
Furthermore, by incorporating learned task embeddings into our model's architecture, we not only improve its results but also enable efficient few-shot learning for new tasks.
With these improvements, our model sets a new standard in the field, offering significantly more precise and robust instruction-based image editing capabilities than existing models.

\section{Related Work}
\label{sec:prev}

The emergence of high-performing text-to-image diffusion models~\cite{Rombach2021HighResolutionIS,imagen,dalle2,gafni2022make} facilitated the development of effective text-based image editing methods.
Such methods usually employ aligned and detailed descriptions of the input and edited image to perform a specific edit. 
Prompt-to-Prompt~(P2P)~\cite{prompt2prompt} injects the input caption attention maps to the target caption attentions maps. 
Null-Text Inversion~\cite{Mokady2022NulltextIF} inverts an input image using the null-text embedding to support editing of a real image. 
Plug-and-Play~(PNP)~\cite{pnp} injects spatial features in addition to attention maps and obtains better performance at global editing. 
Imagic~\cite{Kawar2022ImagicTR} finetunes the diffusion model to support complex textual instructions.
EDICT~\cite{Wallace2022EDICTED} suggests an image inversion based on two noise vectors enabling better image reconstruction and textual faithfulness. 
Another class of image editing models, employs an input mask as additional input~\cite{avrahami2022blended,wang2023imagen,xie2023smartbrush}.
Blended Diffusion~\cite{avrahami2022blended} modifies the diffusion step by blending the input image in the unmasked regions.
Imagen Editor~\cite{wang2023imagen} and SmartBrush~\cite{xie2023smartbrush} finetune the text-to-image model to be conditioned on both the input image and mask.
While the text-based image editing methods detailed above enable humans to edit images, they frequently exhibit inconsistent performance and require multiple inputs, such as aligned and detailed descriptions of both the input and target images, or at times, input masks.

To offer a more intuitive and user-friendly interface, and significantly enhance ease of use for humans, InstructPix2Pix~\cite{brooks2023instructpix2pix} introduced an instructable image editing model.
They developed this model by utilizing both GPT-3~\cite{gpt3} and Prompt-to-Prompt~\cite{prompt2prompt}, to generate a large synthetic dataset for instruction-based image editing, and employed the dataset to train an instructable image editing model. 
Unlike InstructPix2Pix which used a synthetic dataset, MagicBrush~\cite{zhang2023magicbrush} developed a manually-annotated instruction-guided image editing dataset by requesting humans to use an online image editing tool~\cite{dalle2_tool}. 
Finetuning InstructPix2Pix on this dataset led to improved image editing capabilities.
However, even though there has been progress and improvement in instruction-based image editing models, we show in Sec.~\ref{sec:exp} that state-of-the-art image editing models still struggle with accurately interpreting and precisely executing editing instructions. 

In this paper, we drastically narrow such performance gaps by leveraging multi-task training and a matching architecture. 
Unlike prior work that solely focuses on image editing~\cite{brooks2023instructpix2pix,zhang2023magicbrush}, we train our model to perform various tasks and learn a very diverse set of capabilities. 
The quality and versatility of our training procedure and dataset, together with our improved architecture for multi-task learning, enables us to make a big leap in performance and differentiates us from prior work in the field. Fig.~\ref{fig:comp_our_images} include several challenging editing samples as examples.

\begin{table}[ht]
\centering
\begin{tabular}{@{}p{\linewidth}@{}}
\toprule
\textbf{1. Region-Based Editing} \\
\begin{minipage}[t]{\linewidth} %
\begin{itemize}[leftmargin=*, nosep, after=\strut]
    \item \textbf{Local}: Substituting one object for another, altering an object's attributes (e.g., ``make it smile'').
    \item \textbf{Remove}: Erasing an object from the image.
    \item \textbf{Add}: Inserting a new object into the image. 
    \item \textbf{Texture}: Altering an object's visual characteristics without affecting its structure (e.g.,  painting over, filling or covering an object).
    \item \textbf{Background}: Changing the scene's background.
\end{itemize}
\end{minipage} \\
\midrule
\textbf{2. Free-Form Editing} \\
\begin{minipage}[t]{\linewidth}
\begin{itemize}[leftmargin=*, nosep, after=\strut]
    \item \textbf{Global}: Edit instructions that affect the entire image, or that can not be described using a mask (e.g., ``let's see it in the summer'').
    \item \textbf{Style}: Change the style of an image.
    \item \textbf{Text Editing}: This involves text-related editing tasks such as adding, removing, swapping text, and altering the text's font and color.
\end{itemize}
\end{minipage} \\
\midrule
\textbf{3. Vision tasks} \\
\begin{minipage}[t]{\linewidth}
\begin{itemize}[leftmargin=*, nosep, after=\strut]
    \item \textbf{Detect}: Identifying and marking a specific object within the image with a rectangle bounding box.
    \item \textbf{Segment}: Isolating and marking an object in the image.
    \item \textbf{Color}: Color adjustments like sharpening and blurring.
    \item \textbf{Image-to-Image Translation}: Tasks that involve bi-directional image type conversion, such as sketch-to-image, depth map-to-image, normal map-to-image, pose-to-image, segmentation map-to-image, and so on.
\end{itemize}
\end{minipage} \\
\bottomrule
\end{tabular}
\caption{Description of the tasks forming the \model dataset.}
\label{tab:task_description}
\vspace{-0.5cm}
\end{table}

\section{Multi-Task Dataset for Image Editing}
\label{sec:data}
Training a robust and accurate image editing model requires a highly diverse dataset of input images, editing instructions, and output edited images. 
However, manually collecting such examples is impractically time-consuming, existing sources on the web (e.g. communities and forums on social media) are limited in size, and publicly available synthetic datasets often lack in diversity or quality. 
Therefore, we construct a new dataset that encompasses \ntasks distinct tasks and ten million examples. 
Each example~$(c_I, c_T, x, i)$ in our dataset, contains an input image~$c_I$, a text instruction~$c_T$, a target image~$x$, and a task index $i$ (out of the \ntasks). 
The following section outlines the process of the data construction. In Sec.~\ref{sec:instruct} we describe the instruction generation process, and in Sec.~\ref{sec:images} the image pairs ($c_I, x$) generation and filtering.

\subsection{Task Categories}
\label{sec:tasks}
The dataset is composed of tasks which are divided into three main categories: region-based editing, free-form editing, and vision tasks. Tab.~\ref{tab:task_description} includes the full list of tasks, and their distribution 
in the train set is visualized in Fig.~\ref{fig:data_dist}.

\subsection{Instruction Generation}
\label{sec:instruct}
To generate editing instructions, we leverage the dialogue-optimized 70 billion parameter Llama 2 variant~\cite{touvron2023llama}. 
We observed that using a single agent to generate the instructions for all tasks leads to a lack of diversity in the dataset. Notably, the LLM exhibits a bias towards particular tasks and instruction phrasings. 
To address this, we utilize in-context learning to create a task-specific agent for each task. 
Concretely, we provide the LLM with a task description, a few task-specific exemplars, and a real image caption.
To increase diversity we sample the exemplars and randomize their order.
Given such input, we expect the LLM to output (1) an editing instruction, (2) an output caption for an ideal output image, and (3) which objects should be updated or added to the original image. 
We refer the readers to Fig.~\ref{fig:llama_learning}-\ref{fig:llama_prompting} for examples of our prompts. Further details on instruction generation are provided in Appendix~\ref{sec:instruction_sup}.

\subsection{Image Pairs Generation}
\label{sec:images}

Our aim is to generate pairs of input and edited images that adhere to the edit instructions and preserve image elements that should remain intact. 
In order to address the unique challenges associated with each task and create a high quality dataset, we develop a novel generation technique for each task. 
In the following subsections, we first detail our core improvements to the Prompt-to-Prompt algorithm, utilized when generating data for Local, Add, Remove, Texture, and Global tasks. Secondly, we describe our post-processing filtering. %
Full details can be found in Appendix~\ref{sec:image_generation_sup}. 
For region-based tasks, see Appendix~\ref{sec:region_dataset_sup}. Free-form tasks are described in detail in Appendix~\ref{sec:free_form_tasks_sup}. Finally, all vision tasks are described in Appendix~\ref{sec:vision_tasks_sup}.

\myparagraph{Grounded Precise Editing.}
\label{sec:grounded_precise}
A crucial prerequisite when creating a pair of input and edited images is to guarantee that the two images differ only in specific elements or locations, while remaining identical in all other aspects.
Previous instruct-based image editing methods~\cite{brooks2023instructpix2pix} rely on Prompt-to-Prompt~(P2P) to build an image-editing dataset. 
P2P injects cross-attention maps from the input image generation to the edited image generation. 
To support local edits, P2P additionally approximates a \textit{mask} of the edited part, based on the cross-attention maps and constrains the edit to this local area. 
P2P relies on word-to-word alignment between the input image caption and the edited image caption~(e.g. "a cat riding a \textit{bicycle}" and "a cat riding a \textit{car}") to produce editing image pairs. %
However, when there is no word-to-word alignment, the resulting mask tends to be imprecise due to its reliance on cross-attention maps.
Furthermore, as word-to-word alignment is not a practical assumption in most of the image editing tasks, this approach often fails to preserve structure and identity.
To address this challenge, we propose a mask extraction method, which is applied before the editing process. Our approach involves: (i) identifying the edited areas from the editing instruction via an LLM and creating corresponding masks before image generation, and (ii) integrating these masks during the editing process to ensure seamless fusion of edited regions with the original image.
Further description of the method is found at Appendix~\ref{sec:mask_based_control_sup}.

Distinct editing challenges, such as adding or removing objects, require tailored solutions. We utilize various techniques, including dilation and Gaussian blurring, to refine the masks. We describe the mask extraction process in detail in Appendix~\ref{sec:mask_extraction_sup}.

\myparagraph{Filtering.}
We employ a comprehensive filtering approach to ensure the fidelity of the dataset. 
This includes: 
(i) using the task predictor~(Sec.~\ref{sec:text_to_task}) to reassign samples with instructions that should belong to another task, (ii) applying CLIP filtering metrics~\cite{brooks2023instructpix2pix}, (iii) employing structure preserving filtering based on the L1 distance between the depth map of the input image and the edited image, and (iv) applying image detectors to validate the presence (in Add task), the absence (in Remove task) or replacement (in Local task) of elements, according to the objects specified in the instruction. 
This process filters out 70\% of the data, resulting in a final dataset of ten million samples.

\begin{figure*}[t]
   \centering
\begin{tabular}{@{\hspace{-2\tabcolsep}}c@{\hspace{-0.3\tabcolsep}}c@{\hspace{-0.3\tabcolsep}}c@{\hspace{-0.3\tabcolsep}}c@{\hspace{-0.3\tabcolsep}}c}

 & \begin{tabular}[x]{@{}c@{}} \footnotesize{Input} \end{tabular}  & \begin{tabular}[x]{@{}c@{}} \footnotesize{\textbf{\model}} \end{tabular} &  \begin{tabular}[x]{@{}c@{}} \footnotesize{InstructPix2Pix} \end{tabular} & \footnotesize{MagicBrush}  \\

\resizebox{!}{26px}{
\begin{tabular}[x]{@{}c@{}} \textit{Make it} \\ \textit{play a} \\ \textit{rainbow}\\ \textit{colored}\\ \textit{trumpet} \end{tabular}} &
\raisebox{-.5\height}{
\includegraphics[width=0.235\linewidth]{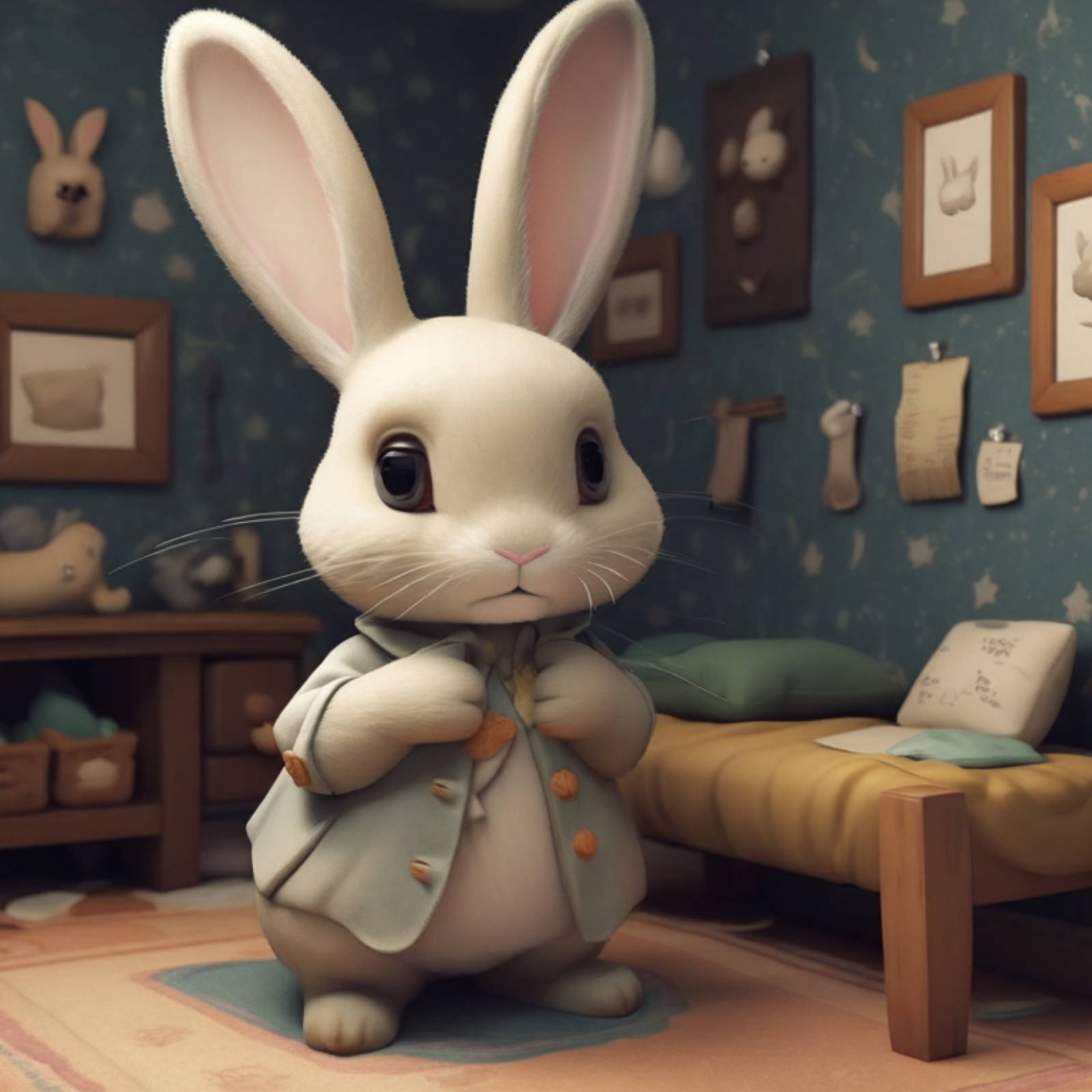}}&
\raisebox{-.5\height}{
\includegraphics[width=0.235\linewidth]{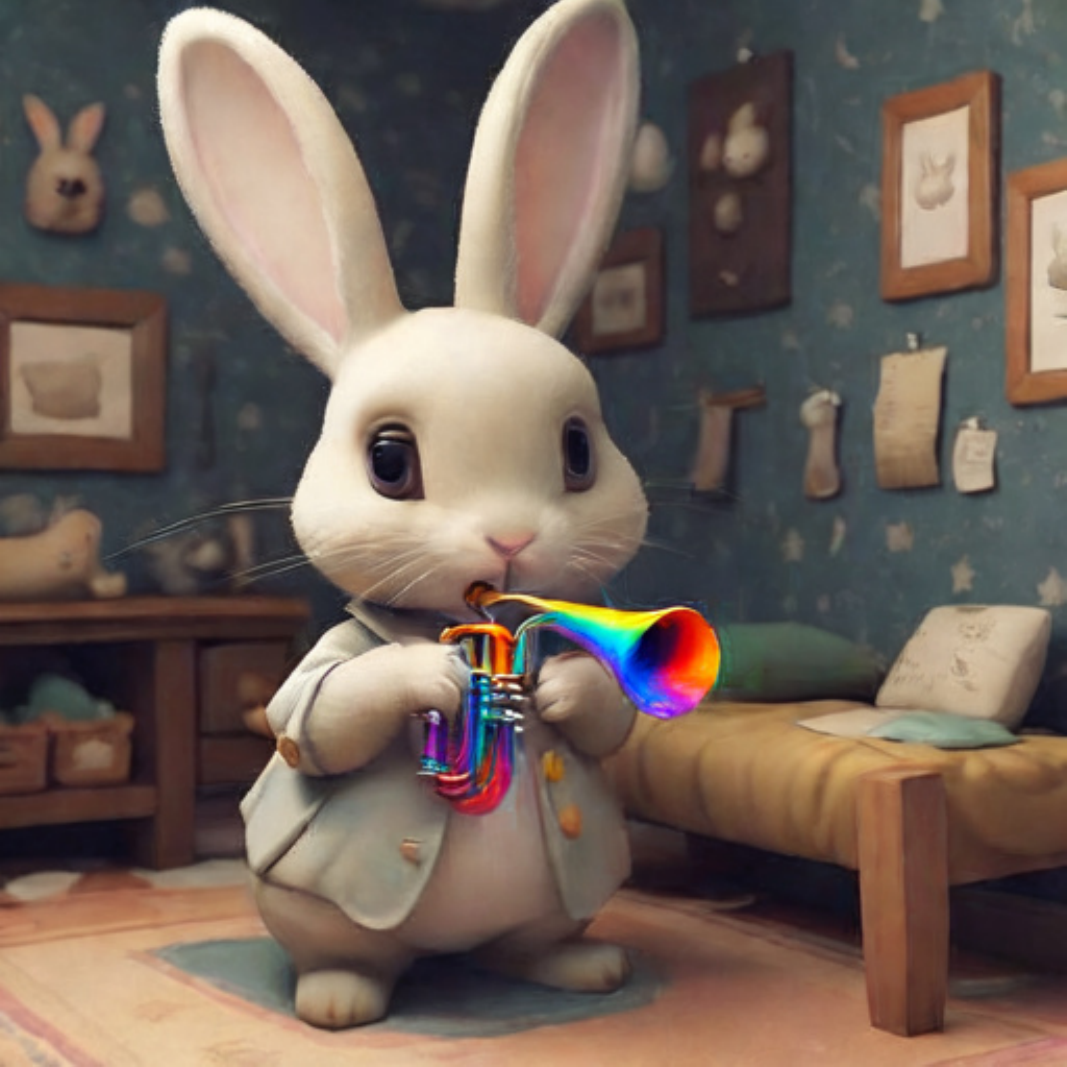}}&
\raisebox{-.5\height}{
\includegraphics[width=0.235\linewidth]{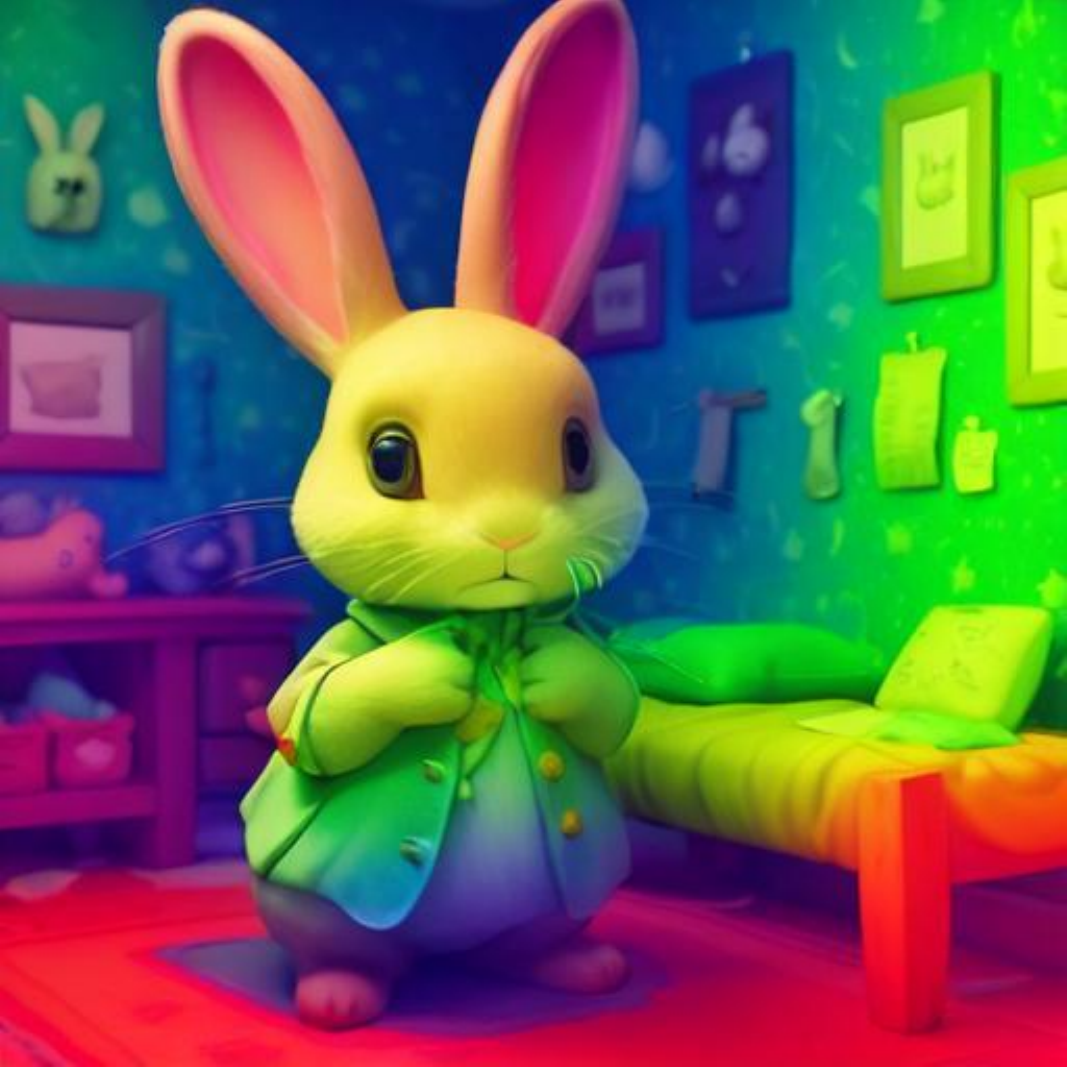 }} &
\raisebox{-.5\height}{
\includegraphics[width=0.235\linewidth]{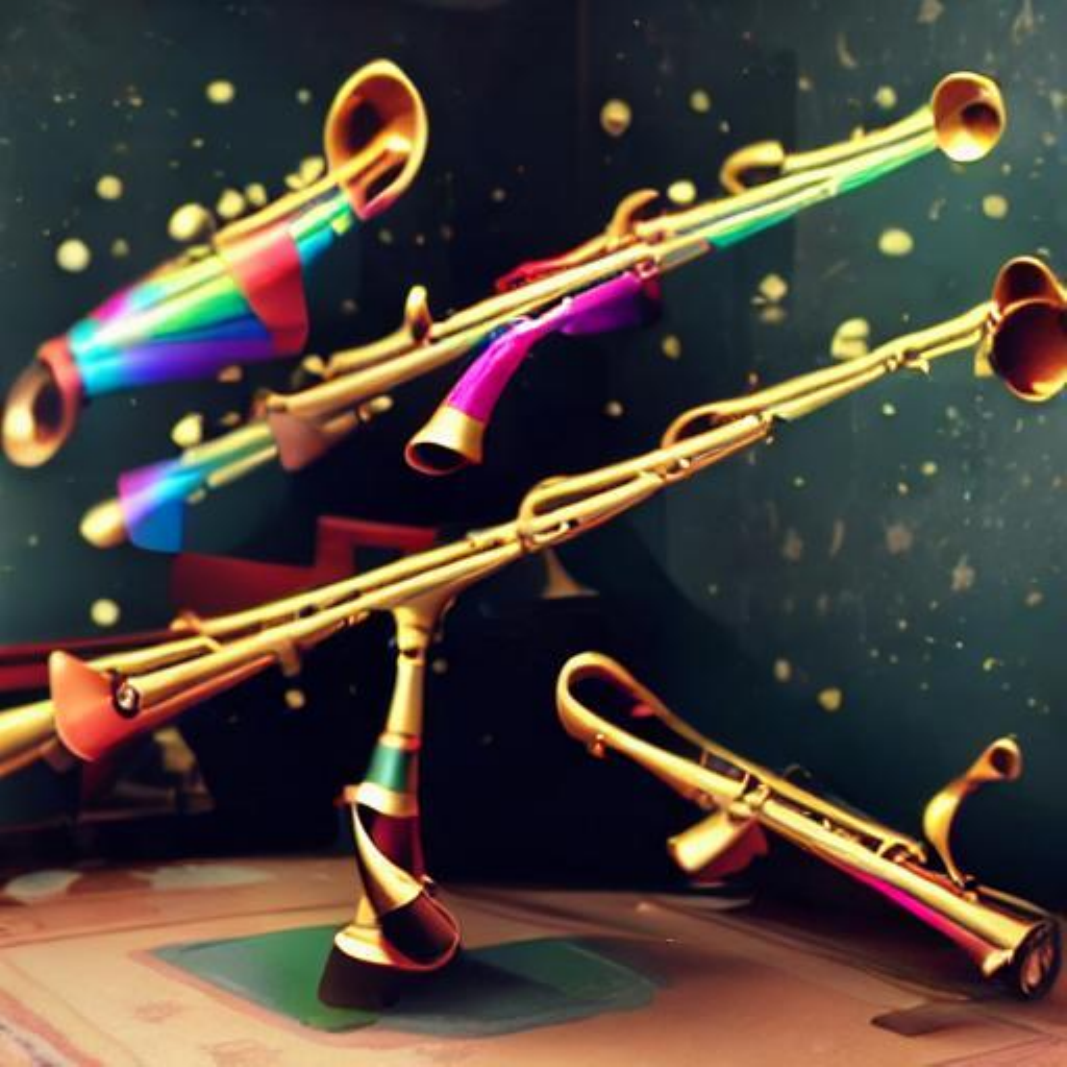}} \\ [5mm]

& \multicolumn{4}{c}{\footnotesize{\textbf{Global \& Texture edit}. Baselines struggle to execute complex instructions that involve both global edits and texture changes.}}\\

\resizebox{!}{22px}{
\begin{tabular}[x]{@{}c@{}}\textit{Add two}\\ \textit{unicorns} \\ \textit{on top of}\\ \textit{the car} \end{tabular}} &
\raisebox{-.5\height}{
\includegraphics[width=0.235\linewidth]{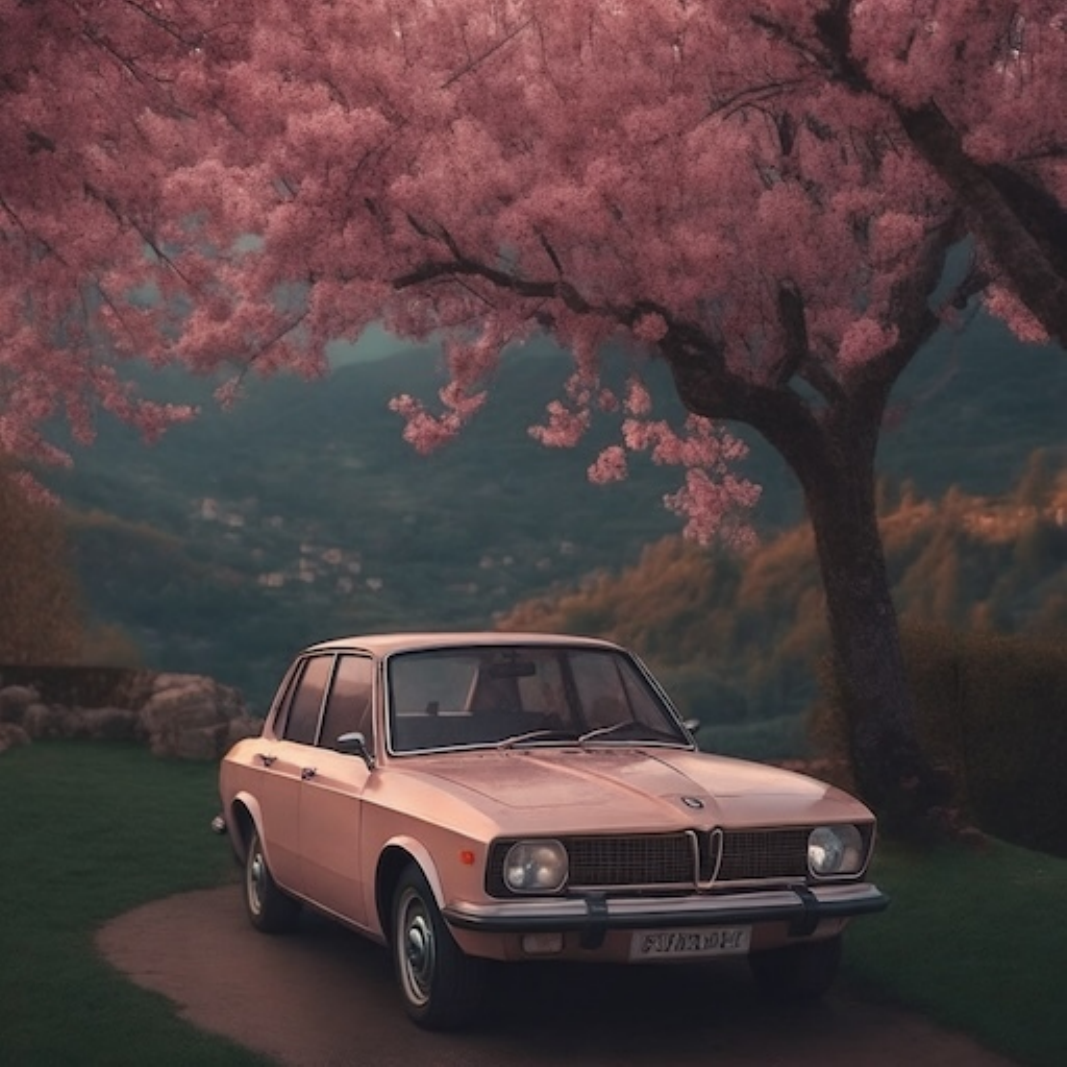}}&
\raisebox{-.5\height}{
\includegraphics[width=0.235\linewidth]{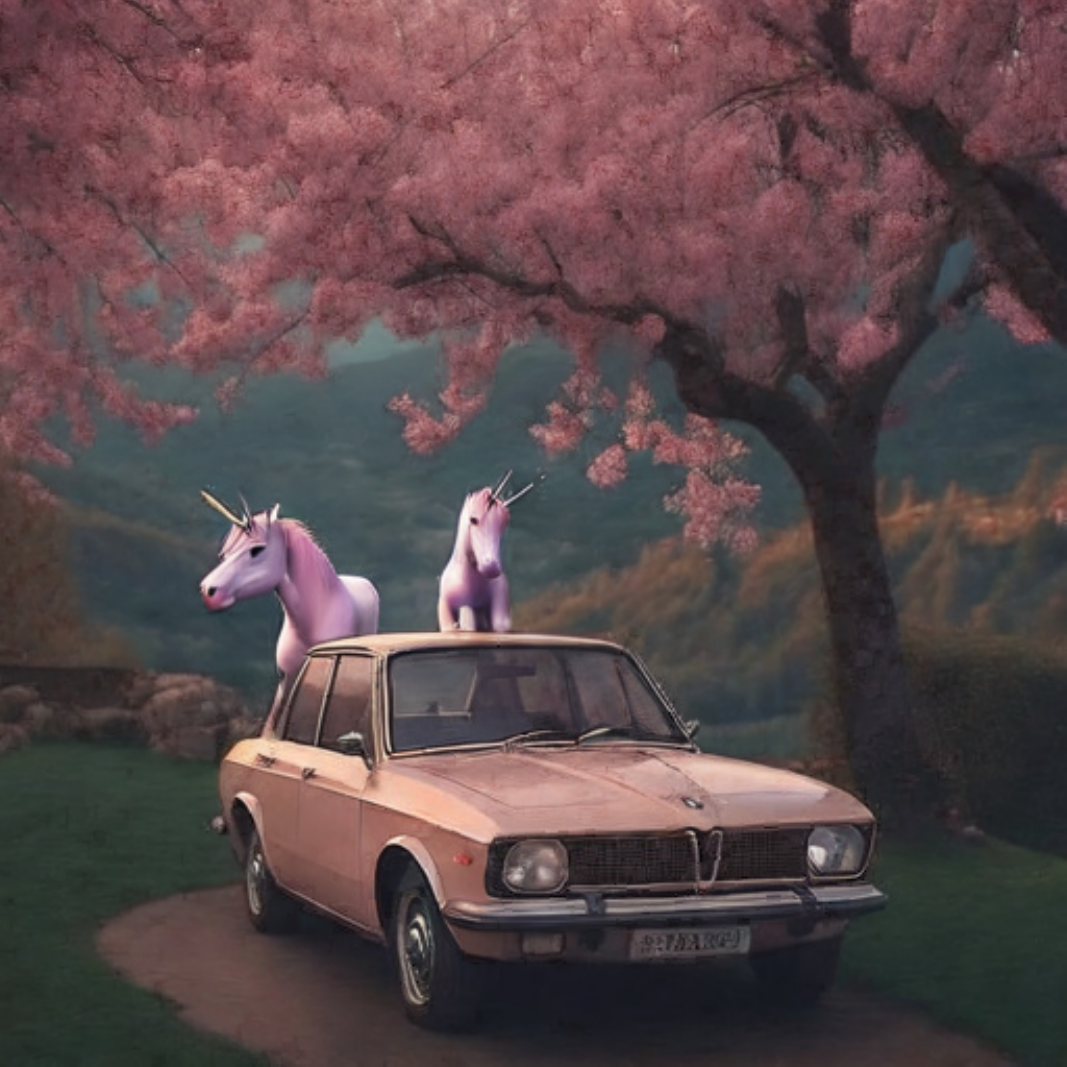}}&
\raisebox{-.5\height}{
\includegraphics[width=0.235\linewidth]{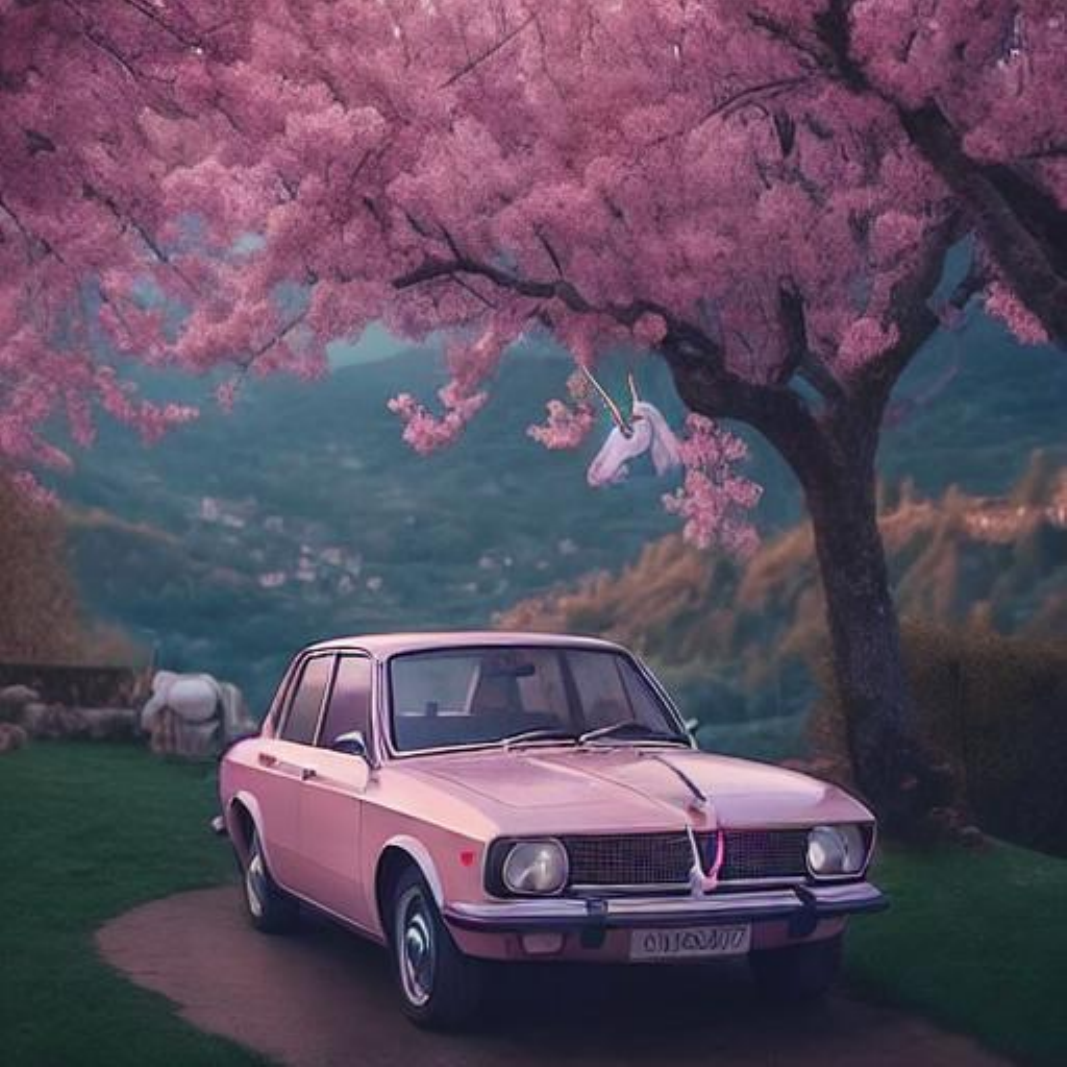 }} &
\raisebox{-.5\height}{
\includegraphics[width=0.235\linewidth]{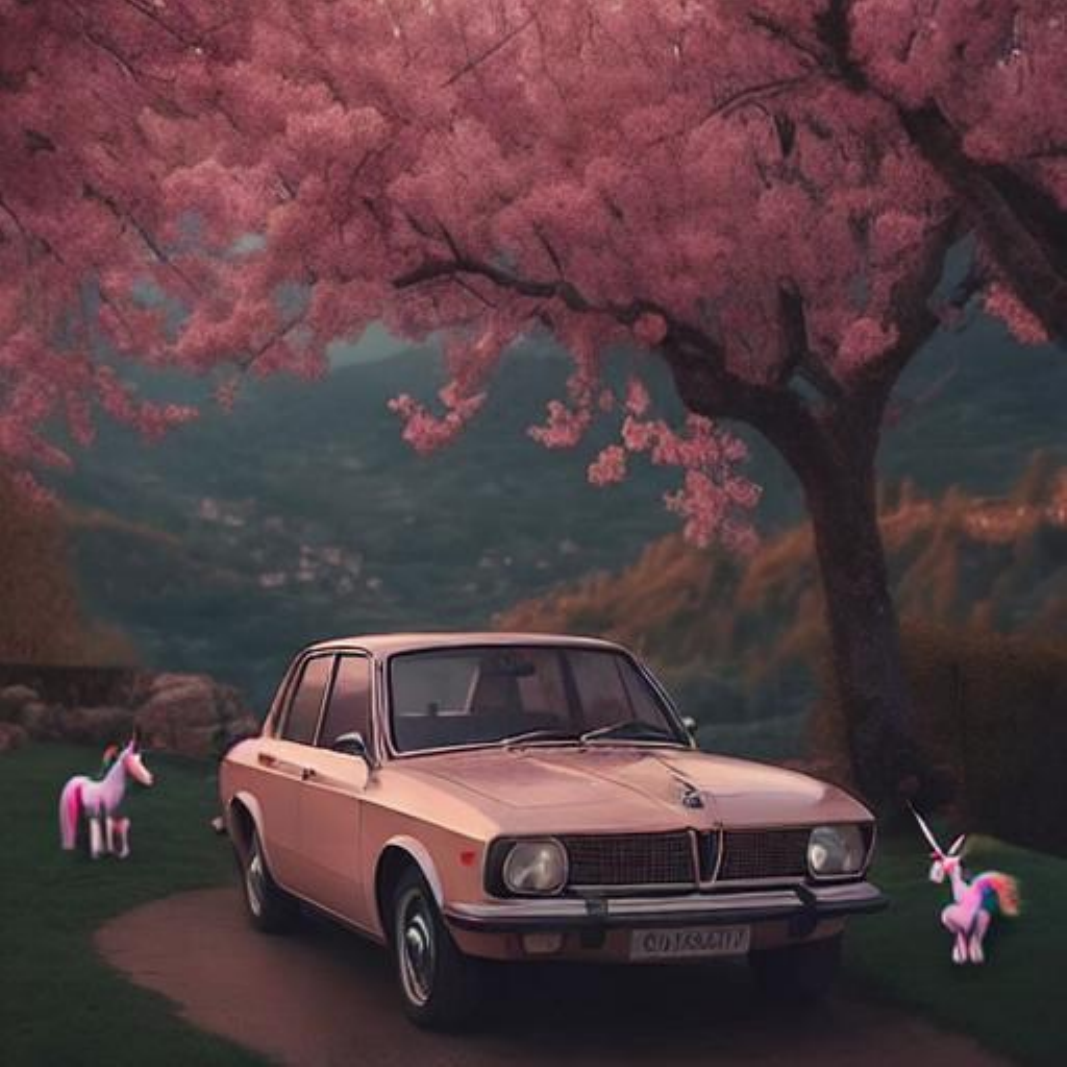}} \\ [5mm]

& \multicolumn{4}{c}{\footnotesize{\textbf{Prepositions \& Counting}. Baselines struggle with relations between objects and number of objects.}}\\

\resizebox{!}{20px}{
\begin{tabular}[x]{@{}c@{}} \textit{Change}\\ \textit{the legs} \\ \textit{to be}\\ \textit{bionic} \end{tabular}} &
\raisebox{-.5\height}{
\includegraphics[width=0.235\linewidth]{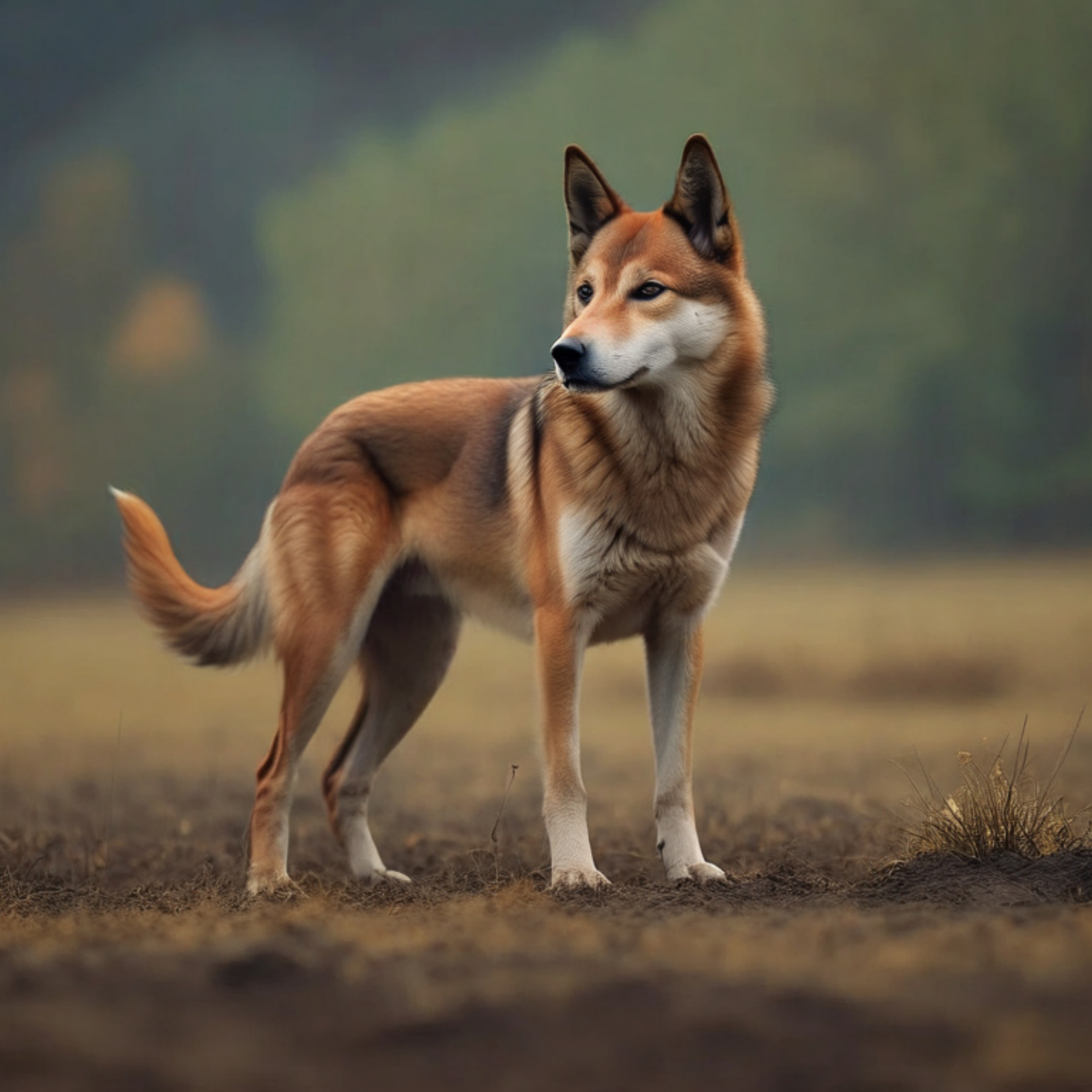}}&
\raisebox{-.5\height}{
\includegraphics[width=0.235\linewidth]{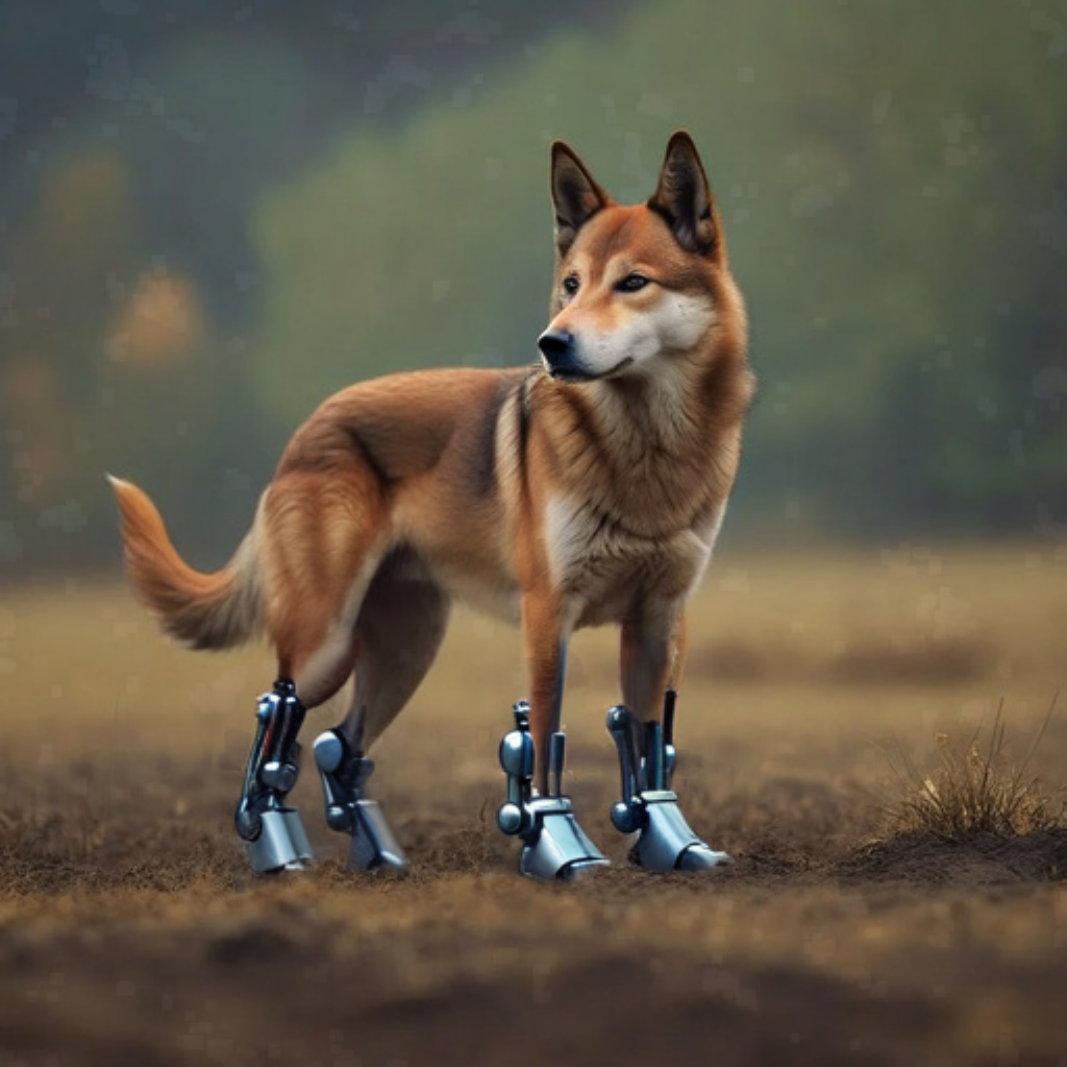}}&
\raisebox{-.5\height}{
\includegraphics[width=0.235\linewidth]{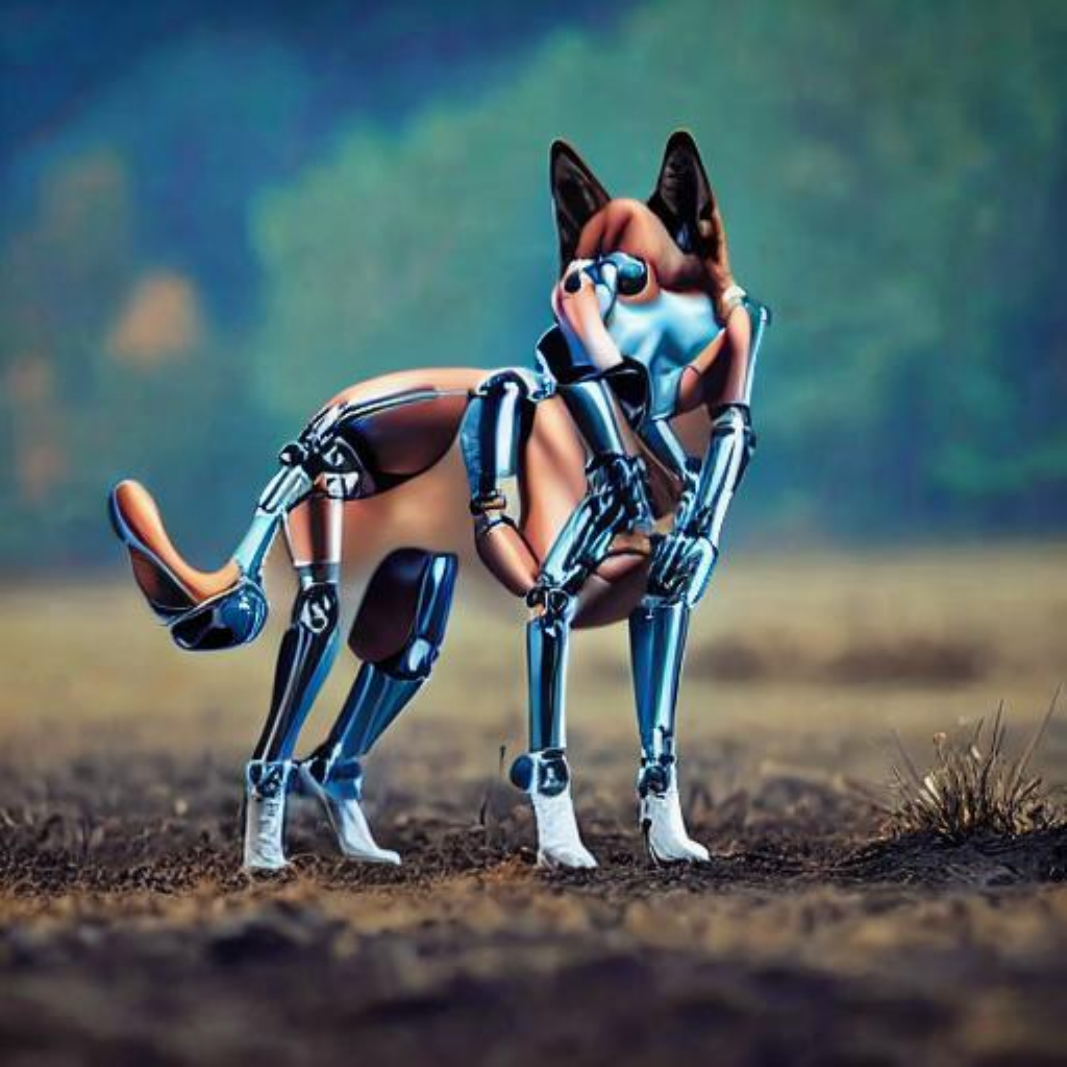 }} &
\raisebox{-.5\height}{
\includegraphics[width=0.235\linewidth]{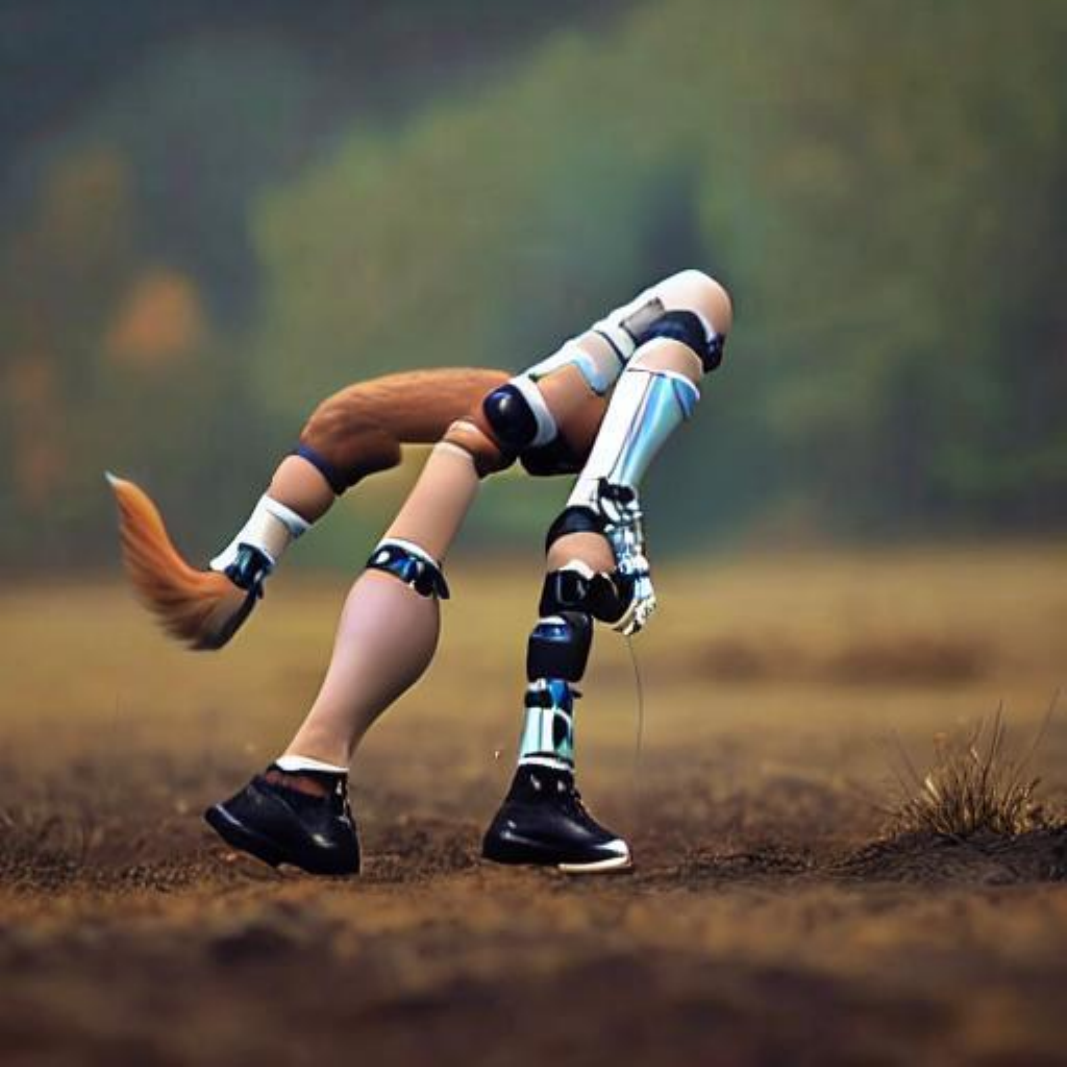}} \\
& \multicolumn{4}{c}{\footnotesize{\textbf{Local}. Baselines struggle to perform intricate local edits.}}\\
    \vspace{-0.6cm} %
\end{tabular}
\caption{Failure cases of baseline instruction-based image editing models.\protect\footnotemark}
\label{fig:comp_our_images} 
    \vspace{-0.5cm} %
\end{figure*}

\section{Method}
\label{sec:method}

\model is a diffusion model designed to multi-task across a broad spectrum of editing tasks.
These include region-based and free-form image editing tasks, as well as traditional computer vision tasks like detection, segmentation, and depth estimation, all of which are formulated as generative tasks. 
As \model is trained on various tasks, a crucial aspect is the ability to identify the semantic edit (e.g., global/local/texture) that needs to be applied, based on the user instruction. 
However, in cases where the instruction is very unique (such as "fix the bumper" in Fig.~\ref{fig:task_cond_ablat}), or when there is ambiguity regarding the edit type (e.g. "Change the sky to be gray" in Fig.~\ref{fig:task_cond_ablat} can be interpreted as both Global edit and Texture edit), the model may encounter difficulty determining the expected edit type. 
To provide the model with a strong condition that will steer the generation process toward the correct task, we propose learning a unique task embedding  for each task, which we integrate into the model.
During training, the task embeddings are learned together with model weights. 
Post training, \model is able to adapt to new tasks via few-shot learning a new task embedding, leaving the rest of the model frozen. 
Last, we introduce a method to preserve the quality of the generated images in multi-turn editing scenarios.
We follow next with a detailed description of each part of our method.
\subsection{Architecture}
Our model builds upon the foundation set by Emu, which is outlined in~\cite{emu}. 
The Emu model is a two-stage approach that begins with a pre-training phase and concludes with a quality fine-tuning stage. The pivotal aspect of the method is that the fine-tuning dataset is relatively small, comprising only a few thousand images, but must be of exceptional quality, often necessitating manual annotation. Emu adapted the latent diffusion model architecture~\cite{rombach2021highresolution} to support high-resolution image generation and incorporated a 16-channel autoencoder with encoder $E$ and decoder $D$. 
In the following section, we adapt the notation of~\cite{brooks2023instructpix2pix}.\footnotetext{The samples depicted in this caption were selected by the authors. They do not cover all scenarios, but are meant to represent some common scenarios the authors encountered.}
A large U-Net, $\epsilon_\theta$, with 2.8 billion parameters, $\theta$, text embeddings from CLIP ViT-L~\cite{radford2021learning} and T5-XXL~\cite{2020t5}, and a substantial pre-training dataset of 1.1 billion images facilitate the model's ability to learn complex semantics and finer details, with a noise-offset strategy contributing to high-contrast and aesthetically pleasing image generation. Given the encoded latent of an image $z=E(x)$, the diffusion process generates a noisy latent $z_t$ where the noise level increases over timesteps $t \in T$.
To convert Emu to an instruction-based image editing model, we condition it on the image to be modified $c_I$ and the instruction $c_T$. \model is trained to minimize the following optimization problem,
\begin{equation}
\min_{\theta}{\mathbb{E}_{y,\epsilon,t}{\left[ \| \epsilon - \epsilon_\theta(z_t,t,E(c_I),c_T) \|_2^2 \right]}}
\label{eq:loss}
\end{equation}
where $\epsilon \in N(0,1)$ is the noise added by the diffusion process and $y=(c_T,c_I,x)$ is a triplet of instruction, input image and target image from the dataset. In practice, we initialize the weights of \model with the weights of Emu. To support the image conditioning, we follow~\cite{brooks2023instructpix2pix} and increase the number of input channels. New weights are initialized to zero. During inference, we perform classifier-free guidance on both image and text conditions. In our experiments we use a scale of $\gamma_I=1.5$ for the image condition and $\gamma_T=5.0$ for the text condition. 
Furthermore, we apply a rescaling of the diffusion scheduler to achieve a zero signal-to-noise ratio (SNR) at the terminal timestamp, as suggested in \cite{lin2023common}. This is crucial in order to avoid any mismatch between the model's training and testing phases. For more implementation details see Sec.~\ref{sec:impl_details}.

\subsection{Learned Task Embeddings}
\label{sec:text_to_task}

To guide the generation process toward the correct task, we learn an embedding vector for each task in the dataset. 
During training, given a sample from our dataset, we use the task index, $i$, to fetch the task's embedding vector, $v_i$, from an embedding table, and optimize it jointly with the model weights. 
We do so by introducing the task embedding $v_i$ as an additional condition to the U-Net, $\epsilon_\theta$.
Concretely, we integrate the task embedding into the U-Net via cross-attention interactions, and by adding it to the timestep embeddings. 
The optimization problem is updated to
\begin{equation}
\min_{\theta,v_1,\dots,v_k}{\mathbb{E}_{\hat{y},\epsilon,t}{\left[ \| \epsilon - \epsilon_\theta(z_t,t,E(c_I),c_T,v_i) \|_2^2 \right]}}
\label{eq:loss_task_emb}
\end{equation}
where $k$ is the total number of tasks in our dataset and $\hat{y}=(c_I,c_T,x,i)$ is a quadruplet of input image, input instruction text, target image, and task index from the dataset.

\begin{figure}[t]
   \centering
\begin{tabular}{@{\hspace{-1\tabcolsep}}c@{\hspace{0.001\tabcolsep}}c@{\hspace{-0.3\tabcolsep}}c@{\hspace{-0.3\tabcolsep}}c}

\resizebox{!}{22px}{
\begin{tabular}[x]{@{}c@{}}(1) \\ \textit{Change} \\ \textit{the sky} \\ \textit{to be} \\ \textit{gray} \end{tabular}}&
\raisebox{-.5\height}{
\includegraphics[width=0.3\linewidth]{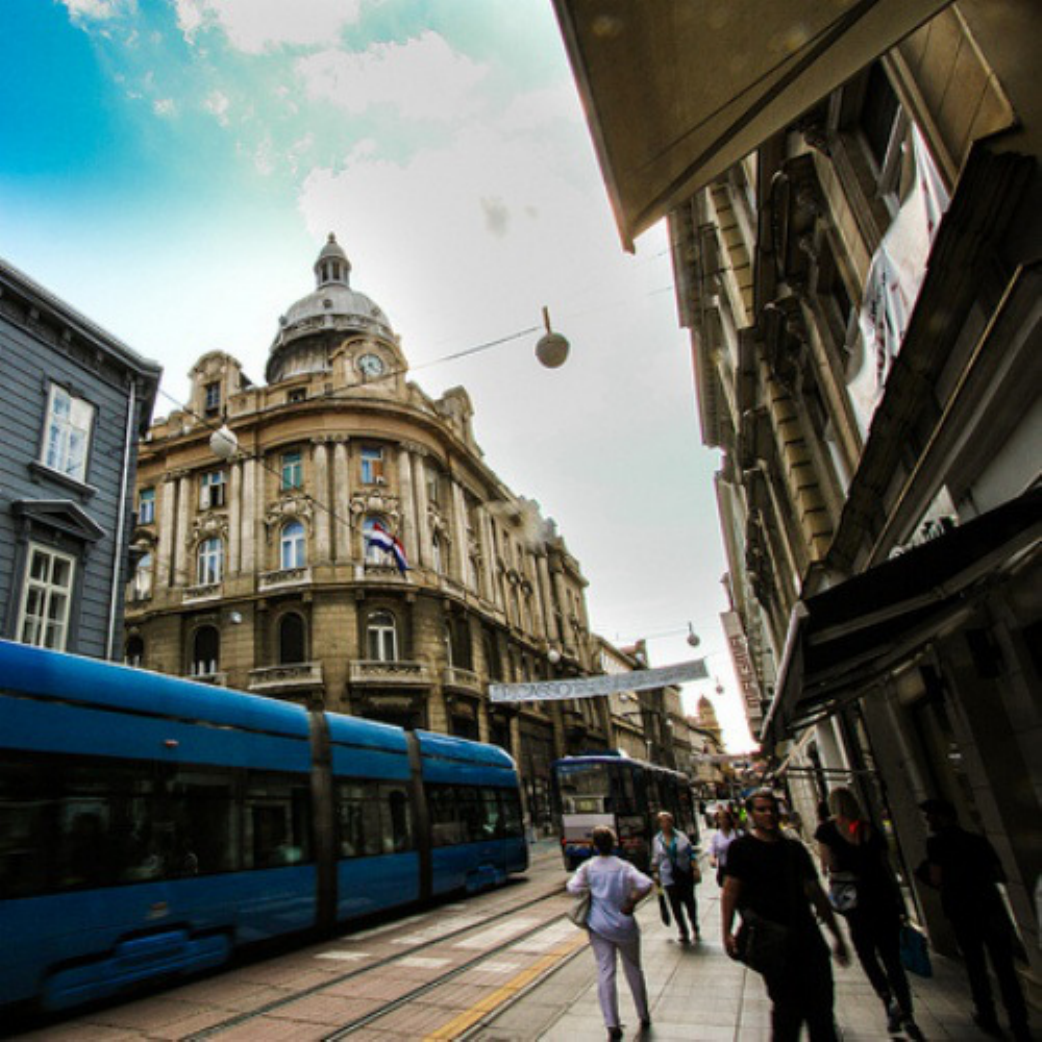}}&
\raisebox{-.5\height}{
\includegraphics[width=0.3\linewidth]{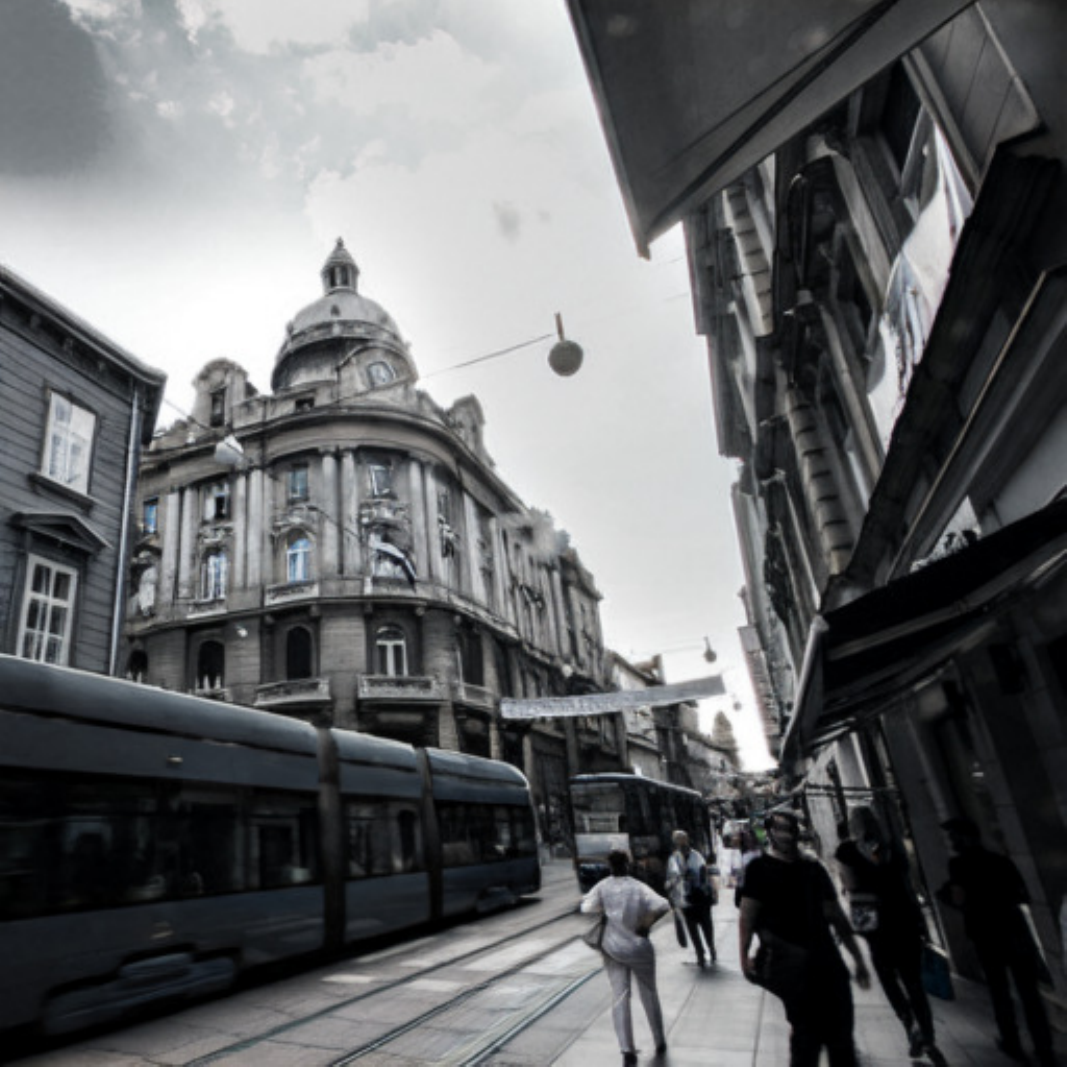}}&
\raisebox{-.5\height}{
\includegraphics[width=0.3\linewidth]{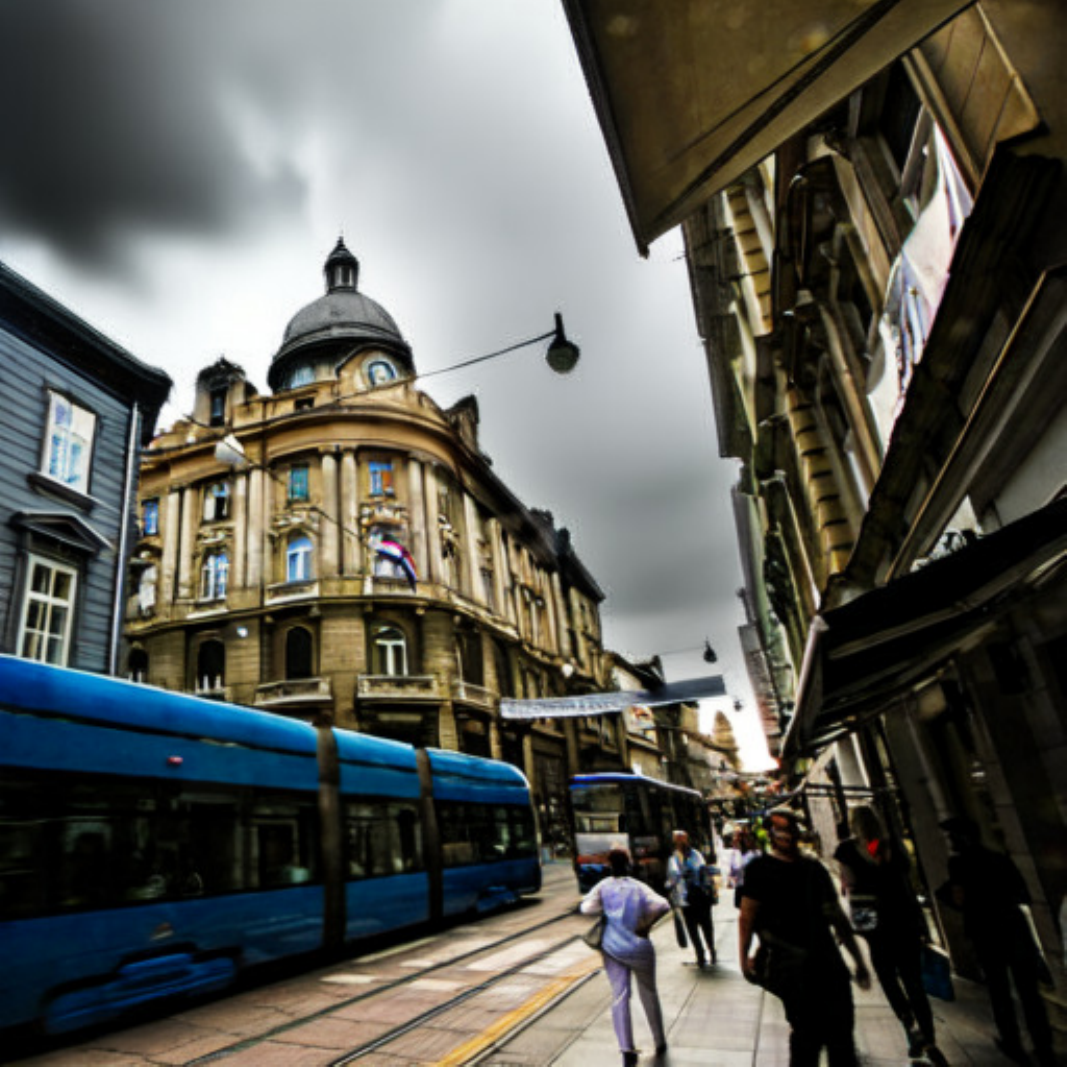}} \\[11.5mm] 

\resizebox{!}{22px}{
\begin{tabular}[x]{@{}c@{}}(2) \\ \textit{Fix the} \\ \textit{bumper} \\ \textit{of the} \\ \textit{vehicle} \end{tabular}}&
\raisebox{-.5\height}{
\includegraphics[width=0.3\linewidth]{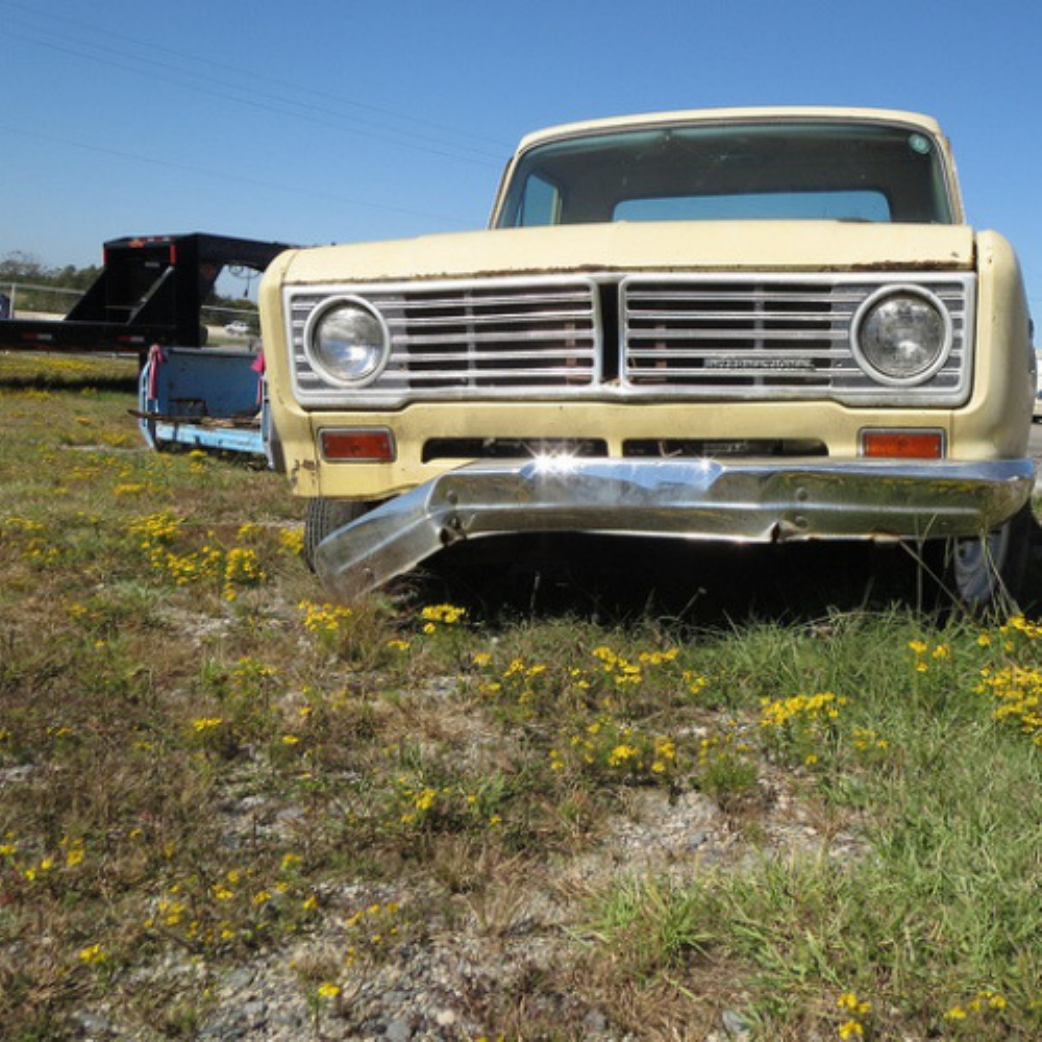}}&
\raisebox{-.5\height}{
\includegraphics[width=0.3\linewidth]{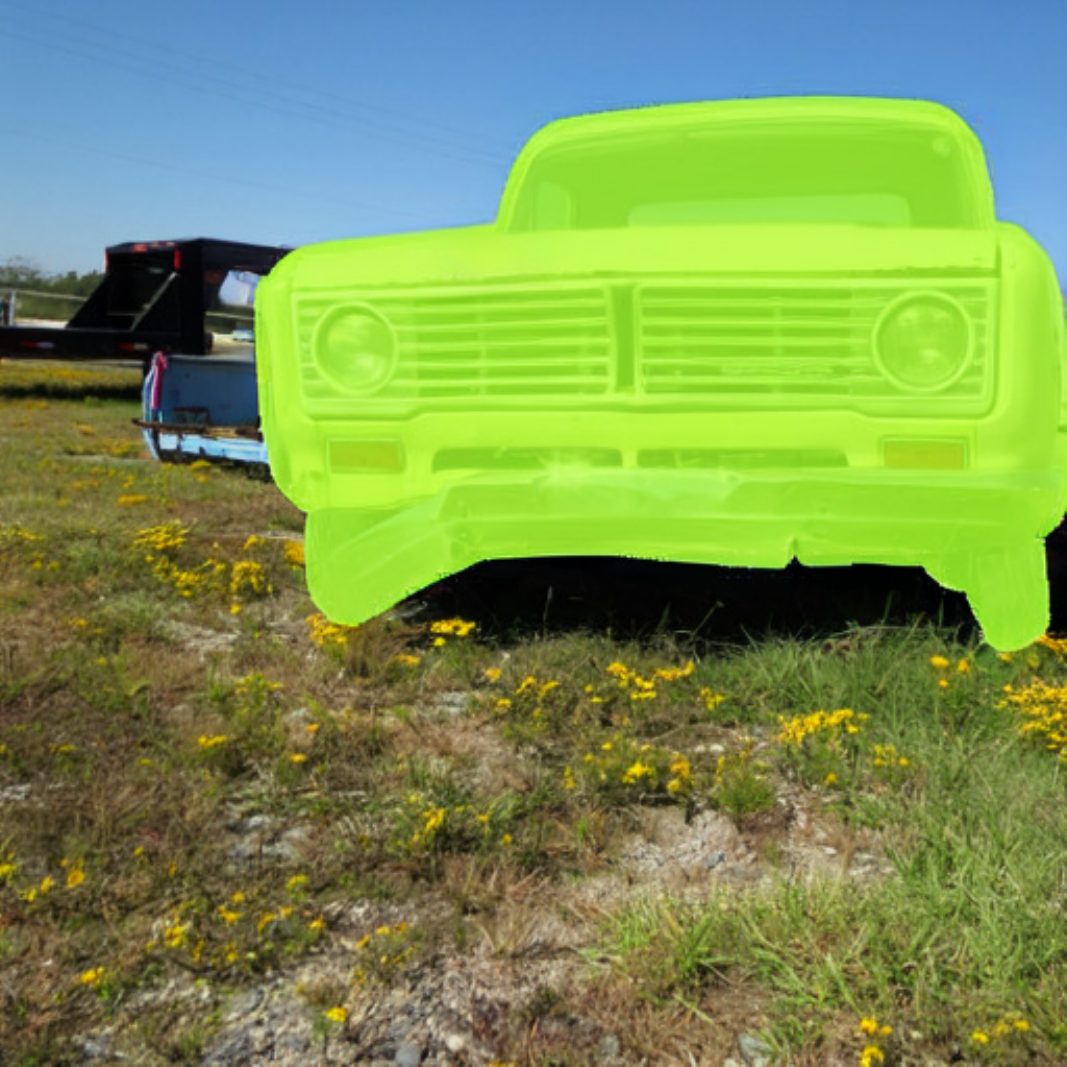}}&
\raisebox{-.5\height}{
\includegraphics[width=0.3\linewidth]{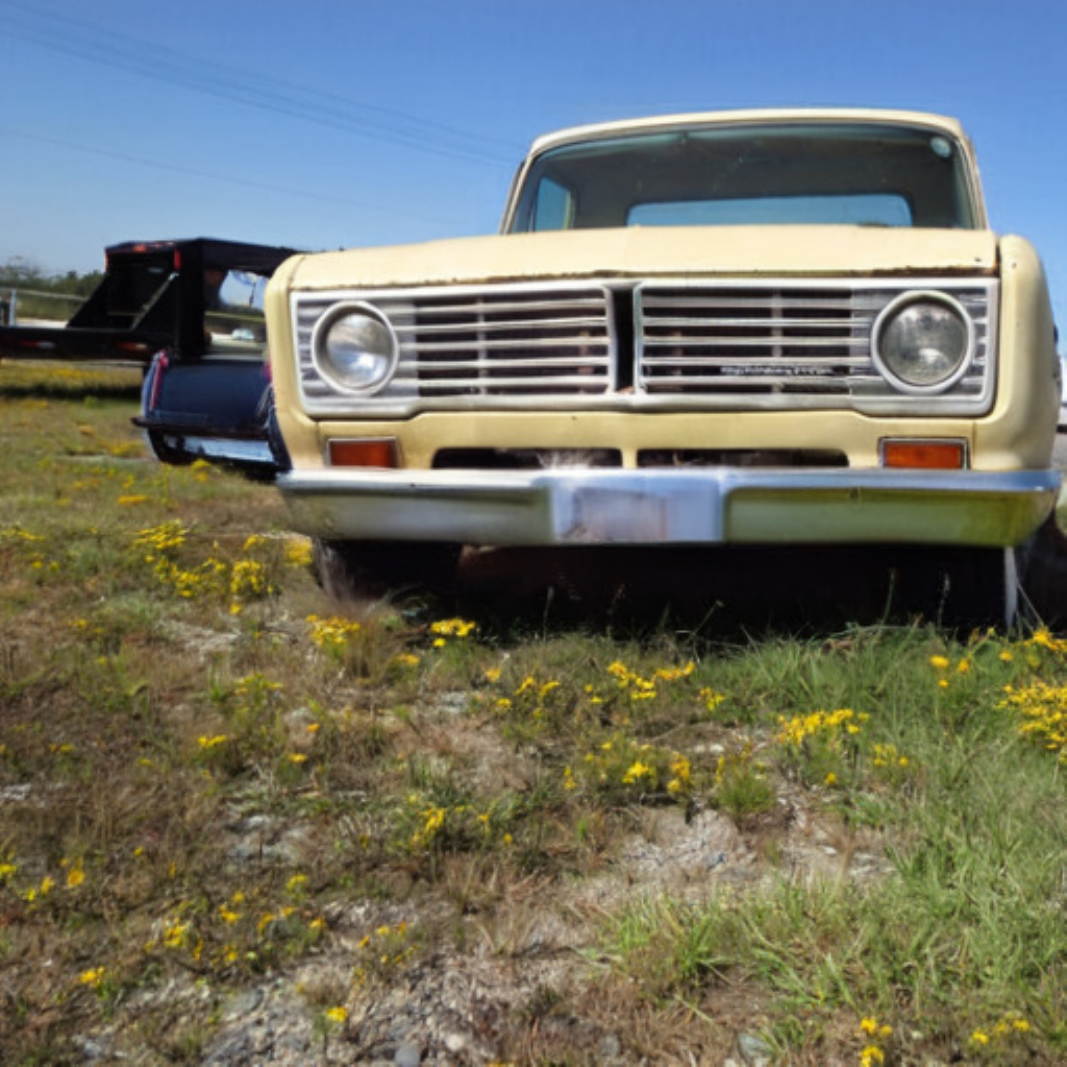}} \\[11.5mm] 

\resizebox{!}{27px}{
\begin{tabular}[x]{@{}c@{}}(3) \\ \textit{Turn the} \\ \textit{television} \\ \textit{into a} \\ \textit{claude} \\ \textit{monet} \\ \textit{painting} \end{tabular}}&
\raisebox{-.5\height}{
\includegraphics[width=0.3\linewidth]{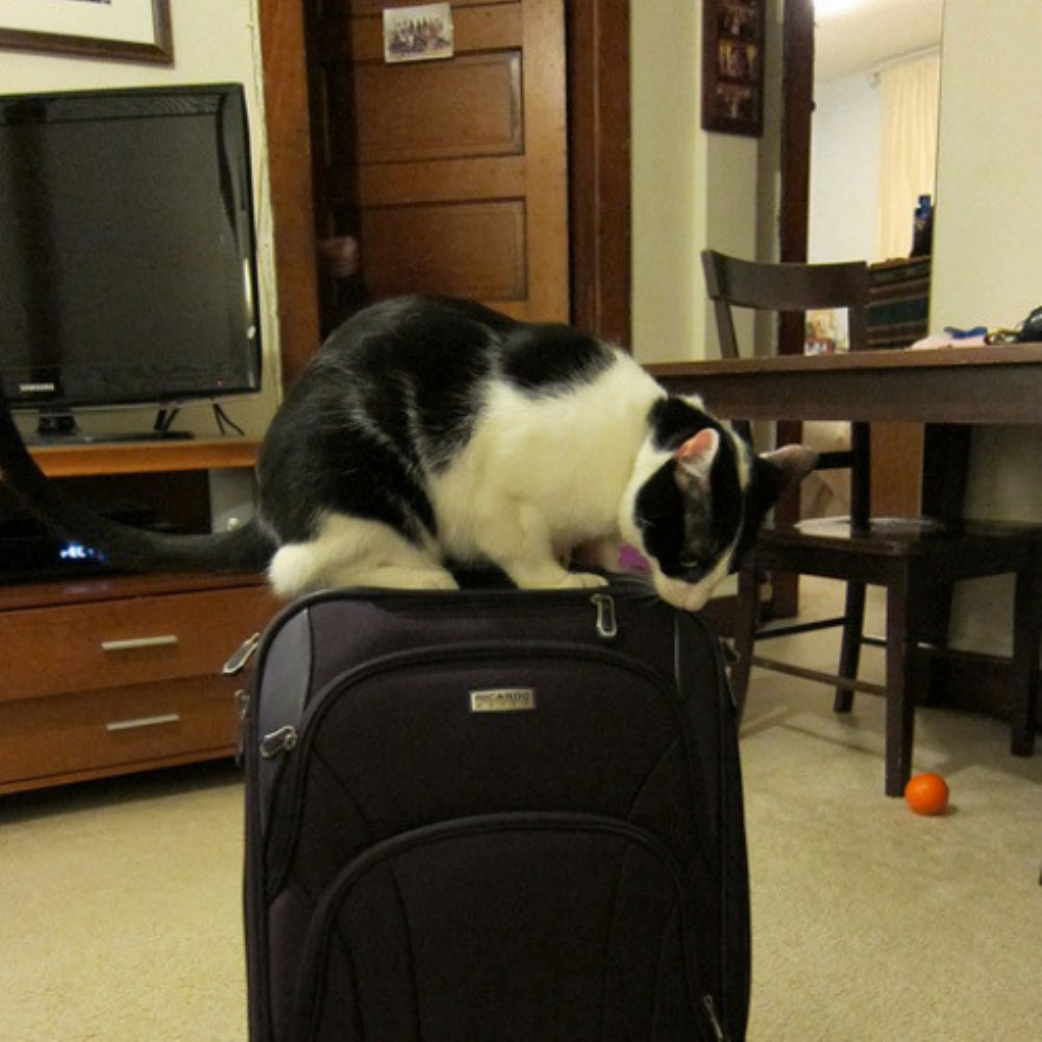}}&
\raisebox{-.5\height}{
\includegraphics[width=0.3\linewidth]{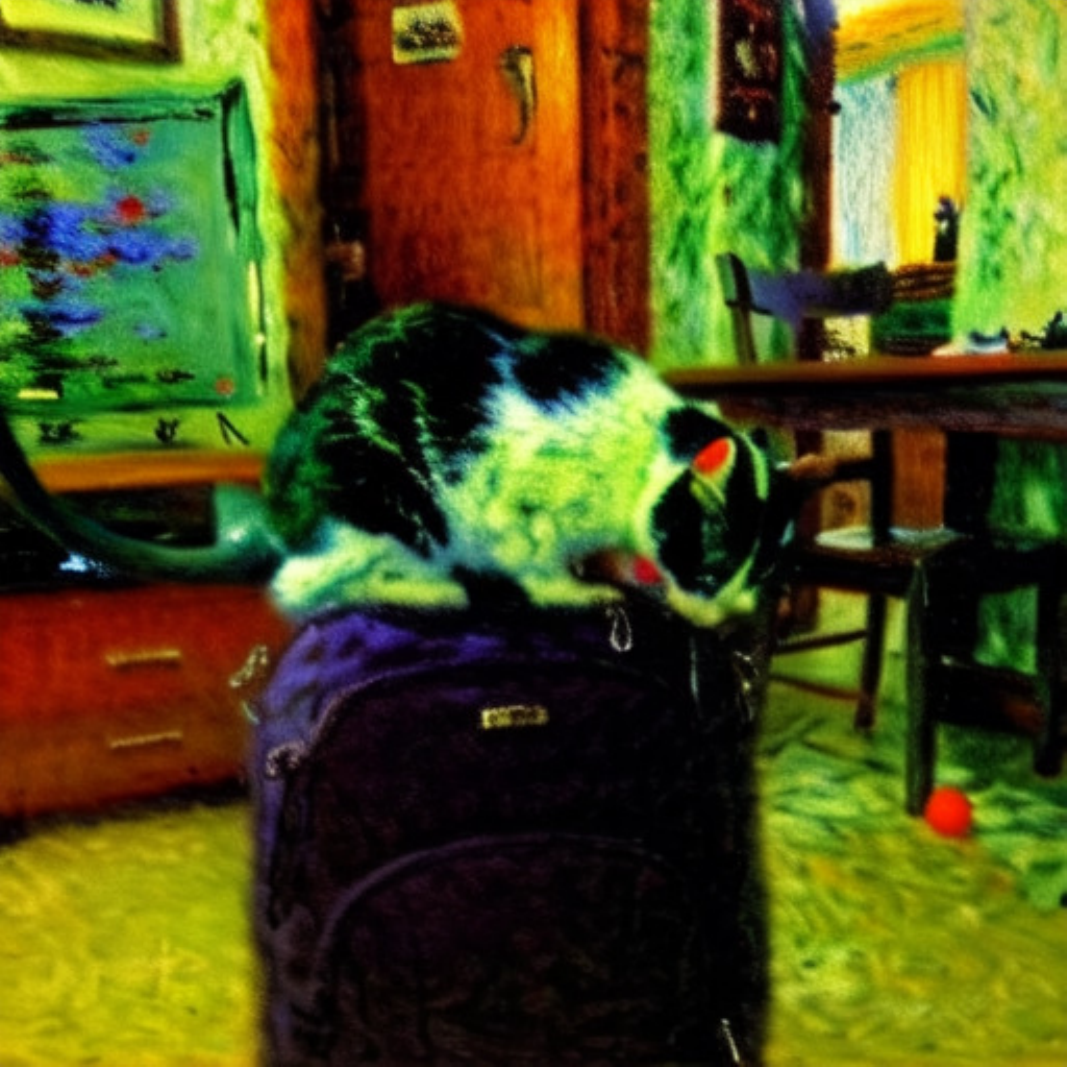}}&
\raisebox{-.5\height}{
\includegraphics[width=0.3\linewidth]{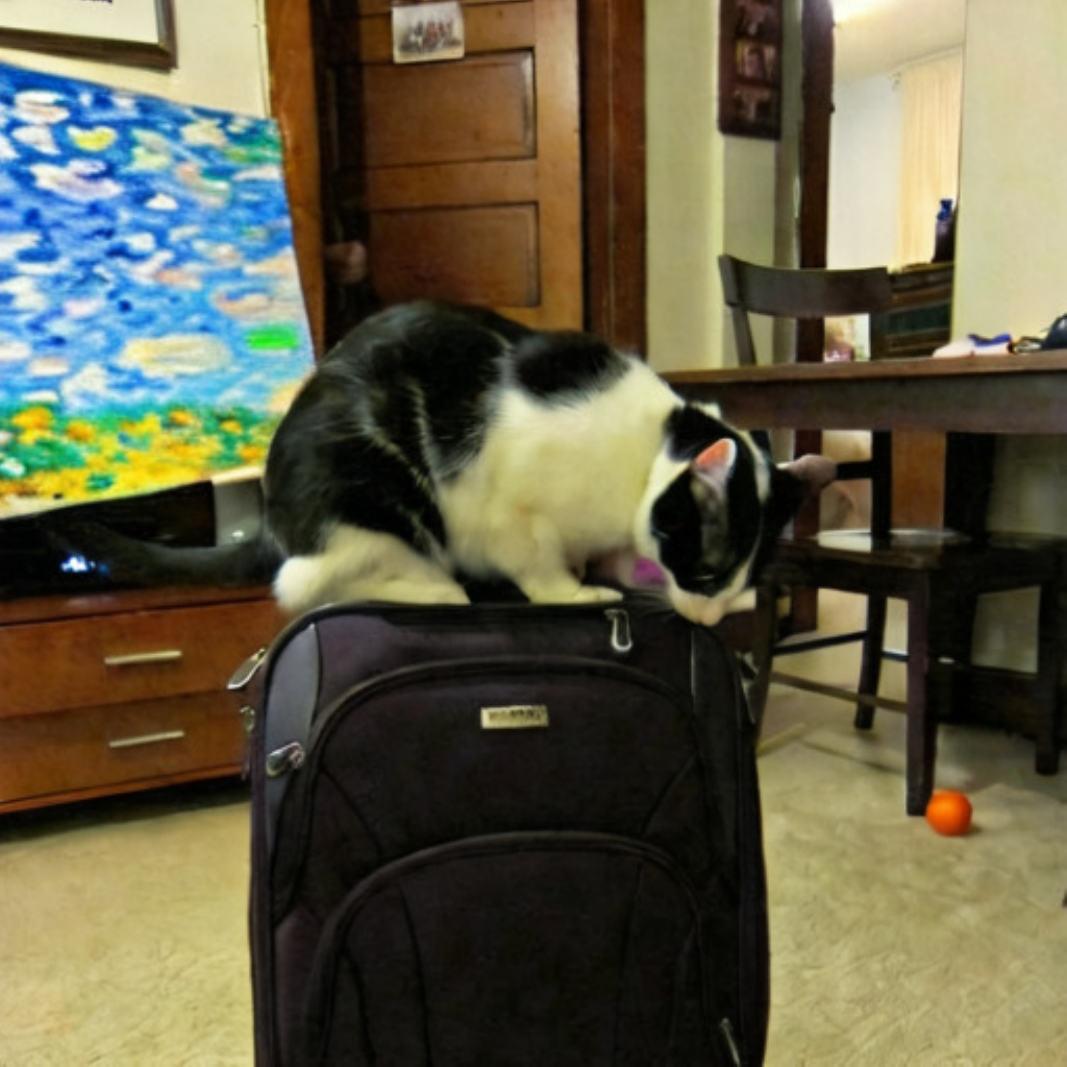}} \\
& \begin{tabular}[x]{@{}c@{}}Input \end{tabular}  & \begin{tabular}[x]{@{}c@{}}Without task emb. \end{tabular} & With task emb.   \\
\end{tabular}
\caption{\textbf{Task embeddings. } Model trained without task embeddings may get confused about the edit type when the instructions are complicated or there is ambiguity regarding the edit type: (1) Global edit (instead of Texture), (2) Segmentation (instead of Global), (3) Style edit (instead of Local).}
\label{fig:task_cond_ablat} 
\vspace{-0.6cm}
\end{figure}

Task-specific conditioning arises from the observation that models lacking such conditioning can become perplexed about the type of edit required, particularly when the instructions are complex or the edit type is ambiguous. For instance, as visualized in Fig.~\ref{fig:task_cond_ablat}, (1) a model without task conditioning might perform a global edit when a texture edit is required, (2) it might opt for segmentation when a global edit is necessary, and (3) it could implement a style edit in situations where a local edit would fit better.

In the inference stage, we predict the task index. 
Specifically, we fine-tune a Flan-T5-XL model to identify the task at hand given the input instruction.

\subsection{Task Inversion}\label{task_inversion}
To enable few-shot learning of new tasks without losing the general abilities of \model, we propose a method for adapting the model without changing the U-Net weights.
Given a few examples of a new task, we learn a new task embedding, $v_{\text{new}}$. 
We freeze the model weights, and adapt it to the task only through the task embedding.
Thus, to fit a new task embedding we solve the following optimization problem:
\begin{equation}
\min_{v_{\text{new}}}{\mathbb{E}_{y,\epsilon,t}{\left[ \| \epsilon - \epsilon_\theta(z_t,t,E(c_I),c_T,v_{\text{new}}) \|_2^2 \right]}}
\label{eq:loss_task}
\end{equation}
where $v_{\text{new}}$ is the learned task embedding. Note that during task inversion $y$ is a triplet belonging to the new task.

The model can then be employed for the new task by conditioning it on the learned task embedding, and it can still handle its original tasks by relying on the initial task embeddings.
In Sec.~\ref{sec:few_shot}, we demonstrate that our model effectively generalizes to novel tasks using this method. 

\subsection{Sequential Edit Thresholding} 
We notice that applying the model repeatedly, in multi-turn editing scenarios, aggregates reconstruction and numerical errors, which translate to noticeable artifacts.
To mitigate this, we add a per-pixel thresholding step after each edit-turn. 
At each step $s$, we use the pixel value in the output image, $c^{s+1}_I$, only if its alteration surpasses a specific threshold.
Otherwise, we keep the pixel value from the input image, $c^{s}_I$. 
Specifically, given an edit turn $s$, we compute the absolute difference image $d=\| c^{s+1}_I - c^s_I \|_1$ over the RGB channel, and apply the following thresholding:

\begin{equation}
c^{s+1}_I = 
  \begin{cases} 
   c^s_I & \text{if } \bar{d} < \alpha, \\
   c^{s+1}_I & \text{otherwise}.
  \end{cases}
\label{eq:pixel_update}
\end{equation}
where, $\bar{d}$ is obtained after passing $d$ through a low pass filter, in order to smooth the transition between previous and current pixels. In practice, we choose $\alpha=0.03$. Please refer to Sec.~\ref{sec:multi_turn_qualitative_resutls} for a qualitative comparison. In Fig.~\ref{fig:cat} we show examples of multi-turn editing.

\setlength{\tabcolsep}{2pt}
\begin{table*}[h!]
\centering
\caption{Comparison with image-editing baselines evaluated on \model test set and MagicBrush test set. For each benchmark we report CLIP, L1, DINO metrics and human ratings. Human evaluation shows the percentage of raters that prefer the results of \model.} 
\label{tab:eval_image_editing}
\scalebox{0.85}{
\centering
\begin{tabular}{lccccc|cc|ccccc|cc}
\toprule
 & \multicolumn{7}{c|}{\model Test set} & \multicolumn{7}{c}{MagicBrush Test Set}\\
\midrule
Method & $\text{CLIP}_{dir}\!\uparrow$ &  $\text{CLIP}_{im}\!\uparrow$ & $\text{CLIP}_{out}\!\uparrow$ &  $\text{L1}\!\downarrow$  & DINO$\uparrow$ & Text & Image & $\text{CLIP}_{dir}\!\uparrow$ &  $\text{CLIP}_{im}\!\uparrow$ & $\text{CLIP}_{out}\!\uparrow$ &  $\text{L1}\!\downarrow$ & DINO$\uparrow$ & Text & Image \\
& & & & & & align. & faith. & & & & & & align. & faith. \\
\midrule
InstructPix2Pix~\cite{brooks2023instructpix2pix} & 0.078 & 0.834 & 0.219 & 0.121 & 0.762 & 77.33 & 76.71
                & 0.115 & 0.837 & 0.245 & 0.093 & 0.767 & 71.79 & 71.60 \\
MagicBrush~\cite{zhang2023magicbrush}      & 0.090 & 0.838 & 0.222 & 0.100 & 0.776 & 74.50 & 74.10
                & 0.123 & 0.883 & 0.261 & 0.058 & 0.871 & 59.54 & 60.39 \\
PnP~\cite{pnp}             & 0.028 & 0.521 & 0.089 & 0.304 & 0.153 & 98.95 & 99.00
                & 0.025 & 0.568 & 0.101 & 0.289 & 0.220 & 97.24 & 96.96 \\
Null-Text Inv.~\cite{Mokady2022NulltextIF} & 0.101 & 0.761 & \textbf{0.236} & \textbf{0.075} & 0.678 & 81.63 & 85.47
                & 0.121 & 0.752 & \textbf{0.263}  & 0.077 & 0.664 & 76.54 & 85.66 \\
Our             & \textbf{0.109} & \textbf{0.859} & 0.231 & 0.094 & \textbf{0.819} & -- & --
                & \textbf{0.135} & \textbf{0.897} & 0.261 & \textbf{0.052} & \textbf{0.879} & -- & -- \\
\bottomrule
\end{tabular}}
\vspace{-0.5cm}
\end{table*}

\section{Experiments}
\label{sec:exp}
Our experiments evaluate the ability of \model to follow user instructions faithfully and preserve the visual fidelity of the original image. %
First, we evaluate the performance of our approach on instruction-based image editing tasks.
Second, we conduct a comprehensive ablation study to assess the effectiveness of our different contributions.
Specifically, we ablate the contribution of the computer vision tasks to the model performance on image editing tasks, the importance of learned task embeddings, and the effect of multi-task learning on instruction-based image editing.  
Further ablation on our data generation pipeline can be found in Appendix~\ref{sec:data_exp}.
Finally, we demonstrate our model's ability to learn new tasks via few-shot learning. 

\subsection{Measures} %
We employ two main measures in our evaluation: edit text alignment and image faithfulness. Specifically, for each pair of input image and editing instruction, we use the following automatic metrics: (i) CLIP~\cite{radford2021learning} text-image direction similarity~($\text{CLIP}_{dir}$) -- measuring agreement between change in captions and the change in images, (ii) CLIP image similarity~($\text{CLIP}_{img}$) -- measuring change between edited and input image, (iii) CLIP output similarity~($\text{CLIP}_{out}$) -- measuring edited image similarity with output caption, (iv) L1 pixel-distance between input and edit image, and (v) DINO~\cite{caron2021emerging} similarity between the DINO embeddings of input and edited images. 
With the exception of the L1 distance, where lower values indicate better performance, higher values in all other measures signify better results.
A low L1 distance translates to small changes in image's pixel values. A high DINO and $\text{CLIP}_{img}$ similarity score, suggests semantic similarity between the images. For region-based edits, high image similarity scores indicate the edits were precise. For free-form edits, high similarity scores indicate image structure preservation. $\text{CLIP}_{dir}$ and $\text{CLIP}_{out}$ measure how well the model followed the instruction.

In addition, we asked human raters to evaluate the text alignment and image faithfulness. In each evaluation scenario, raters are presented with two modified images alongside the original input image and instruction, and are presented with two questions: (i) Image Faithfulness: which image better preserves elements in the input image, and (ii) Text Alignment: which image best follows the instruction.

\subsection{Evaluation} 
\label{sec:eval}
Throughout the paper, we report results on the MagicBrush test set~\cite{zhang2023magicbrush} and the \model benchmark.  
In the following section, we describe our motivation for creating this benchmark, and detail its curation process.
To date, there are two main benchmarks for evaluating instruction-based image editing capabilities. 
First, the InstructPix2Pix benchmark~\cite{brooks2023instructpix2pix}, which is intrinsically biased due to its reliance on \textit{generated} Stable Diffusion~\cite{Rombach2021HighResolutionIS} input images, and GPT-3~\cite{gpt3} \textit{generated} instructions. 
Consequently, it is unclear whether its results will truly mirror the performance on \textit{real} input images, with \textit{genuine} user instructions.

Unlike InstructPix2Pix, the second benchmark, MagicBrush~\cite{zhang2023magicbrush}, uses a diverse set of authentic input images from the MS-COCO benchmark~\cite{coco,coco2}, and annotator-defined instructions.
Nonetheless, this dataset also suffers from inherent bias.
During data collection, annotators were directed to use the DALLE-2 image editing platform~\cite{dalle2_tool} to generate the edited images.
Thus, this benchmark is biased towards editing instructions that the DALLE-2 editor can successfully follow, which may compromise both its diversity and complexity.

\myparagraph{\model Benchmark.} To collect a dataset with reduced bias and of higher diversity, we take a different approach. 
We first define seven different categories of potential image editing operations: background alteration (Background), comprehensive image changes (Global), style alteration (Style), object removal (Remove), object addition (Add), localized modifications (Local), and color/texture alterations (Texture).
Then, we utilize the diverse set of input images from the MagicBrush benchmark~\cite{zhang2023magicbrush}, and for each editing operation, we task crowd workers to devise relevant, creative, and challenging instructions.
Moreover, to increase the quality of the collected examples, we apply a post-verification stage, in which crowd workers filter examples with irrelevant instructions.
Finally, to support evaluation for methods that require input and output captions~\cite{prompt2prompt,pnp}, we additionally collect an input caption and output caption. 
When doing so, we ask annotators to ensure that the captions capture both important elements in the image, and elements that should change based on the instruction.    
See Sec.~\ref{sec:instruction_based_image_editing_benchmark} for examples of our benchmark, which we publicly release to support better evaluation of instruction-based image editing models, and more details on the benchmark curation process.

\subsection{Baseline Comparisons}
We compare our model against two instruction-based image editing baseline models: InstructPix2Pix~\cite{brooks2023instructpix2pix}, and MagicBrush~\cite{zhang2023magicbrush}, which is a variant of InstructPix2Pix that was fine-tuned on the MagicBrush dataset.
Additionally, we compare our model against two text-based image editing methods: PNP~\cite{pnp} and Null-Text Inversion modification of P2P~\cite{prompt2prompt, Mokady2022NulltextIF}.
Unlike instruction-based models, these works expect image descriptions. Therefore, we provide them with access to the input caption and output caption.
Do note, providing these methods with access to the ground-truth captions could potentially offer an edge over instruction-based models, since the automatic metrics also rely on these captions.
Tab.~\ref{tab:eval_image_editing} shows our results versus the baselines. 
The findings indicate that human raters consistently prefer \model over all baselines by a large margin.
Furthermore, apart from Null-Text Inversion, which as explained above, utilizes the ground-truth captions during inference, our approach outperforms the existing baselines on the automatic metrics. We provide qualitative comparisons in Fig.~\ref{fig:comp_our_images}, Fig.\ref{fig:comp_our_images_sup}-\ref{fig:comp_our_images_sup2}, and Fig.\ref{fig:comp_our_test}-\ref{fig:comp_our_test2}.
For performance on vision tasks, see Sec.~\ref{sec:vision_eval_sup}.

\subsection{Ablations}

\label{sec:task_condition}

\myparagraph{Computer Vision Tasks Enhance Image Editing Tasks.}
Here we demonstrate the importance of the vision tasks to \model performance on image editing tasks. For this, we trained two additional models on all tasks except: (i) detect and segment tasks, and (ii) image-to-image translation tasks. 
As we show in Tab.~\ref{tab:cv_helps}, adding the detection and segmentation tasks improves the model performance in region-based editing tasks.
Additionally, we observe that image-to-image translation tasks improve the performance in free-form editing tasks. 
We hypothesize that the recognition tasks improve the model's recognition capabilities, leading to more accurate and precise localized modifications. 
Similarly, image-to-image tasks assist the model in understanding the entire image structure,  thereby enhancing its capabilities for global operations.

\myparagraph{Contribution of Learned Task Embeddings.} We compare three variants of \model: (i) conditioned on the ground-truth task embedding, (ii) conditioned on the task embedding, as predicted by the task predictor described in Sec.~\ref{sec:text_to_task}, and (iii) without conditioning on the task type.\footnote{(iii) was trained without learned task embeddings. %
}
Tab.~\ref{table:ablation} shows the results on the validation set of our benchmark.  
As can be seen, conditioning on the task type boosts the model's performance. %
Furthermore, our task predictor closes the gap with the ground-truth conditioned model. 
Qualitatively, we observe that without conditioning on the task type the model may perform the wrong editing operation (Fig.~\ref{fig:task_cond_ablat}). 
In Fig.~\ref{fig:task_cond_control}, we demonstrate the effect of manipulating the task while keeping the instruction and input image fixed. 
As can be seen, changing the task embedding directly influences the task executed by the model.

\setlength{\tabcolsep}{2pt}
\begin{table}[t]
  \centering
  \caption{Learned task embeddings ablation on our validation set. We compare variations of \model: without task type condition, with predicted task type, and with ground-truth task type.}
  \label{table:ablation}
  \scalebox{0.85}{
  \begin{tabular}{l|ccccc}
  \toprule\noalign{\smallskip}
   Method & $\text{CLIP}_{dir}\!\uparrow$ &  $\text{CLIP}_{im}\!\uparrow$ & $\text{CLIP}_{out}\!\uparrow$ &  $\text{L1}\!\downarrow$  & DINO$\uparrow$ \\
  \noalign{\smallskip}
  \hline
  w/o task emb. & 0.104 & 0.843  & 0.227 & 0.109 &  0.792 \\
  with pred. task  & 0.117 & 0.850  & 0.231 & 0.103 &  0.809 \\
  with gt task & 0.119 & 0.852 & 0.231 & 0.100 & 0.811   \\ 
 \bottomrule
  \end{tabular}}
  \vspace{-2mm}
\end{table}

\myparagraph{Multi-Task versus Expert Models.}
We hypothesize that training a single model on a diverse range of tasks, encompassing both image editing and computer vision, leads to enhanced performance in each individual task, outperforming models that are specifically trained for a single task. 
To validate this hypothesis, we train an expert model for each task, and compare its performance to ours on that particular task. In this ablation, we train the models for half the steps of our complete model. %
Results provided in Tab.~\ref{table:experts} show that the multi-task model is superior to all experts models.

\myparagraph{Influence of Number of Tasks.}
Here, we ablate the number of tasks participating in the multi-task training scheme. 
In Fig.~\ref{fig:agg_fig} we report the average $\text{CLIP}_{dir}$ on the Style and Texture tasks when iteratively excluding other tasks, and training a model on the new tasks list.
As can be seen, augmenting the model with additional tasks leads to improved performance, even in tasks which are not directly associated with the added ones.
For instance, the Background task enhances the model's performance on the Texture and Style tasks.

\setlength{\tabcolsep}{2pt}
\begin{table}[t!]
\centering
\caption{Contribution of computer vision tasks. Human evaluation is shown as a percentage of majority votes in favor of our model.}
\label{tab:cv_helps}
\scalebox{0.95}{
\centering
\begin{tabular}{lcc|cc}
\toprule
 & \multicolumn{2}{c}{Region-Based} &  \multicolumn{2}{c}{Free-form}\\
\midrule
\small{Method} & Text &  Image & Text & Image   \\
& align.  & faith. & align. & faith. \\
\midrule
without detect/segment    & \textbf{60.0} & \textbf{60.2} &  52.3 & 51.5 \\
without im2im translation  & 50.2 & 49.0 & \textbf{58.0} & \textbf{60.1}  \\
\bottomrule
\end{tabular}}
\vspace{-0.5cm}
\end{table}

\begin{figure}[t]
   \centering
   \begin{tabular}%
{@{\hspace{-4\tabcolsep}}c@{\hspace{-2\tabcolsep}}c@{\hspace{-2\tabcolsep}}c}
      \resizebox{!}{28px}{
         \begin{tabular}[x]{@{}c@{}} (i) \\ \textit{Incorporate}\\ \textit{a bee into} \\ \textit{the bag's} \\ \textit{pattern and} \\ \textit{detect it} \end{tabular}
      } &
      \raisebox{-.5\height}{
         \includegraphics[width=0.4\linewidth]{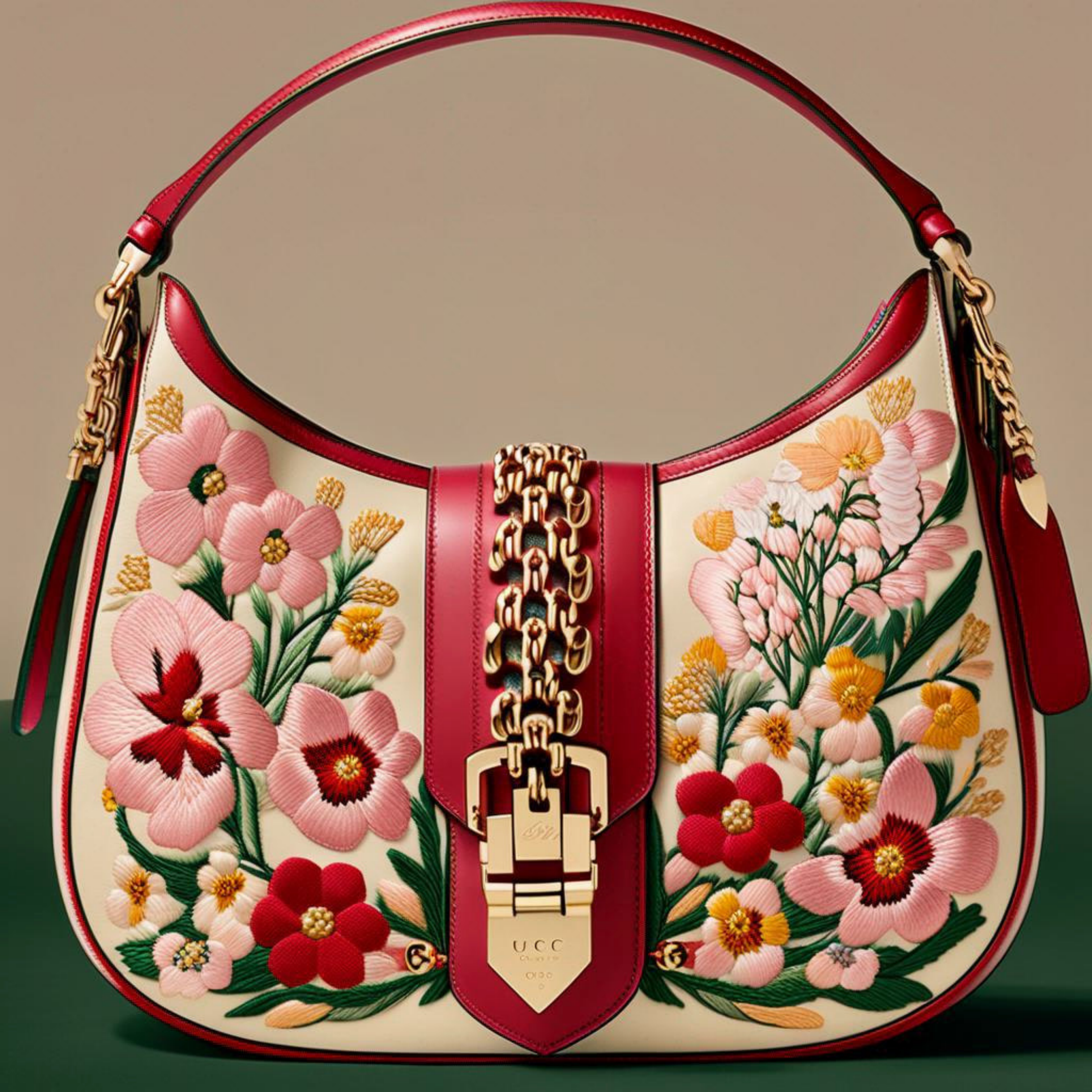}
      } &
      \raisebox{-.5\height}{
         \includegraphics[width=0.4\linewidth]{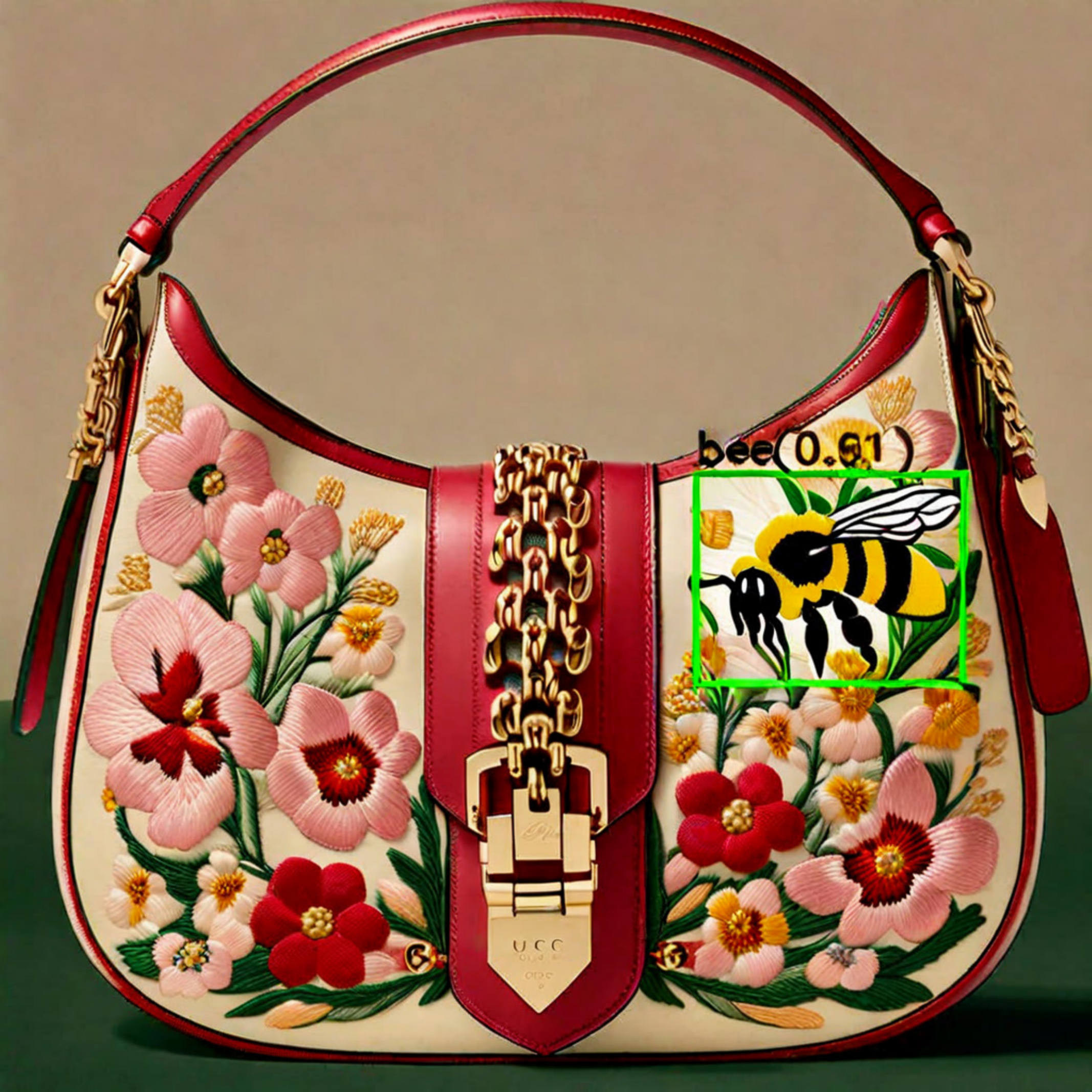}
      } \\ [16mm]
      \resizebox{!}{20px}{
         \begin{tabular}[x]{@{}c@{}} (ii) \\ \textit{Mark the} \\ \textit{gift}\\ \textit{bags} \end{tabular}
      } &
      \raisebox{-.5\height}{
         \includegraphics[width=0.4\linewidth]{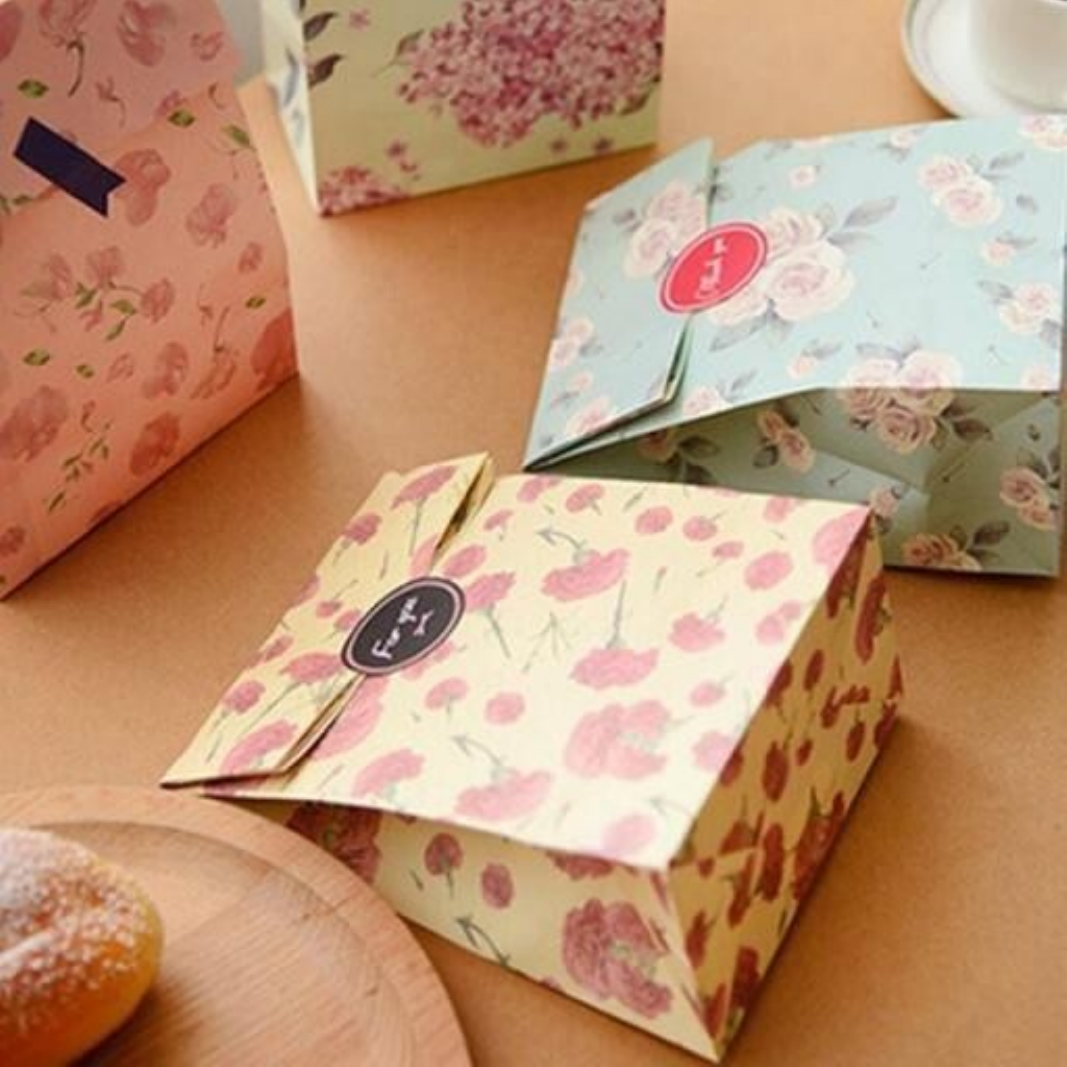}
      } &
      \raisebox{-.5\height}{
         \includegraphics[width=0.4\linewidth]{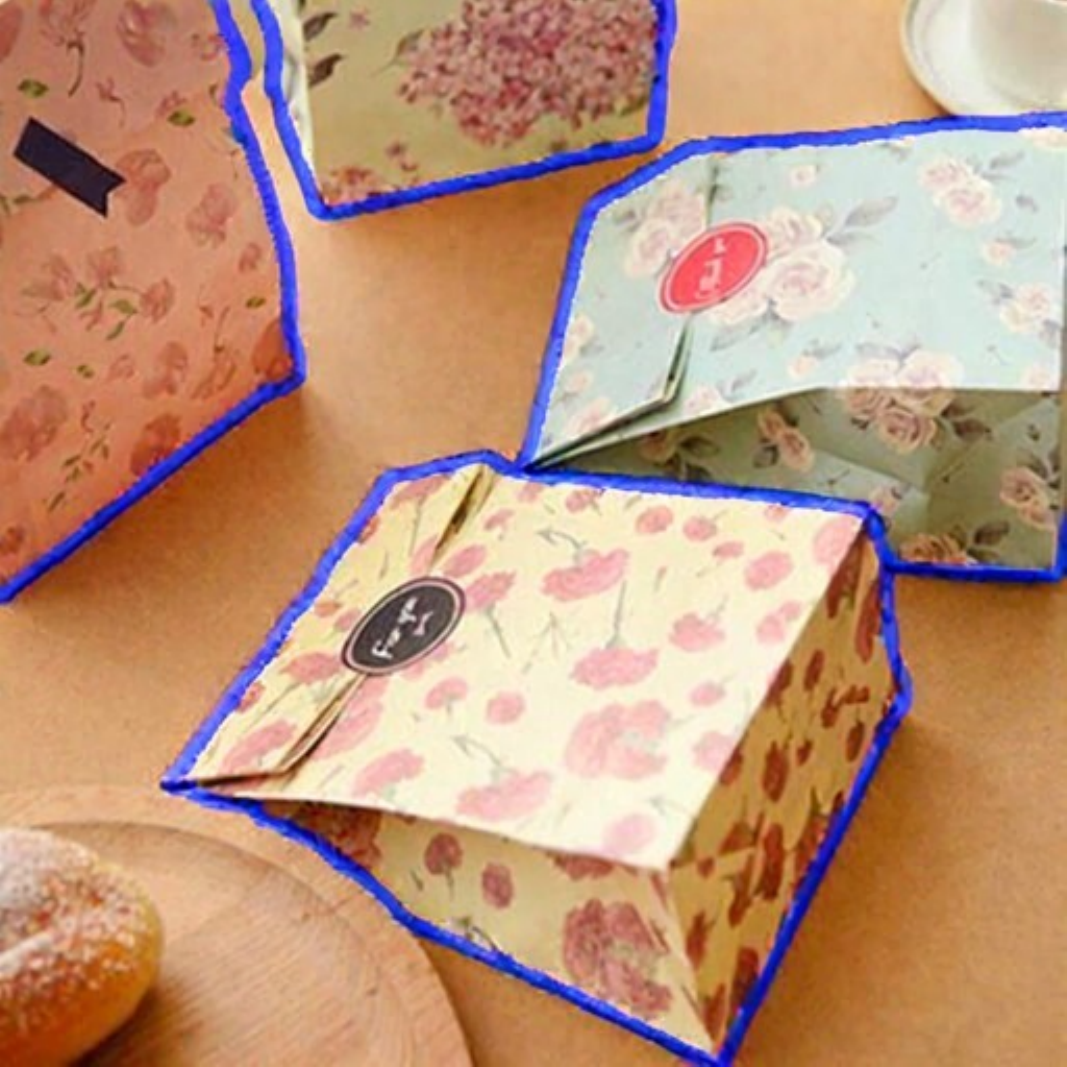}
      } \\
      & \begin{tabular}[x]{@{}c@{}}Input \end{tabular}  & \begin{tabular}[x]{@{}c@{}}\model \end{tabular} \\
   \end{tabular}
   \caption{Generations on unseen tasks with task inversion. (i) composition of add and detect tasks, (ii) object contour detection.}
   \label{fig:few_shot_examples}
\end{figure}

\subsection{Few-Shot Learning of New Tasks}
\label{sec:few_shot}
\begin{figure}
\centering
    \hspace{-0.8cm}
    \subfloat[Inpainting]
    	{\includegraphics[width=0.25\textwidth]
    	{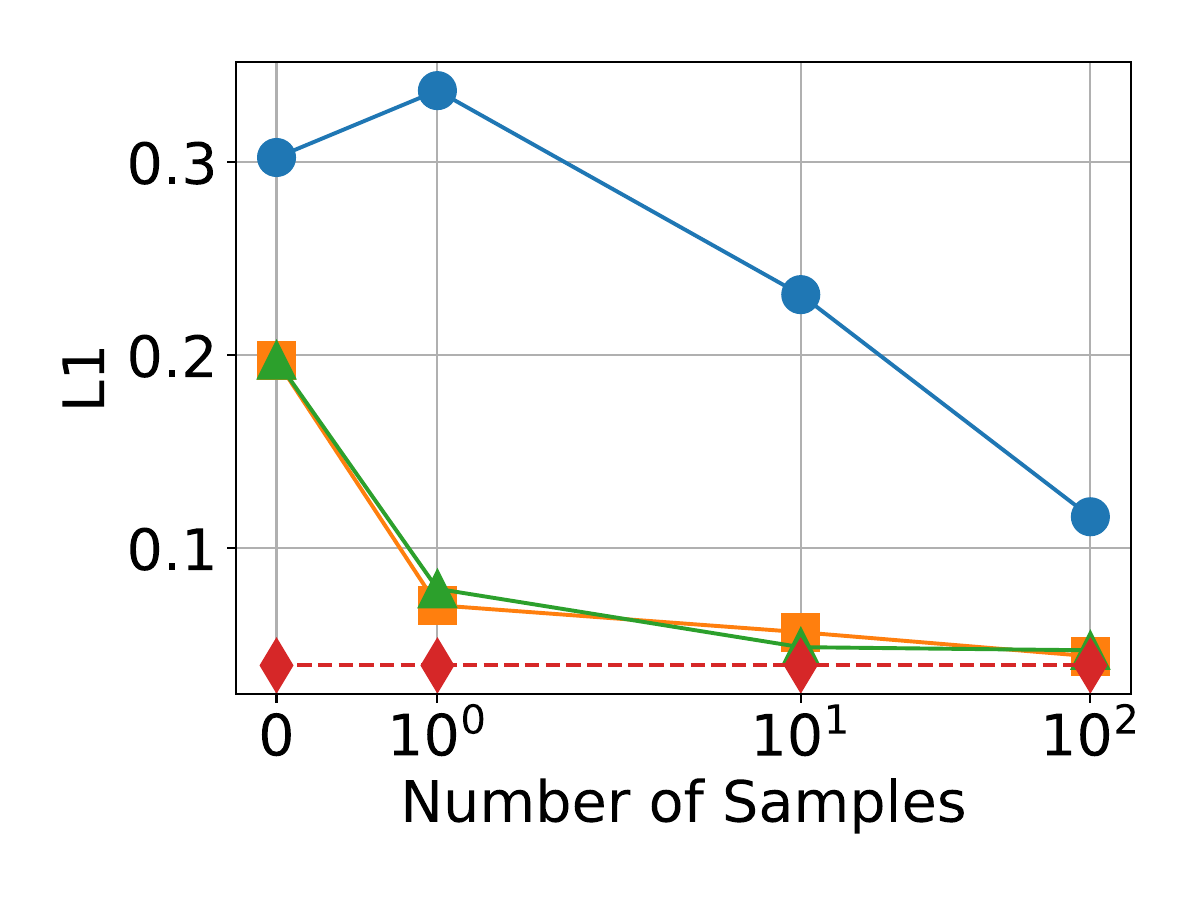}}
    \hspace{-0.2cm}
    \subfloat[Add + Detect]
    	{\includegraphics[width=0.25\textwidth]
    	{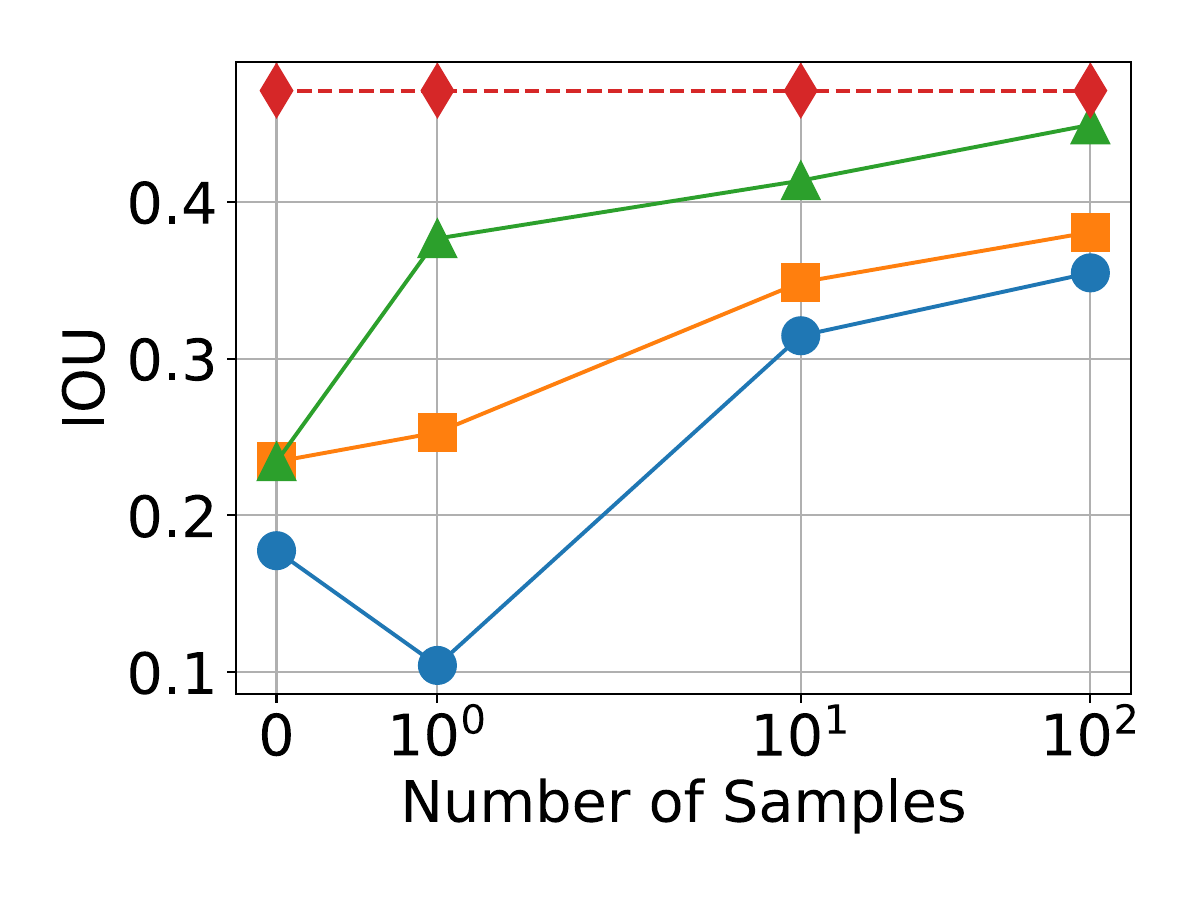}}
    \caption{\label{fig:few-shot}Few-shot performance for different tasks over 1, 10, and 100 samples. Each line represents a different training setting: Emu finetune (Blue, $\bigcirc$), \model finetune (Orange, $\Box$), task inversion (Green, $\triangle$), all compared to an upper-bound expert trained on 100k samples (Red dashed line, $\diamondsuit$).}
    \label{fig:few-shot_sup}
    \vspace{-0.5cm}
\end{figure}

Finally, we evaluate our model's ability to generalize by testing it in a few-shot scenario with previously unseen tasks. 
We test its generalization performance across the following tasks: (i) super-resolution (x4), (ii) object contour detection, (iii) mask-based inpainting, and (iv) a composite task formed by combining two tasks from our dataset: add and detect. 
For each task, we assess the model's performance when trained with 0, 1, and 100 examples, with the 0-example case being equivalent to a zero-shot setting. 
We compare three baselines: (i) Scratch -- \model initialized with Emu's weights, trained on the examples, (ii) Task Inversion -- \model with task inversion~(Sec.~\ref{task_inversion}), and (iii) Finetune -- \model where we finetune all of the model's weights on the examples. As an upper-bound expert, we train the first baseline on 100,000 examples. Note that, the "Task Inversion" and "Finetune" baselines were trained on the multi-task dataset whereas the “Scratch” baseline was not. 
As can be seen in Fig.~\ref{fig:few-shot}, fine-tuning with a single example is enough to significantly enhance the performance, while training from scratch results in overfitting. Moreover, utilizing 100 samples nearly achieves expert-level performance, implying that the model can effectively generalize to novel tasks.
Additionally, we observe that task inversion is comparable to a full finetune. This suggests that the model already possesses all the necessary information and can simply be "queried" with a new task embedding to produce the desired output. 
For performance and generation examples on additional tasks see~\cref{fig:few-shot_supp,fig:few_shot_examples,fig:few_shot_examples_sup}.

\section{Conclusion}
\model presents a step change in instructable image editing capabilities, primarily due to its unique training on both recognition and generation tasks. This dual-focus approach significantly enhances the model's comprehension of natural language instructions, enabling it to accurately execute a wide array of editing operations. Its ability to generalize to new tasks like image inpainting and super-resolution with minimal examples further demonstrates its versatility and advanced understanding.
Additionally, our framework has the potential for further integration with a multimodal LLM in future work. This enhancement could be especially useful for editing tasks that necessitate more intricate reasoning from the input image, like counting objects or undertaking complex, highly detailed tasks.

\ifreview
\else
\section*{Acknowledgements}   
We extend our gratitude to the following people for their contributions (alphabetical order):
Andrew Brown,
Ankit Ramchandani,
Guan Pang,
Ishan Misra,
Mannat Singh,
Ning Zhang,
Parveen Krishnan,
Peizhao Zhang,
Peter Vajda,
Rohit Girdhar,
Roshan Sumbaly,
Tong Xiao,
Vladan Petrovic,
Xide Xia.
\fi

{
    \small
    \bibliographystyle{ieeenat_fullname}
    \bibliography{main}
}

\clearpage
\setcounter{page}{1}
\maketitlesupplementary

\section{Data}
\label{sec:data_sup}

\begin{figure}[ht!]
    \centering
    \includegraphics[width=1.0\linewidth]{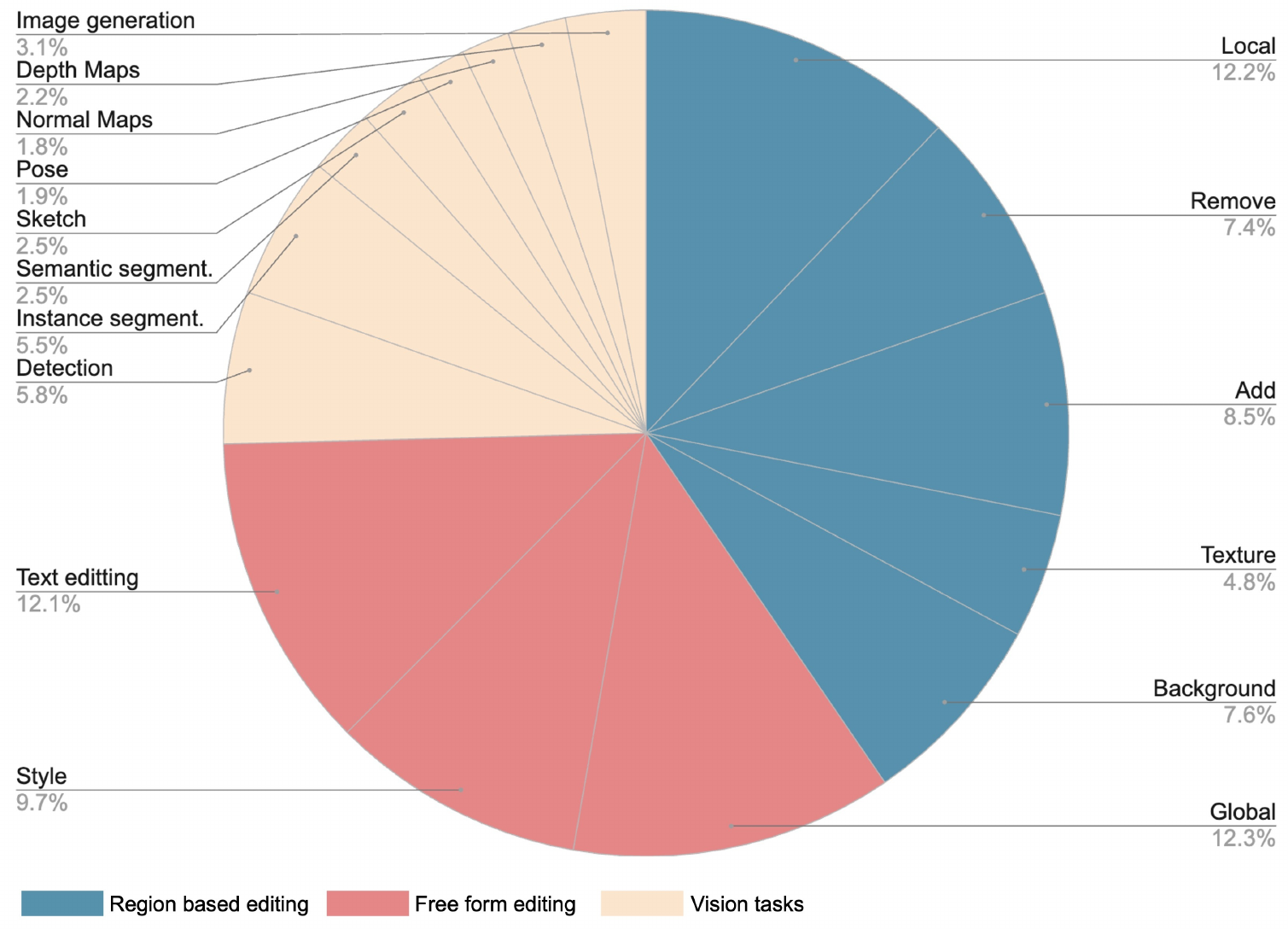}
    \caption{Distribution of the tasks in our training dataset.}
    \label{fig:data_dist}
\end{figure}

Fig.~\ref{fig:data_dist} shows the tasks composing our dataset and their distribution.

\subsection{Instruction Generation}
\label{sec:instruction_sup}
We generate instructions utilizing the dialogue-optimized 70B parameter Llama 2 variant. We use a temperature of 0.9 and set the top-p value to 0.9.
We employ LLM in-context learning to generate instructions. Figs.~\ref{fig:llama_learning}-\ref{fig:llama_prompting} demonstrate the prompts used for task Add. A similar approach is used for the remaining tasks. We instruct the LLM to generate instructions similar to, but diverse from, the examples provided.

To achieve this, we supply the LLM with the following: (1) a system message describing the input and output formats, (2) an introduction message in which we outline the problem and the goal for each key in the output, and, (3) a historical context of the conversation with the LLM containing examples for possible outputs.
We then prompt the LLM with a new input caption and ask it to provide a new instruction.
To encourage more variance and randomness in the LLM-generated instructions, we perform the following on the historical context: (1) shuffling between examples, (2) randomly sampling 60\% of the examples, and, (3) randomly changing the verbs in the examples from a set of words.

\subsection{Image Pairs Generation}
\label{sec:image_generation_sup}
Below we describe in detail our image generation methods for all the tasks. The image pair generation phase uses an image caption, 
and the corresponding output caption, "original object", and "edited object" that the LLM generated in the instruction generation phase.
\subsubsection{Grounded Precise Editing}
\label{sec:mask_based_control_sup}
As described in Sec.~\ref{sec:grounded_precise}, we integrate the mask $m$ of the edited area, during the editing process, to ensure seamless blending of edited regions with the original image. We call this operation \textit{mask-based attention control}. Blending is defined as follows: $x_t \cdot m + (1-m) \cdot y_t$, where $x_t$ is the noisy edited image in step $t$, and, $y_t$ is the noisy version of the input image in step $t$.
In the first $\textit{blend}_{s}$ percent of the steps we replace each of the noisy generated images with the corresponding noisy version of the input images. In the rest of the steps we use blending. The purpose of this, is to ensure structure preservation between the input and the edited image.
We continue by following P2P and inject the self attention layers on all of the tokens. Cross attention layers are injected on the common tokens between the input and output captions. We denote by $\mathcal{N_\textit{c}}$, and $\mathcal{N_\textit{s}}$ the portion of steps where we share cross attention and self attention maps, correspondingly. 
\subsubsection{Mask Extraction}
\label{sec:mask_extraction_sup}
Region-based editing includes all the editing instructions that perform changes to the image in a limited region, leaving the rest of the image unchanged. To adjust a particular object or location while preserving the rest of the details, we utilize a mask of the local area in the editing process. 
We utilize DINO~\cite{Liu2023GroundingDM} to detect the area that needs to be masked, using the "original object" and "edited object" fields that were generated in the previous stage (Sec.~\ref{sec:instruction_sup}).

\textit{Dilation, Gaussian Blurring and Bounding Box Masks:} 
We observe that when utilizing mask-based attention control to generate an edited image, it often replaces the object with a similar object type instead of removing it. For example, when masking the region around a dog, we confine the editing to that specific area, resulting in the generation of a new variation of the dog.
We address this issue by creating three different types of masks. The first employs the original precise mask, created by DINO and SAM~\cite{Kirillov2023SegmentA}. The second involves expanding the mask beyond the added object through dilation and then refining it using Gaussian blurring. Finally, the third approach uses the bounding box around the object (created by DINO), thereby eliminating the constraints of a specific shape.
We generate multiple images, each with a different mask, and then filter for the best image. Our filtering is described in Sec.~\ref{sec:images}.

\textit{Possessive words:} In some cases the "original object" and "edited object" generated by the LLM contain possessive words (e.g, "a dog's tail"). We observe that, in many cases, DINO struggles to detect the object in these cases. 
To this end, we employ an additional prompting to the LLM to identify the object without possession.
\subsubsection{Region-Based Editing Tasks}
\label{sec:region_dataset_sup}

\paragraph{Local/Texture} 
Given the input caption, we first generate the input image. Then, we utilize the "original object" (as described in Sec.~\ref{sec:image_generation_sup}) to extract the local mask (using Sec.~\ref{sec:mask_extraction_sup}). Lastly, we apply masked-based attention control using the obtained mask to generate the edited image. We repeat this entire process for 10 iterations, where in each iteration, we sample the guidance scale from $[4,8]$, $\mathcal{N_\textit{c}}$ and $\mathcal{N_\textit{s}}$ from $[0.3,0.9]$, and $\textit{blend}_{s}$ from $[0.02, 0.2]$.

\paragraph{Add} Extracting the mask of the "edited object" (the object that was added in this case) is not possible in advance because the object does not exist in the input image. To overcome this challenge, we address this as follows:
\begin{enumerate}
    \item We generate the output image $y$ using the output caption. Note that the image $y$ contains the "edited object".
    \item The mask $m$ of the "edited object" in $y$ is extracted.
    \item We apply the mask-based attention control to generate the input image $x$ using the input caption, the image $y$ and the mask $m$
\end{enumerate}

The main problem with this approach is that in certain instances, we generate a different version of the object, instead of eliminating it, as described in Sec~\ref{sec:mask_extraction_sup}. 

\paragraph{Remove.} The process of generating data for Remove task is similar to the one of Add task. The only difference is that we first generate the image $x$ (using the input caption), then extract the mask $m$ of the object to remove, and finally generate the image $y$ using the output caption, image $x$ and the mask $m$.  

\paragraph{Background.} Given an input image, input caption and the edited object (in this case, the alternative background), we first extract the background mask. To eliminate artifacts in the contour, we apply minimum filter which extends the background mask and then smooth it using Gaussian filtering. Next, we provide the image and the resulting mask as input to an inpainting model, which creates a new background. Lastly, we blend the input image and the  edited image in the mask region. We generate 10 edited images, with different noise and guidance scale, and pick the best according to CLIP metrics.

 \subsubsection{Free-Form Editing Tasks}
 \label{sec:free_form_tasks_sup}
 \paragraph{Global.} The global task includes editing instructions that are not restricted to a specific area. Therefore, we generate the image pairs using mask-based attention control with a blank mask. 
 $\textit{blend}_{s}$ is sampled from $[0.1, 0.2]$ to encourage better image faithfulness.
 We sample $\mathcal{N_\textit{c}}$ and $\mathcal{N_\textit{s}}$ from $[0.4,0.9]$. 
\paragraph{Style.} We use Plug-and-Play (PNP)~\cite{pnp} to generate the stylized edited images. The goal of this task is to alter the image style according to the editing instruction while preserving the image structure. 
We apply PNP on the real input images using DDIM inversion. For each sample, we generate 10 edited images, each with the following parameters sampled: guidance scale sampled from $[6.5,10.0]$, $\mathcal{N_\textit{s}}$ from $[0.5,1.0]$, and, the portion of spatial features to share is set to 0.8. 
\paragraph{Text Editing.} 
The text editing task includes adding text to the image, removing text from the image, and replacing one text with the other. In addition, we allow the user to choose the font and the color of the added text. %
We generate a mask, $m$, of the text found in the input image, $x$, using OCR~\cite{du2020pp}. We utilize mask $m$ to inpaint the image, denote the new image $y$. For adding text, we use $y$ as the input image and $x$ as the edited image. For removing text and replacing text, we use the reverse. When replacing text, we overlay the inpainted region in image $y$ with a text in a specific font and color.

\subsubsection{Vision tasks}
\label{sec:vision_tasks_sup}

\paragraph{Detect/Segment.} Given an input image, we detect the "edited object" using DINO. To formalize detection as a generative task, we create a new image $y$ by drawing the detected bounding box. For segmentation, we paint the detected object pixels. %
\paragraph{Color.} We define the Color task as a modification to the overall colors of an image. We generate samples by applying the following filters: (1) color filters - randomly changing the brightness, contrast, saturation and hue of an image, (2) blurring - applying random-sized Gaussian kernels, and (3) sharpening and defocusing.
\paragraph{Image-to-Image Translation} Tasks that involve bi-directional mapping from conditioning images to target images. For instance, sketch-to-image and image-to-sketch. We follow~\cite{zhang2023adding}, to generate depth maps, segmentation maps, human poses, normal maps and sketches.

\setlength{\tabcolsep}{2pt}
\begin{table}[h!]
  \centering
  \caption{\textbf{Data generation pipeline evaluation.} We compare our data generation pipeline with that of InstructPix2Pix. We also report the automatic metrics on the InstructPix2Pix training dataset. }
  \label{table:data_p2p_comp}
  \scalebox{0.85}{
  \begin{tabular}{c|l|ccccc}
  \toprule\noalign{\smallskip}
  Task & Method & $\text{CLIP}_{dir}$ &  $\text{CLIP}_{im}$ & $\text{CLIP}_{out}$ &  $\text{L1}\!\downarrow$  & DINO$\uparrow$  \\
  \noalign{\smallskip}
  \hline
  \noalign{\smallskip}
   \multirow{2}{*}{Local} & IP2P & 0.329 & 0.922  & 0.270 & 0.046 & 0.917 \\
  & Our & 0.402 & 0.927 & 0.289 & 0.029 &  0.908 \\
  \hline
  \noalign{\smallskip}
   \multirow{ 2}{*}{Texture} & IP2P & 0.282 & 0.876  & 0.297 & 0.189 & 0.671 \\
   & Our & 0.373 & 0.957  & 0.296 & 0.033 & 0.923 \\
     \hline
  \noalign{\smallskip}
   \multirow{2}{*}{Remove} & IP2P  &  0.204 & 0.818 & 0.254 & 0.067 & 0.755 \\
  & Our & 0.279 & 0.913 & 0.266 & 0.046 &  0.841 \\
  \hline
  \noalign{\smallskip}
  \multirow{2}{*}{Add} & IP2P &  0.263 & 0.897  & 0.278 & 0.157 & 0.934 \\
  & Our & 0.318 & 0.962 & 0.304 & 0.007 & 0.925 \\
  \hline
  \noalign{\smallskip}
   \multirow{2}{*}{Global} & IP2P & 0.281 & 0.916 & 0.276 & 0.103 & 0.845\\
  & Our & 0.315 & 0.919 & 0.289 & 0.081 &  0.869 \\
  \hline
  \noalign{\smallskip}
   \multirow{2}{*}{Background} & IP2P  &  0.106 & 0.829  & 0.271 & 0.082 & 0.725 \\
  & Our & 0.214 & 0.843 & 0.283 & 0.201 &  0.771 \\
  \midrule
  \multicolumn{2}{c|}{IP2P Dataset}  & 0.172 & 0.855 & 0.271 &0.119 & 0.809 \\
  \bottomrule
  \end{tabular}}
  \vspace{-4mm}
\end{table}

\section{Dataset Evaluation}
\label{sec:data_exp}
In Sec.~\ref{sec:data} we introduce our dataset generation pipeline, which includes  methods that address the unique difficulties associated with each particular task. 
In this section we compare our approach with that of InstructPix2Pix.
We begin by sampling 6,000 random samples from the same distribution of Sec.~\ref{sec:data}, following the instruction generation stage.
Hence, each sample contains the input image, input caption, editing instruction, output caption, and the edited objects. 
We then generate image pairs using both our data generation pipeline, and that of InstructPix2Pix, which employs Prompt-to-Prompt and CLIP-based filtering.
In Tab.~\ref{table:data_p2p_comp} we report automatic metrics comparing the outputs of each pipeline.
As can be seen, our method for data generations outperforms that of InstructPix2Pix (IP2P) on all the tasks.
Additionally, to isolate the effect of our instruction generation stage, we also directly evaluate the InstructPix2Pix training dataset, which also underperforms when compared to ours.

\setlength{\tabcolsep}{5pt}
\begin{table}[ht]
\centering
\caption{Number of images per task and split in our Image Editing Benchmark} 
\label{tab:benchmark_dist}
\centering
\begin{tabular}{lcc}
\toprule
Task & Validation set & Test set \\
\midrule 
Add & 264 & 533 \\
Background & 266 & 373 \\
Color & 262 & 519 \\
Global & 220 & 219 \\
Remove & 264 & 531 \\
Local & 256 & 446 \\
Style & 227 & 434 \\
\bottomrule
\end{tabular}
\end{table}

\setlength{\tabcolsep}{4pt}
\begin{table*}[h]
  \centering
  \caption{Comparison of \model to task-specific experts on image-editing tasks. We report automatic metrics and  human preference ratings. Human evaluation (\%) is shown as a percentage of majority votes in favor of our multi-task model compared to an expert model.}
  \label{table:experts}
  \scalebox{0.85}{
  \begin{tabular}{c|l|ccccc|cc}
  \toprule\noalign{\smallskip}
  Task & Method & $\text{CLIP}_{dir}\!\uparrow$ &  $\text{CLIP}_{img}\!\uparrow$ & $\text{CLIP}_{out}\!\uparrow$ &  $\text{L1}\!\downarrow$  & DINO$\uparrow$ & Text & Image \\
  & & & & & & & Align. & Faith. \\
  \noalign{\smallskip}
  \hline
   \multirow{2}{*}{Local} & Expert & 0.139 & 0.879  & 0.244 & 0.057 &  0.841 & - & - \\
  & Our & 0.142 & 0.885  & 0.252 & 0.047 &  0.891 & 57.5 & 56.9 \\
  \hline
  \noalign{\smallskip}
   \multirow{2}{*}{Global} & Expert & 0.106 &  0.820 &  0.227 & 0.096 & 0.823 & - & - \\
  & Our & 0.118 & 0.852 & 0.235 & 0.072 &0.847  & 58.4 & 62.6 \\
     \hline
  \noalign{\smallskip}
   \multirow{2}{*}{Add} & Expert & 0.119 &  0.851 & 0.237 & 0.059  &  0.828 & - & - \\
  & Our & 0.123 & 0.917 & 0.240 & 0.036 & 0.892 & 61.1 & 59.6 \\
  \hline
  \noalign{\smallskip}
   \multirow{2}{*}{Background} & Expert & 0.145 & 0.689 & 0.229 & 0.240  & 0.560 & - & - \\
  & Our & 0.157 & 0.852 & 0.240 & 0.223 & 0.586 & 64.3 & 62.5 \\
  \hline
  \noalign{\smallskip}
  \end{tabular}}
  \vspace{-2mm}
\end{table*}

\section{Image Editing Benchmark}
\label{sec:instruction_based_image_editing_benchmark}
We take the images from the MagicBrush benchmark ~\cite{zhang2023magicbrush} and undergo a three-step annotation process utilizing crowd workers: (i) instruction generation, (ii) instruction filtering, and, (iii) caption annotation. In the first step, three crowd workers are assigned to generate an instruction for each (image, task) pair. Moving to the second stage, five different crowd workers classify each (image, instruction) pair's task type and whether the instruction is relevant to the image. Instructions with at least one irrelevant annotation are then filtered out, and for the remaining ones, the task is determined through majority voting among the five workers. At this juncture, we select, at most, a single instruction for each (image, task) pair to preserve the benchmark's diversity. 

Finally, we task crowd workers with annotating two captions for each remaining (image, instruction) pair - one for the image, and one for the desired image after having edited it. This facilitates automatic evaluation using the methodologies outlined in \cite{prompt2prompt,pnp}. Throughout this annotation phase, workers are presented with the input image and instruction, and are tasked with providing captions that faithfully describe the image while aligning with the given instruction. See Tab.~\ref{tab:benchmark_dist} for the number of images per task and split in our benchmark.

\begin{figure}[t]
    \centering
    \includegraphics[width=1.0\linewidth]{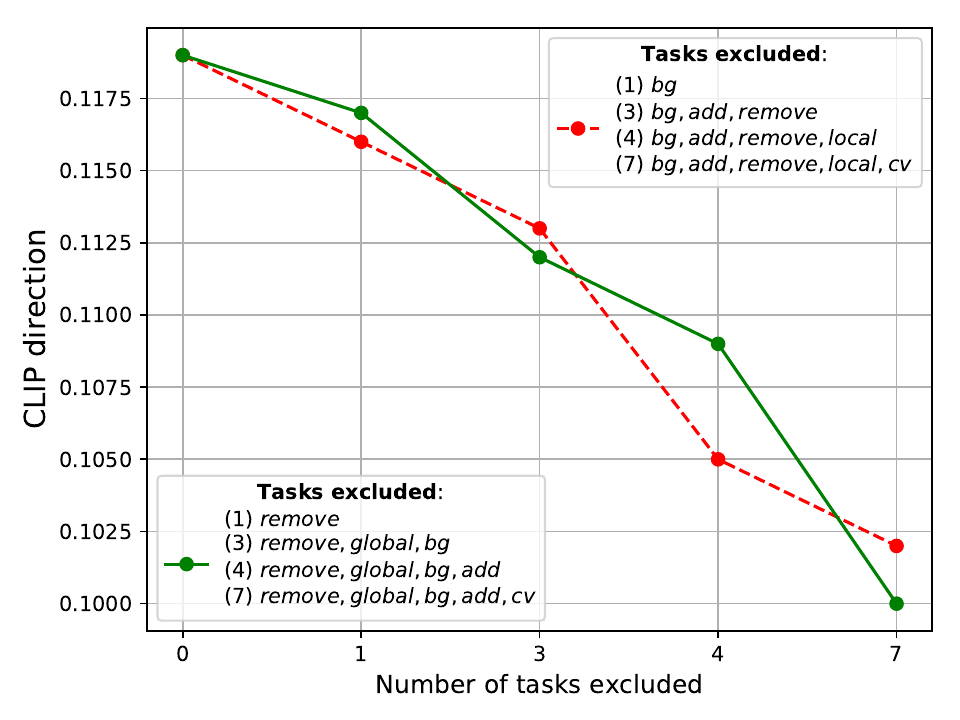}
    \caption{
    Ablation on the model performance ($\text{CLIP}_{dir}$) on Style and Texture tasks as we progressively exclude tasks that don't fall within these categories. %
    }
    \label{fig:agg_fig}
    \vspace{-2mm}
\end{figure}
\section{Additional Results}

\subsection{Performance on Vision Tasks} 
\label{sec:vision_eval_sup}
We also evaluate the performance of our model on tasks other than edit, specifically: detection, segmentation, and depth estimation. We report: (i) Mean Average Precision~(mAP@0.5) on MS-COCO~\cite{coco} for detection task, (ii) Mean Intersection over Union~(mIoU) on ADE20K~\cite{zhou2017scene, zhou2019semantic} for segmentation task, and, (iii) Root Mean Square Error~(RMSE) on NYUv2~\cite{nyuv2} for monocular depth estimation. \model was not trained on those datasets, therefore, we report zero-shot results on both tasks, see Tab.~\ref{tab:eval_cv}.

\setlength{\tabcolsep}{1pt}
\begin{table}[ht!]
\centering
\caption{\model performance on vision tasks. For object detection we use mAP@0.5, for segmentation we use mIoU, and, for depth estimation we use RMSE.} 
\label{tab:eval_cv}
\scalebox{1.0}{
\centering
\begin{tabular}{lccc}
\toprule
Method & Object & Semantic & Depth  \\ 
& Detection~$\!\uparrow$ & Segmentation~$\!\uparrow$ & Estimation~$\!\downarrow$ \\ 
\midrule
\model                                             & 61.467 & 50.028 & 0.246   \\
\bottomrule
\end{tabular}}
\end{table}

 \subsection{Controlling the Task Embedding}
\begin{figure}[ht]
\centering
\scalebox{0.90}{   
\begin{tabular}{@{\hspace{-5\tabcolsep}}c@{\hspace{-0.3\tabcolsep}}c@{\hspace{-0.3\tabcolsep}}c@{\hspace{-0.3\tabcolsep}}c@{\hspace{-0.3\tabcolsep}}c}

& \begin{tabular}[x]{@{}c@{}} \footnotesize{Input} \end{tabular} & \begin{tabular}[x]{@{}c@{}} \footnotesize{Predicted} \end{tabular} & \begin{tabular}[x]{@{}c@{}} \footnotesize{Global} \end{tabular} & \footnotesize{Text}   \\ 
 \resizebox{!}{11px}{
\begin{tabular}[x]{@{}c@{}}\textit{Add} \\ \textit{pink} \end{tabular}}&
\raisebox{-.5\height}{
\includegraphics[width=0.25\linewidth]{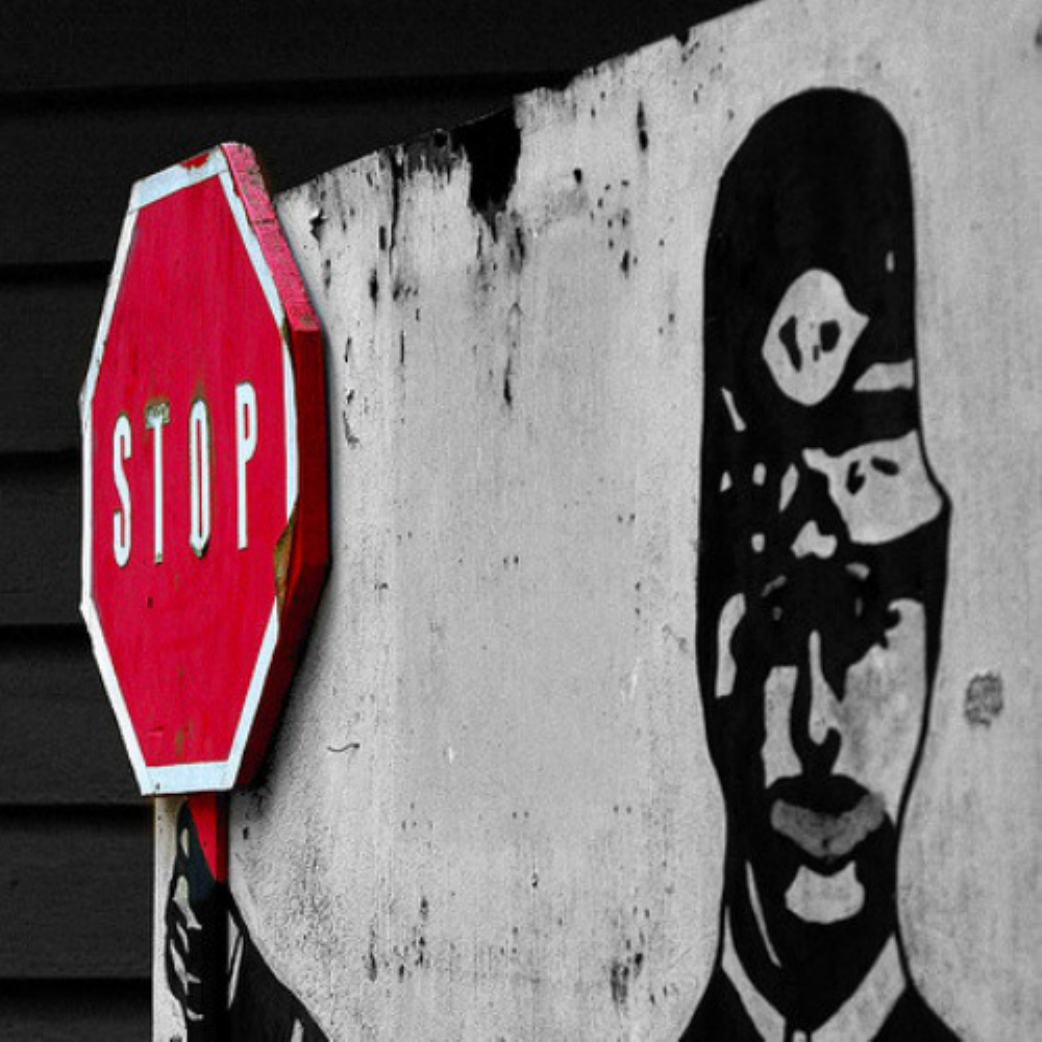}}&
\raisebox{-.5\height}{
\includegraphics[width=0.25\linewidth]{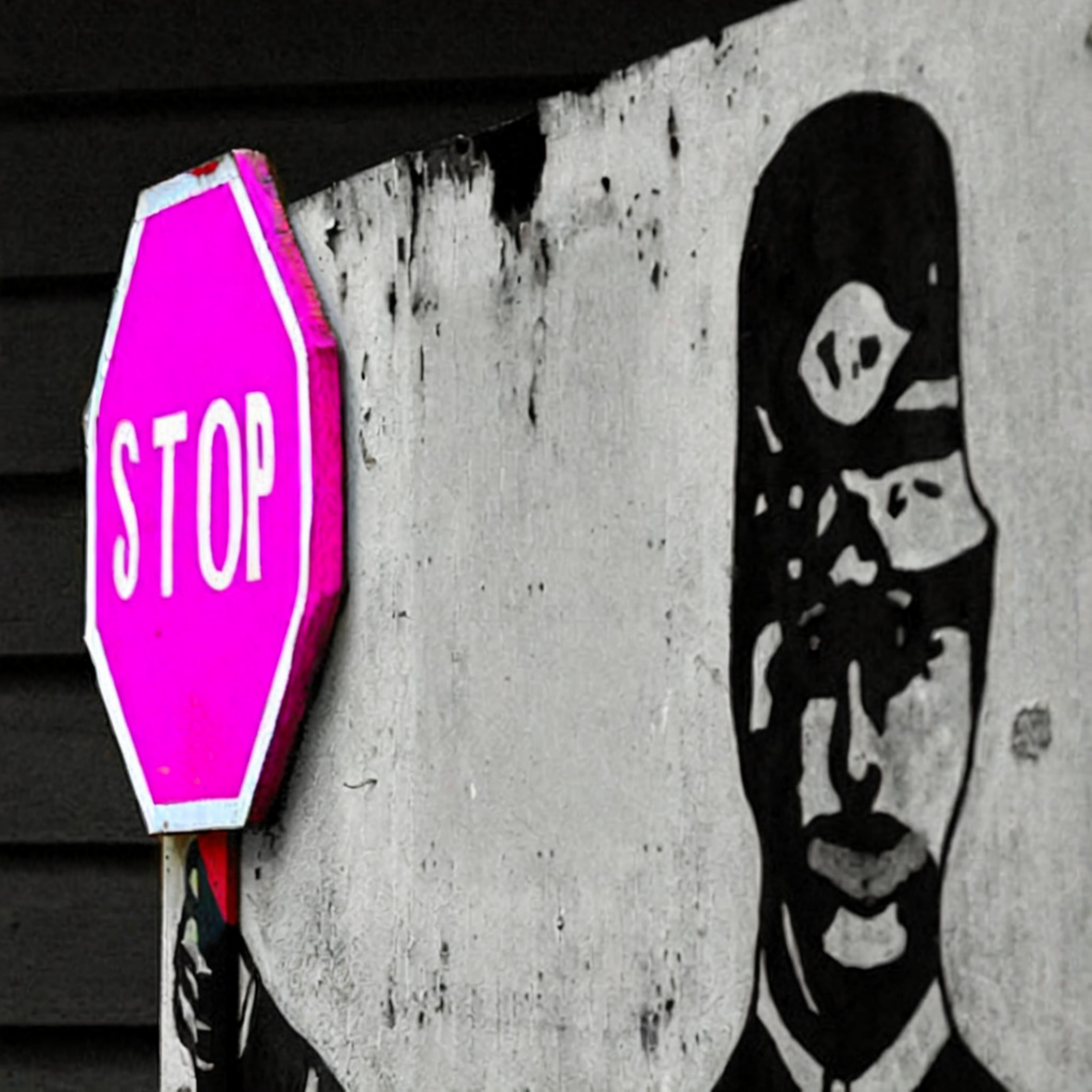}}&
\raisebox{-.5\height}{
\includegraphics[width=0.25\linewidth]{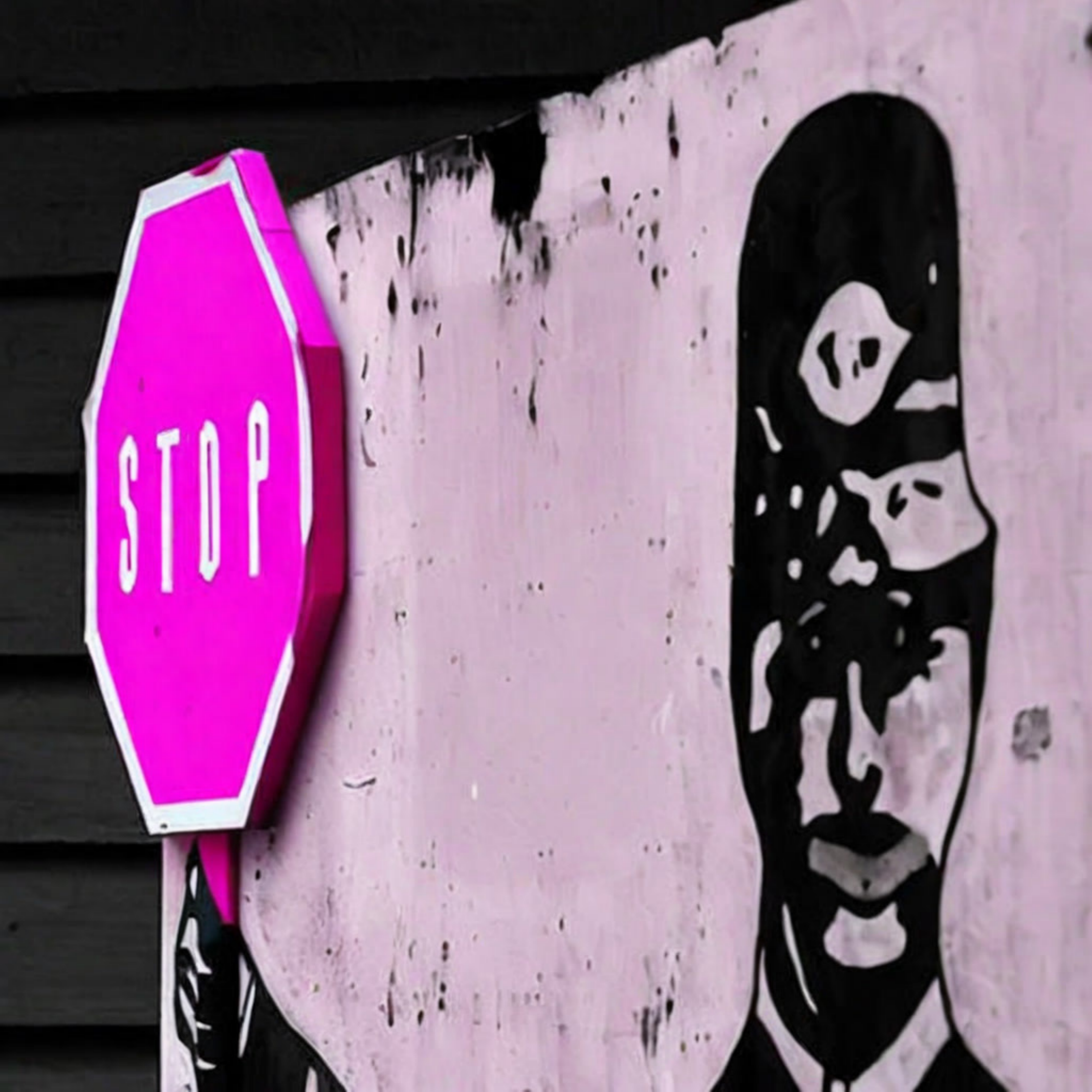}}&
\raisebox{-.5\height}{
\includegraphics[width=0.25\linewidth]{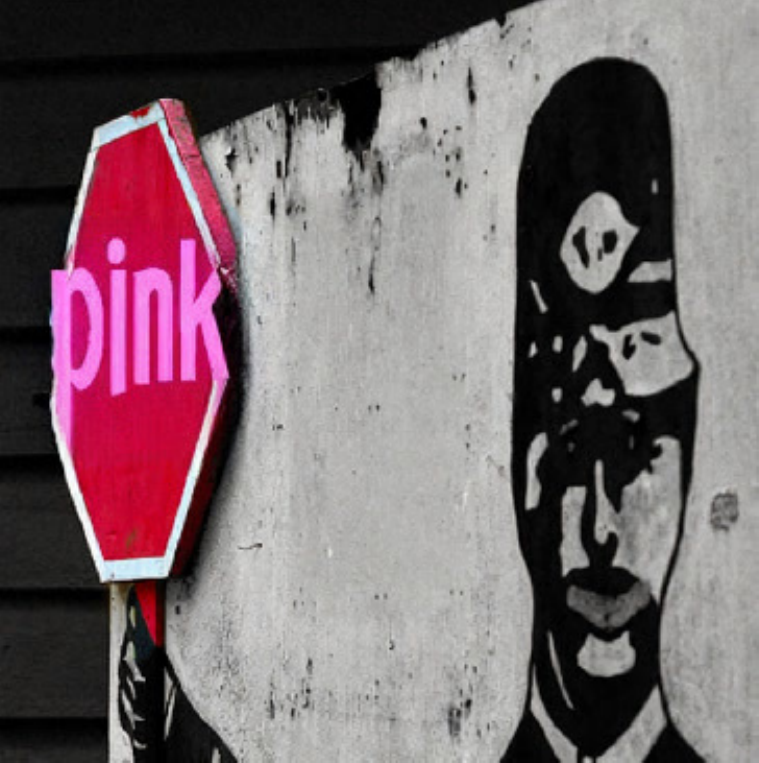}} \\ [8mm]

& \begin{tabular}[x]{@{}c@{}} \footnotesize{Input} \end{tabular} & \begin{tabular}[x]{@{}c@{}}\footnotesize{Predicted} \end{tabular} & \begin{tabular}[x]{@{}c@{}} \footnotesize{Local} \end{tabular} & \footnotesize{Background}   \\
 \resizebox{!}{14px}{
\begin{tabular}[x]{@{}c@{}}\textit{Make} \\ \textit{it} \\ \textit{cartoon} \end{tabular}}&
\raisebox{-.5\height}{
\includegraphics[width=0.25\linewidth]{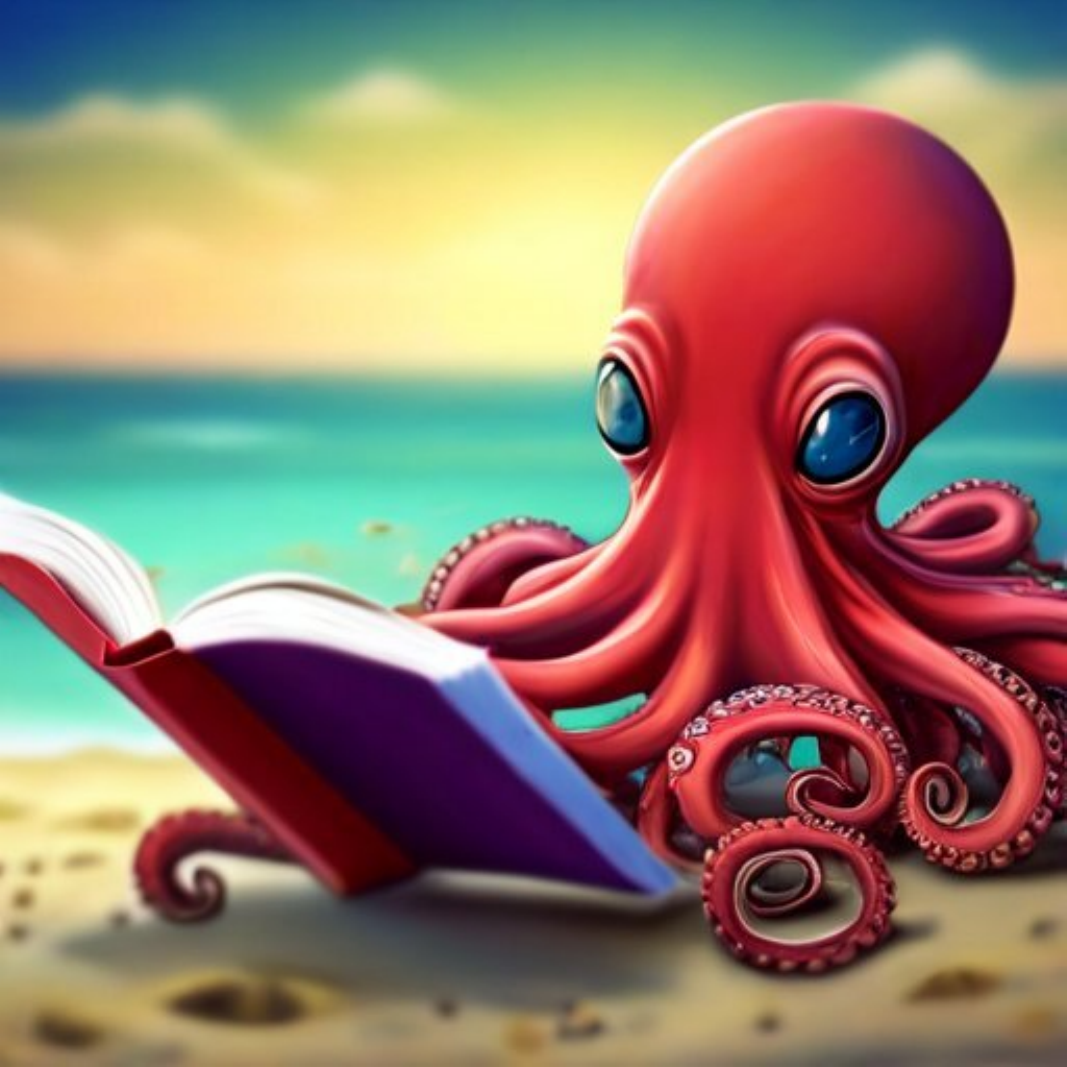}}&
\raisebox{-.5\height}{
\includegraphics[width=0.25\linewidth]{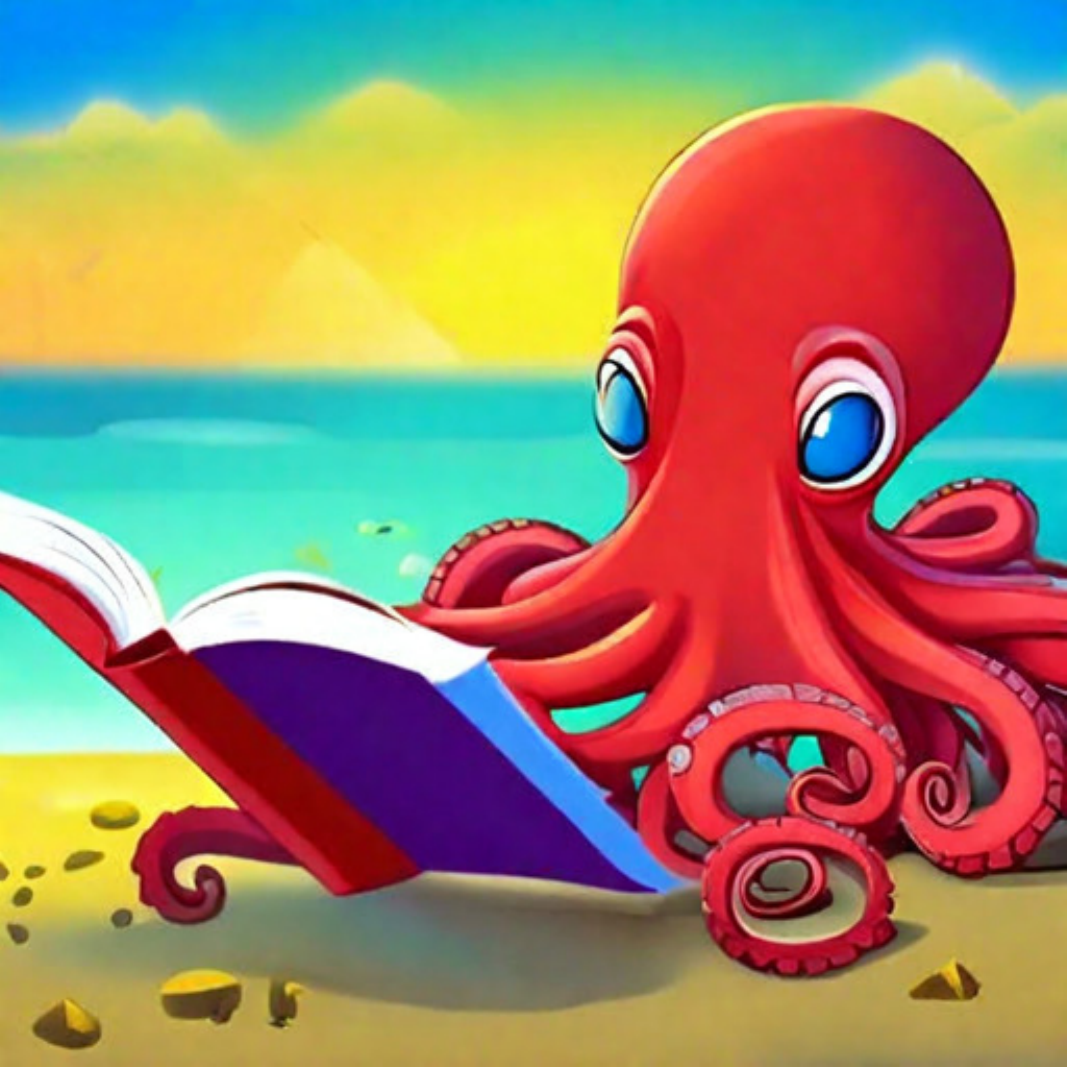}}&
\raisebox{-.5\height}{
\includegraphics[width=0.25\linewidth]{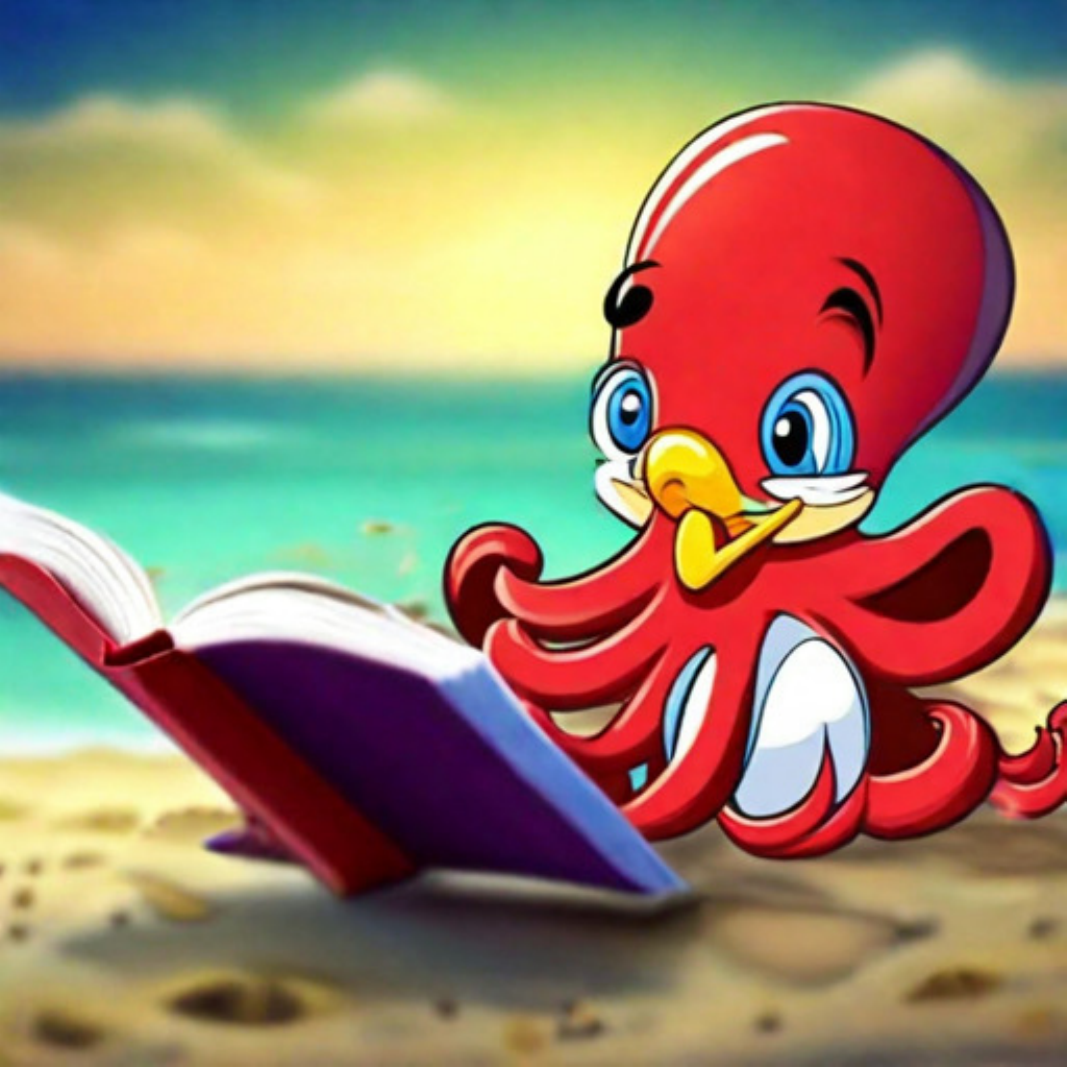}}&
\raisebox{-.5\height}{
\includegraphics[width=0.25\linewidth]{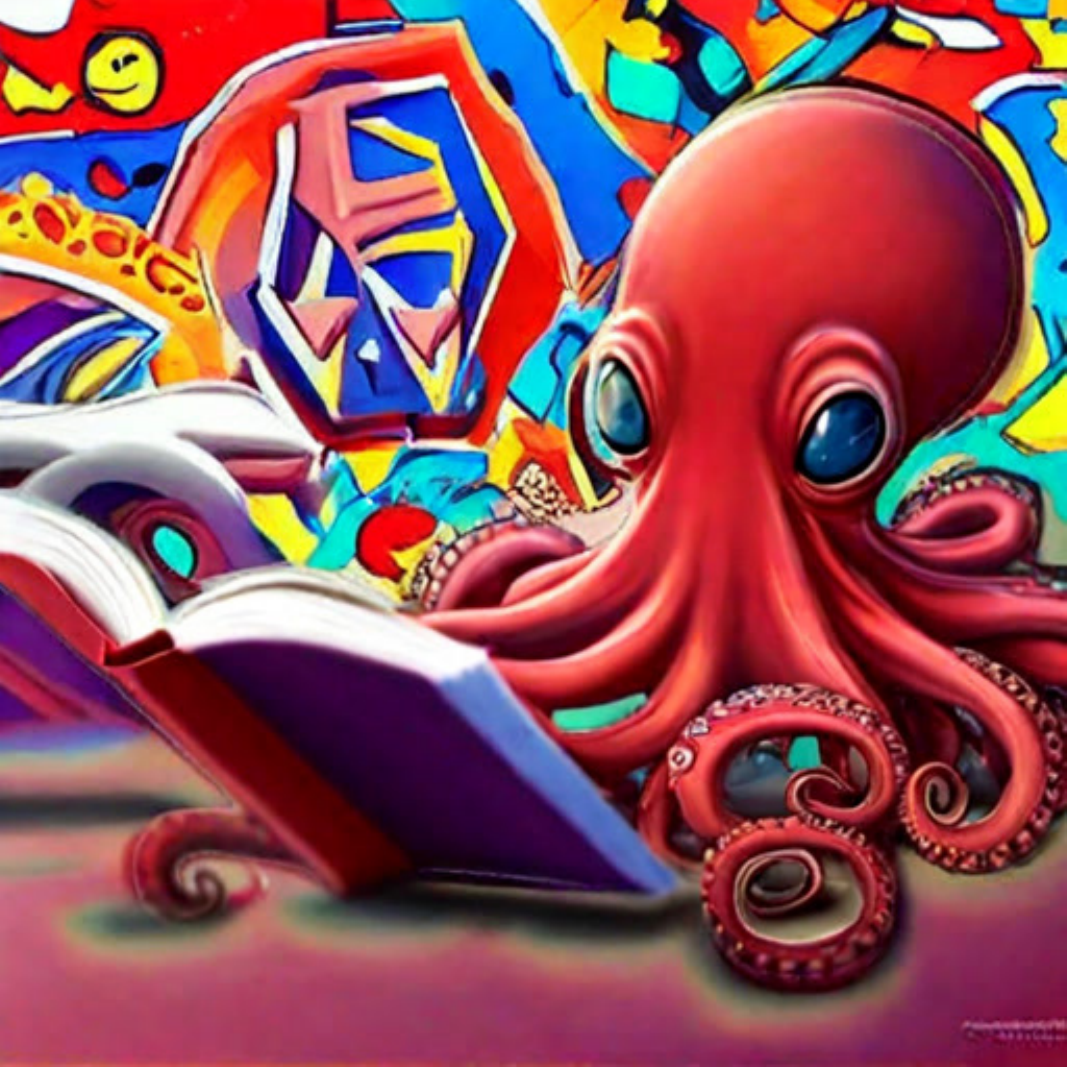}} \\ [8mm]

& \begin{tabular}[x]{@{}c@{}}\footnotesize{Input} \end{tabular} & \begin{tabular}[x]{@{}c@{}}\footnotesize{Predicted} \end{tabular} & \begin{tabular}[x]{@{}c@{}} \footnotesize{Style} \end{tabular} & \footnotesize{Segment}   \\
 \resizebox{!}{18px}{
\begin{tabular}[x]{@{}c@{}}\textit{Change} \\ \textit{to} \\ \textit{living} \\ \textit{room} \end{tabular}}&
\raisebox{-.5\height}{
\includegraphics[width=0.25\linewidth]{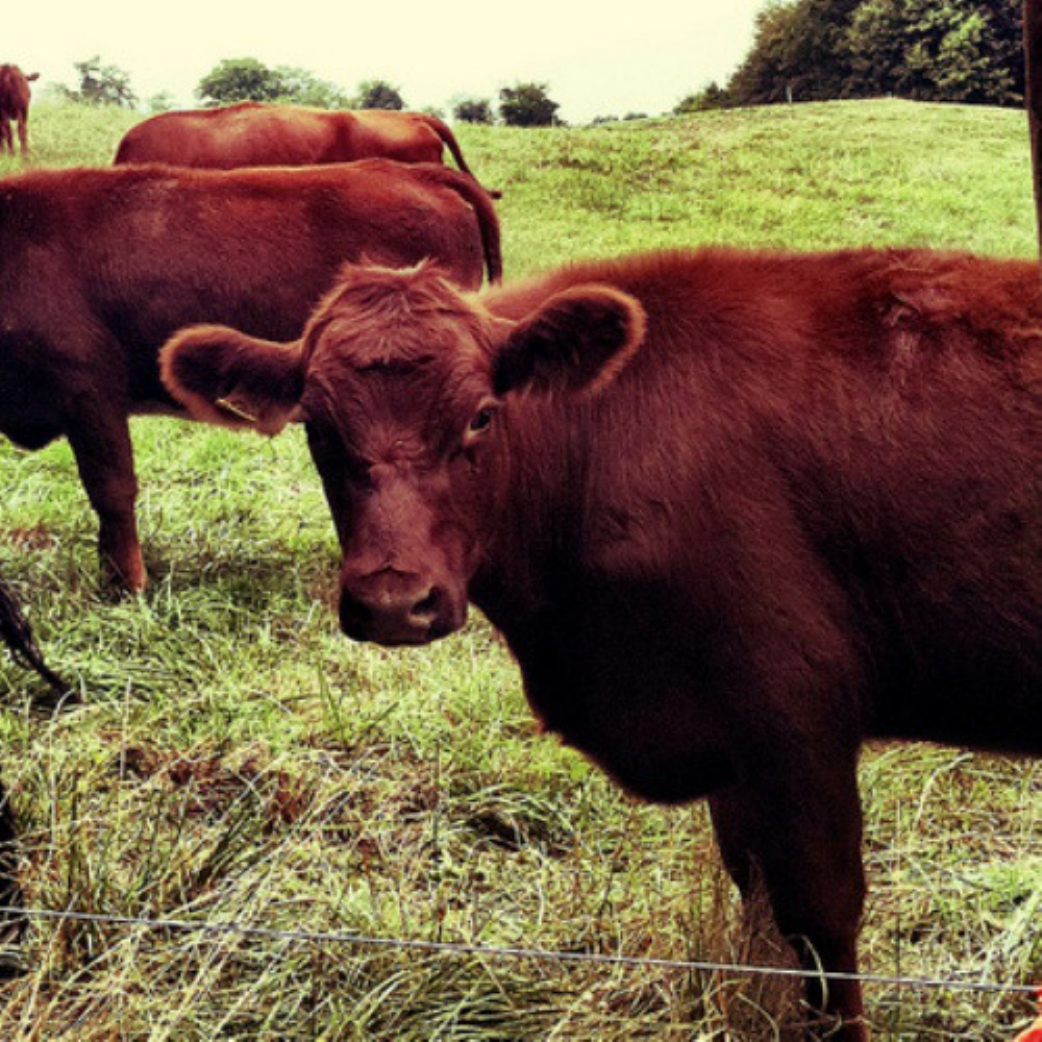}}&
\raisebox{-.5\height}{
\includegraphics[width=0.25\linewidth]{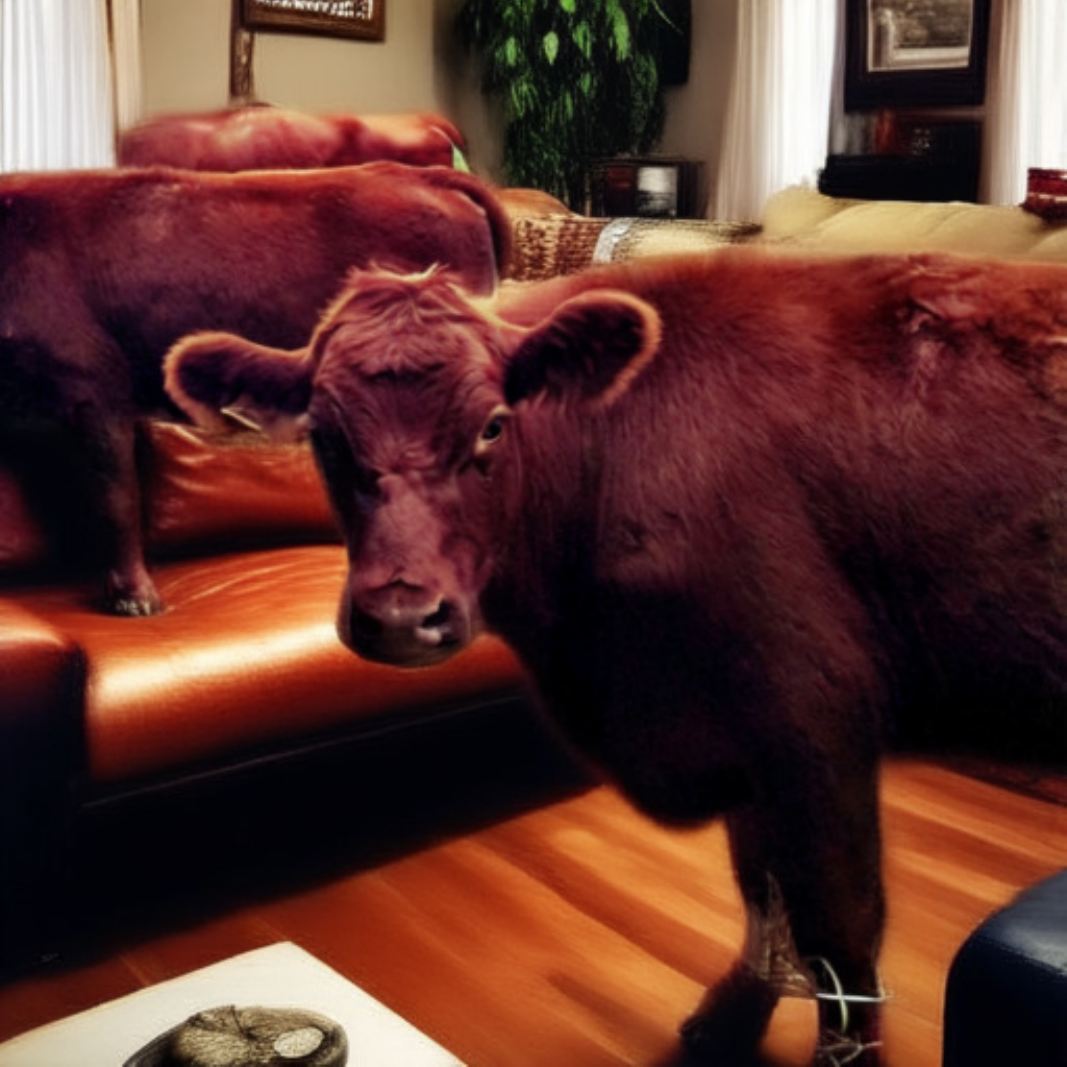}}&
\raisebox{-.5\height}{
\includegraphics[width=0.25\linewidth]{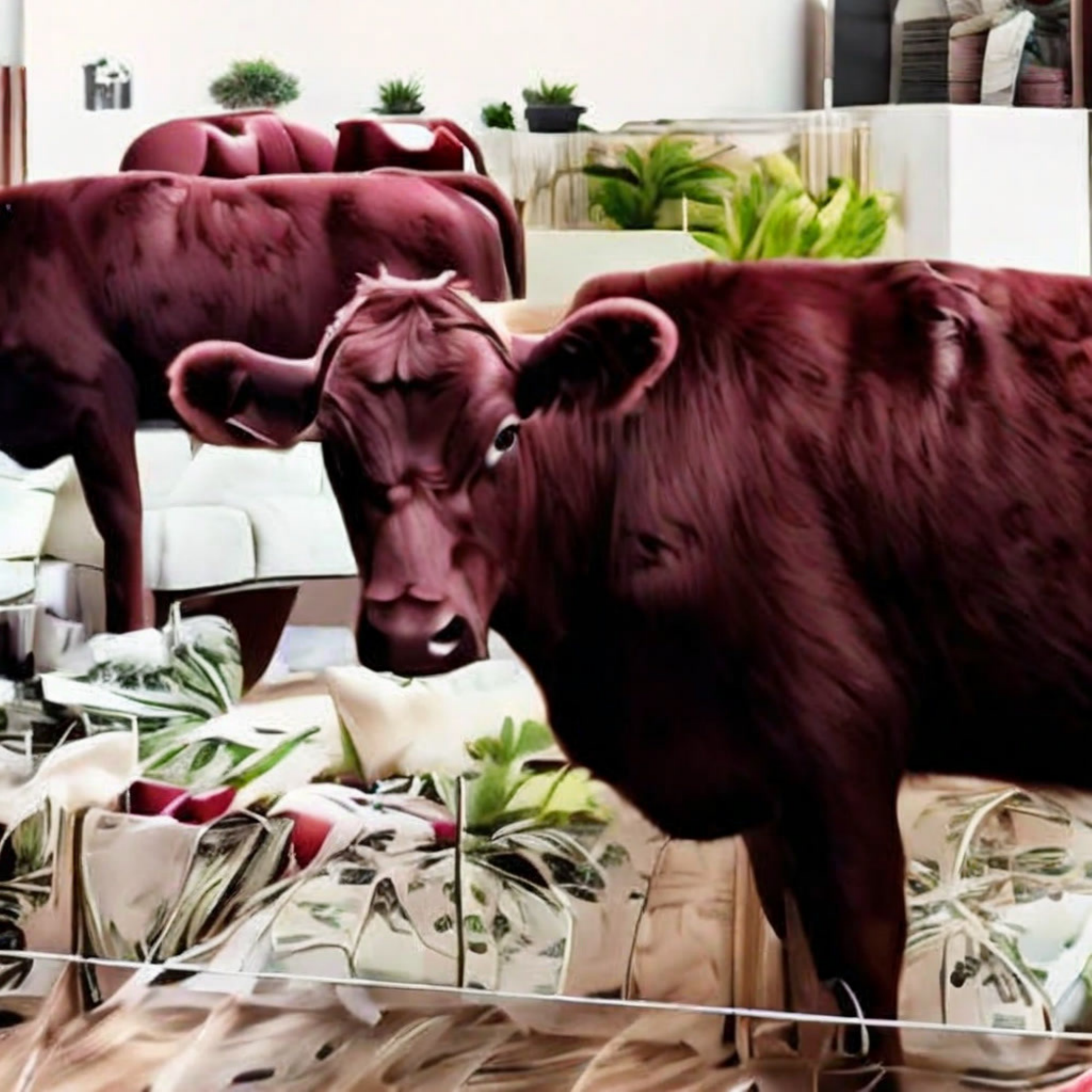}}&
\raisebox{-.5\height}{
\includegraphics[width=0.25\linewidth]{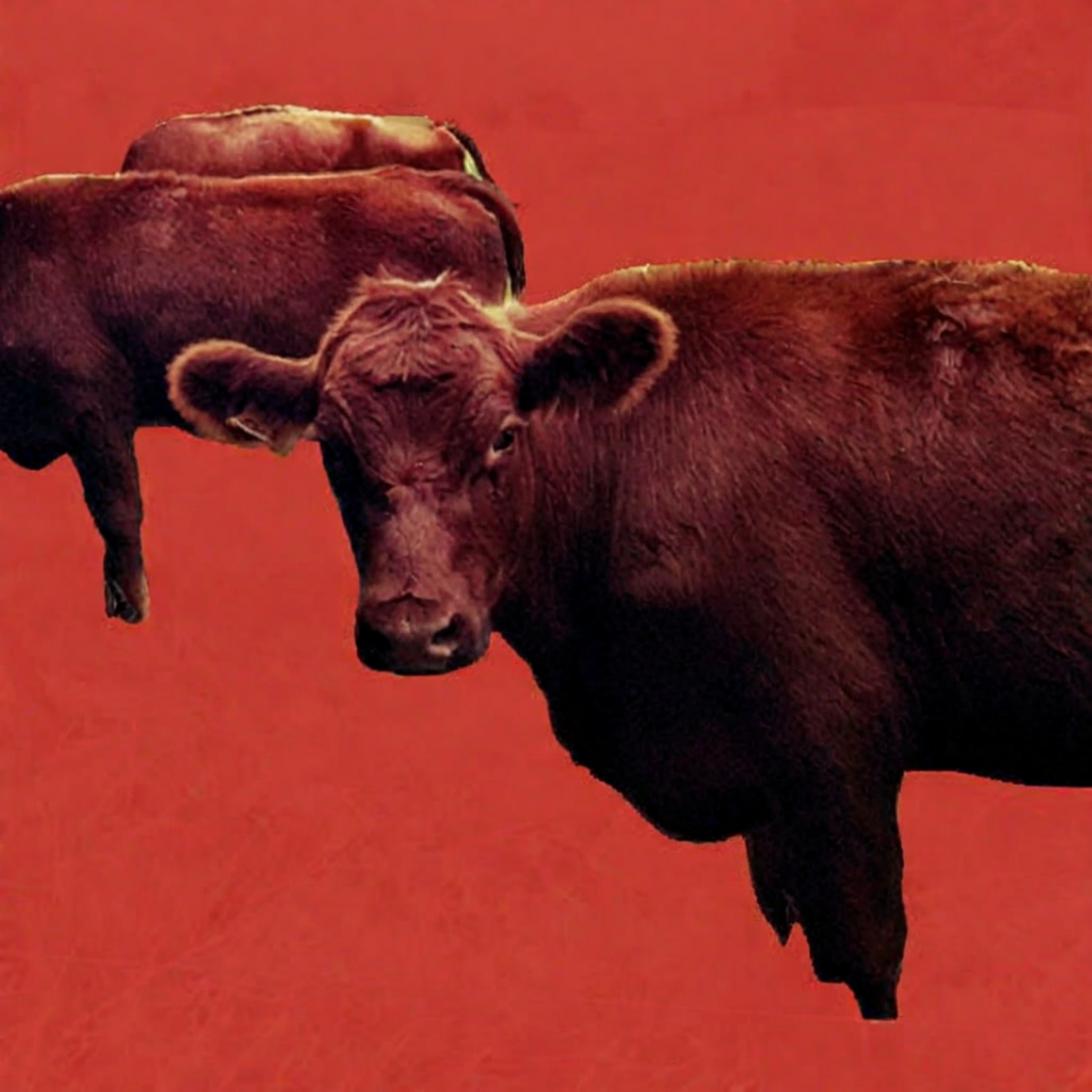}} \\

\end{tabular}
}
\caption{Controlling the Task Embedding. For each sample, we present the edited image using the task predicted by the task predictor. In addition, we present the edited image generated using the same input image and instruction, but with different task embeddings. For instance, in the first row we generate the edited image using the predicted task (Add), Global task, and Text task.}
\label{fig:task_cond_control} 
\end{figure}

As depicted in Fig.~\ref{fig:task_cond_control}, altering the task embedding controls the task executed by the model, resulting in different generations for a given instruction. 

\subsection{Influence of Number of Tasks}
We report results for the ablation of the number of tasks in Fig.~\ref{fig:agg_fig}.

\subsection{Few-Shot Learning of New Tasks}
\begin{figure}[h!]
\centering
    \subfloat[Super Resolution]
        {\includegraphics[width=0.24\textwidth]
    	{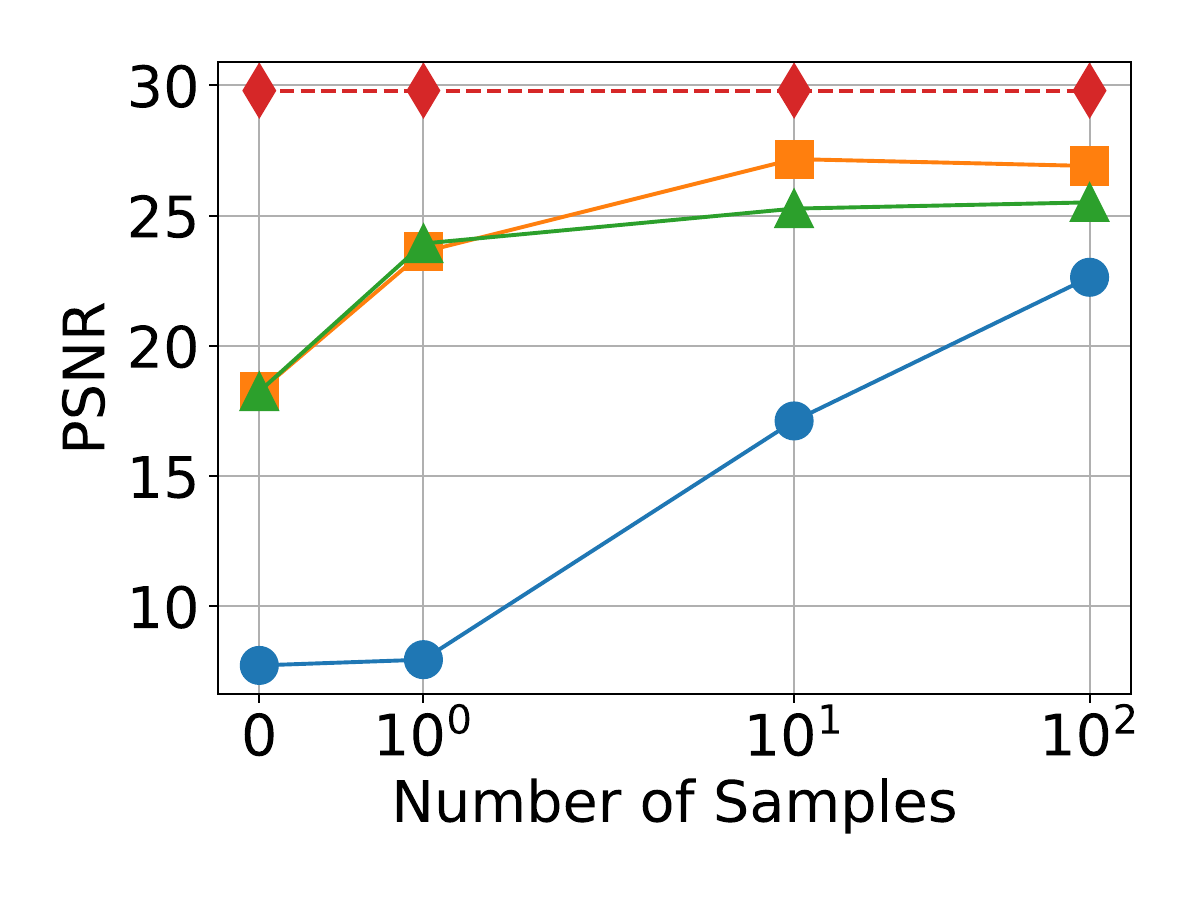}}
    \subfloat[Contour Detection]
        {\includegraphics[width=0.24\textwidth]
    	{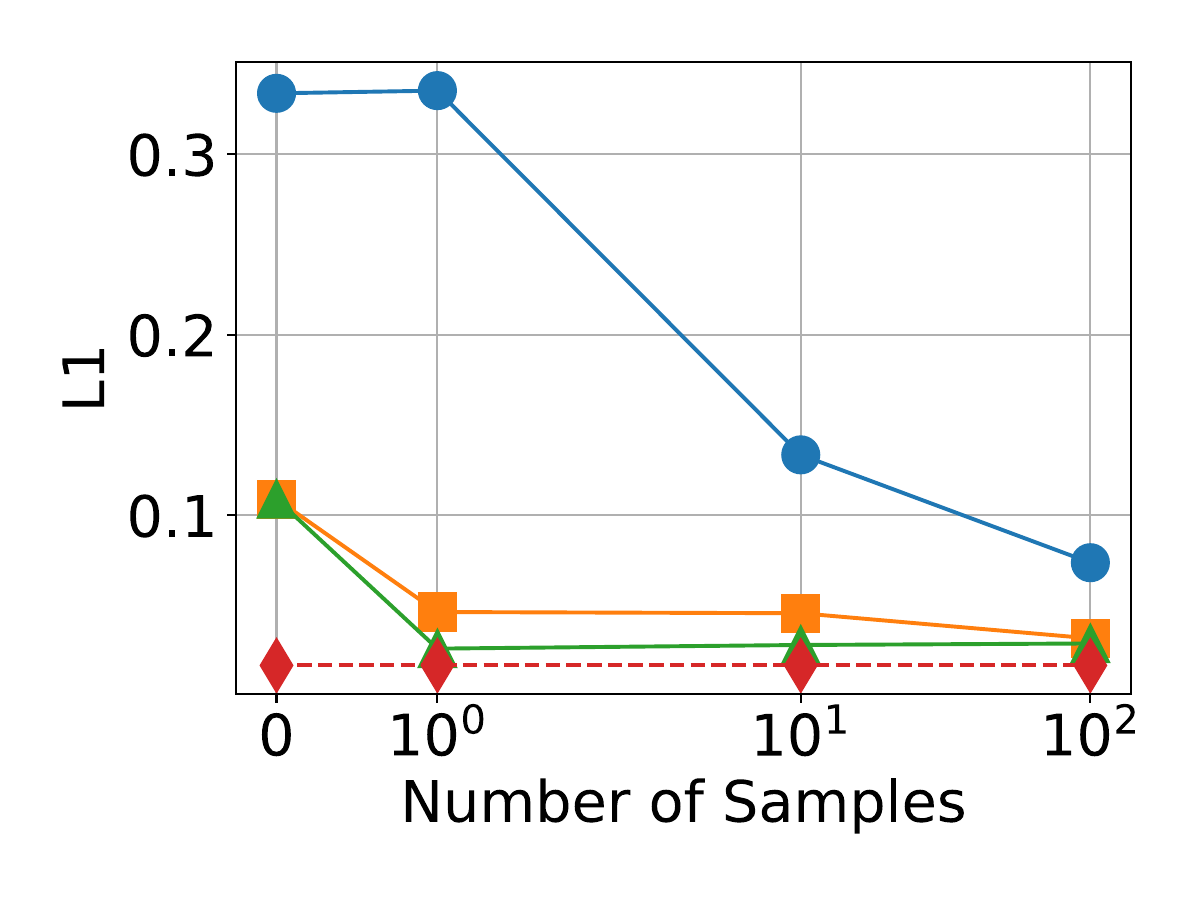}}
    \caption{\label{fig:few-shot_supp}Few-shot performance for different tasks over 1, 10, and 100 samples. Each line represents a different training setting: Emu finetune (Blue, $\bigcirc$), \model finetune (Orange, $\Box$), task inversion (Green, $\triangle$), all compared to an upper-bound expert trained on 100k samples (Red dashed line, $\diamondsuit$).}
\end{figure}

\begin{figure*}[h]
   \centering
\begin{tabular}{@{\hspace{-10\tabcolsep}}c@{\hspace{0.001\tabcolsep}}c@{\hspace{-0.3\tabcolsep}}c@{\hspace{-0.3\tabcolsep}}c@{\hspace{-1\tabcolsep}}c@{\hspace{0.001\tabcolsep}}c@{\hspace{-0.3\tabcolsep}}c@{\hspace{-0.3\tabcolsep}}c}

\resizebox{!}{21px}{
\begin{tabular}[x]{@{}c@{}} \textit{Add} \\ \textit{sunglasses} \\ \textit{and detect} \\ \textit{them} \end{tabular}}&
\raisebox{-.5\height}{
\includegraphics[width=0.2\linewidth]{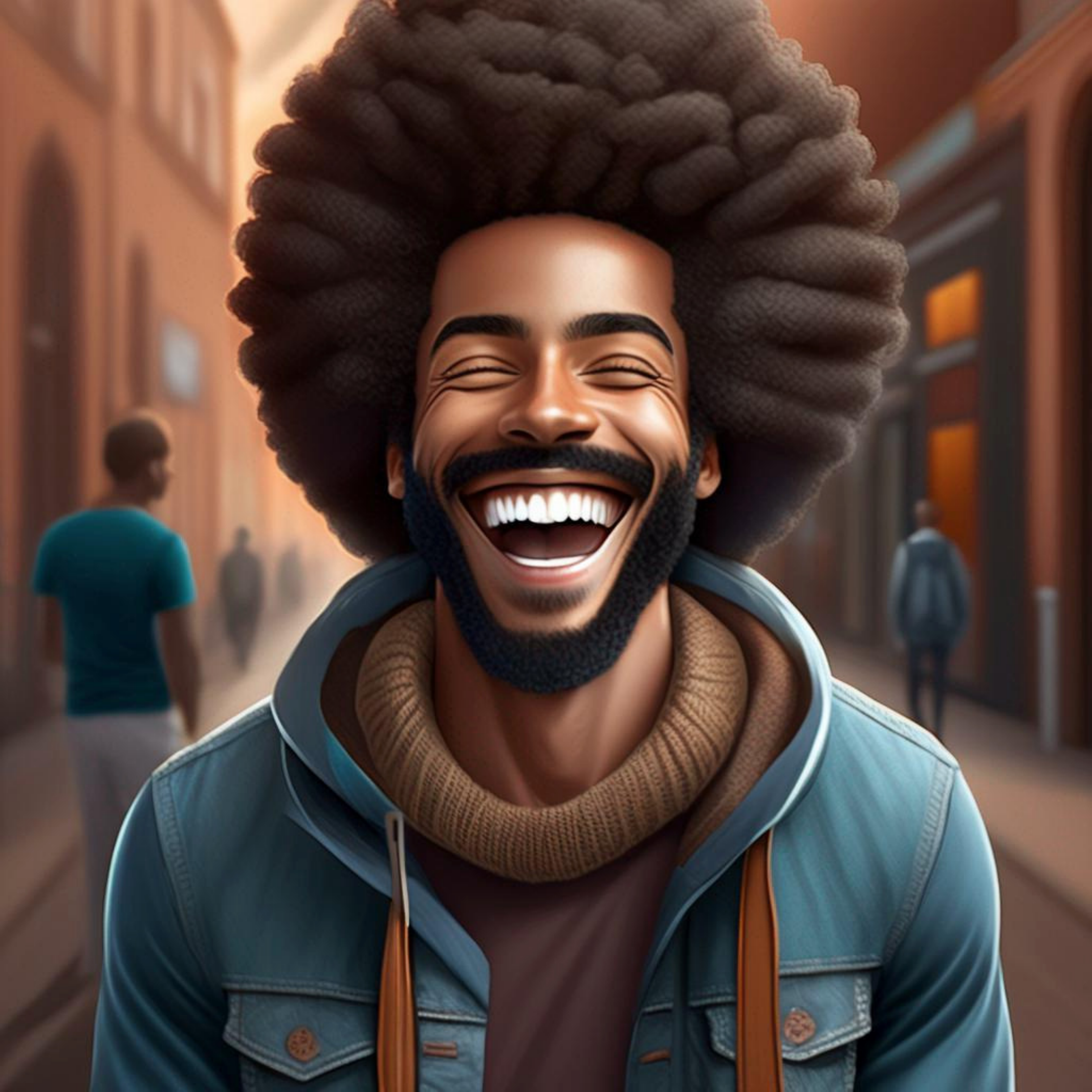}}&
\raisebox{-.5\height}{
\includegraphics[width=0.2\linewidth]{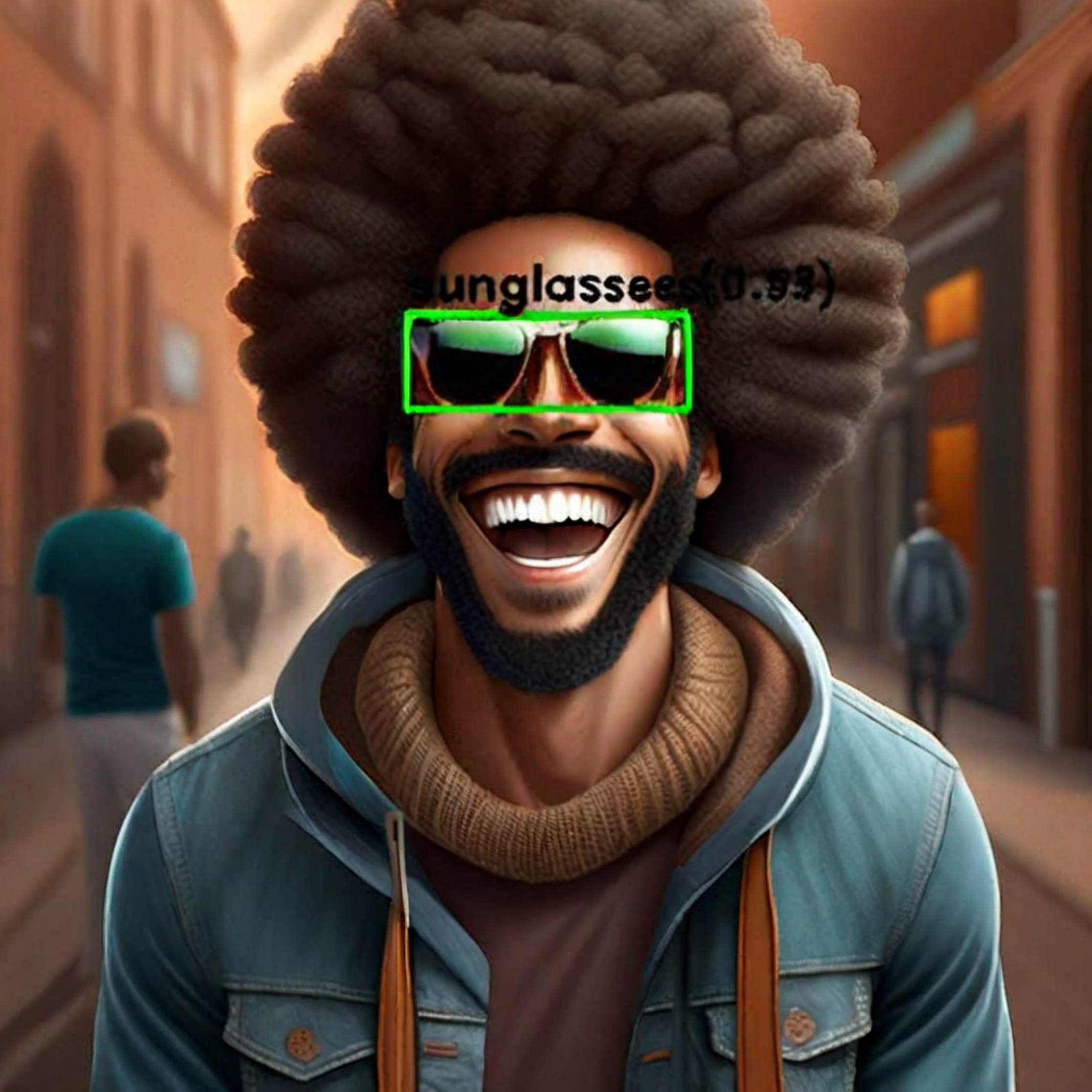}}&

\resizebox{!}{21px}{
\begin{tabular}[x]{@{}c@{}} \textit{Add a} \\ \textit{butterfly} \\ \textit{and detect} \\ \textit{it} \end{tabular}}&
\raisebox{-.5\height}{
\includegraphics[width=0.2\linewidth]{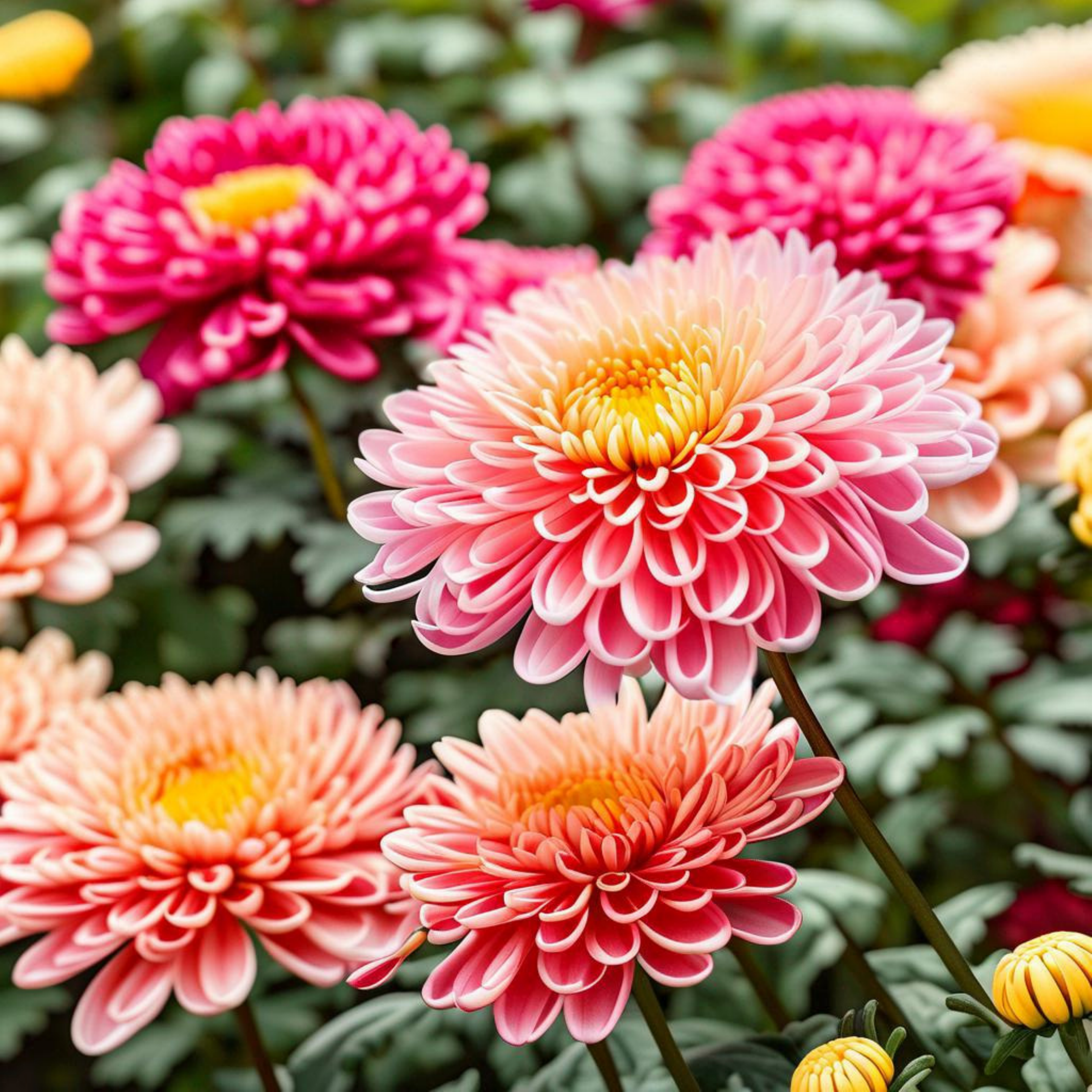}}&
\raisebox{-.5\height}{
\includegraphics[width=0.2\linewidth]{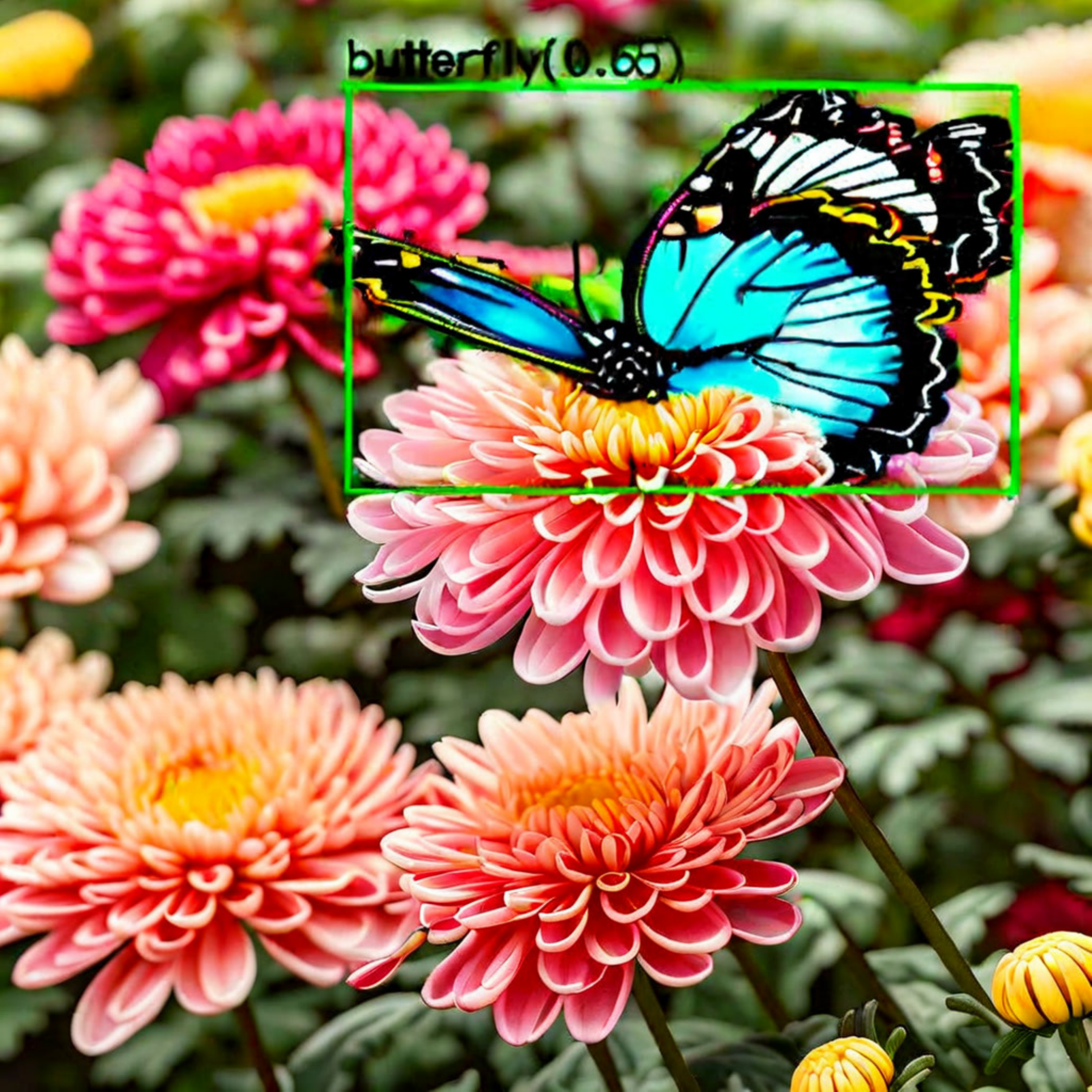}}&
\\[9mm] 

\resizebox{!}{30px}{
\begin{tabular}[x]{@{}c@{}} \textit{Add pumpkins} \\ \textit{next to the} \\ \textit{squirrel and} \\ \textit{change the} \\ \textit{style to} \\ \textit{3D rendering} \end{tabular}}&
\raisebox{-.5\height}{
\includegraphics[width=0.2\linewidth]{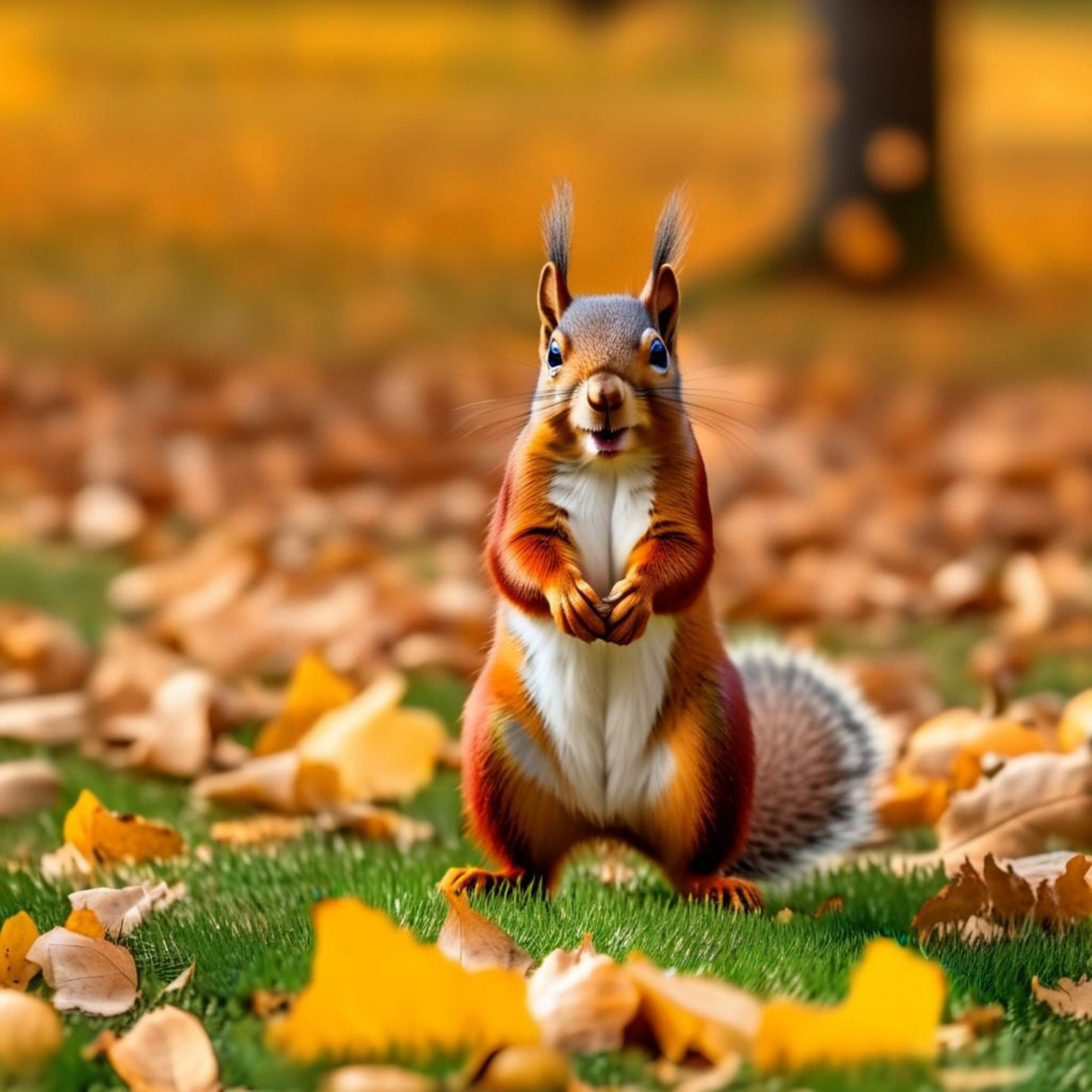}}&
\raisebox{-.5\height}{
\includegraphics[width=0.2\linewidth]{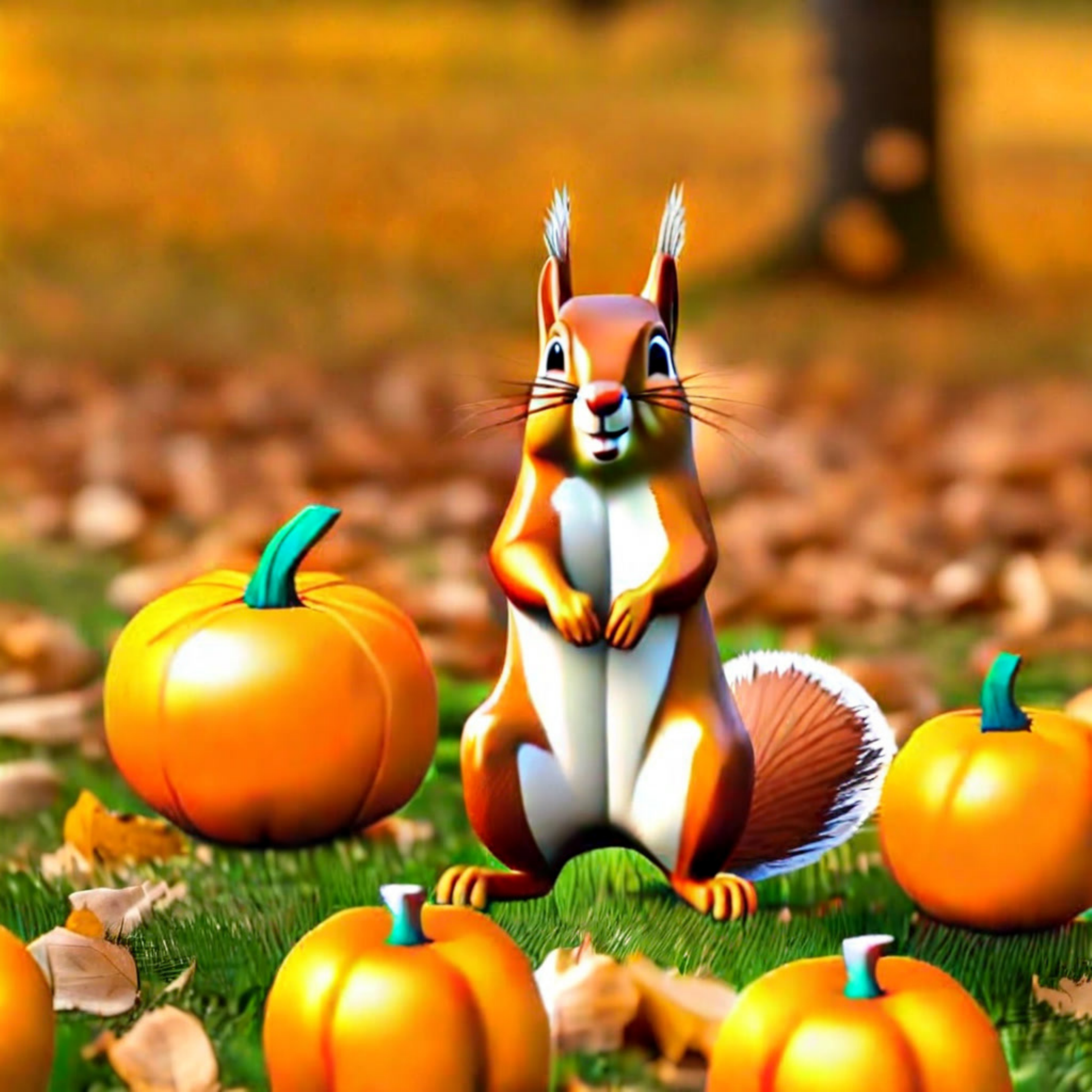}}&

\resizebox{!}{30px}{
\begin{tabular}[x]{@{}c@{}} \textit{Include a} \\ \textit{flock of birds} \\ \textit{flying and} \\ \textit{make it a} \\ \textit{vintage} \\ \textit{photograph}.
 \end{tabular}}&
\raisebox{-.5\height}{
\includegraphics[width=0.2\linewidth]{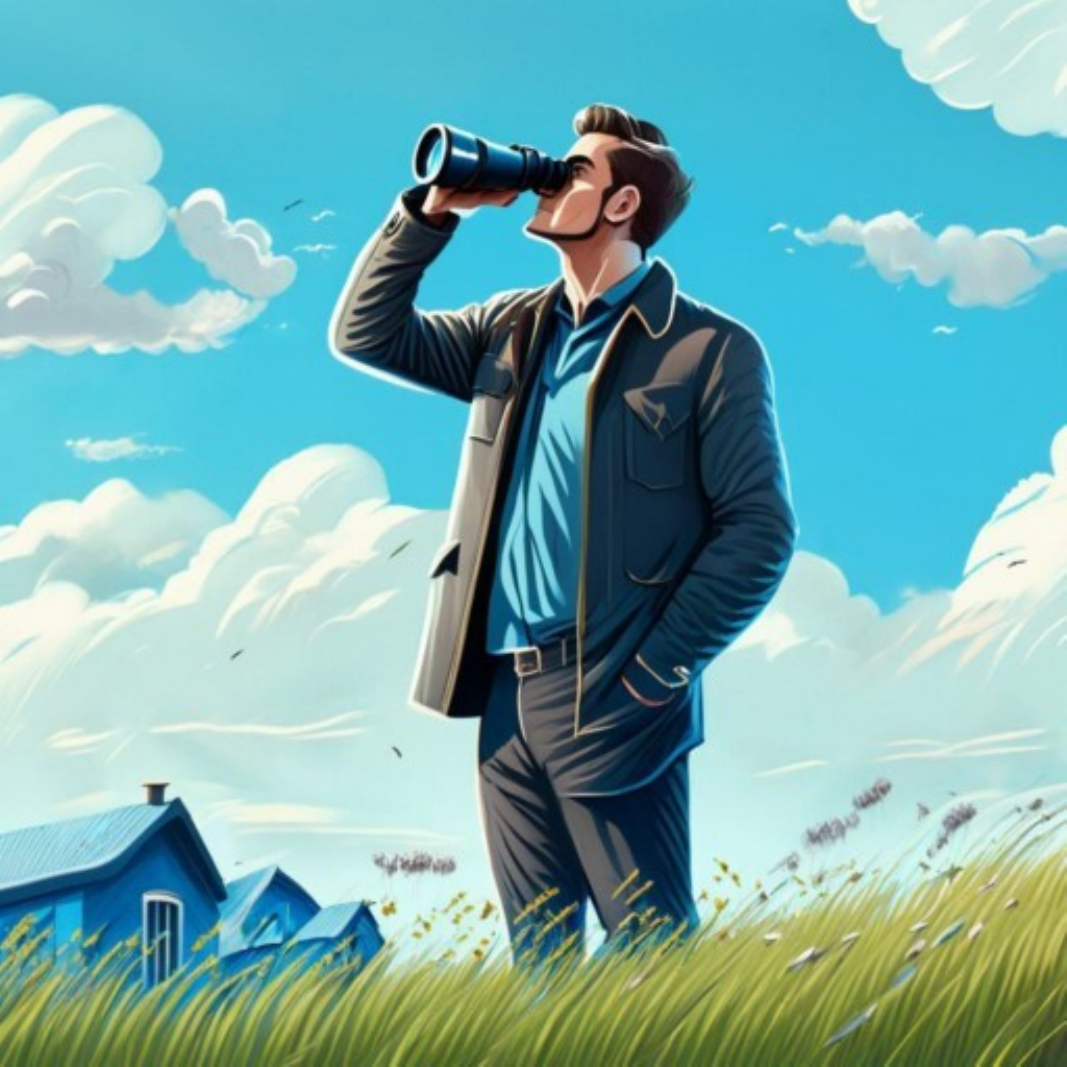}}&
\raisebox{-.5\height}{
\includegraphics[width=0.2\linewidth]{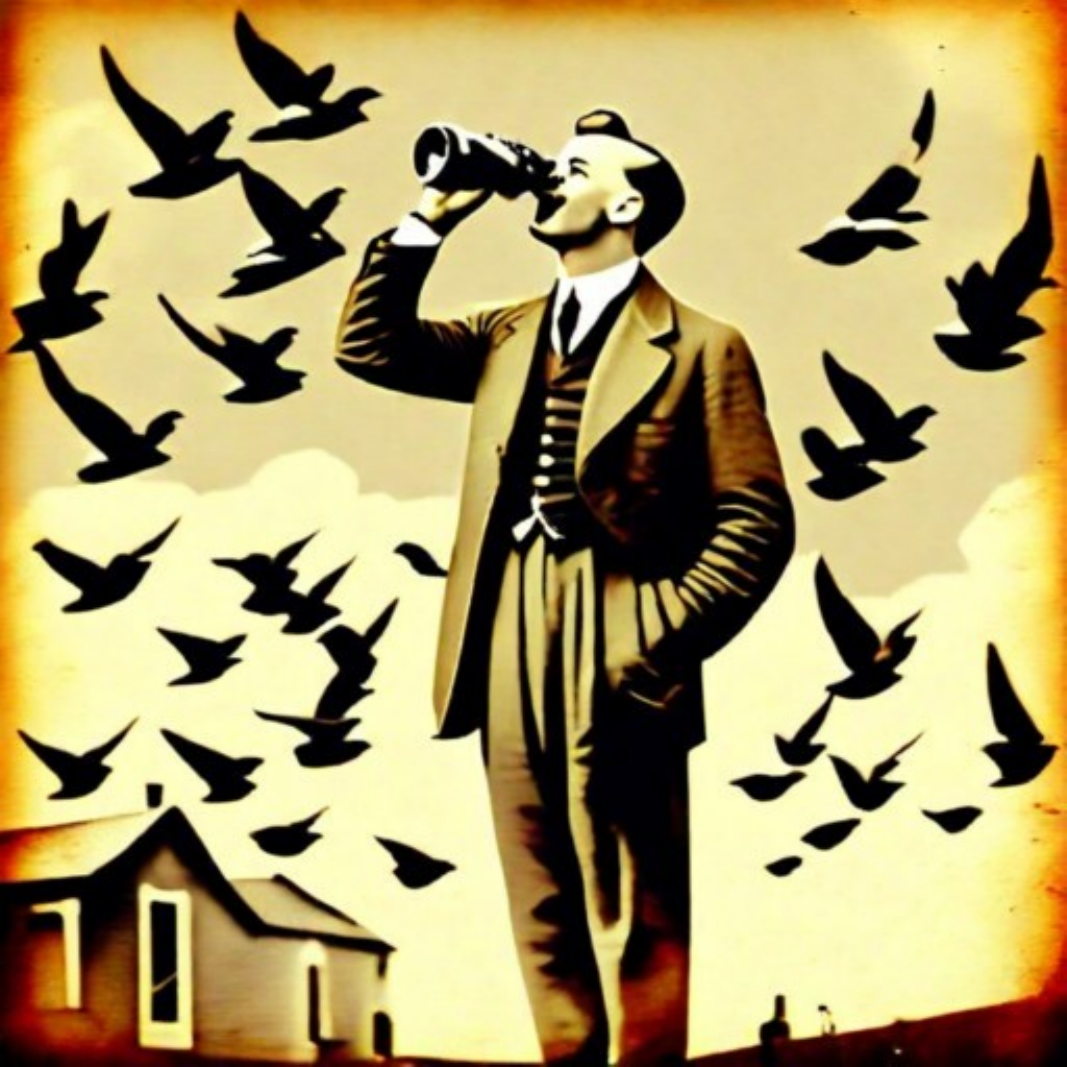}}&
\\[18mm]

\resizebox{!}{28px}{
\begin{tabular}[x]{@{}c@{}} \textit{Fill the} \\ \textit{missing} \\ \textit{pixels:} \\ \textit{halloween} \\ \textit{giant chocolate} \\ \textit{chip cookie} \end{tabular}}&
\raisebox{-.5\height}{
\includegraphics[width=0.2\linewidth]{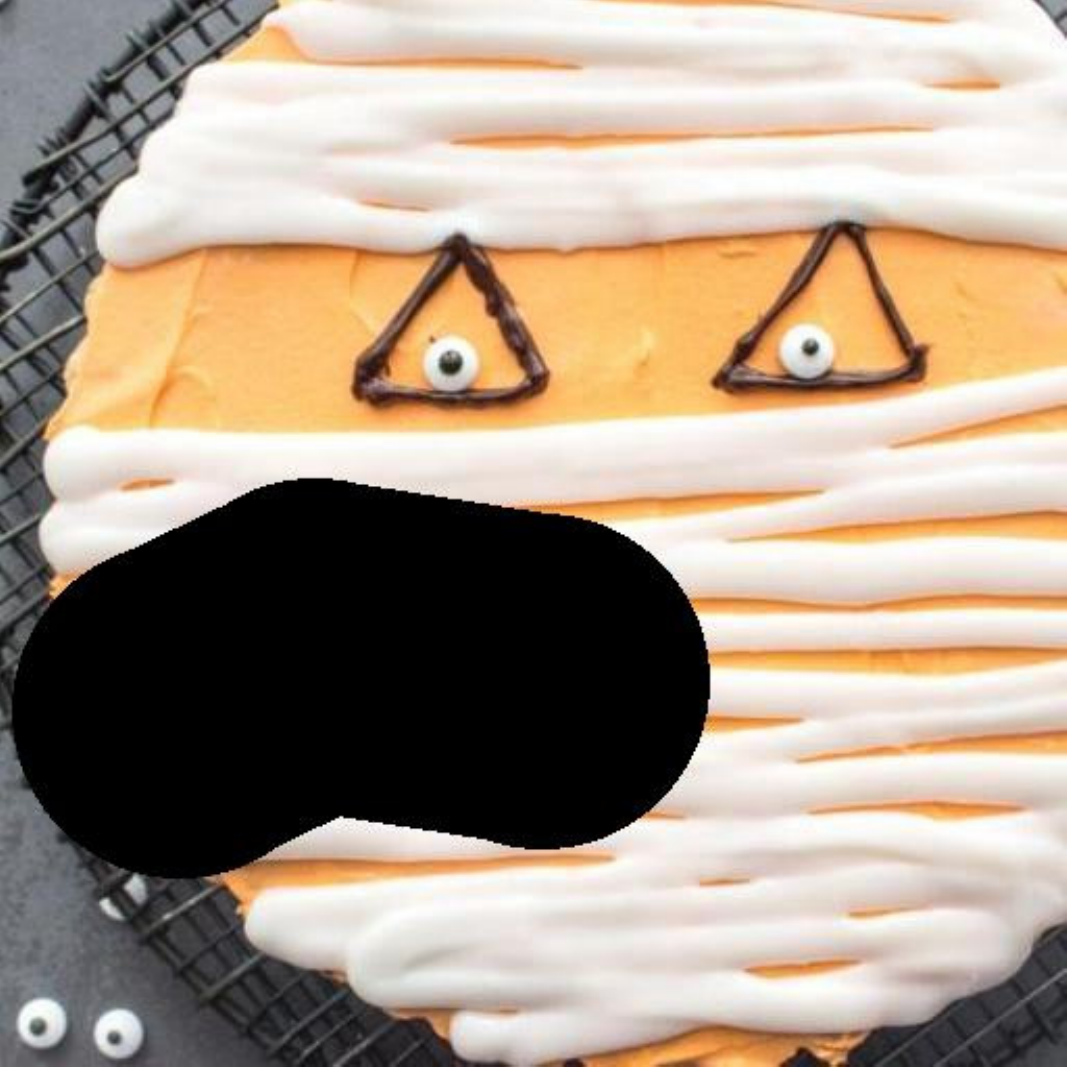}}&
\raisebox{-.5\height}{
\includegraphics[width=0.2\linewidth]{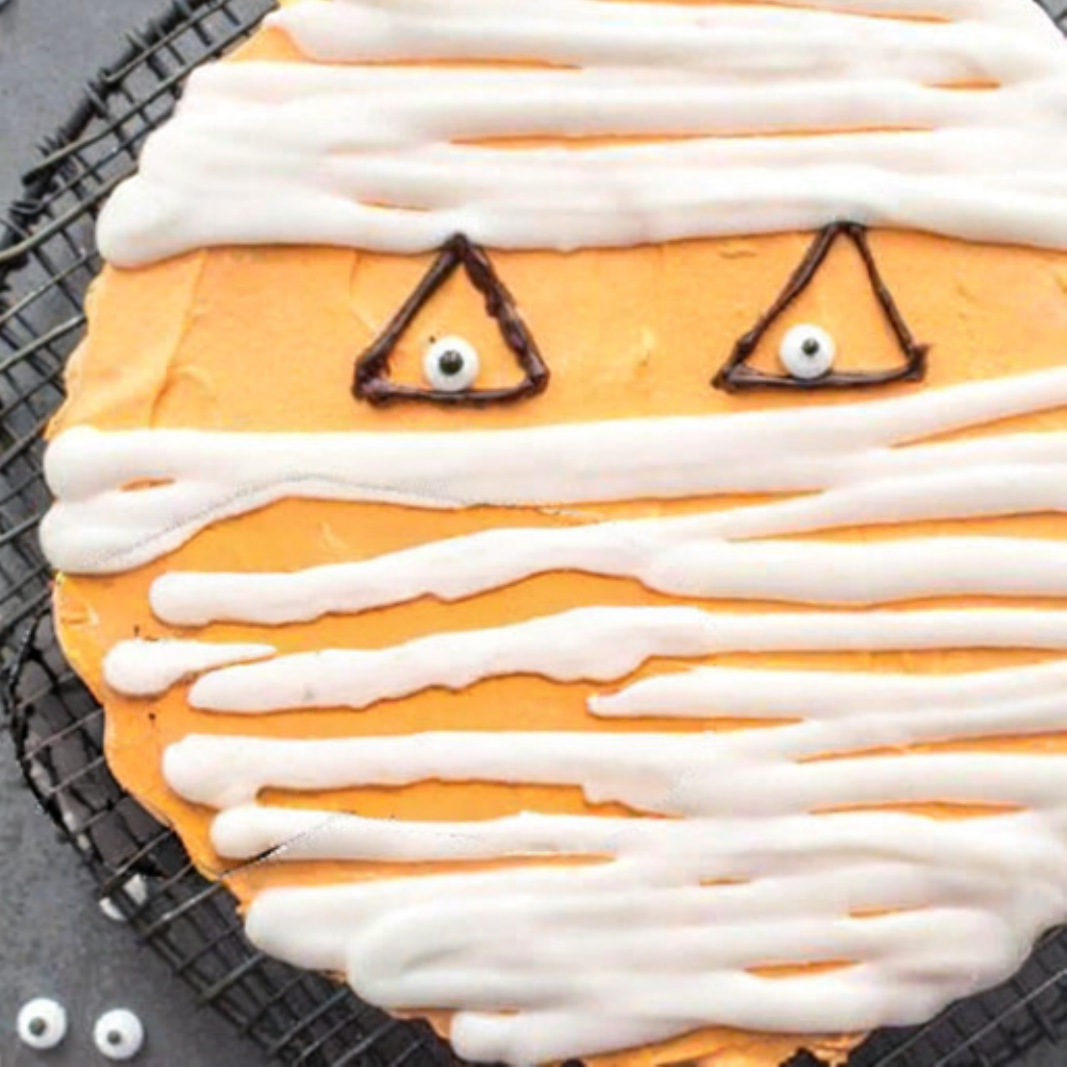}}&

\resizebox{!}{28px}{
\begin{tabular}[x]{@{}c@{}} \textit{Fill the} \\ \textit{missing} \\ \textit{pixels:} \\ \textit{a view of} \\ \textit{a city} \\ \textit{from afar} \end{tabular}}&
\raisebox{-.5\height}{
\includegraphics[width=0.2\linewidth]{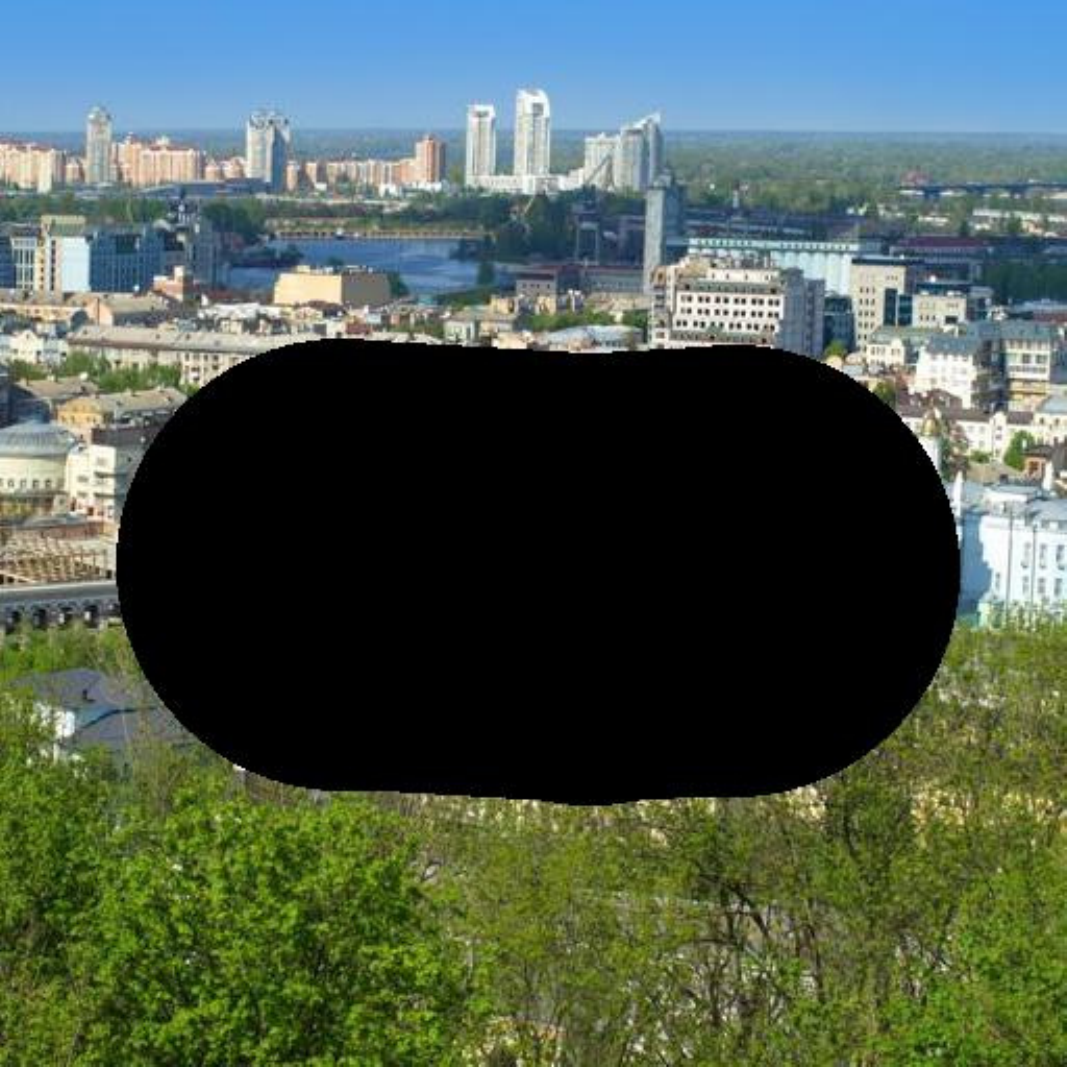}}&
\raisebox{-.5\height}{
\includegraphics[width=0.2\linewidth]{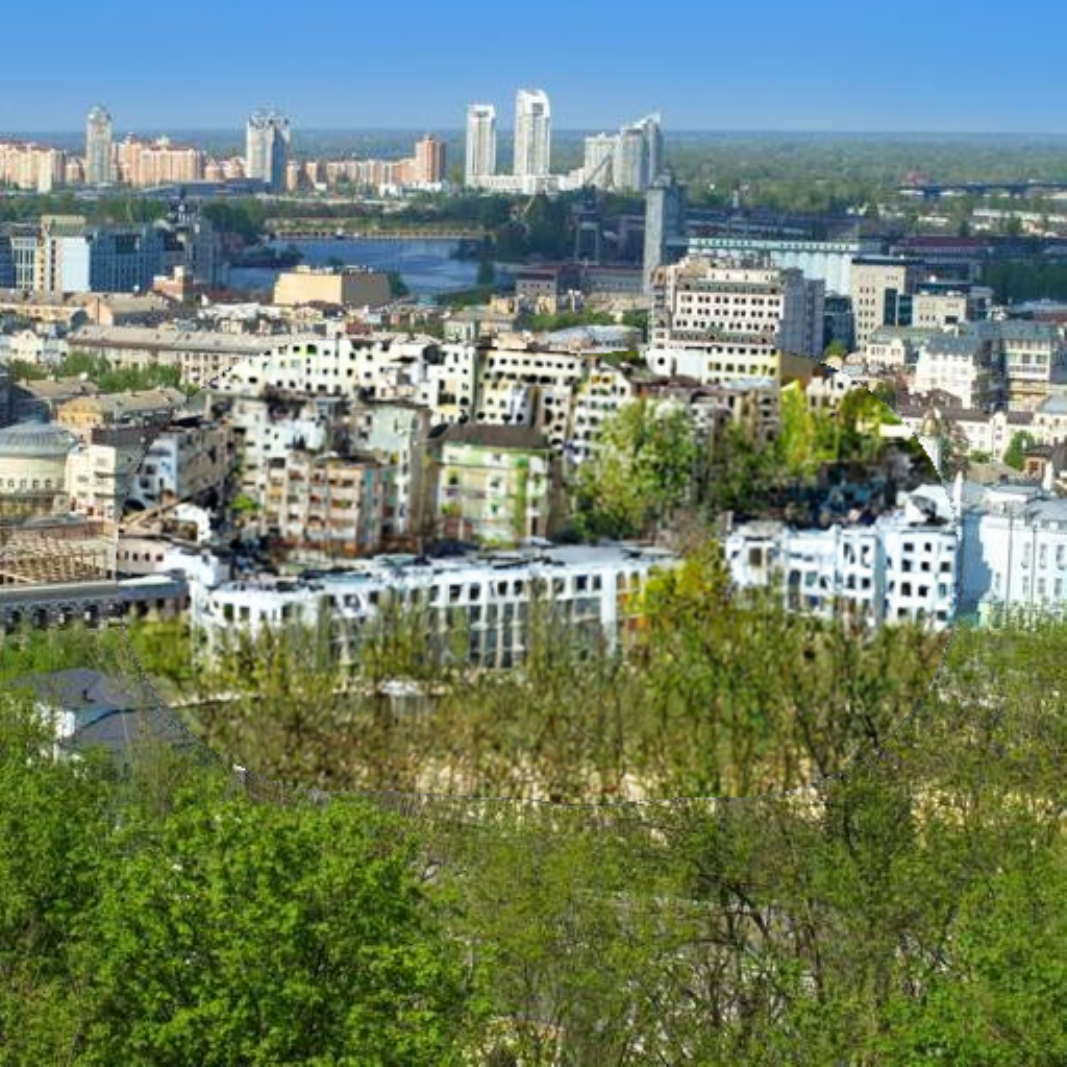}}&
\\[9mm]

\resizebox{!}{18px}{
\begin{tabular}[x]{@{}c@{}} \textit{Mark the} \\  \textit{watch} \\  \textit{faces} \end{tabular}}&
\raisebox{-.5\height}{
\includegraphics[width=0.2\linewidth]{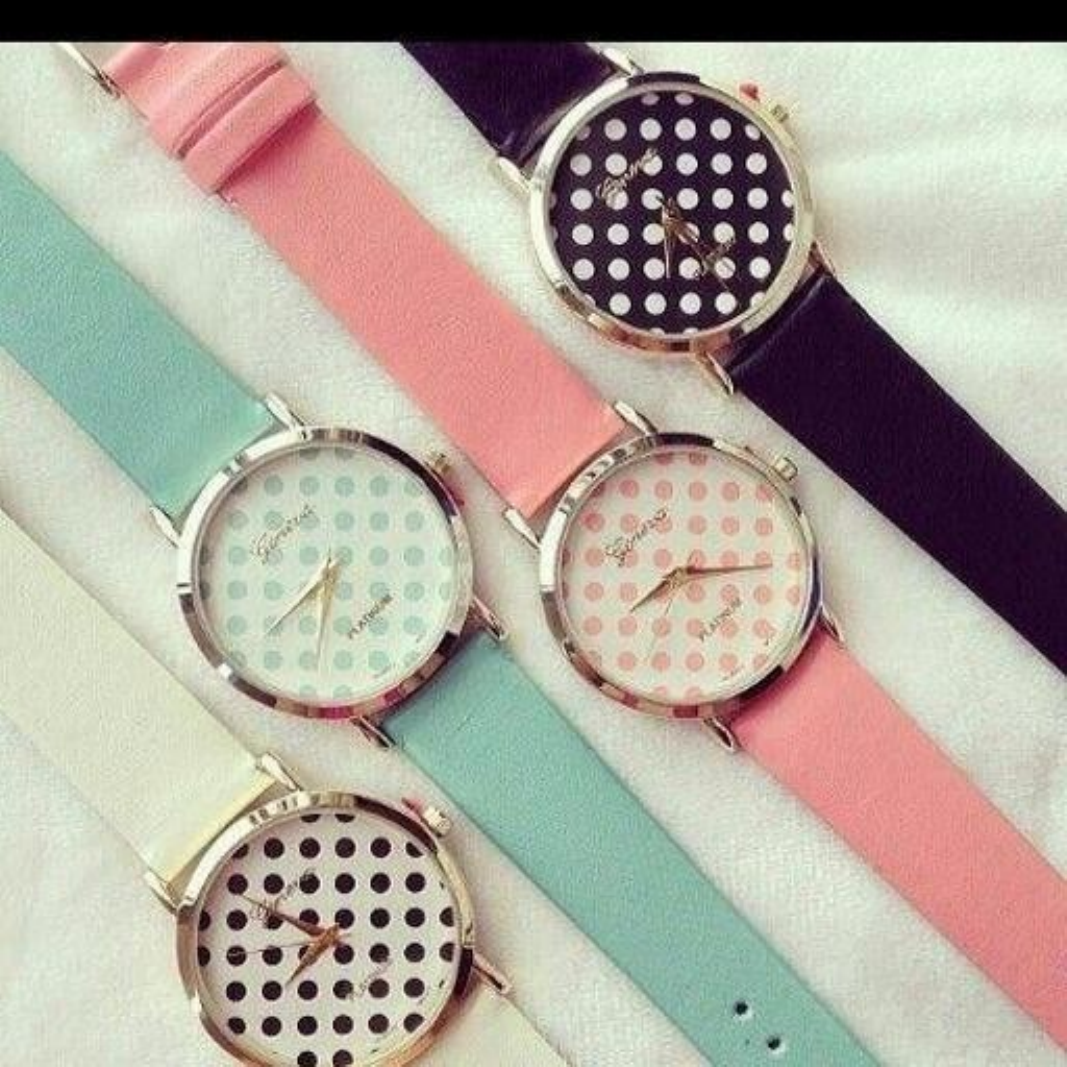}}&
\raisebox{-.5\height}{
\includegraphics[width=0.2\linewidth]{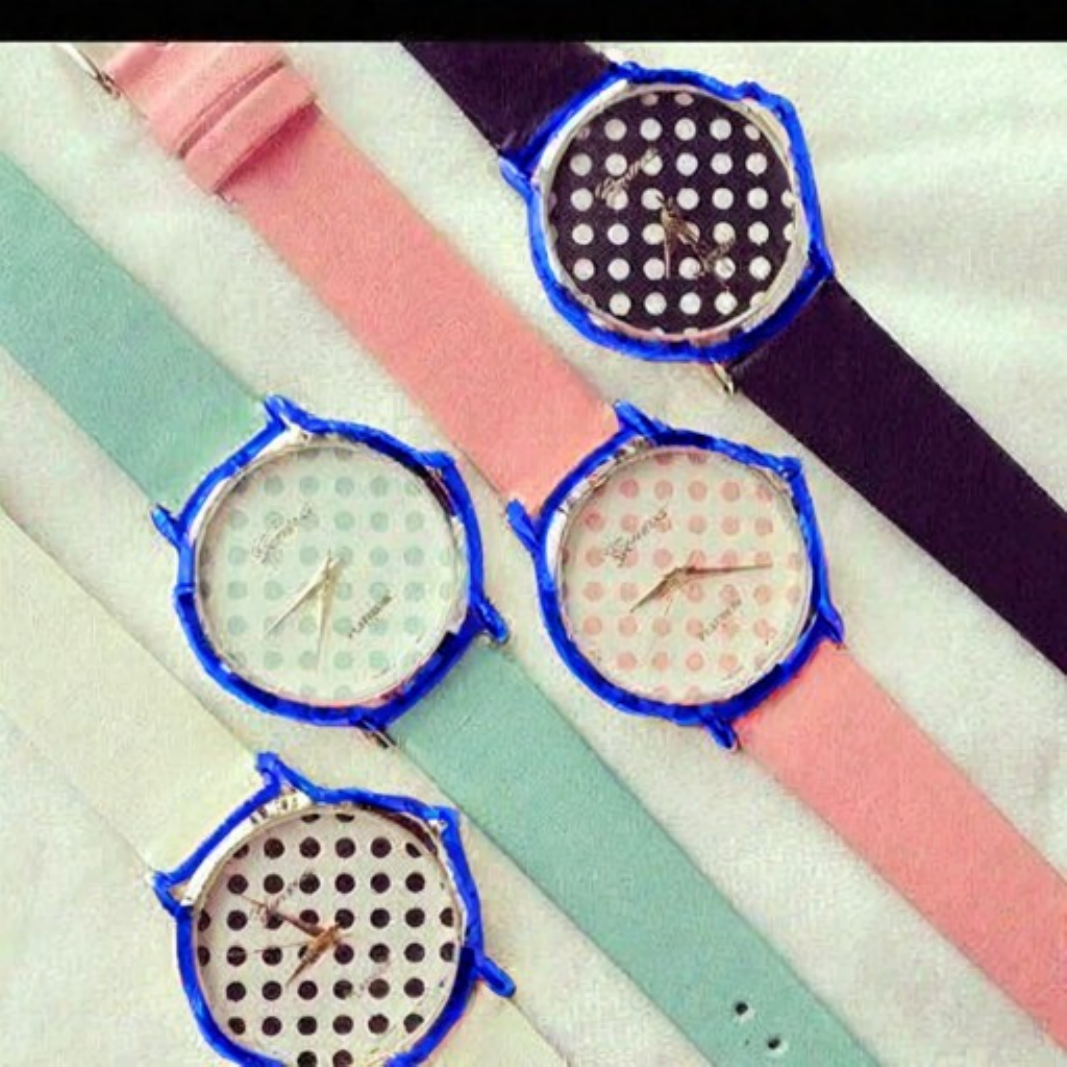}} &

\resizebox{!}{18px}{
\begin{tabular}[x]{@{}c@{}} \textit{Mark} \\ \textit{the} \\ \textit{furniture} \end{tabular}}&
\raisebox{-.5\height}{
\includegraphics[width=0.2\linewidth]{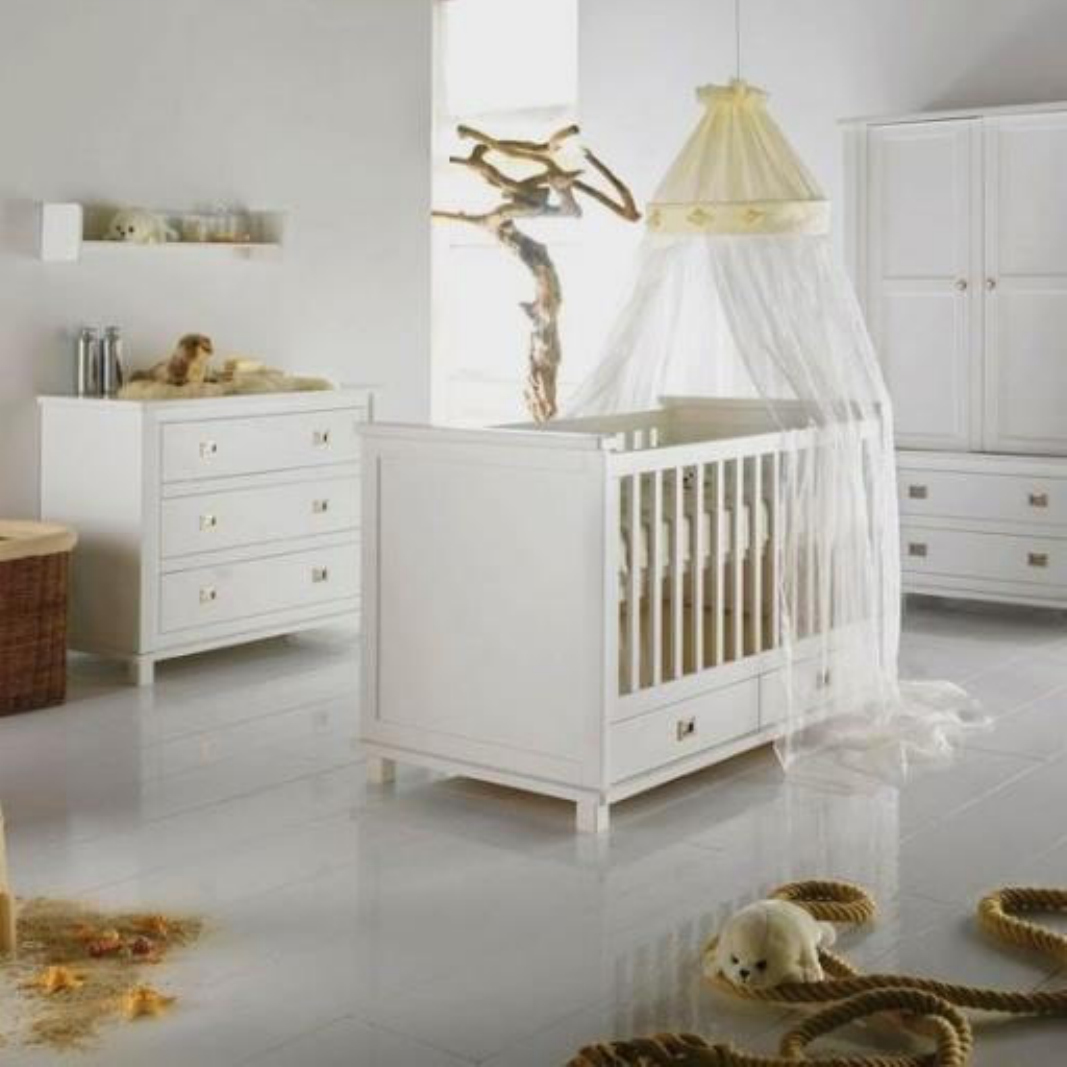}}&
\raisebox{-.5\height}{
\includegraphics[width=0.2\linewidth]{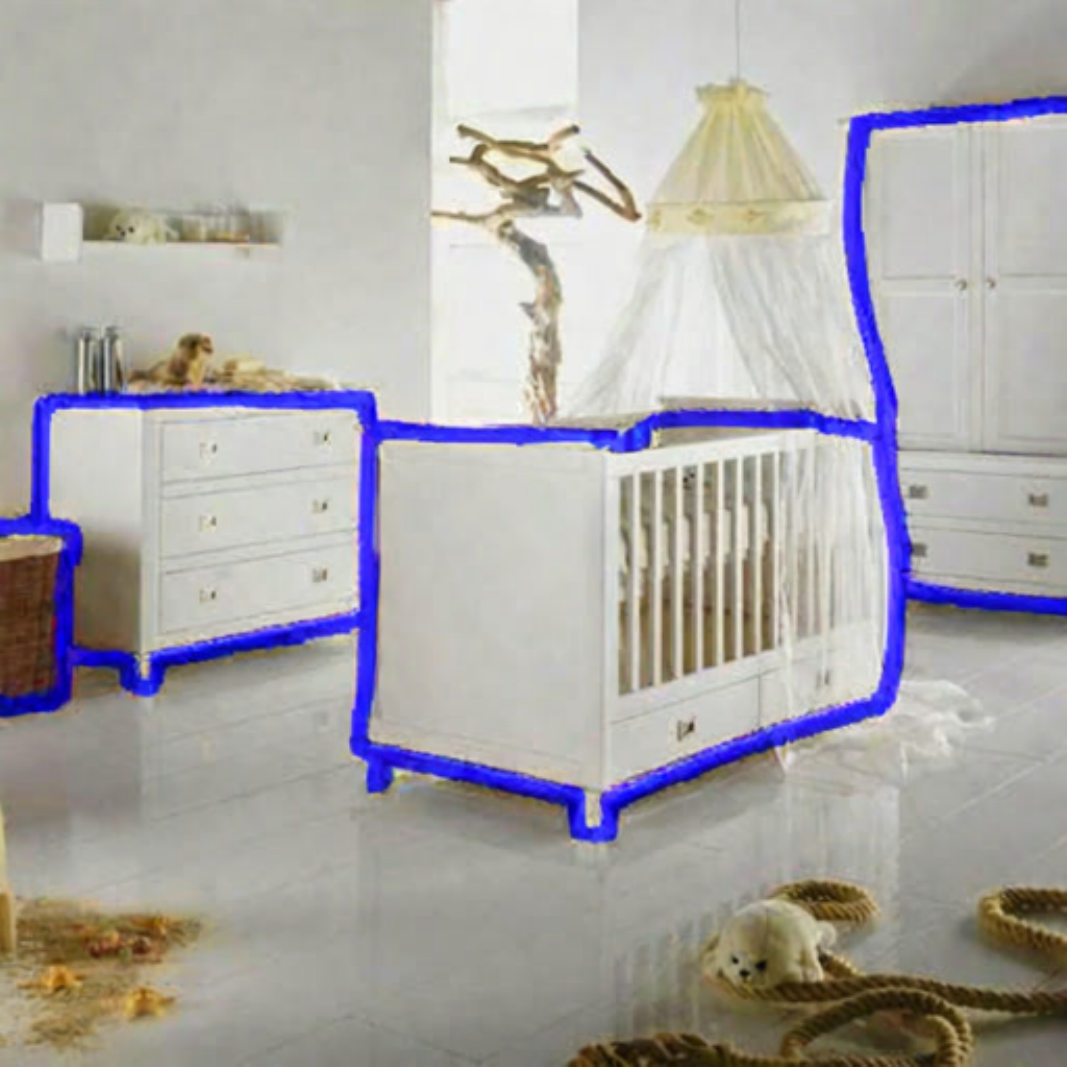}}& \\ [9mm]

\resizebox{!}{18px}{
\begin{tabular}[x]{@{}c@{}} \textit{Upsample} \\ \textit{the} \\ \textit{resolution}\end{tabular}}&
\raisebox{-.5\height}{
\includegraphics[width=0.2\linewidth]{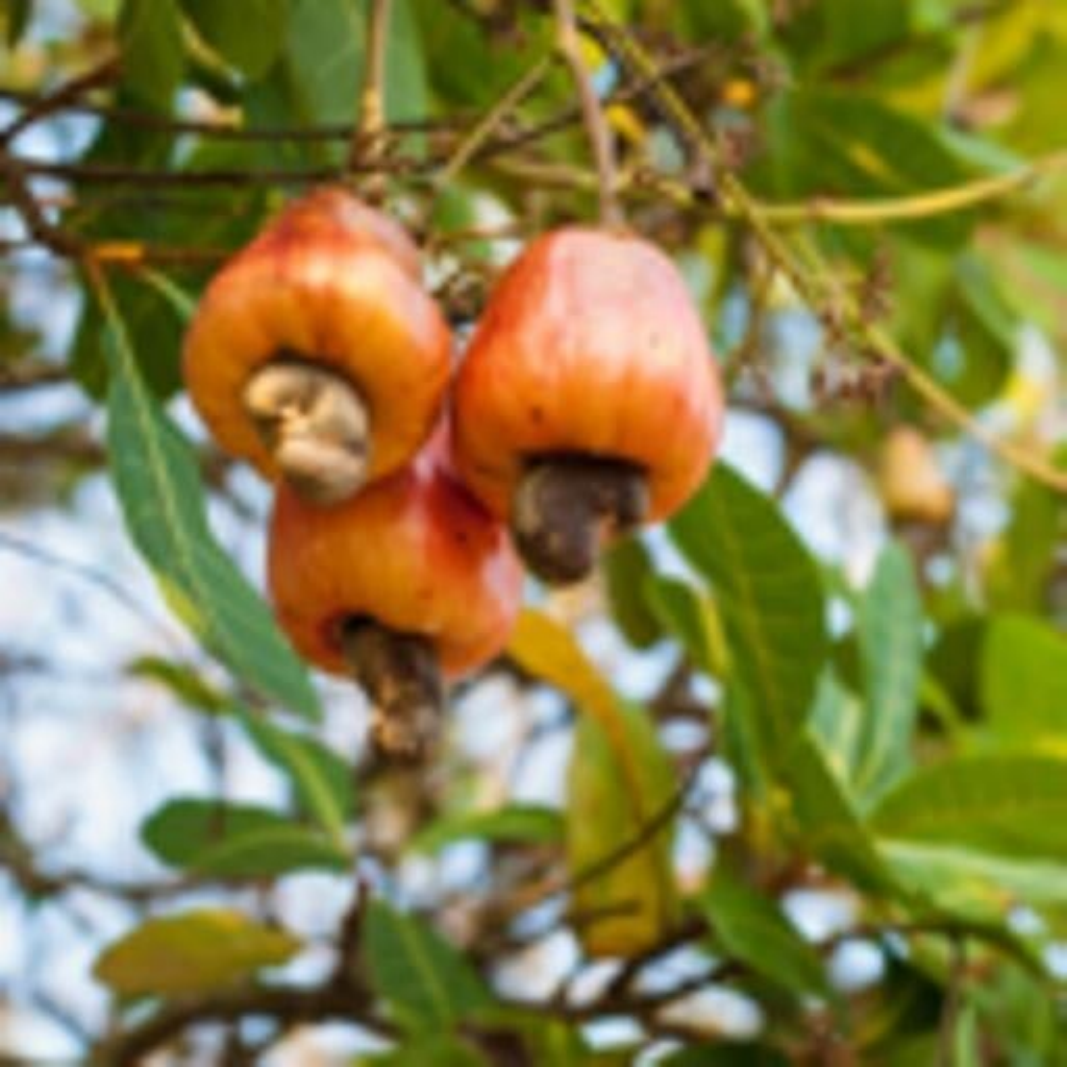}}&
\raisebox{-.5\height}{
\includegraphics[width=0.2\linewidth]{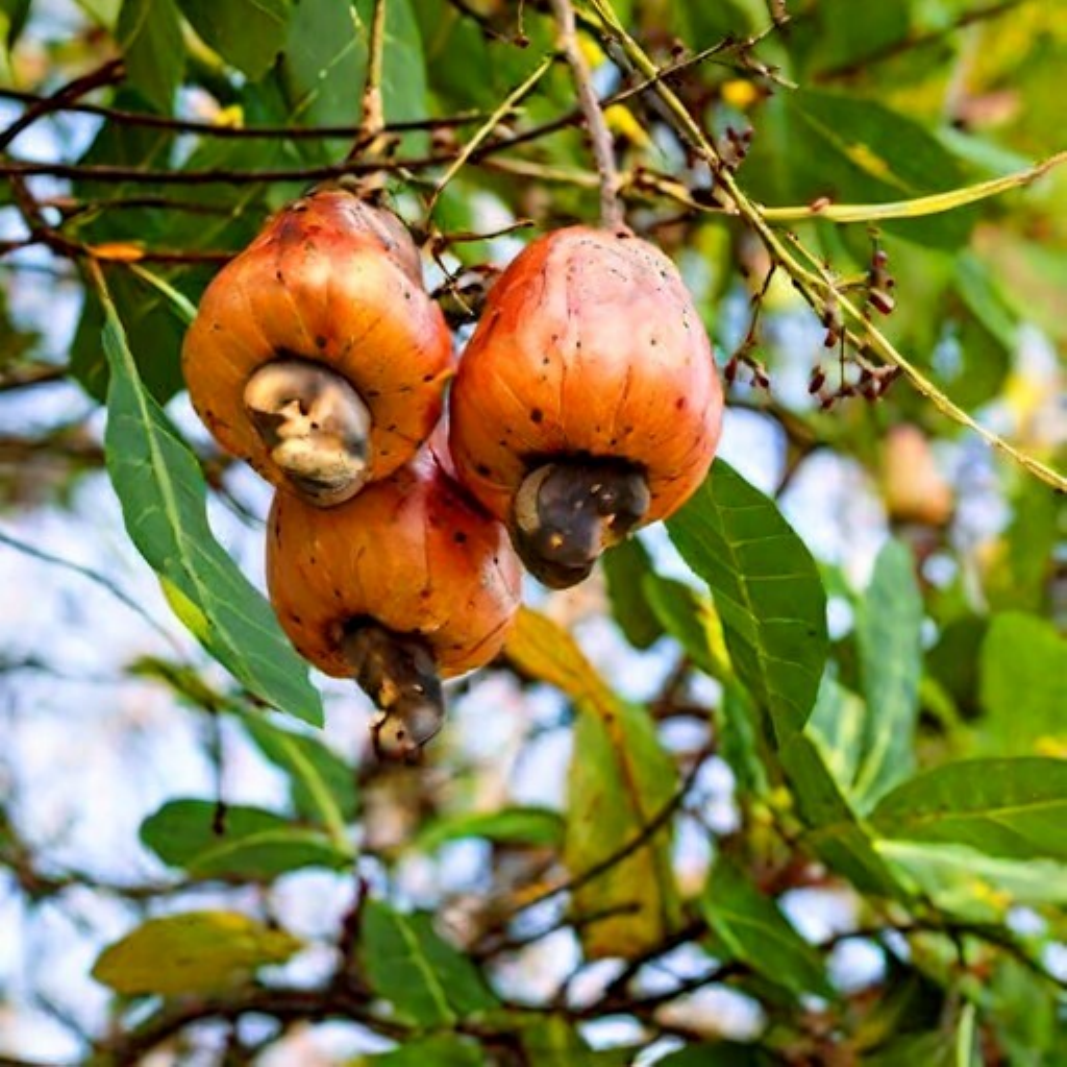}} &

\resizebox{!}{18px}{
\begin{tabular}[x]{@{}c@{}} \textit{Upsample} \\ \textit{the} \\ \textit{resolution}\end{tabular}}&
\raisebox{-.5\height}{
\includegraphics[width=0.2\linewidth]{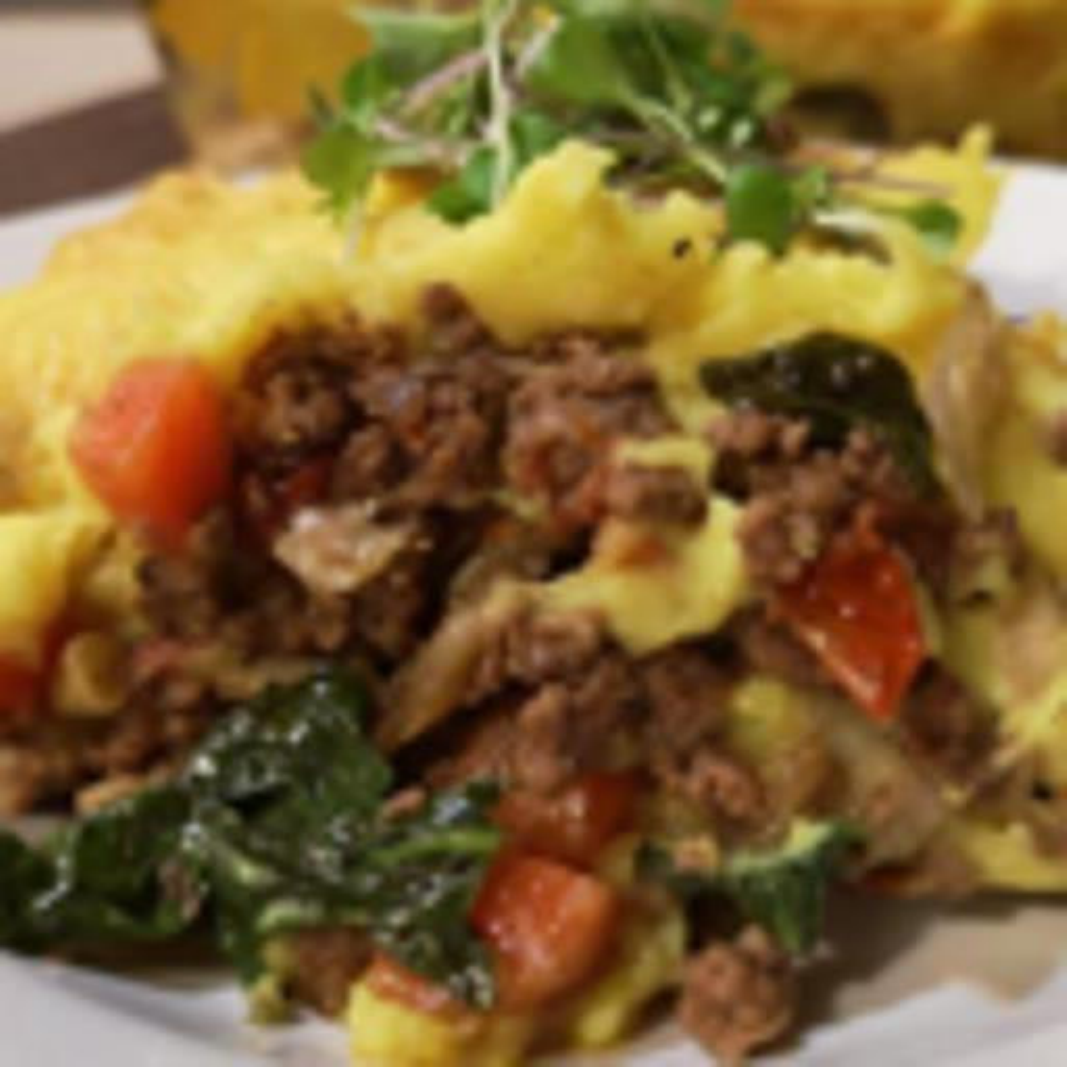}}&
\raisebox{-.5\height}{
\includegraphics[width=0.2\linewidth]{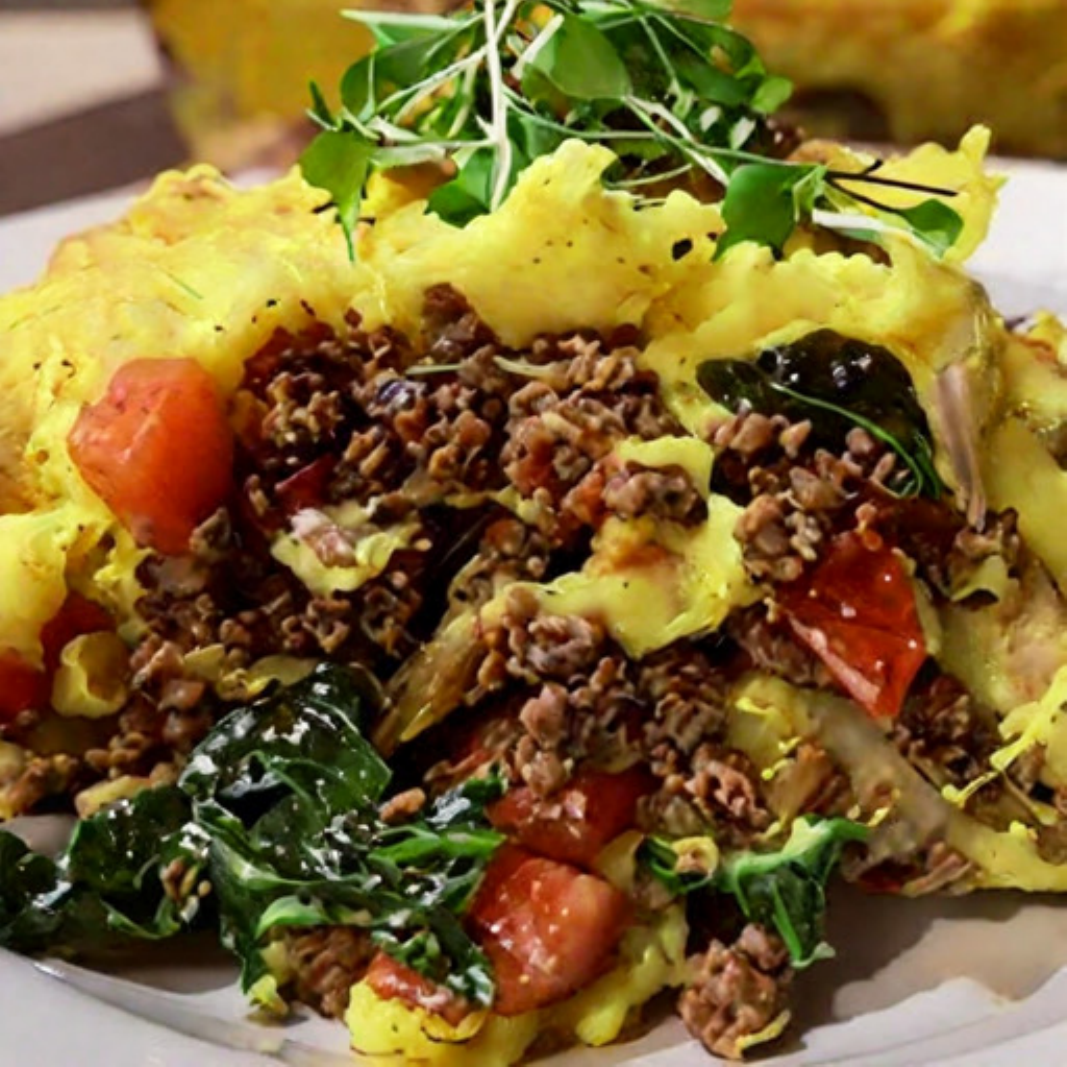}}& \\

& \begin{tabular}[x]{@{}c@{}}Input \end{tabular}  & \begin{tabular}[x]{@{}c@{}}Emu Edit \end{tabular} & &\begin{tabular}[x]{@{}c@{}}Input \end{tabular}  & \begin{tabular}[x]{@{}c@{}}Emu Edit \end{tabular}
\end{tabular}
\caption{Generations of our model on unseen tasks with task inversion. From top to bottom: (i) composition of add and detect tasks, (ii) composition of add and style tasks, (iii) image in-painting, (iv) contour detection, (v) super-resolution.
}
\label{fig:few_shot_examples_sup}
\end{figure*}

Fig.~\ref{fig:few_shot_examples_sup} illustrates generation examples produced by our model for various tasks learned in a few-shot setting. Additionally, the performance results of our model on the tasks of super resolution and contour detection are presented in Fig.~\ref{fig:few-shot_supp}.

\begin{figure*}[t]
   \centering
\begin{tabular}{@{\hspace{-9\tabcolsep}}c@{\hspace{-0.3\tabcolsep}}c@{\hspace{-0.3\tabcolsep}}c@{\hspace{-0.3\tabcolsep}}c@{\hspace{-0.3\tabcolsep}}c@{\hspace{-0.3\tabcolsep}}c@{\hspace{-0.3\tabcolsep}}c@{\hspace{-0.3\tabcolsep}}c@{\hspace{-0.3\tabcolsep}}c}
& \makecell{\resizebox{!}{4px}{A dog playing guitar} \\ \resizebox{!}{4px}{on the beach}}  & \resizebox{!}{4px}{Turn to an electric guitar} & \resizebox{!}{4px}{Make the sea wavy} &  \resizebox{!}{4px}{Change dog color to white}  &  \resizebox{!}{4px}{Turn guitar to red}  & \resizebox{!}{4px}{Add the word "Hello"}  & \resizebox{!}{4px}{Replace stone with sea shell}  & \resizebox{!}{4px}{Make it cloudy}  \\

\resizebox{!}{8px}{
\begin{tabular}[x]{@{}c@{}} \footnotesize{$\alpha=0$} \end{tabular}}&
\raisebox{-.5\height}{
\includegraphics[width=0.11\linewidth]{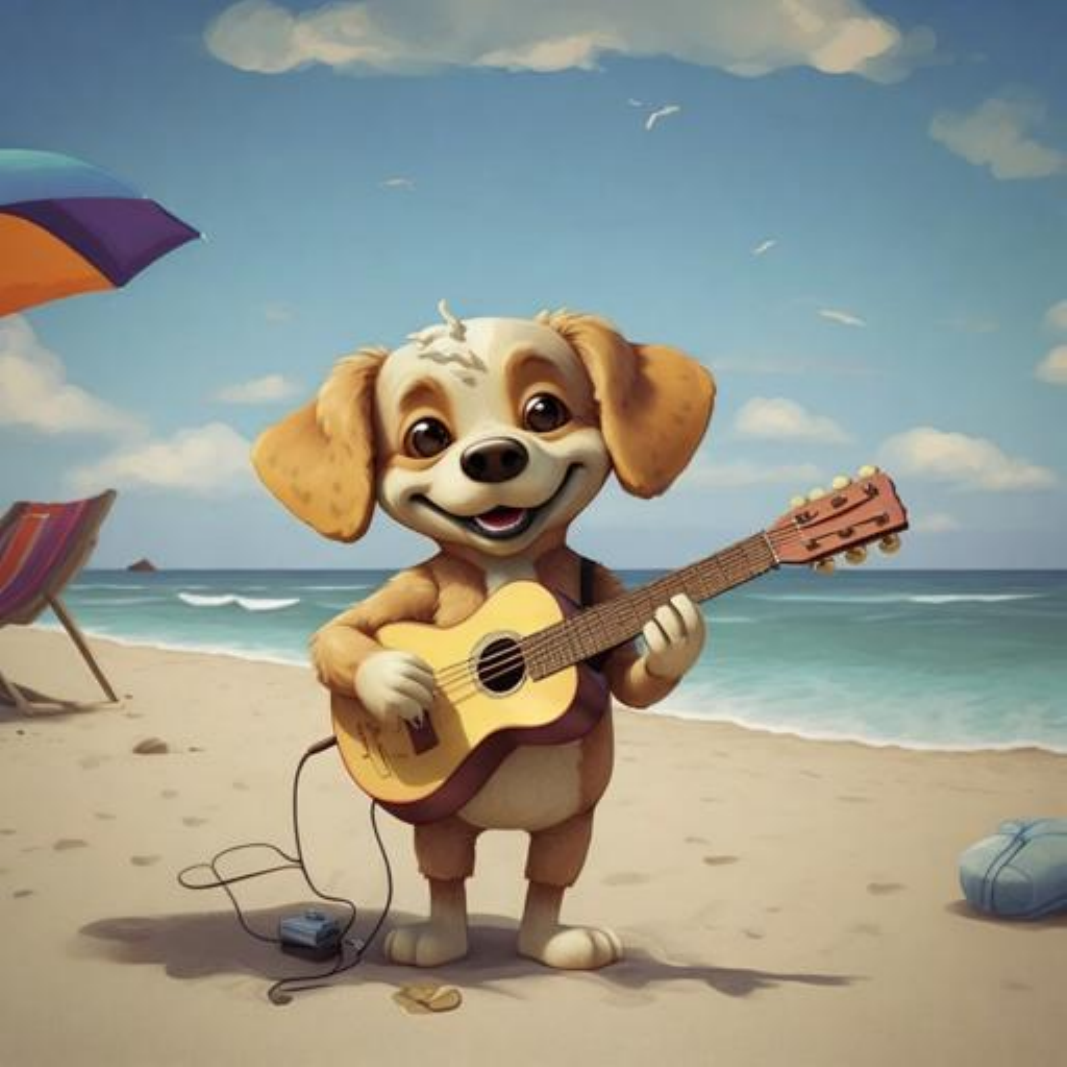}}&
\raisebox{-.5\height}{
\includegraphics[width=0.11\linewidth]{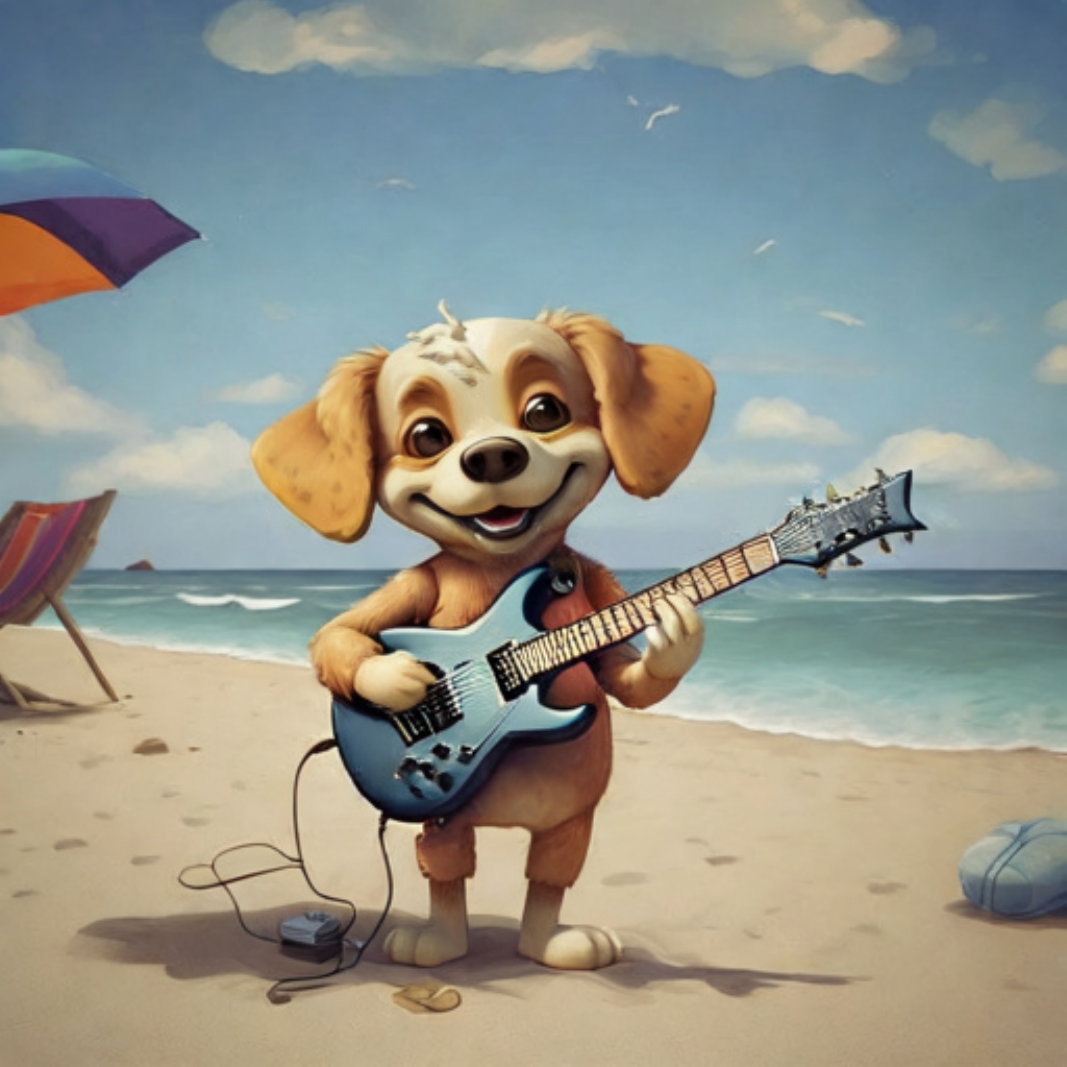}}&
\raisebox{-.5\height}{
\includegraphics[width=0.11\linewidth]{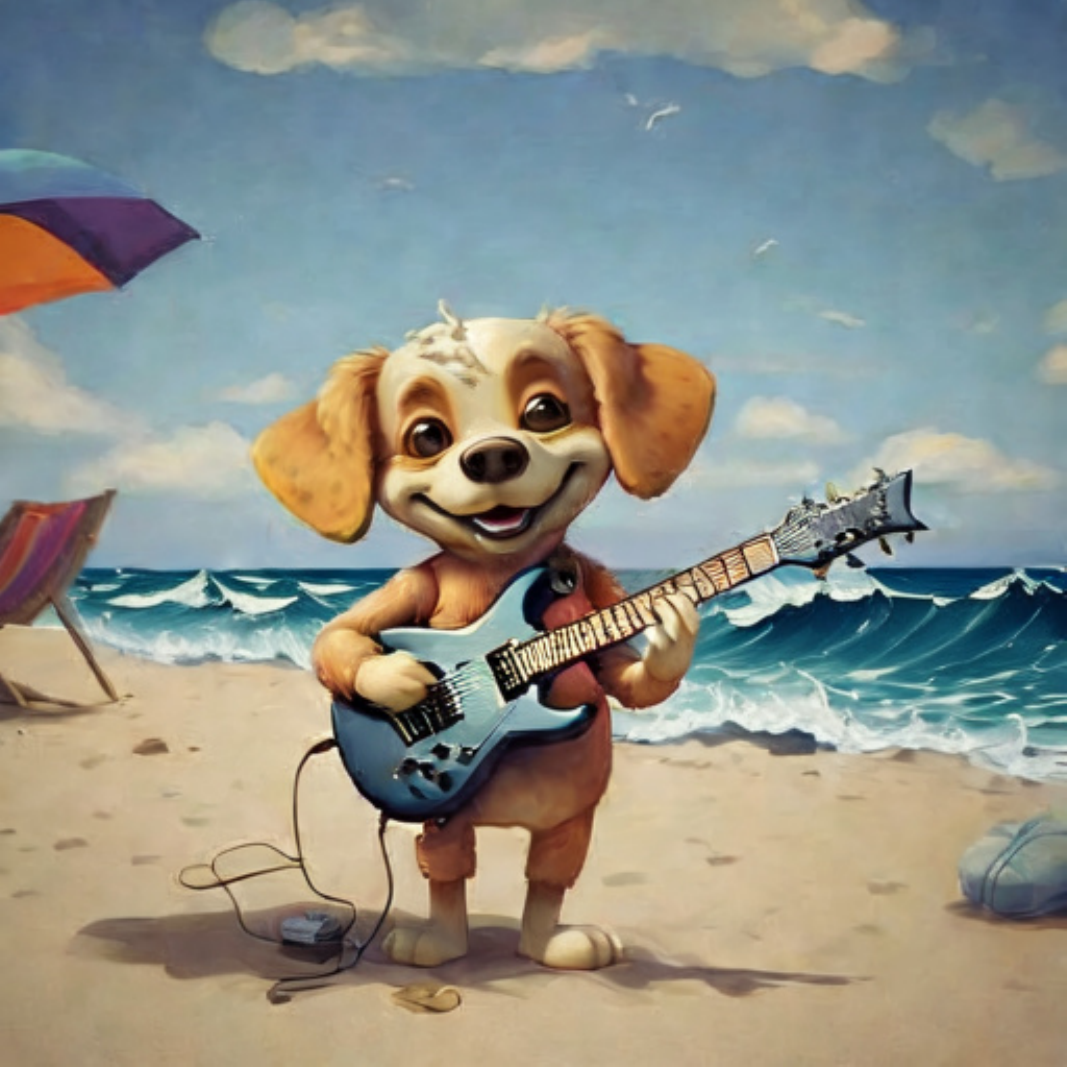}}&
\raisebox{-.5\height}{
\includegraphics[width=0.11\linewidth]{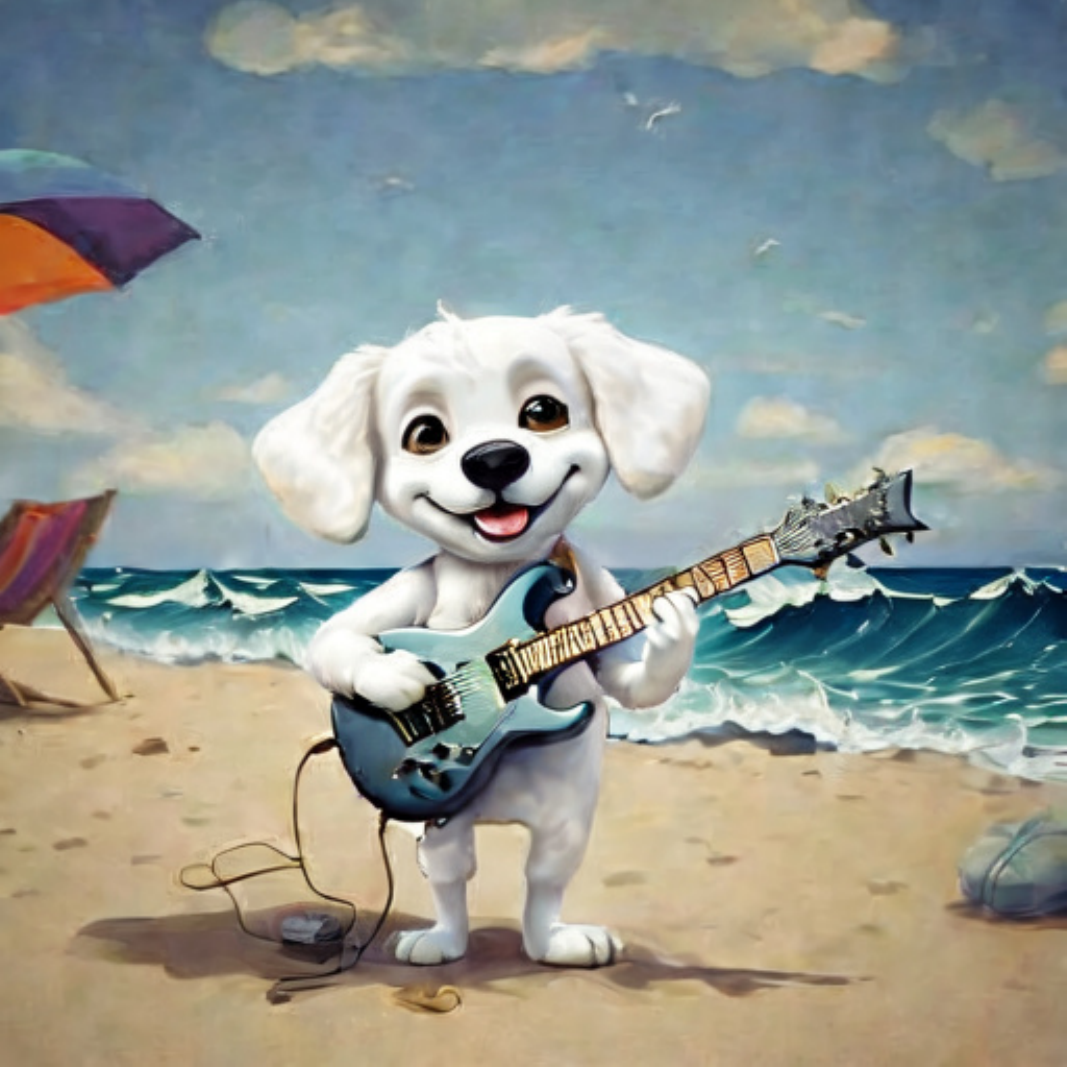}}&
\raisebox{-.5\height}{
\includegraphics[width=0.11\linewidth]{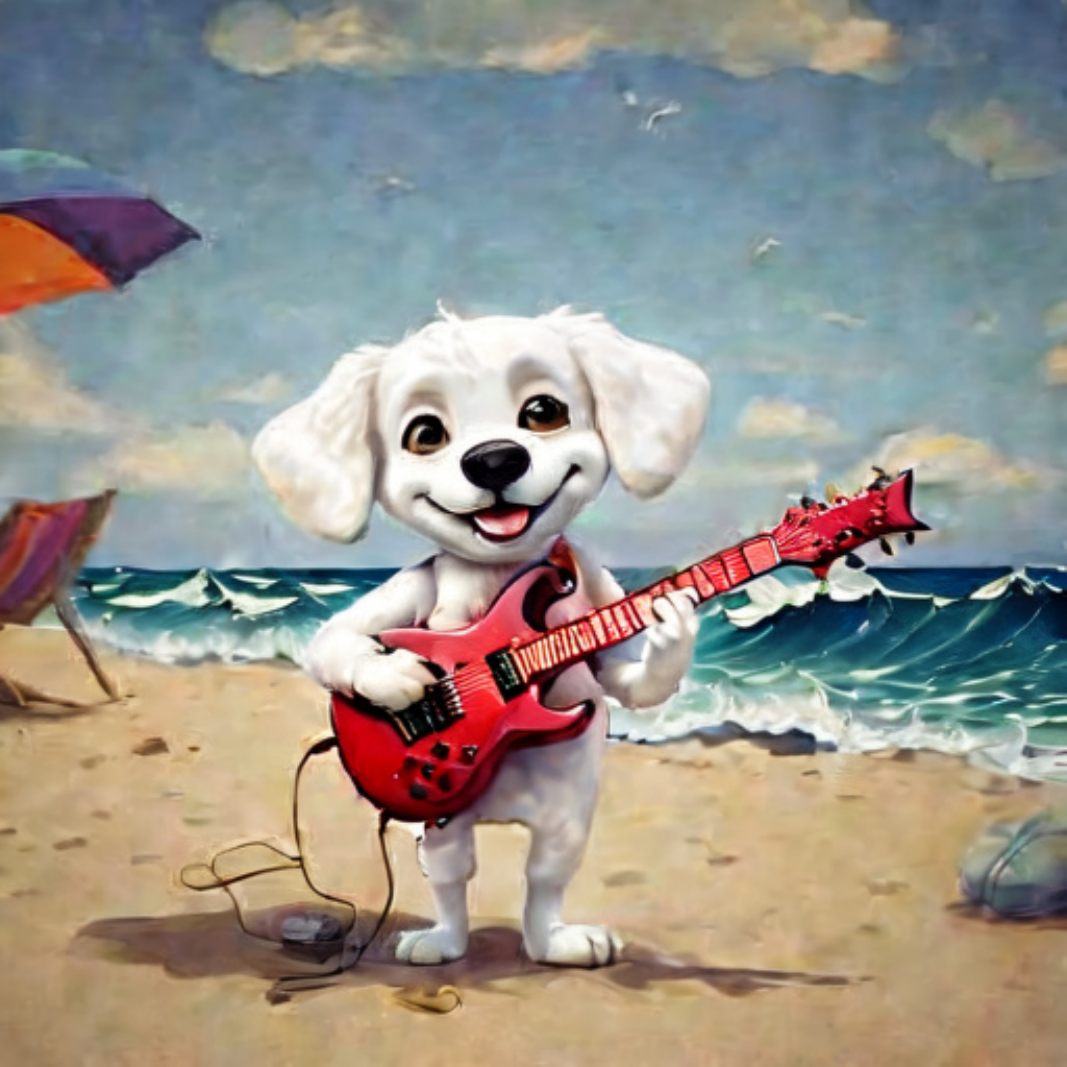}}&
\raisebox{-.5\height}{
\includegraphics[width=0.11\linewidth]{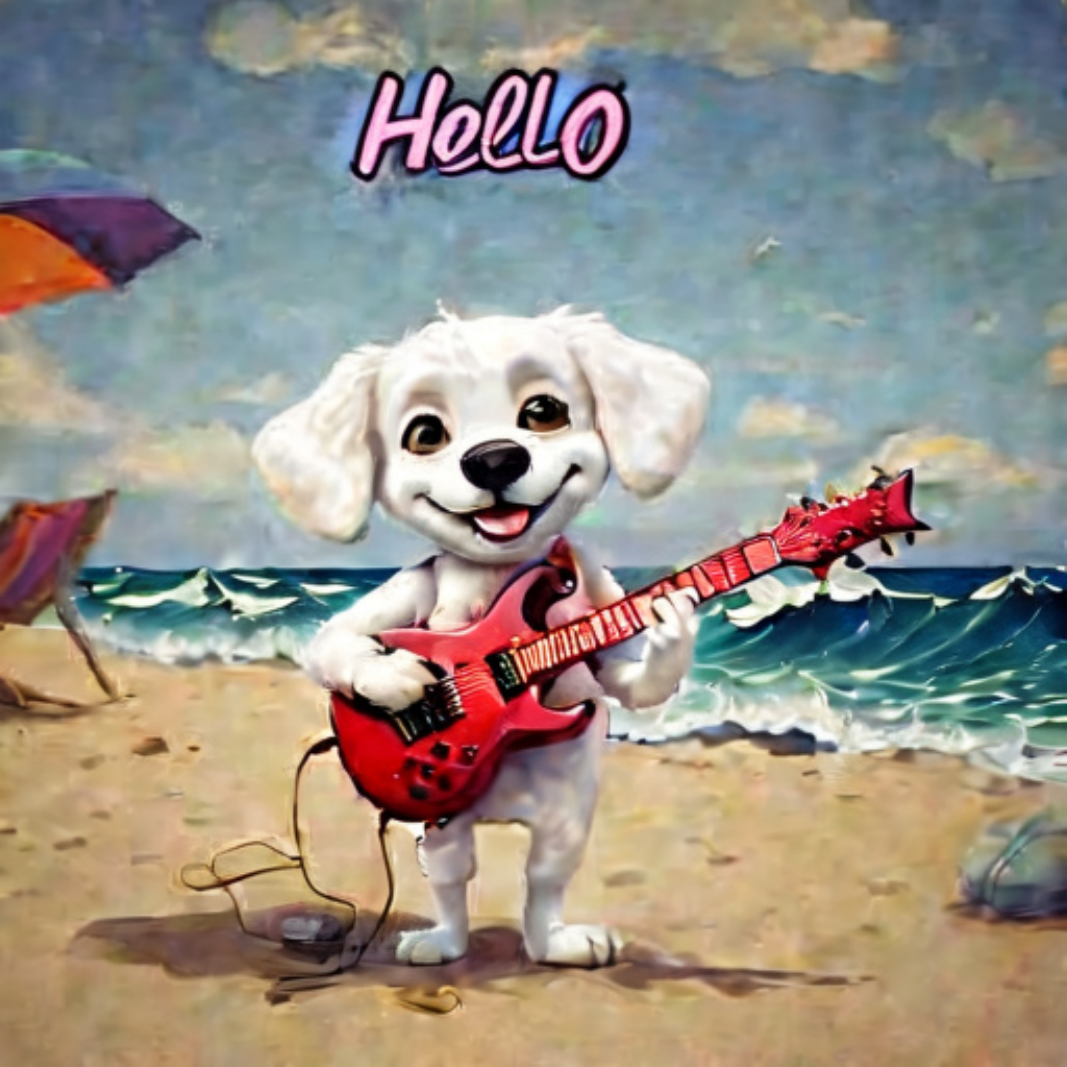}}&
\raisebox{-.5\height}{
\includegraphics[width=0.11\linewidth]{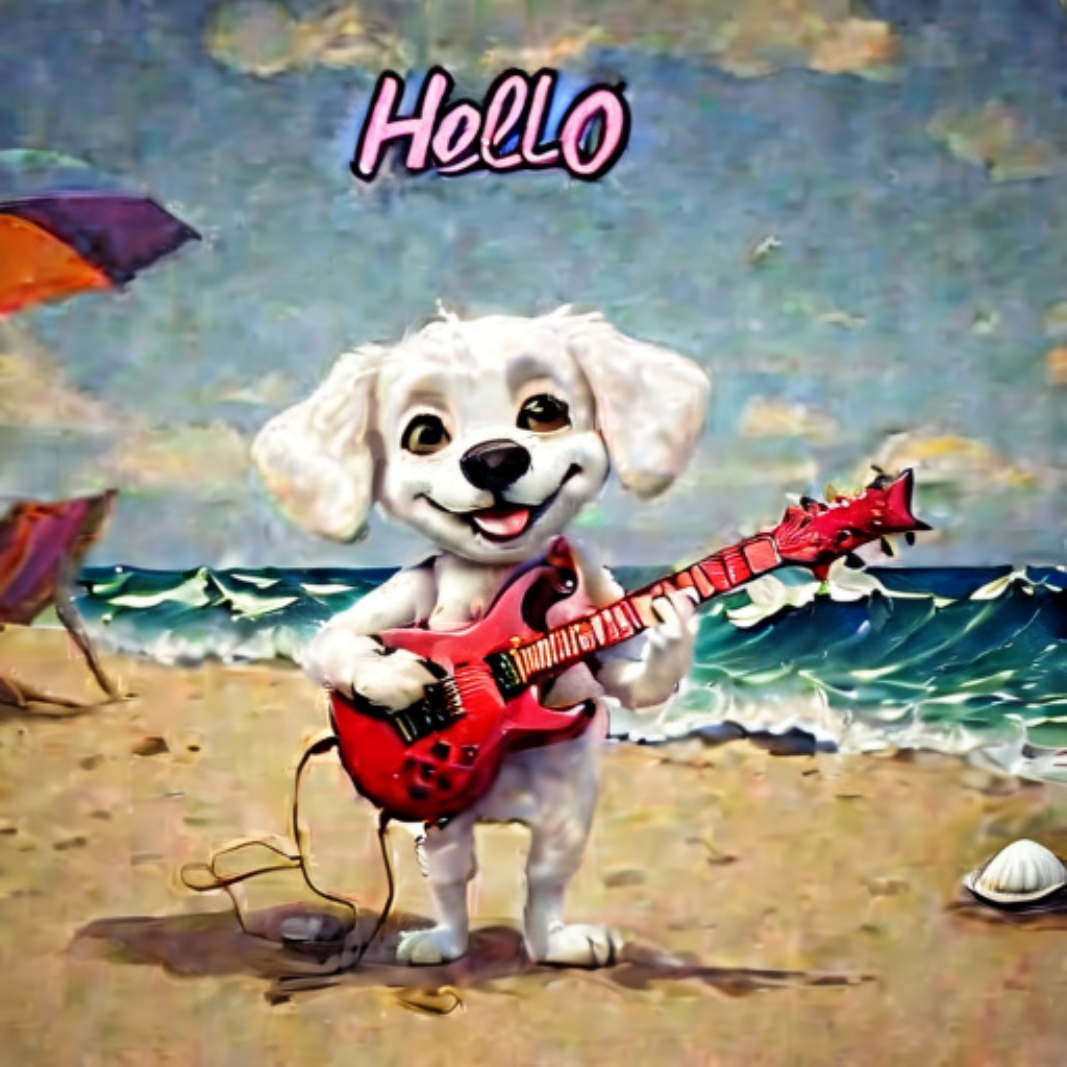}}&
\raisebox{-.5\height}{
\includegraphics[width=0.11\linewidth]{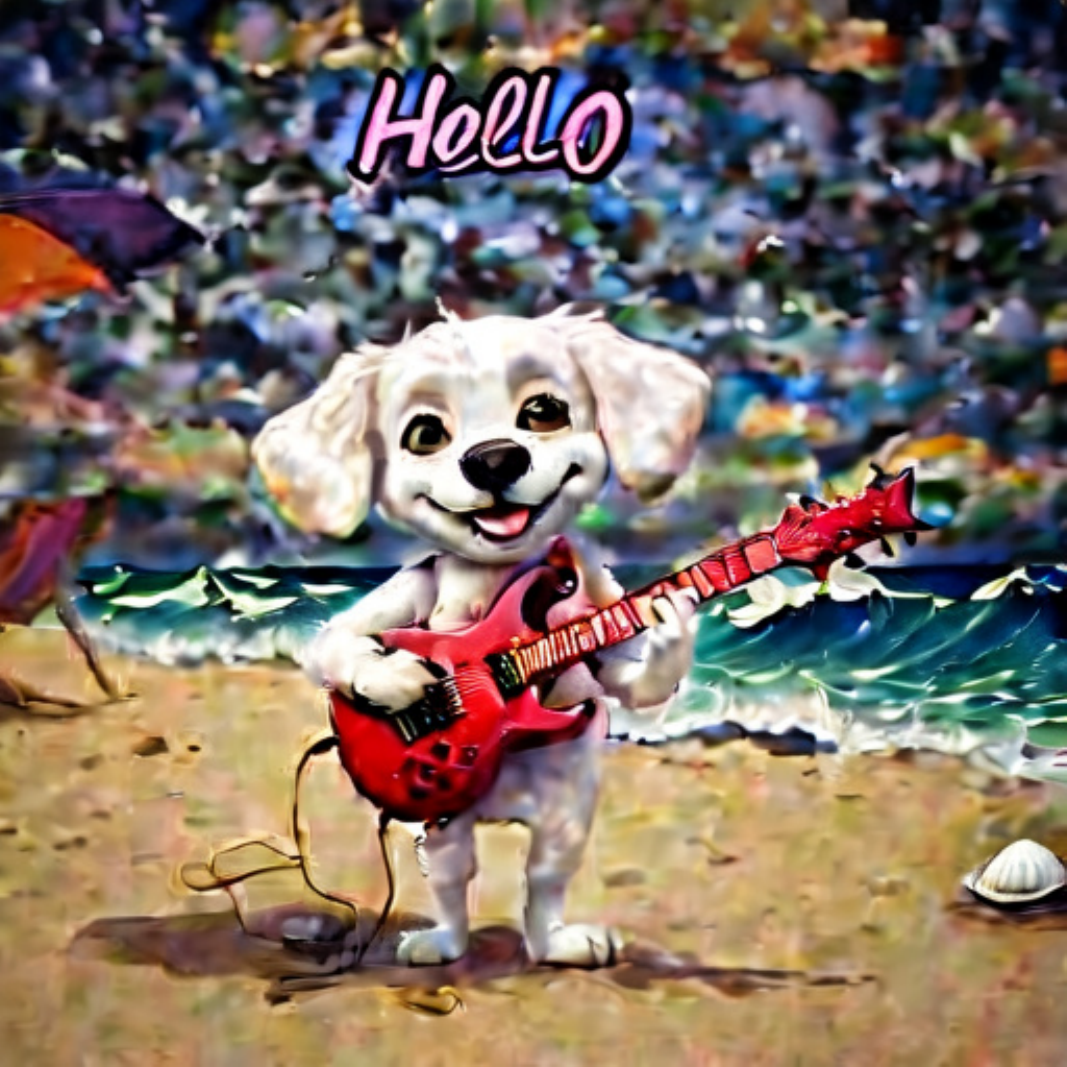}} \\ [13mm]

\resizebox{!}{8px}{
\begin{tabular}[x]{@{}c@{}} \footnotesize{$\alpha=0.01$} \end{tabular}}&
\raisebox{-.5\height}{
\includegraphics[width=0.11\linewidth]{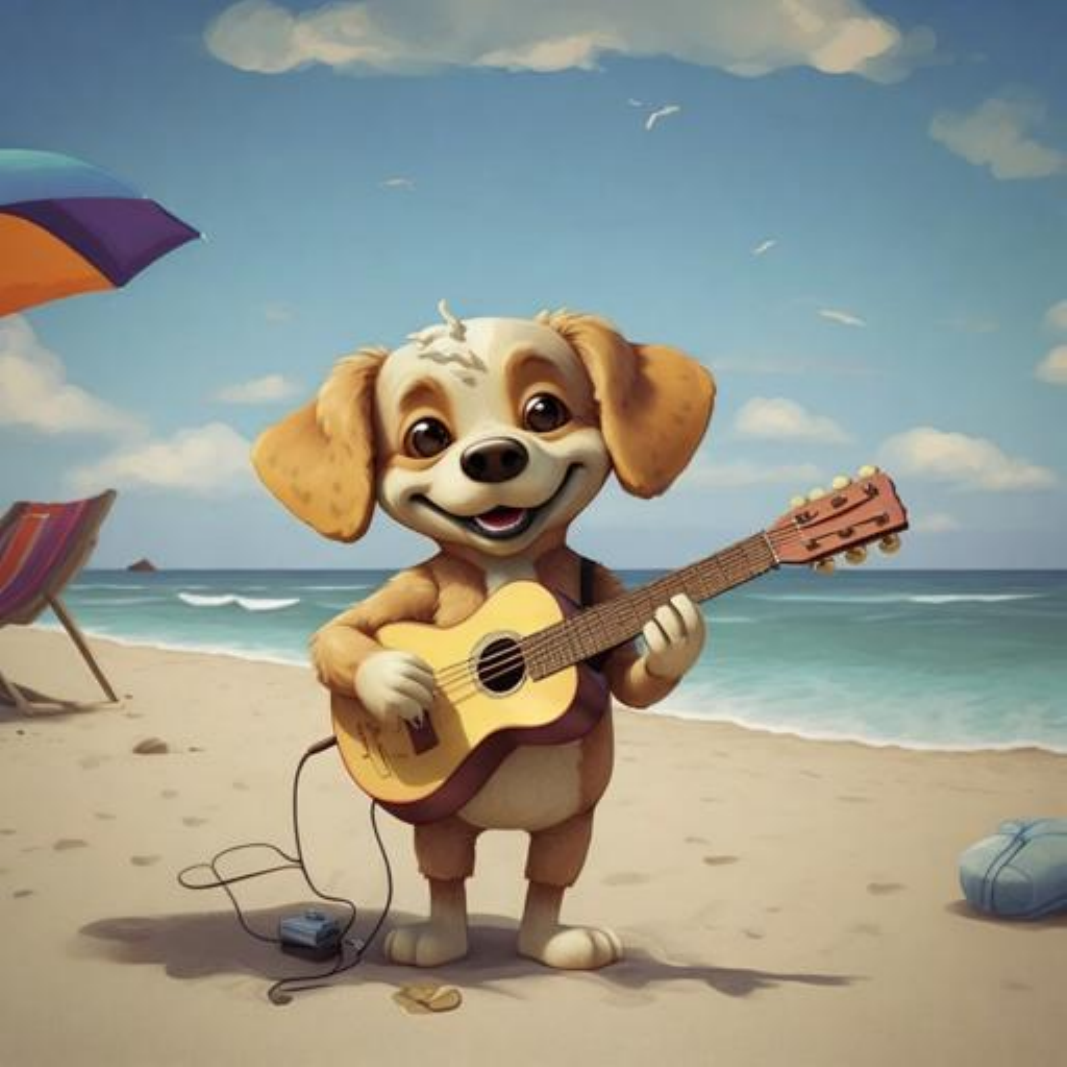}}&
\raisebox{-.5\height}{
\includegraphics[width=0.11\linewidth]{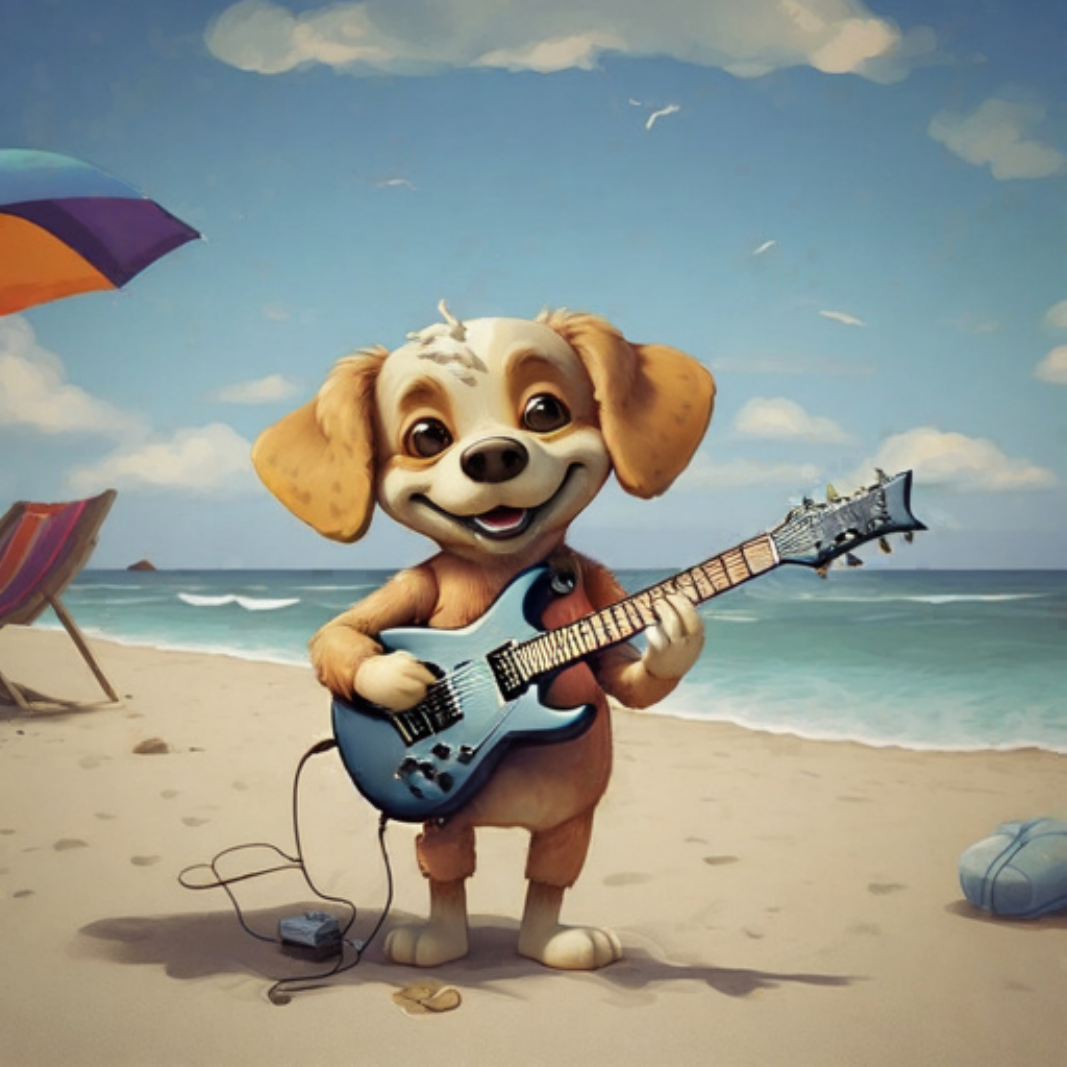}}&
\raisebox{-.5\height}{
\includegraphics[width=0.11\linewidth]{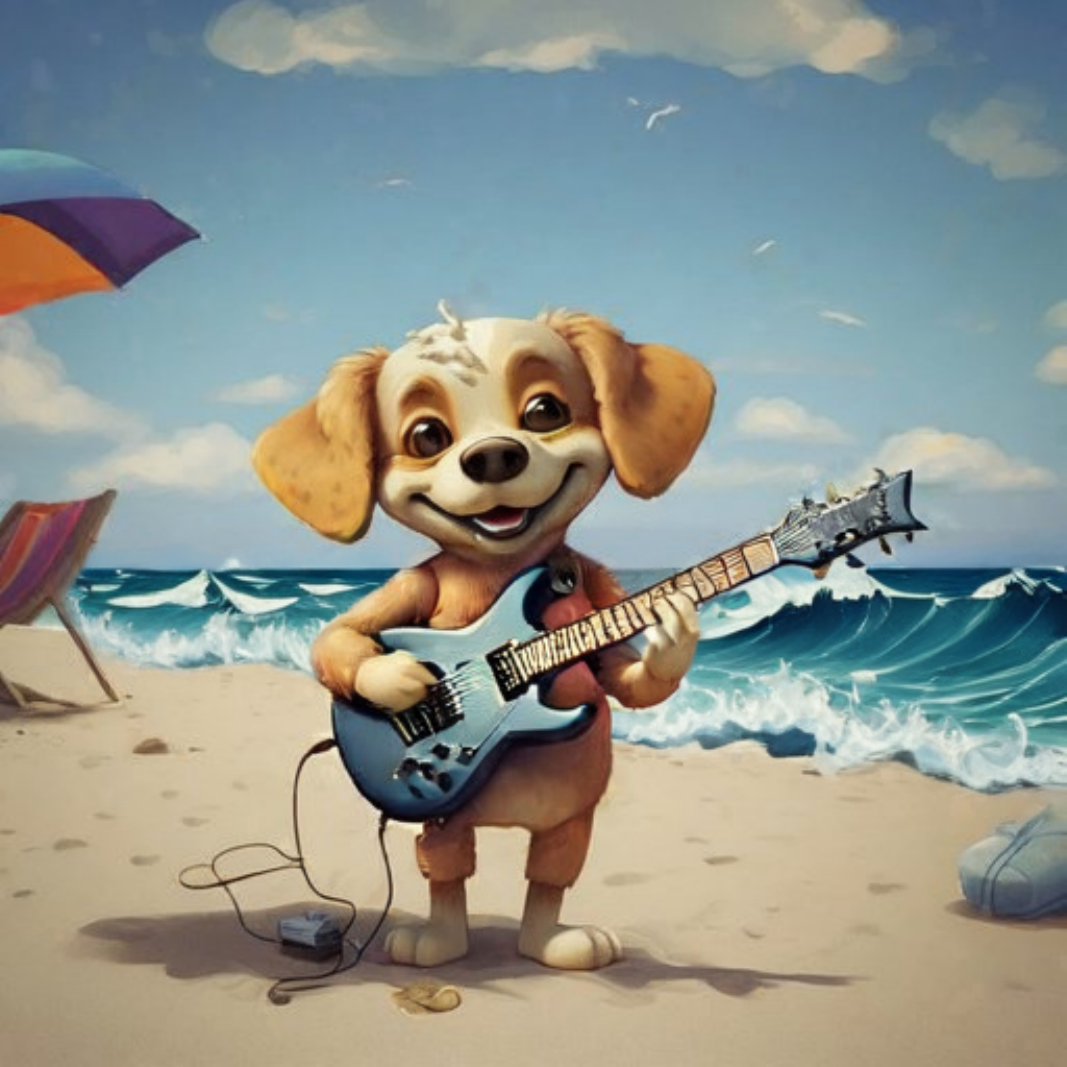}}&
\raisebox{-.5\height}{
\includegraphics[width=0.11\linewidth]{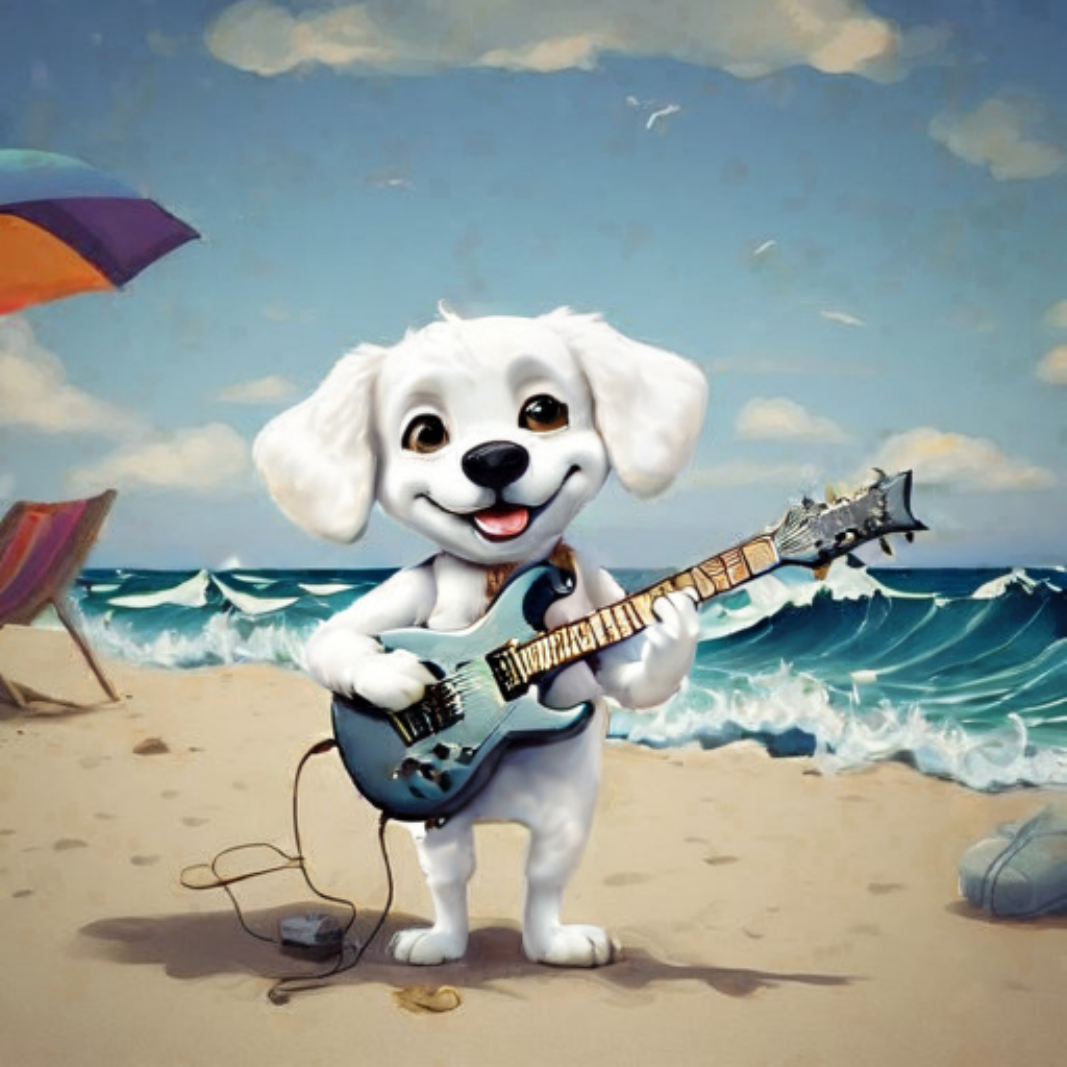}}&
\raisebox{-.5\height}{
\includegraphics[width=0.11\linewidth]{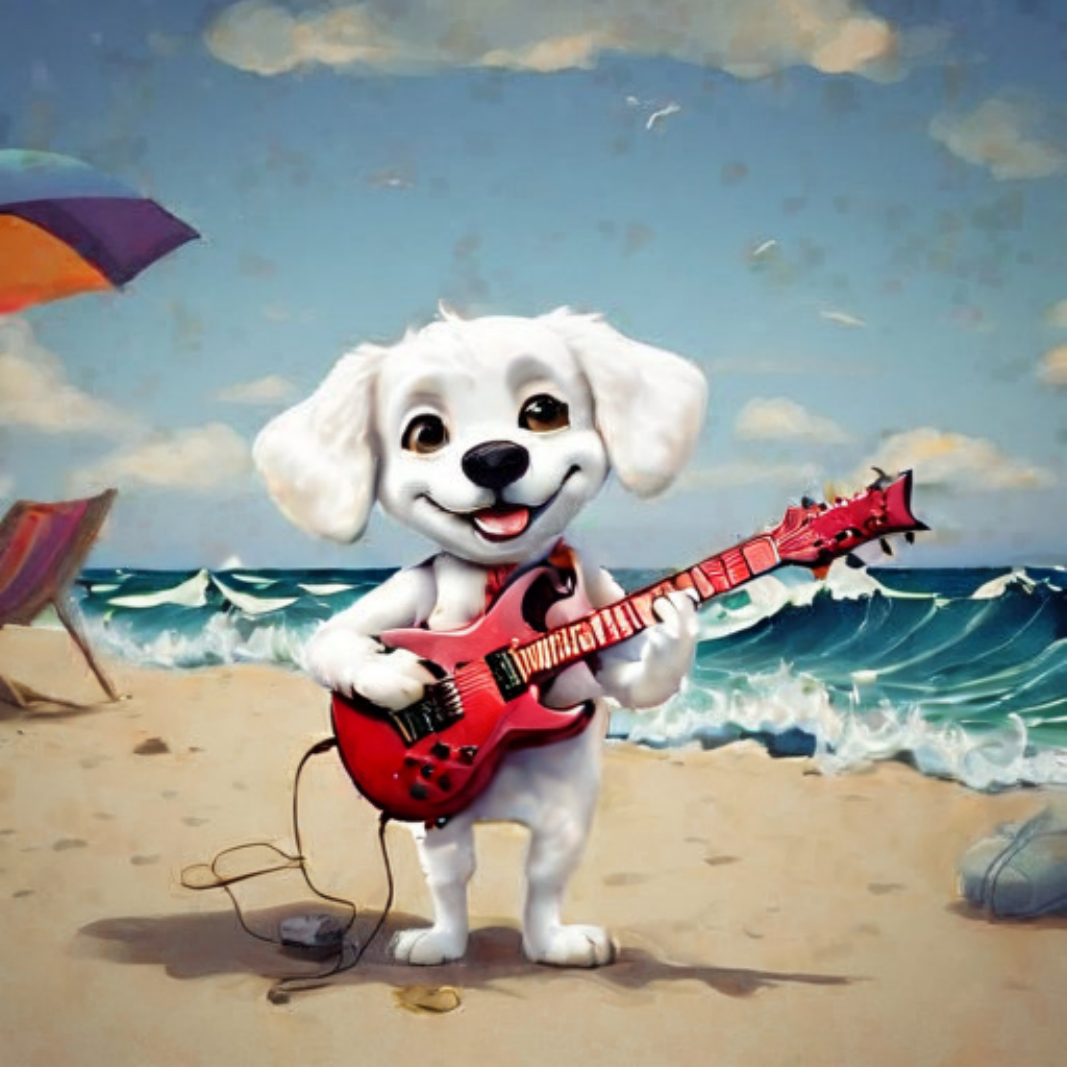}}&
\raisebox{-.5\height}{
\includegraphics[width=0.11\linewidth]{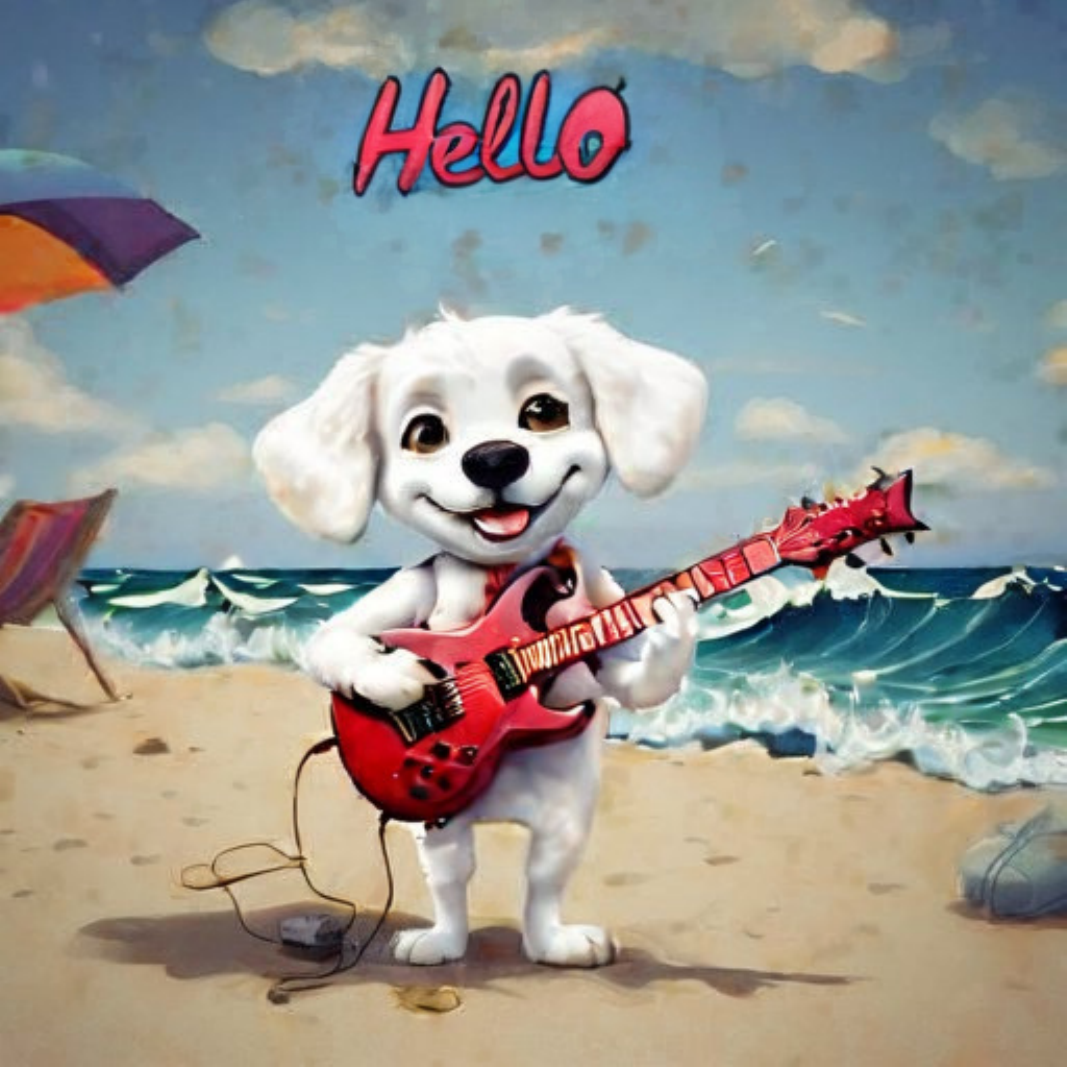}}&
\raisebox{-.5\height}{
\includegraphics[width=0.11\linewidth]{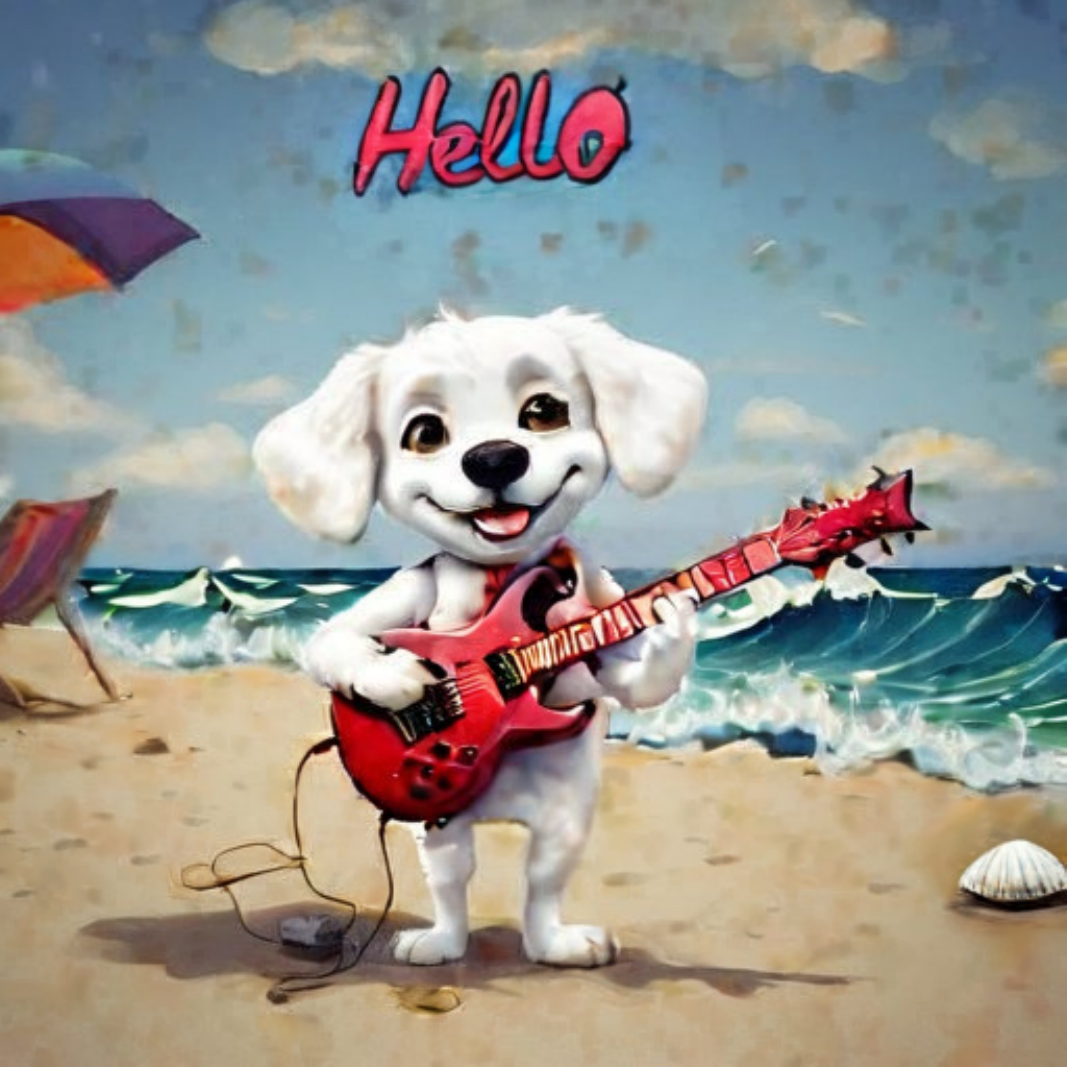}}&
\raisebox{-.5\height}{
\includegraphics[width=0.11\linewidth]{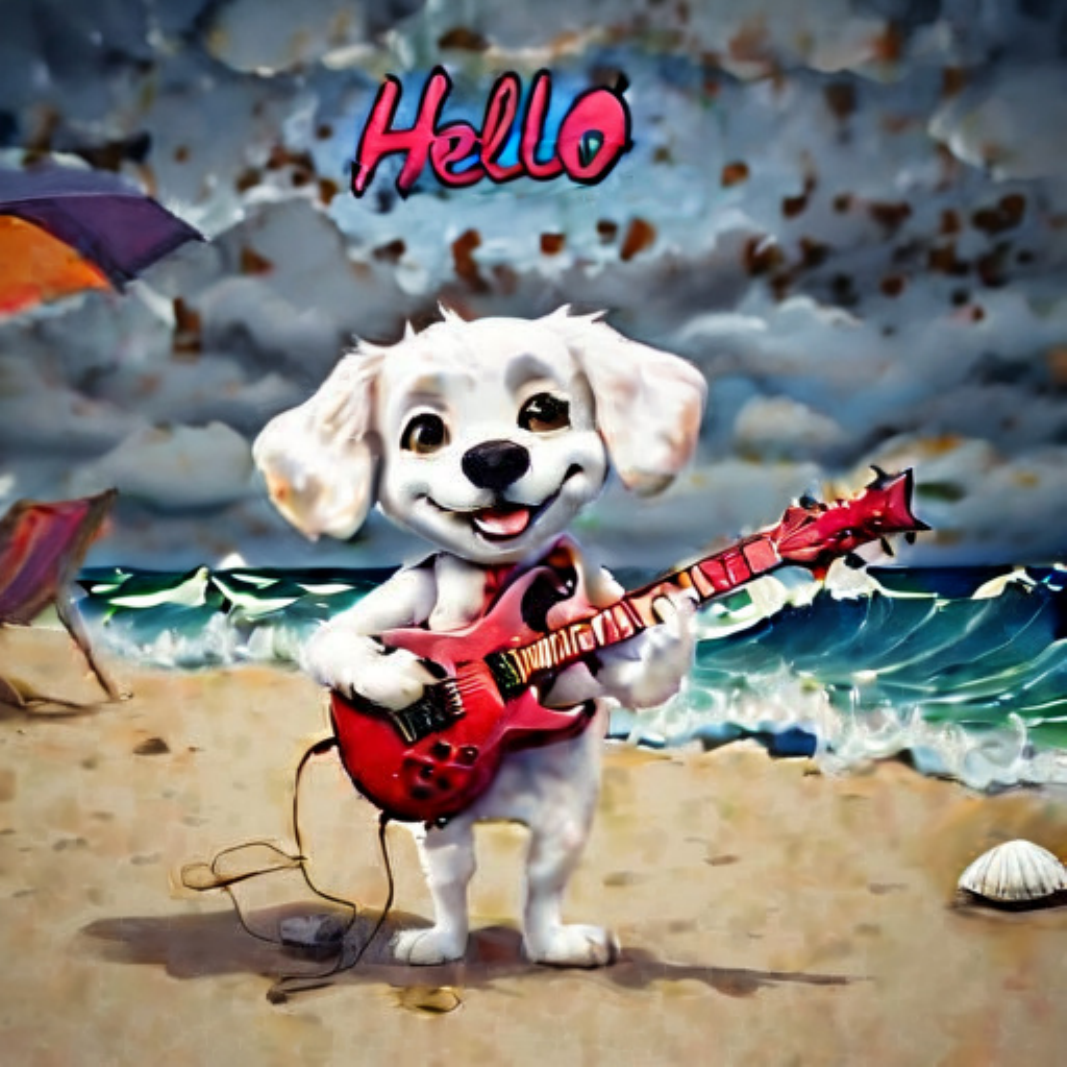}} \\ [13mm]

\resizebox{!}{8px}{
\begin{tabular}[x]{@{}c@{}} \footnotesize{$\alpha=0.025$} \end{tabular}}&
\raisebox{-.5\height}{
\includegraphics[width=0.11\linewidth]{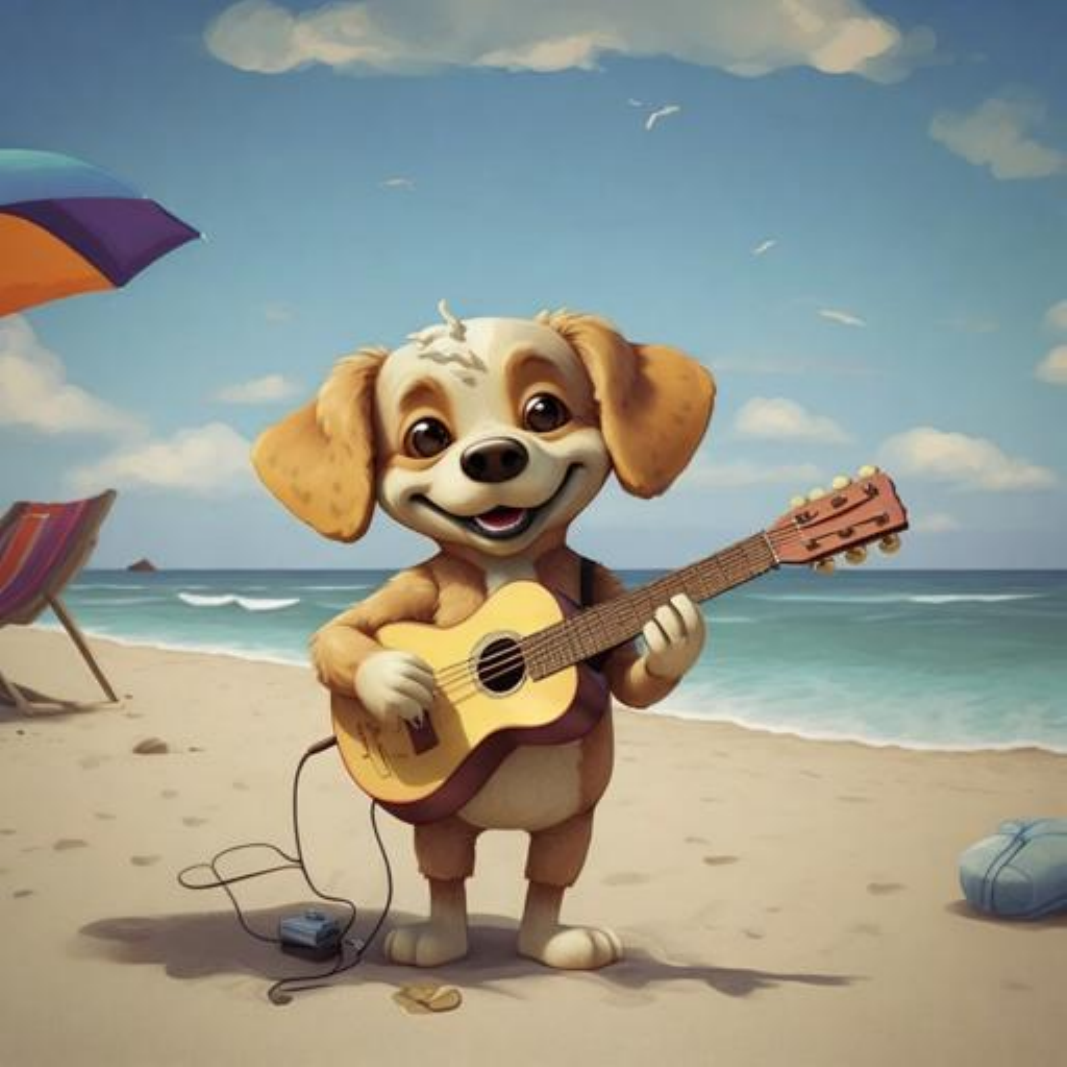}}&
\raisebox{-.5\height}{
\includegraphics[width=0.11\linewidth]{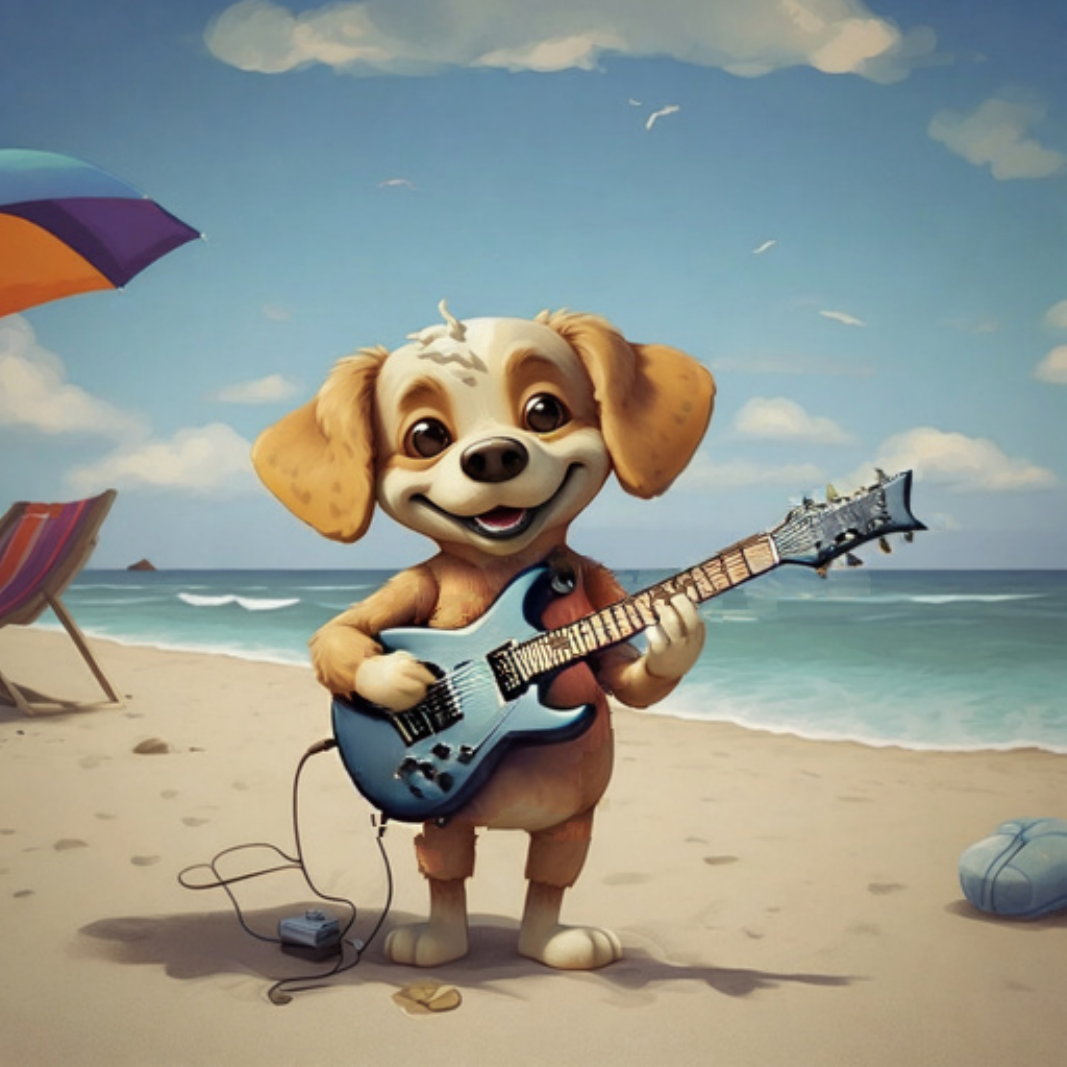}}&
\raisebox{-.5\height}{
\includegraphics[width=0.11\linewidth]{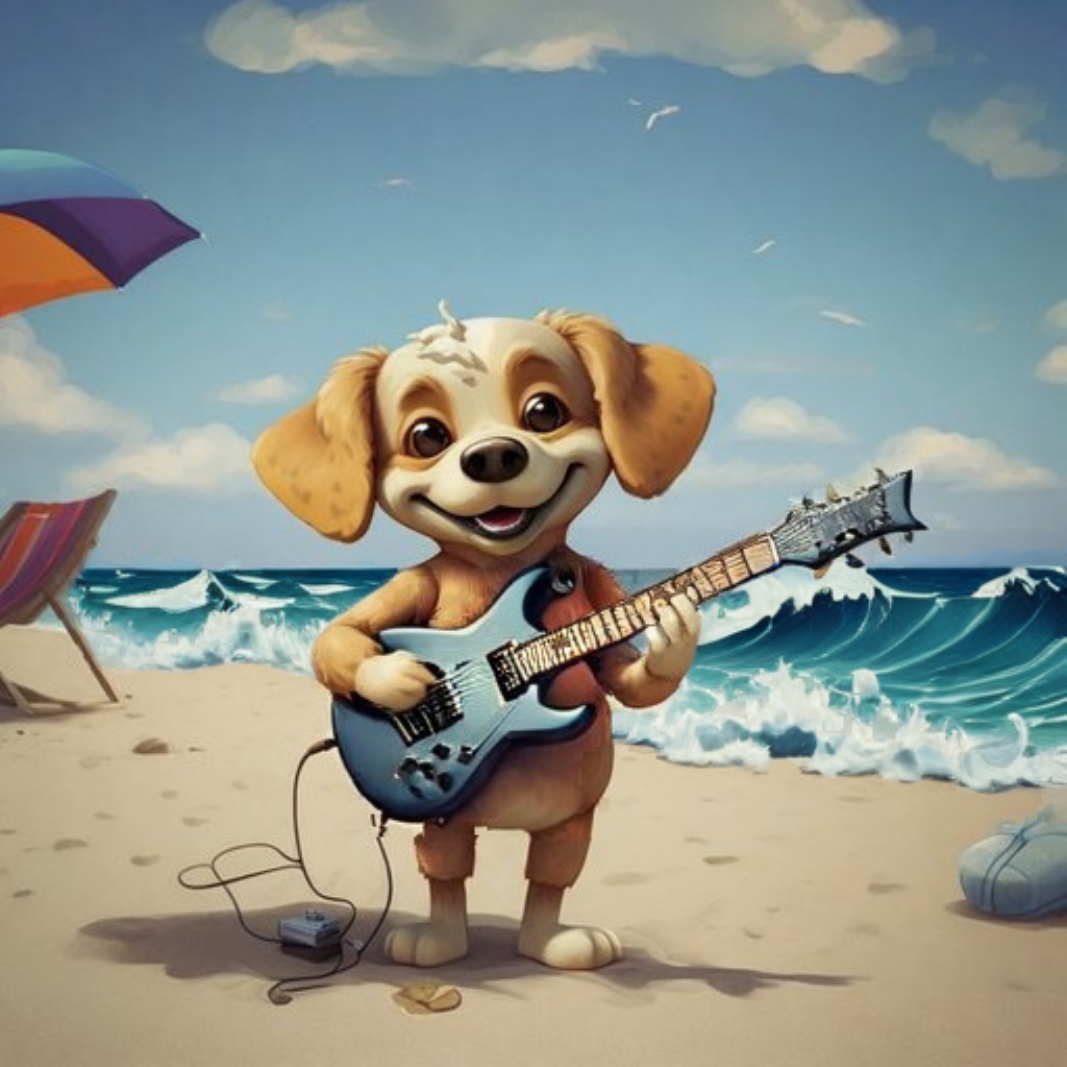}}&
\raisebox{-.5\height}{
\includegraphics[width=0.11\linewidth]{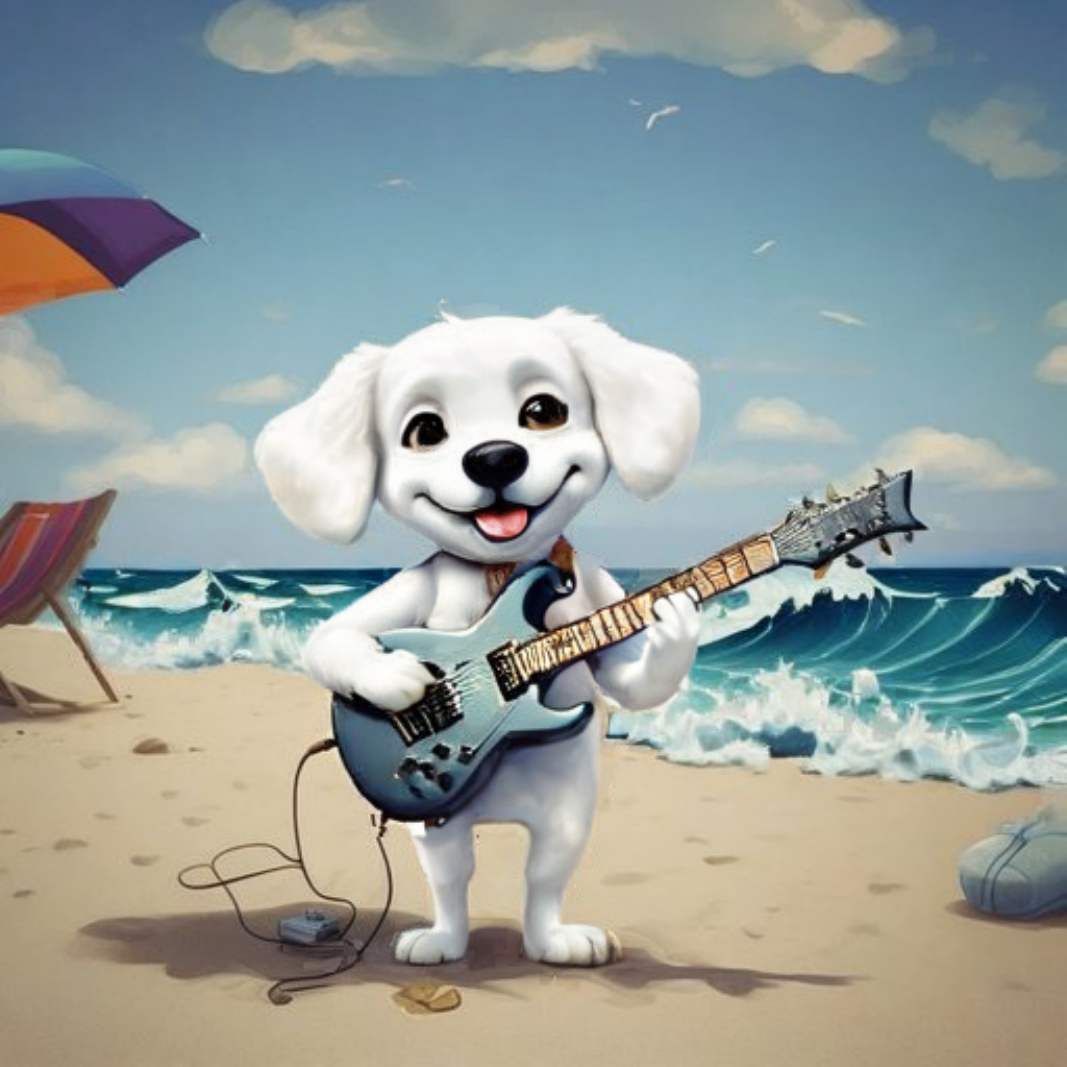}}&
\raisebox{-.5\height}{
\includegraphics[width=0.11\linewidth]{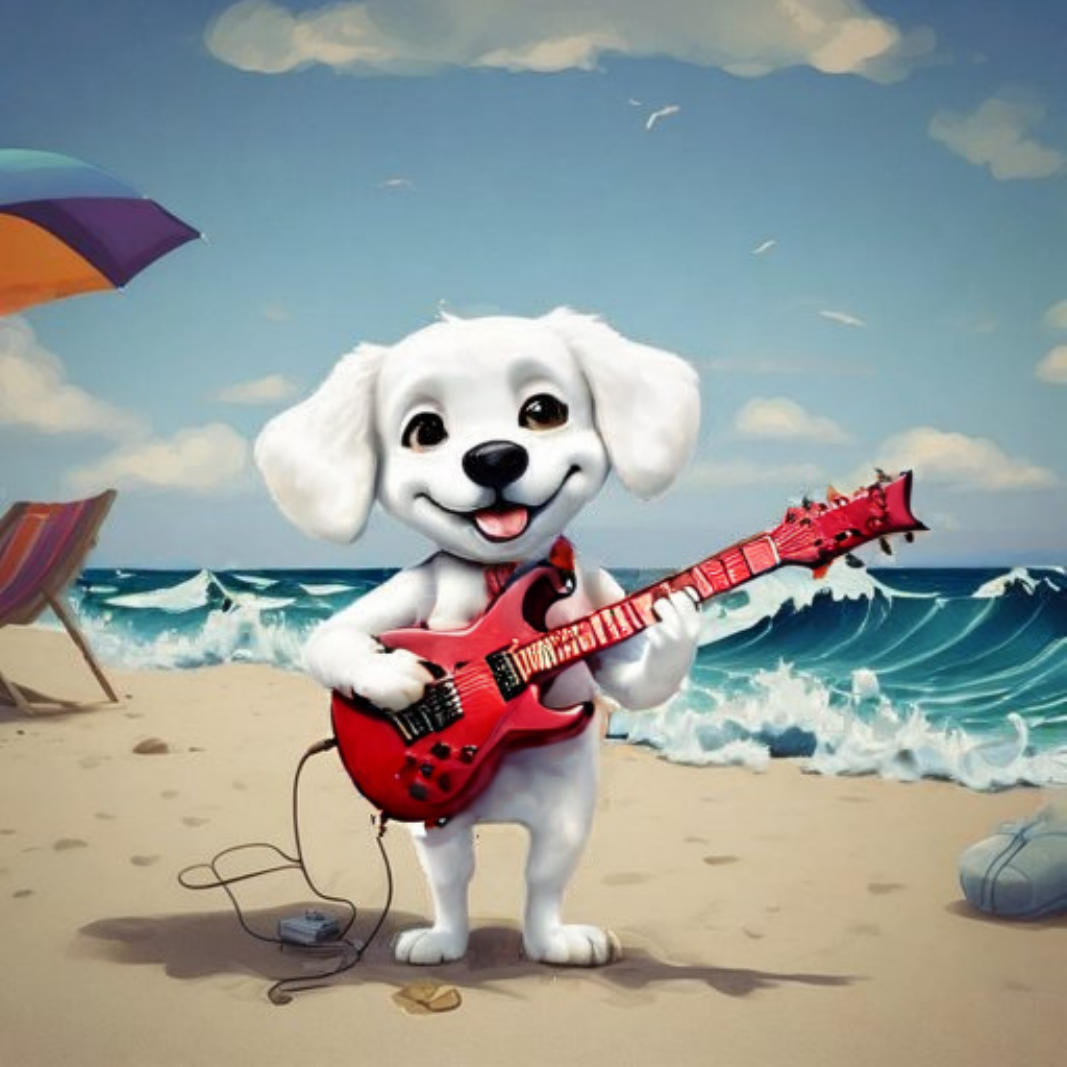}}&
\raisebox{-.5\height}{
\includegraphics[width=0.11\linewidth]{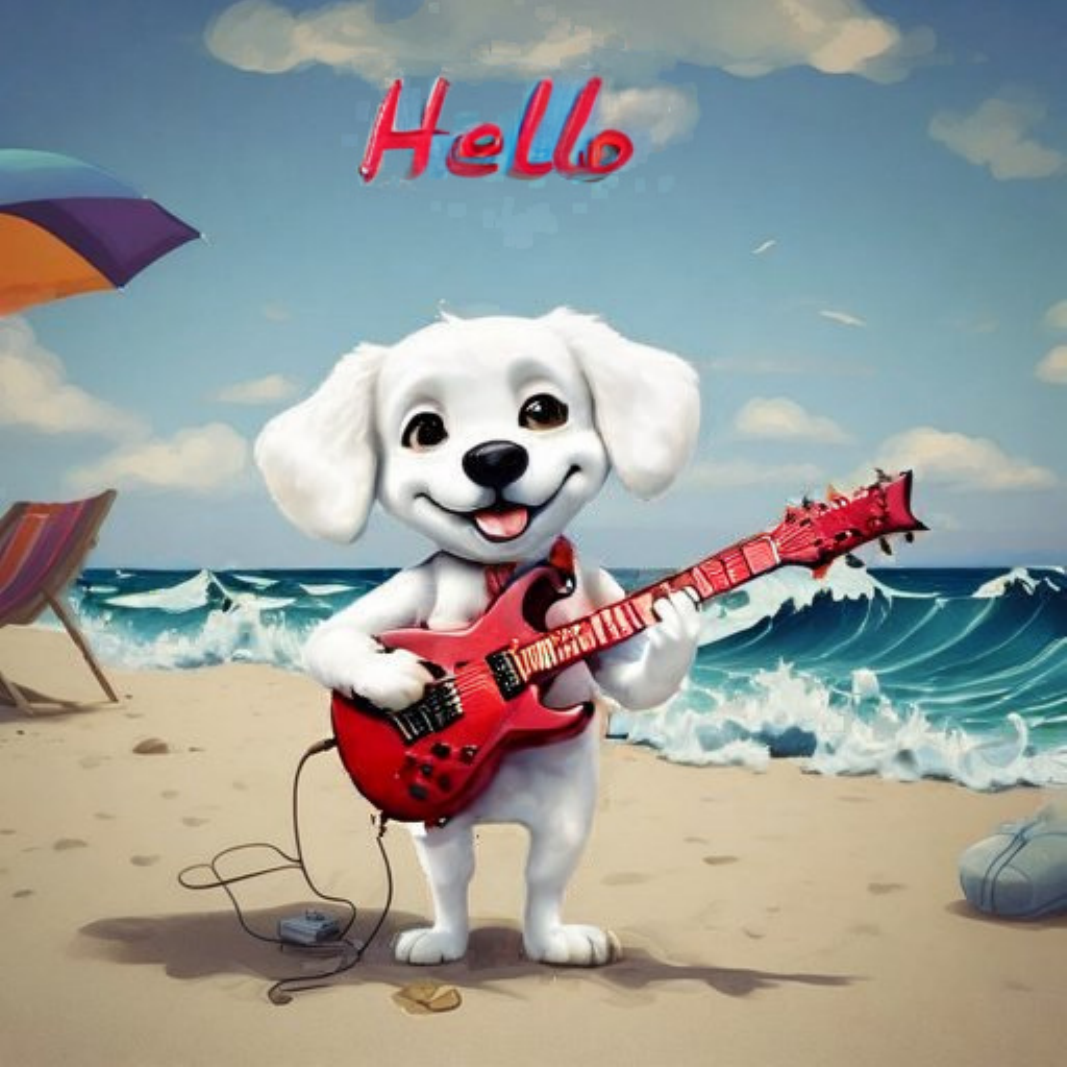}}&
\raisebox{-.5\height}{
\includegraphics[width=0.11\linewidth]{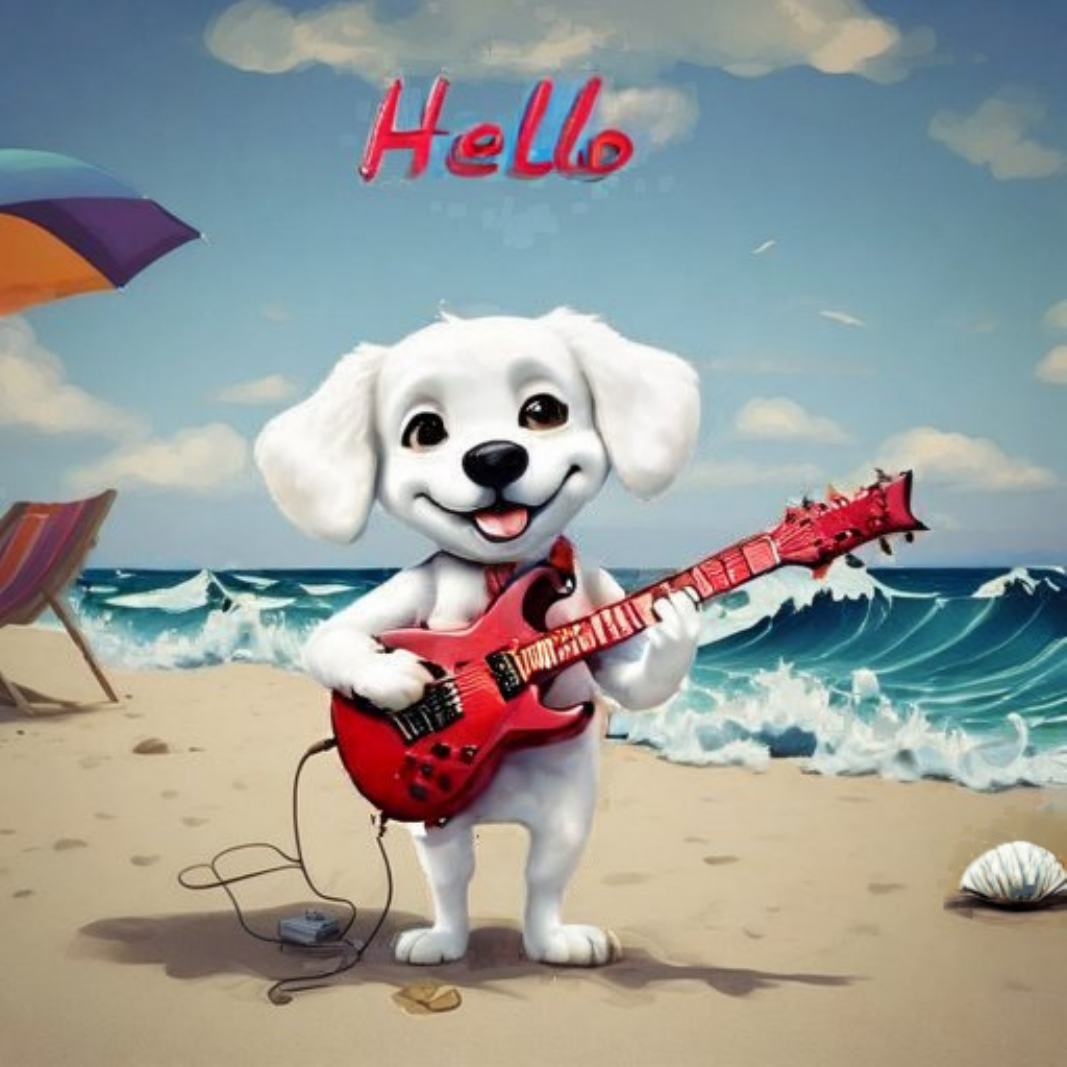}}&
\raisebox{-.5\height}{
\includegraphics[width=0.11\linewidth]{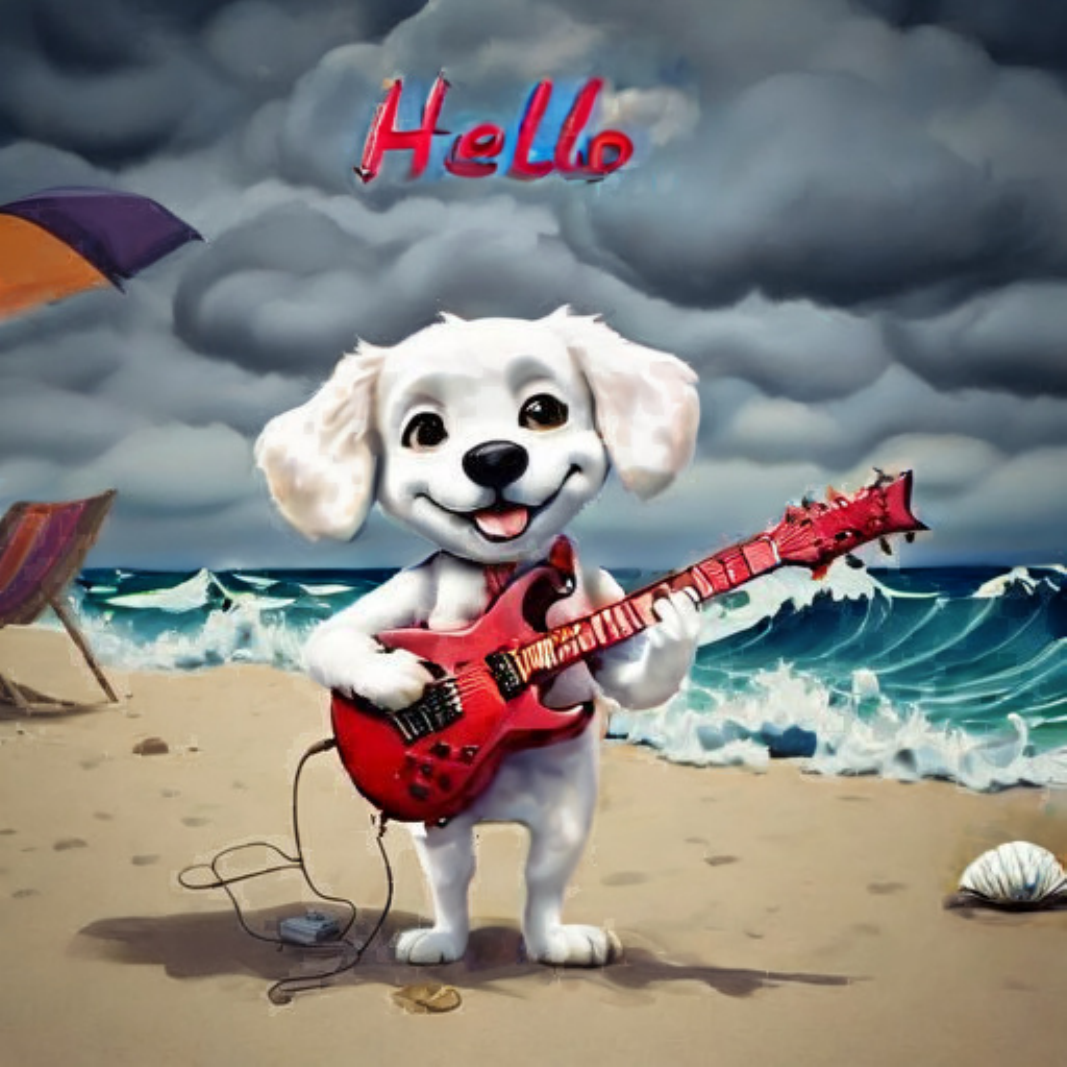}} \\ [13mm]

\resizebox{!}{8px}{
\begin{tabular}[x]{@{}c@{}} \footnotesize{$\alpha=0.05$} \end{tabular}}&
\raisebox{-.5\height}{
\includegraphics[width=0.11\linewidth]{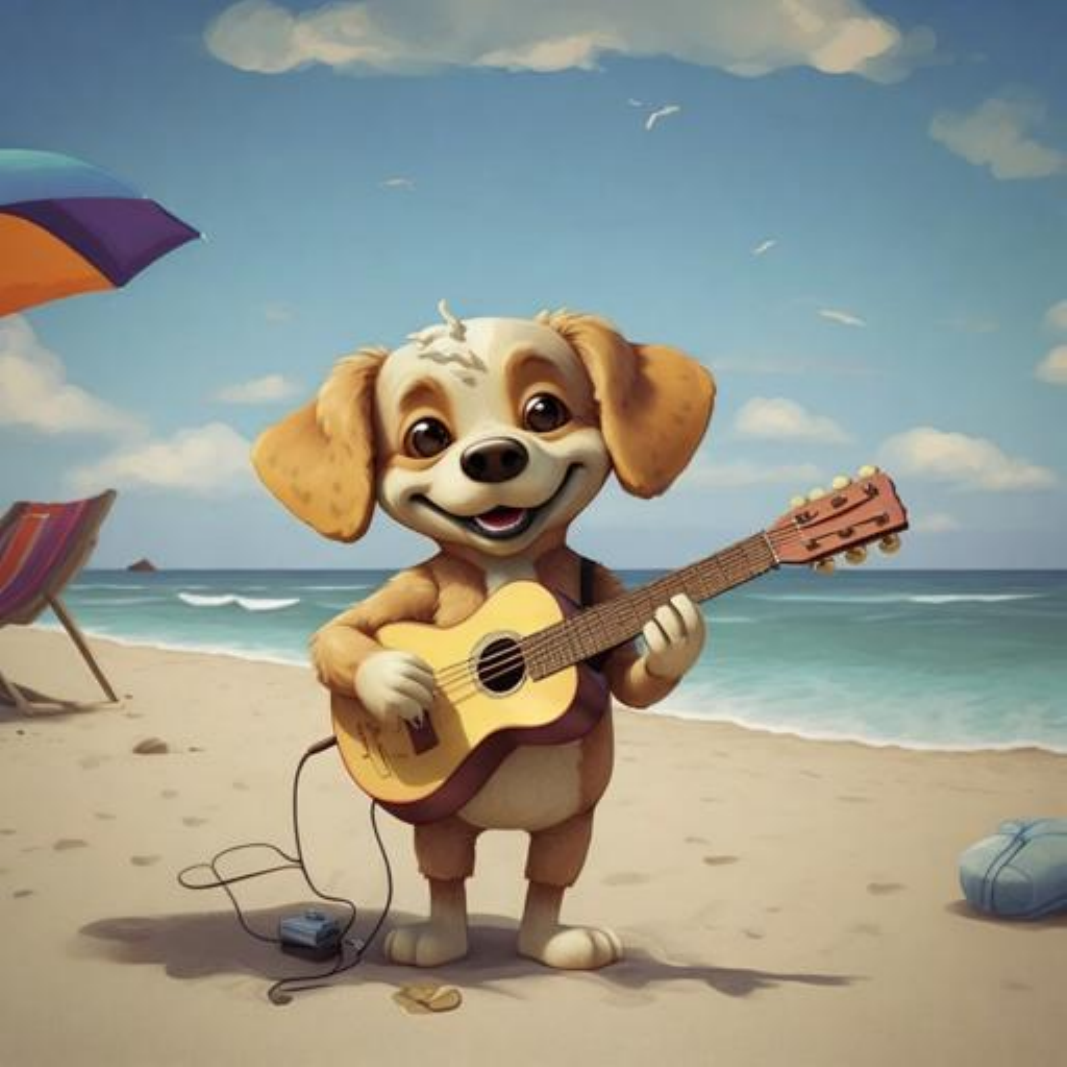}}&
\raisebox{-.5\height}{
\includegraphics[width=0.11\linewidth]{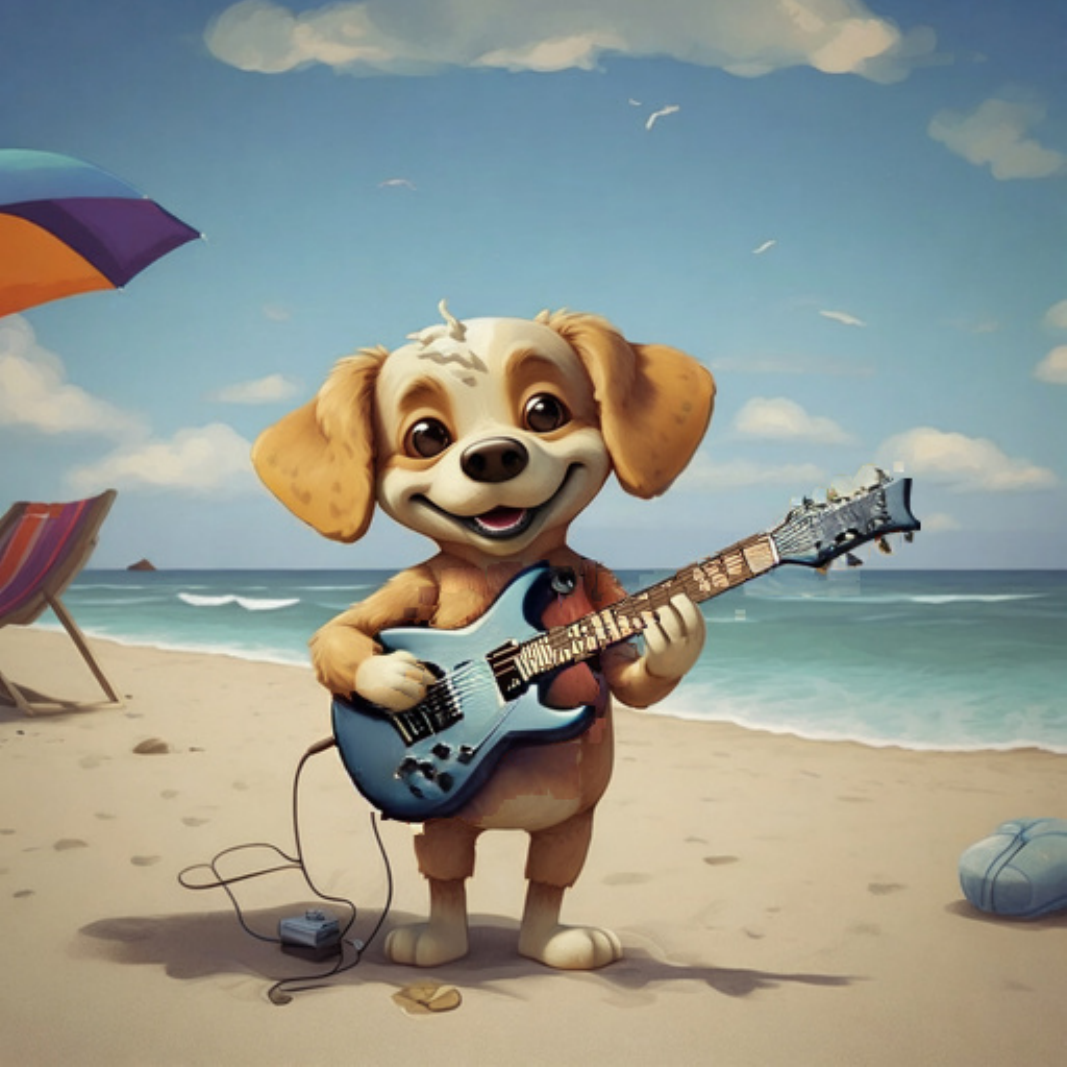}}&
\raisebox{-.5\height}{
\includegraphics[width=0.11\linewidth]{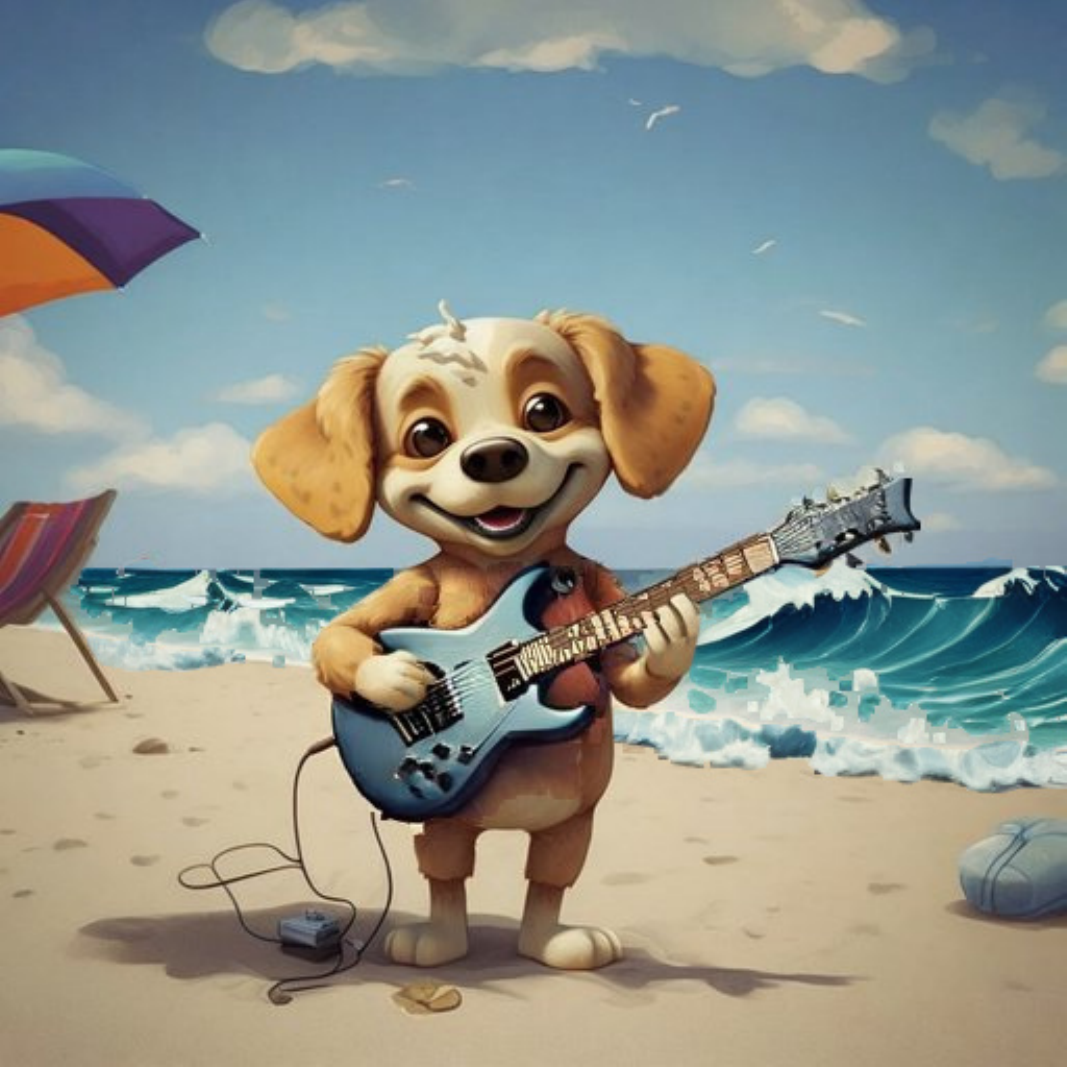}}&
\raisebox{-.5\height}{
\includegraphics[width=0.11\linewidth]{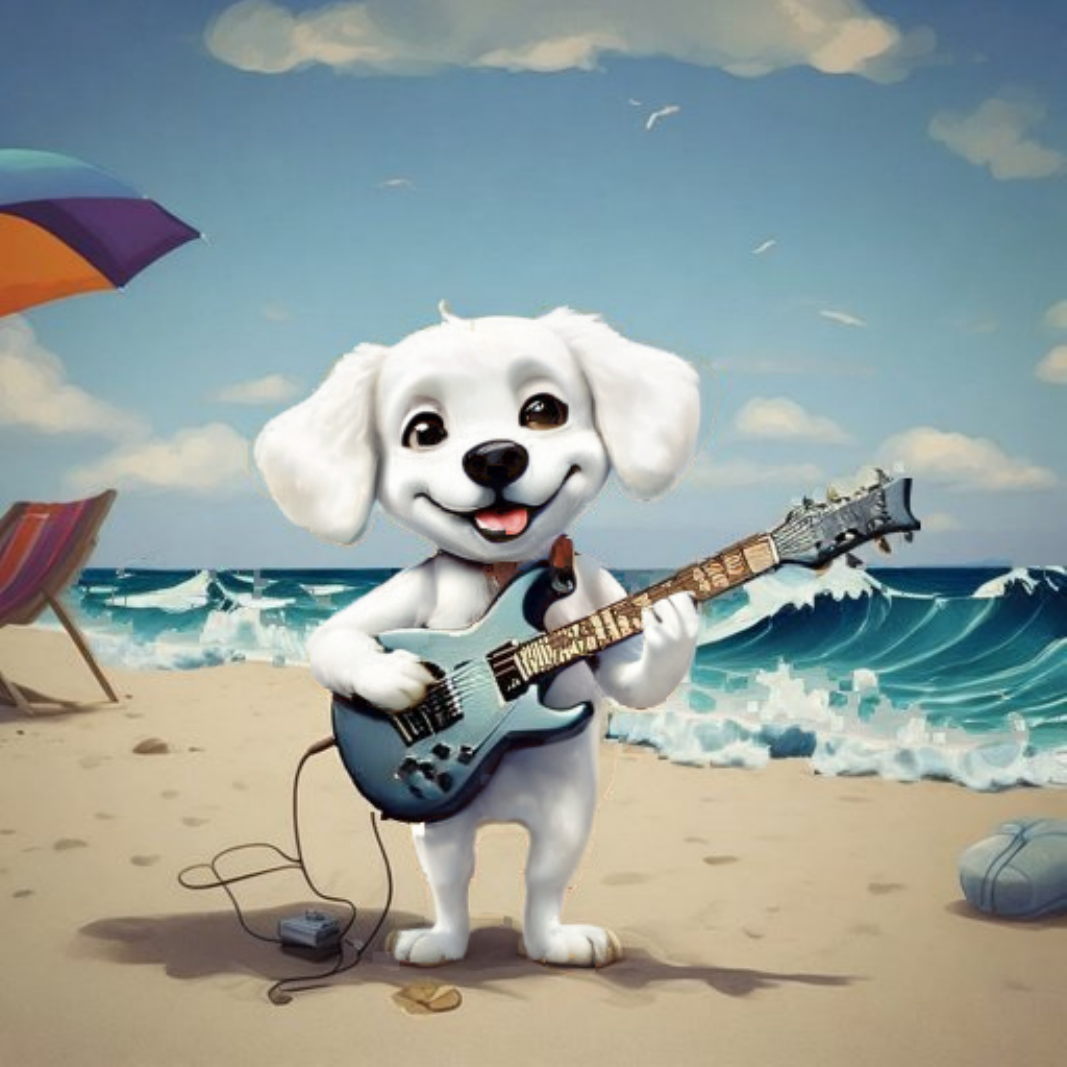}}&
\raisebox{-.5\height}{
\includegraphics[width=0.11\linewidth]{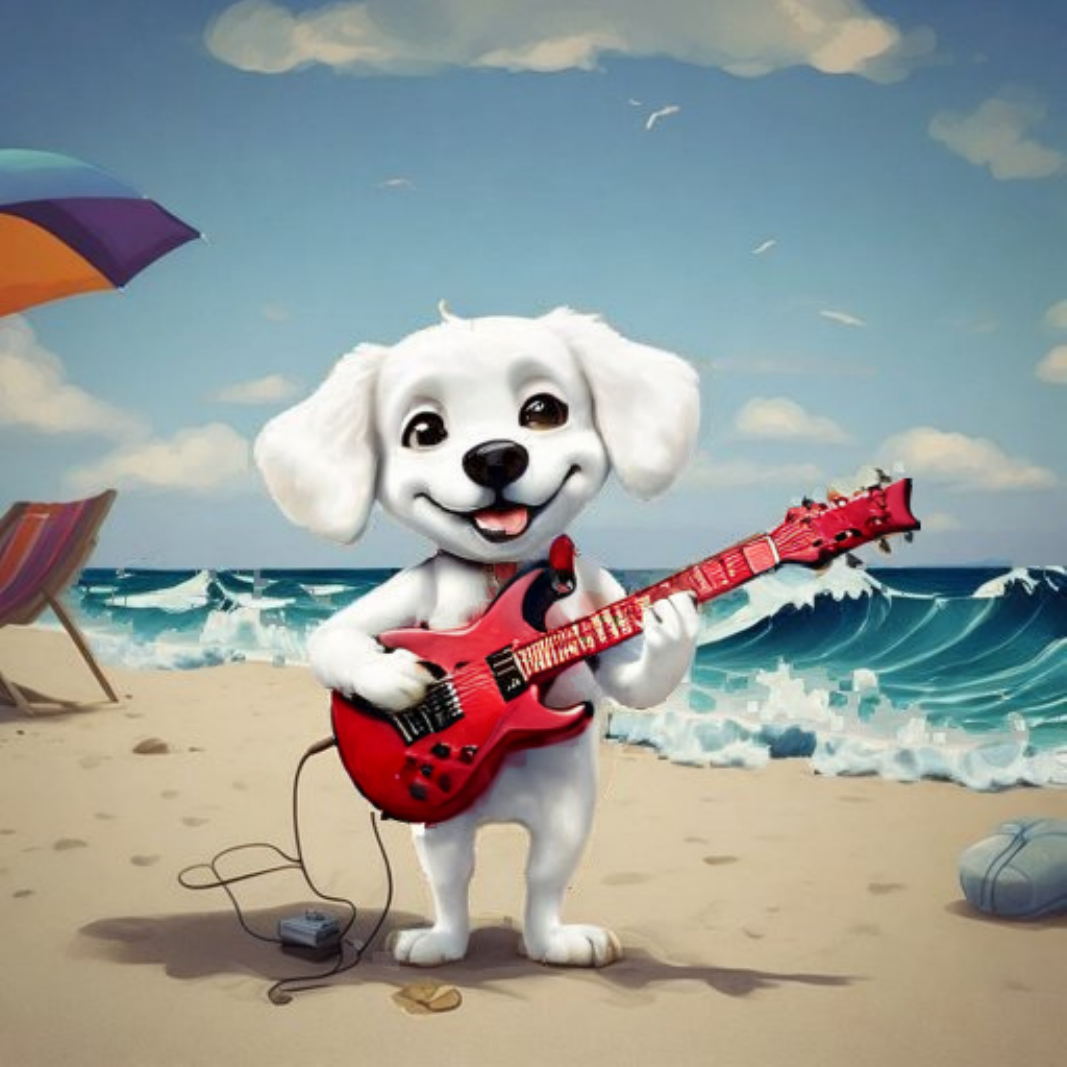}}&
\raisebox{-.5\height}{
\includegraphics[width=0.11\linewidth]{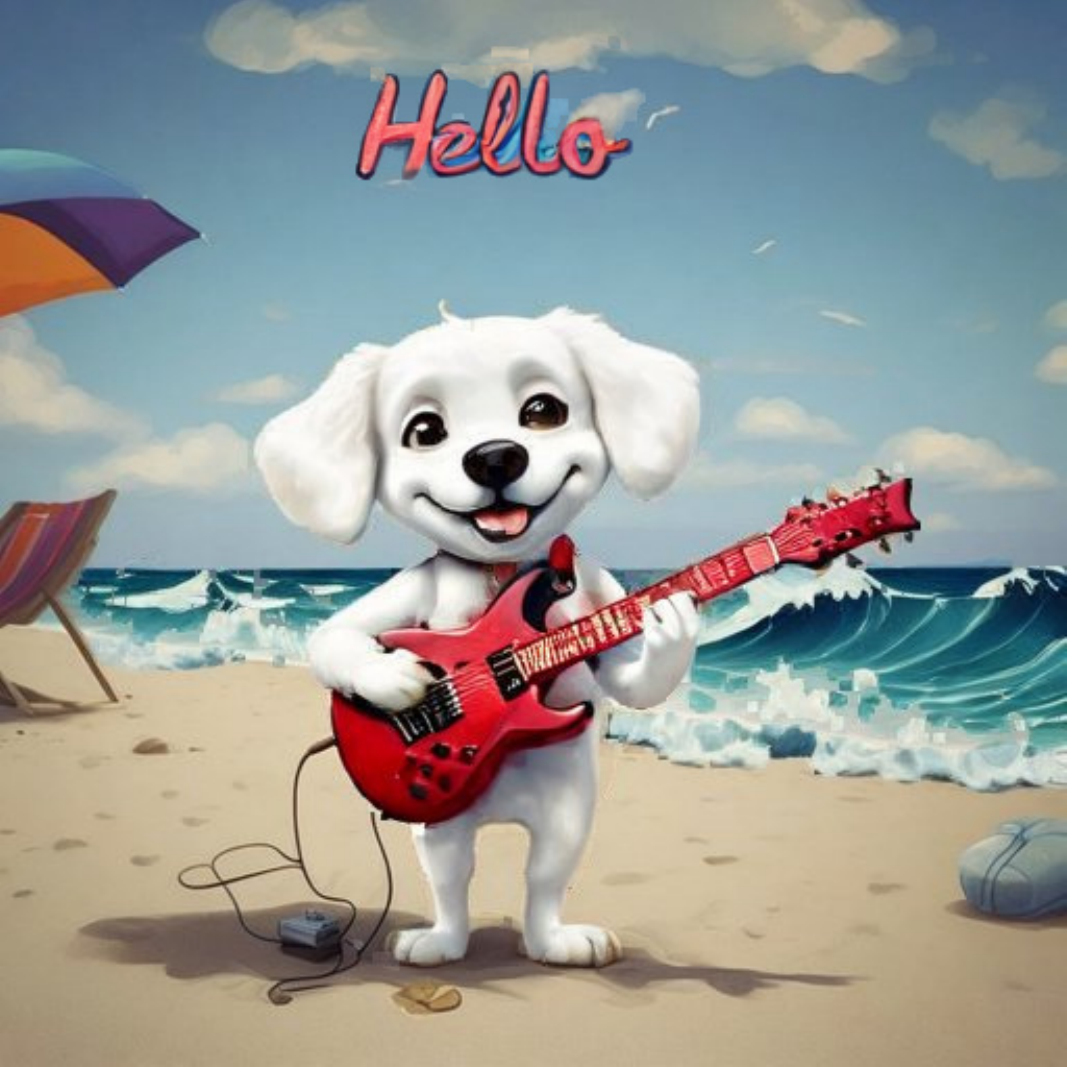}}&
\raisebox{-.5\height}{
\includegraphics[width=0.11\linewidth]{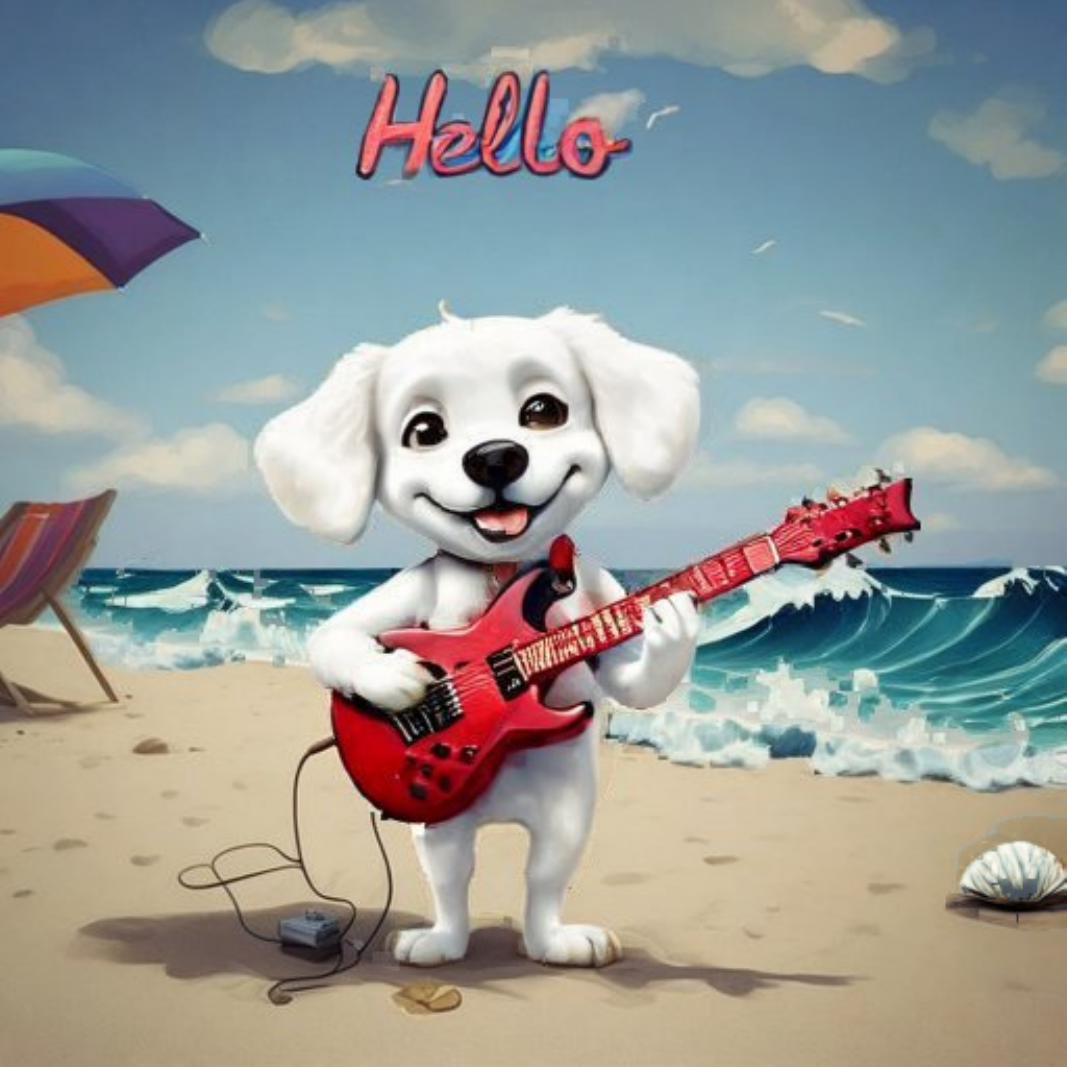}}&
\raisebox{-.5\height}{
\includegraphics[width=0.11\linewidth]{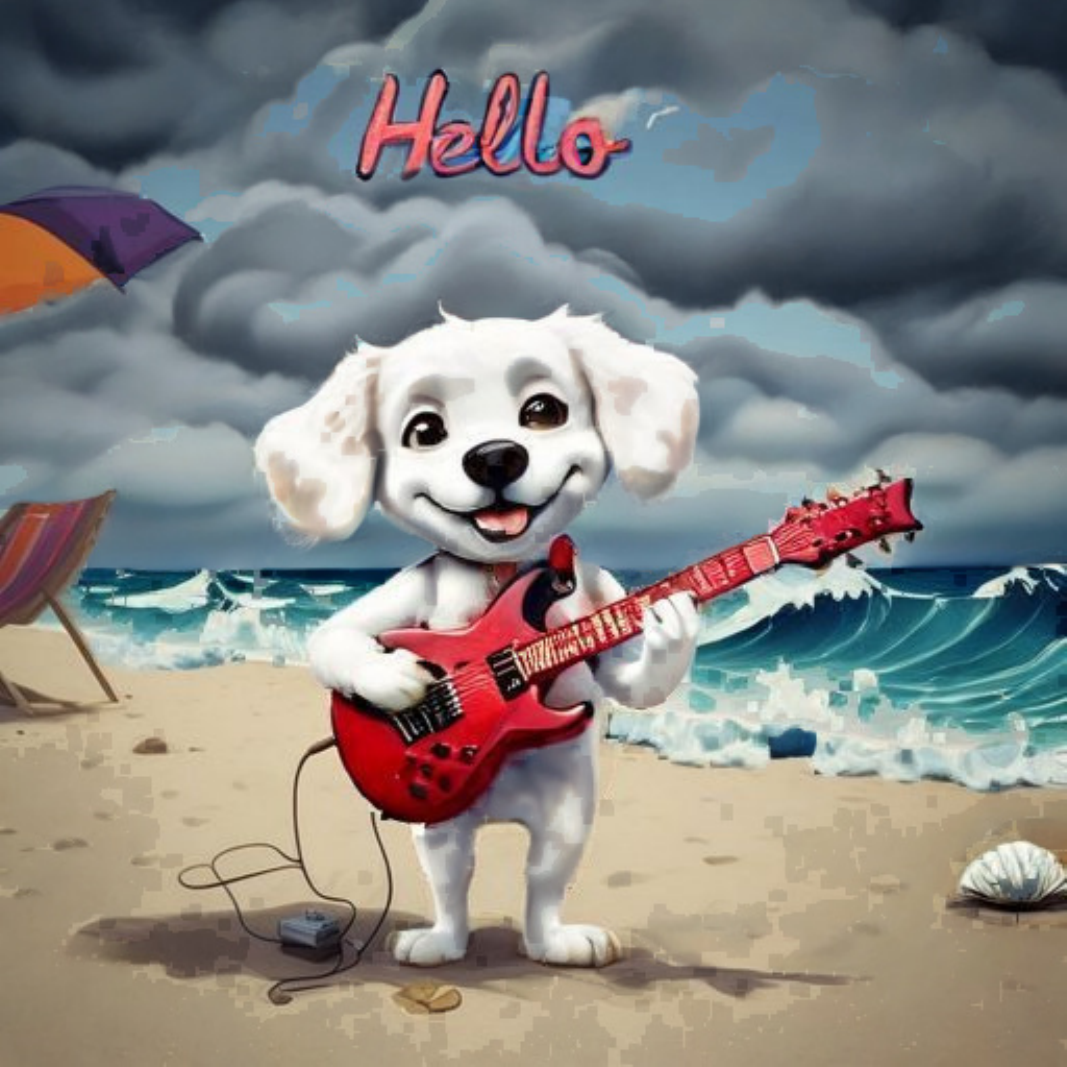}} \\ [13mm]

\resizebox{!}{8px}{
\begin{tabular}[x]{@{}c@{}} \footnotesize{$\alpha=0.1$} \end{tabular}}&
\raisebox{-.5\height}{
\includegraphics[width=0.11\linewidth]{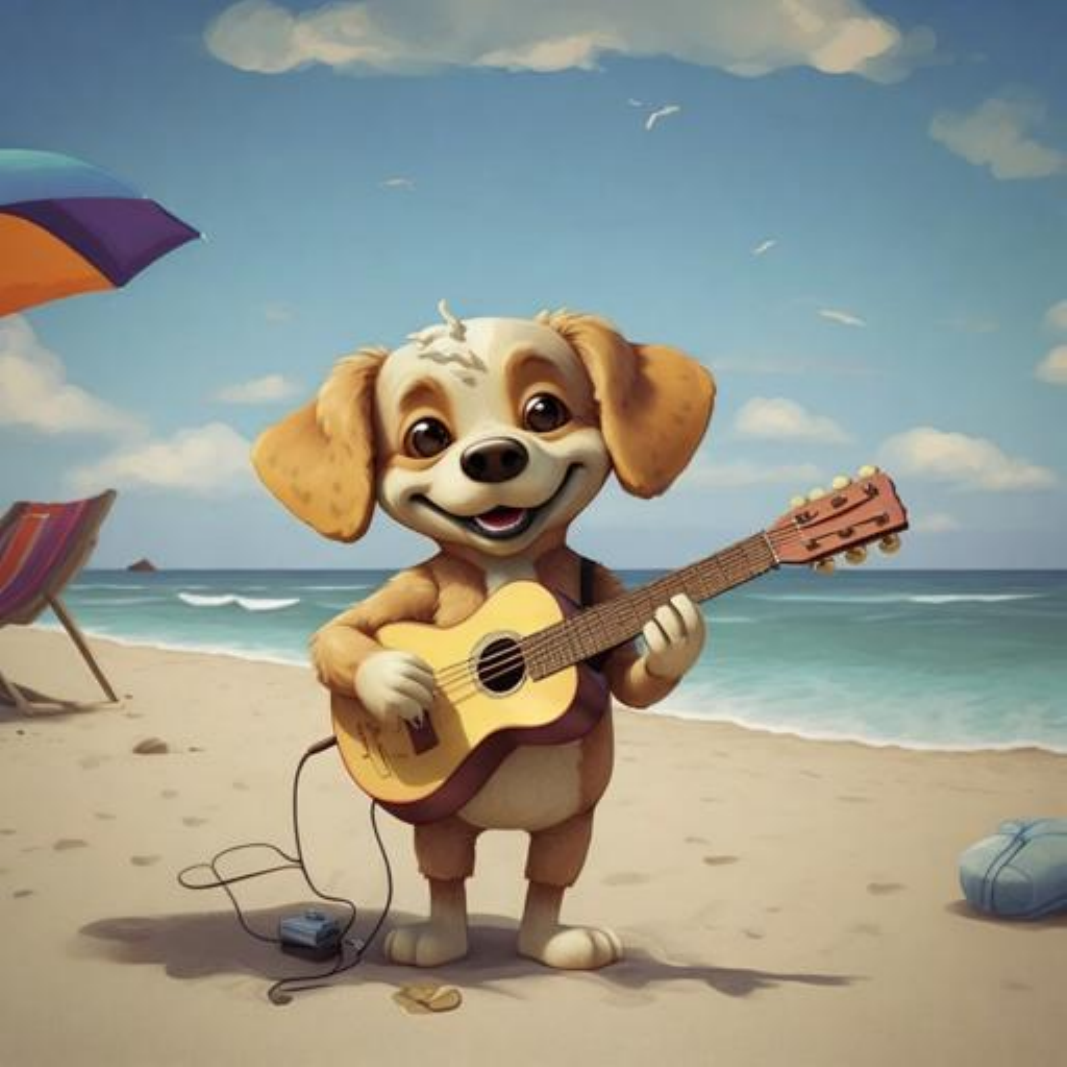}}&
\raisebox{-.5\height}{
\includegraphics[width=0.11\linewidth]{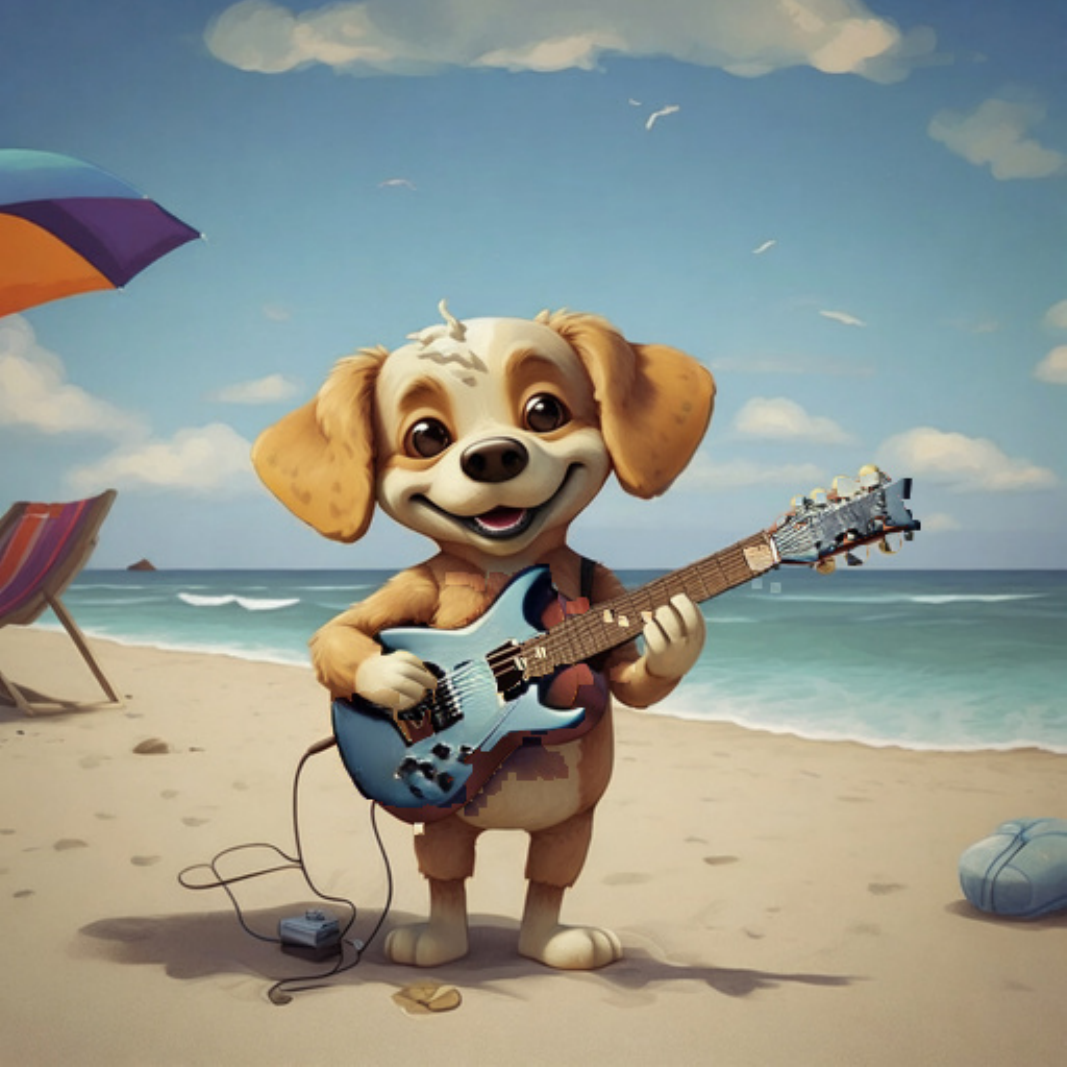}}&
\raisebox{-.5\height}{
\includegraphics[width=0.11\linewidth]{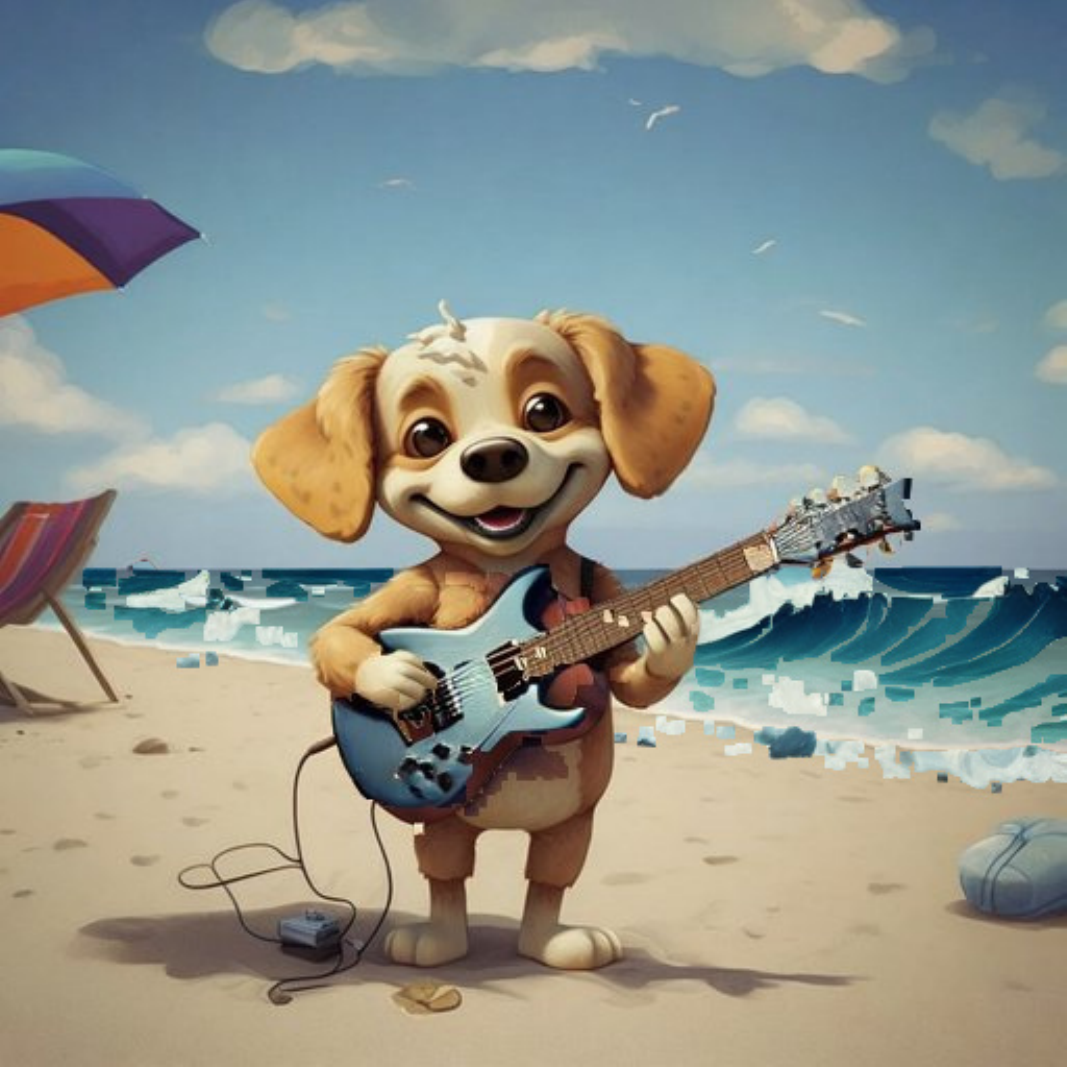}}&
\raisebox{-.5\height}{
\includegraphics[width=0.11\linewidth]{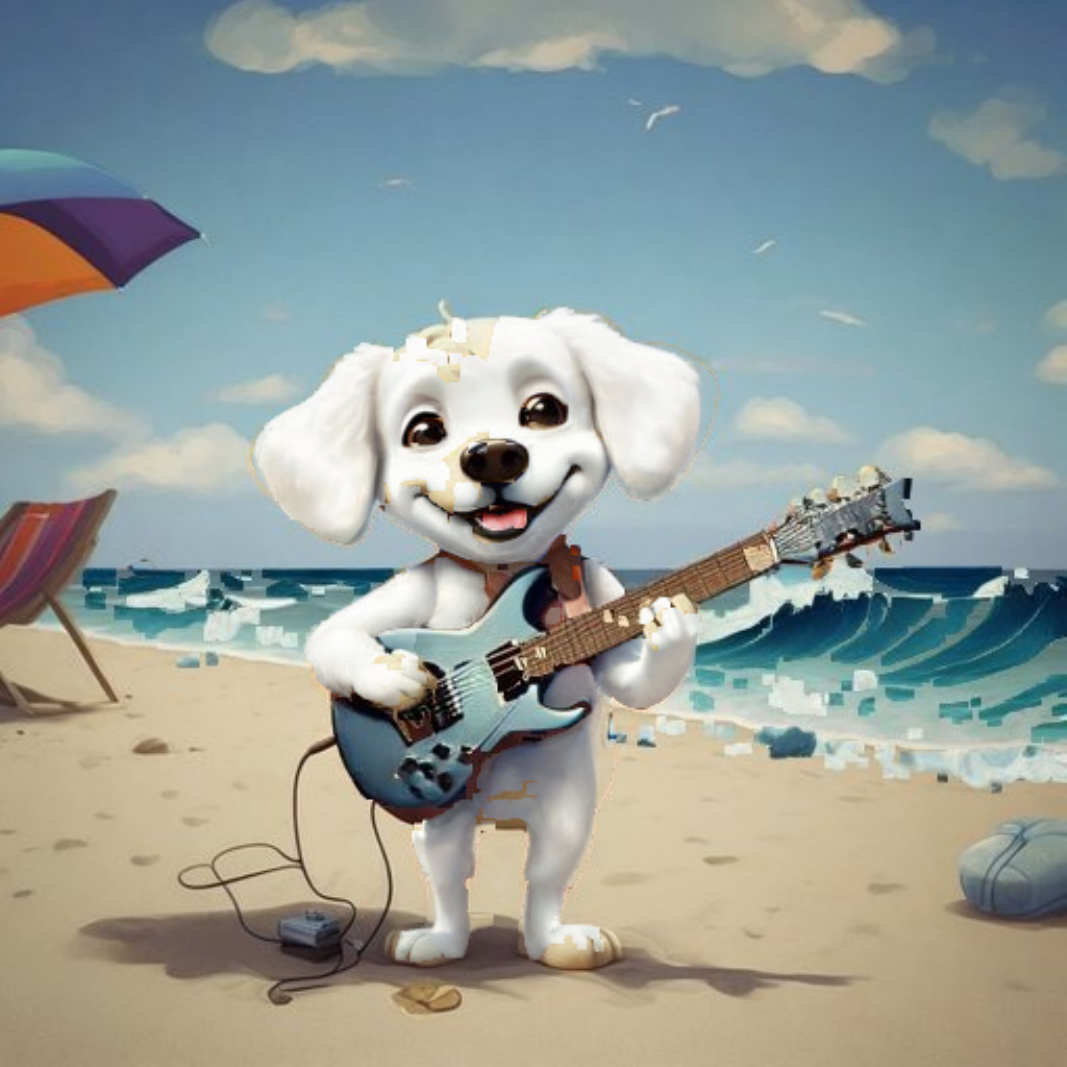}}&
\raisebox{-.5\height}{
\includegraphics[width=0.11\linewidth]{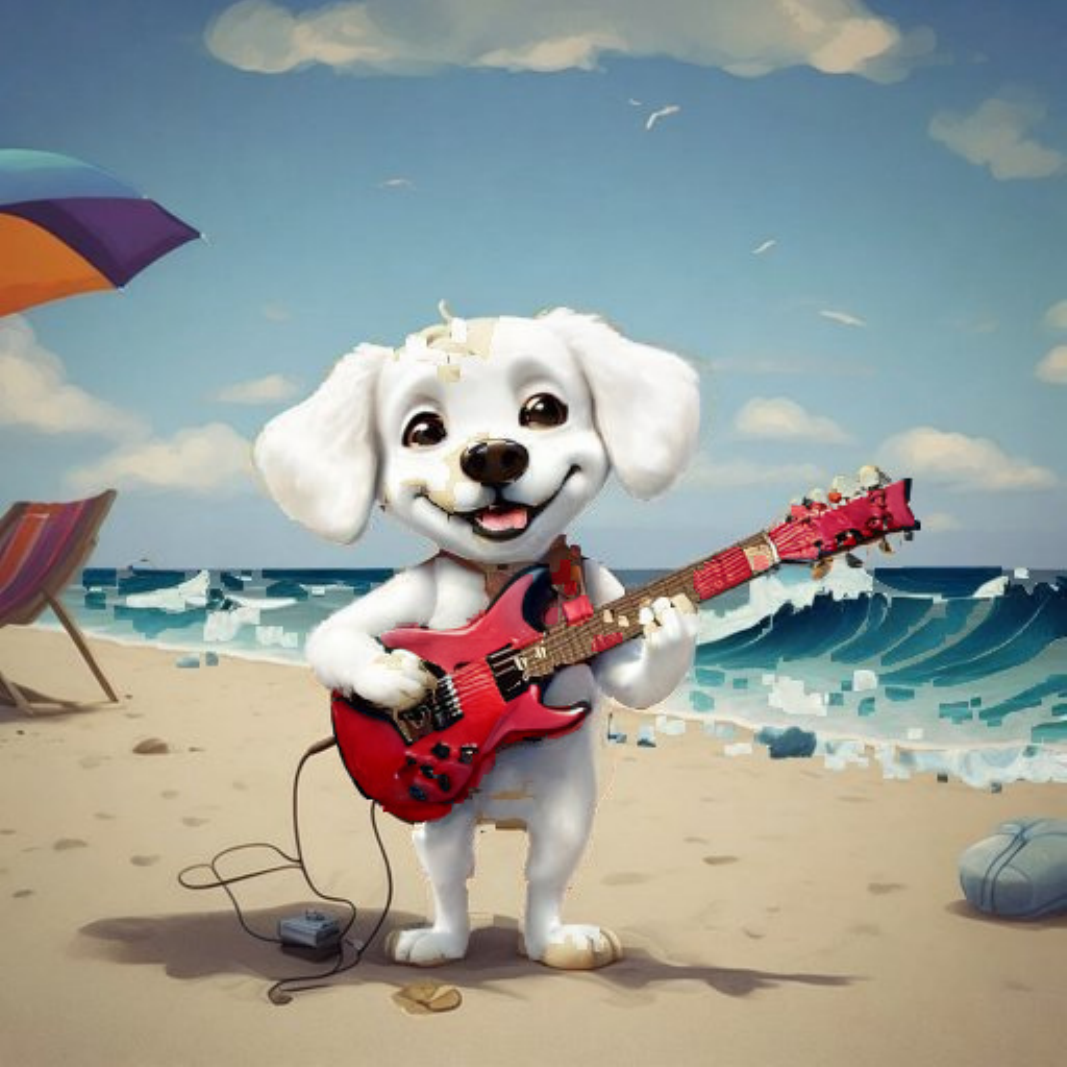}}&
\raisebox{-.5\height}{
\includegraphics[width=0.11\linewidth]{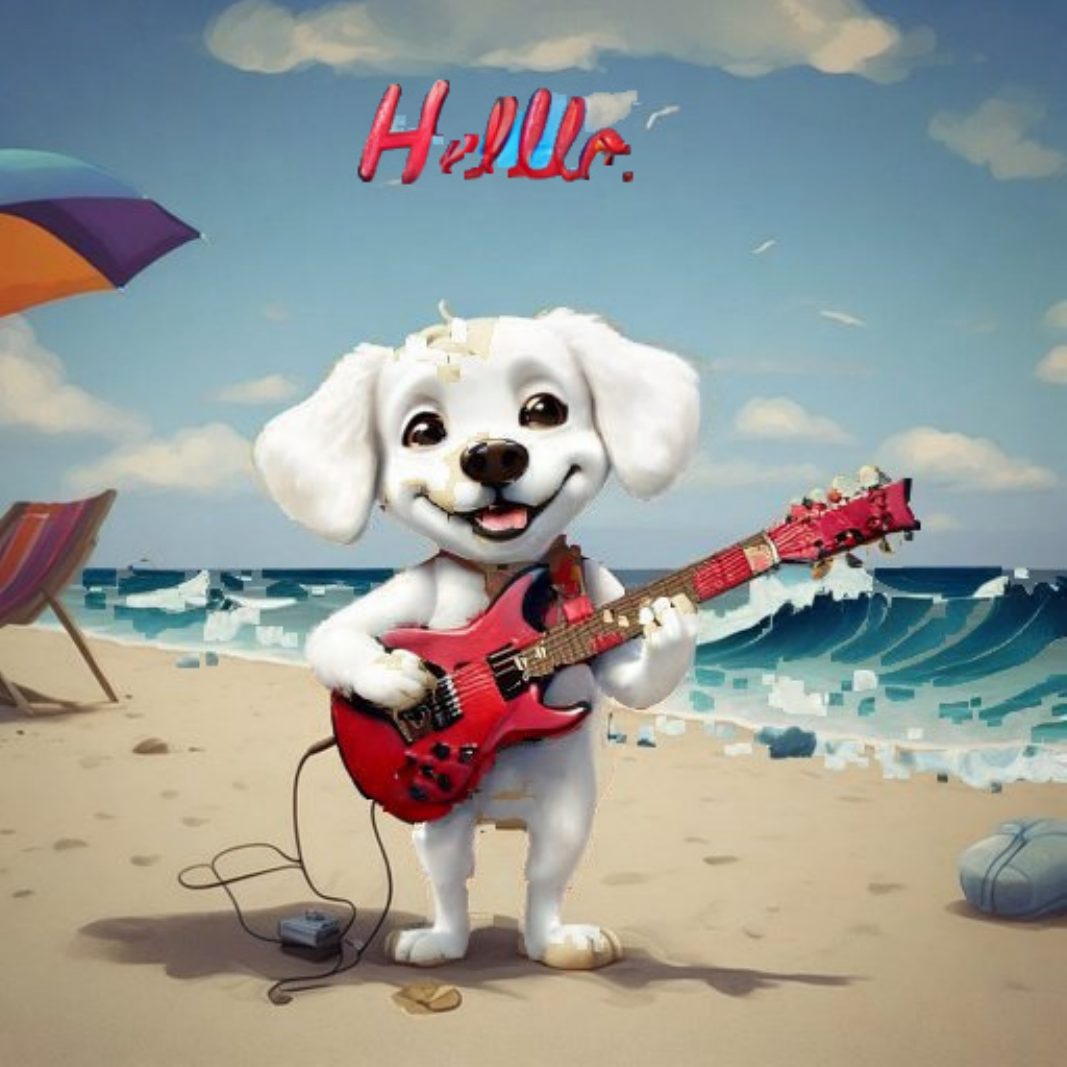}}&
\raisebox{-.5\height}{
\includegraphics[width=0.11\linewidth]{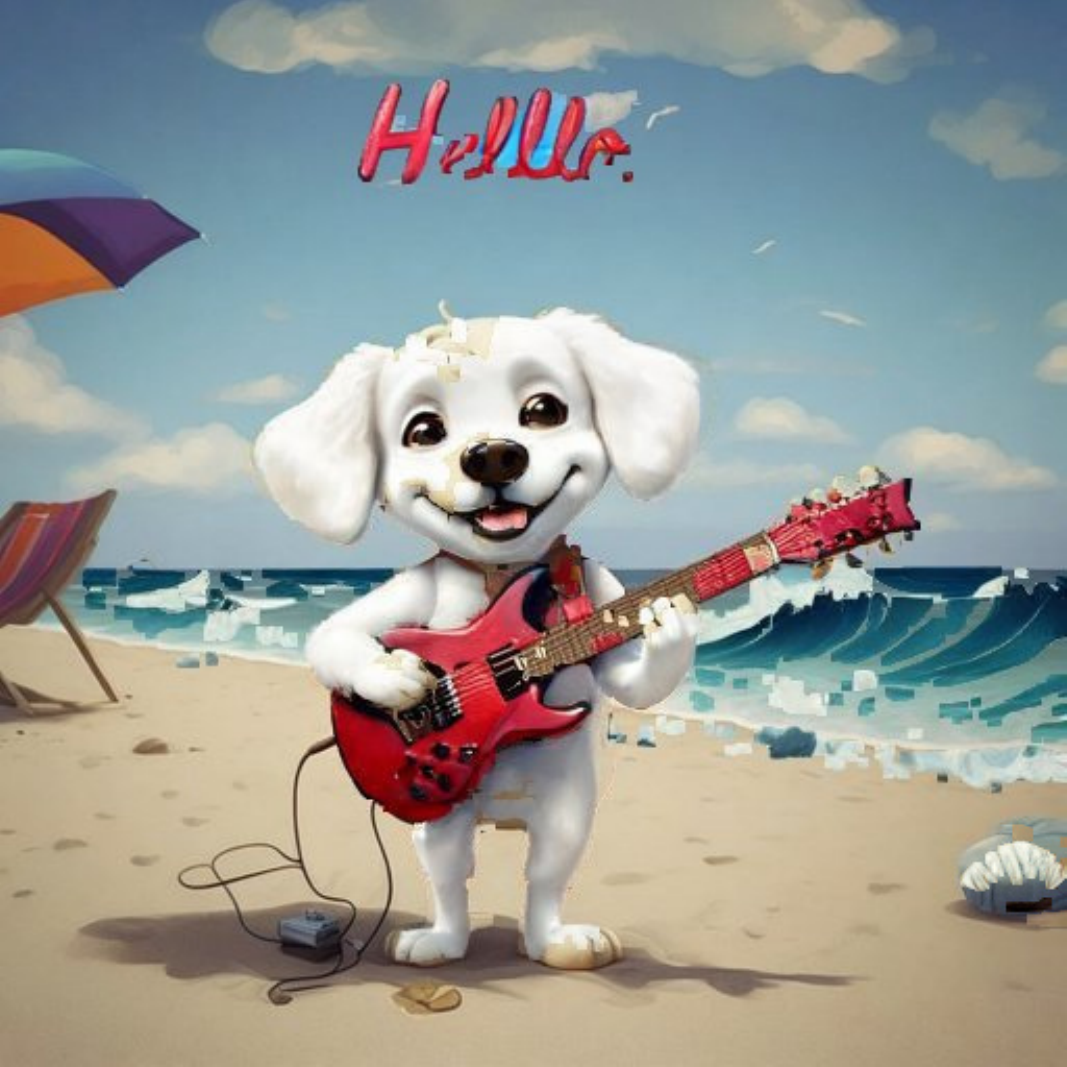}}&
\raisebox{-.5\height}{
\includegraphics[width=0.11\linewidth]{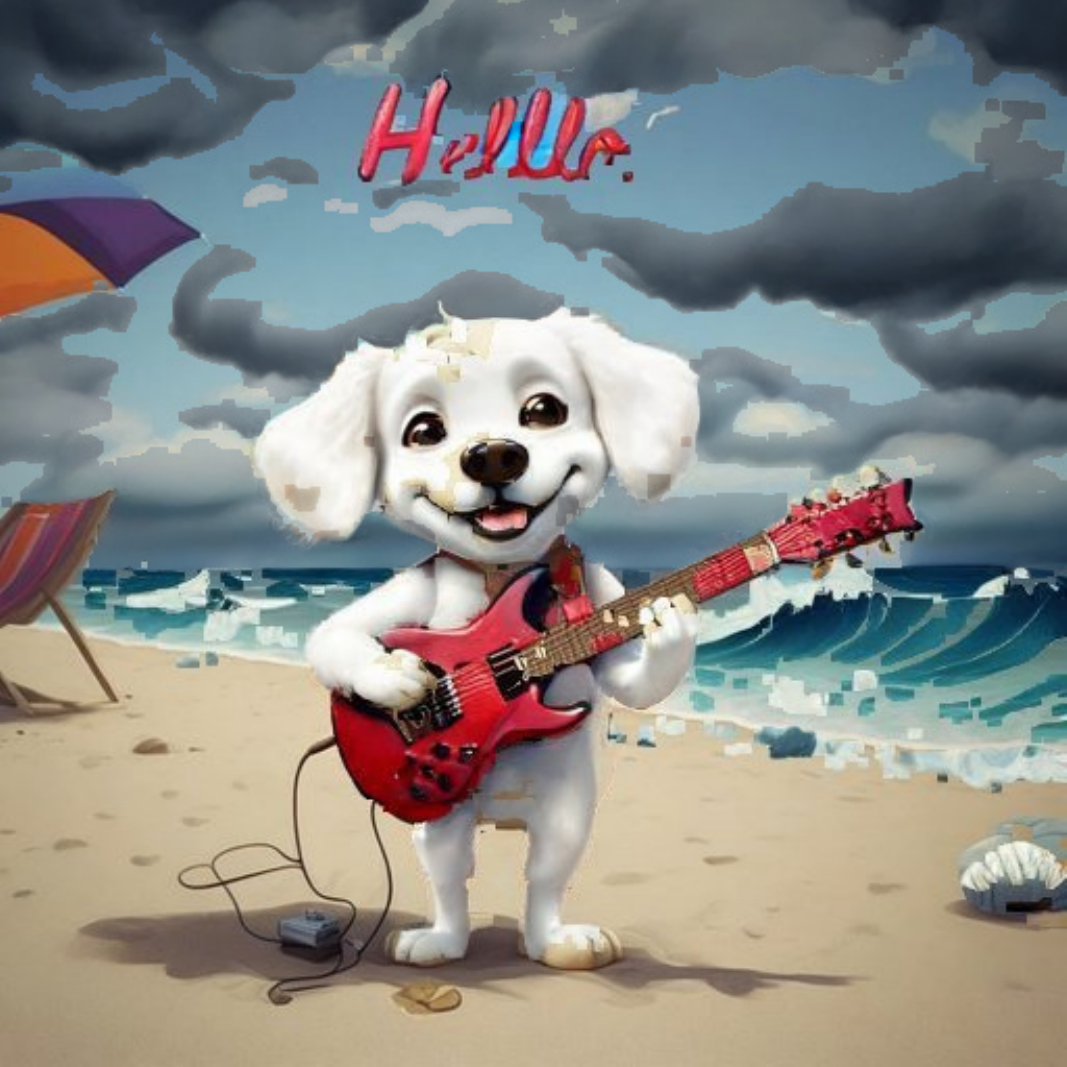}} \\ [13mm]

\resizebox{!}{8px}{
\begin{tabular}[x]{@{}c@{}} \footnotesize{$\alpha=0.25$} \end{tabular}}&
\raisebox{-.5\height}{
\includegraphics[width=0.11\linewidth]{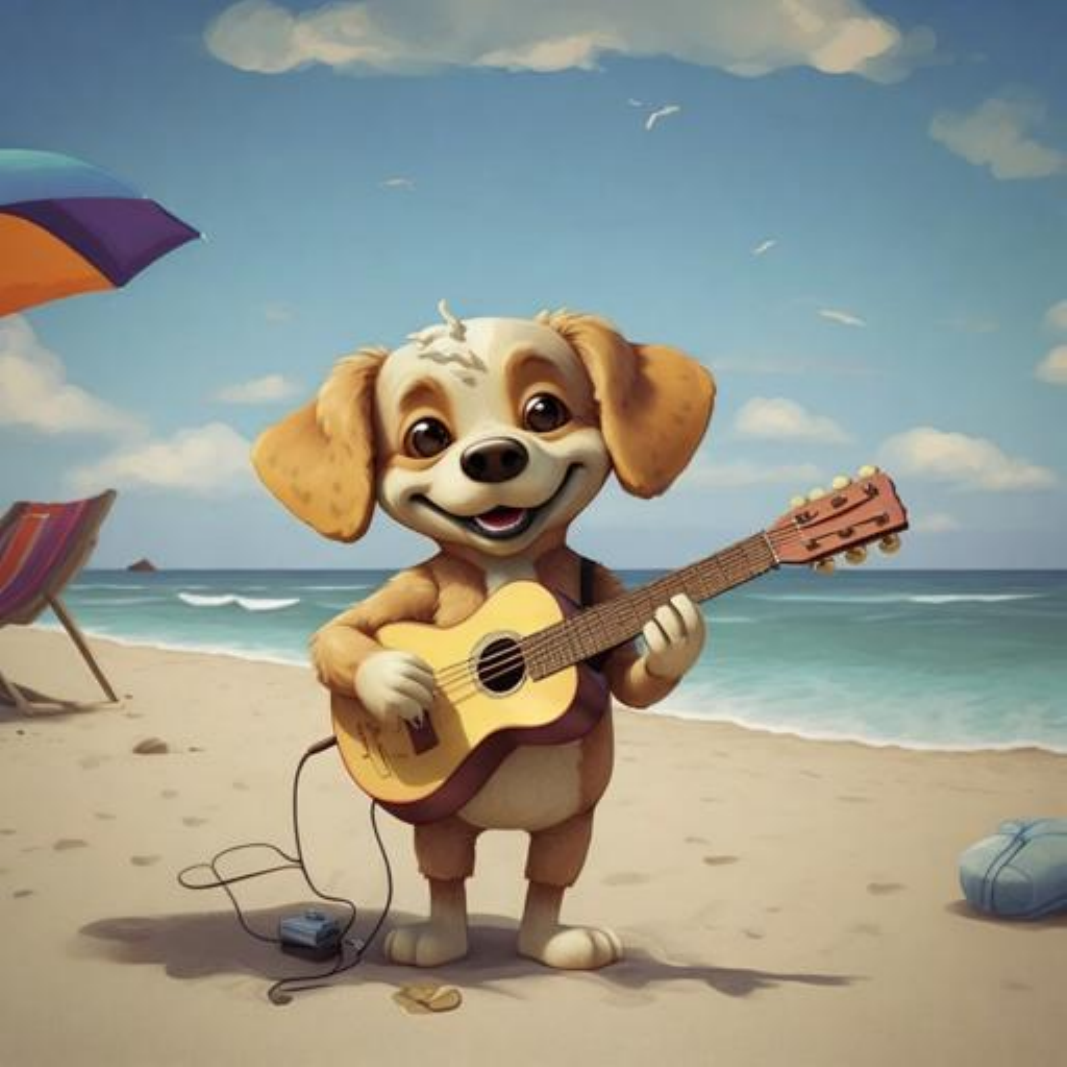}}&
\raisebox{-.5\height}{
\includegraphics[width=0.11\linewidth]{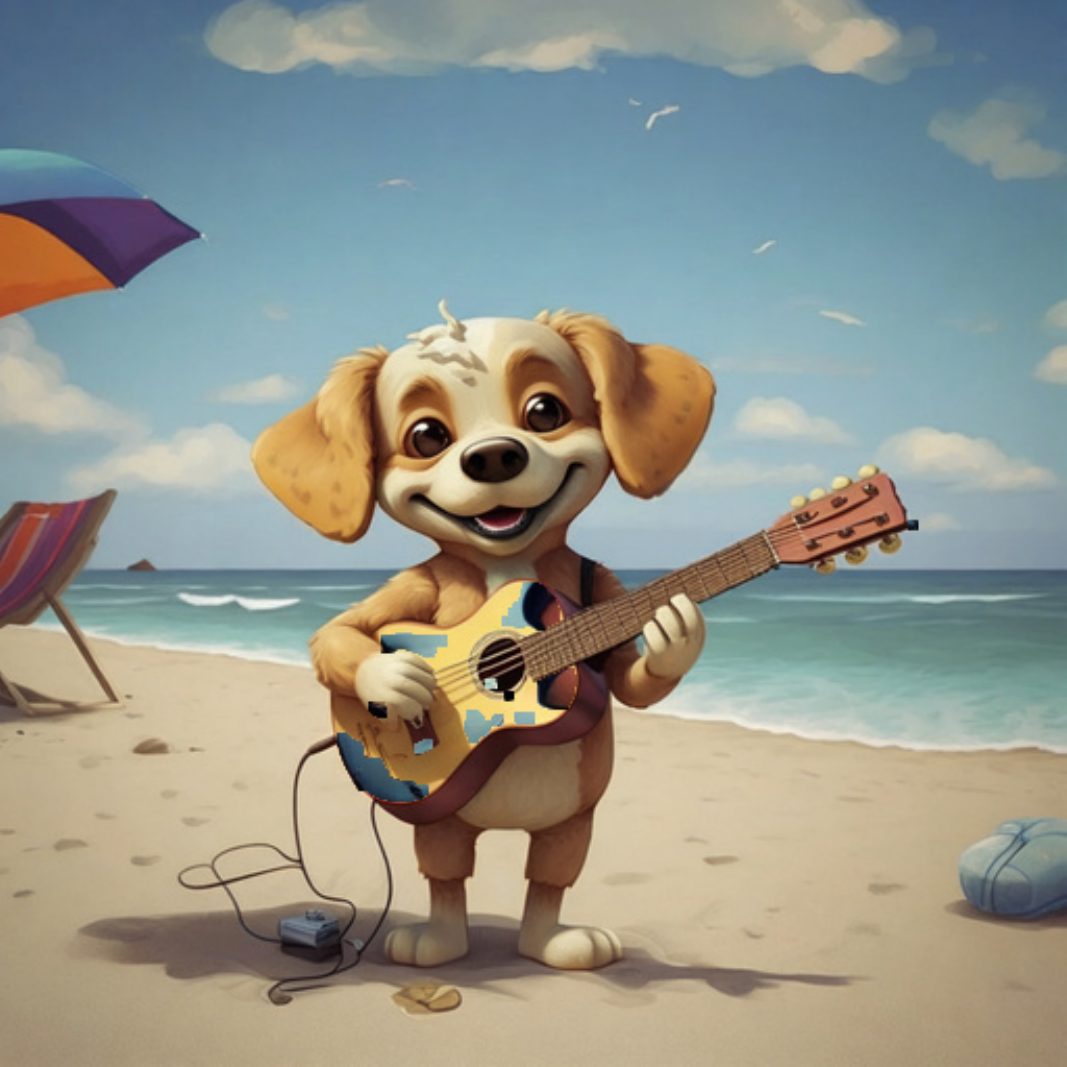}}&
\raisebox{-.5\height}{
\includegraphics[width=0.11\linewidth]{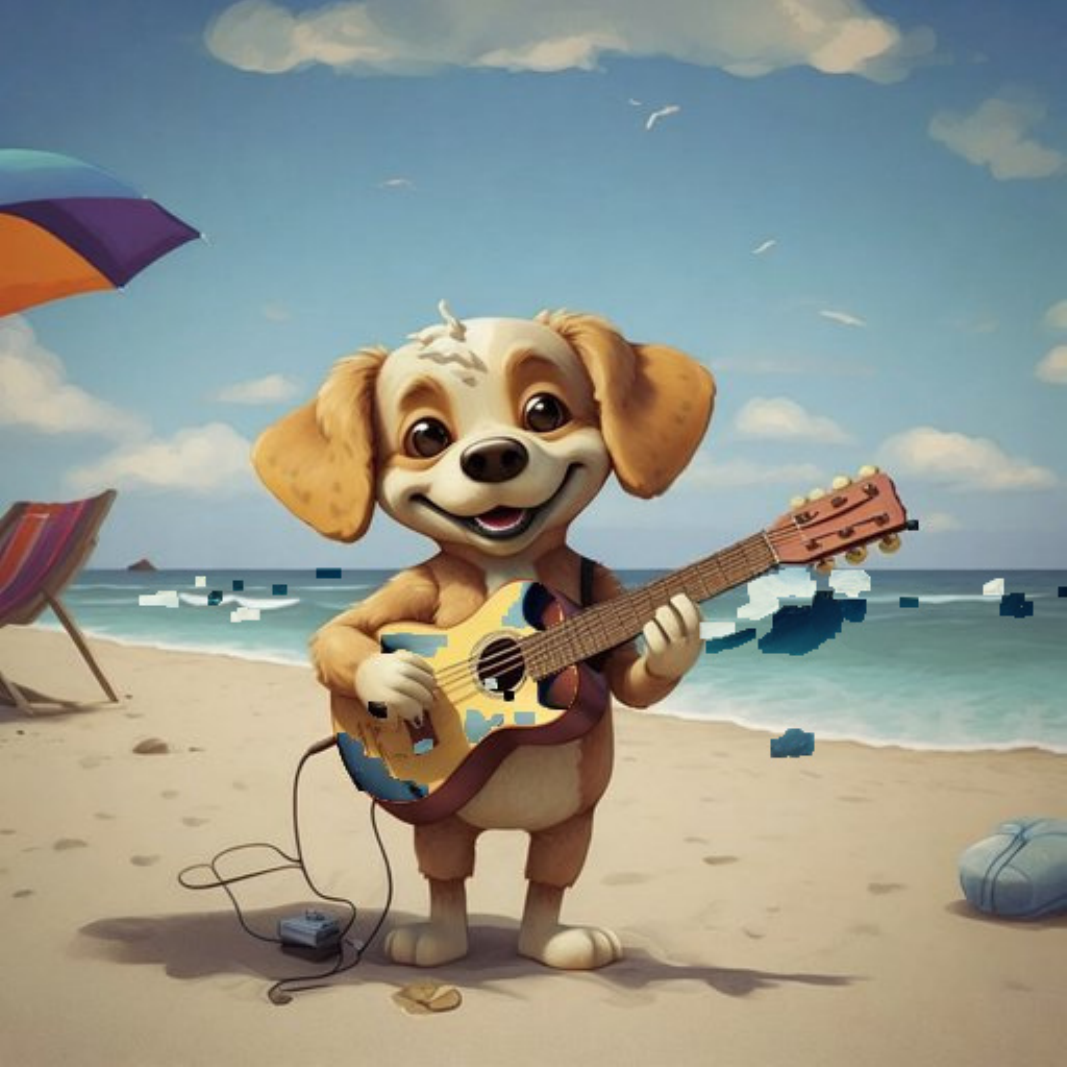}}&
\raisebox{-.5\height}{
\includegraphics[width=0.11\linewidth]{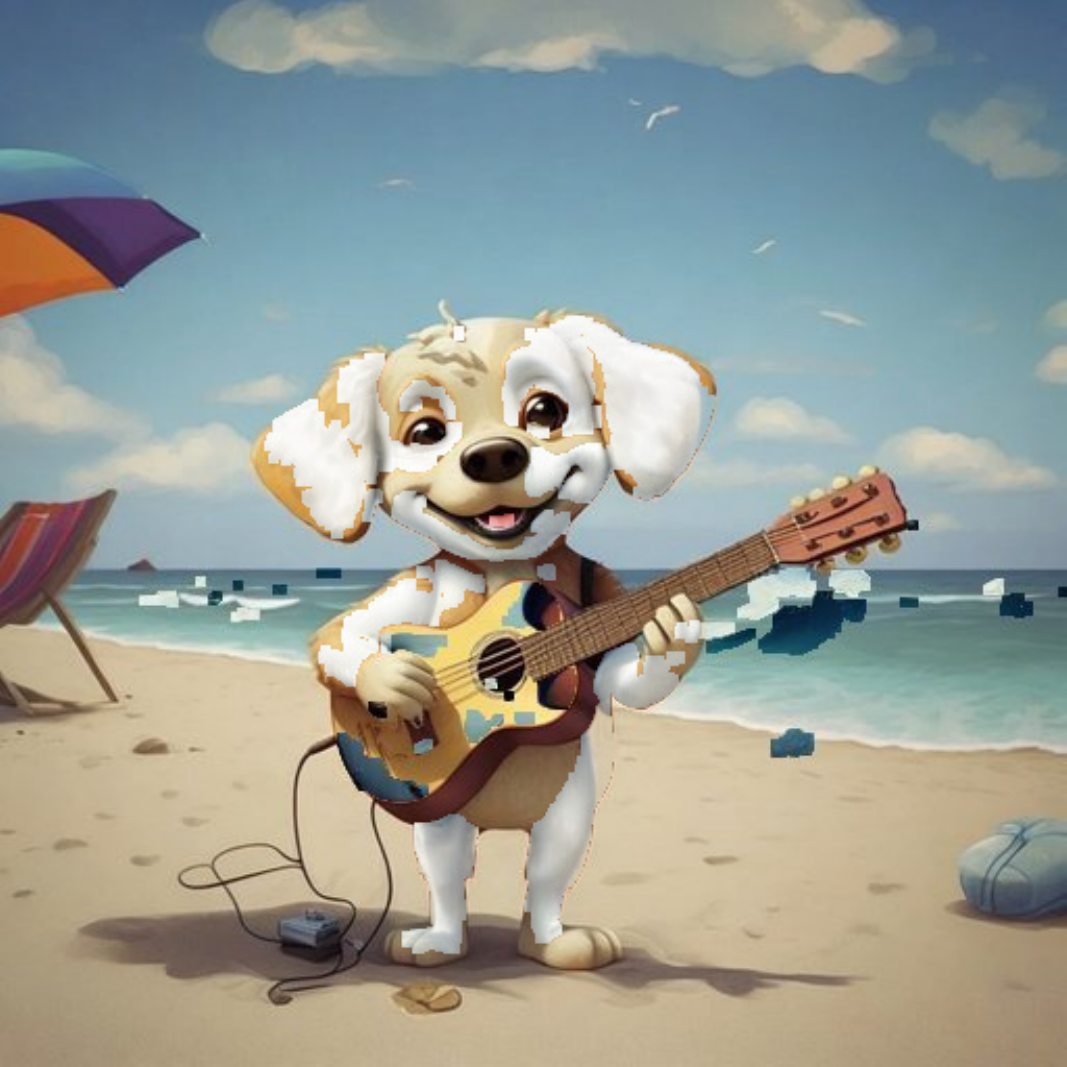}}&
\raisebox{-.5\height}{
\includegraphics[width=0.11\linewidth]{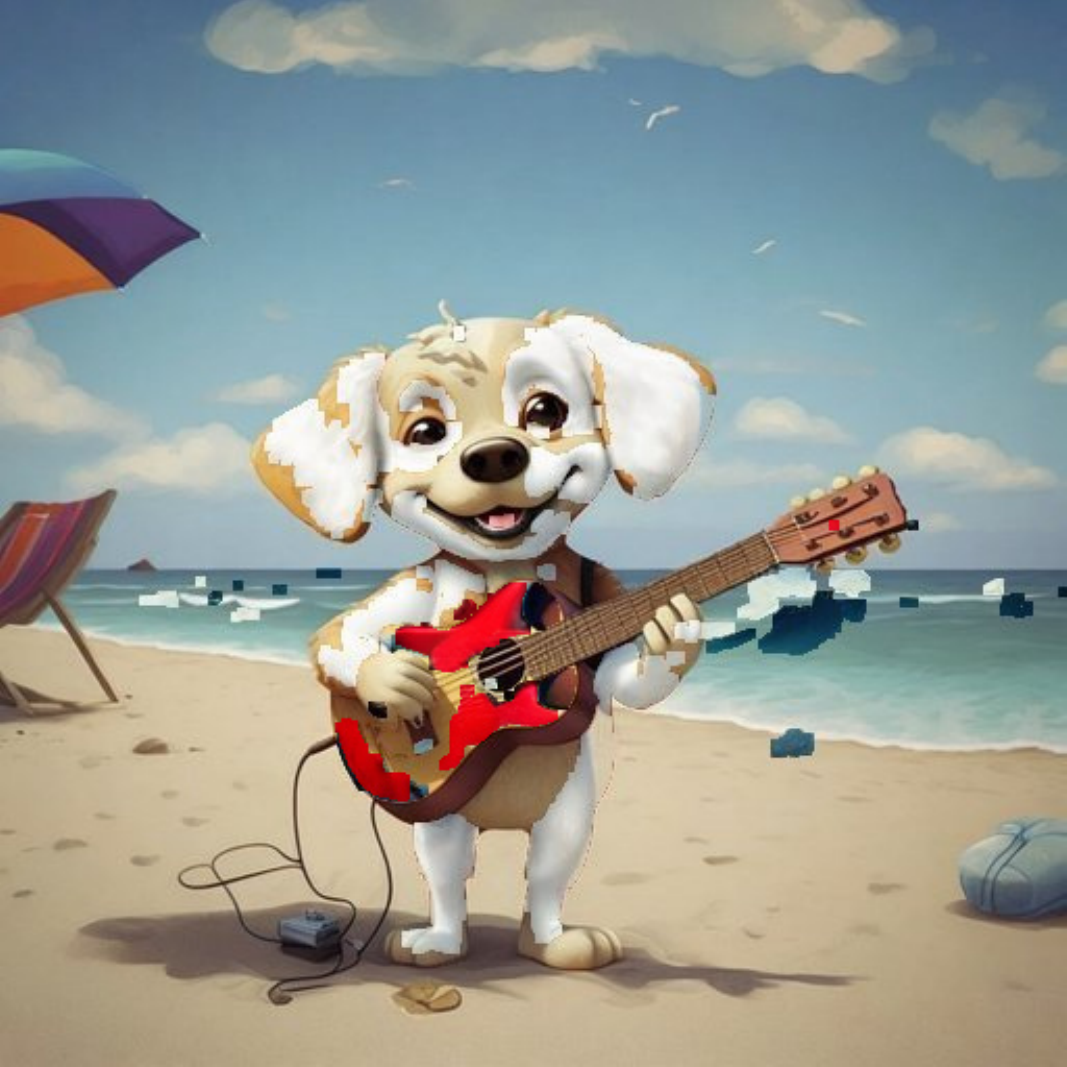}}&
\raisebox{-.5\height}{
\includegraphics[width=0.11\linewidth]{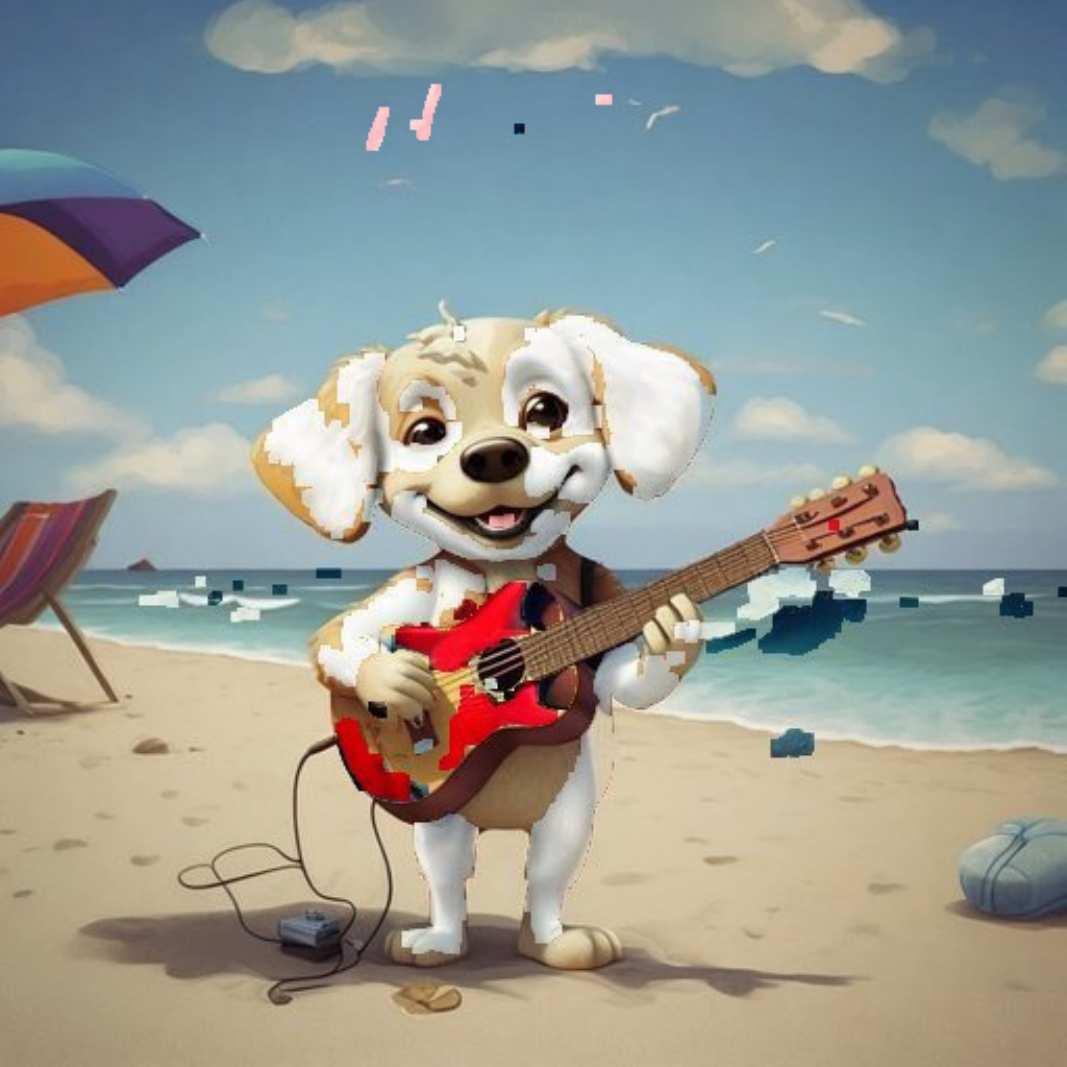}}&
\raisebox{-.5\height}{
\includegraphics[width=0.11\linewidth]{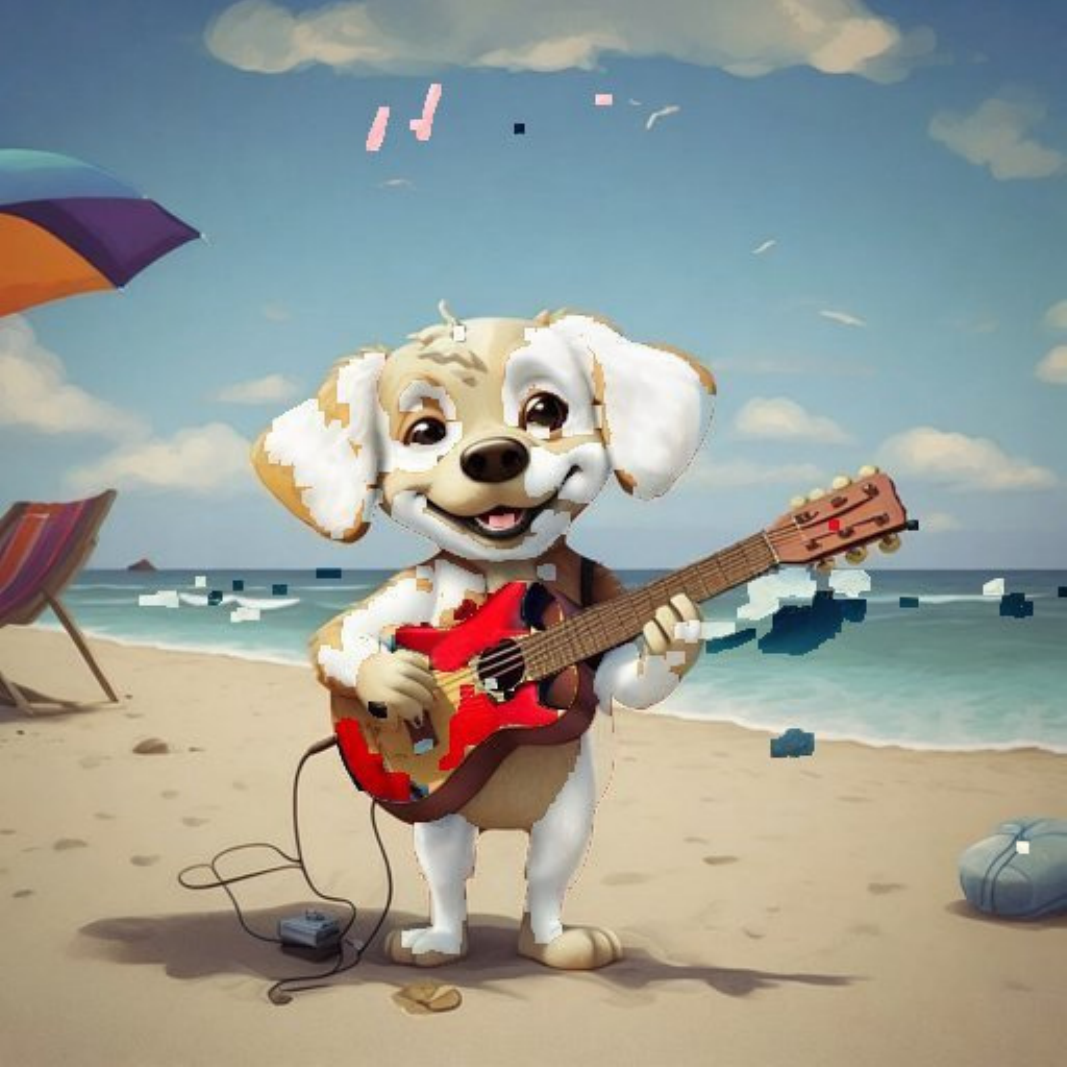}}&
\raisebox{-.5\height}{
\includegraphics[width=0.11\linewidth]{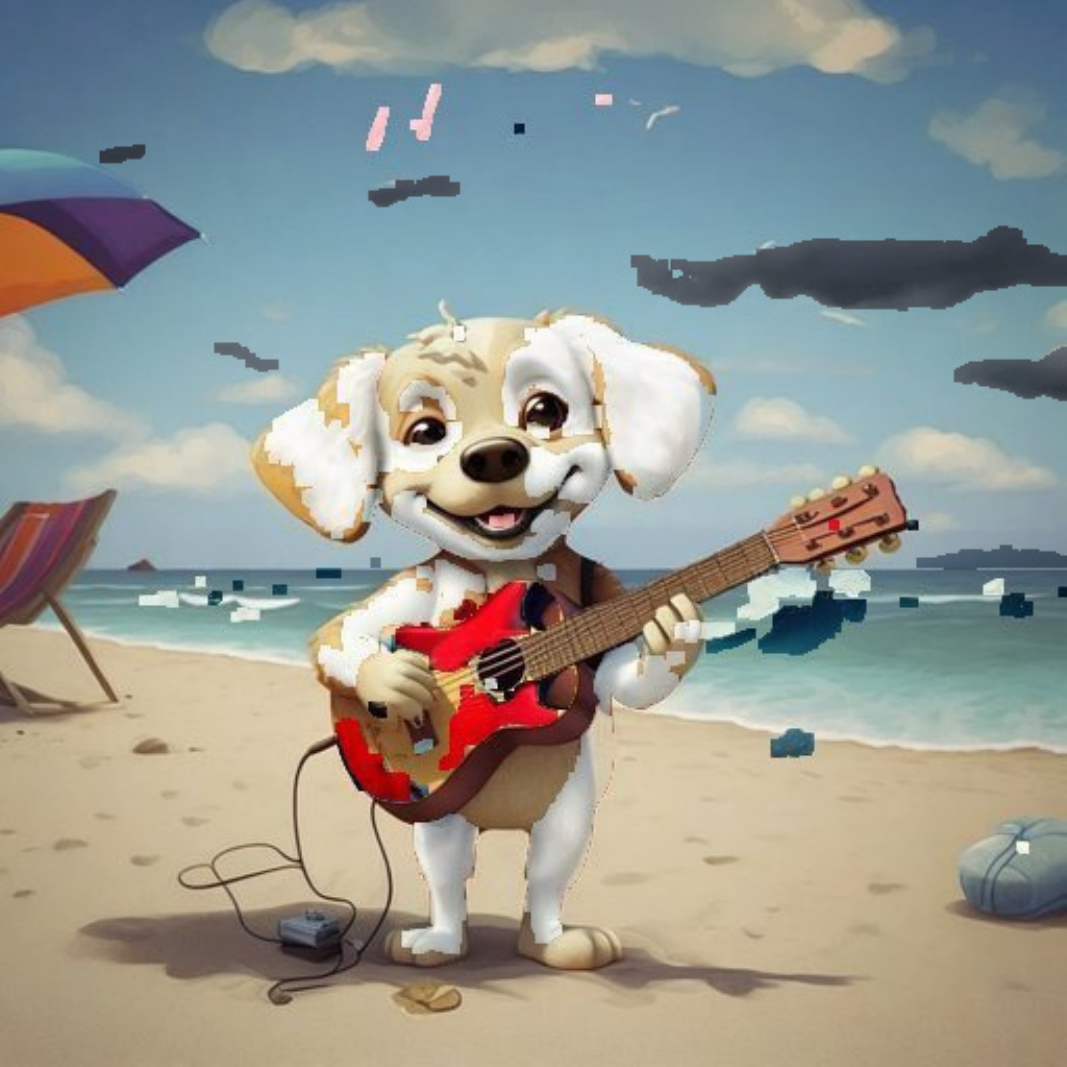}} \\ [13mm]

\resizebox{!}{8px}{
\begin{tabular}[x]{@{}c@{}} \footnotesize{$\alpha=0.5$} \end{tabular}}&
\raisebox{-.5\height}{
\includegraphics[width=0.11\linewidth]{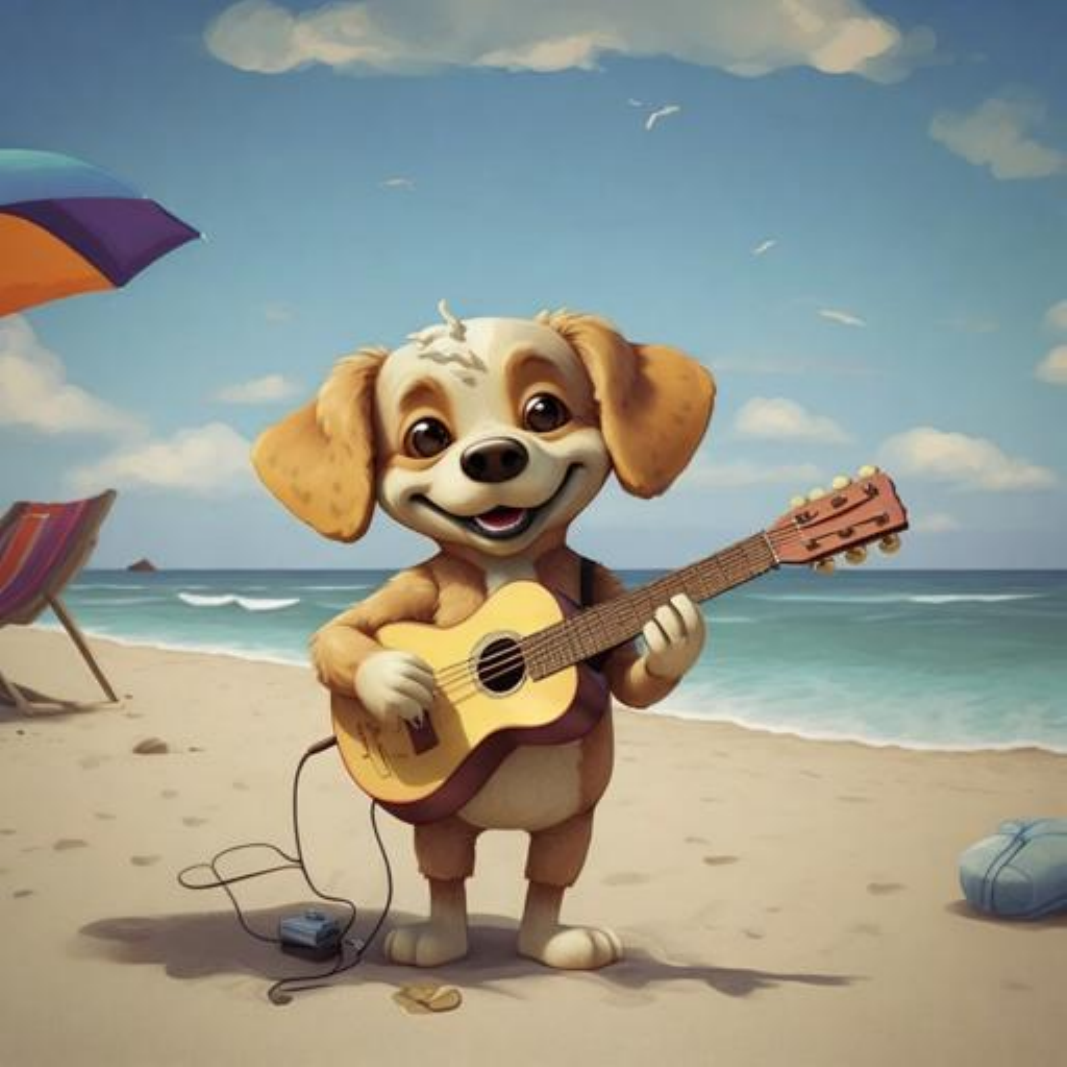}}&
\raisebox{-.5\height}{
\includegraphics[width=0.11\linewidth]{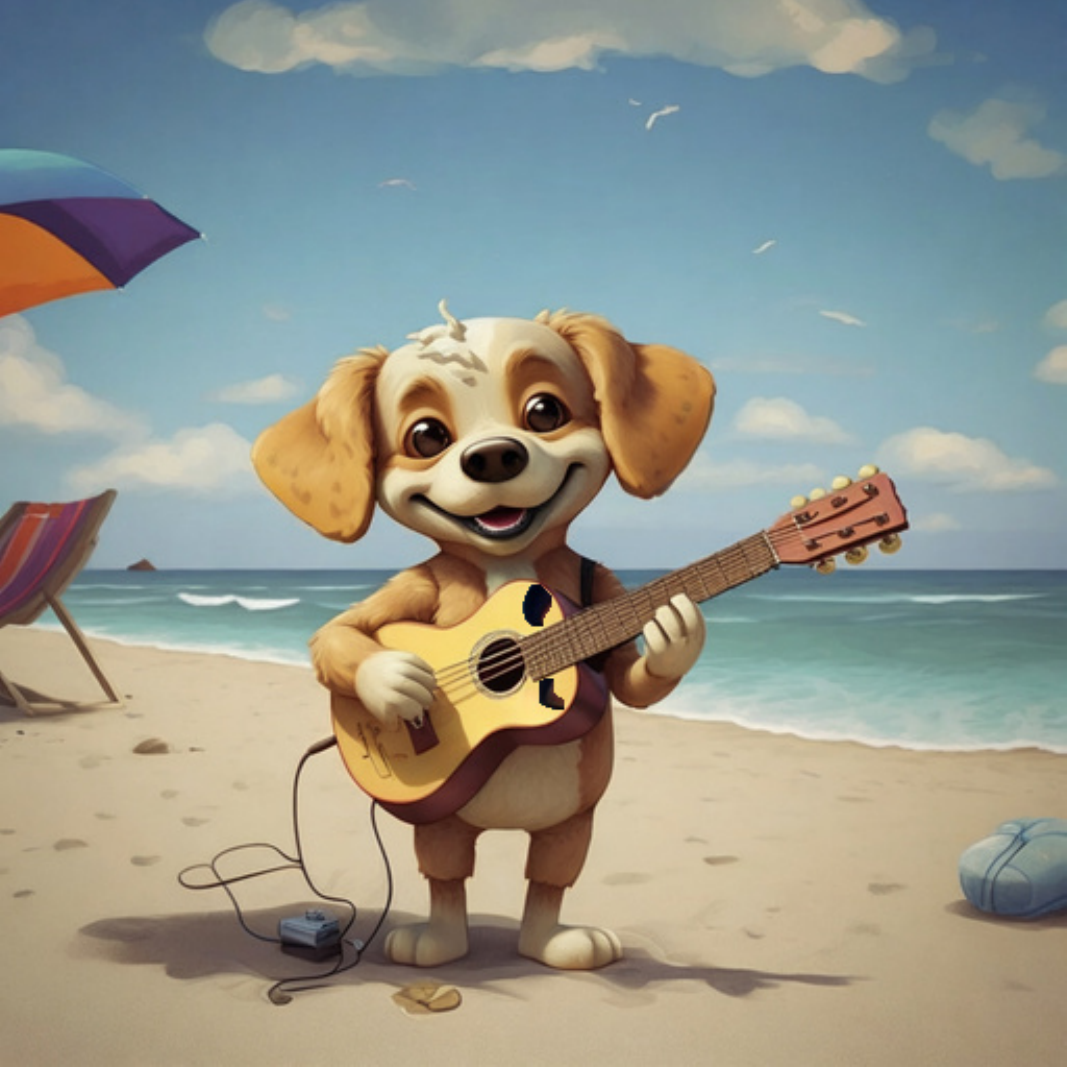}}&
\raisebox{-.5\height}{
\includegraphics[width=0.11\linewidth]{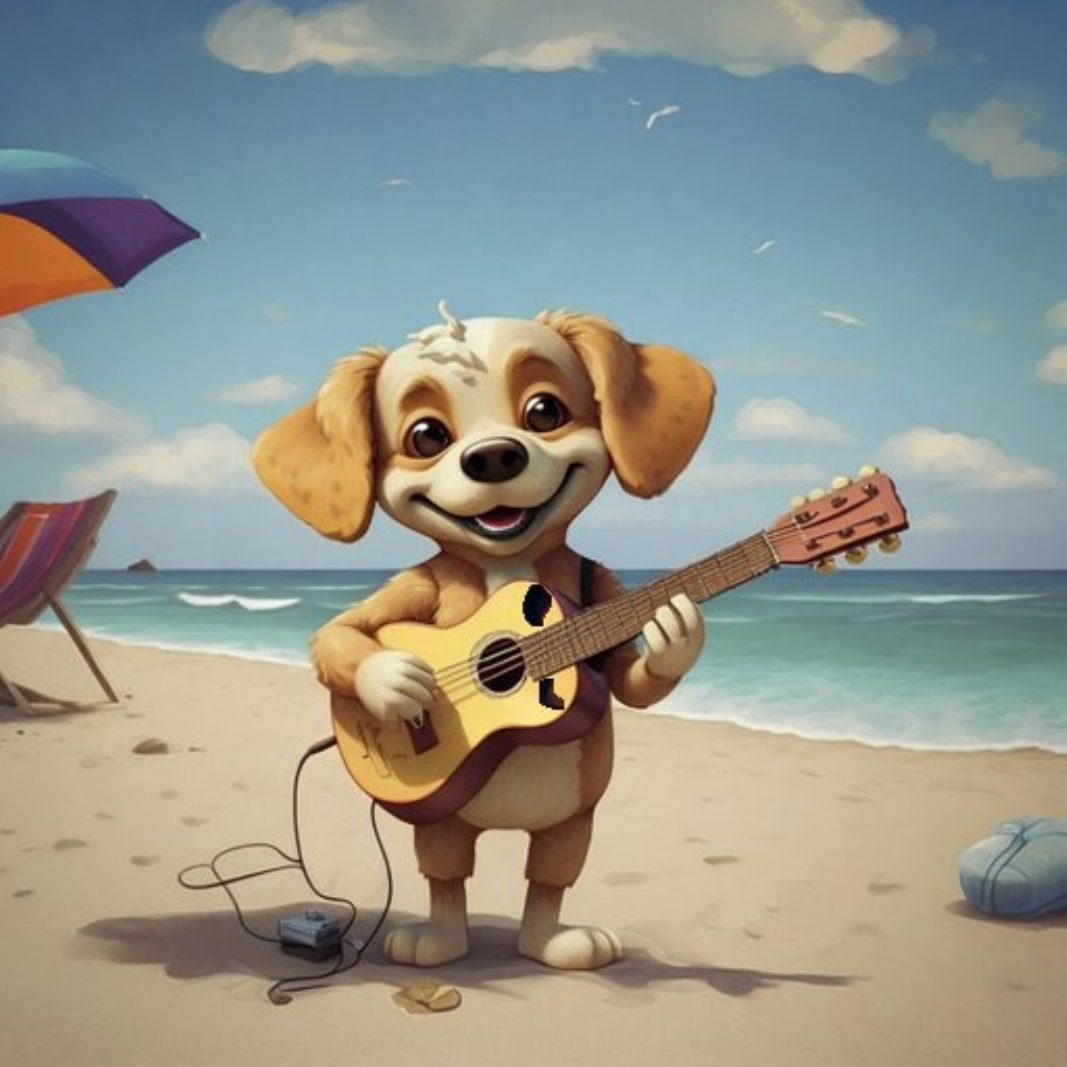}}&
\raisebox{-.5\height}{
\includegraphics[width=0.11\linewidth]{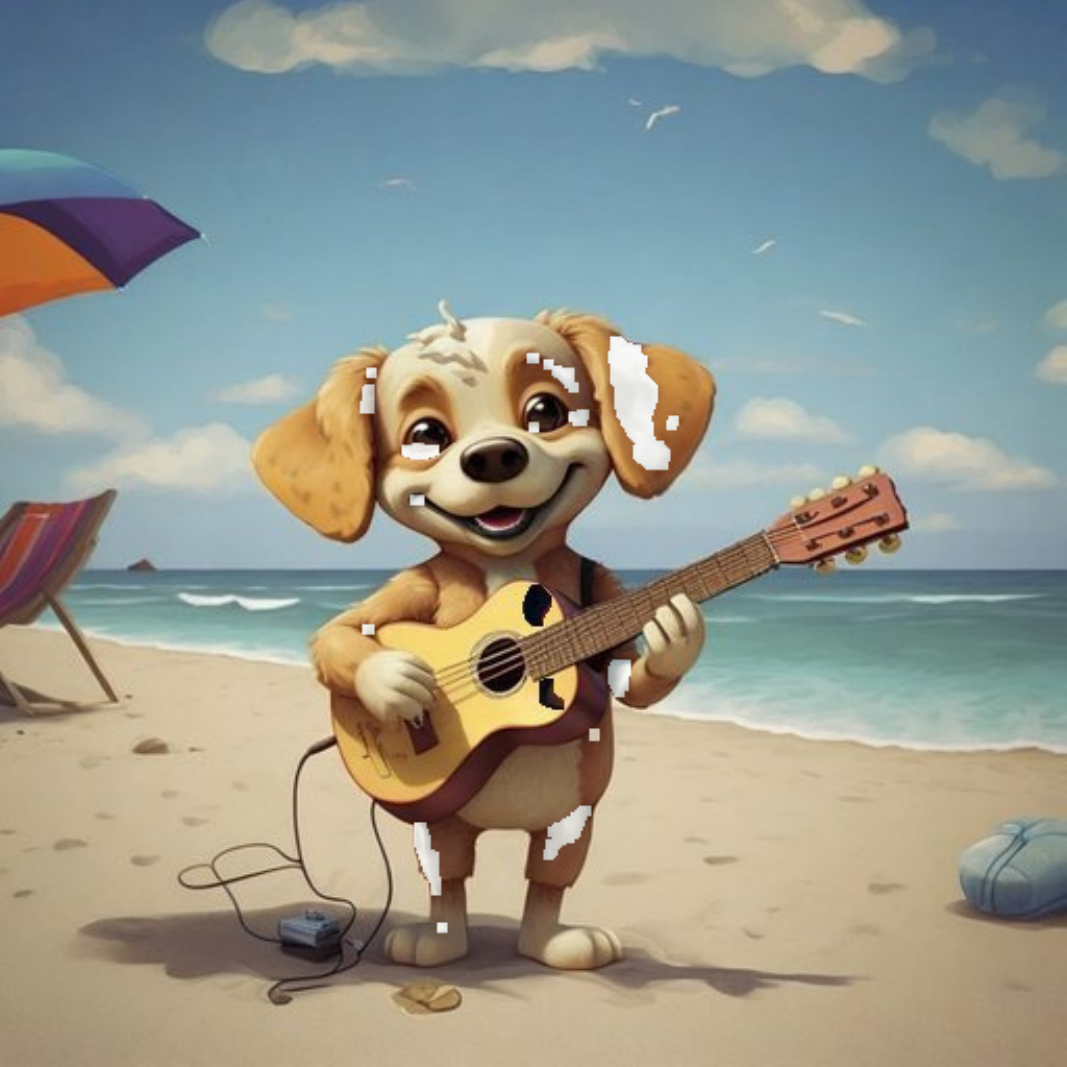}}&
\raisebox{-.5\height}{
\includegraphics[width=0.11\linewidth]{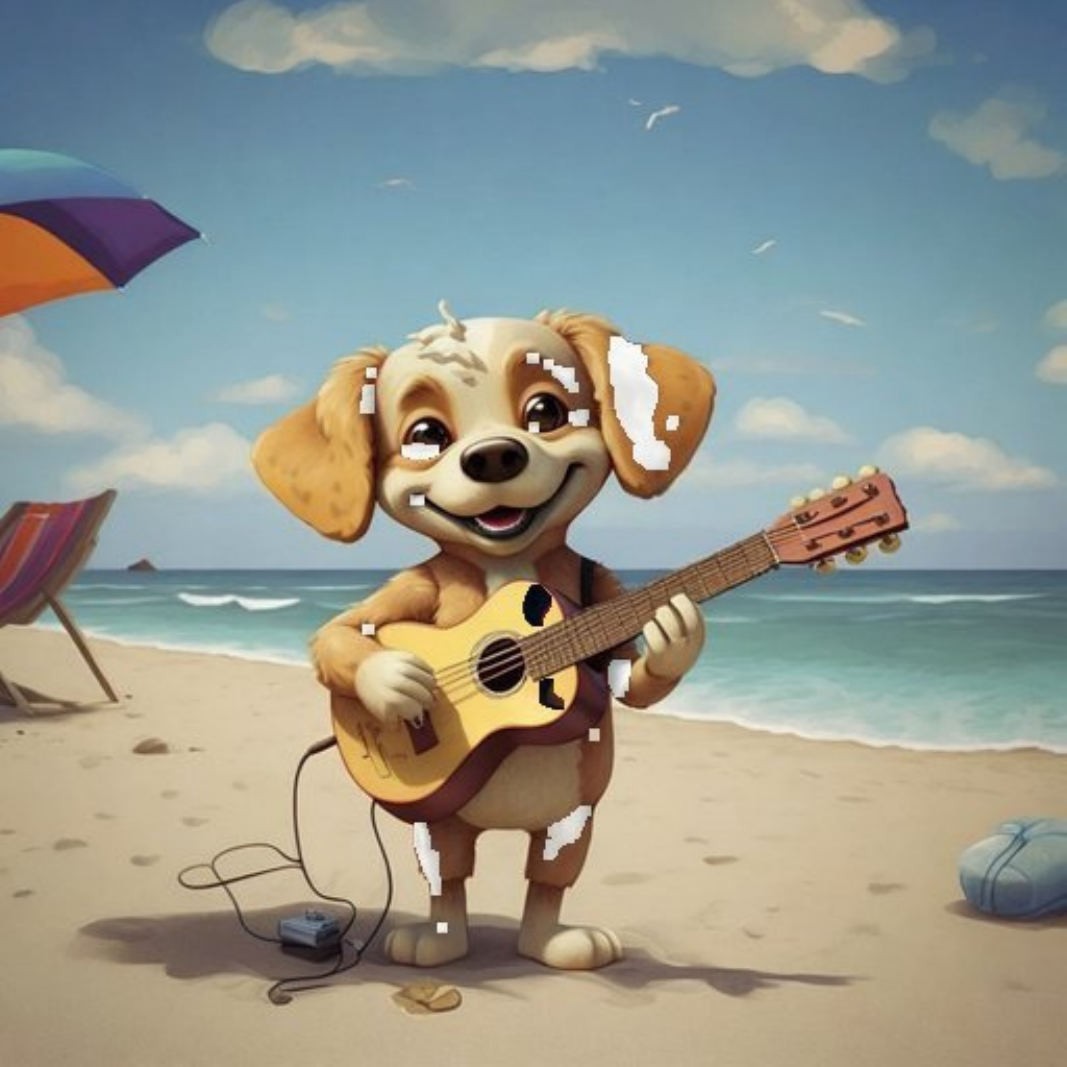}}&
\raisebox{-.5\height}{
\includegraphics[width=0.11\linewidth]{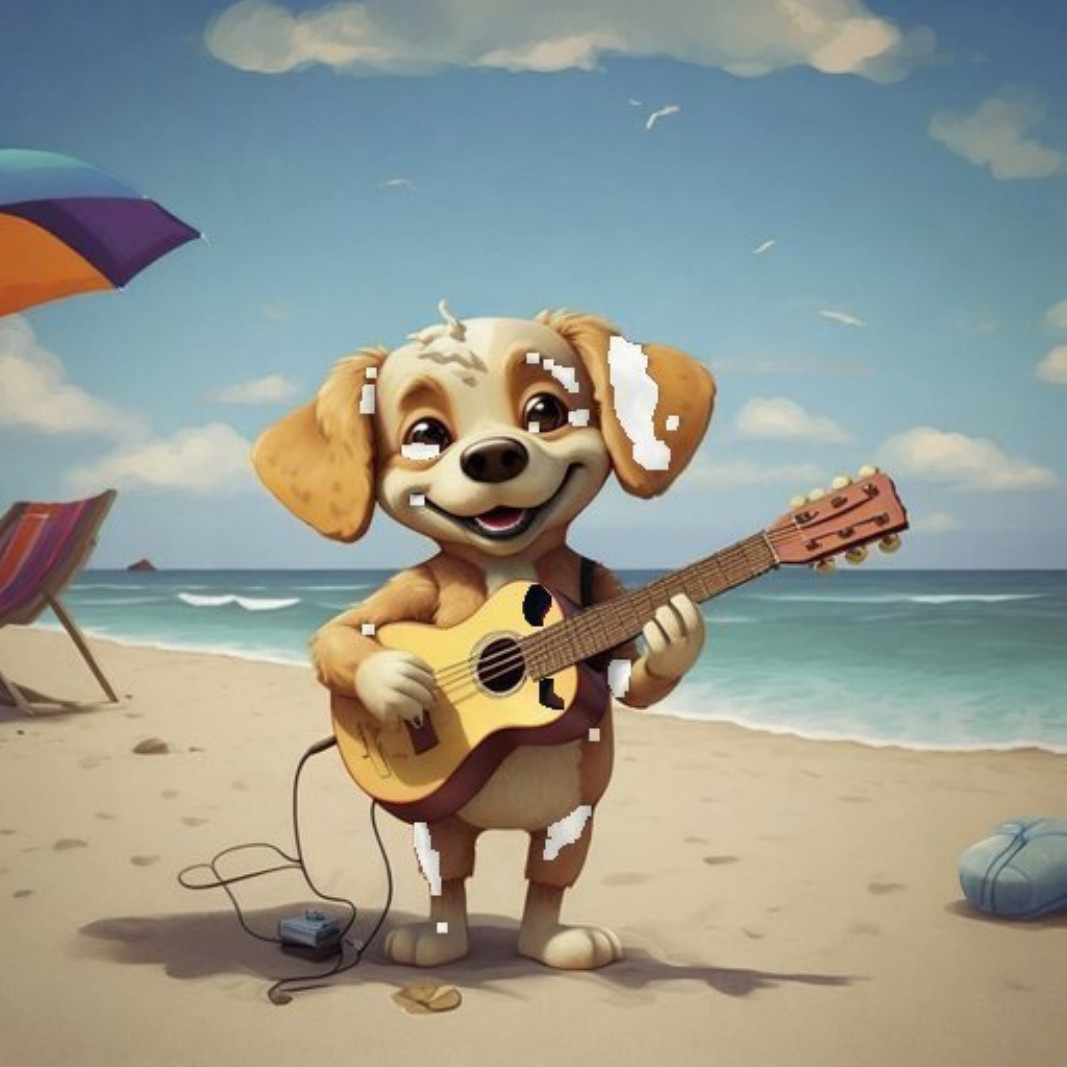}}&
\raisebox{-.5\height}{
\includegraphics[width=0.11\linewidth]{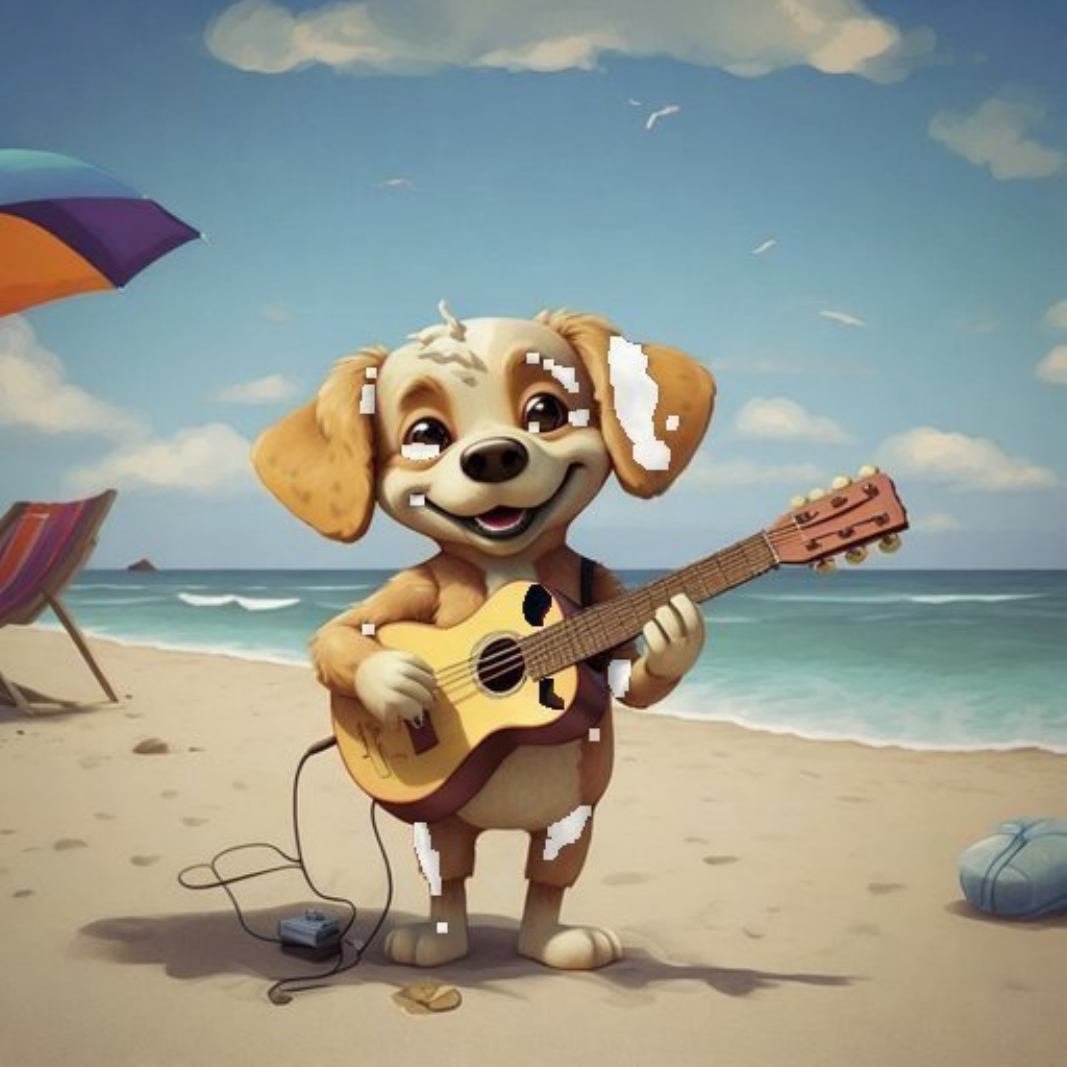}}&
\raisebox{-.5\height}{
\includegraphics[width=0.11\linewidth]{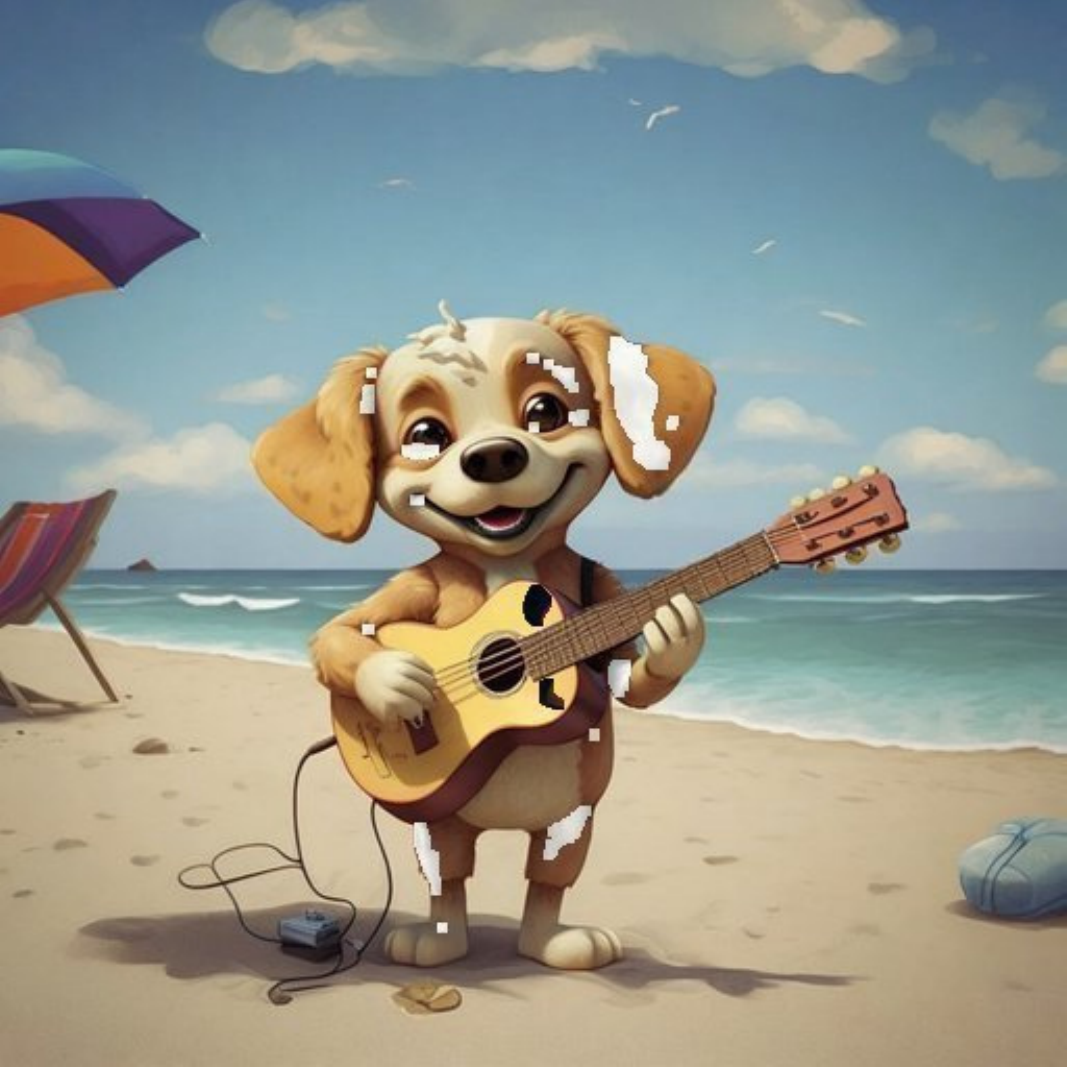}}

\end{tabular}
\caption{Effect of Sequential Edit Thresholding during sequential edits (from left to right) with different $\alpha$ values.}
\label{fig:multiturn} 
\end{figure*}

\subsection{Qualitative Comparisons with Existing Approaches}
\begin{figure*}[t]
   \centering
\begin{tabular}{@{\hspace{-8\tabcolsep}}c@{\hspace{-0.3\tabcolsep}}c@{\hspace{-0.3\tabcolsep}}c@{\hspace{-0.3\tabcolsep}}c@{\hspace{-0.3\tabcolsep}}c}
& \begin{tabular}[x]{@{}c@{}}Input \end{tabular}  & \begin{tabular}[x]{@{}c@{}} \model \end{tabular} &  \begin{tabular}[x]{@{}c@{}} InstructPix2Pix \end{tabular} & MagicBrush  \\
\resizebox{!}{16px}{
\begin{tabular}[x]{@{}c@{}} Give\\ him \\ sneakers \end{tabular}}&
\raisebox{-.5\height}{
\includegraphics[width=0.22\linewidth]{imgs/comp_chosen_samples/mouse/mouse.pdf}}&
\raisebox{-.5\height}{
\includegraphics[width=0.22\linewidth]{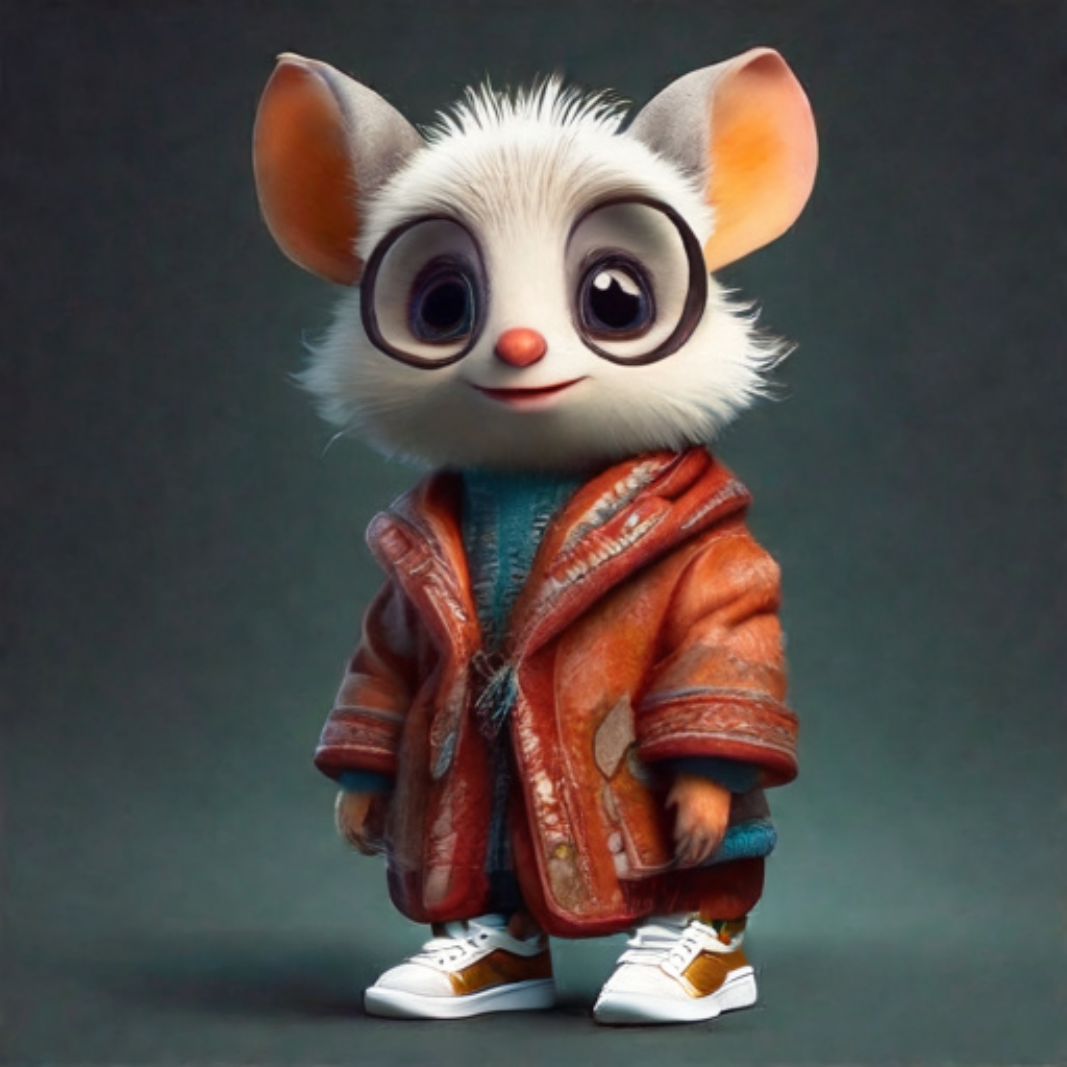}}&
\raisebox{-.5\height}{
\includegraphics[width=0.22\linewidth]{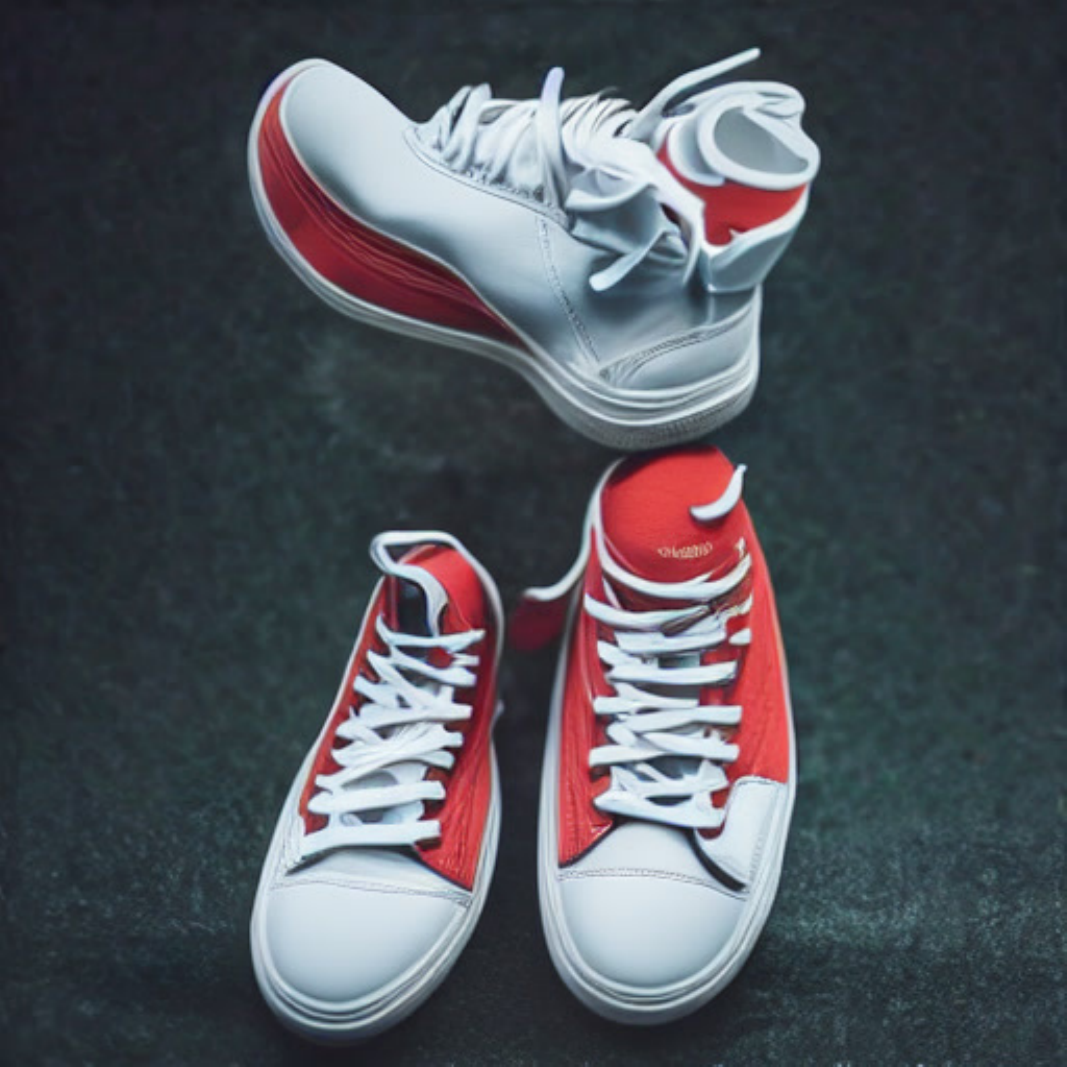 }} &
\raisebox{-.5\height}{
\includegraphics[width=0.22\linewidth]{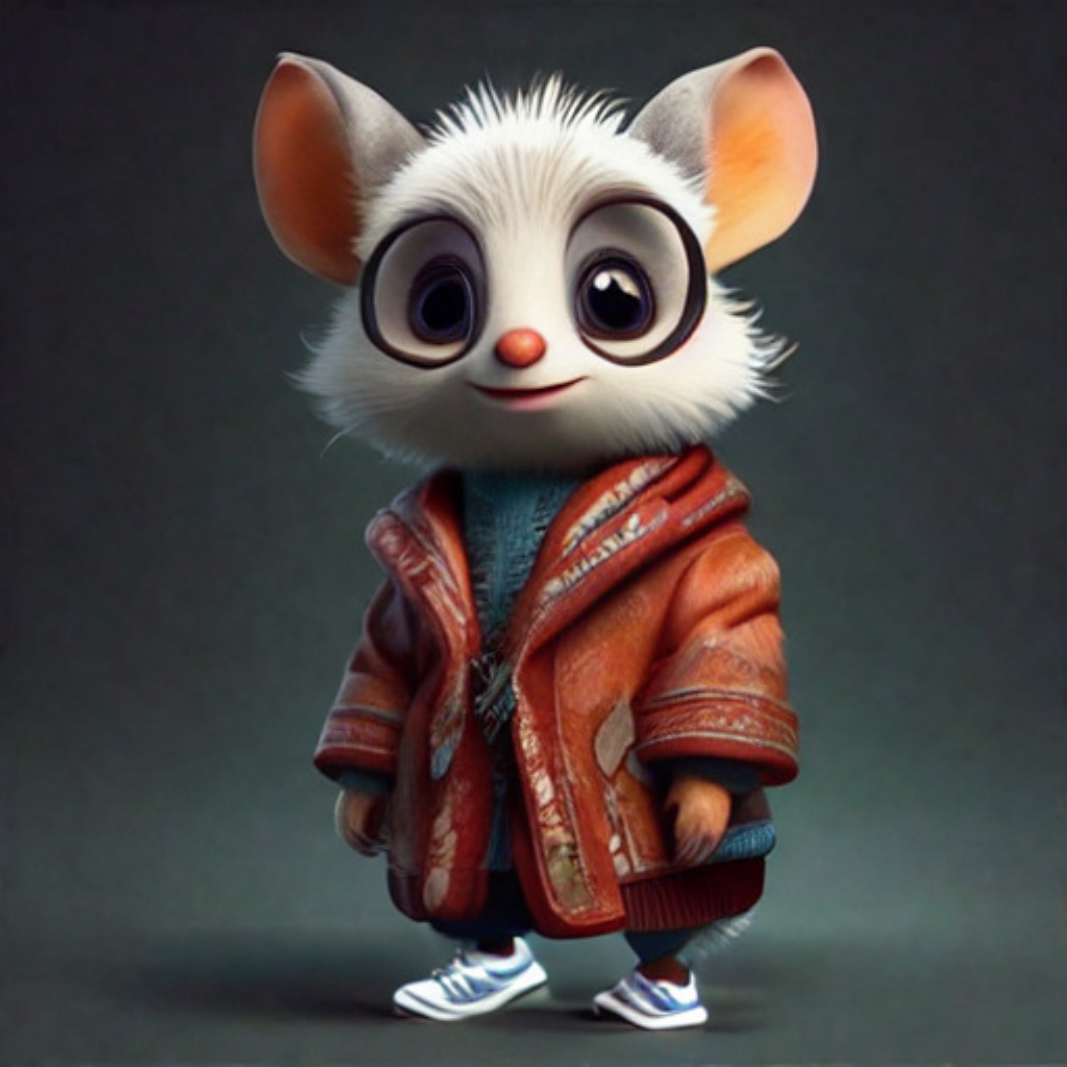}} \\ [11mm]

\resizebox{!}{16px}{
\begin{tabular}[x]{@{}c@{}}Put a\\ big smile \\ on his face \end{tabular}}&
\raisebox{-.5\height}{
\includegraphics[width=0.22\linewidth]{imgs/comp_chosen_samples/mouse/mouse.pdf}}&
\raisebox{-.5\height}{
\includegraphics[width=0.22\linewidth]{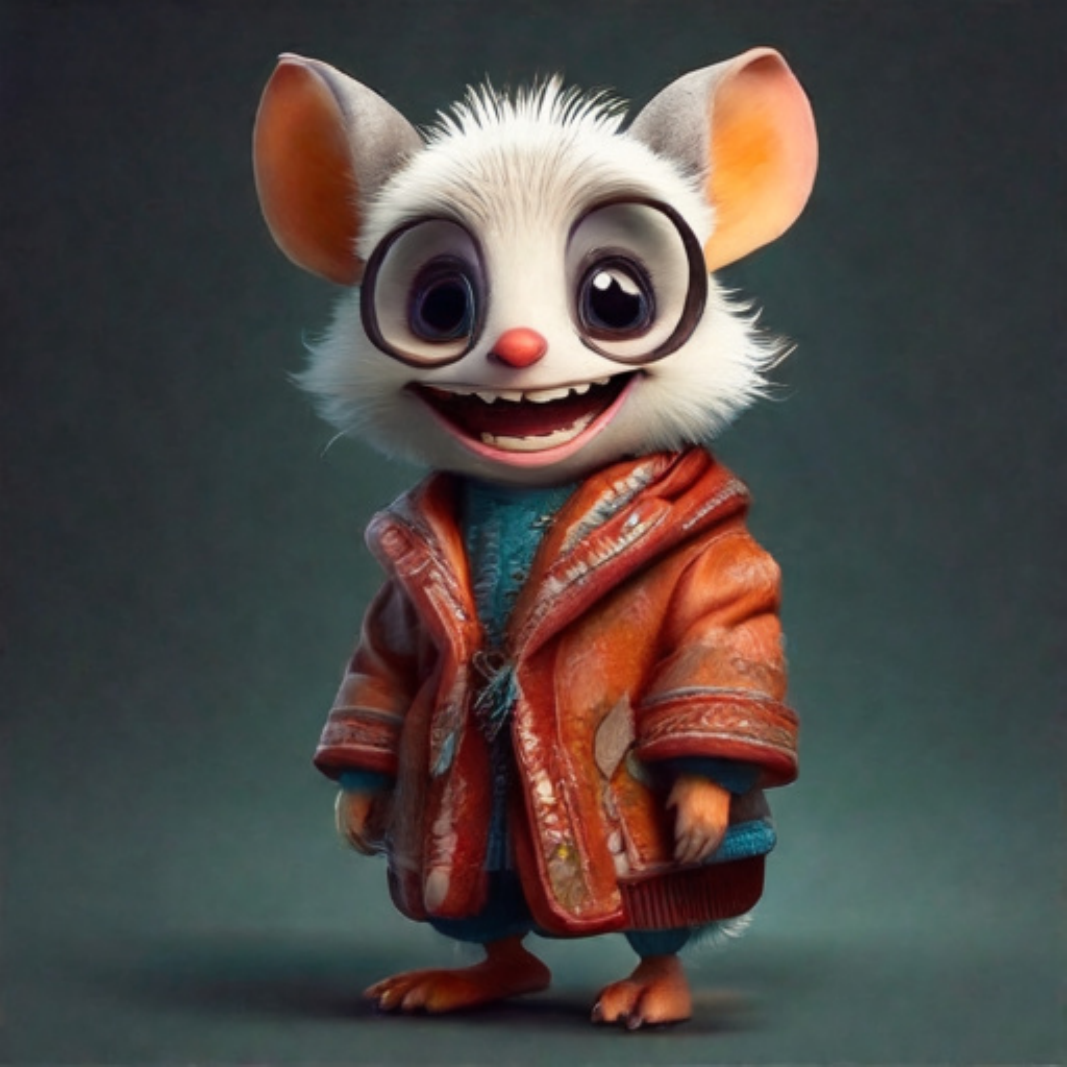}}&
\raisebox{-.5\height}{
\includegraphics[width=0.22\linewidth]{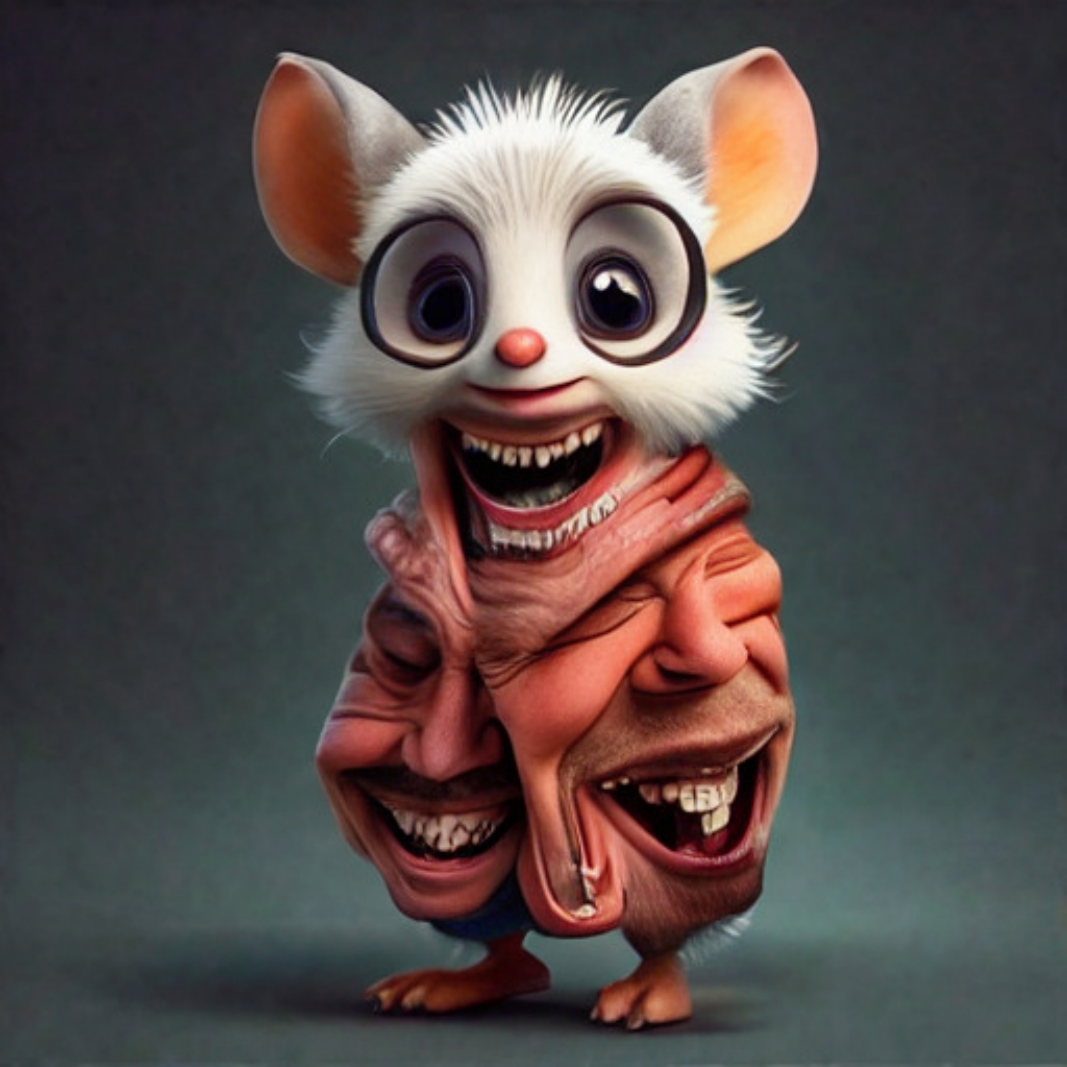 }} &
\raisebox{-.5\height}{
\includegraphics[width=0.22\linewidth]{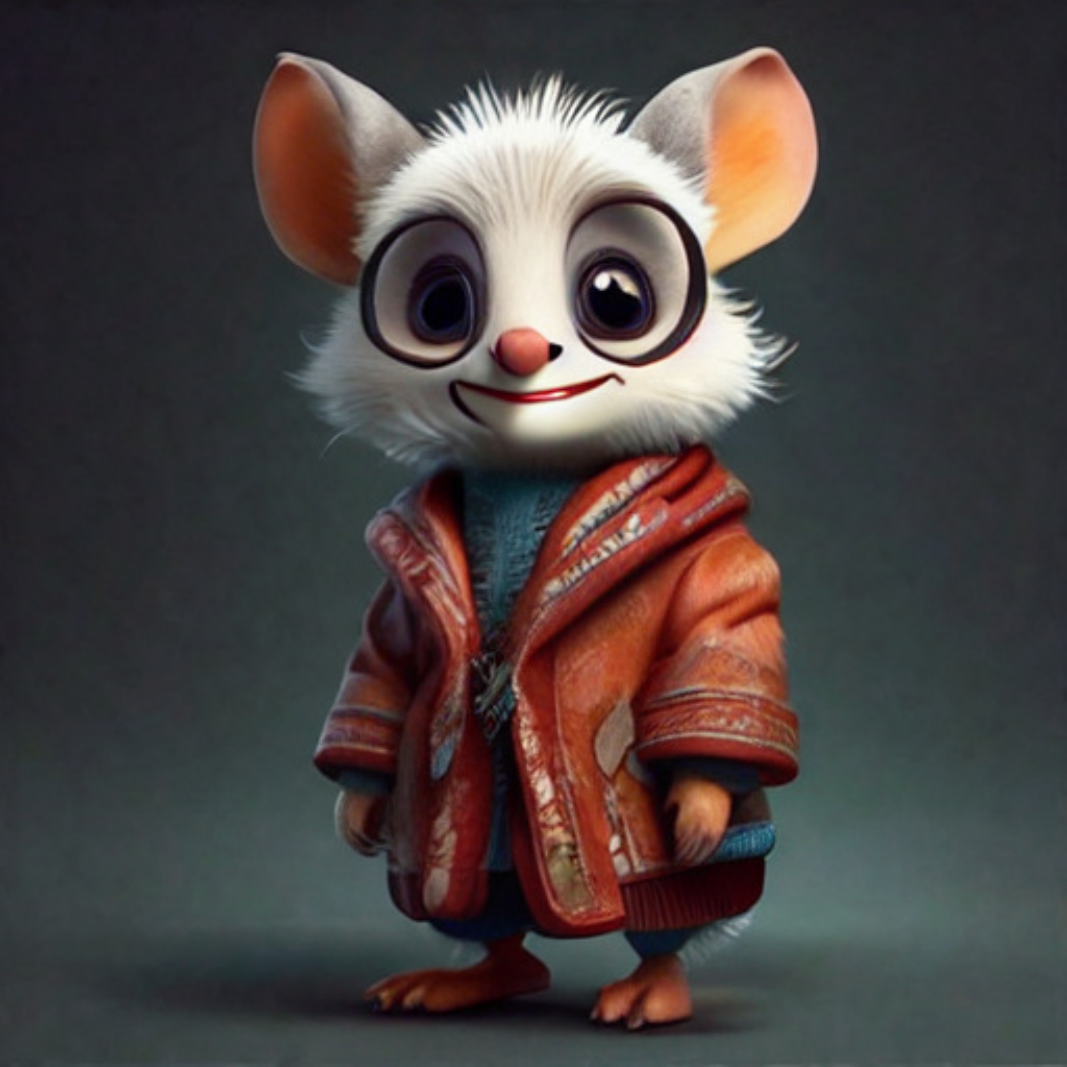}} \\ [11mm]

\resizebox{!}{16px}{
\begin{tabular}[x]{@{}c@{}}Replace\\ nose with \\chicken beak \end{tabular}}&
\raisebox{-.5\height}{
\includegraphics[width=0.22\linewidth]{imgs/comp_chosen_samples/mouse/mouse.pdf}}&
\raisebox{-.5\height}{
\includegraphics[width=0.22\linewidth]{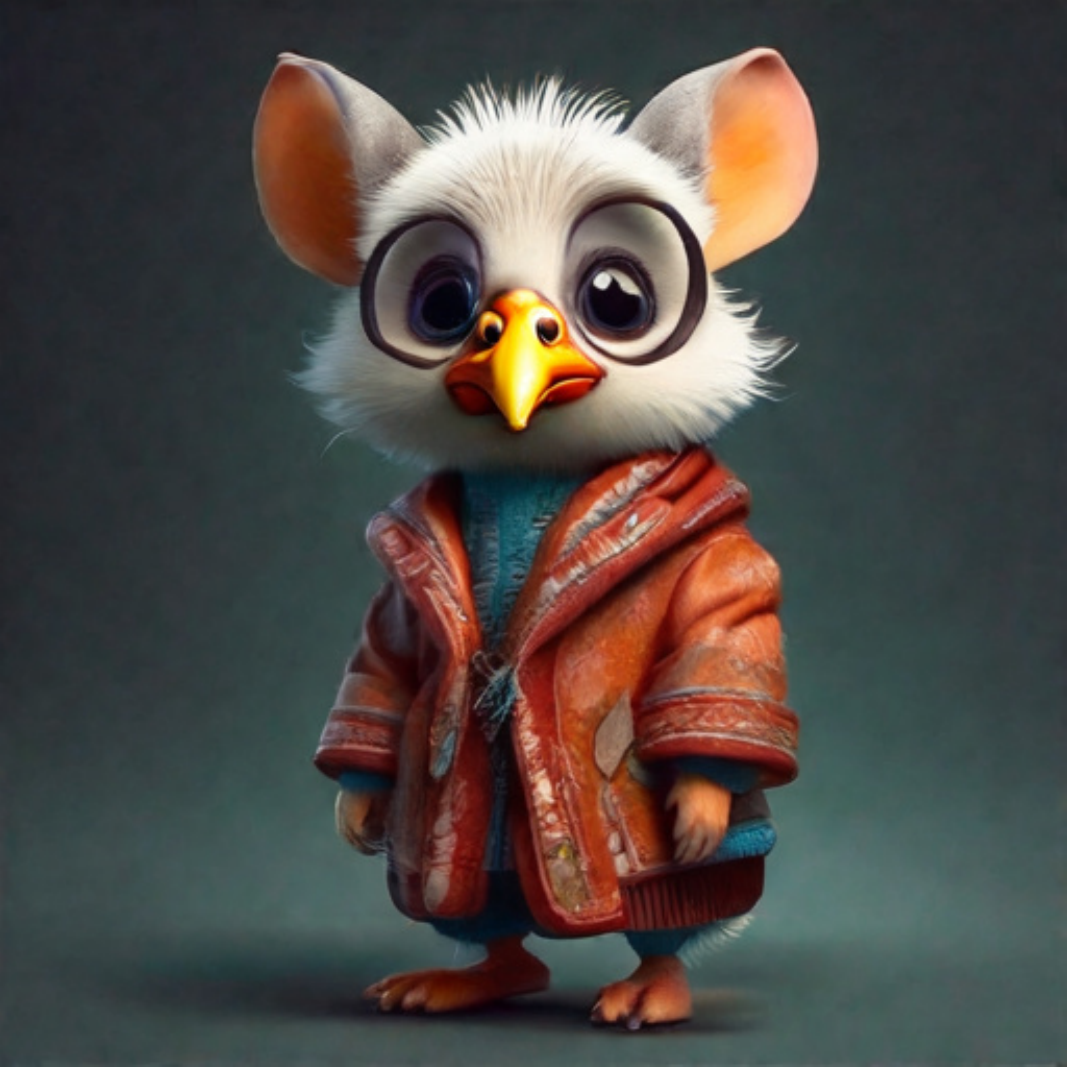}}&
\raisebox{-.5\height}{
\includegraphics[width=0.22\linewidth]{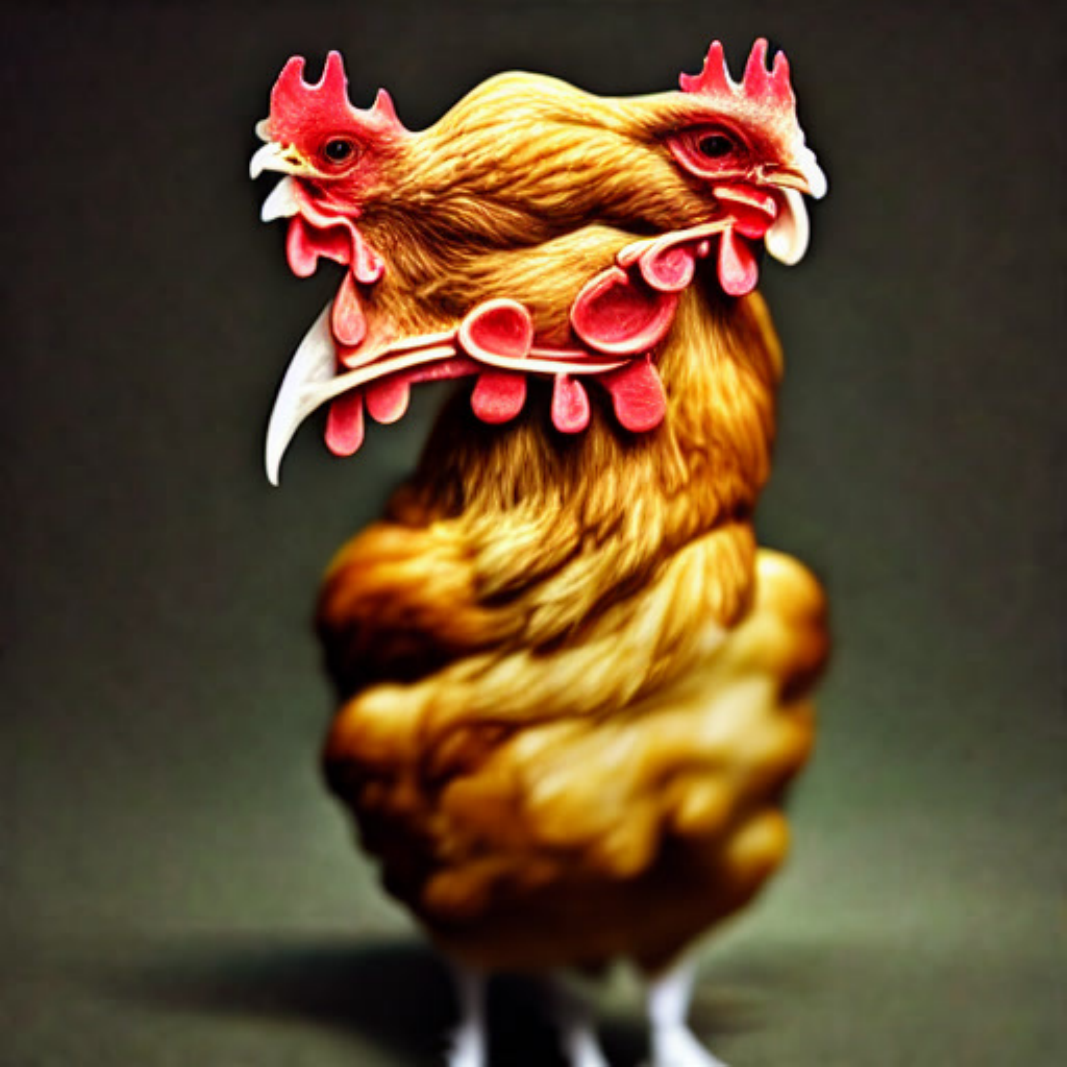 }} &
\raisebox{-.5\height}{
\includegraphics[width=0.22\linewidth]{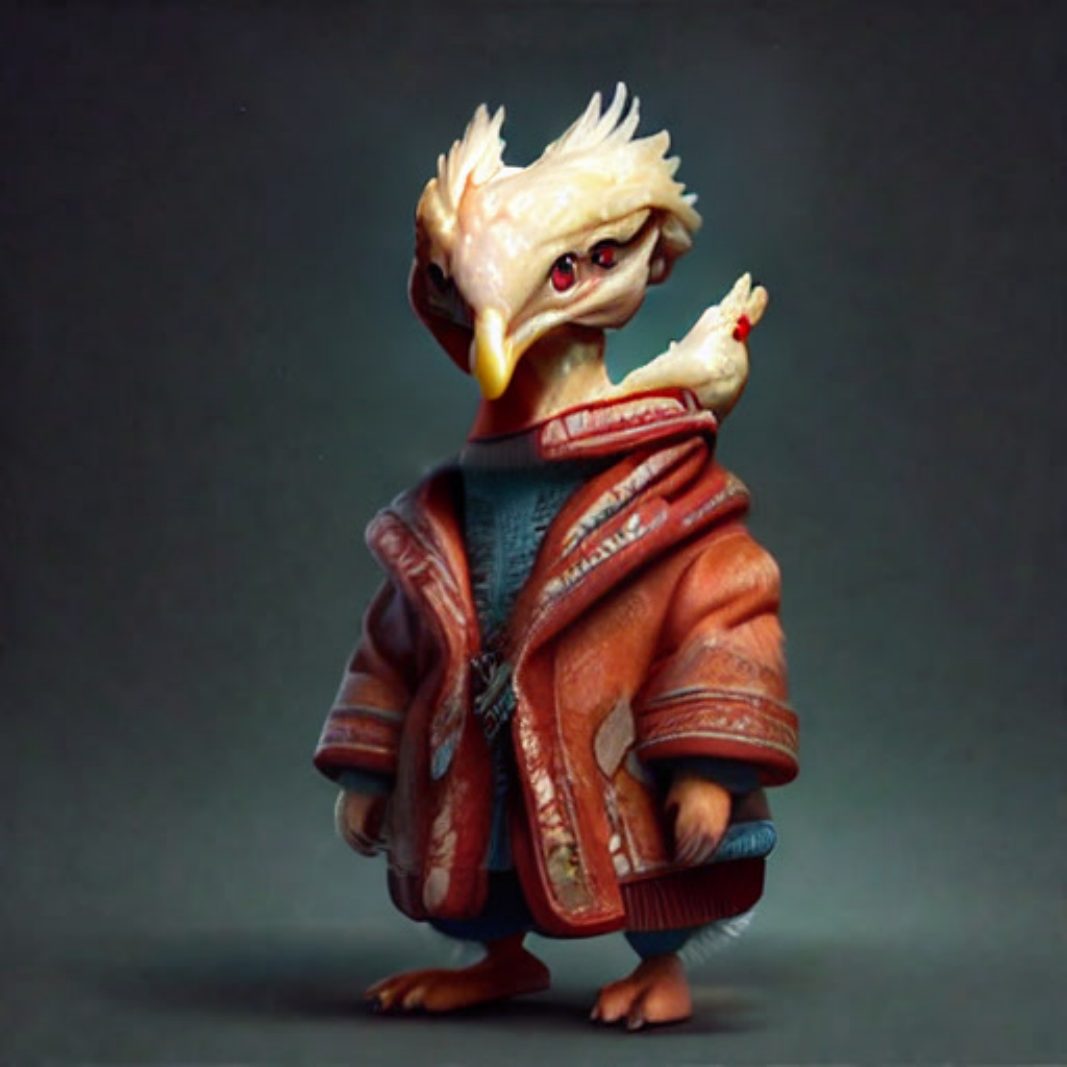}} \\ [20mm]

\resizebox{!}{20px}{
\begin{tabular}[x]{@{}c@{}}Change his \\ color to \\ linear blue \\ gradient \end{tabular}}&
\raisebox{-.5\height}{
\includegraphics[width=0.22\linewidth]{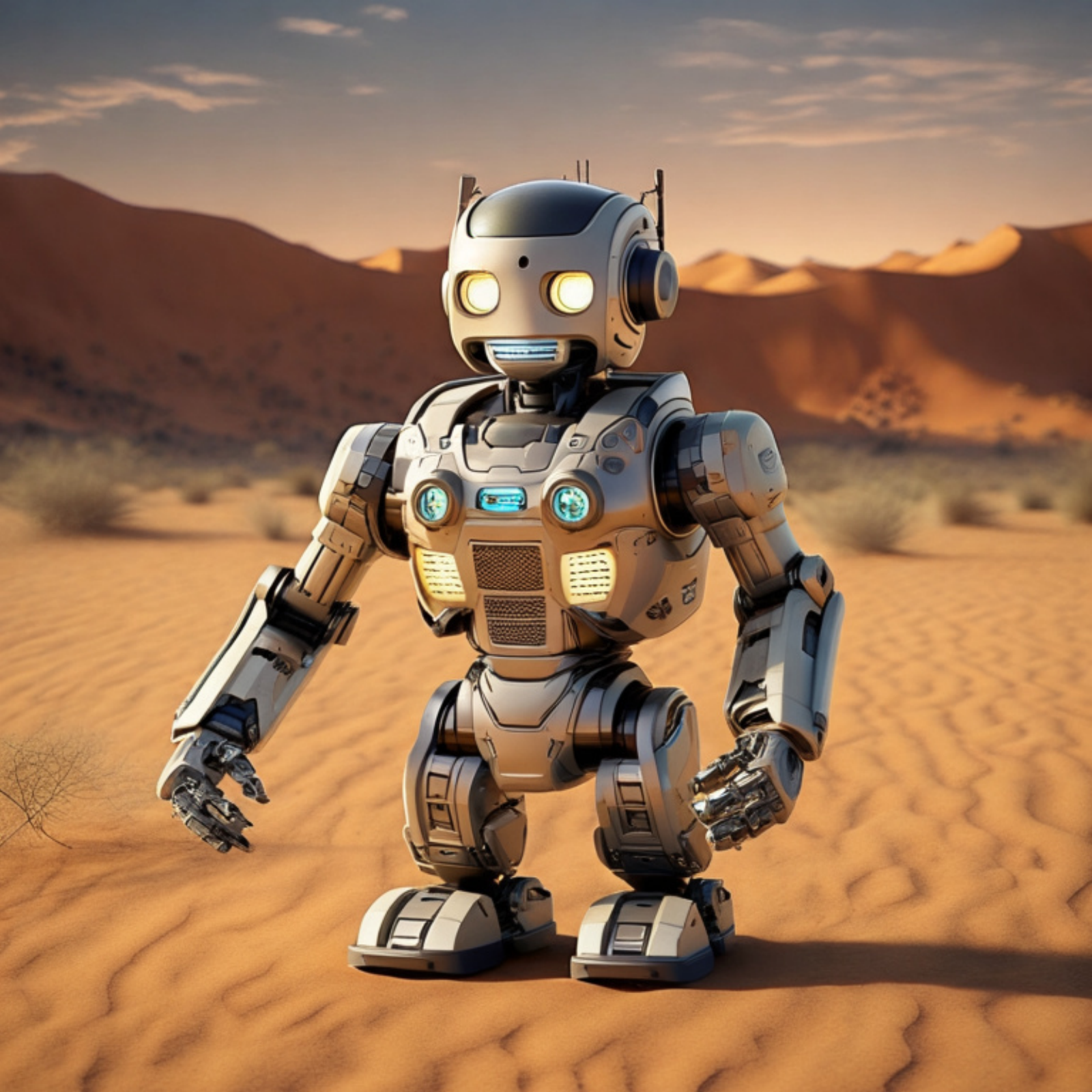}}&
\raisebox{-.5\height}{
\includegraphics[width=0.22\linewidth]{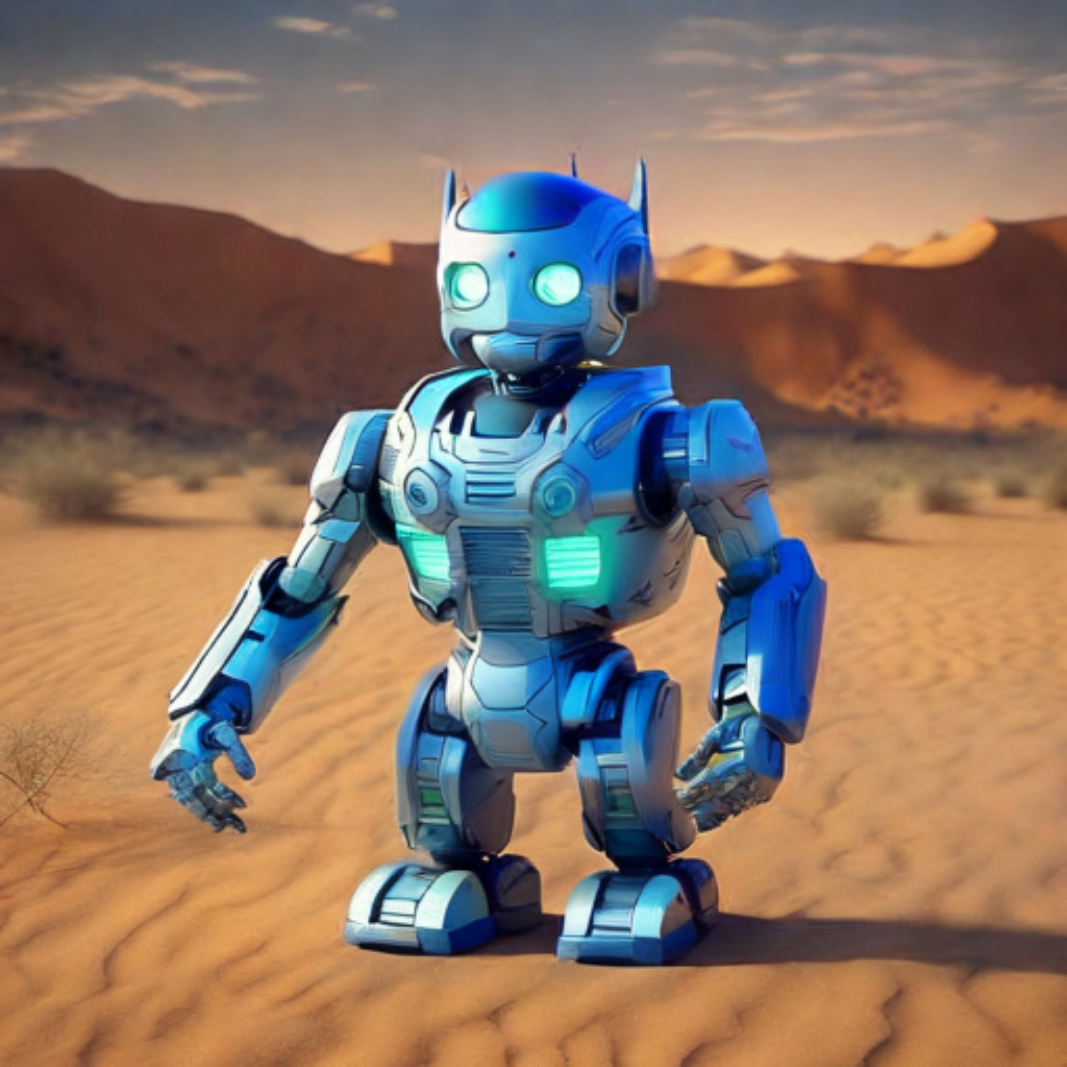}}&
\raisebox{-.5\height}{
\includegraphics[width=0.22\linewidth]{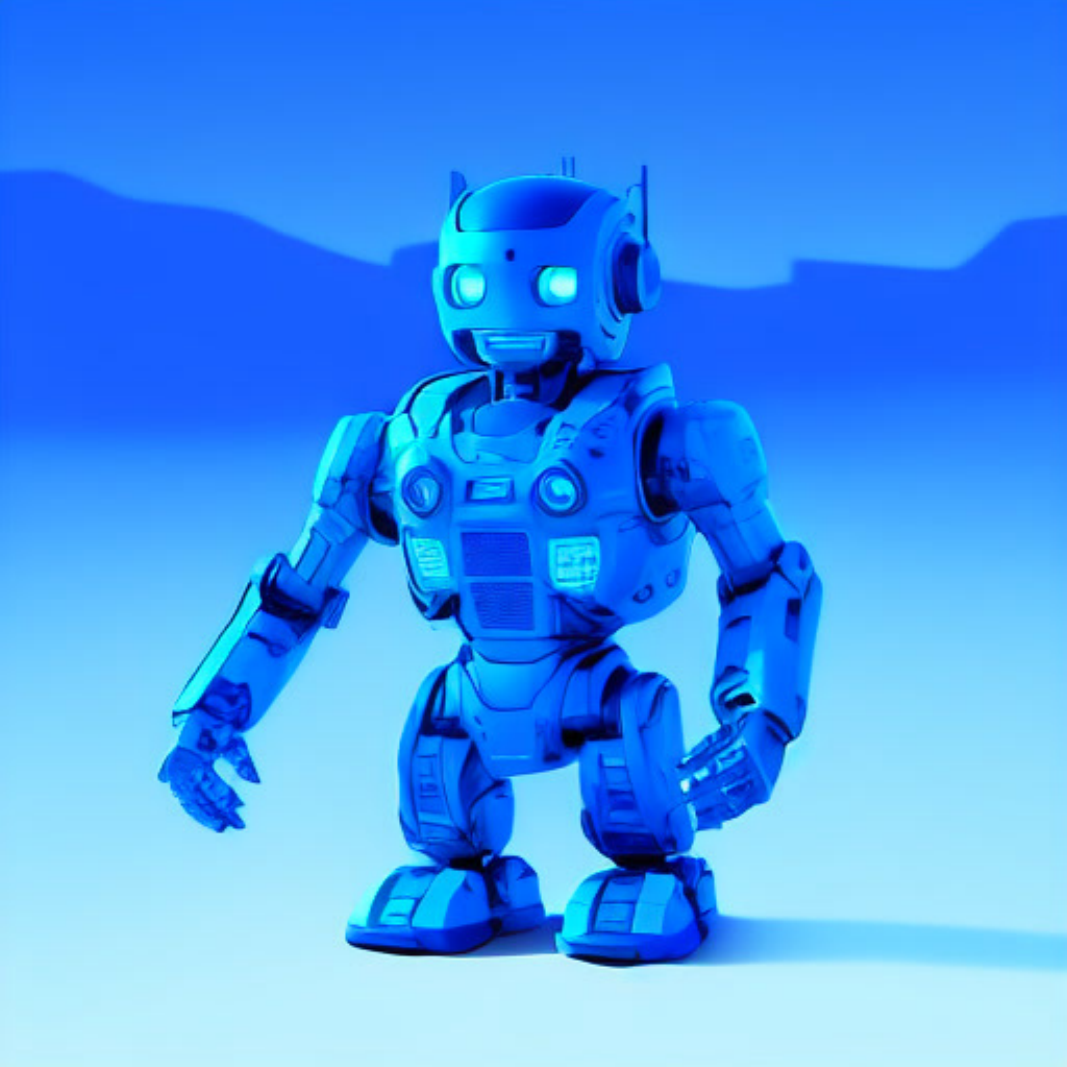 }} &
\raisebox{-.5\height}{
\includegraphics[width=0.22\linewidth]{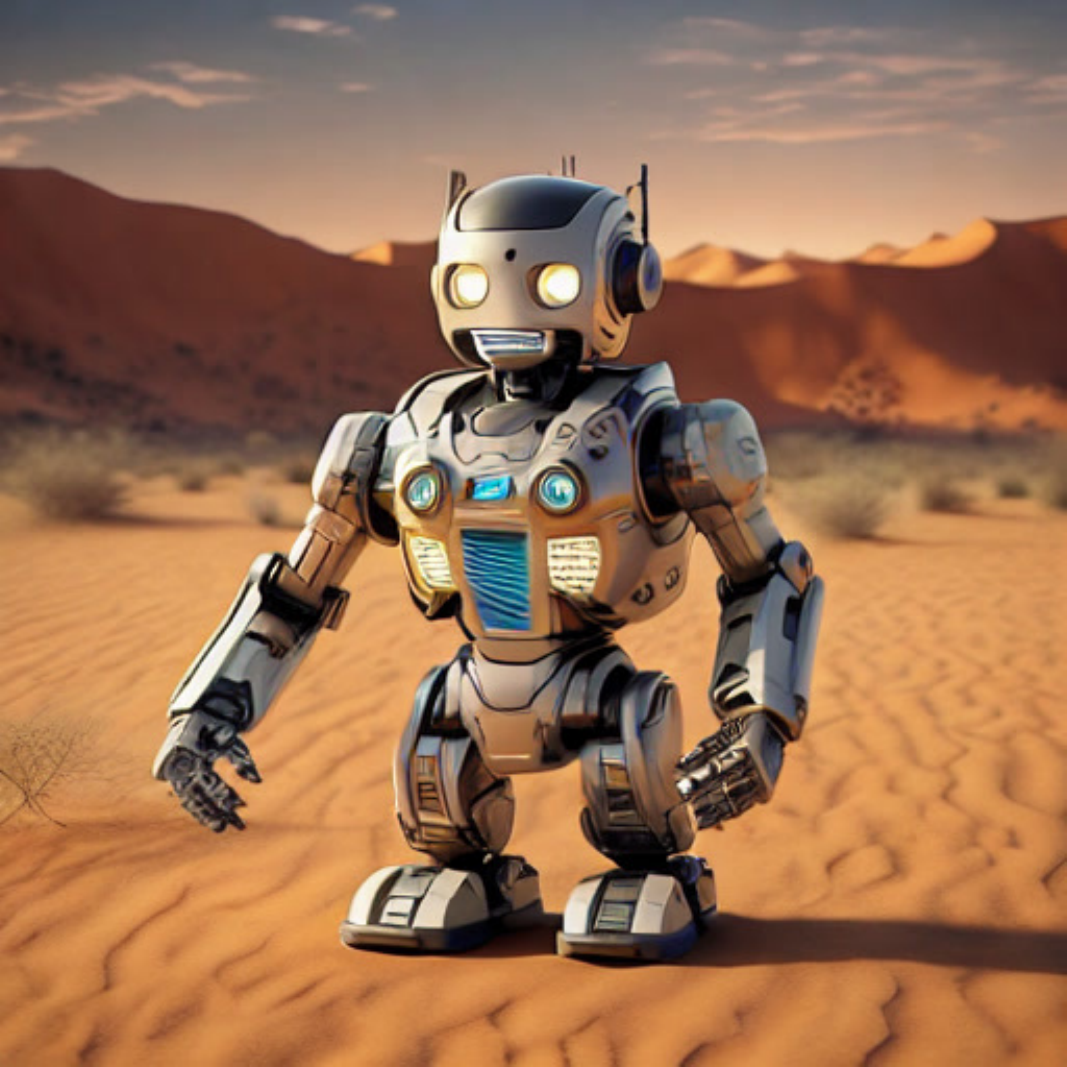}} \\ [11mm]

\resizebox{!}{16px}{
\begin{tabular}[x]{@{}c@{}}Add \\ him \\ wings \end{tabular}}&
\raisebox{-.5\height}{
\includegraphics[width=0.22\linewidth]{imgs/comp_chosen_samples/robot/robot.pdf}}&
\raisebox{-.5\height}{
\includegraphics[width=0.22\linewidth]{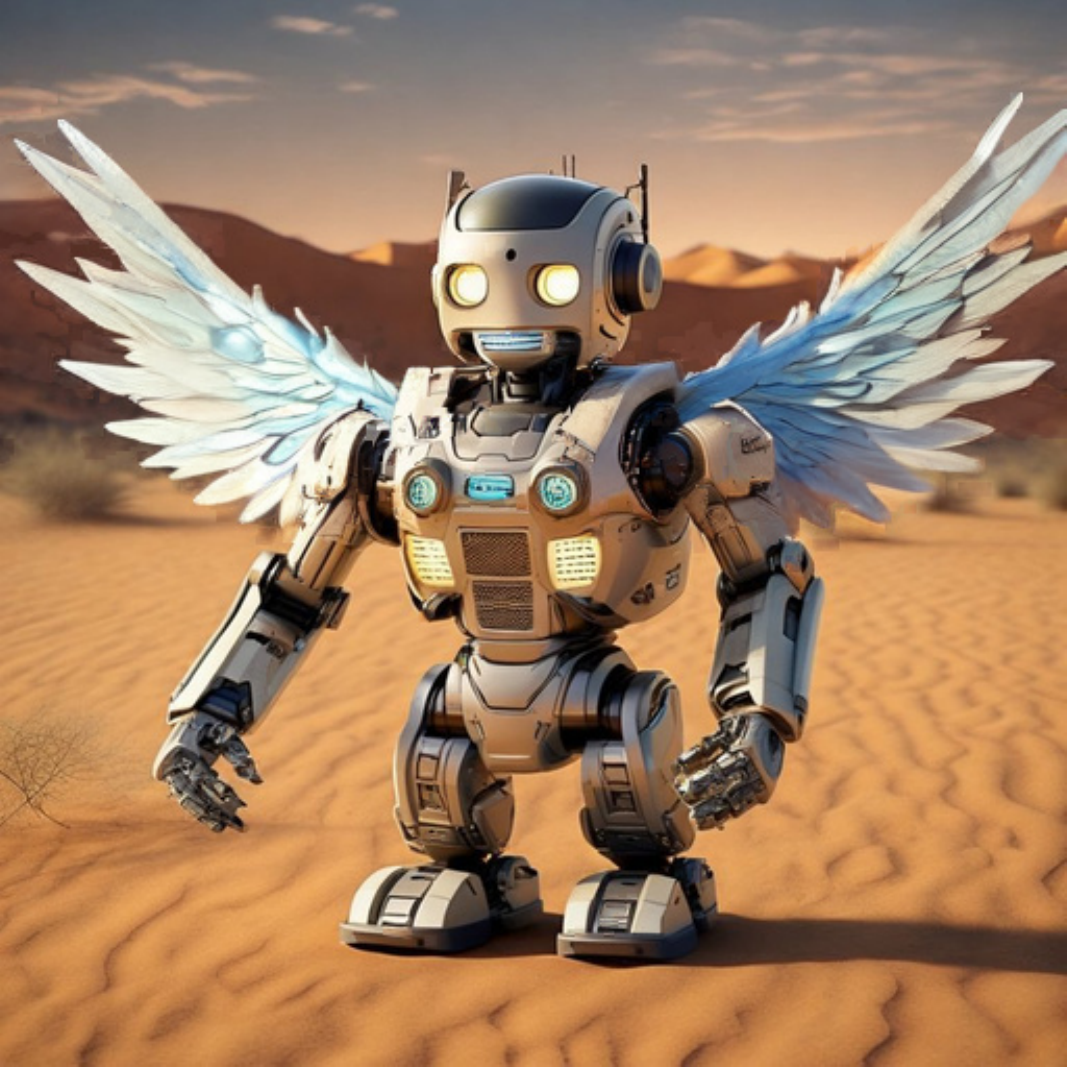}}&
\raisebox{-.5\height}{
\includegraphics[width=0.22\linewidth]{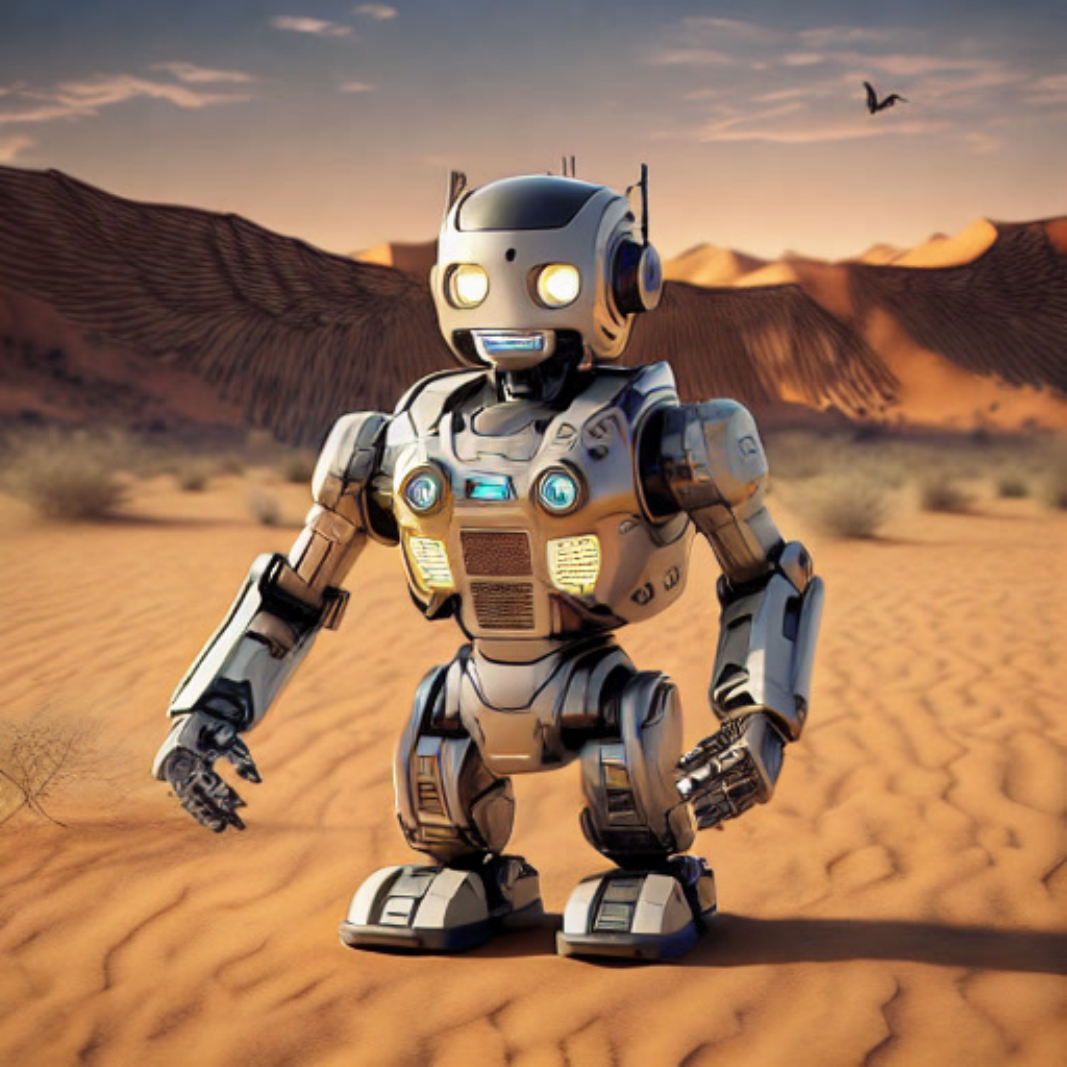 }} &
\raisebox{-.5\height}{
\includegraphics[width=0.22\linewidth]{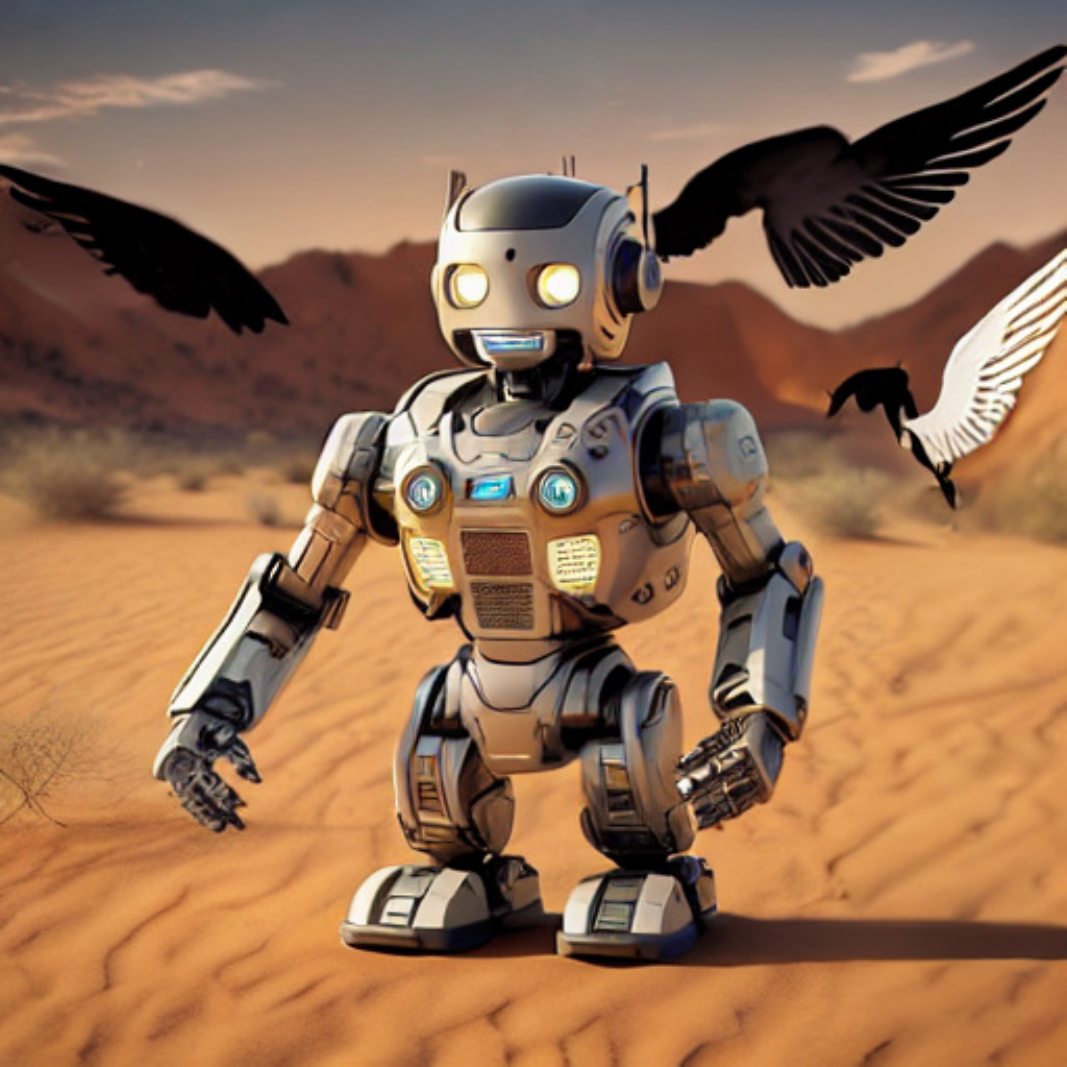}} \\ [11mm]

\resizebox{!}{20px}{
\begin{tabular}[x]{@{}c@{}}Set the \\ background \\ to VR \\ world \end{tabular}}&
\raisebox{-.5\height}{
\includegraphics[width=0.22\linewidth]{imgs/comp_chosen_samples/robot/robot.pdf}}&
\raisebox{-.5\height}{
\includegraphics[width=0.22\linewidth]{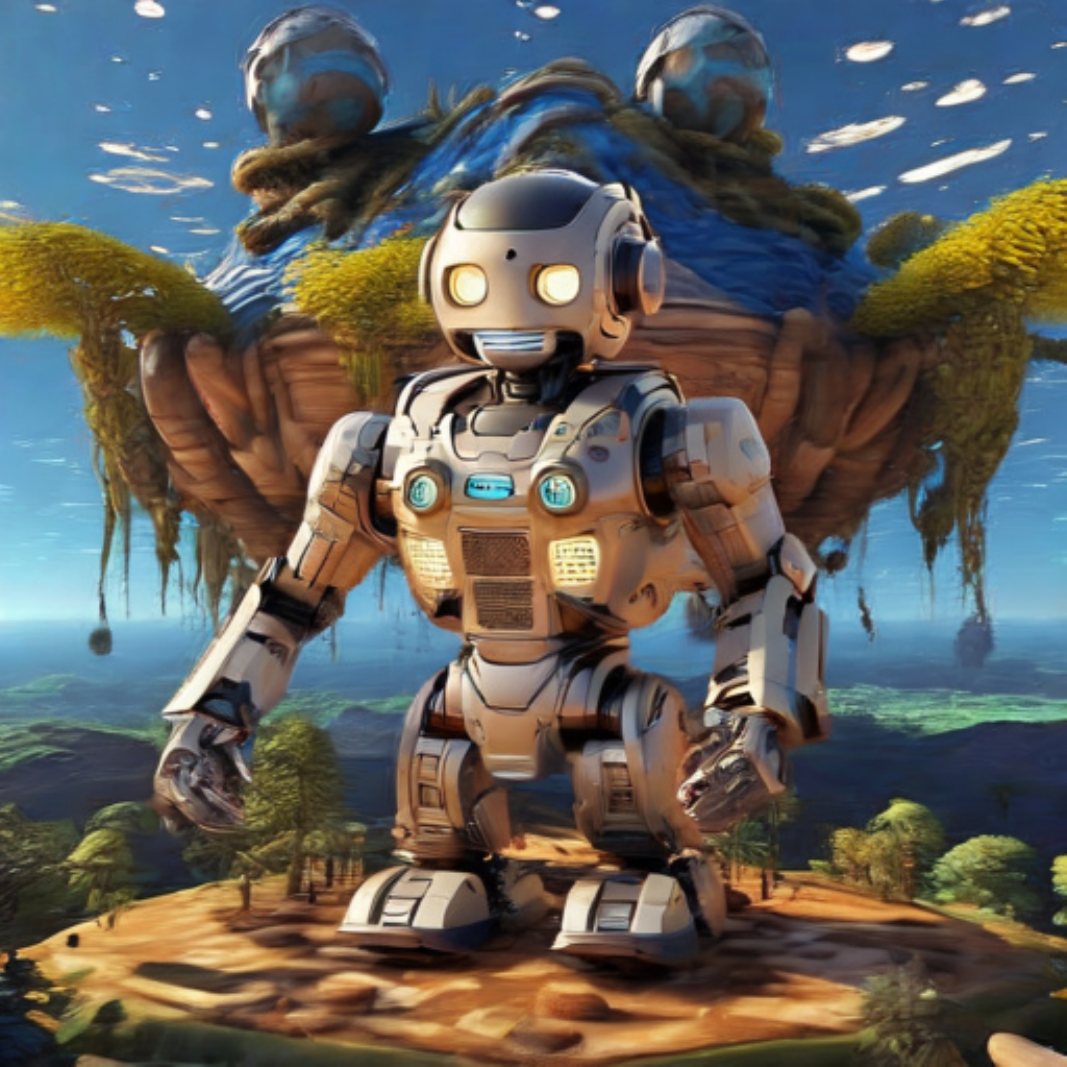}}&
\raisebox{-.5\height}{
\includegraphics[width=0.22\linewidth]{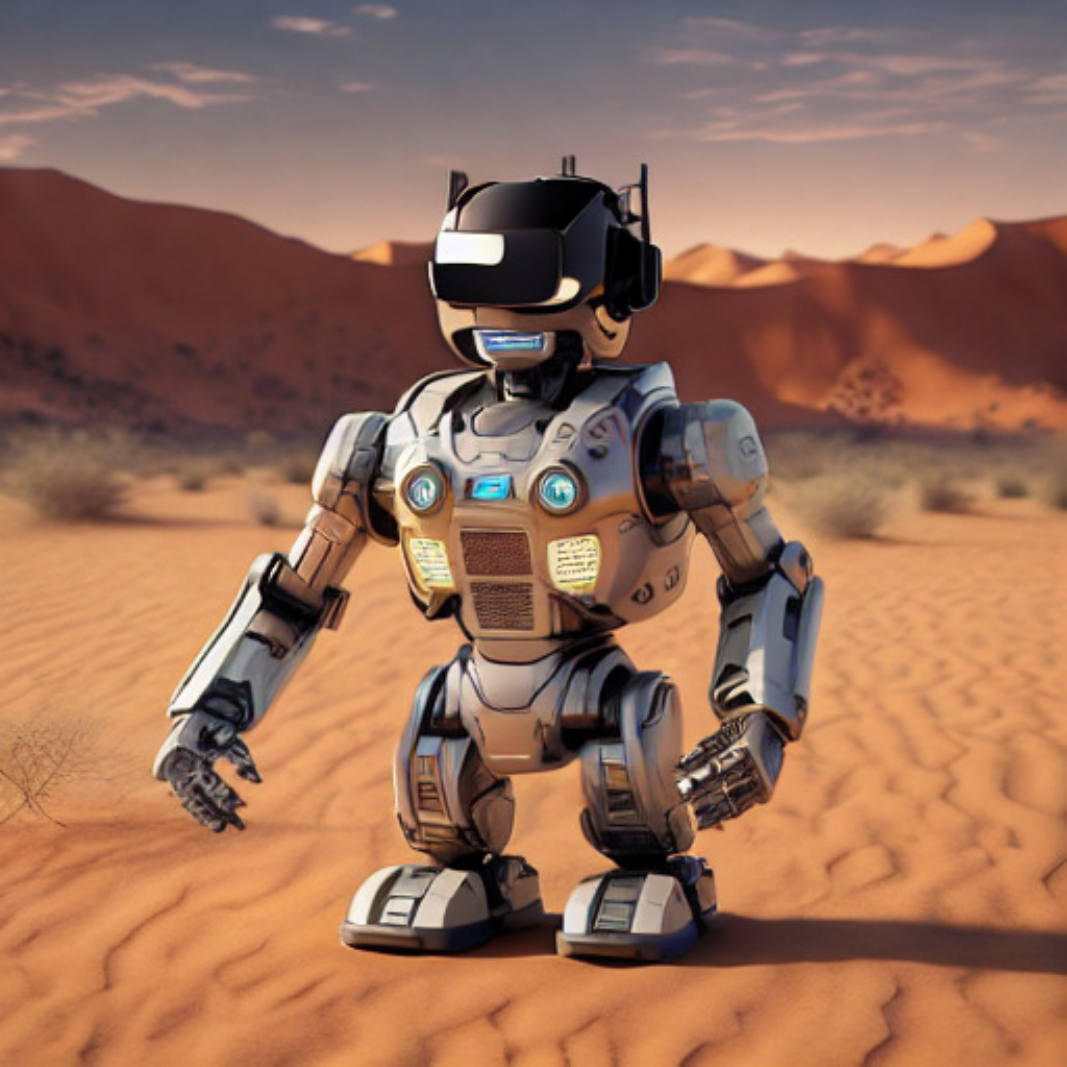 }} &
\raisebox{-.5\height}{
\includegraphics[width=0.22\linewidth]{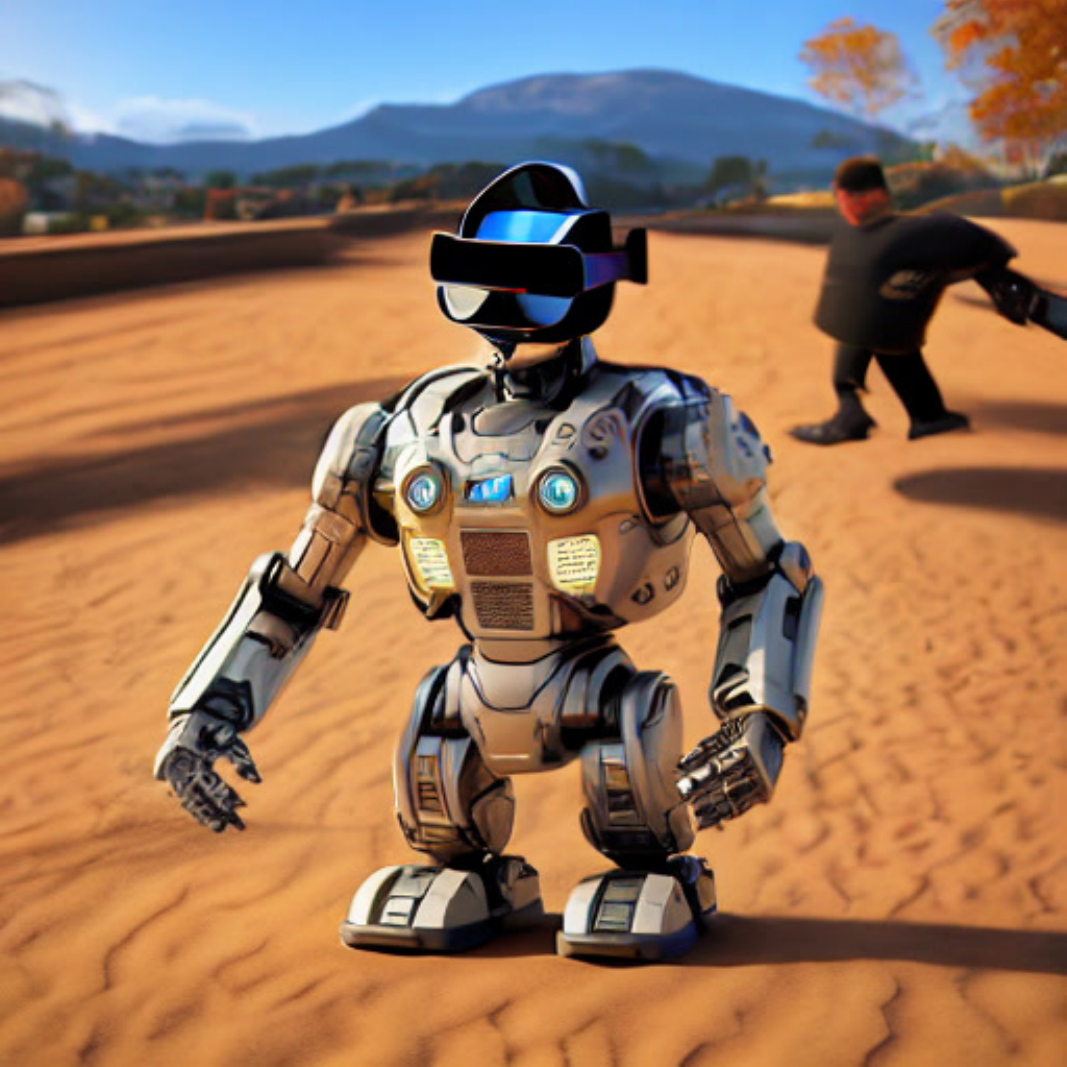}} \\ 

\end{tabular}
\caption{Qualitative comparison with baselines.}
\label{fig:comp_our_images_sup} 
\end{figure*}

\begin{figure*}[t]
   \centering
\begin{tabular}{@{\hspace{-8\tabcolsep}}c@{\hspace{-0.3\tabcolsep}}c@{\hspace{-0.3\tabcolsep}}c@{\hspace{-0.3\tabcolsep}}c@{\hspace{-0.3\tabcolsep}}c}
& \begin{tabular}[x]{@{}c@{}}Input \end{tabular}  & \begin{tabular}[x]{@{}c@{}} \model \end{tabular} &  \begin{tabular}[x]{@{}c@{}} InstructPix2Pix \end{tabular} & MagicBrush  \\
\resizebox{!}{16px}{
\begin{tabular}[x]{@{}c@{}} Replace Emu \\ with \\ peacock \end{tabular}}&
\raisebox{-.5\height}{
\includegraphics[width=0.22\linewidth]{imgs/comp_chosen_samples/emu/emu.pdf}}&
\raisebox{-.5\height}{
\includegraphics[width=0.22\linewidth]{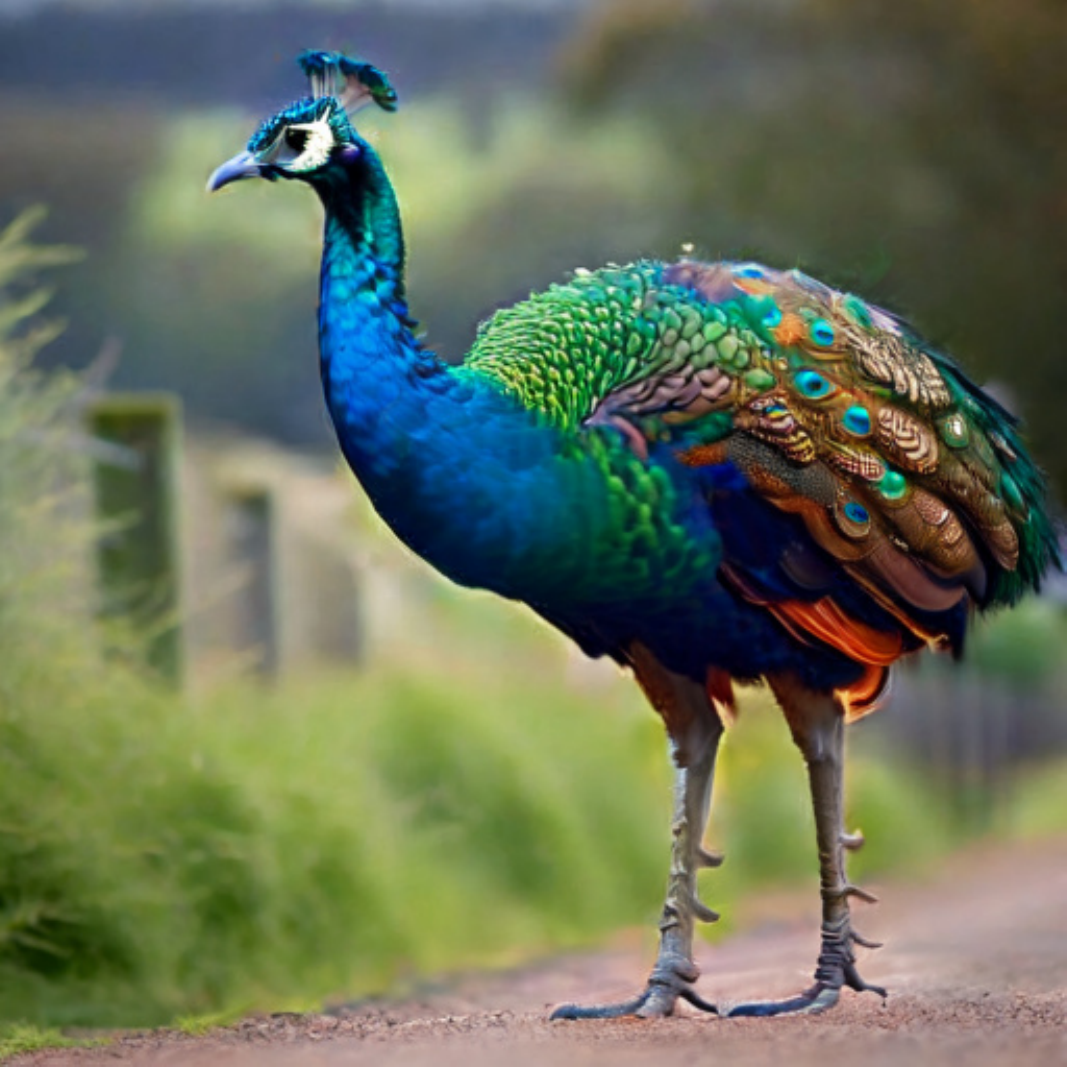}}&
\raisebox{-.5\height}{
\includegraphics[width=0.22\linewidth]{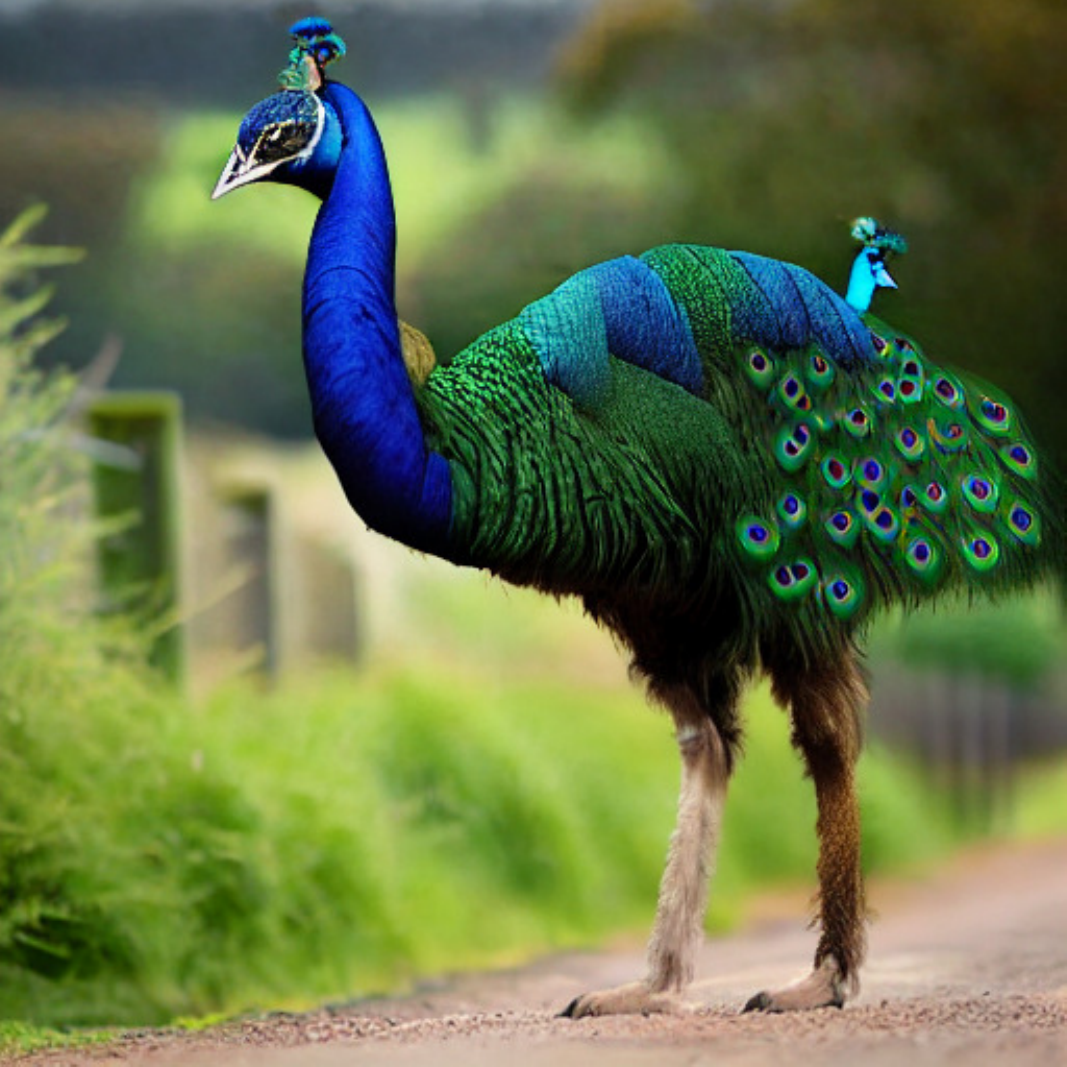 }} &
\raisebox{-.5\height}{
\includegraphics[width=0.22\linewidth]{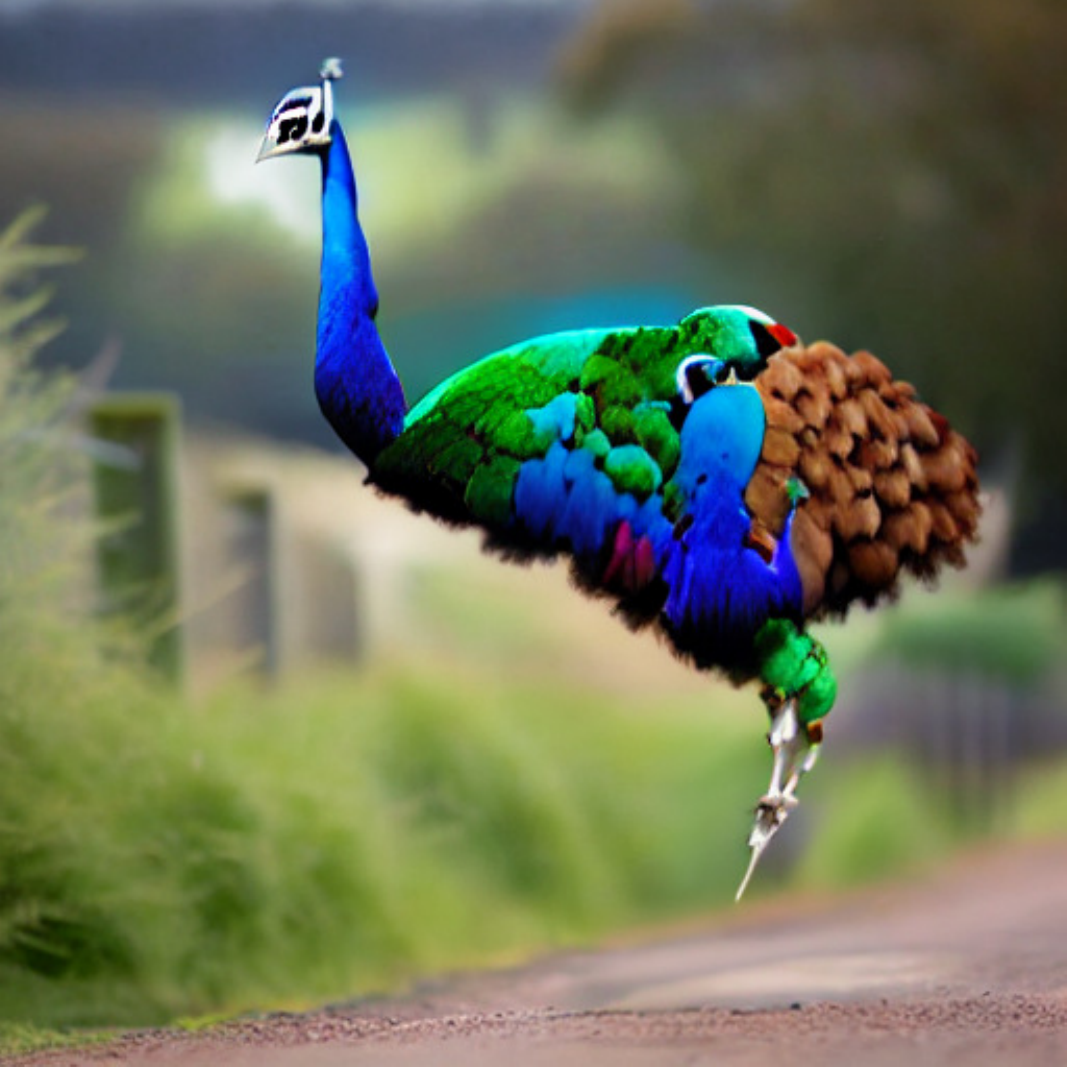}} \\ [11mm]

\resizebox{!}{16px}{
\begin{tabular}[x]{@{}c@{}}Make it \\ a Bansky \\ painting \end{tabular}}&
\raisebox{-.5\height}{
\includegraphics[width=0.22\linewidth]{imgs/comp_chosen_samples/emu/emu.pdf}}&
\raisebox{-.5\height}{
\includegraphics[width=0.22\linewidth]{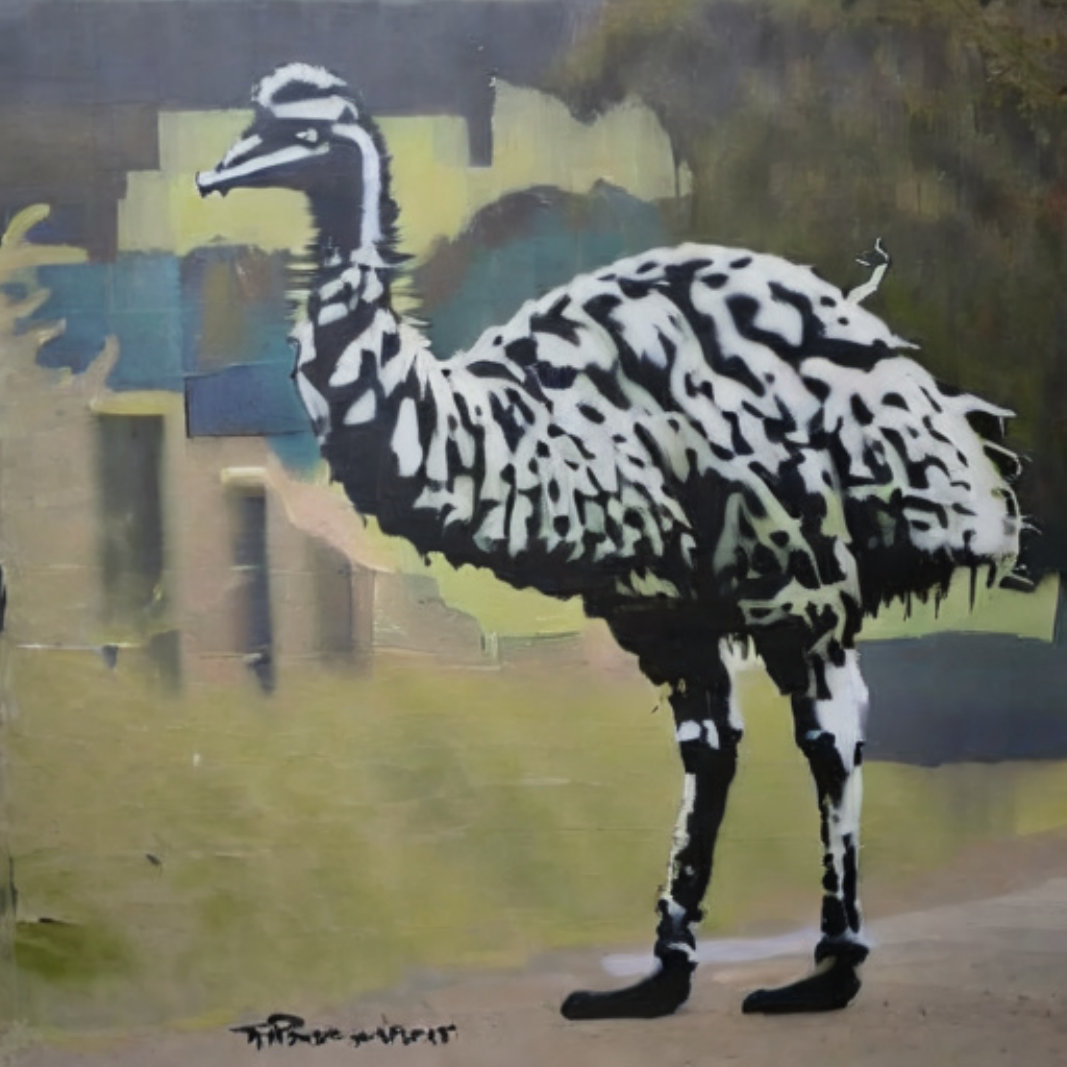}}&
\raisebox{-.5\height}{
\includegraphics[width=0.22\linewidth]{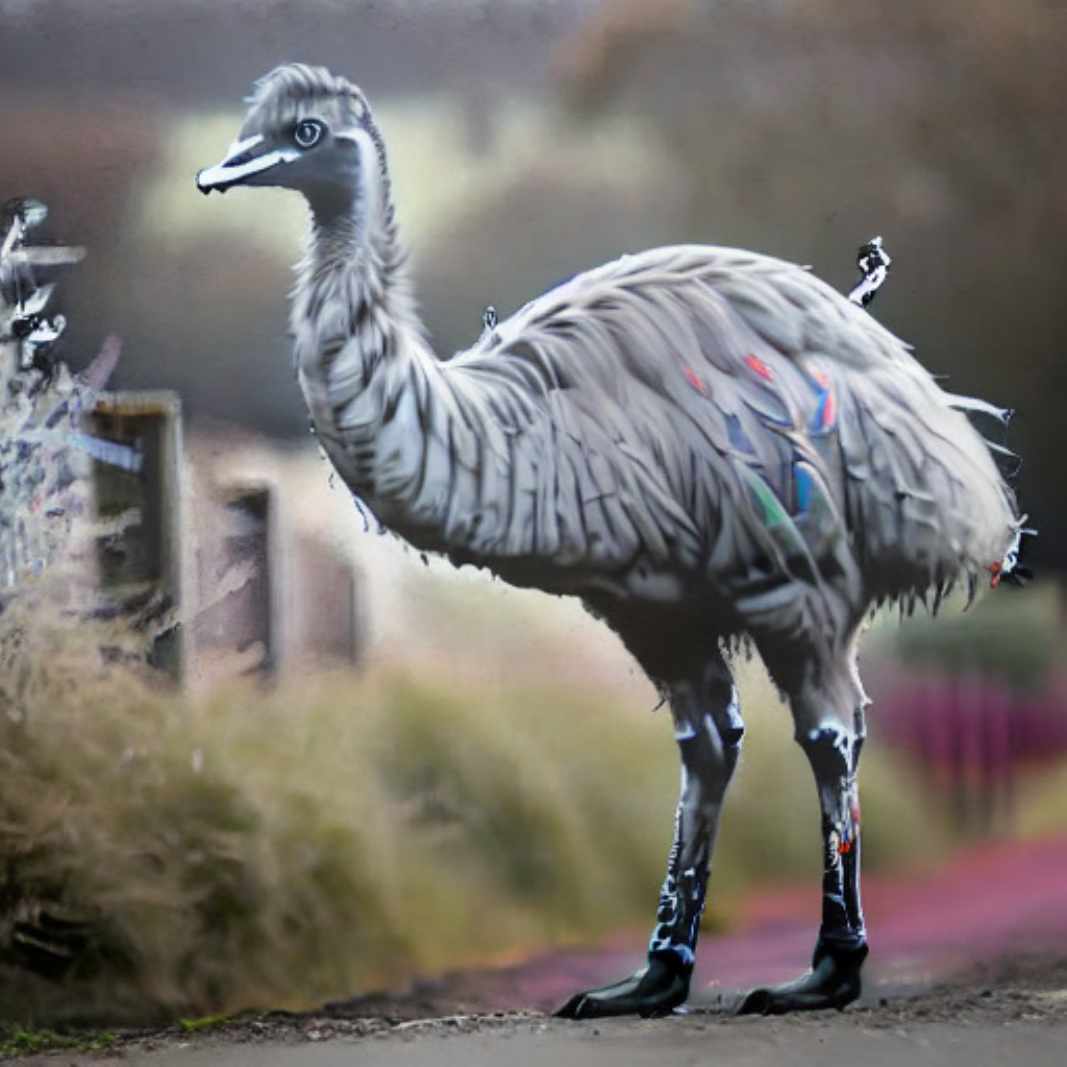 }} &
\raisebox{-.5\height}{
\includegraphics[width=0.22\linewidth]{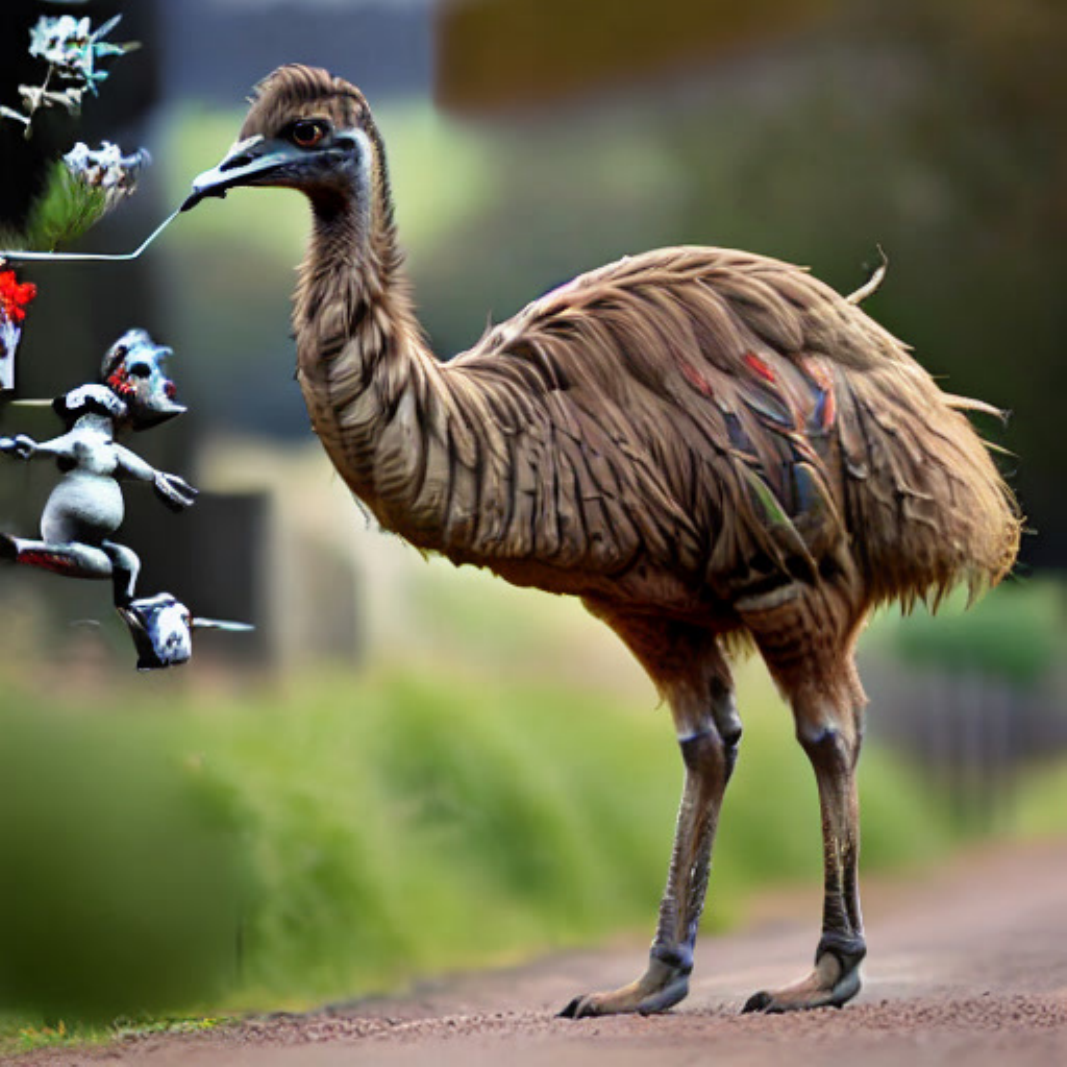}} \\ [20mm]

\resizebox{!}{16px}{
\begin{tabular}[x]{@{}c@{}} Cover the \\ house with \\ candies \end{tabular}}&
\raisebox{-.5\height}{
\includegraphics[width=0.22\linewidth]{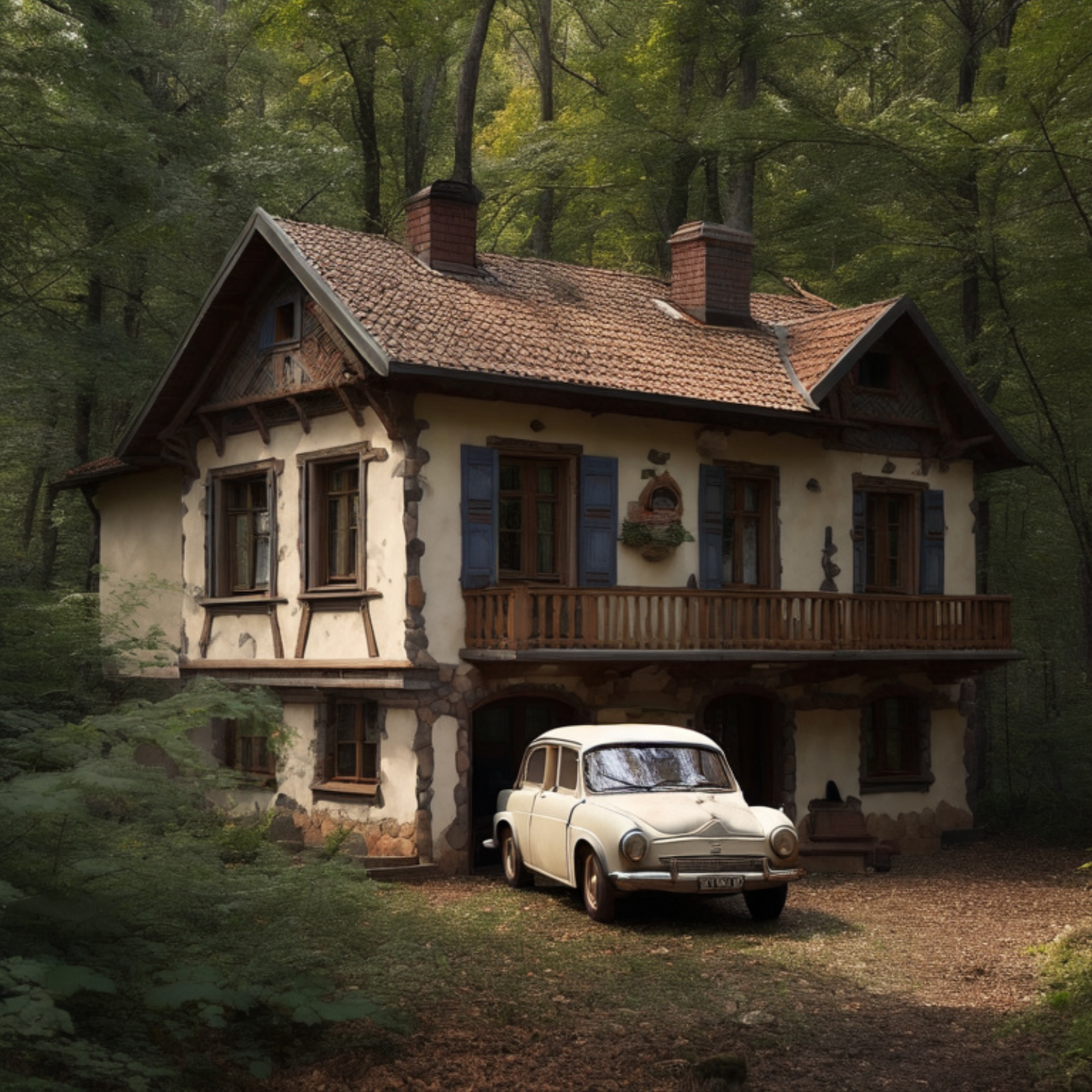}}&
\raisebox{-.5\height}{
\includegraphics[width=0.22\linewidth]{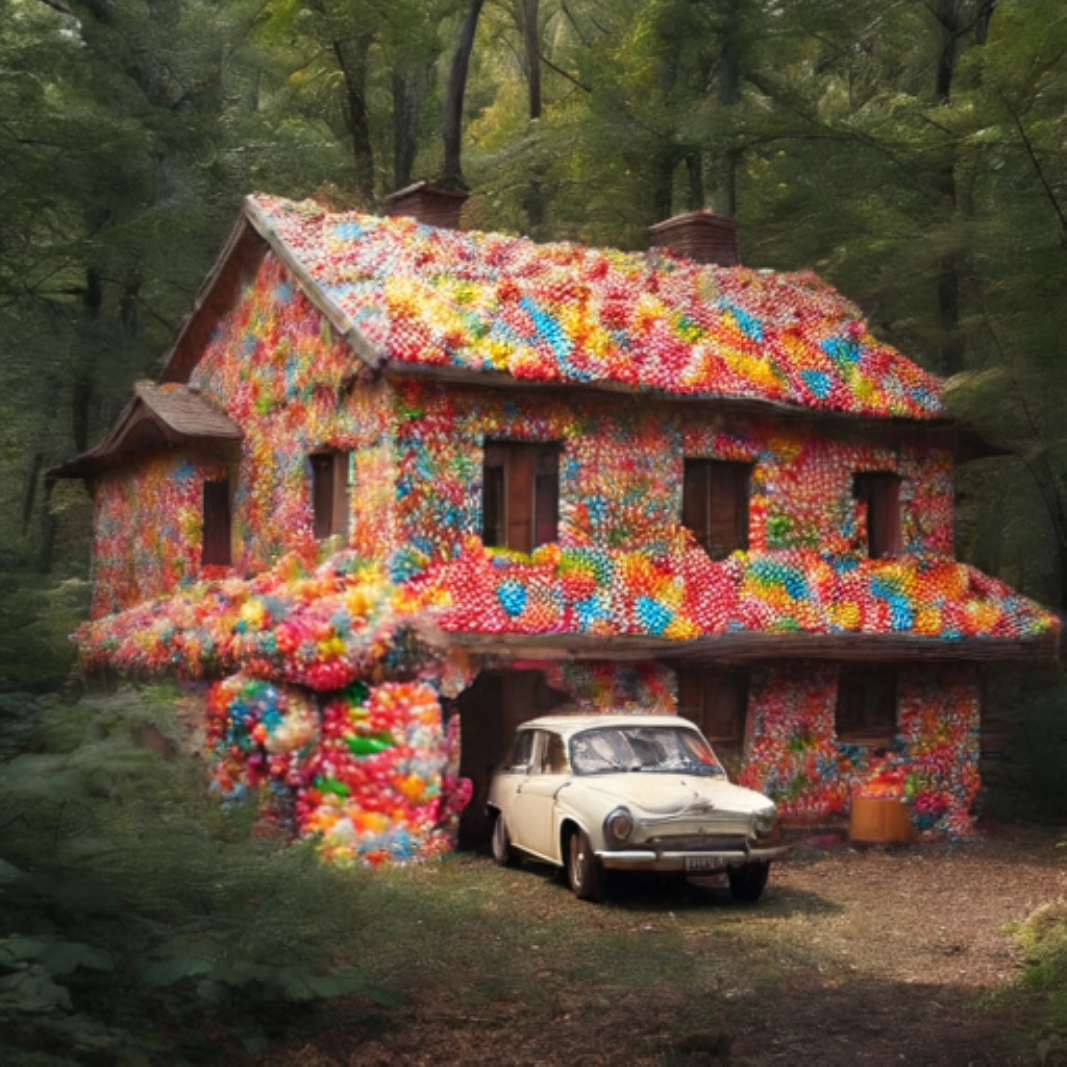}}&
\raisebox{-.5\height}{
\includegraphics[width=0.22\linewidth]{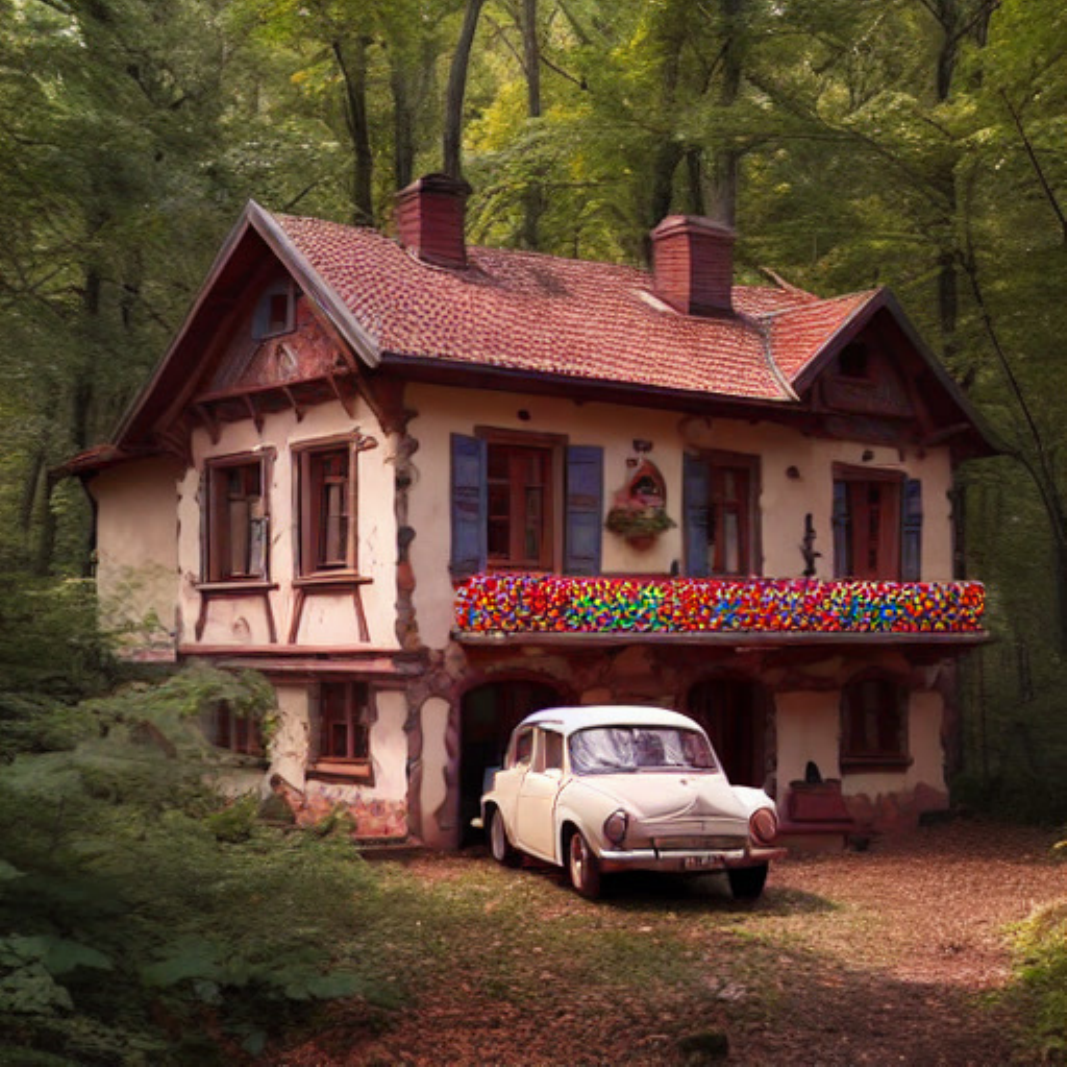 }} &
\raisebox{-.5\height}{
\includegraphics[width=0.22\linewidth]{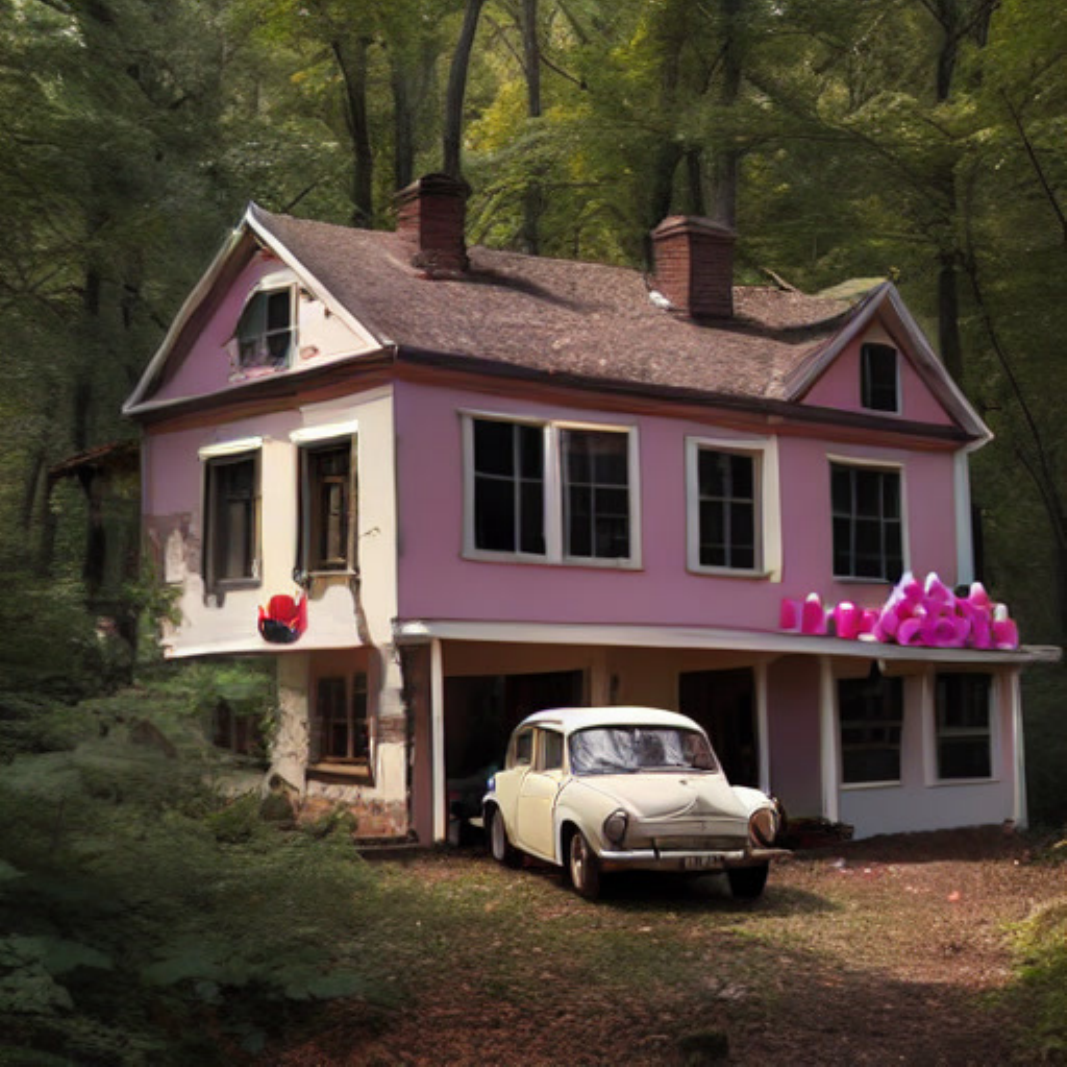}} \\ [11mm]

\resizebox{!}{16px}{
\begin{tabular}[x]{@{}c@{}} Remove \\ the \\ car \end{tabular}}&
\raisebox{-.5\height}{
\includegraphics[width=0.22\linewidth]{imgs/comp_chosen_samples/house/house.pdf}}&
\raisebox{-.5\height}{
\includegraphics[width=0.22\linewidth]{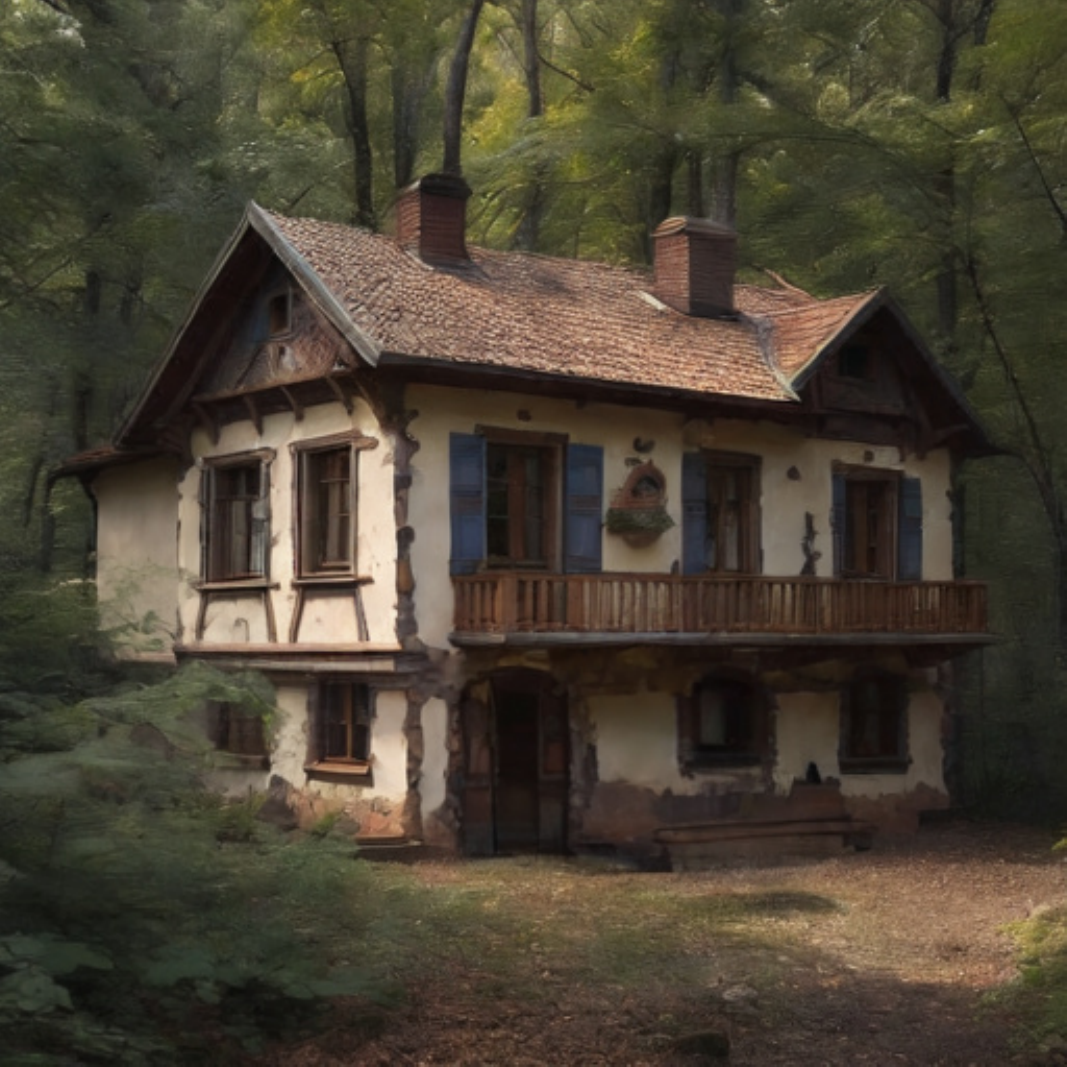}}&
\raisebox{-.5\height}{
\includegraphics[width=0.22\linewidth]{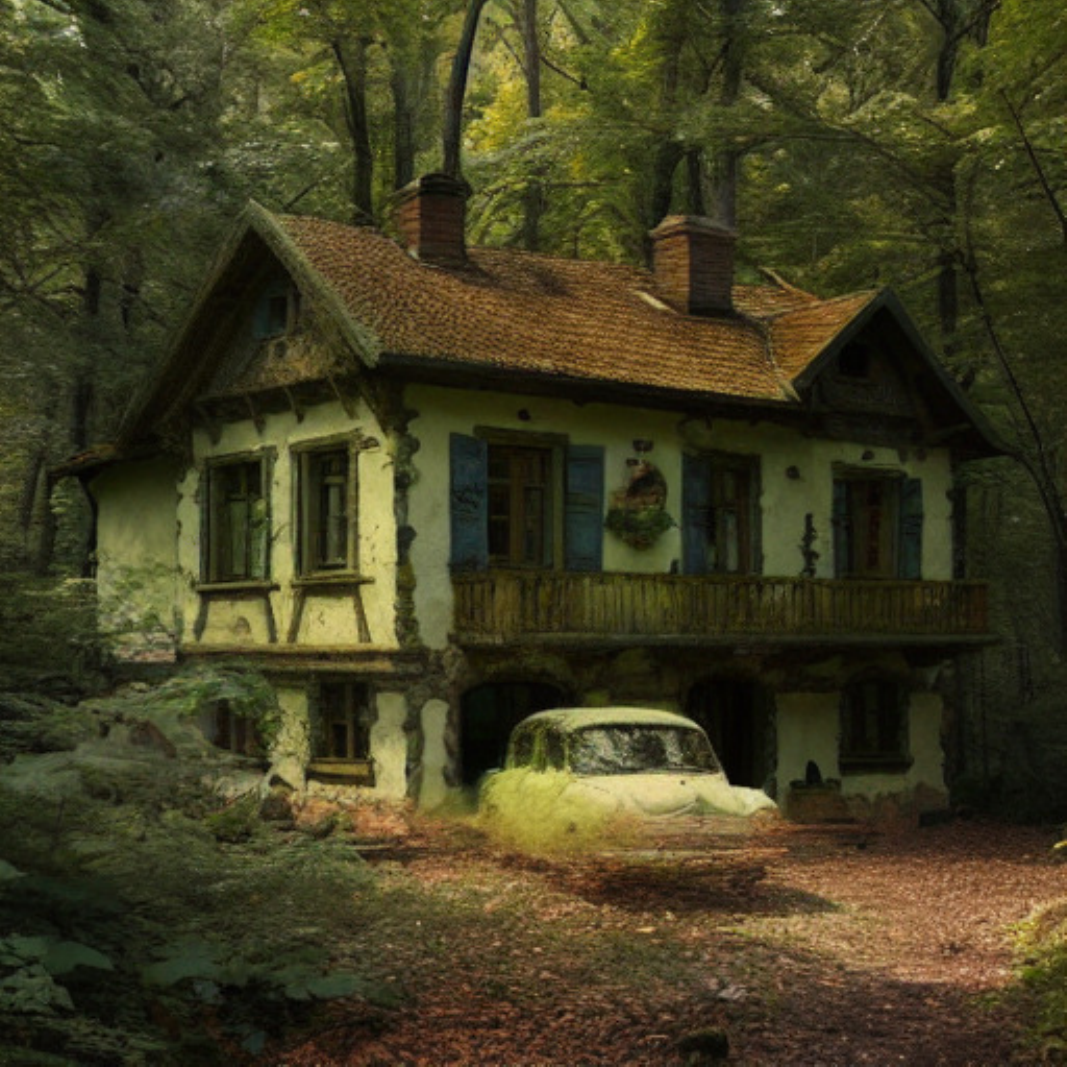 }} &
\raisebox{-.5\height}{
\includegraphics[width=0.22\linewidth]{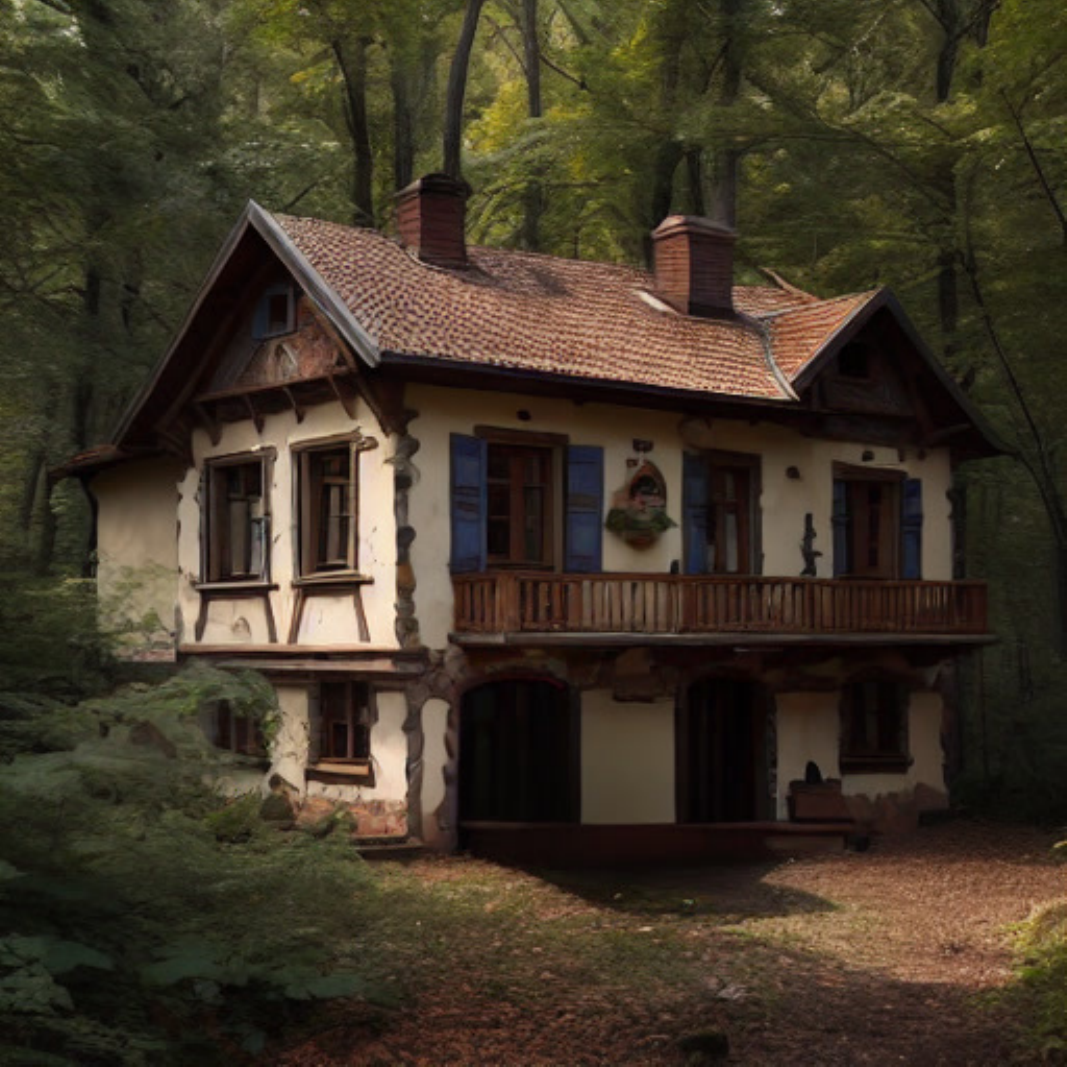}} 

\end{tabular}
\caption{Qualitative comparison with baselines.}
\label{fig:comp_our_images_sup2} 
\end{figure*}

\begin{figure*}[h]
\centering
\scalebox{0.45}{
\begin{tabular}{@{\hspace{-15\tabcolsep}}c@{\hspace{0.1\tabcolsep}}c@{\hspace{0.5\tabcolsep}}c@{\hspace{0.2\tabcolsep}}c@{\hspace{0.2\tabcolsep}}c@{\hspace{0.2\tabcolsep}}c}

\resizebox{!}{40px}{
\begin{tabular}[x]{@{}c@{}}Let the \\ keyboard \\ be yellow \end{tabular}}&
\raisebox{-.5\height}{
\includegraphics[width=0.4\linewidth]{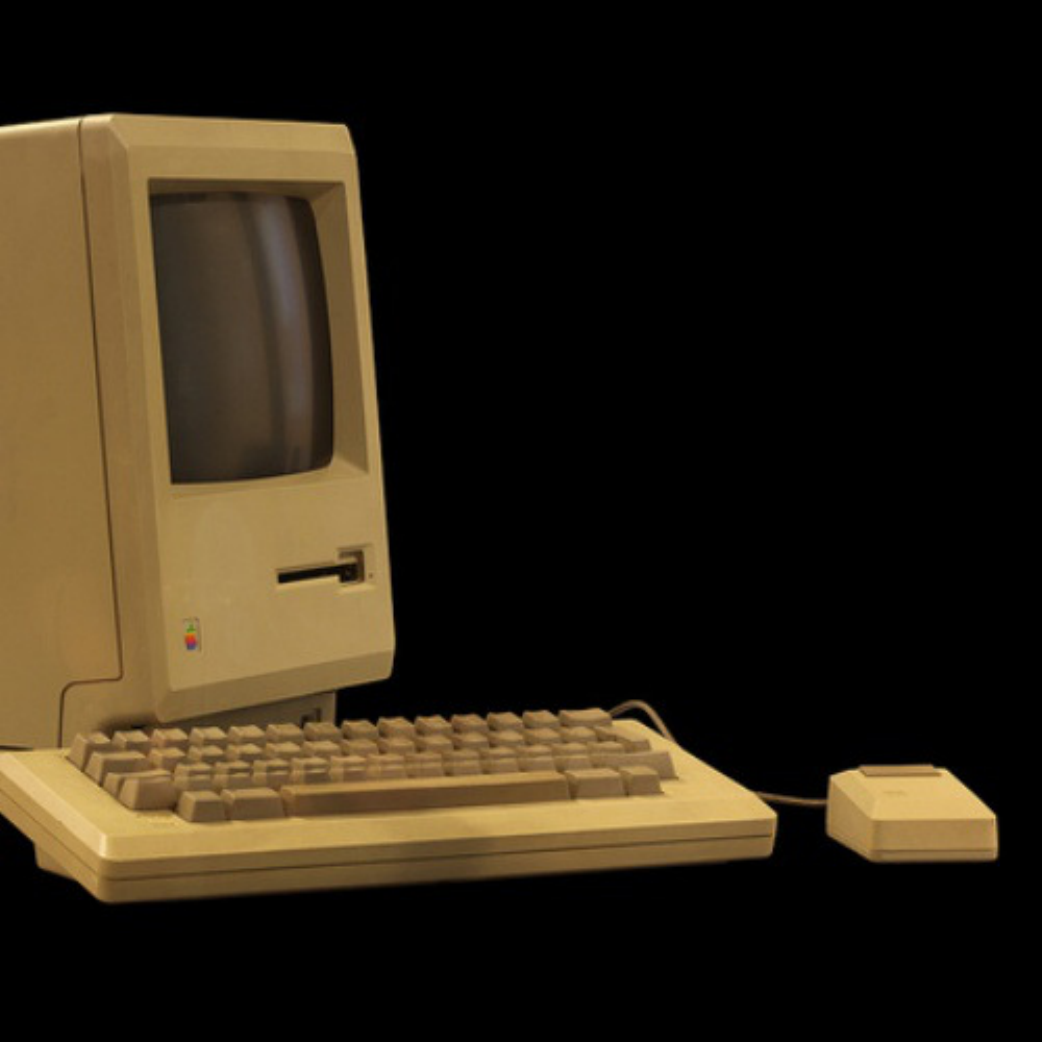}}&
\raisebox{-.5\height}{
\includegraphics[width=0.4\linewidth]{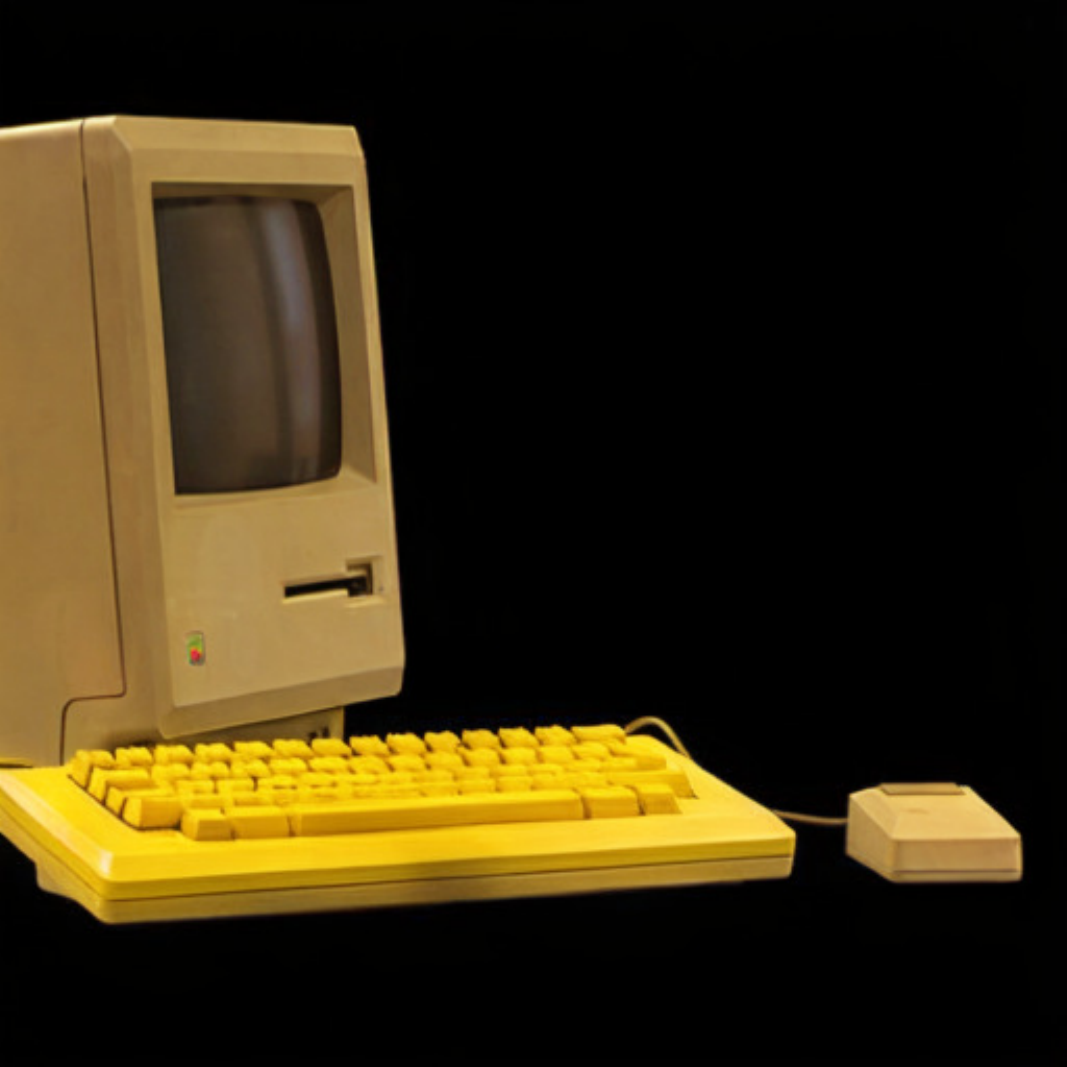}}&
\raisebox{-.5\height}{
\includegraphics[width=0.4\linewidth]{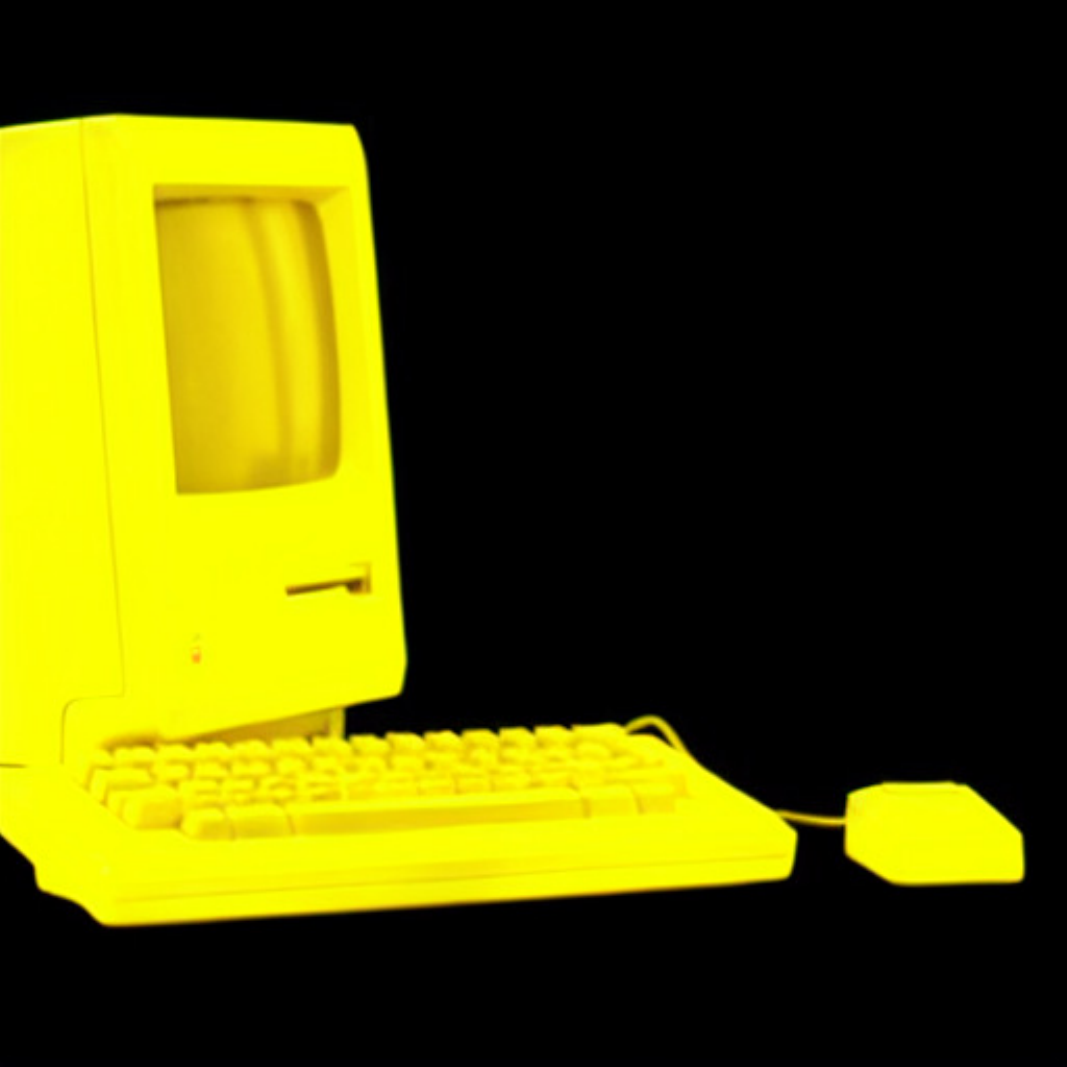}}&
\raisebox{-.5\height}{
\includegraphics[width=0.4\linewidth]{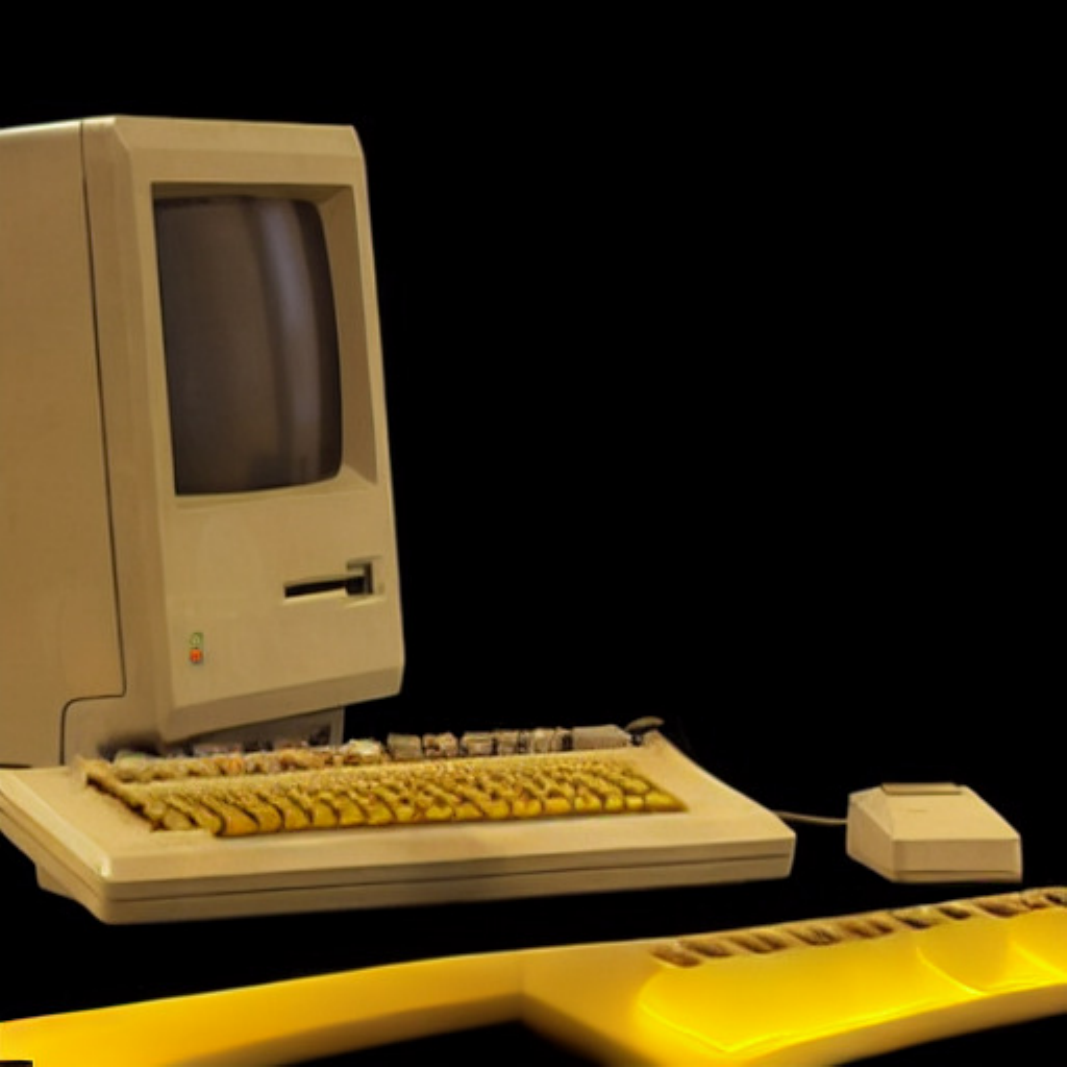}}&
\raisebox{-.5\height}{
\includegraphics[width=0.4\linewidth]{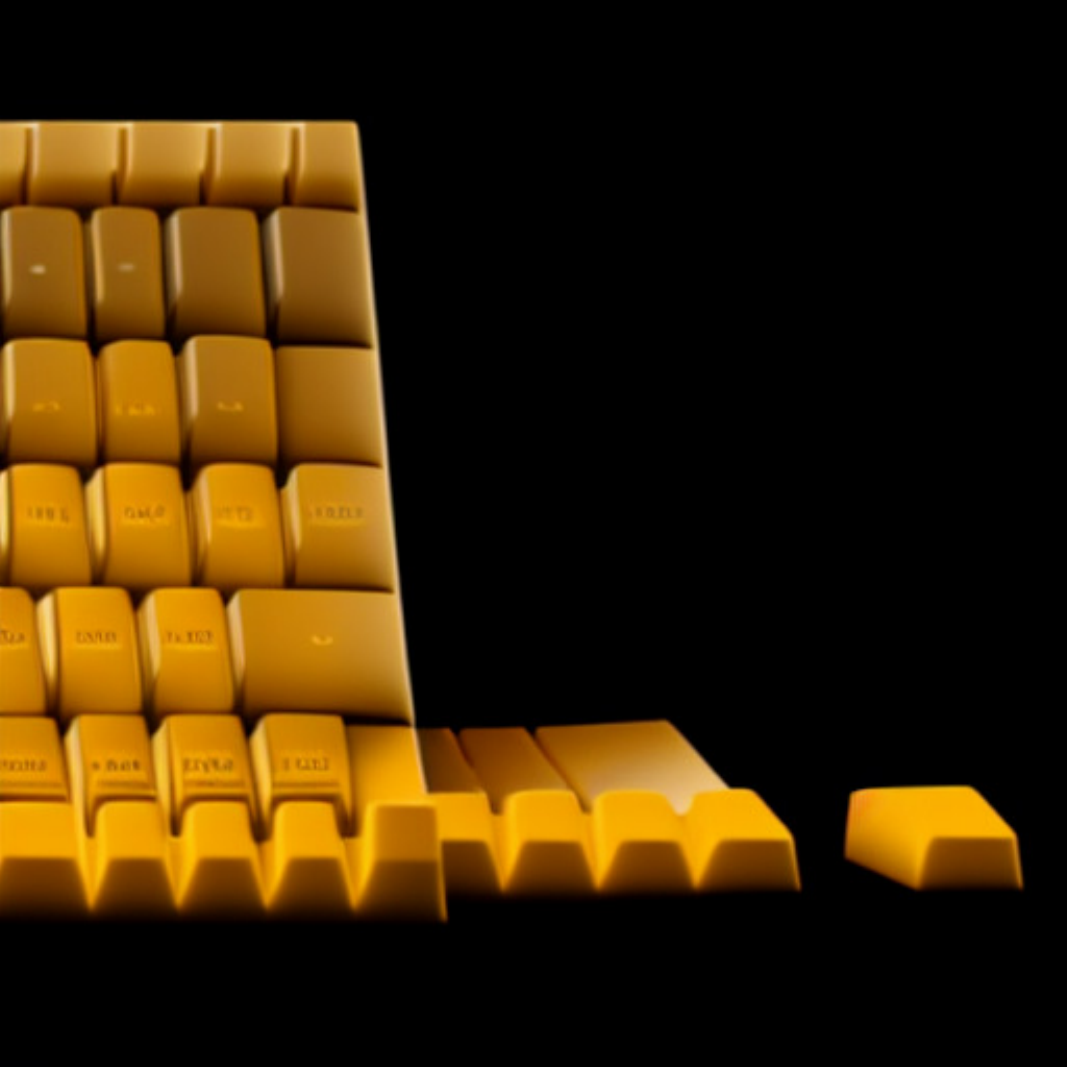}} \\

\resizebox{!}{40px}{
\begin{tabular}[x]{@{}c@{}}Remove the \\ forks from the \\ shelf\end{tabular}}&
\raisebox{-.5\height}{
\includegraphics[width=0.4\linewidth]{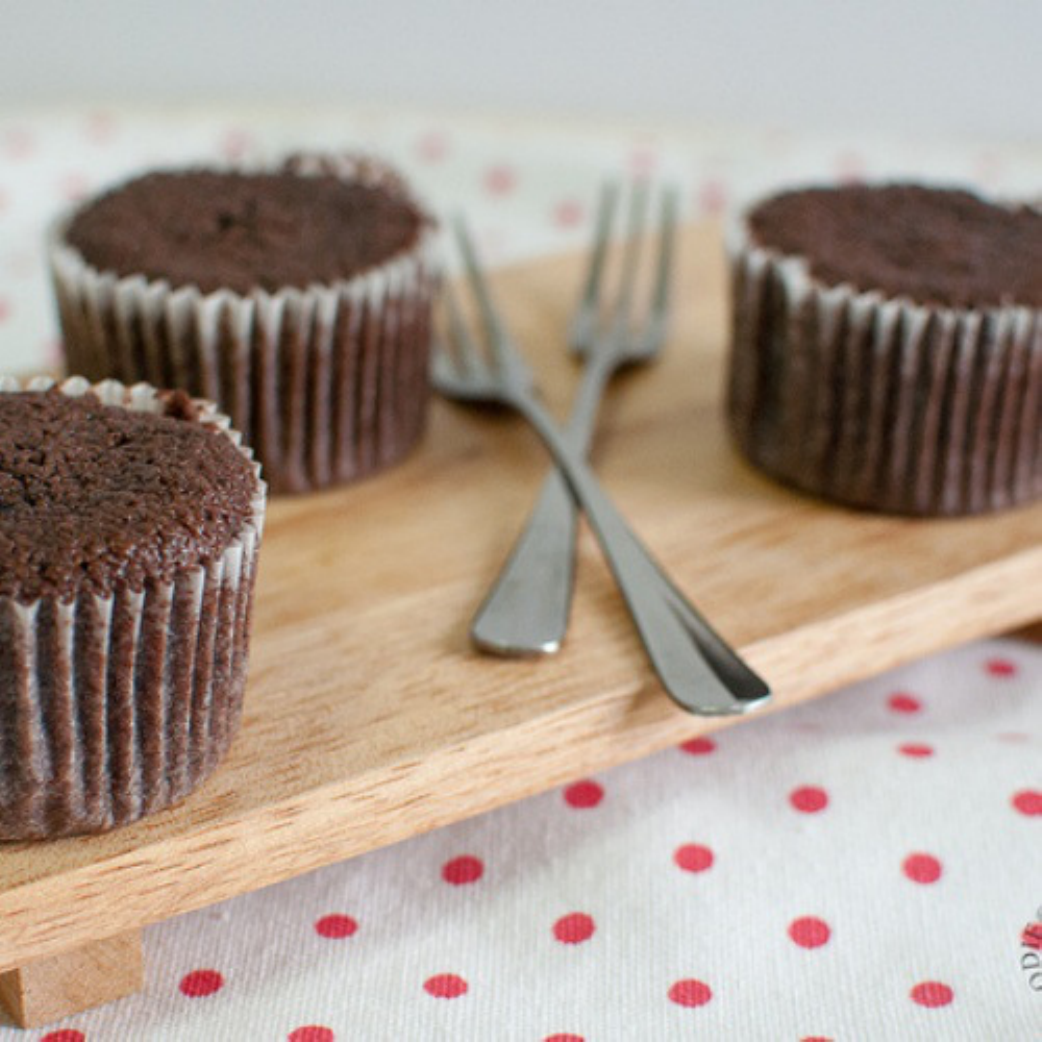}}&
\raisebox{-.5\height}{
\includegraphics[width=0.4\linewidth]{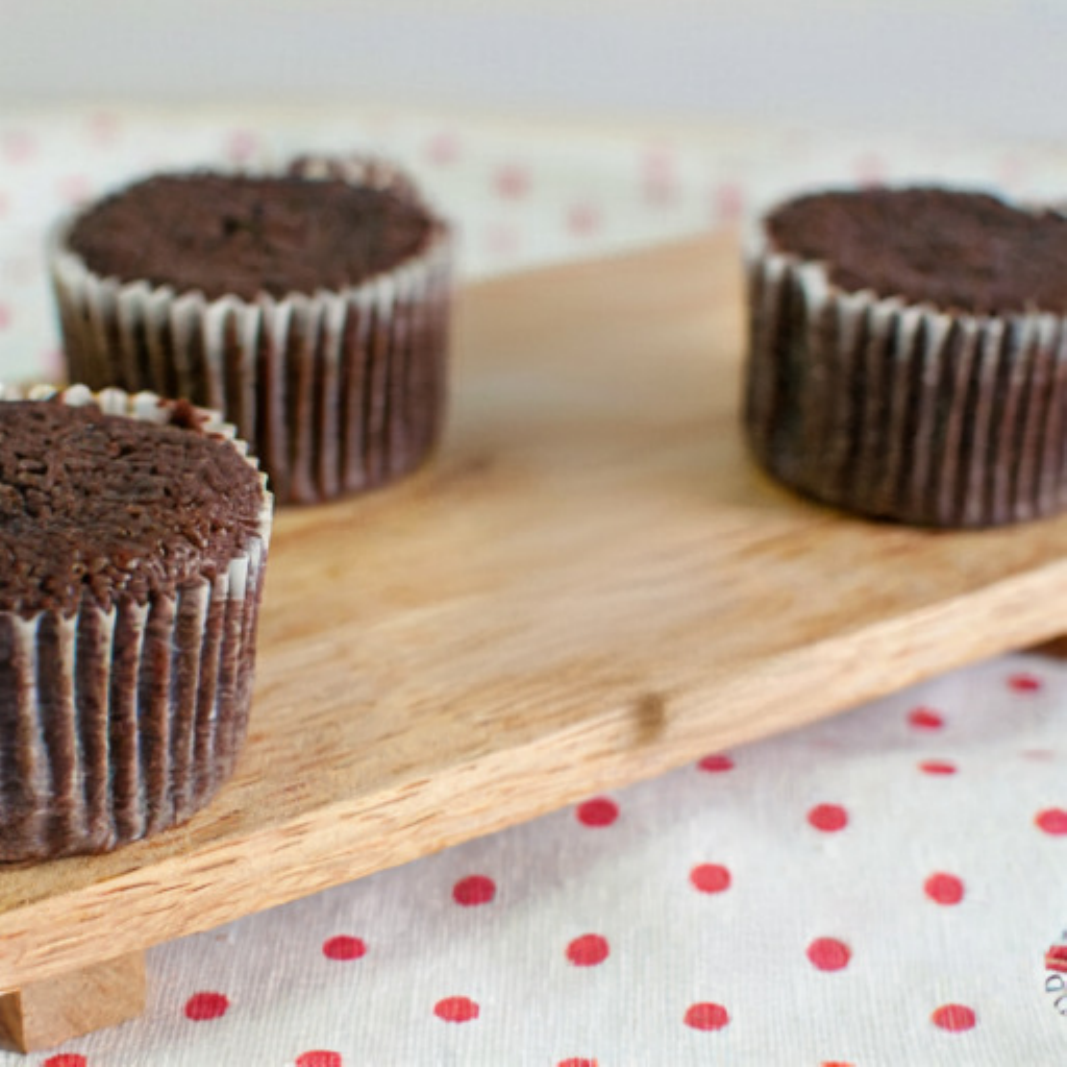}}&
\raisebox{-.5\height}{
\includegraphics[width=0.4\linewidth]{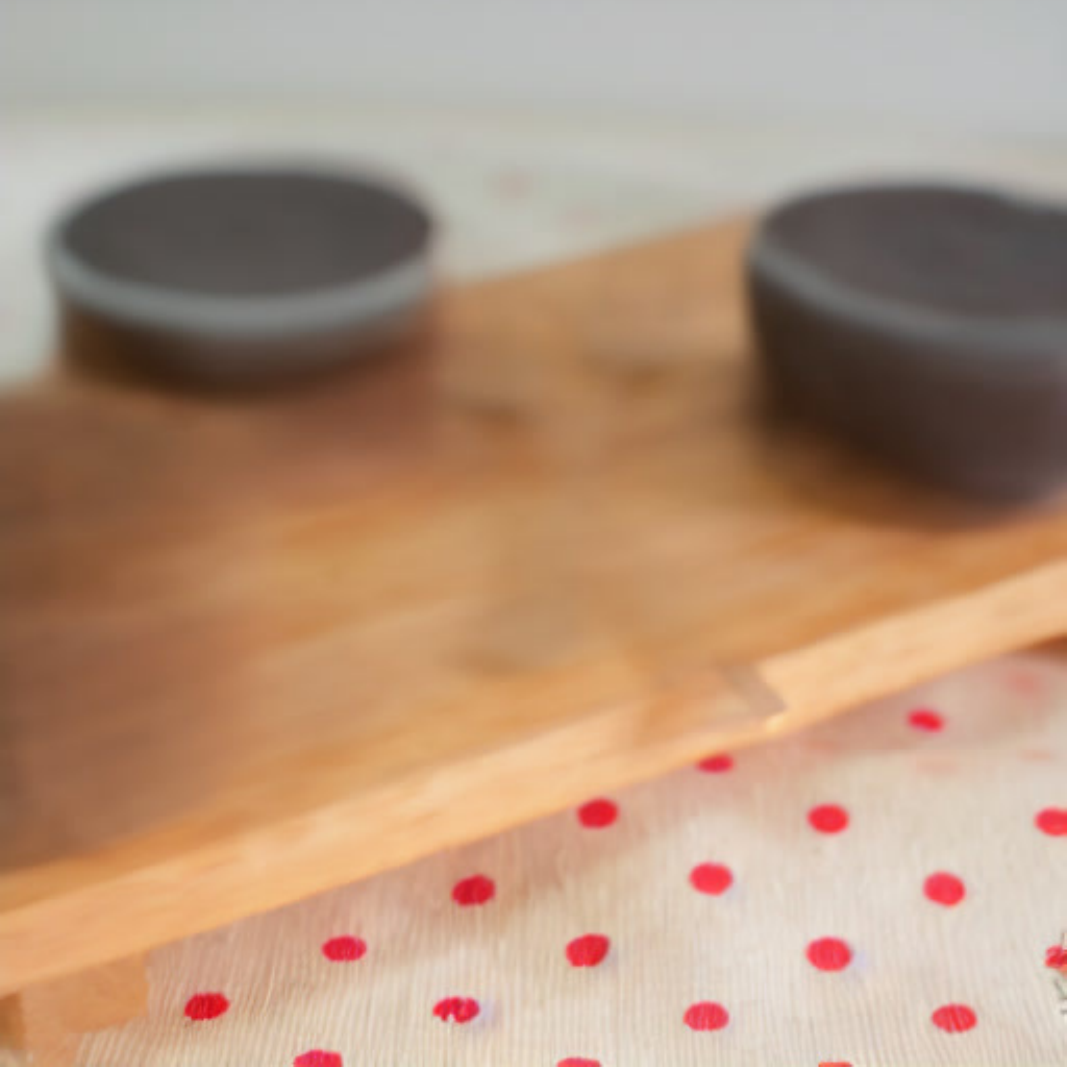}}&
\raisebox{-.5\height}{
\includegraphics[width=0.4\linewidth]{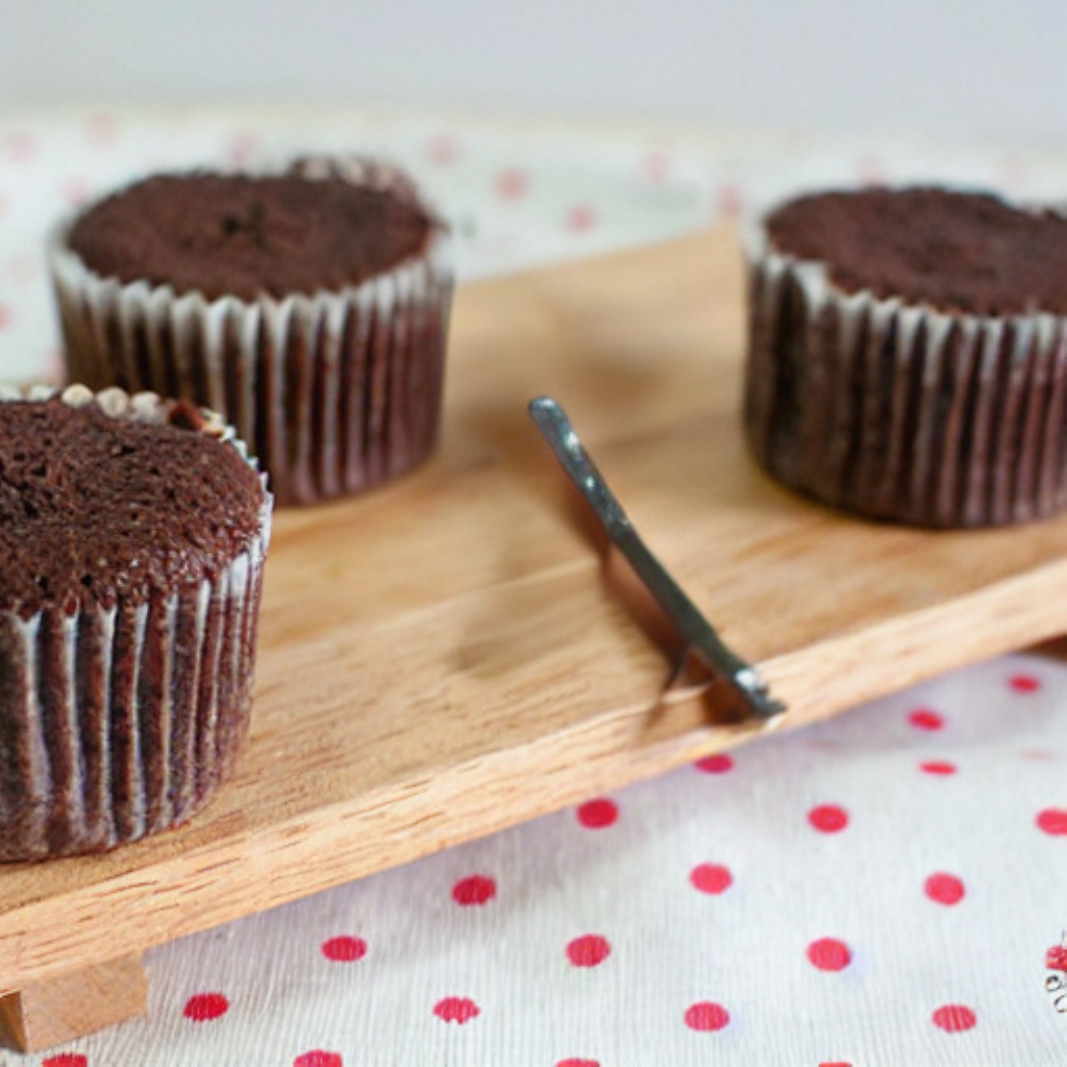}}&
\raisebox{-.5\height}{
\includegraphics[width=0.4\linewidth]{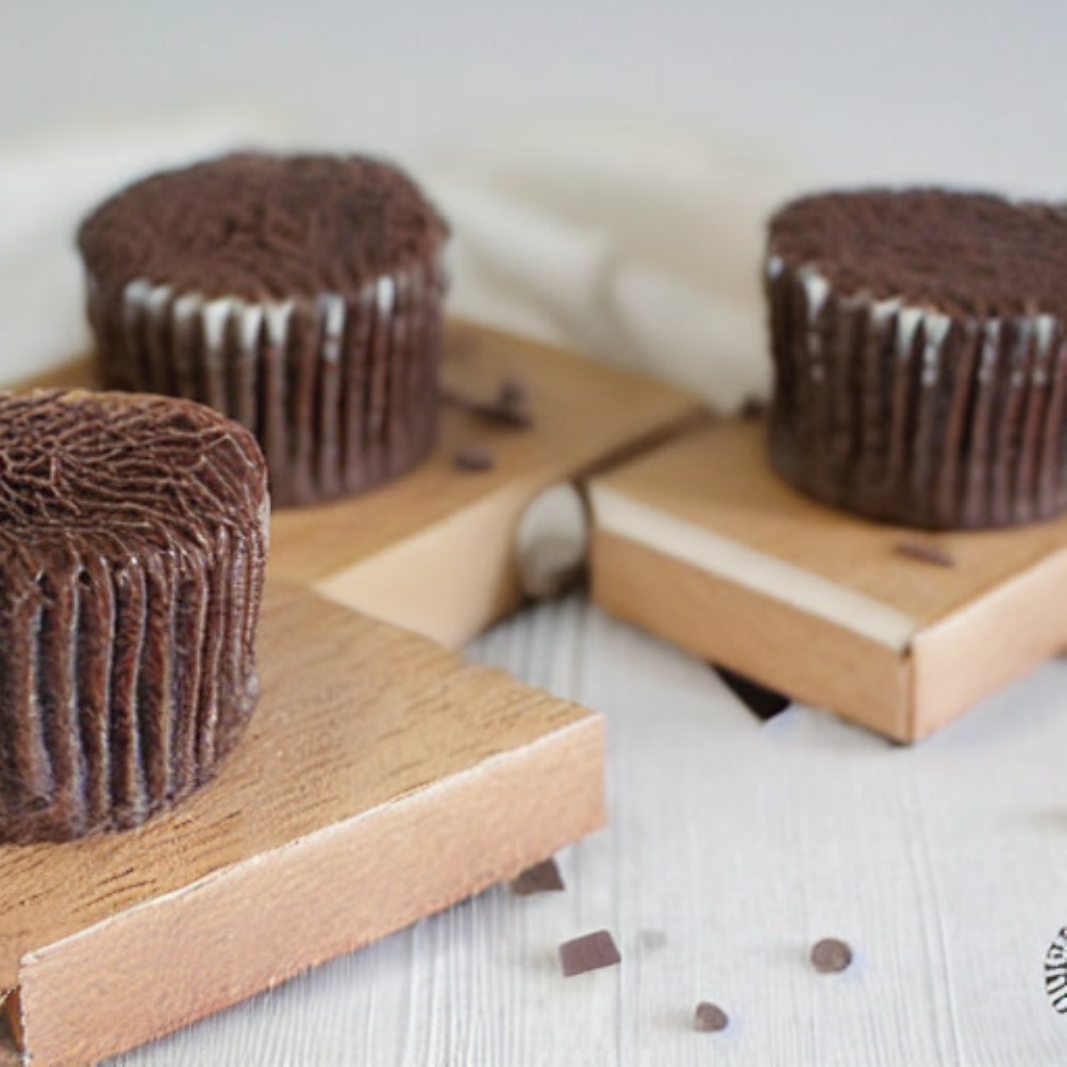}} \\

\resizebox{!}{40px}{
\begin{tabular}[x]{@{}c@{}}Add a green\\ bowl on the \\branch \end{tabular}}&
\raisebox{-.5\height}{
\includegraphics[width=0.4\linewidth]{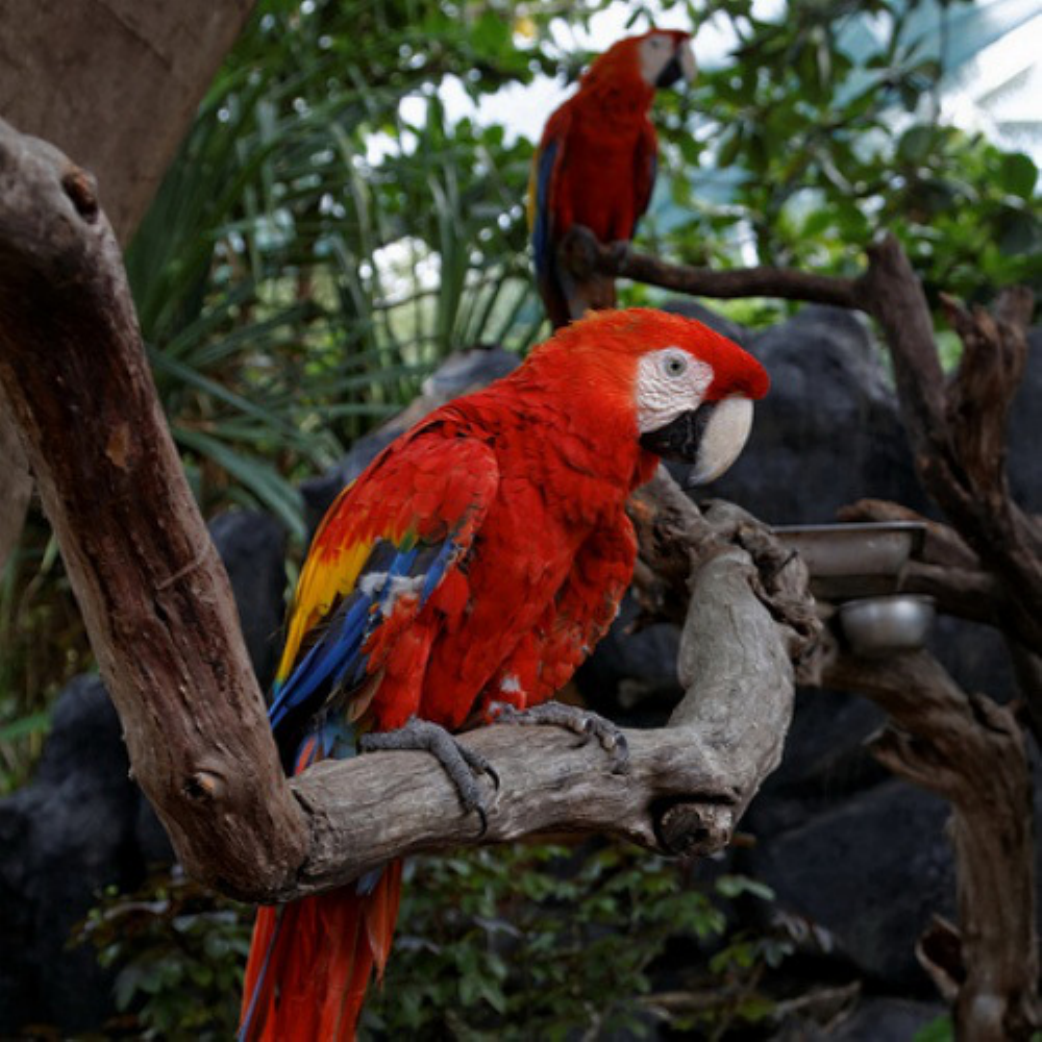}}&
\raisebox{-.5\height}{
\includegraphics[width=0.4\linewidth]{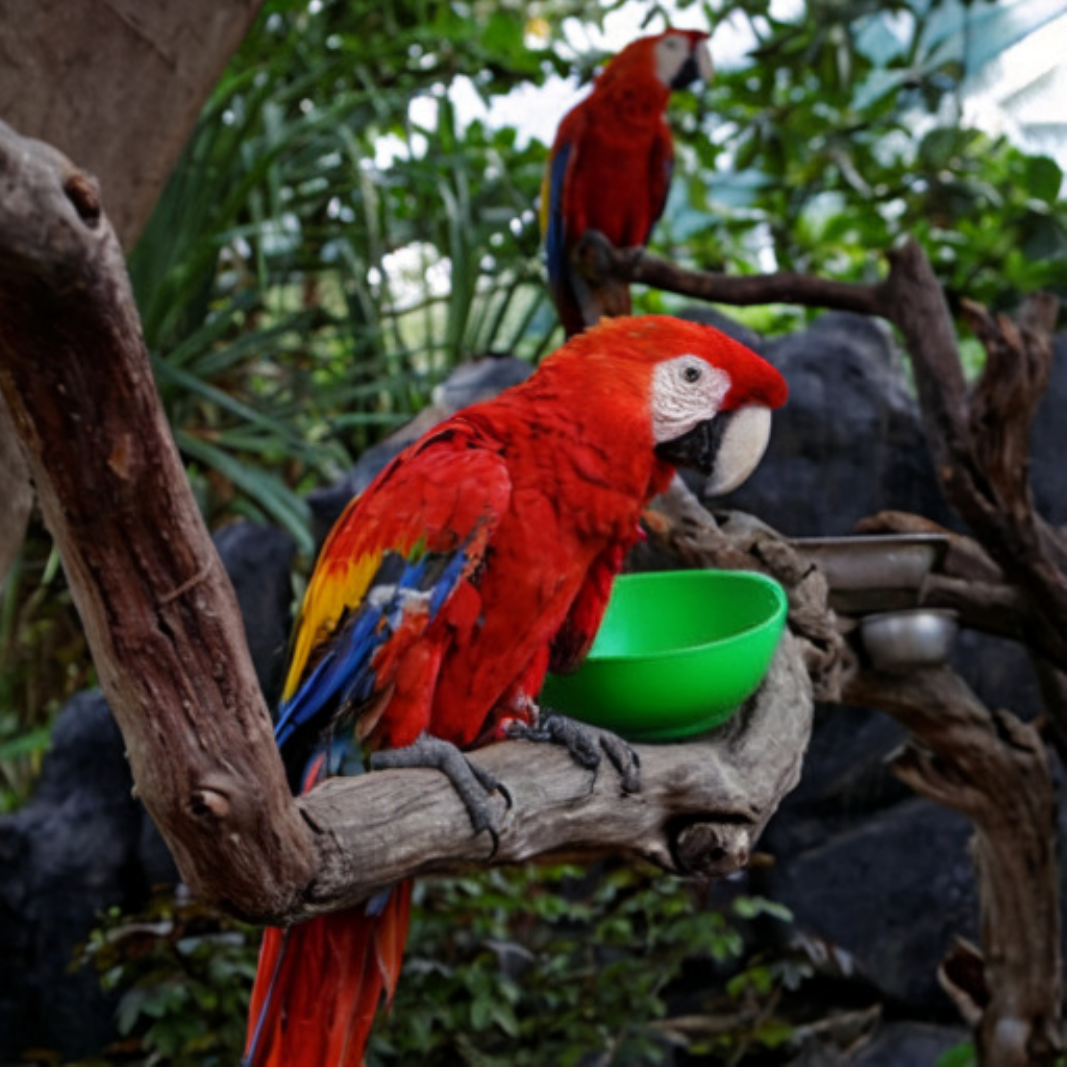}}&
\raisebox{-.5\height}{
\includegraphics[width=0.4\linewidth]{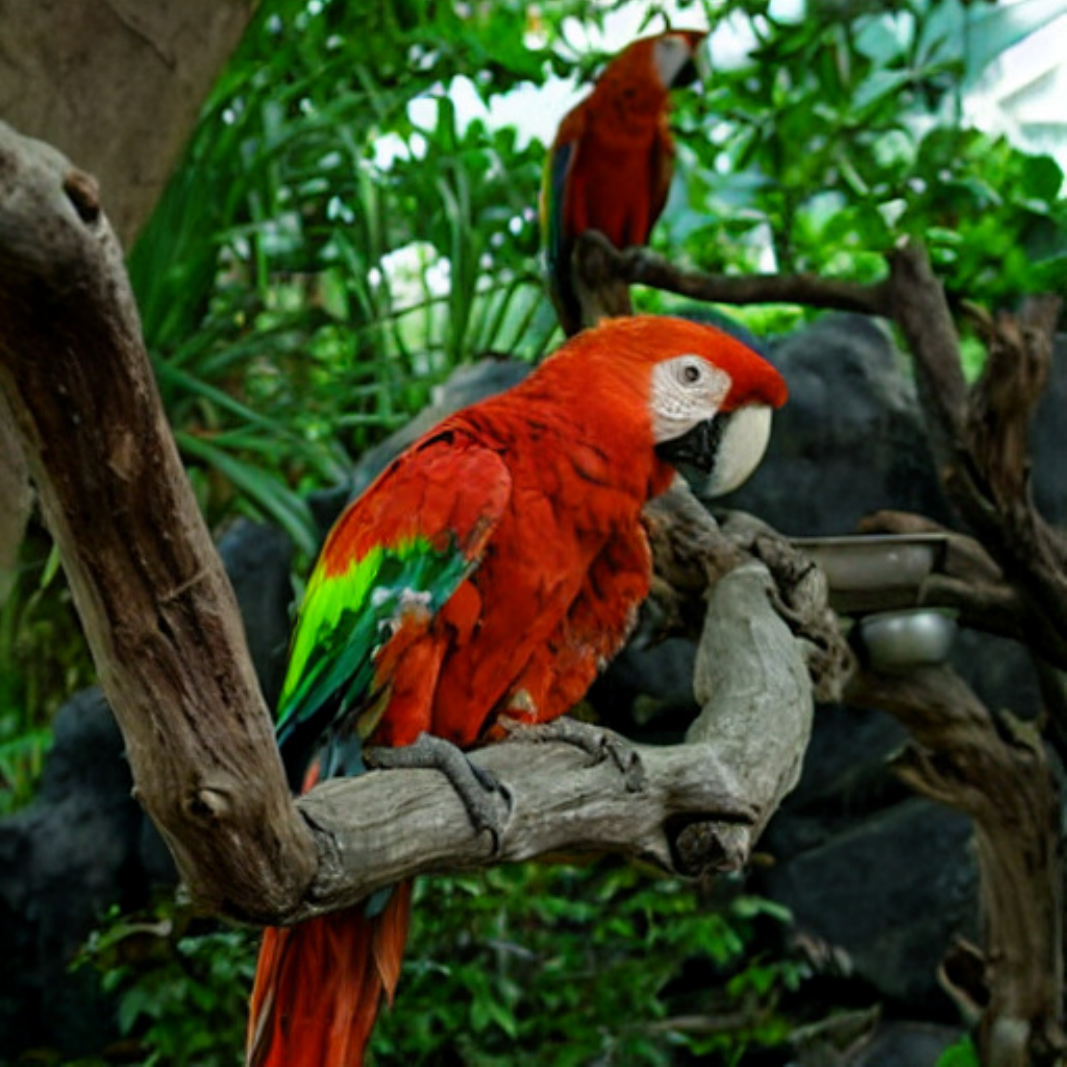}}&
\raisebox{-.5\height}{
\includegraphics[width=0.4\linewidth]{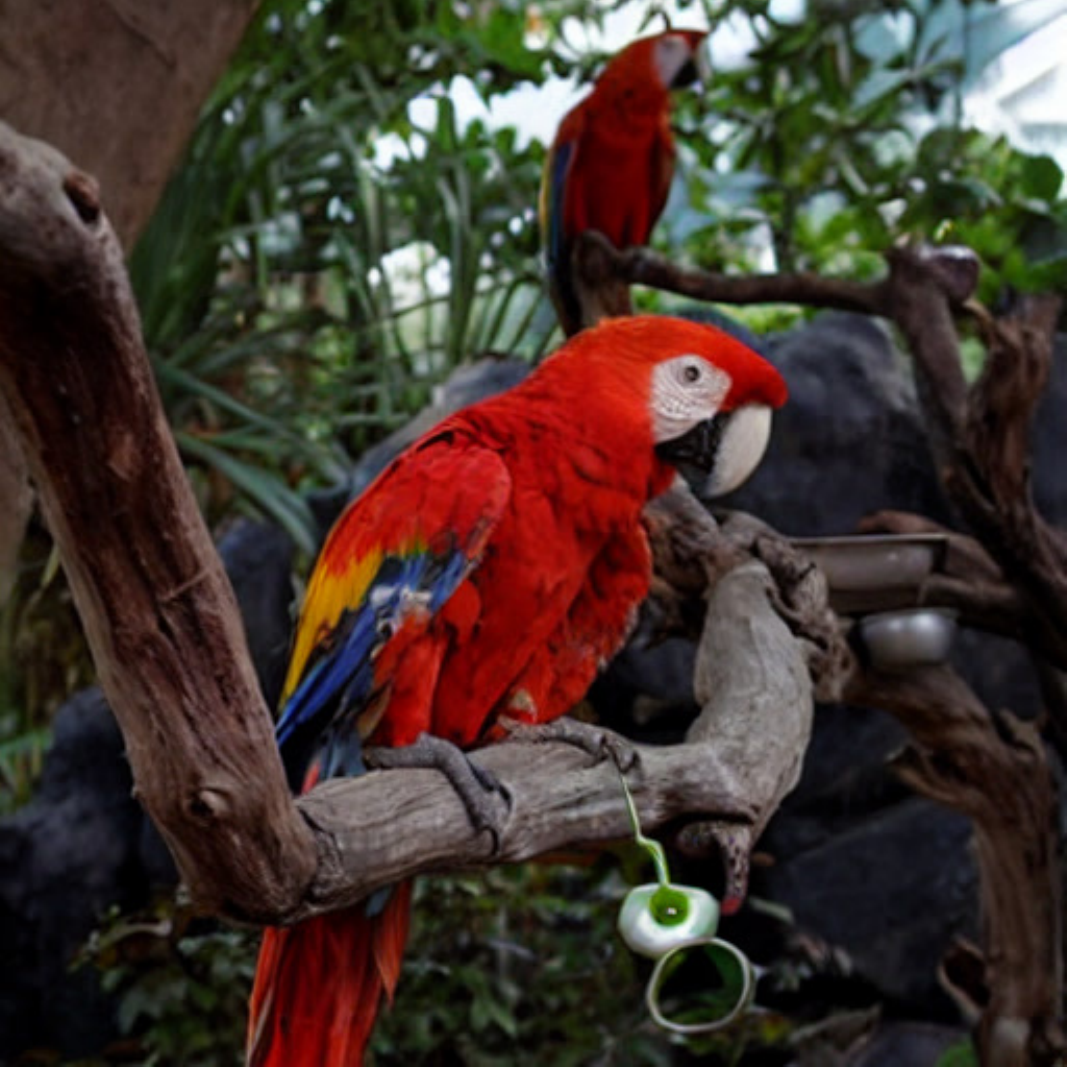}}&
\raisebox{-.5\height}{
\includegraphics[width=0.4\linewidth]
{imgs/comp_our_test/22_magic.pdf}} \\

\resizebox{!}{50px}{
\begin{tabular}[x]{@{}c@{}}Add the word \\ 'hi' in  graffiti \\ font to  the side \\of the truck \end{tabular}}&
\raisebox{-.5\height}{
\includegraphics[width=0.4\linewidth]{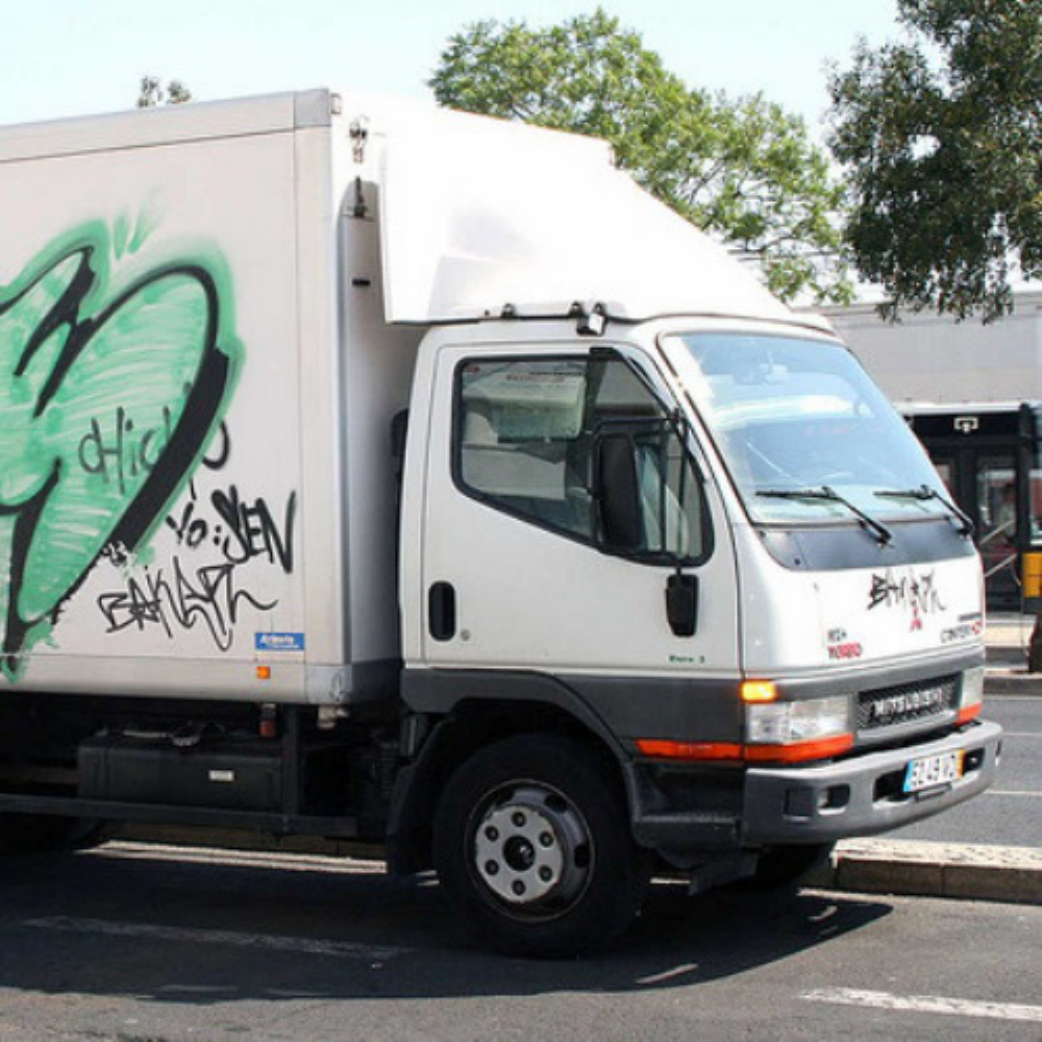}}&
\raisebox{-.5\height}{
\includegraphics[width=0.4\linewidth]{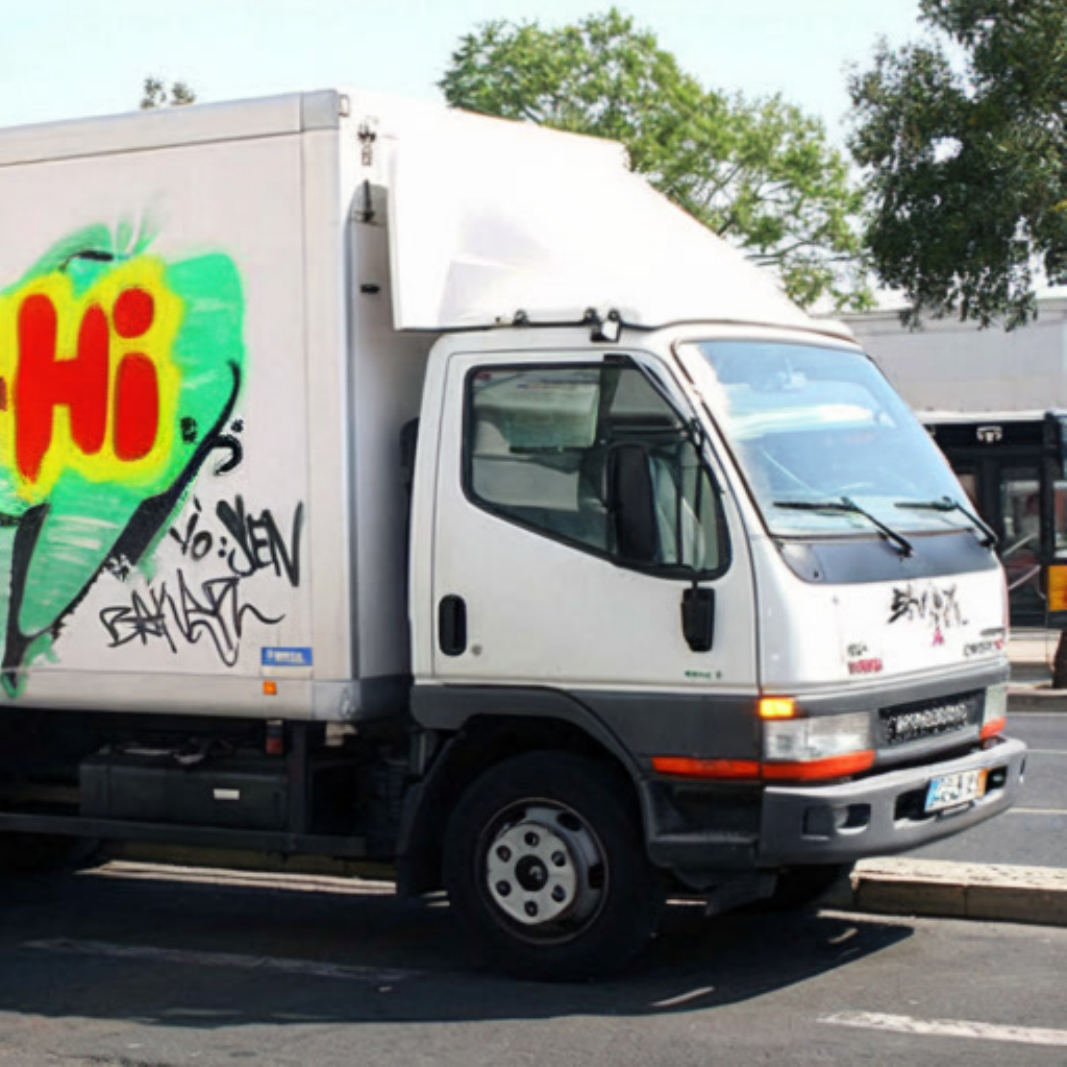}}&
\raisebox{-.5\height}{
\includegraphics[width=0.4\linewidth]{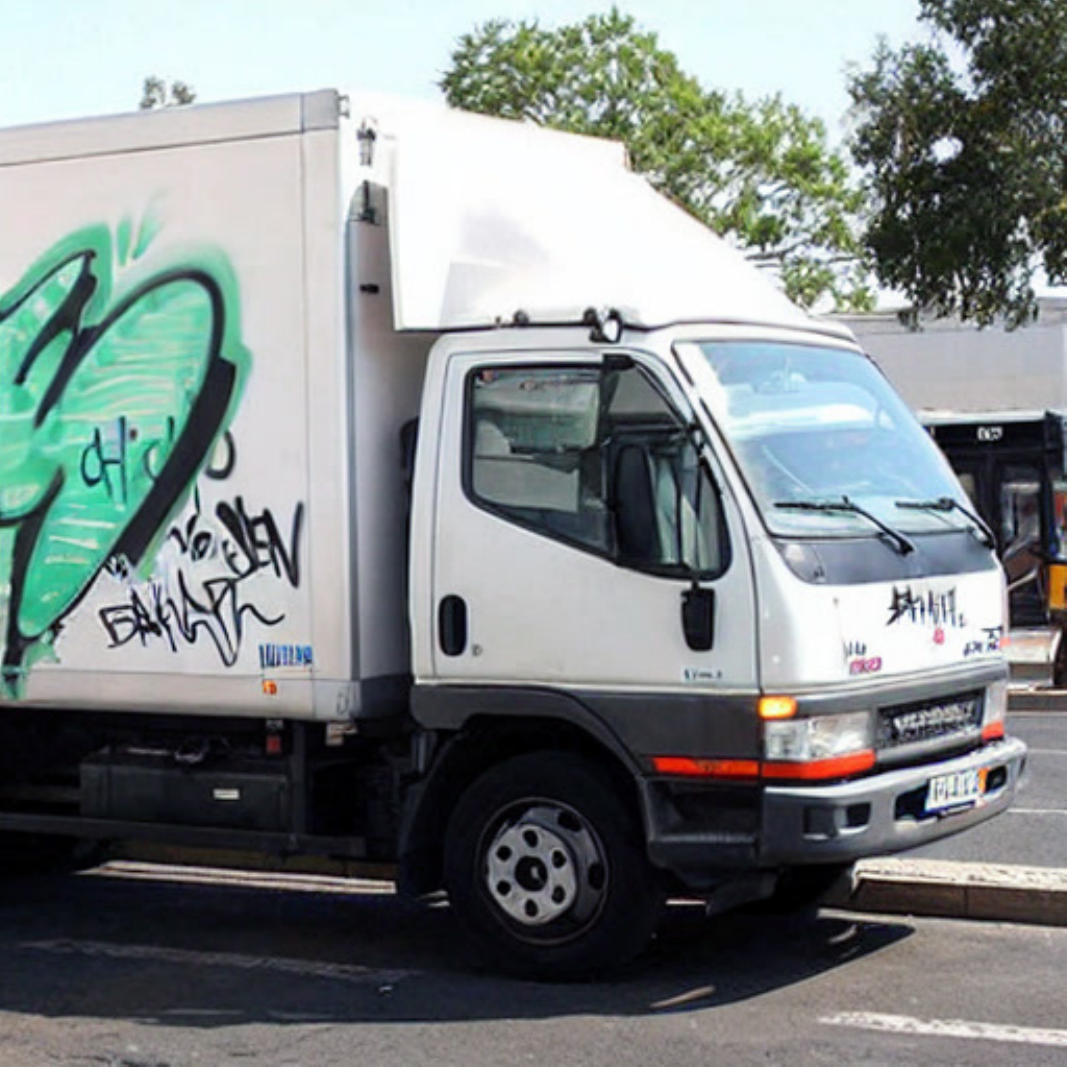}}&
\raisebox{-.5\height}{
\includegraphics[width=0.4\linewidth]{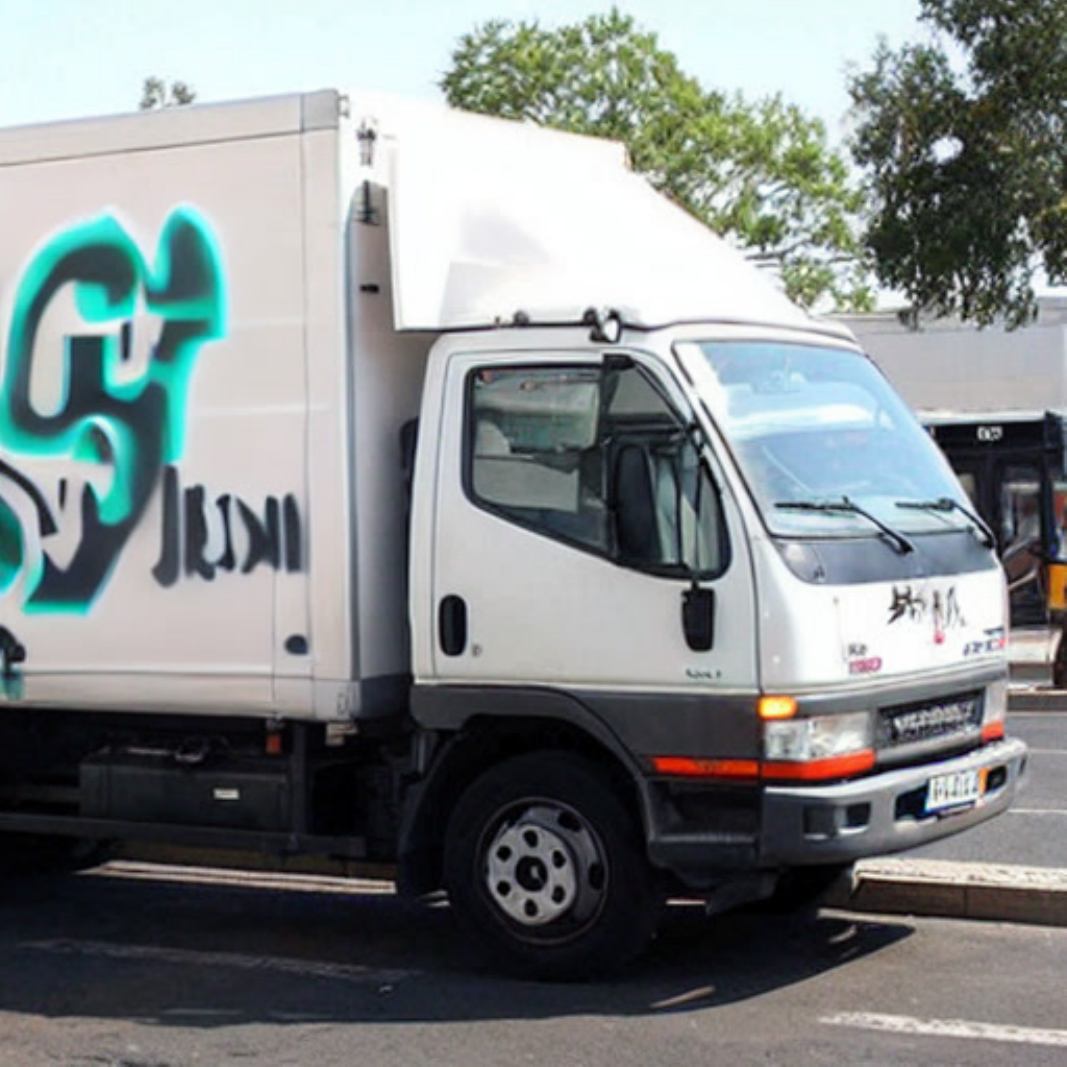}}&
\raisebox{-.5\height}{
\includegraphics[width=0.4\linewidth]{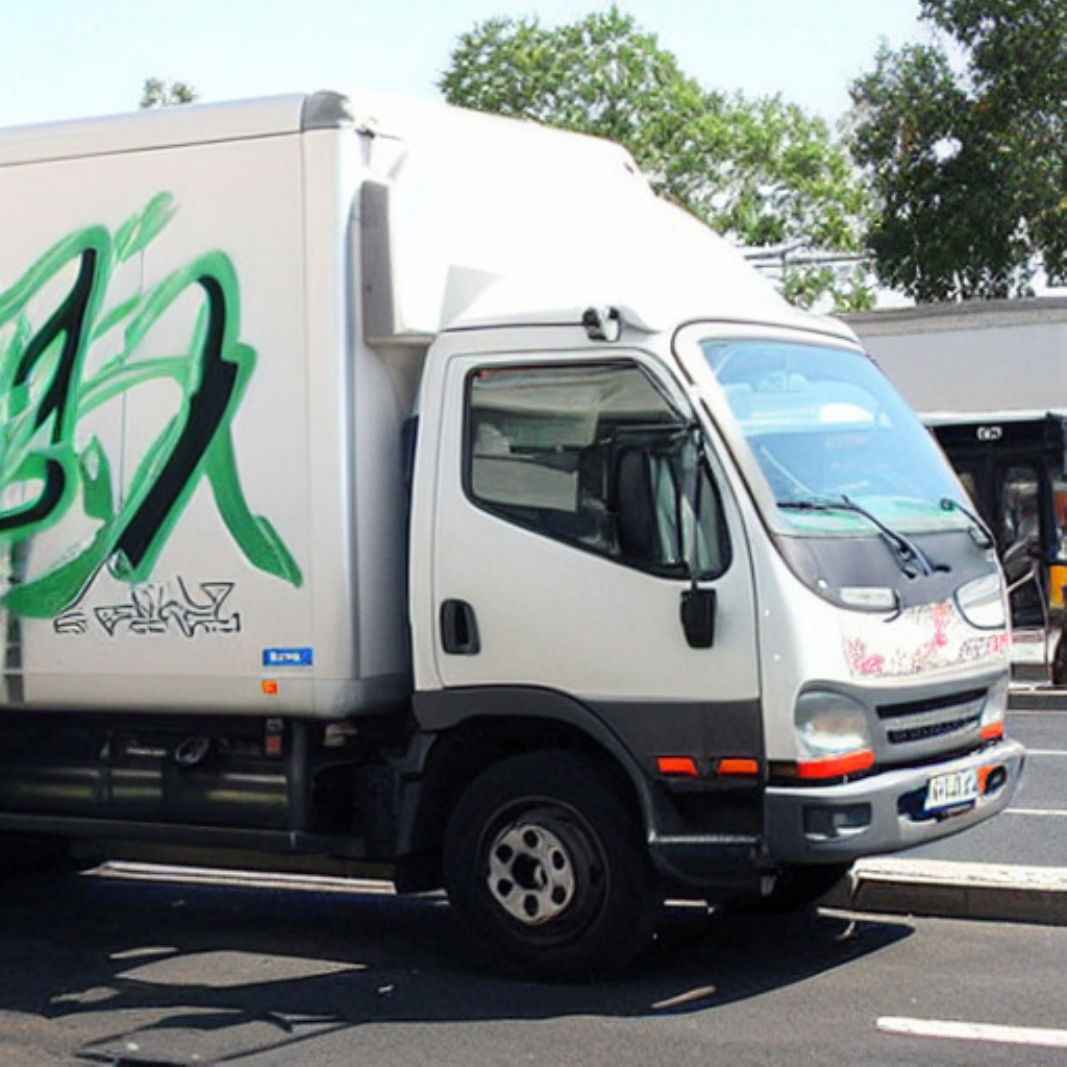}} \\

\resizebox{!}{40px}{
\begin{tabular}[x]{@{}c@{}} Make the \\donut an \\ apple \end{tabular}}&
\raisebox{-.5\height}{
\includegraphics[width=0.4\linewidth]{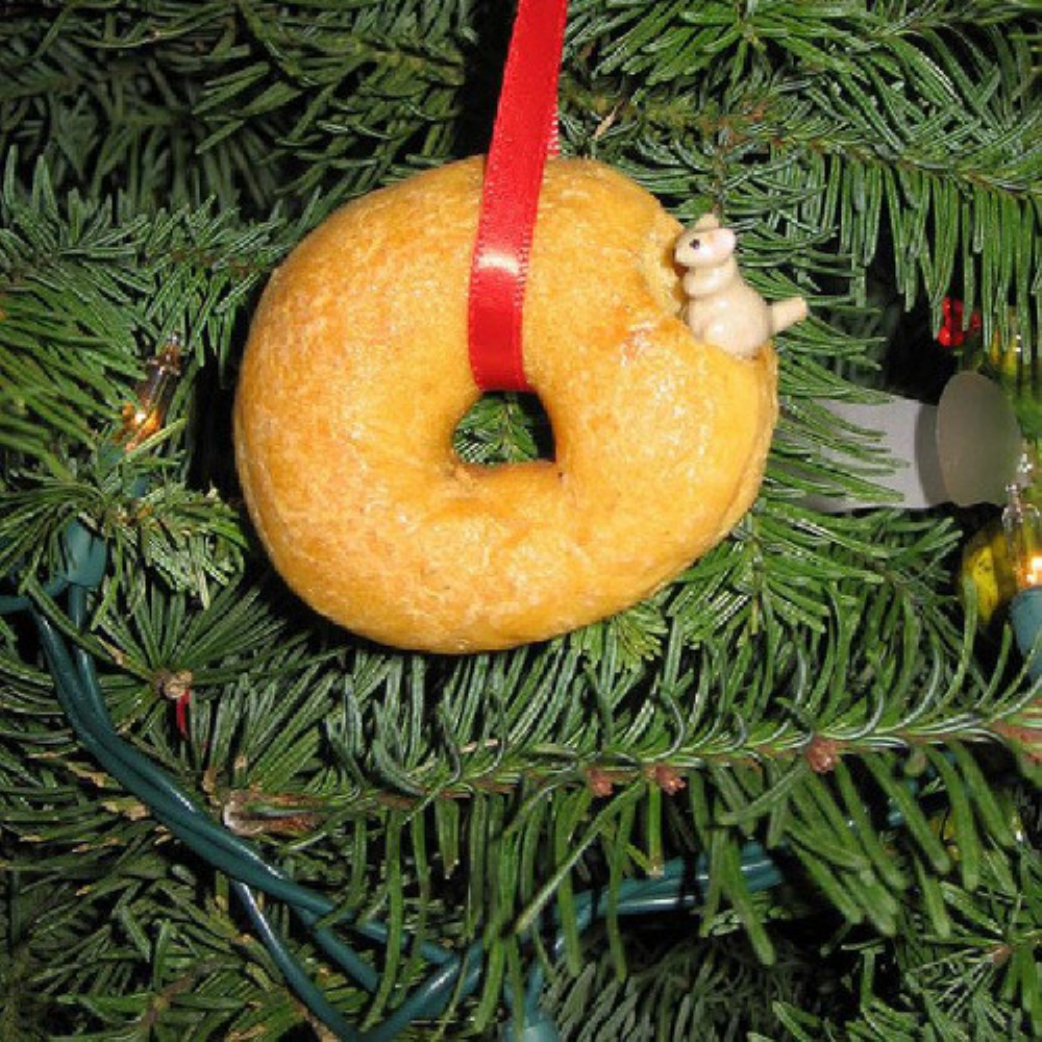}}&
\raisebox{-.5\height}{
\includegraphics[width=0.4\linewidth]{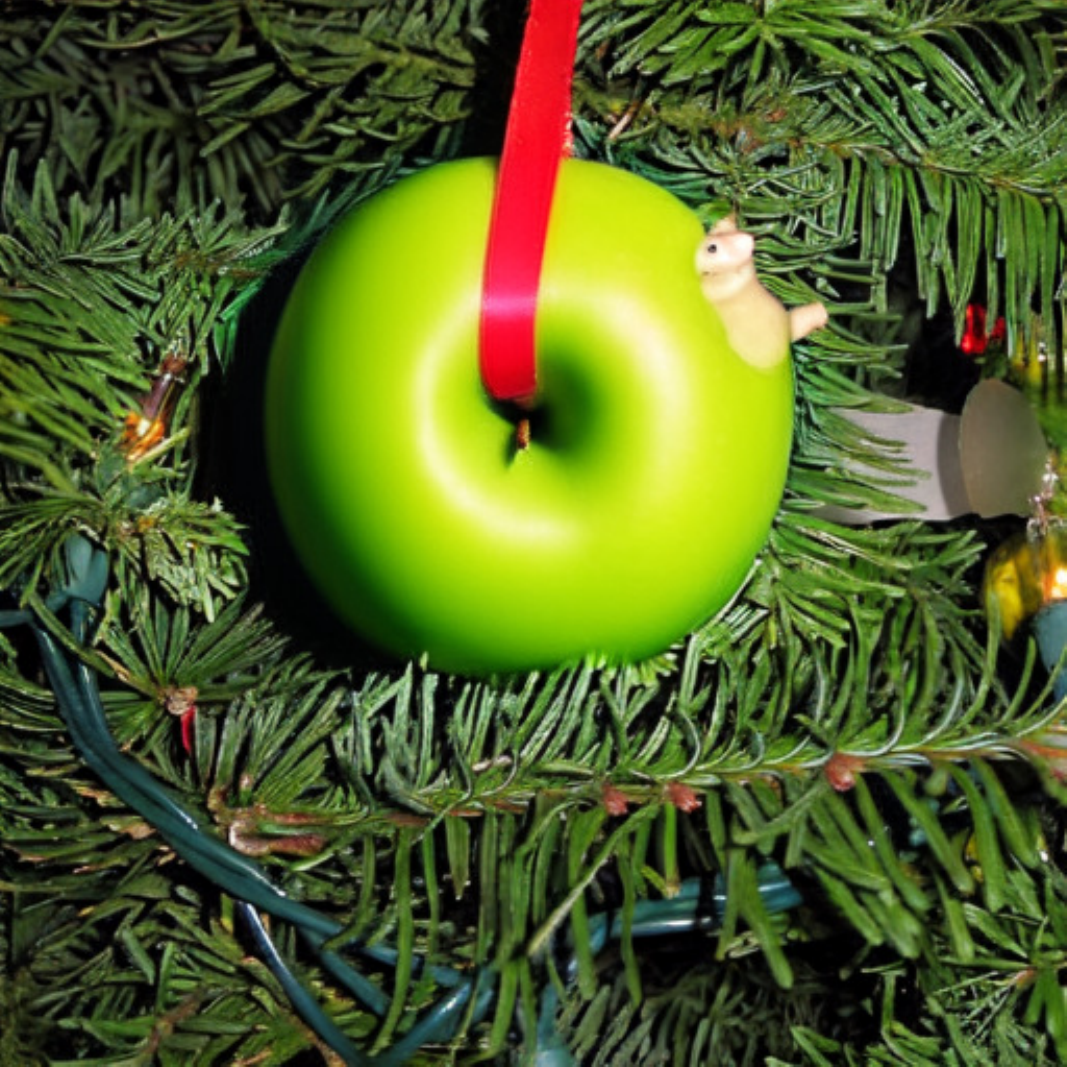}}&
\raisebox{-.5\height}{
\includegraphics[width=0.4\linewidth]{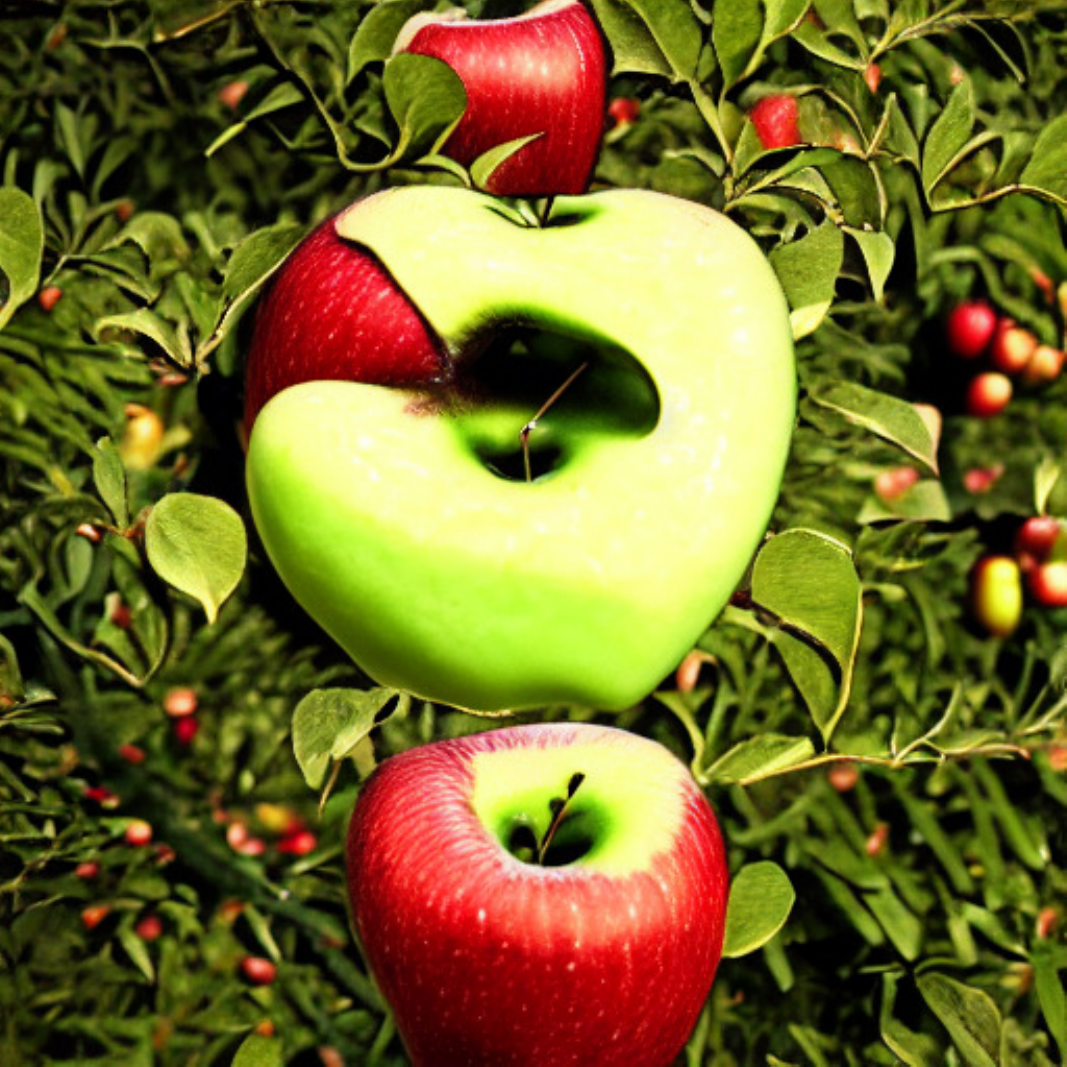}}&
\raisebox{-.5\height}{
\includegraphics[width=0.4\linewidth]{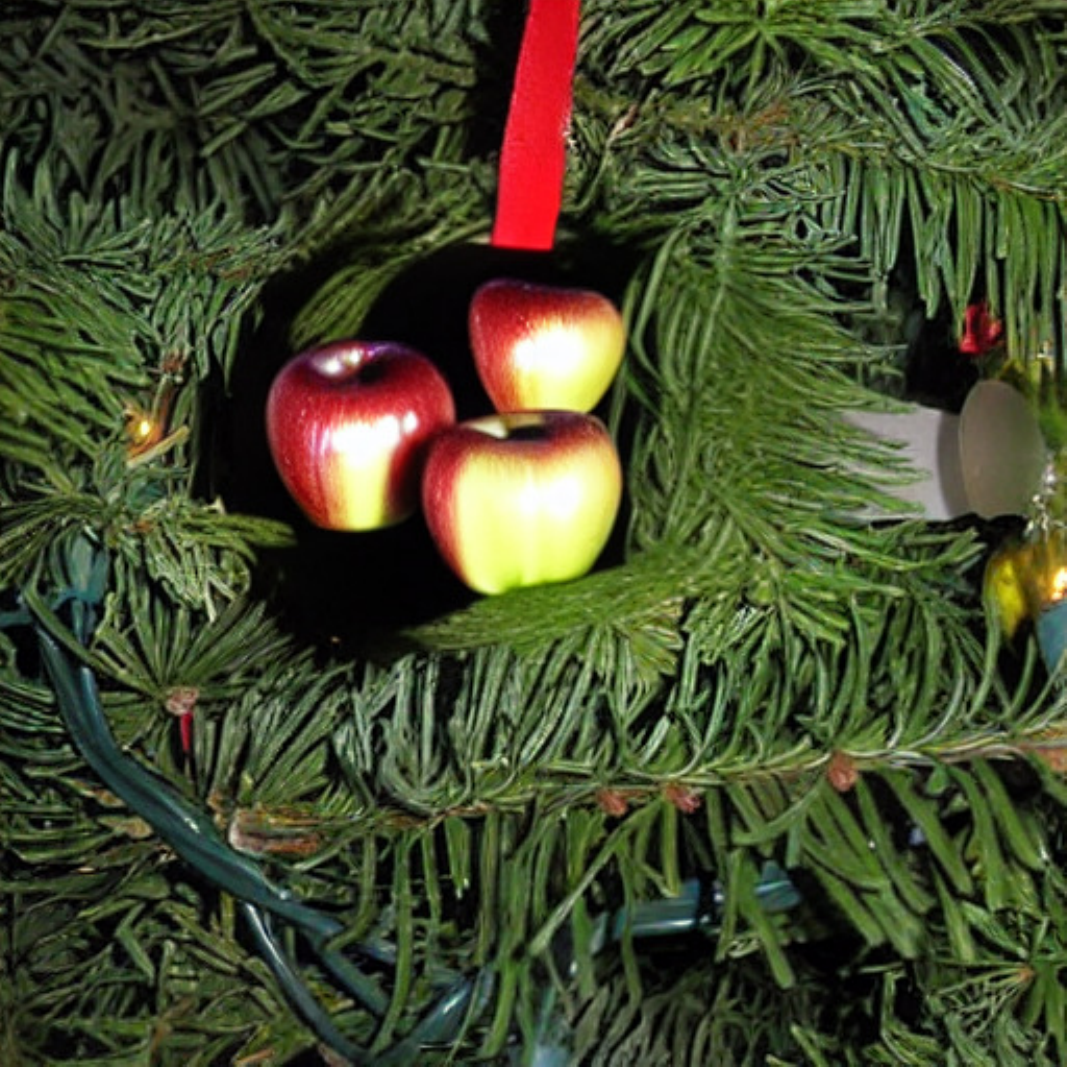}}&
\raisebox{-.5\height}{
\includegraphics[width=0.4\linewidth]{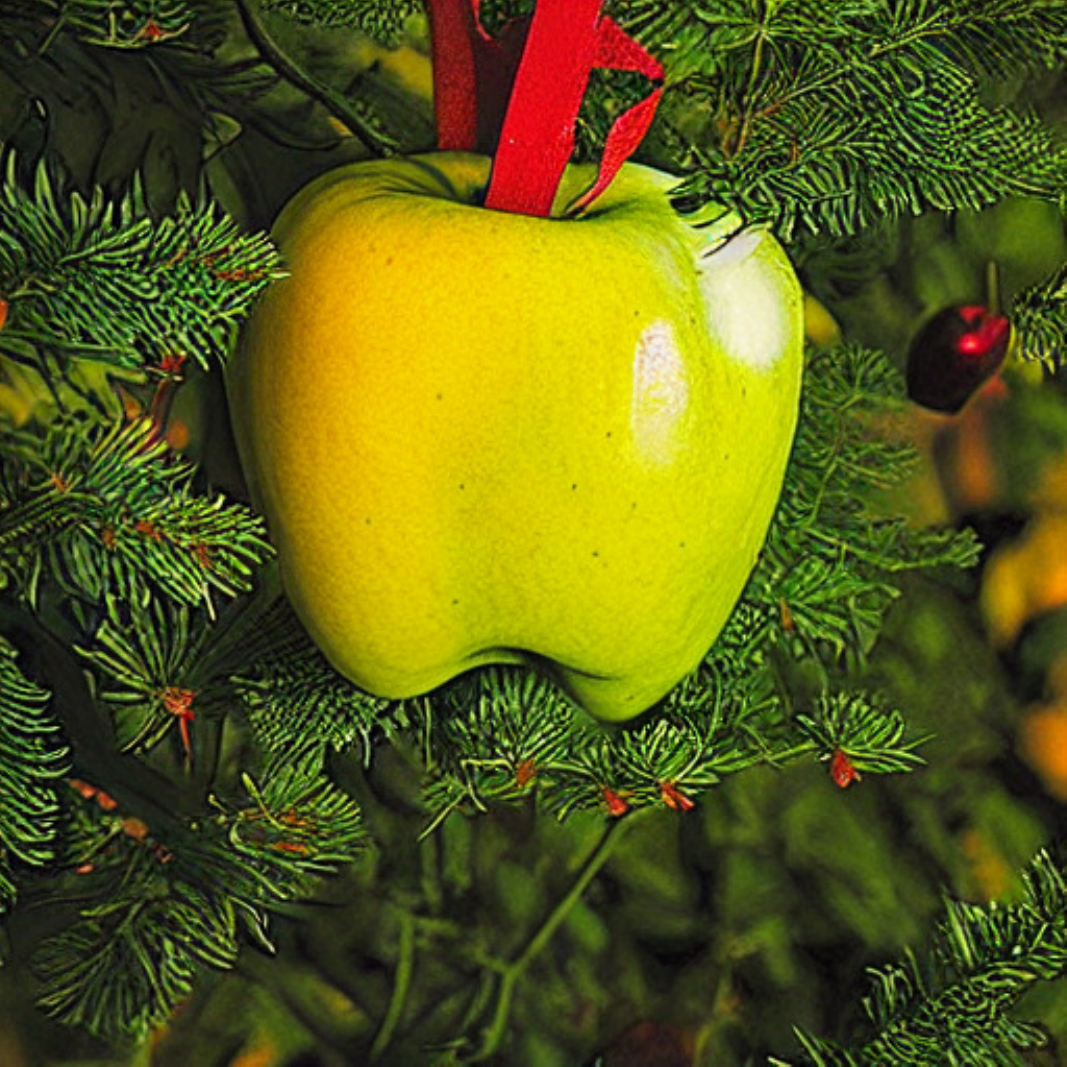}} \\

\resizebox{!}{38px}{
\begin{tabular}[x]{@{}c@{}}Turn the \\umbrella into \\ a palm tree\end{tabular}}&
\raisebox{-.5\height}{
\includegraphics[width=0.4\linewidth]{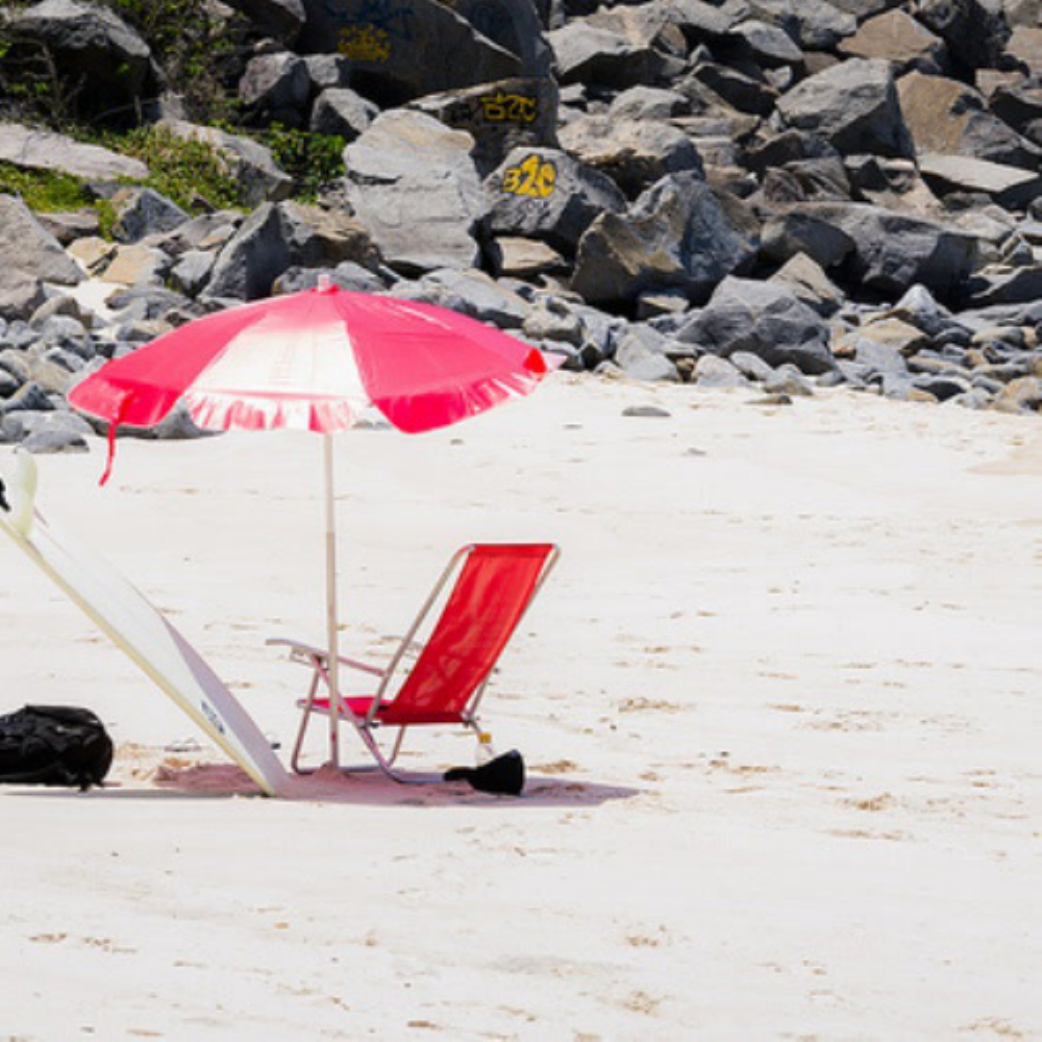}}&
\raisebox{-.5\height}{
\includegraphics[width=0.4\linewidth]{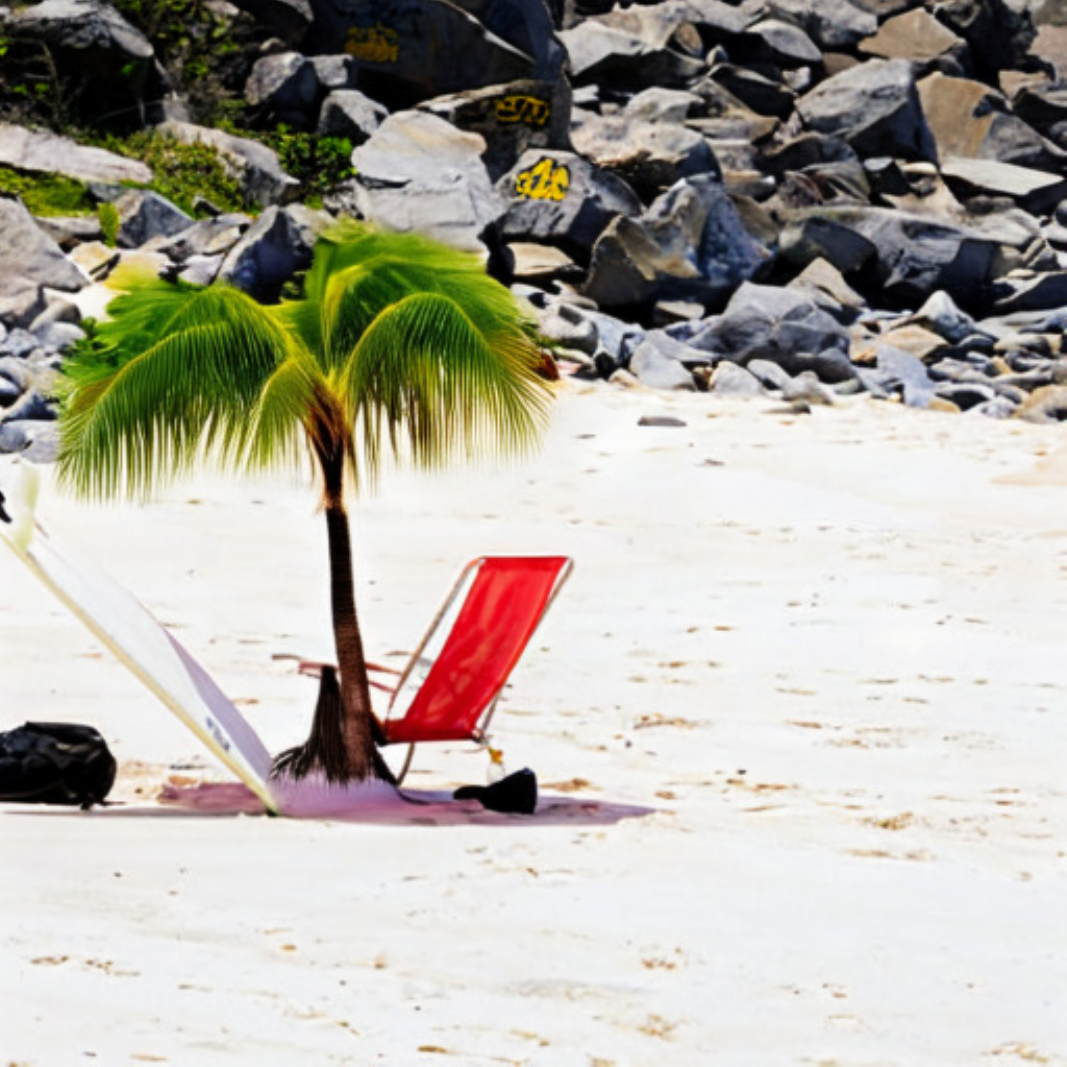}}&
\raisebox{-.5\height}{
\includegraphics[width=0.4\linewidth]{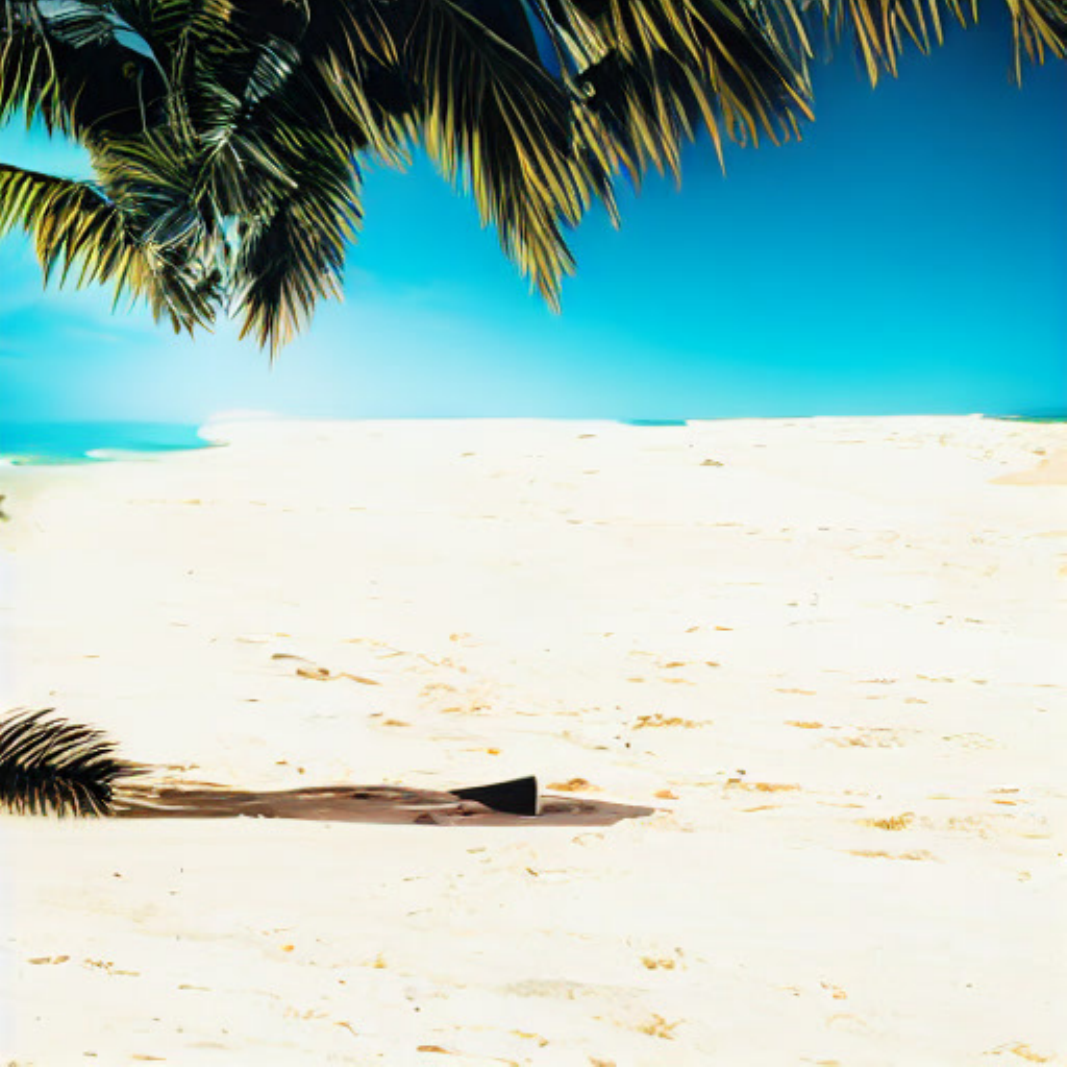}}&
\raisebox{-.5\height}{
\includegraphics[width=0.4\linewidth]{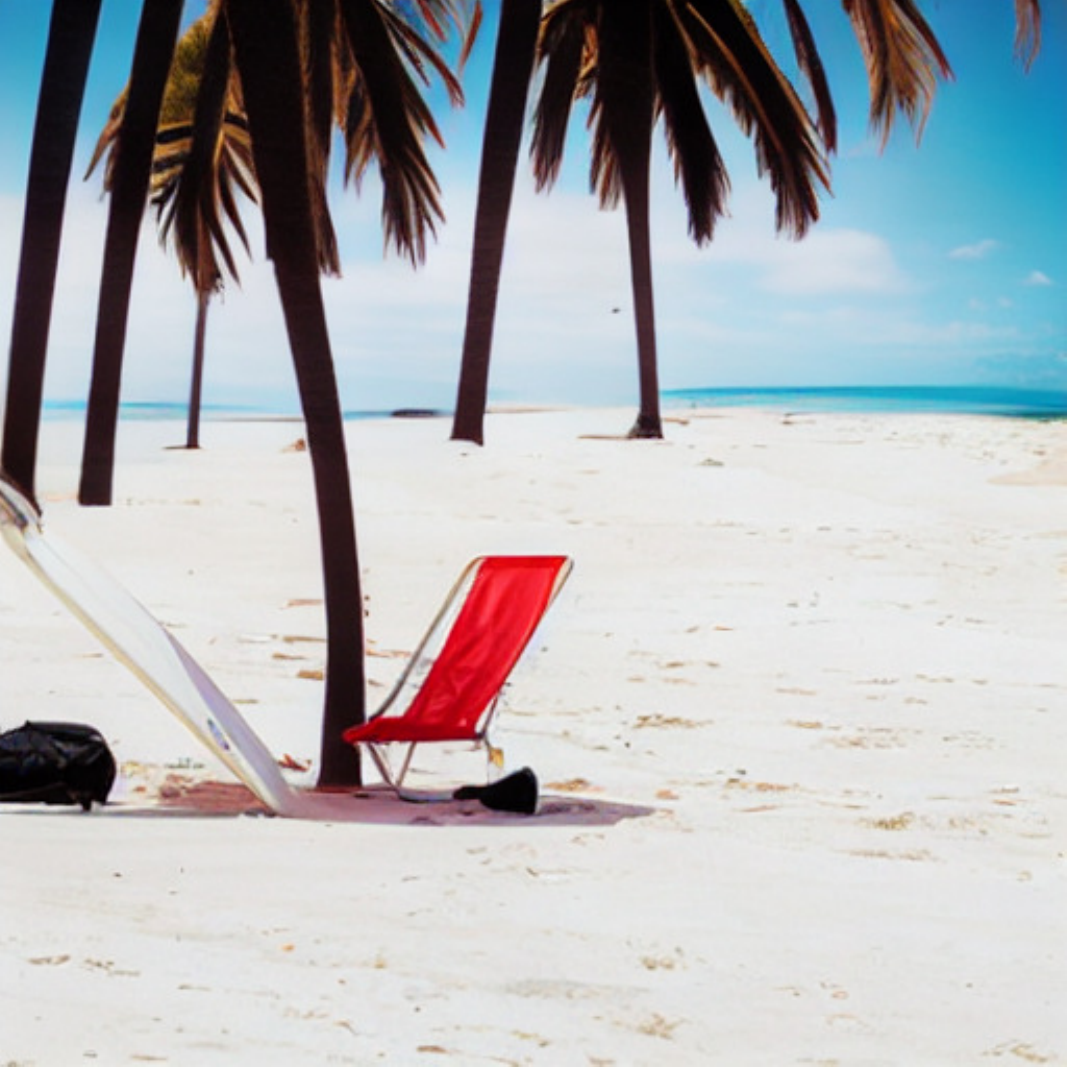}}&
\raisebox{-.5\height}{
\includegraphics[width=0.4\linewidth]{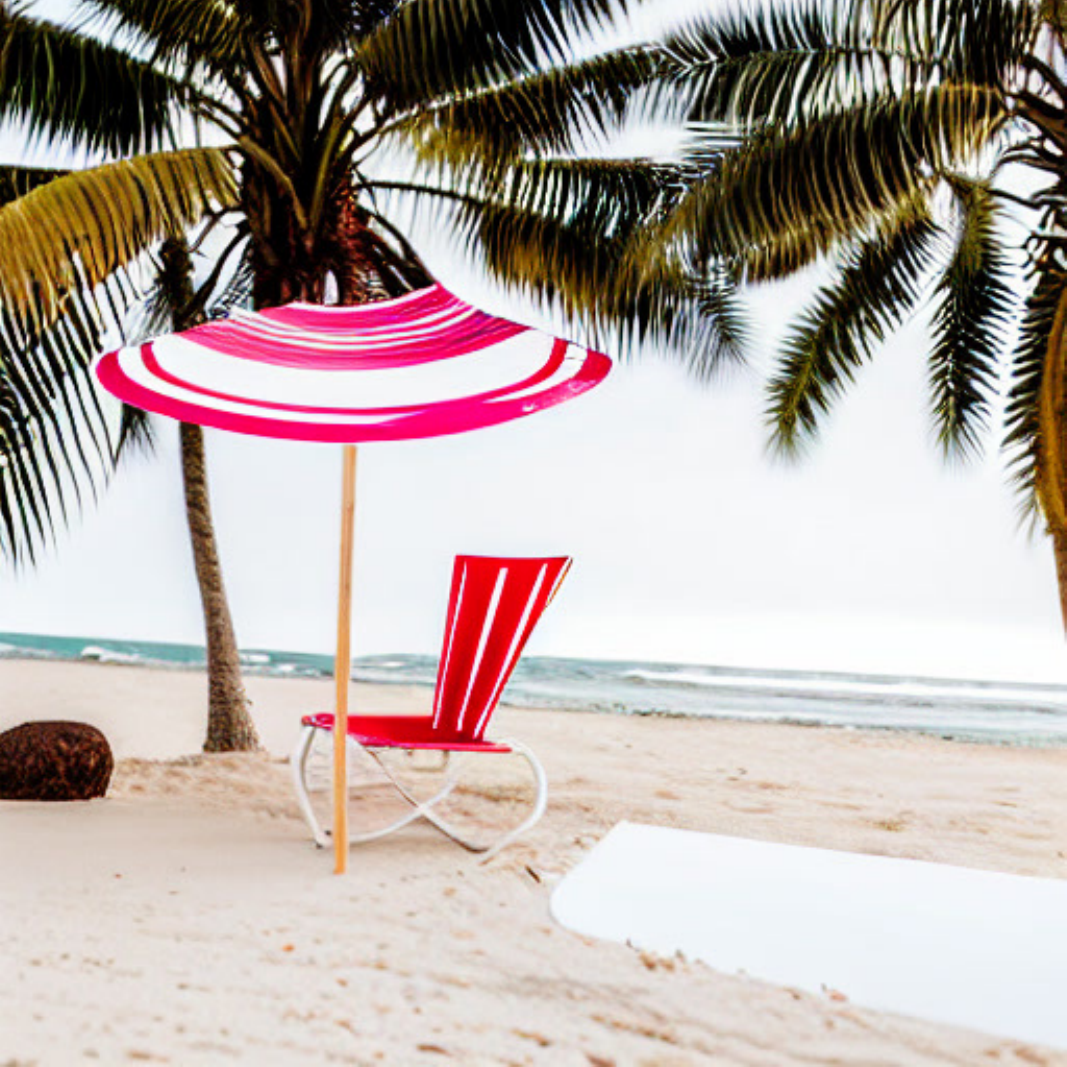}} \\

\resizebox{!}{60px}{
\begin{tabular}[x]{@{}c@{}}Add a dog \\ chasing its own \\ tail in the \\ middle of the \\ carpeted room. \end{tabular}}&
\raisebox{-.5\height}{
\includegraphics[width=0.4\linewidth]{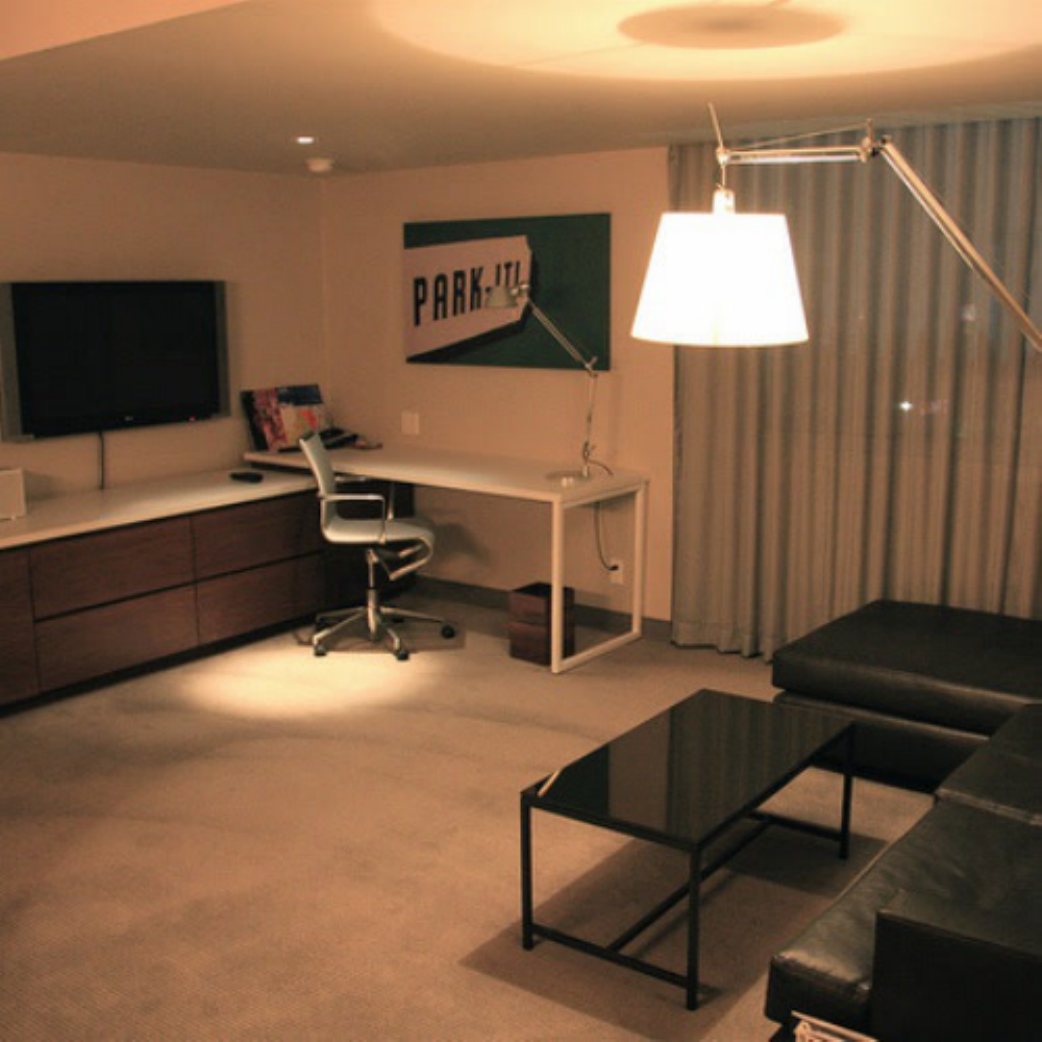}}&
\raisebox{-.5\height}{
\includegraphics[width=0.4\linewidth]{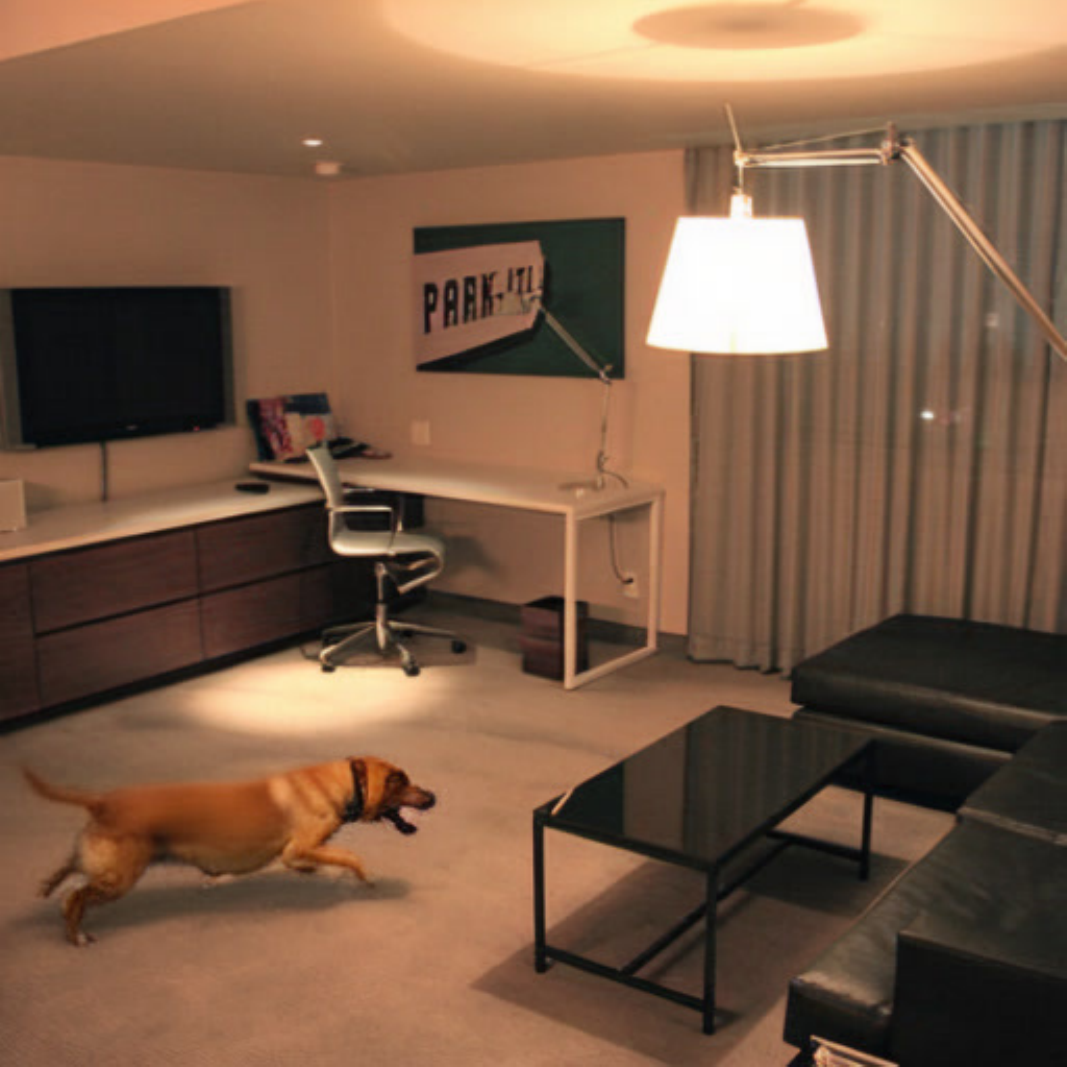}}&
\raisebox{-.5\height}{
\includegraphics[width=0.4\linewidth]{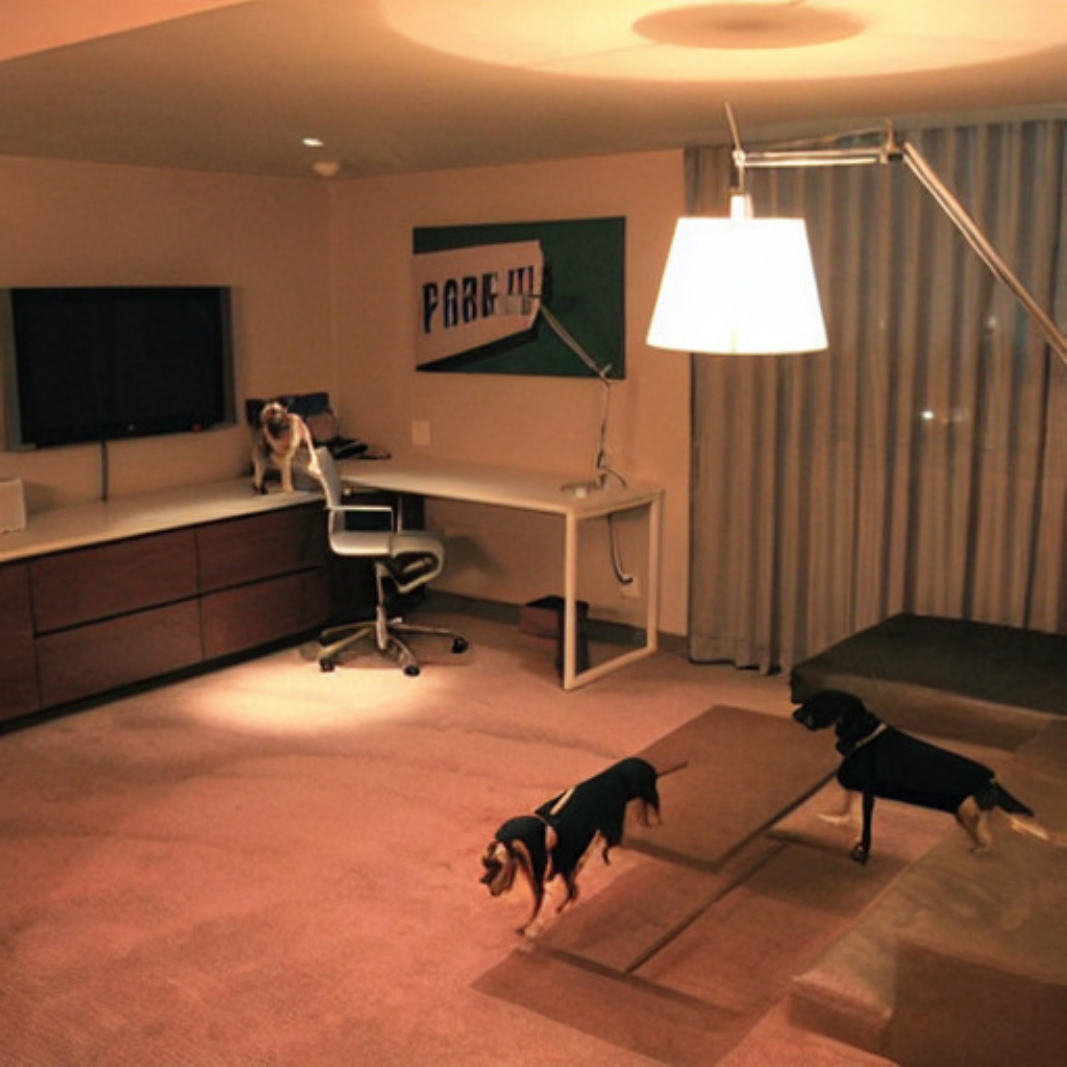}}&
\raisebox{-.5\height}{
\includegraphics[width=0.4\linewidth]{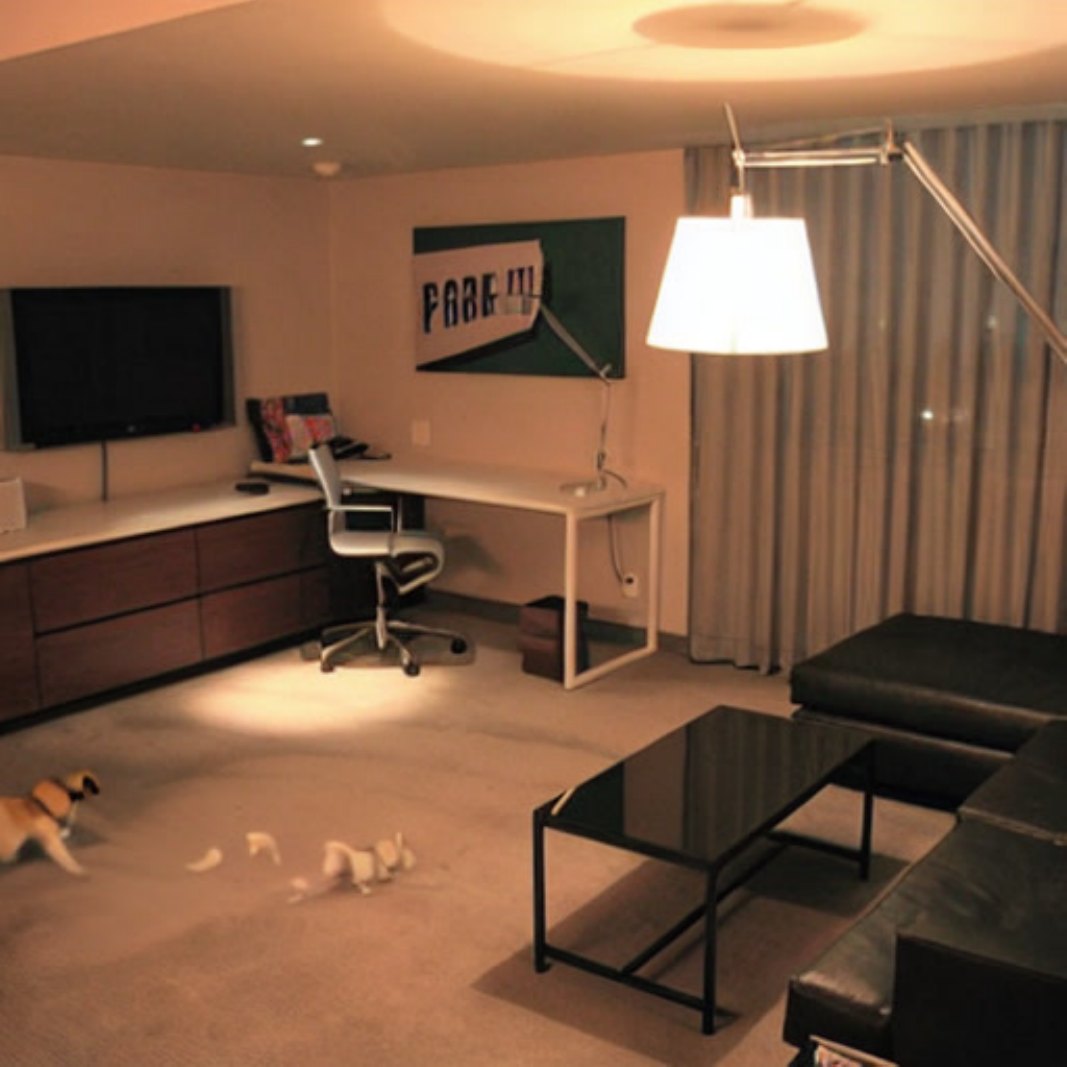}}&
\raisebox{-.5\height}{
\includegraphics[width=0.4\linewidth]{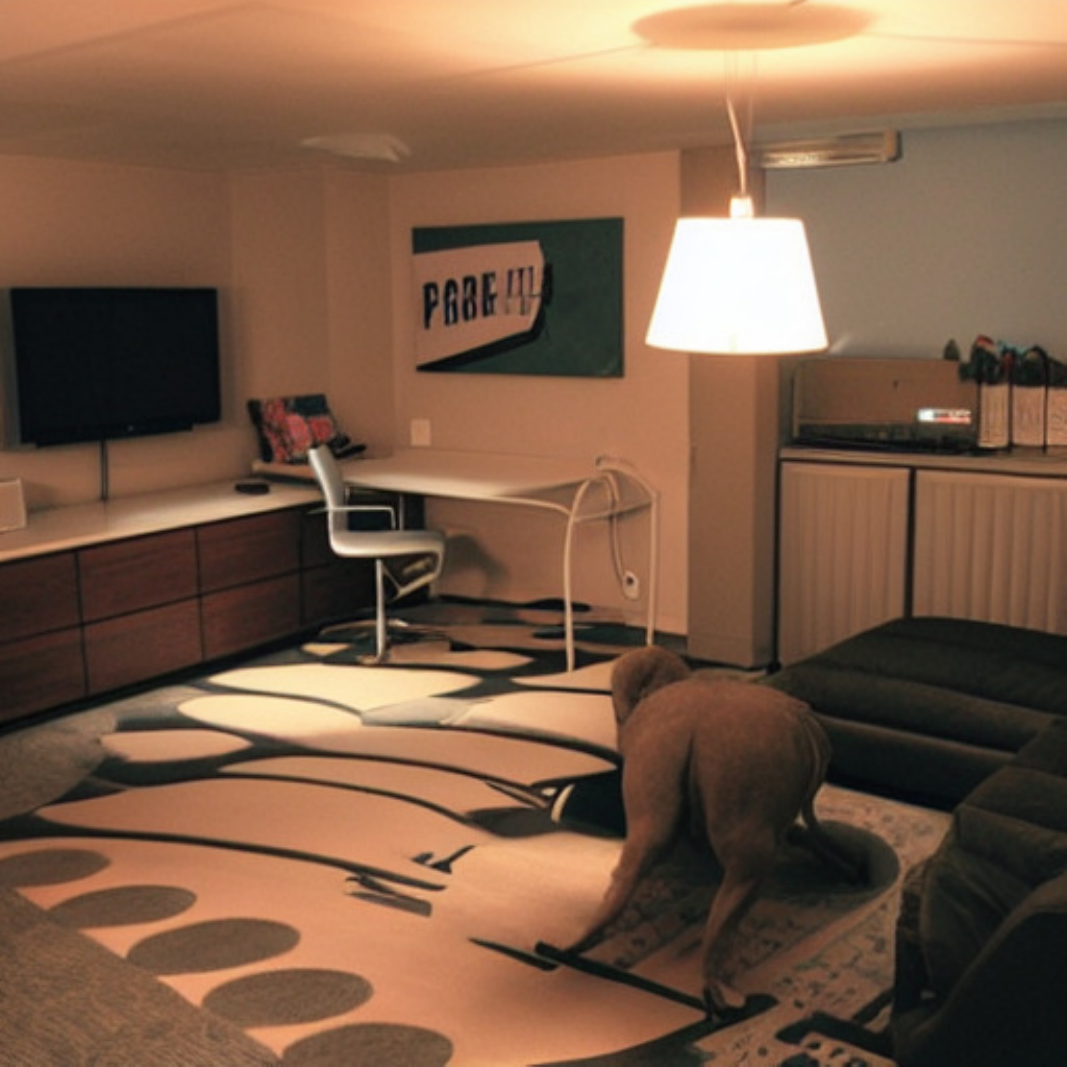}} \\

& \resizebox{!}{18px}{\begin{tabular}[x]{@{}c@{}}Original\end{tabular}}  & \resizebox{!}{18px}{\begin{tabular}[x]{@{}c@{}} \textbf{\model} \end{tabular}} & \resizebox{!}{18px}{\begin{tabular}[x]{@{}c@{}}InstructPix2Pix \end{tabular}} & \resizebox{!}{18px}{\begin{tabular}[x]{@{}c@{}} MagicBrush \end{tabular}} & \resizebox{!}{18px}
{\begin{tabular}[x]{@{}c@{}}P2P \end{tabular}}
\\
\end{tabular}
}
\caption{Qualitative comparison of our model to baselines on \model Test Set.}
\label{fig:comp_our_test} %
\end{figure*}

\begin{figure*}[h]
   \centering
\scalebox{0.45}{
\begin{tabular}{@{\hspace{-20\tabcolsep}}c@{\hspace{0.1\tabcolsep}}c@{\hspace{0.5\tabcolsep}}c@{\hspace{0.2\tabcolsep}}c@{\hspace{0.2\tabcolsep}}c@{\hspace{0.2\tabcolsep}}c}

\resizebox{!}{50px}{
\begin{tabular}[x]{@{}c@{}}Turn the \\ refrigerator into \\a bookshelf with \\ books\end{tabular}}&
\raisebox{-.5\height}{
\includegraphics[width=0.4\linewidth]{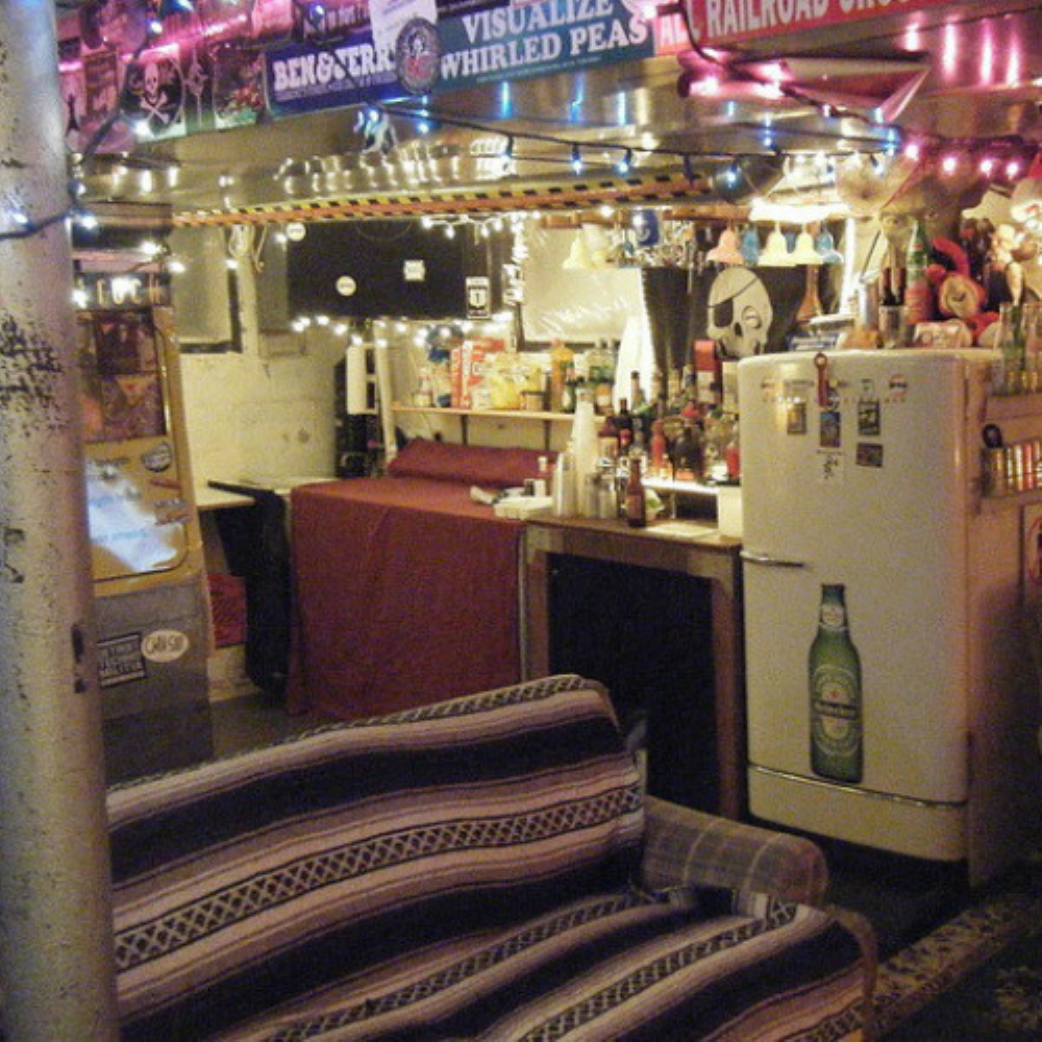}}&
\raisebox{-.5\height}{
\includegraphics[width=0.4\linewidth]{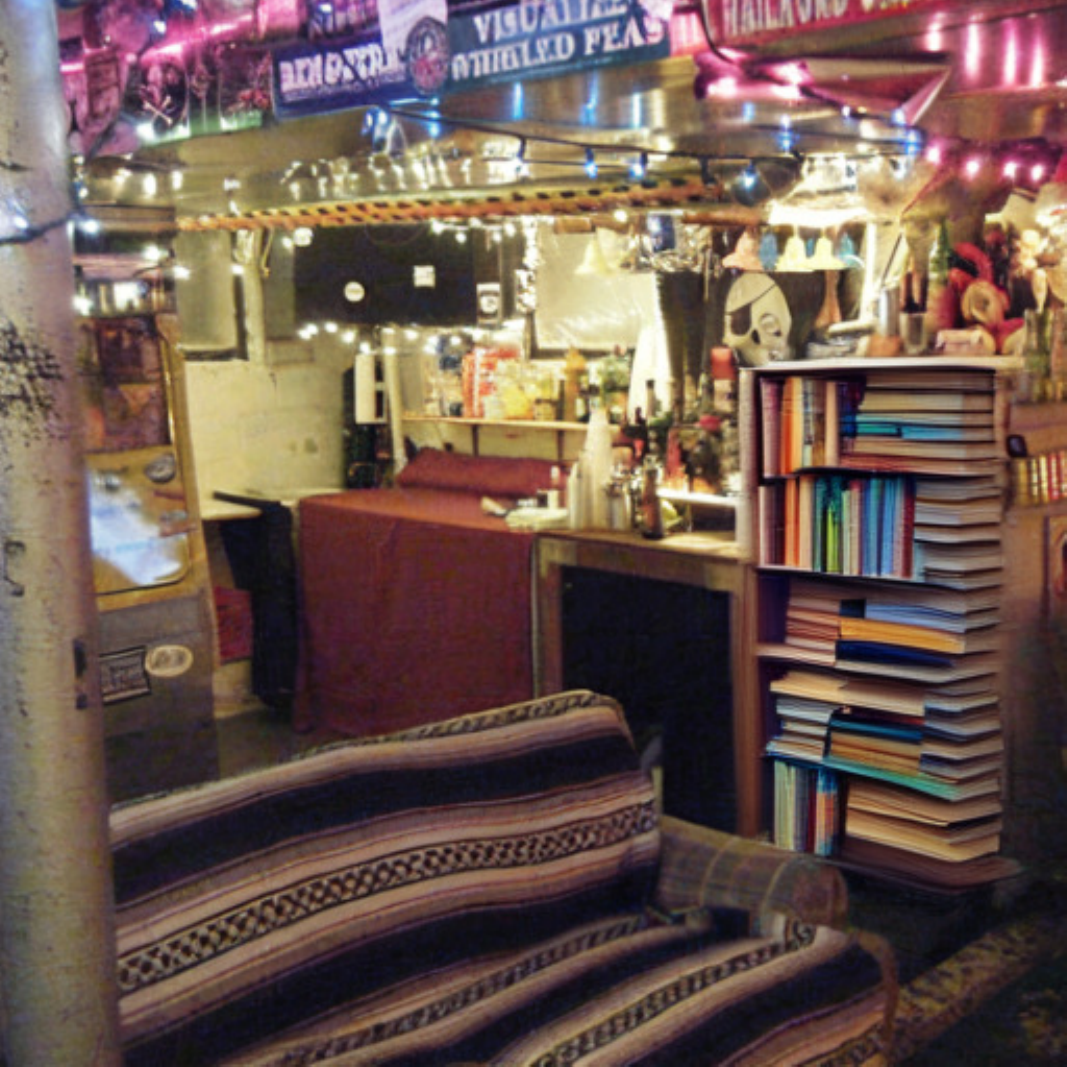}}&
\raisebox{-.5\height}{
\includegraphics[width=0.4\linewidth]{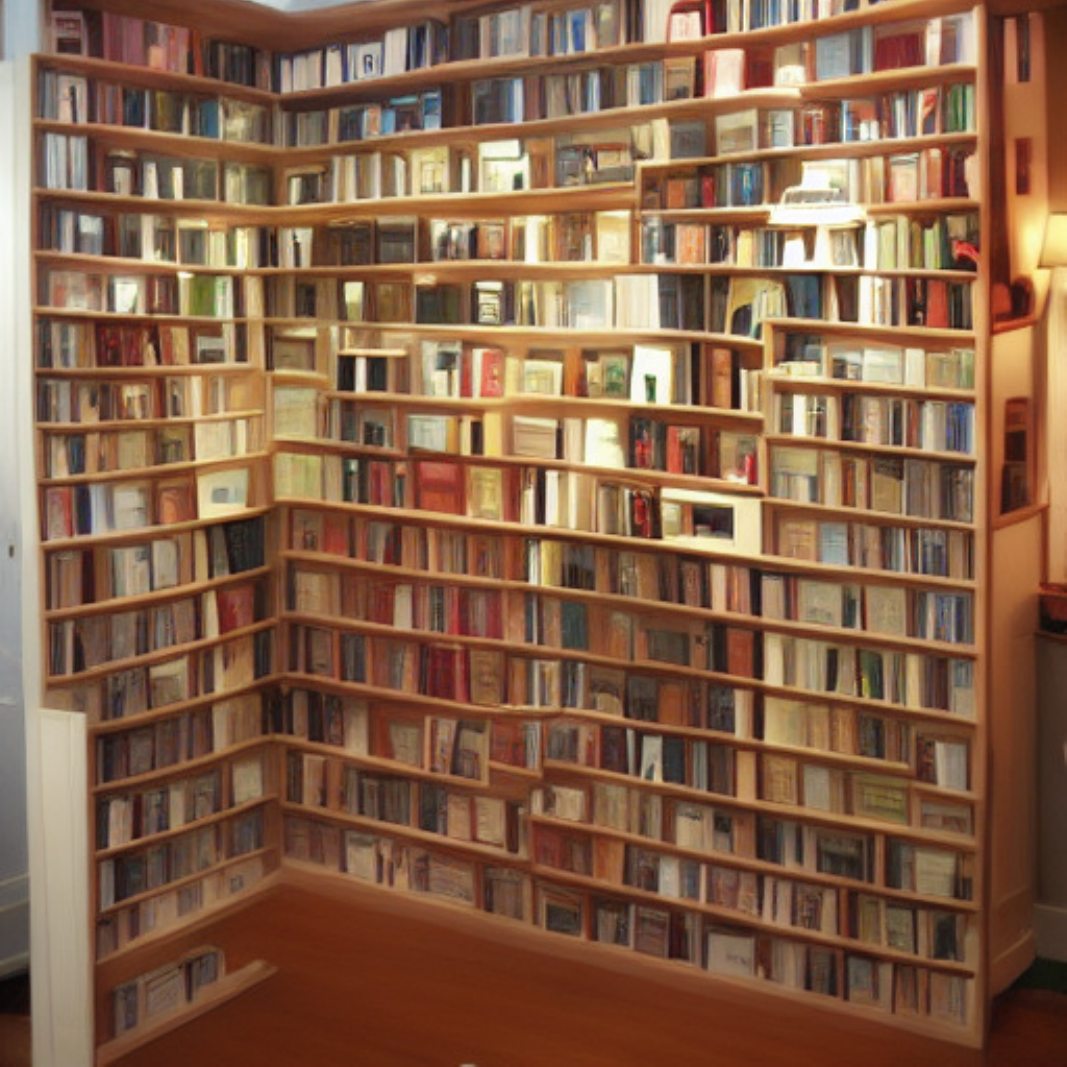}}&
\raisebox{-.5\height}{
\includegraphics[width=0.4\linewidth]{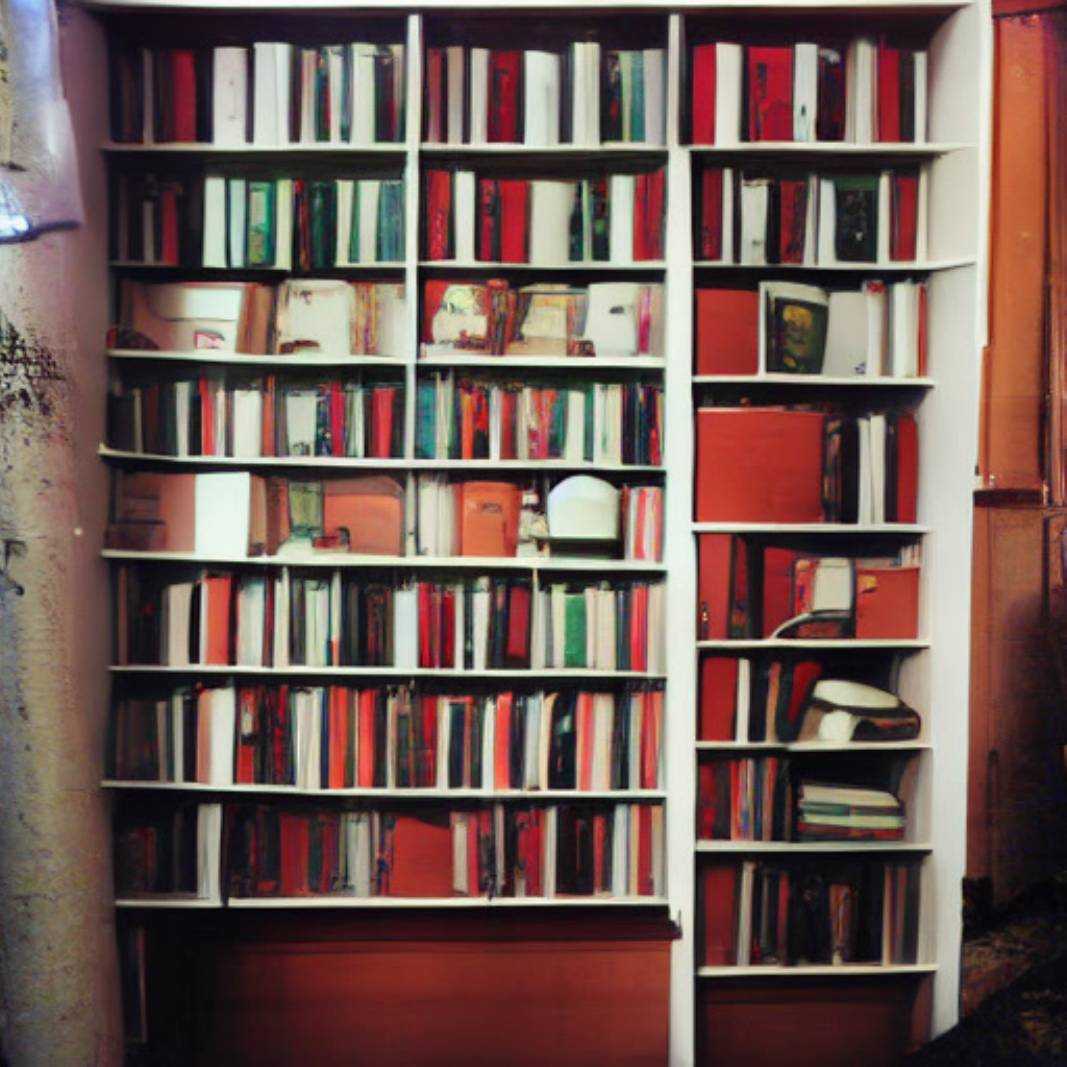}}&
\raisebox{-.5\height}{
\includegraphics[width=0.4\linewidth]{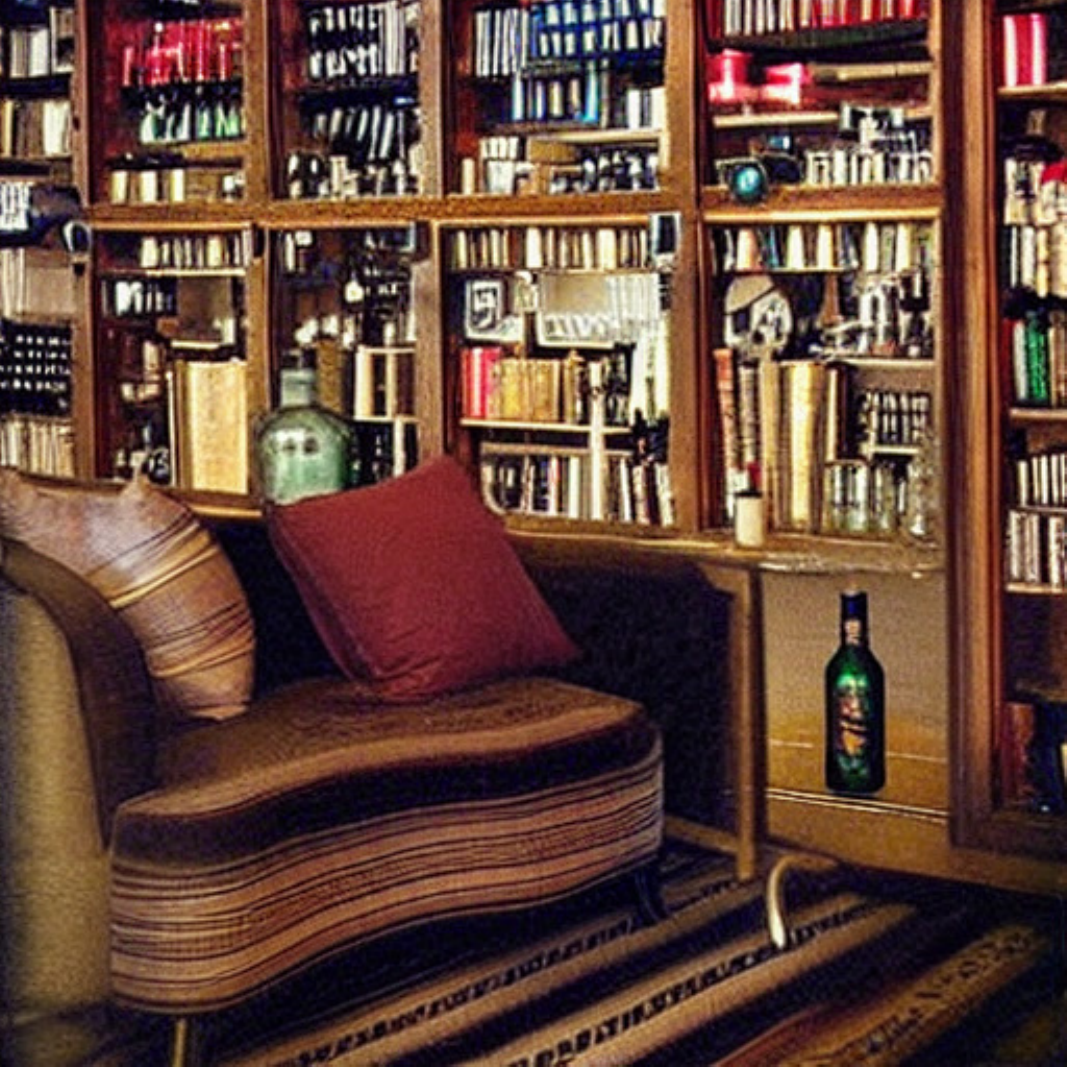}} \\

\resizebox{!}{40px}{
\begin{tabular}[x]{@{}c@{}} Change the image \\ to have a 1970s \\ pop art style.\end{tabular}}&
\raisebox{-.5\height}{
\includegraphics[width=0.4\linewidth]{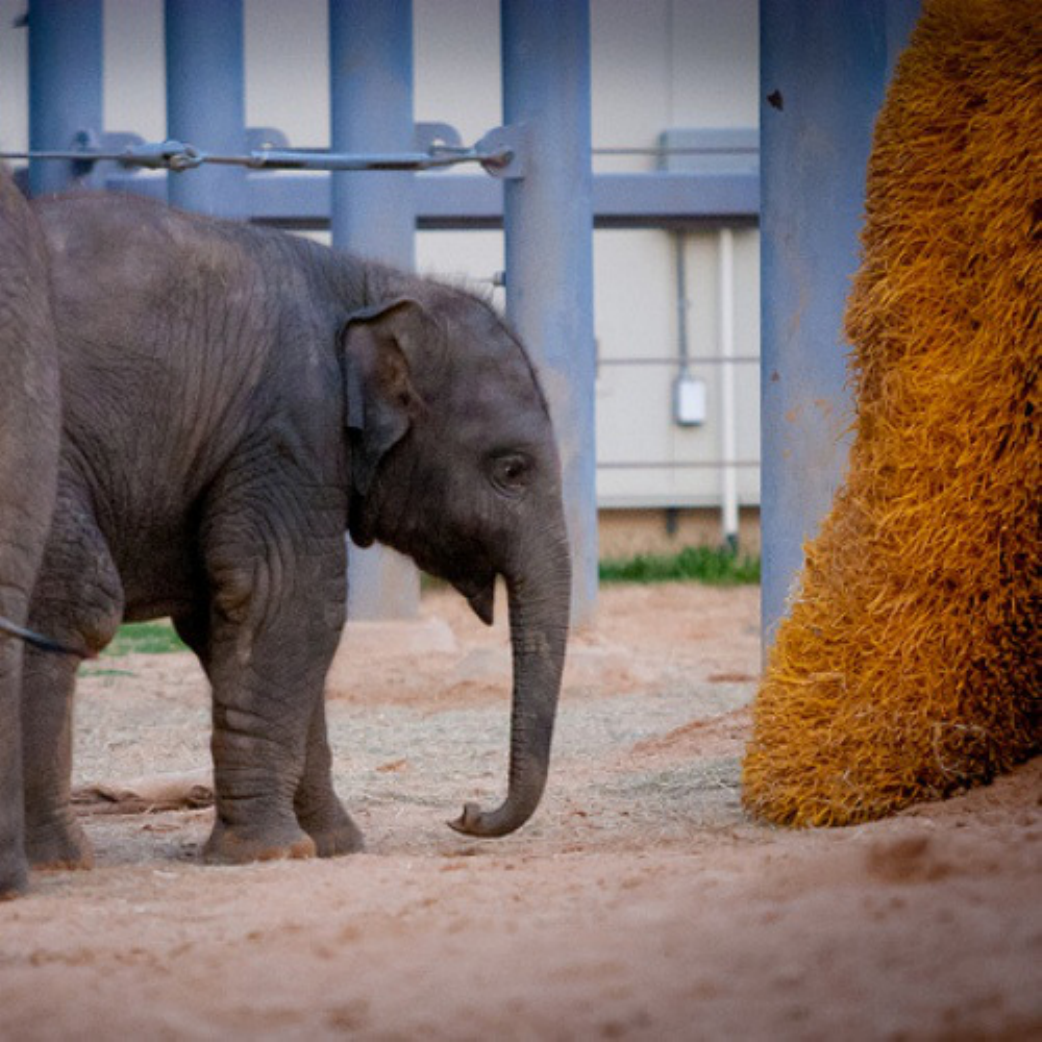}}&
\raisebox{-.5\height}{
\includegraphics[width=0.4\linewidth]{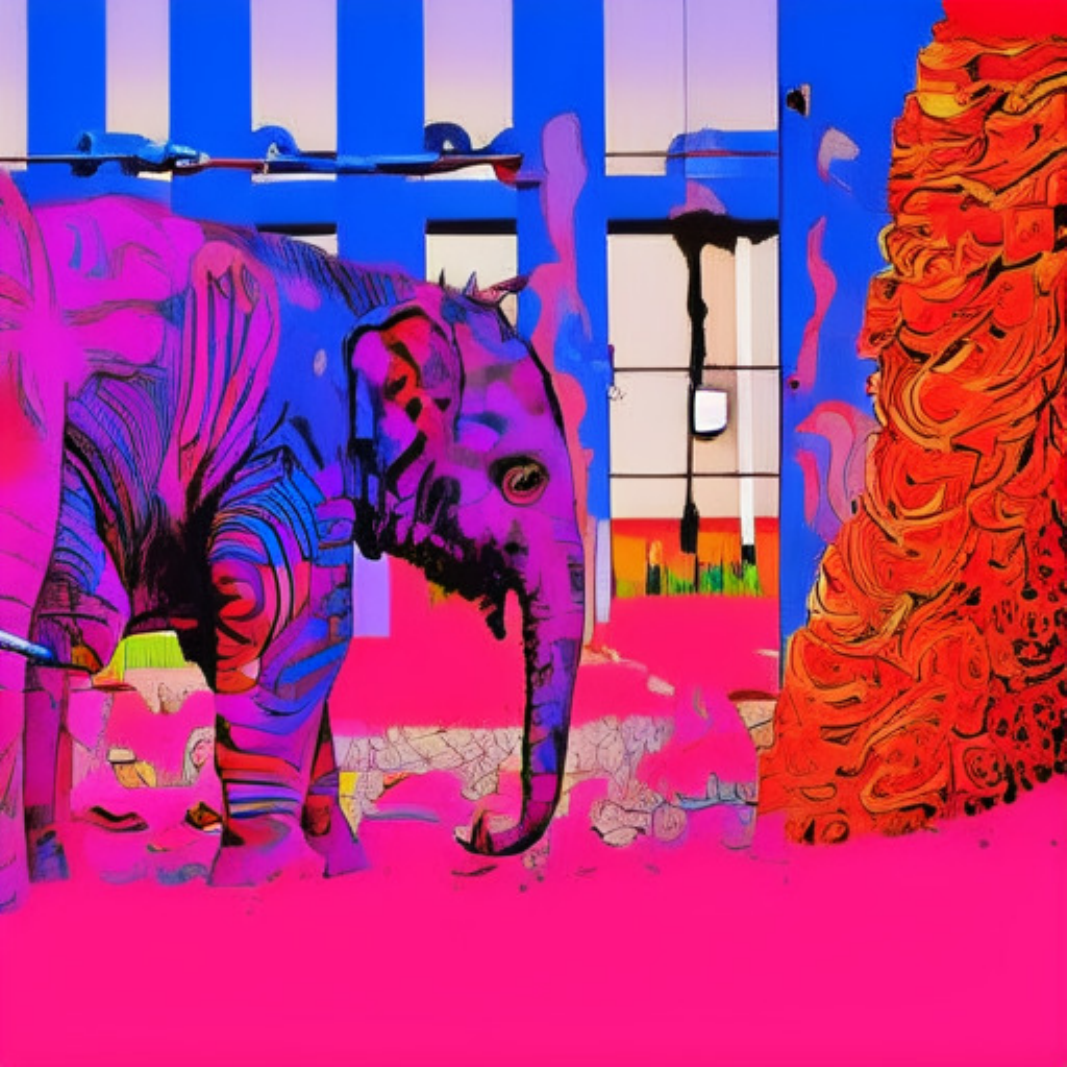}}&
\raisebox{-.5\height}{
\includegraphics[width=0.4\linewidth]{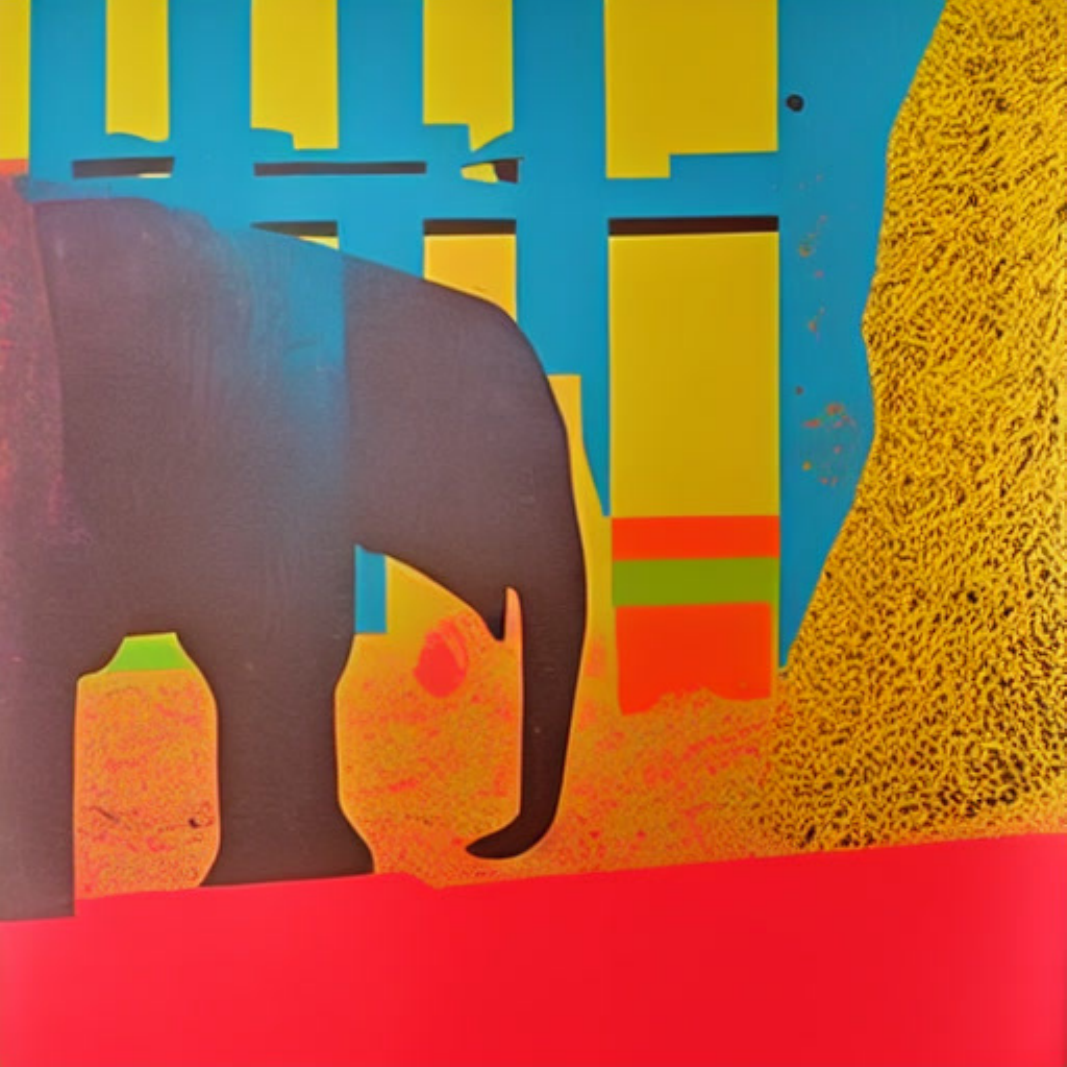}}&
\raisebox{-.5\height}{
\includegraphics[width=0.4\linewidth]{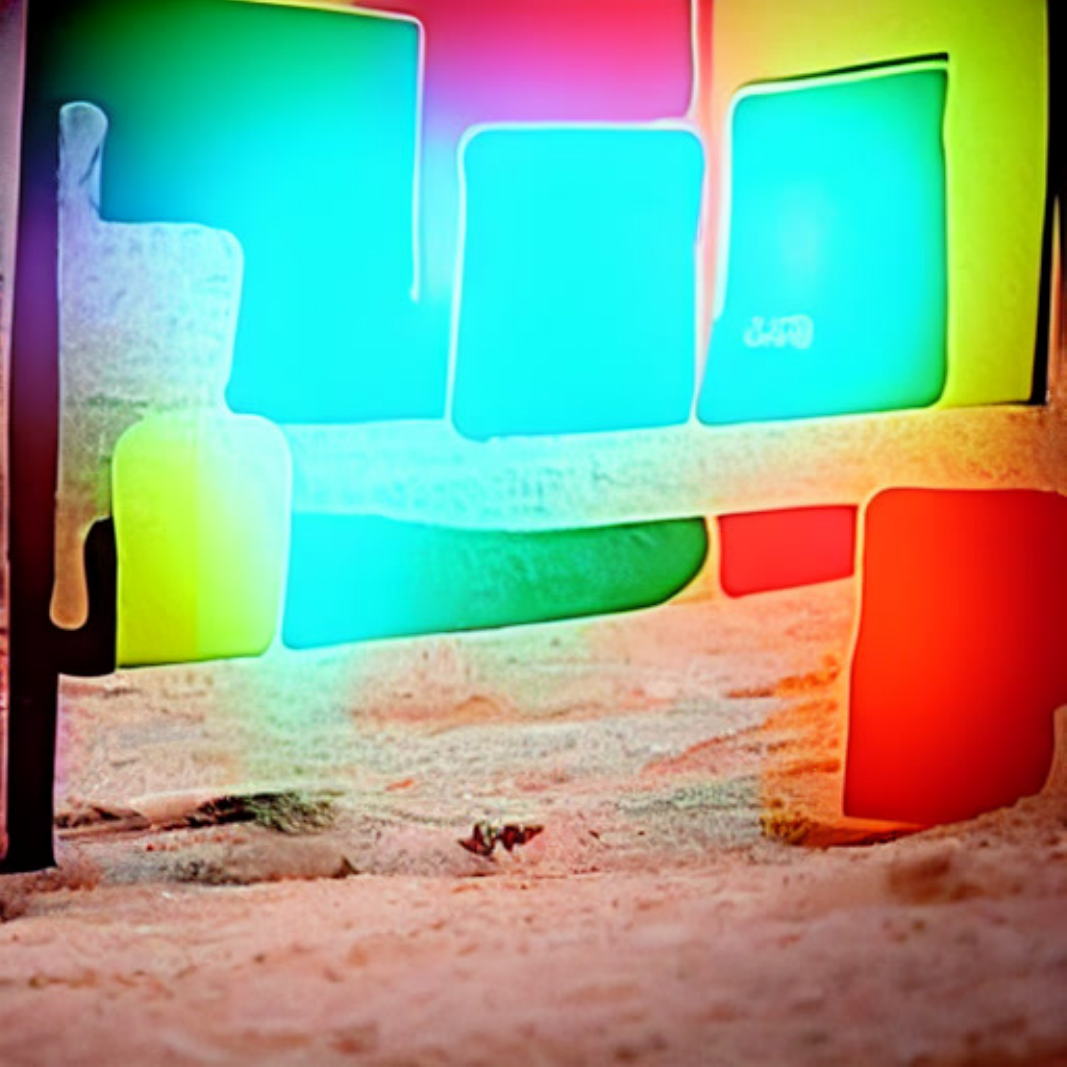}}&
\raisebox{-.5\height}{
\includegraphics[width=0.4\linewidth]{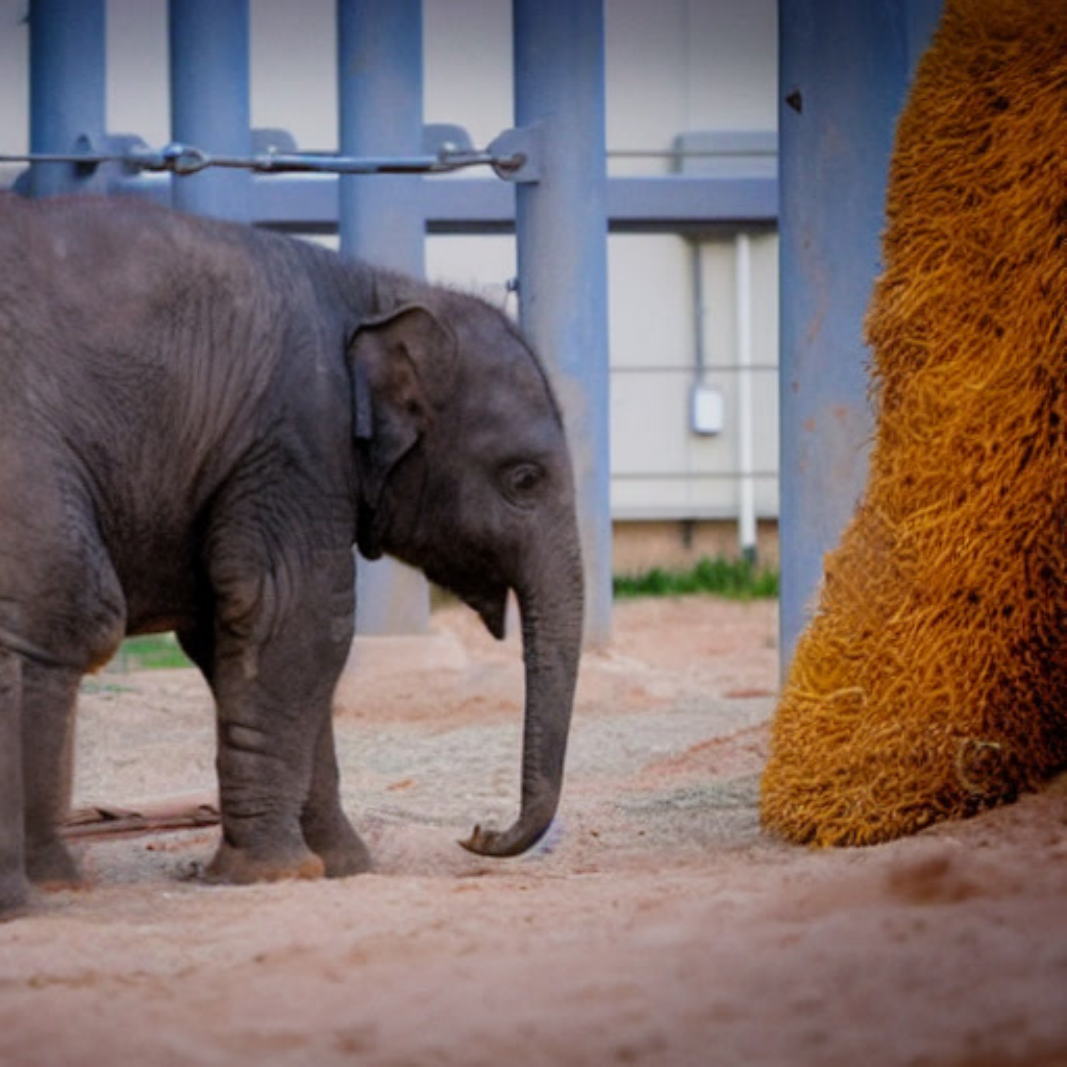}} \\

\resizebox{!}{40px}{
\begin{tabular}[x]{@{}c@{}} Remove the \\ Christmas trees \\ on the table\end{tabular}}&
\raisebox{-.5\height}{
\includegraphics[width=0.4\linewidth]{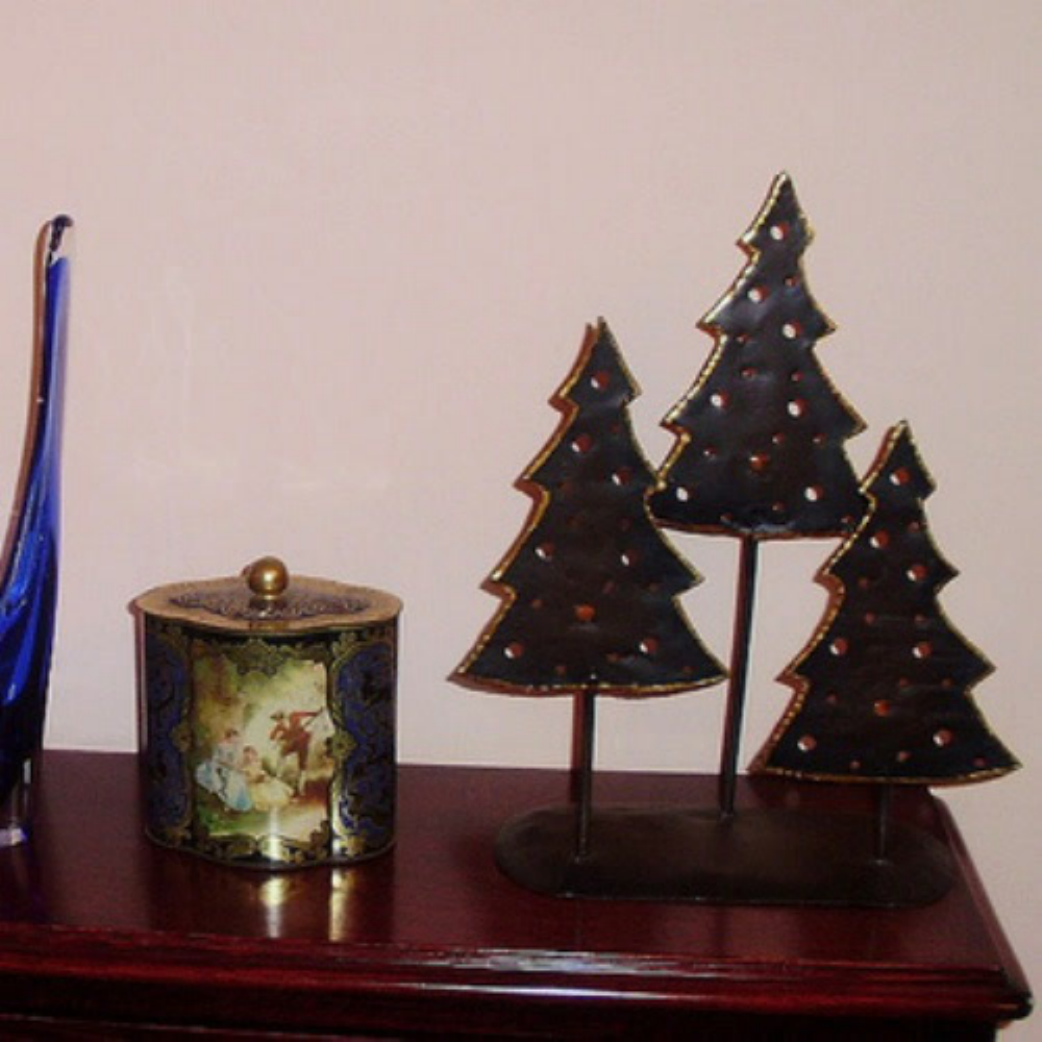}}&
\raisebox{-.5\height}{
\includegraphics[width=0.4\linewidth]{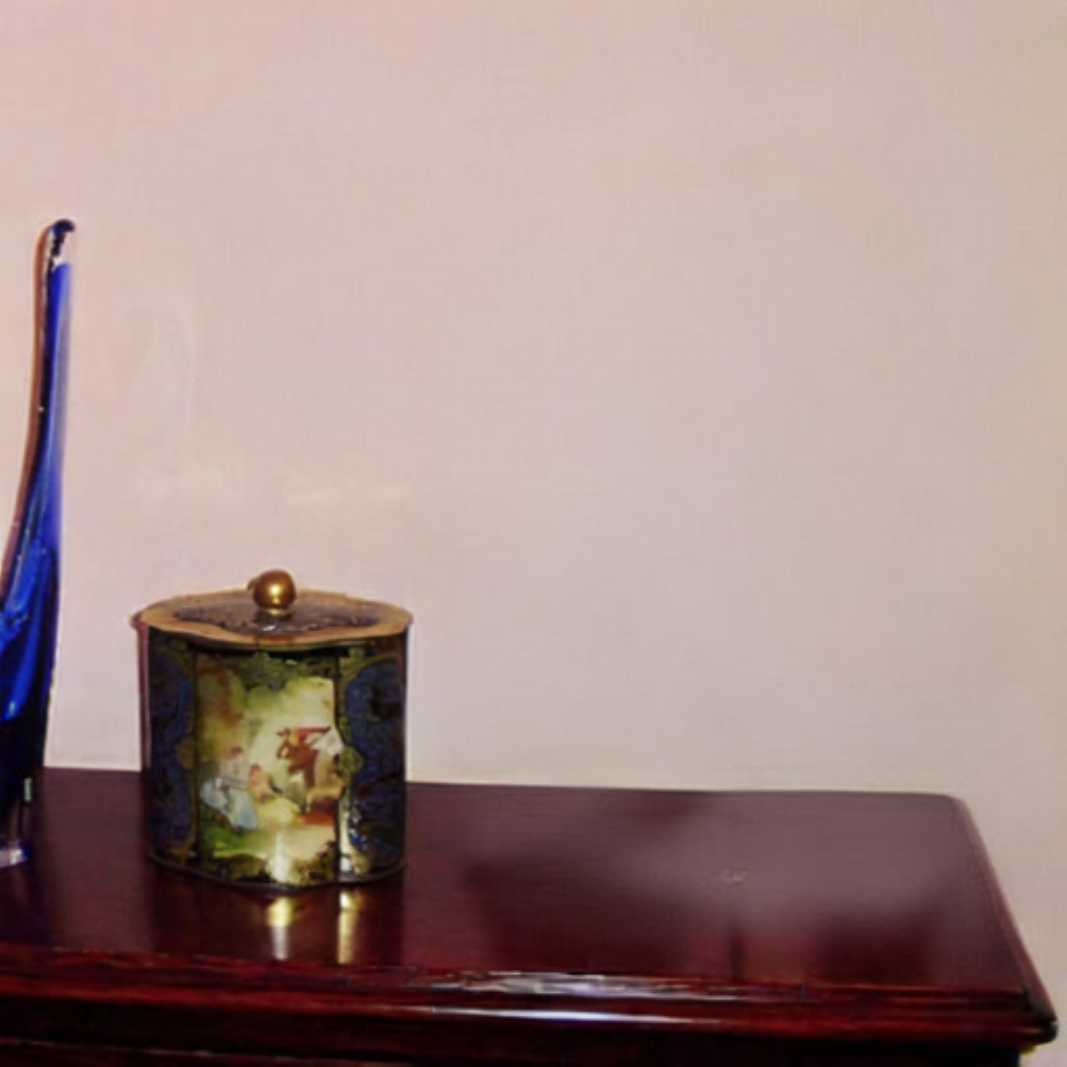}}&
\raisebox{-.5\height}{
\includegraphics[width=0.4\linewidth]{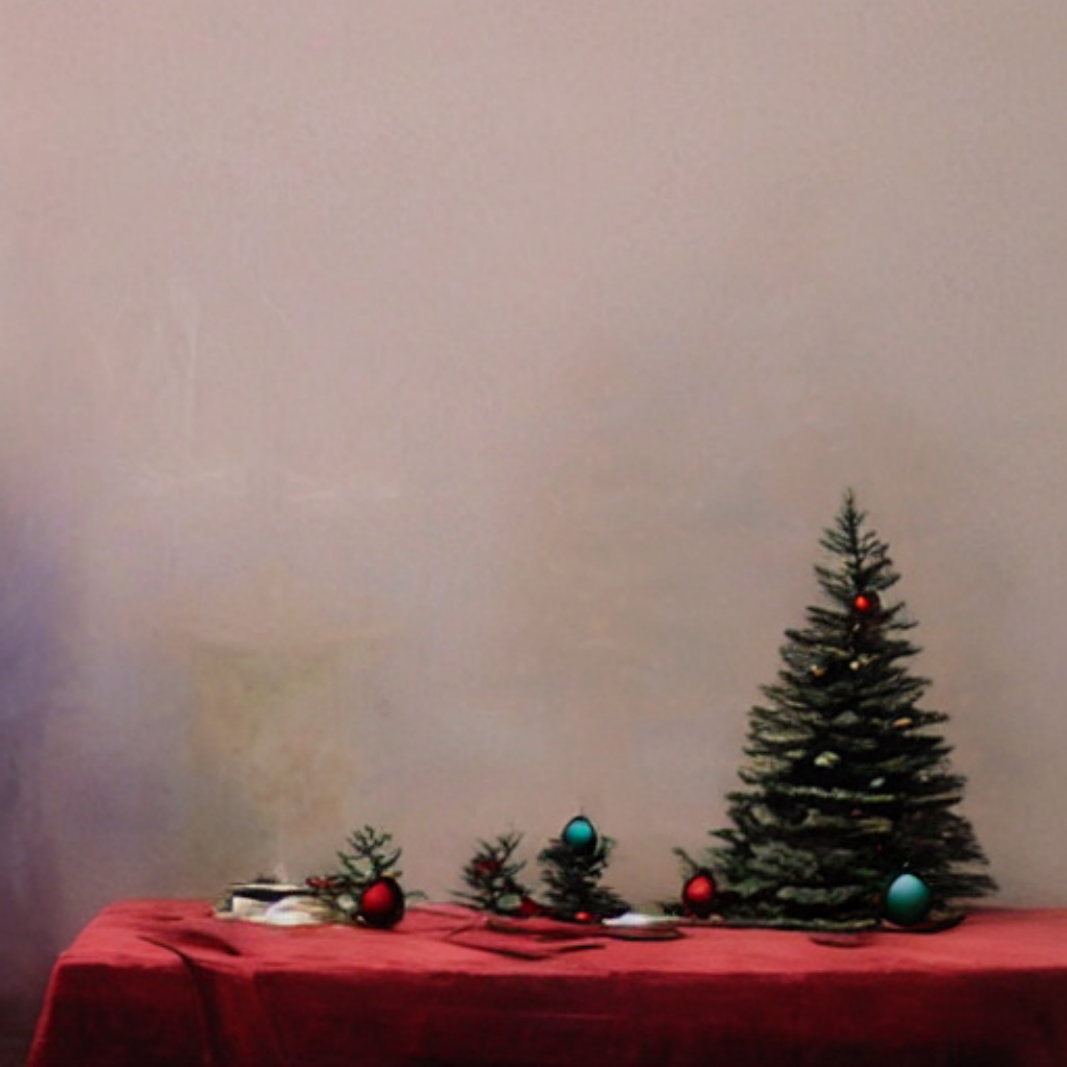}}&
\raisebox{-.5\height}{
\includegraphics[width=0.4\linewidth]{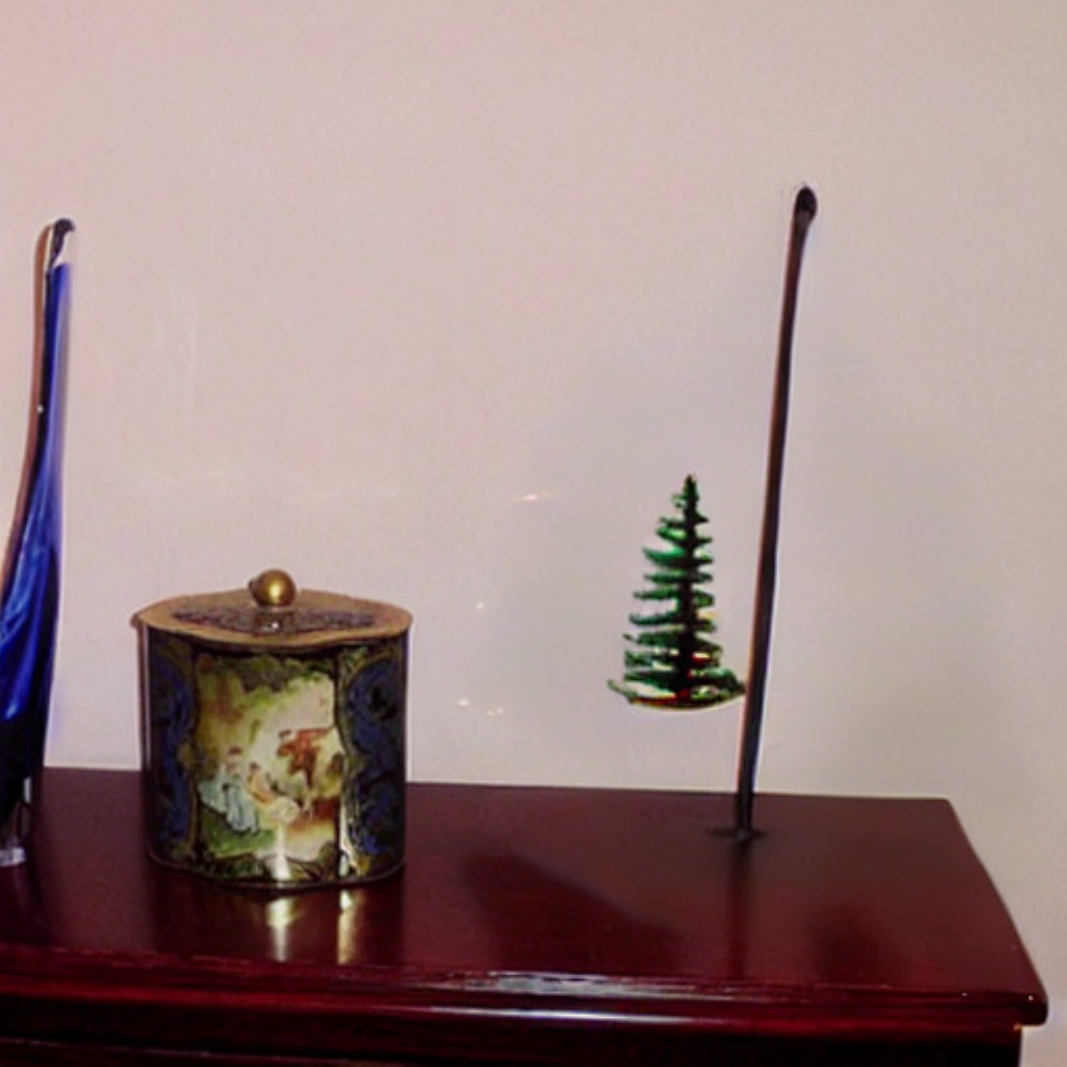}}&
\raisebox{-.5\height}{
\includegraphics[width=0.4\linewidth]{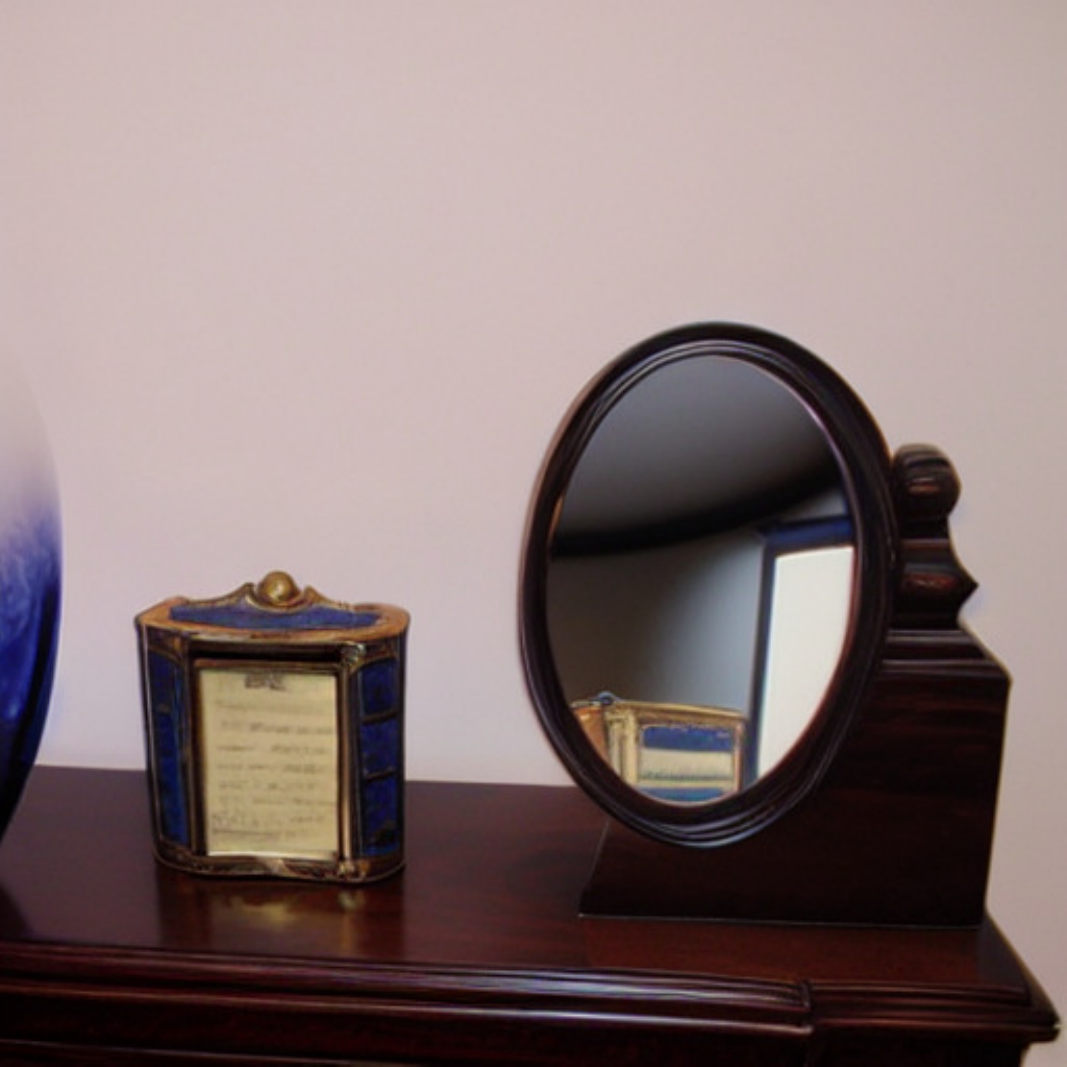}} \\

\resizebox{!}{40px}{
\begin{tabular}[x]{@{}c@{}} Put Stone Henge \\ as the background \\ of the scene. \end{tabular}}&
\raisebox{-.5\height}{
\includegraphics[width=0.4\linewidth]{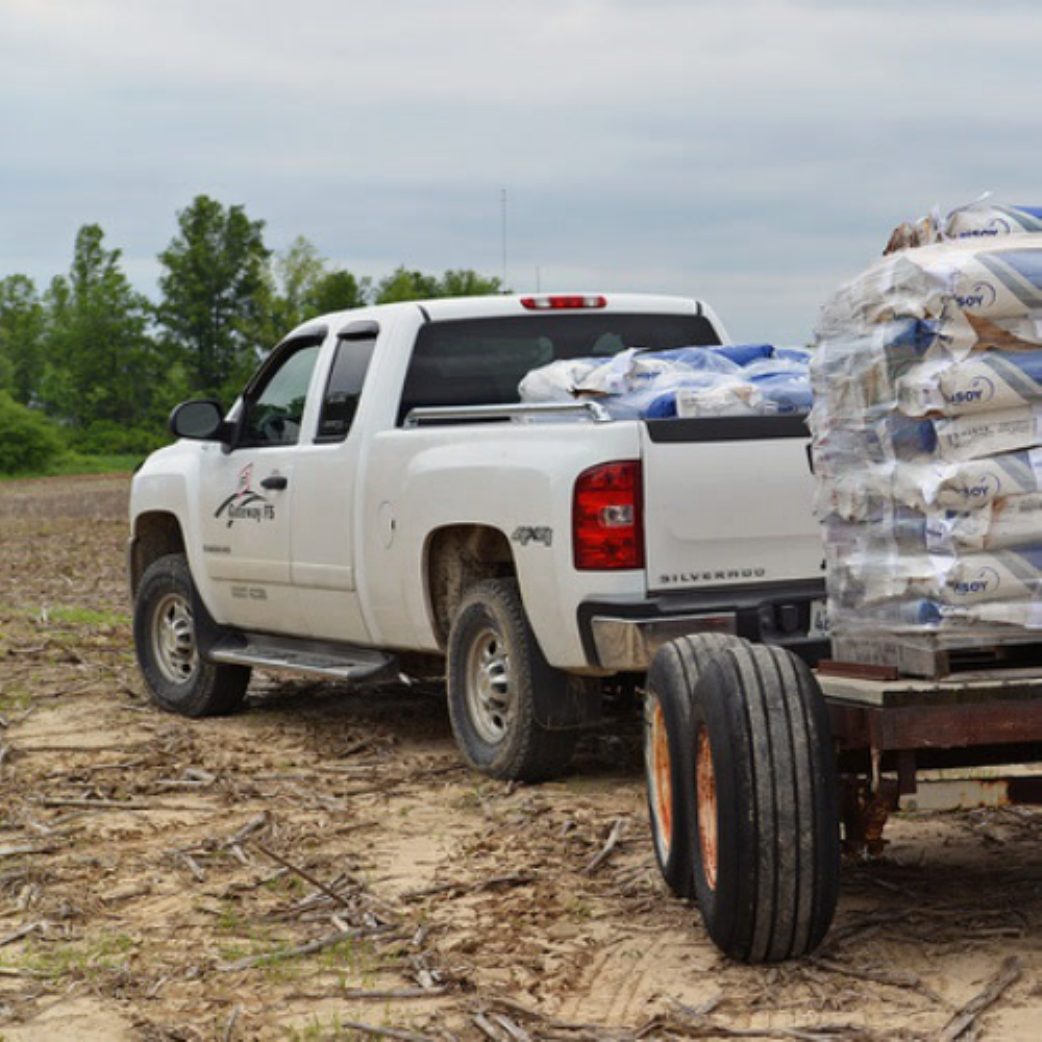}}&
\raisebox{-.5\height}{
\includegraphics[width=0.4\linewidth]{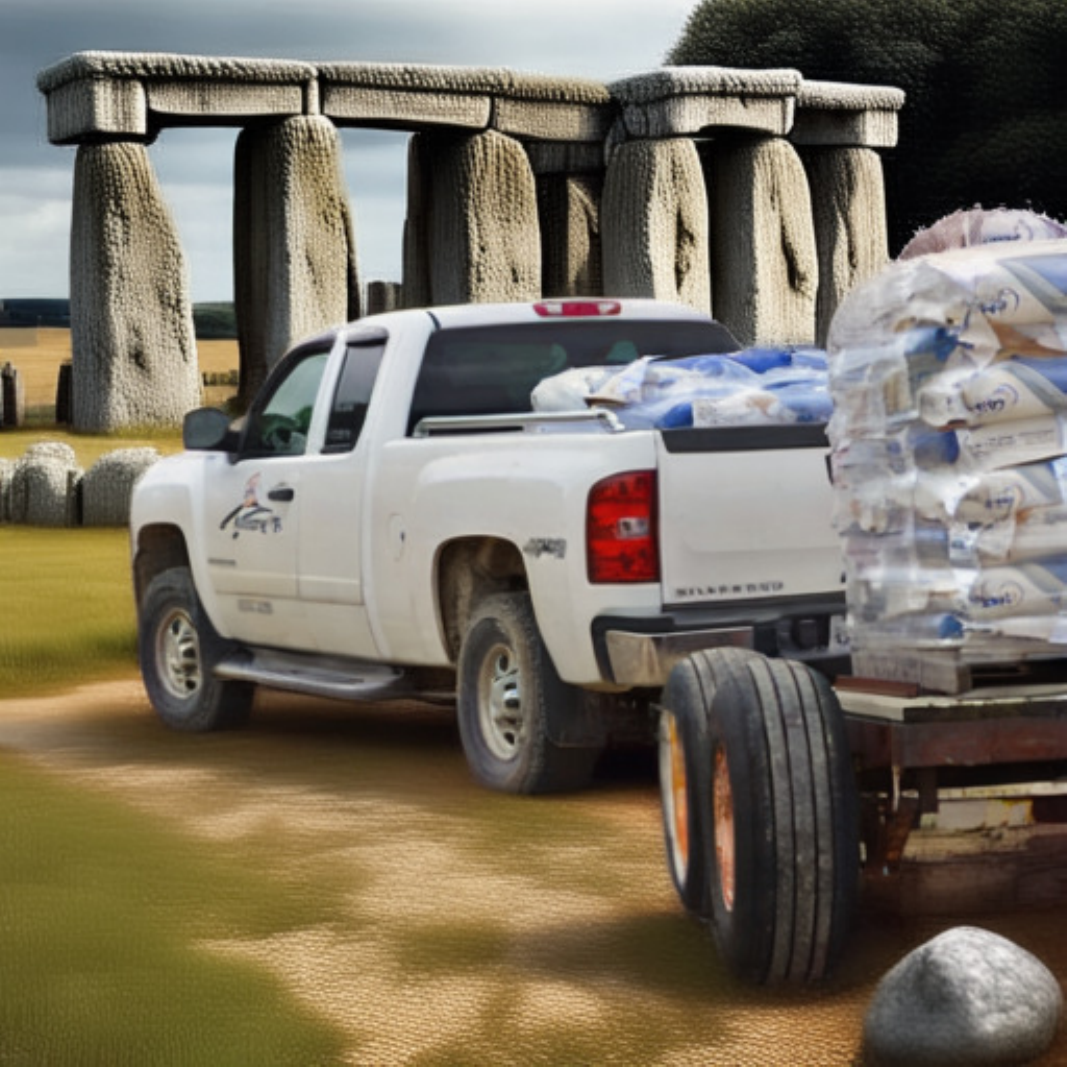}}&
\raisebox{-.5\height}{
\includegraphics[width=0.4\linewidth]{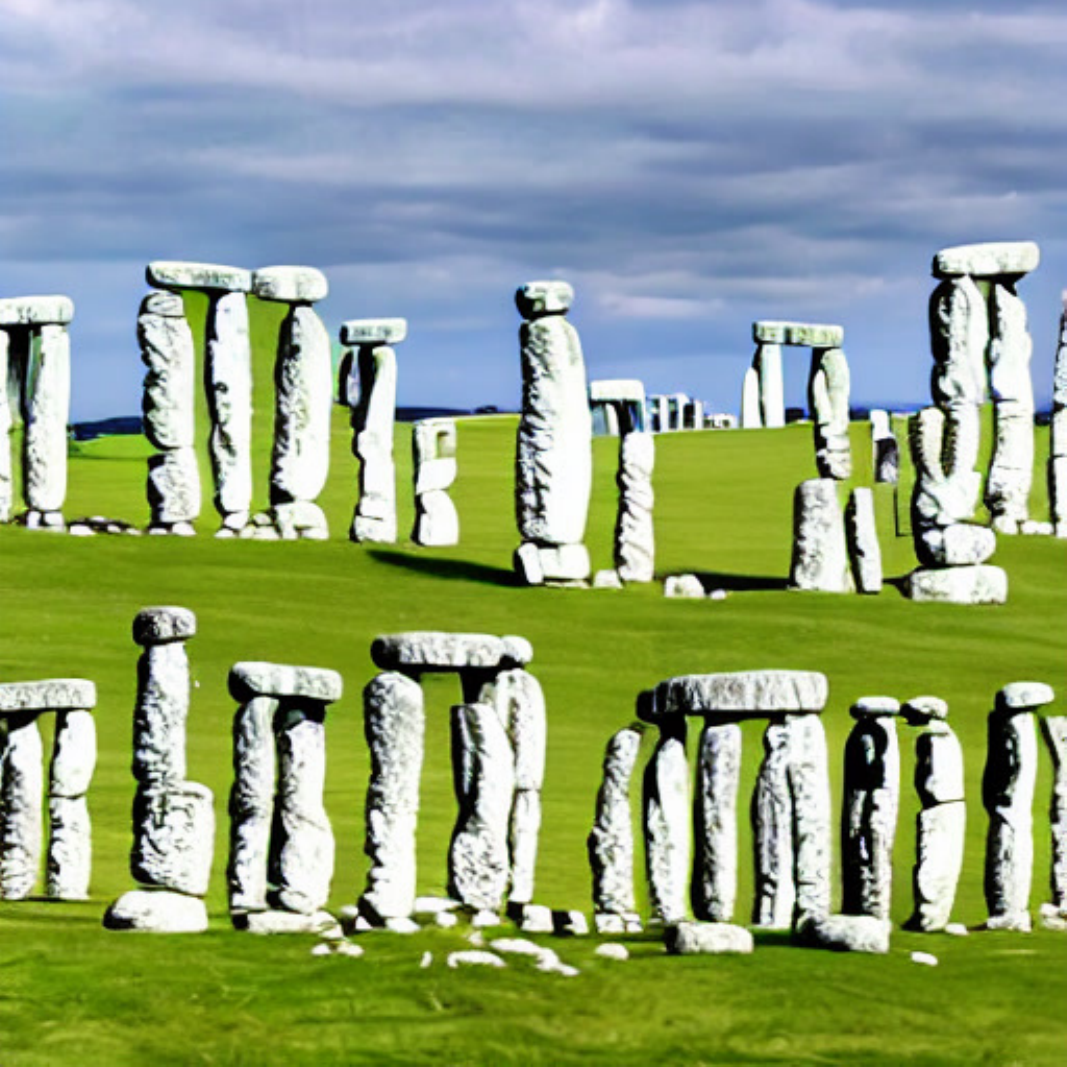}}&
\raisebox{-.5\height}{
\includegraphics[width=0.4\linewidth]{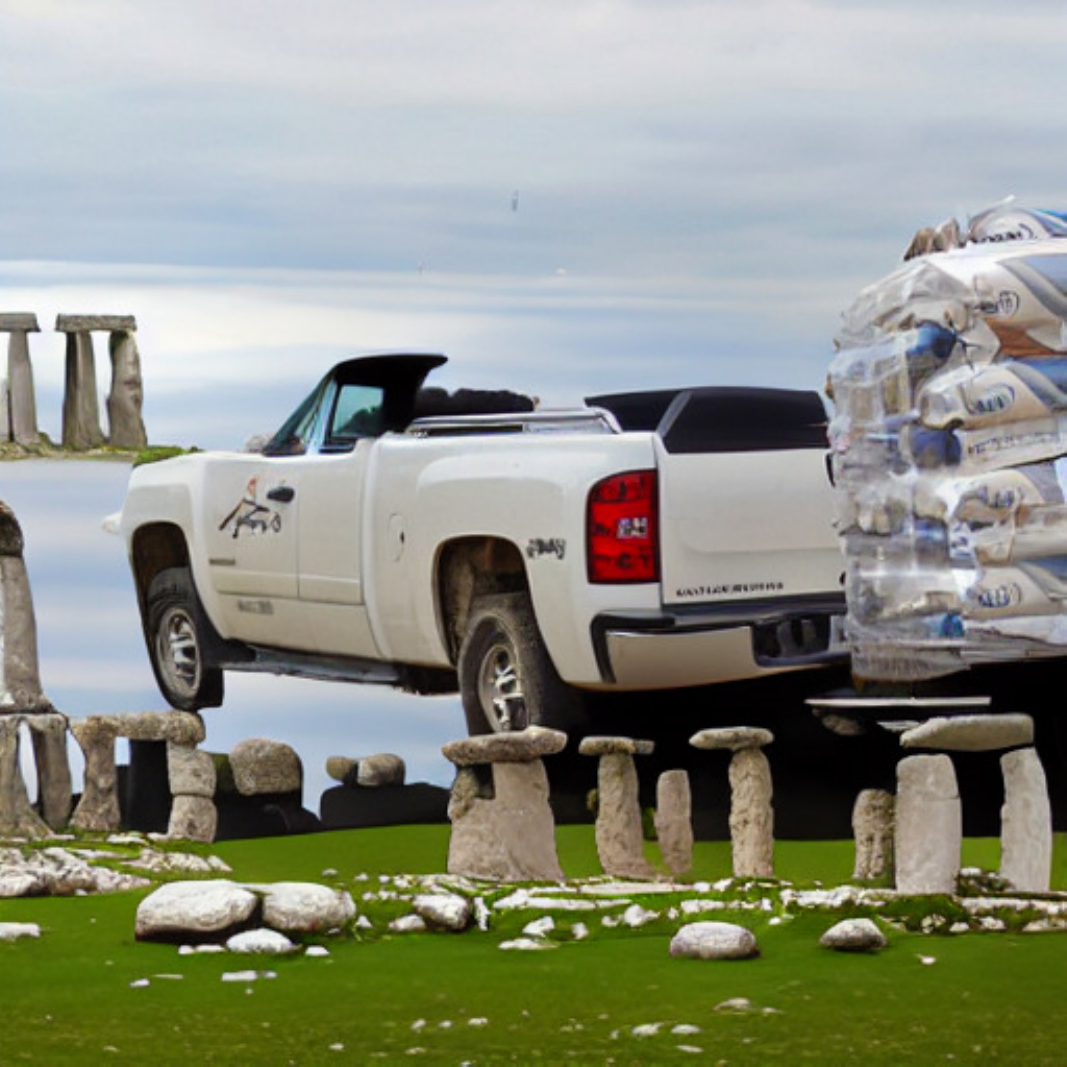}}&
\raisebox{-.5\height}{
\includegraphics[width=0.4\linewidth]{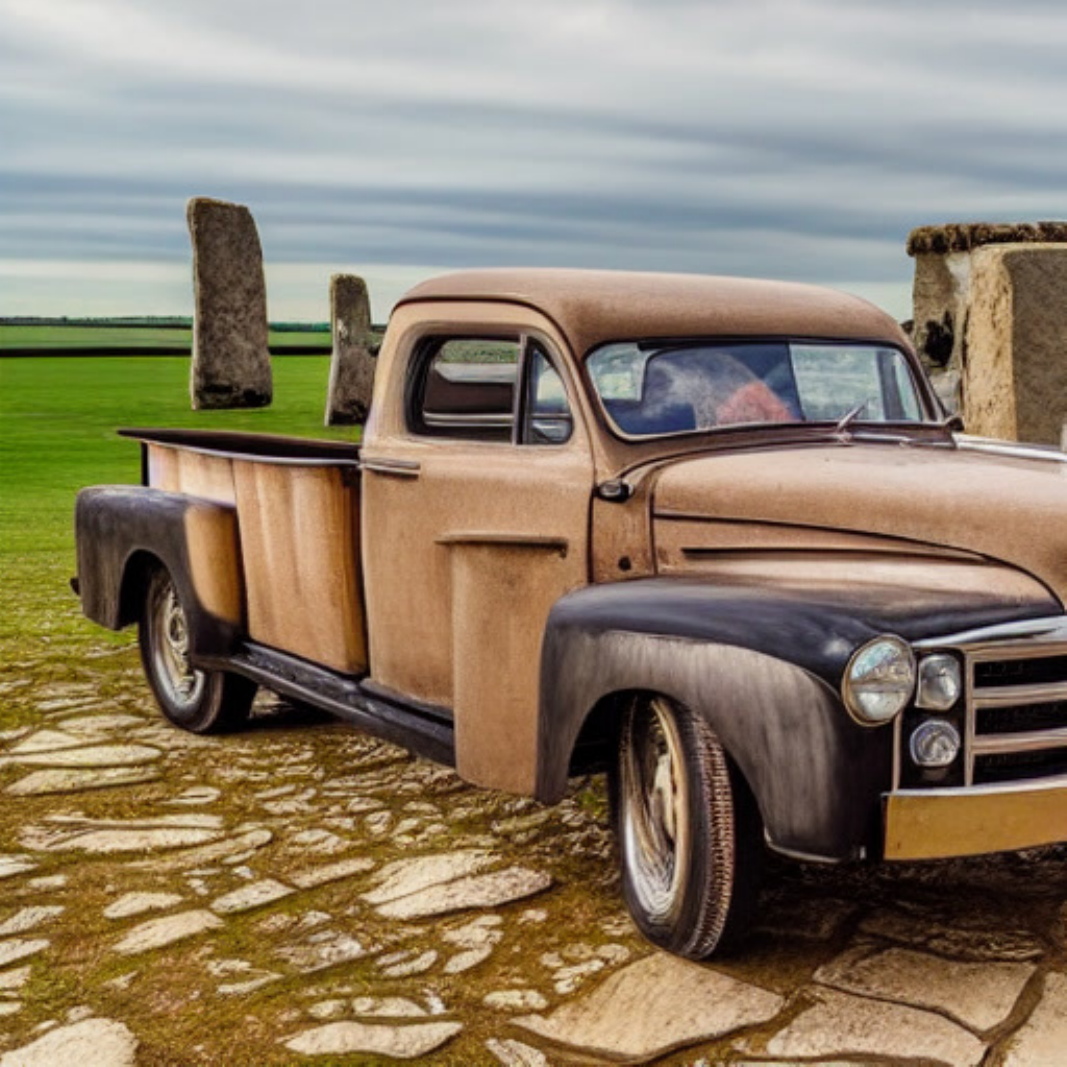}} \\

\resizebox{!}{50px}{
\begin{tabular}[x]{@{}c@{}} Change the \\color of the \\ lighthouse into\\  completely red.\end{tabular}}&
\raisebox{-.5\height}{
\includegraphics[width=0.4\linewidth]{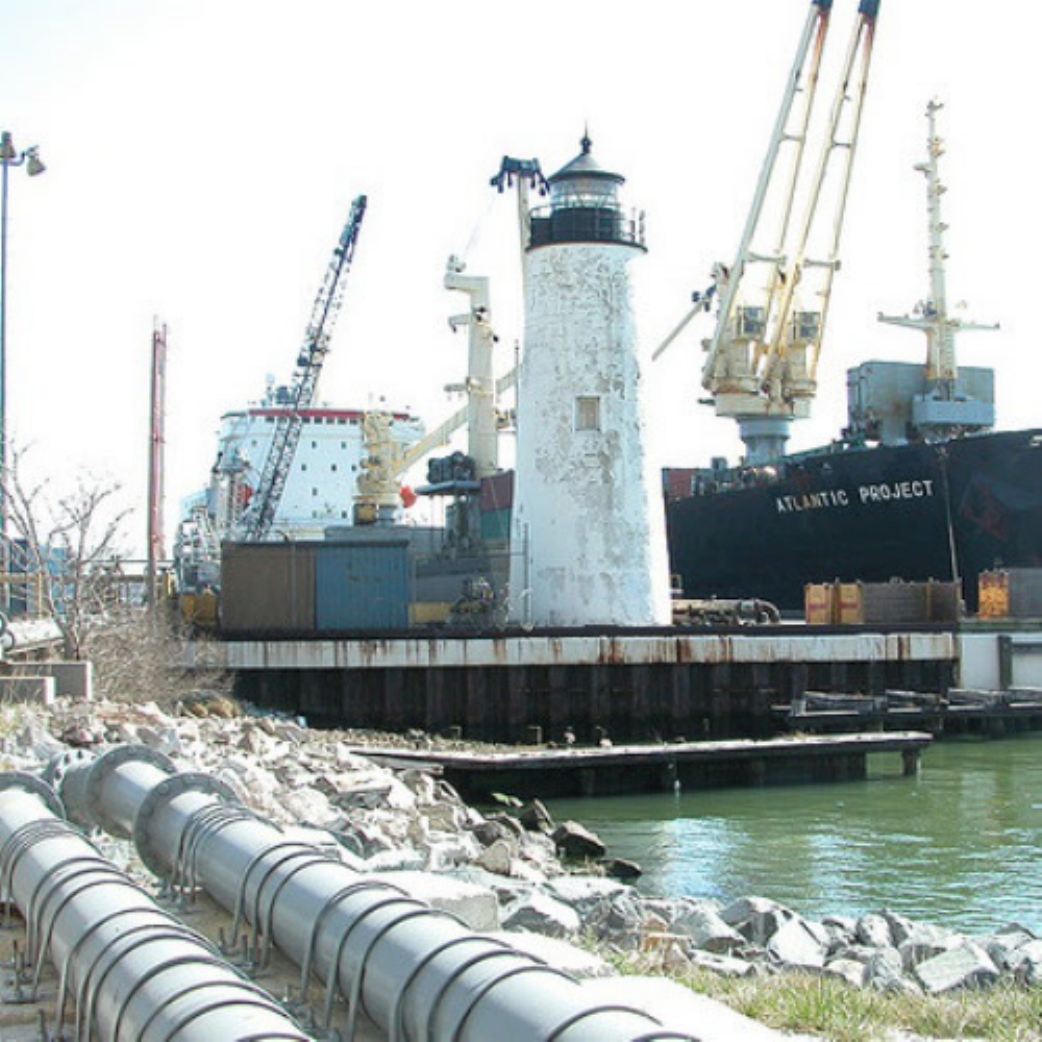}}&
\raisebox{-.5\height}{
\includegraphics[width=0.4\linewidth]{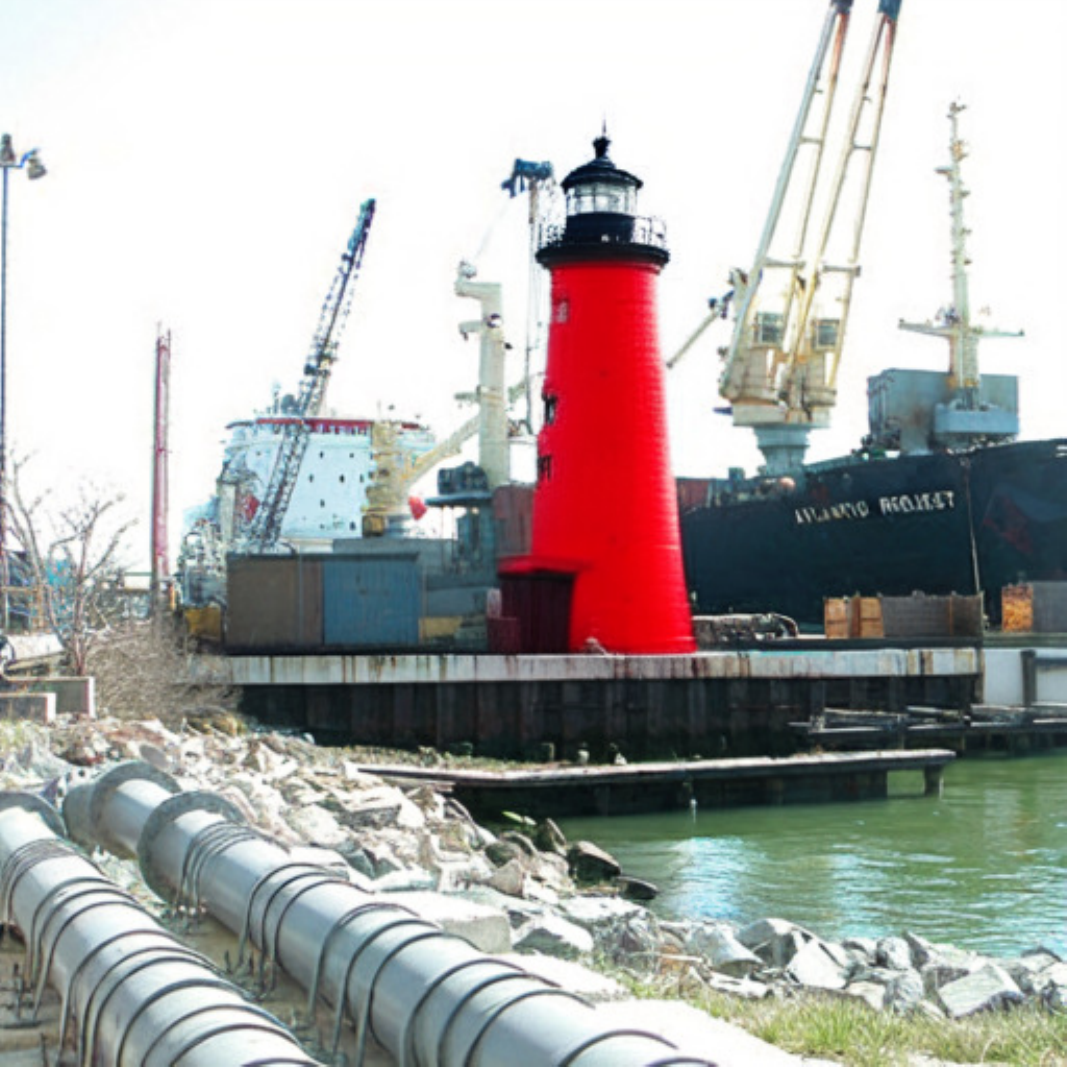}}&
\raisebox{-.5\height}{
\includegraphics[width=0.4\linewidth]{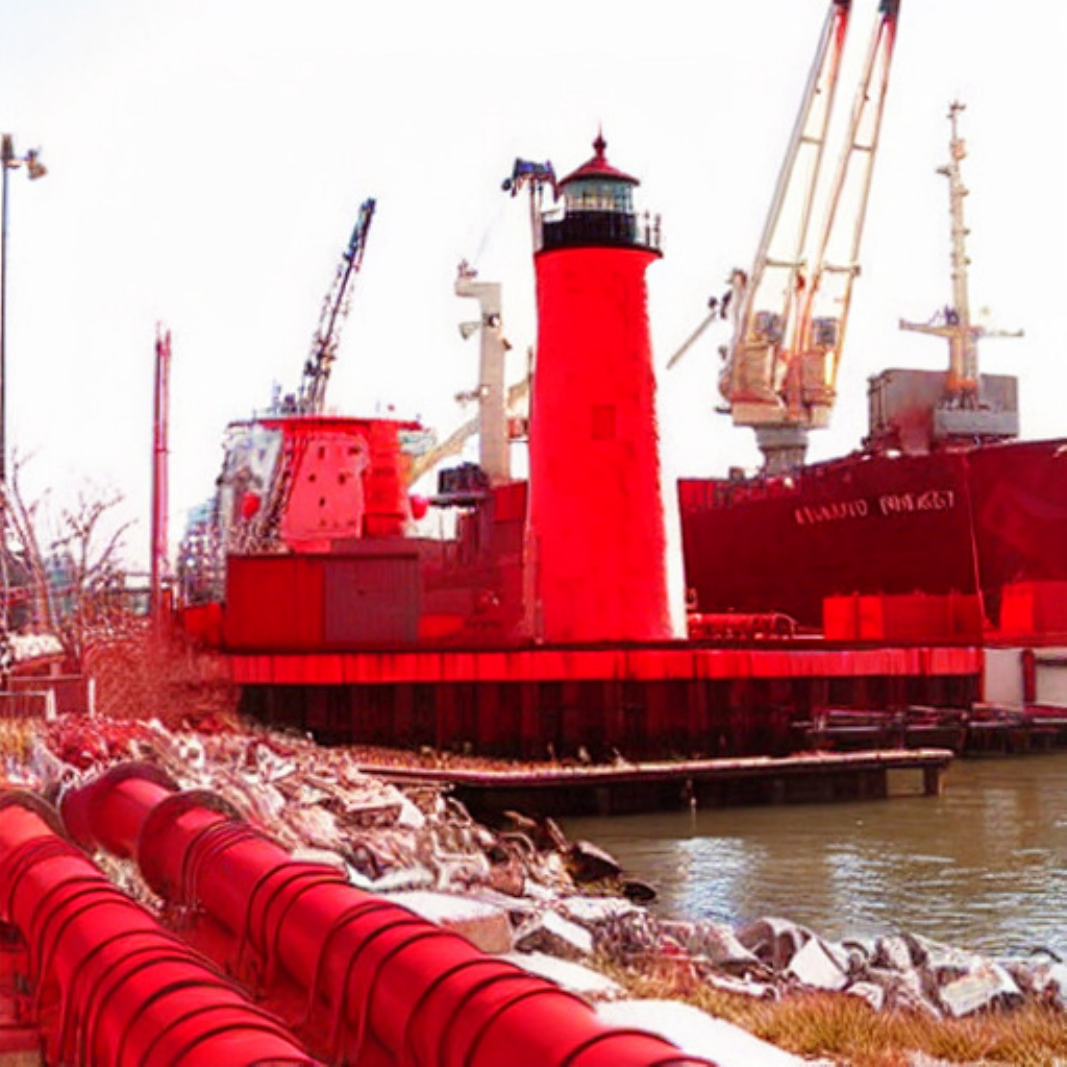}}&
\raisebox{-.5\height}{
\includegraphics[width=0.4\linewidth]{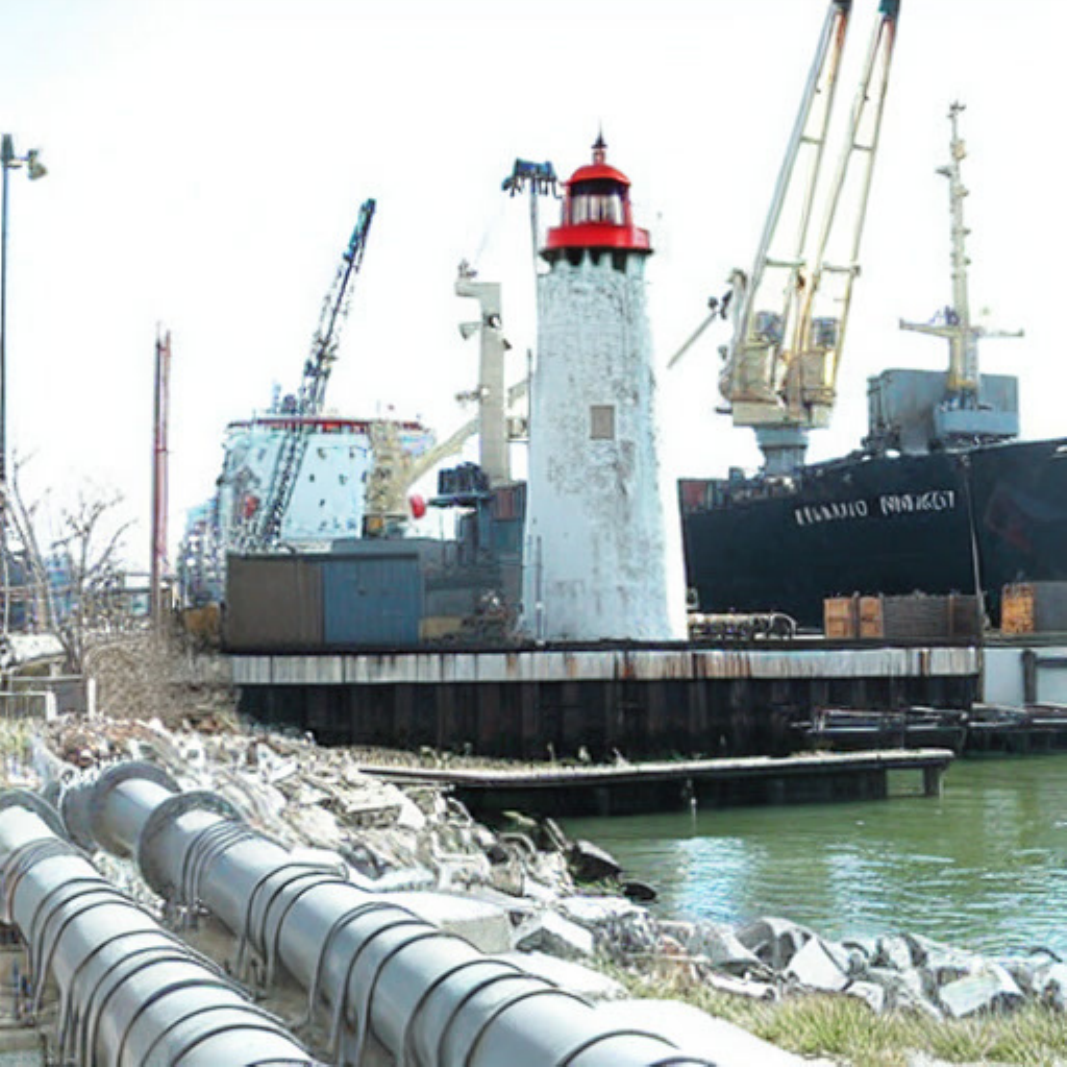}}&
\raisebox{-.5\height}{
\includegraphics[width=0.4\linewidth]{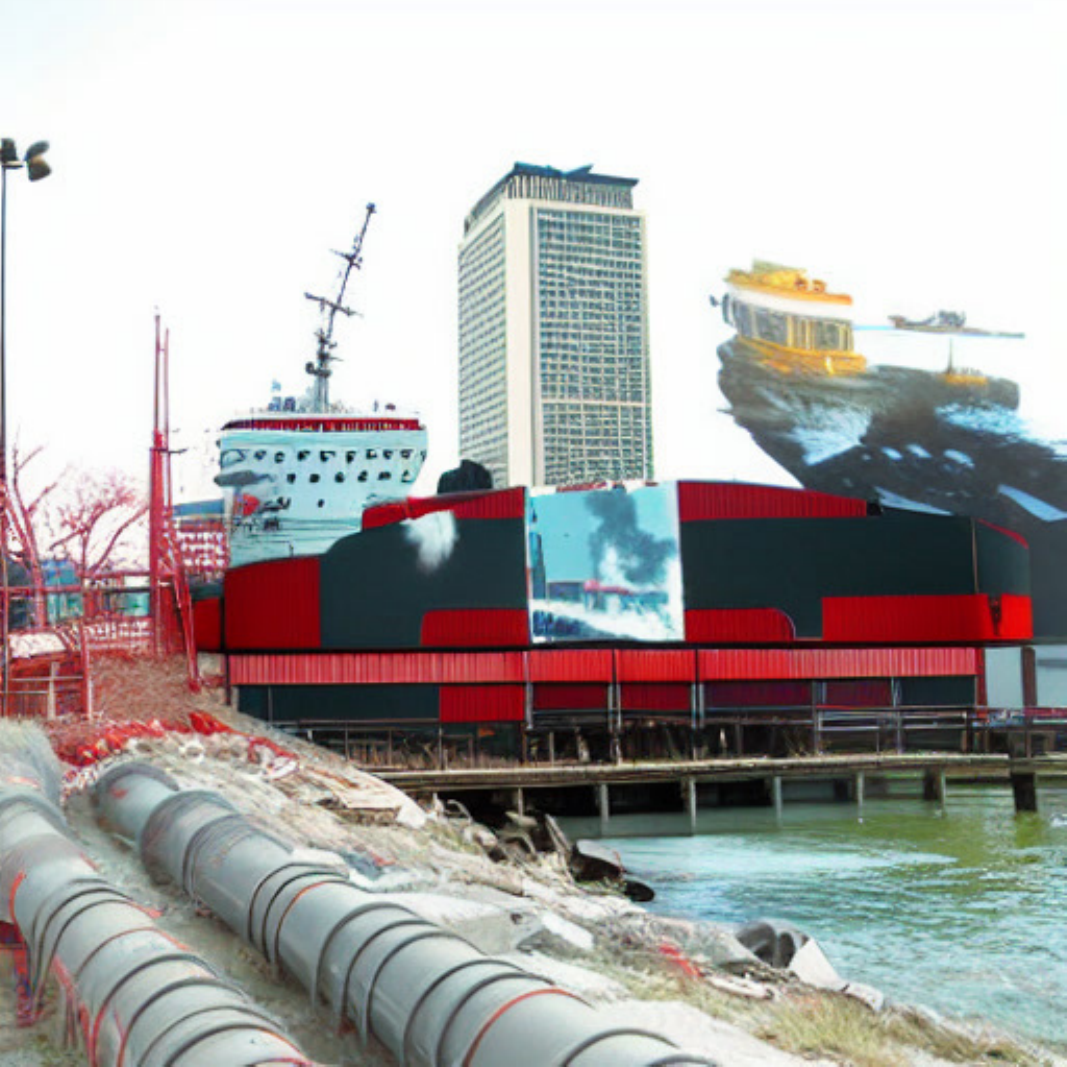}} \\

\resizebox{!}{30px}{
\begin{tabular}[x]{@{}c@{}} Remove the \\ curtains. \end{tabular}}&
\raisebox{-.5\height}{
\includegraphics[width=0.4\linewidth]{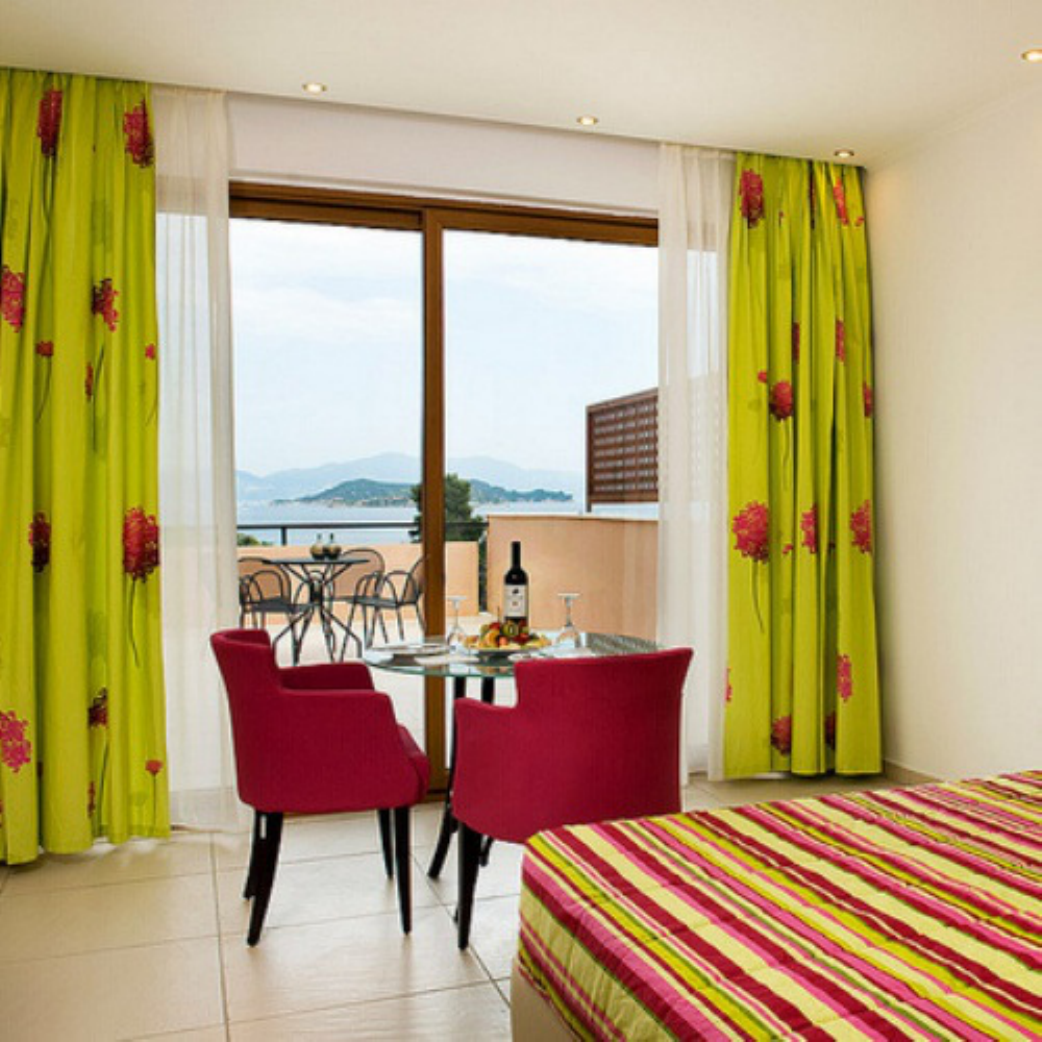}}&
\raisebox{-.5\height}{
\includegraphics[width=0.4\linewidth]{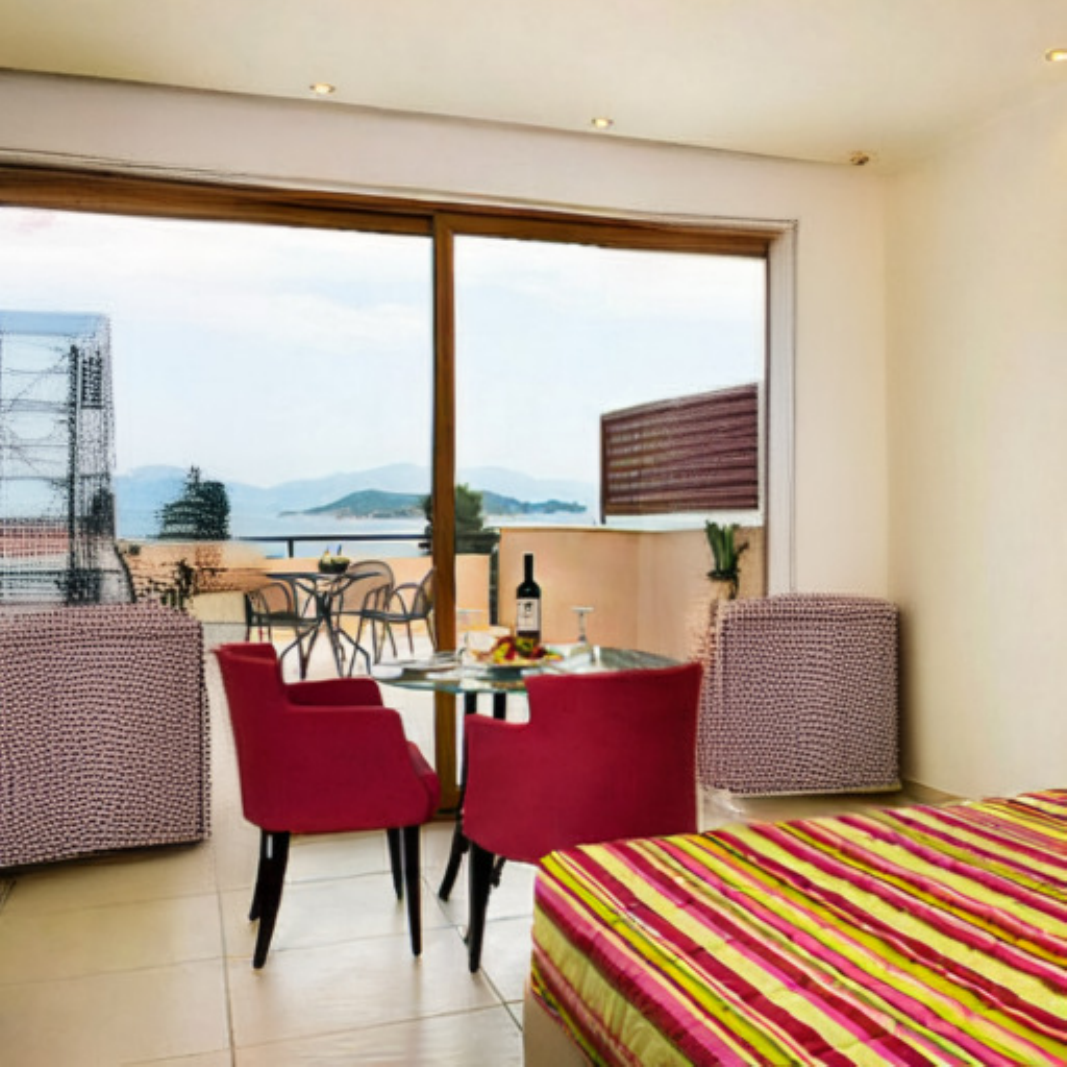}}&
\raisebox{-.5\height}{
\includegraphics[width=0.4\linewidth]{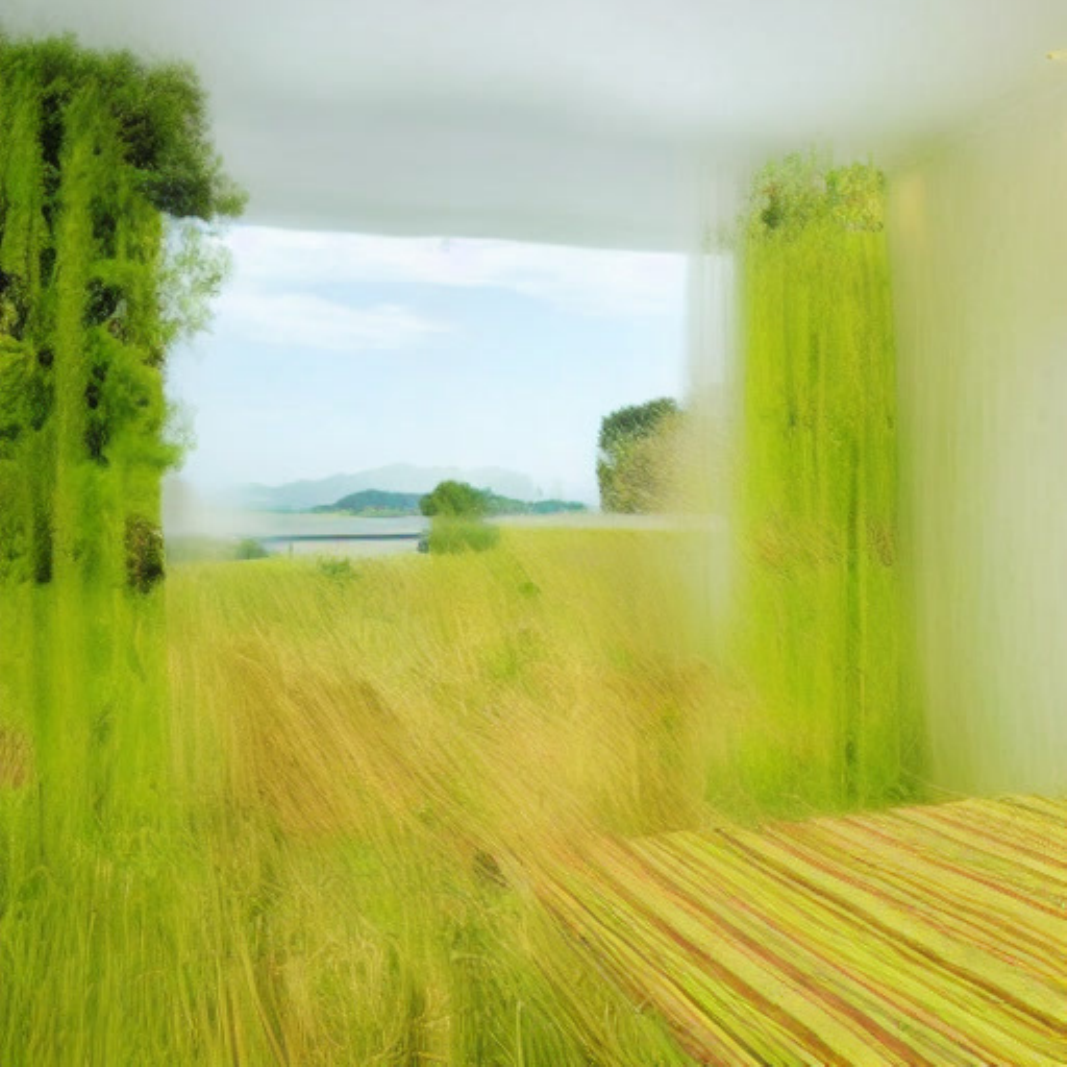}}&
\raisebox{-.5\height}{
\includegraphics[width=0.4\linewidth]{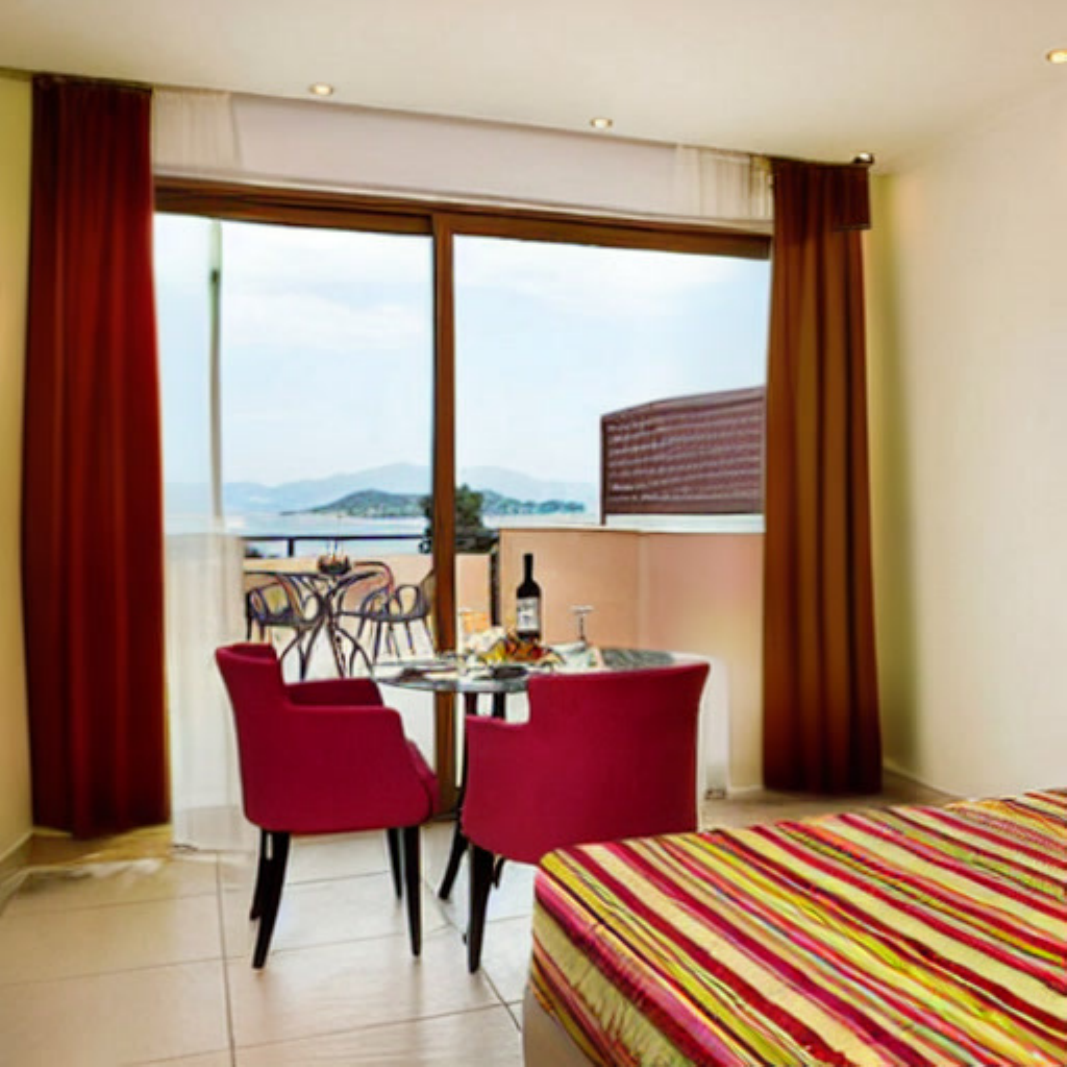}}&
\raisebox{-.5\height}{
\includegraphics[width=0.4\linewidth]{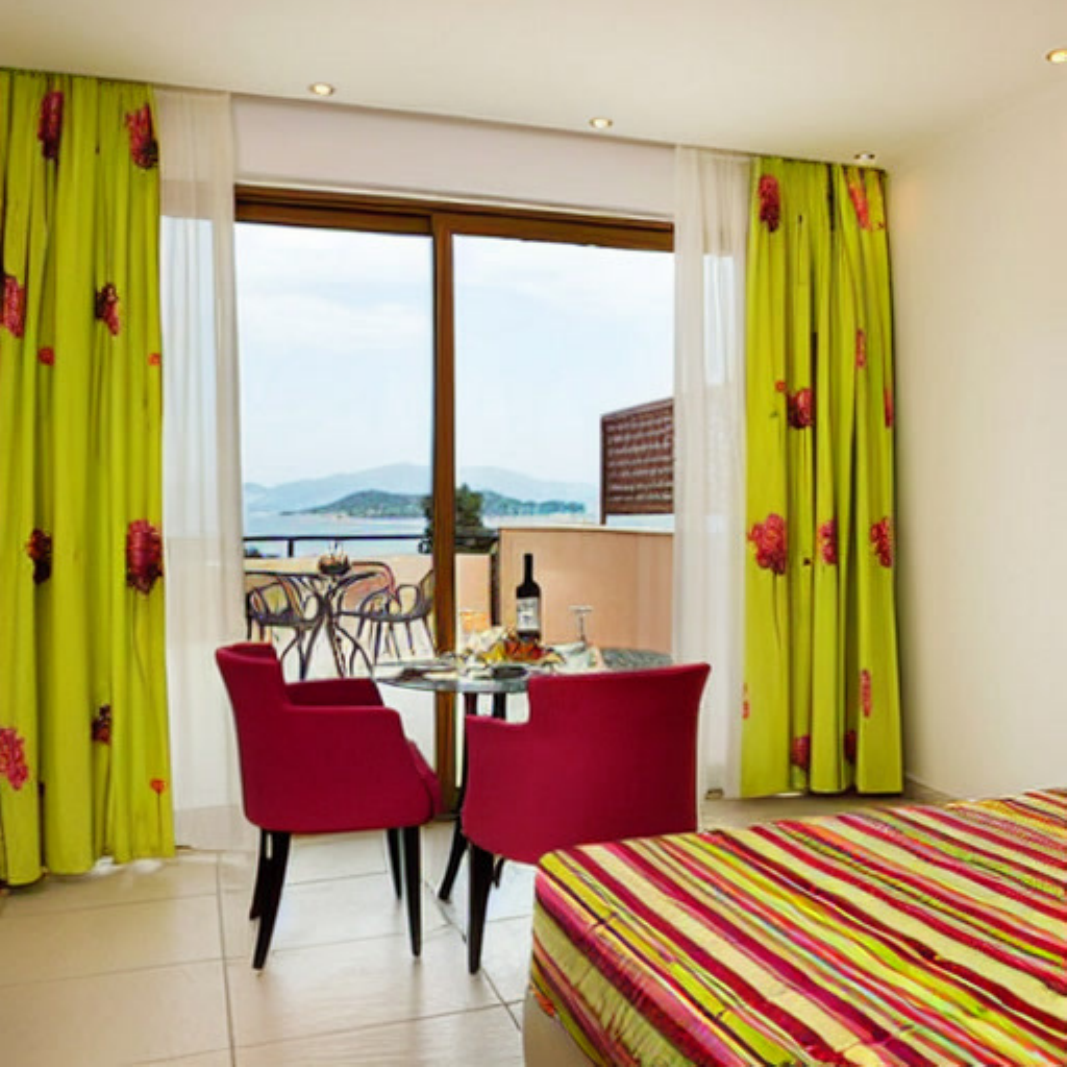}} \\

\resizebox{!}{40px}{
\begin{tabular}[x]{@{}c@{}} Add the word \\ 'sky' in white \\ to the sky. \end{tabular}}&
\raisebox{-.5\height}{
\includegraphics[width=0.4\linewidth]{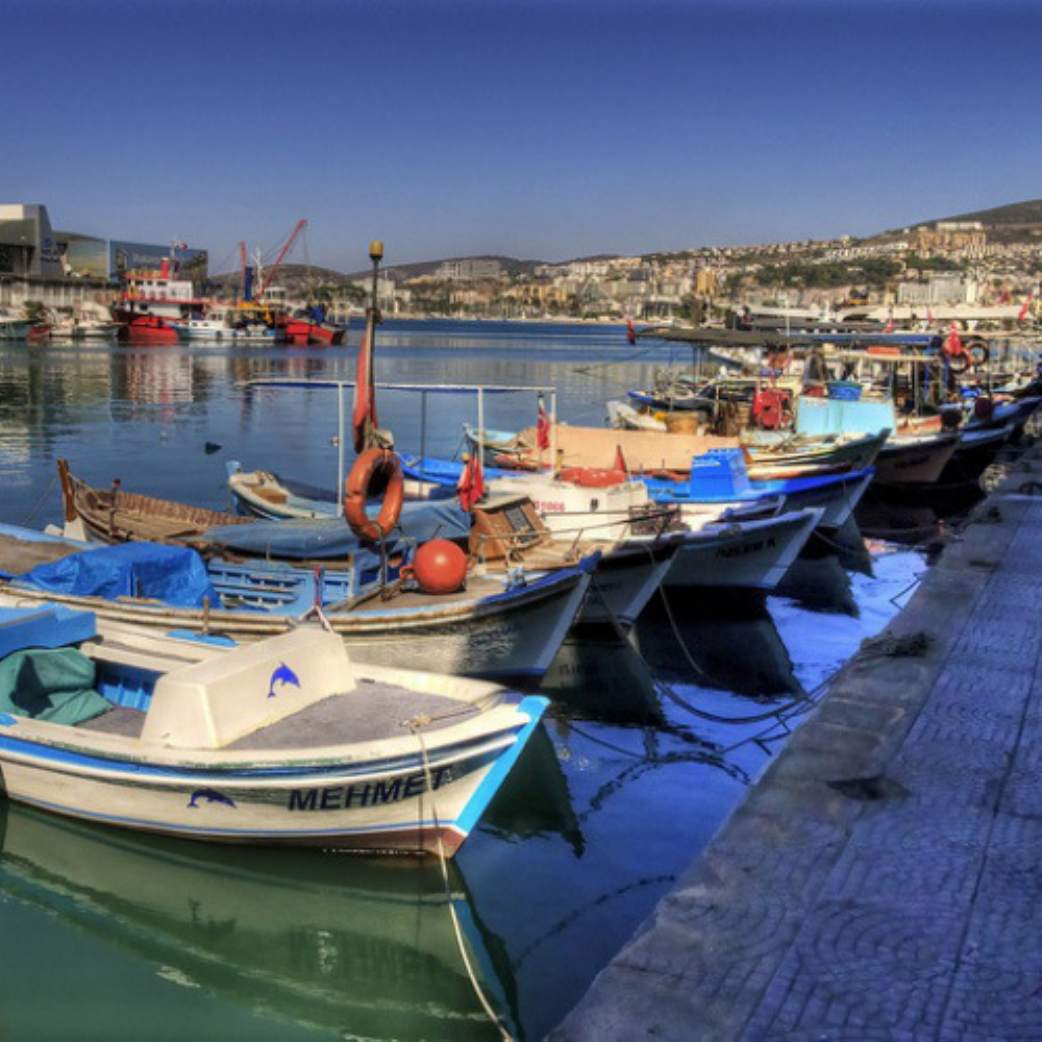}}&
\raisebox{-.5\height}{
\includegraphics[width=0.4\linewidth]{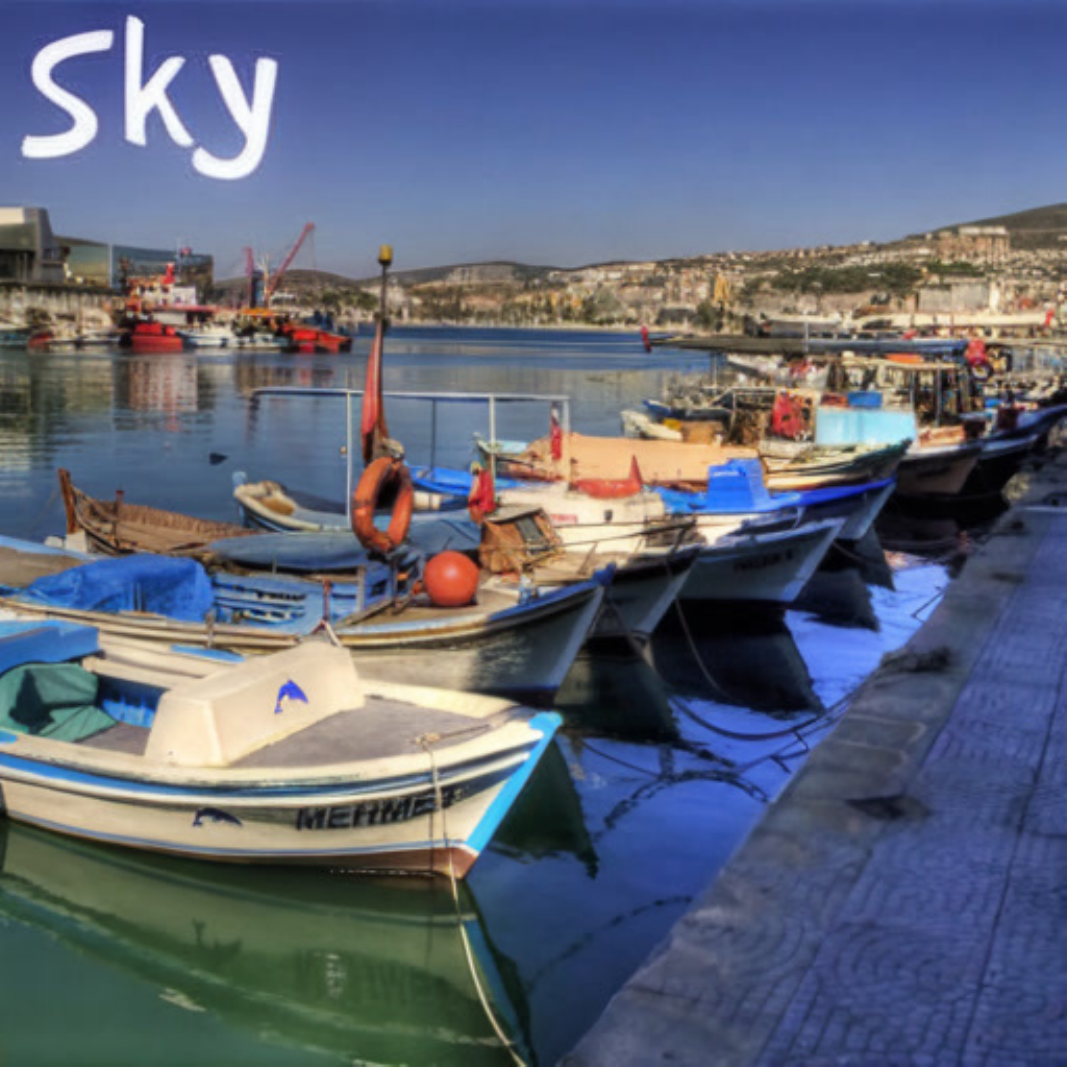}}&
\raisebox{-.5\height}{
\includegraphics[width=0.4\linewidth]{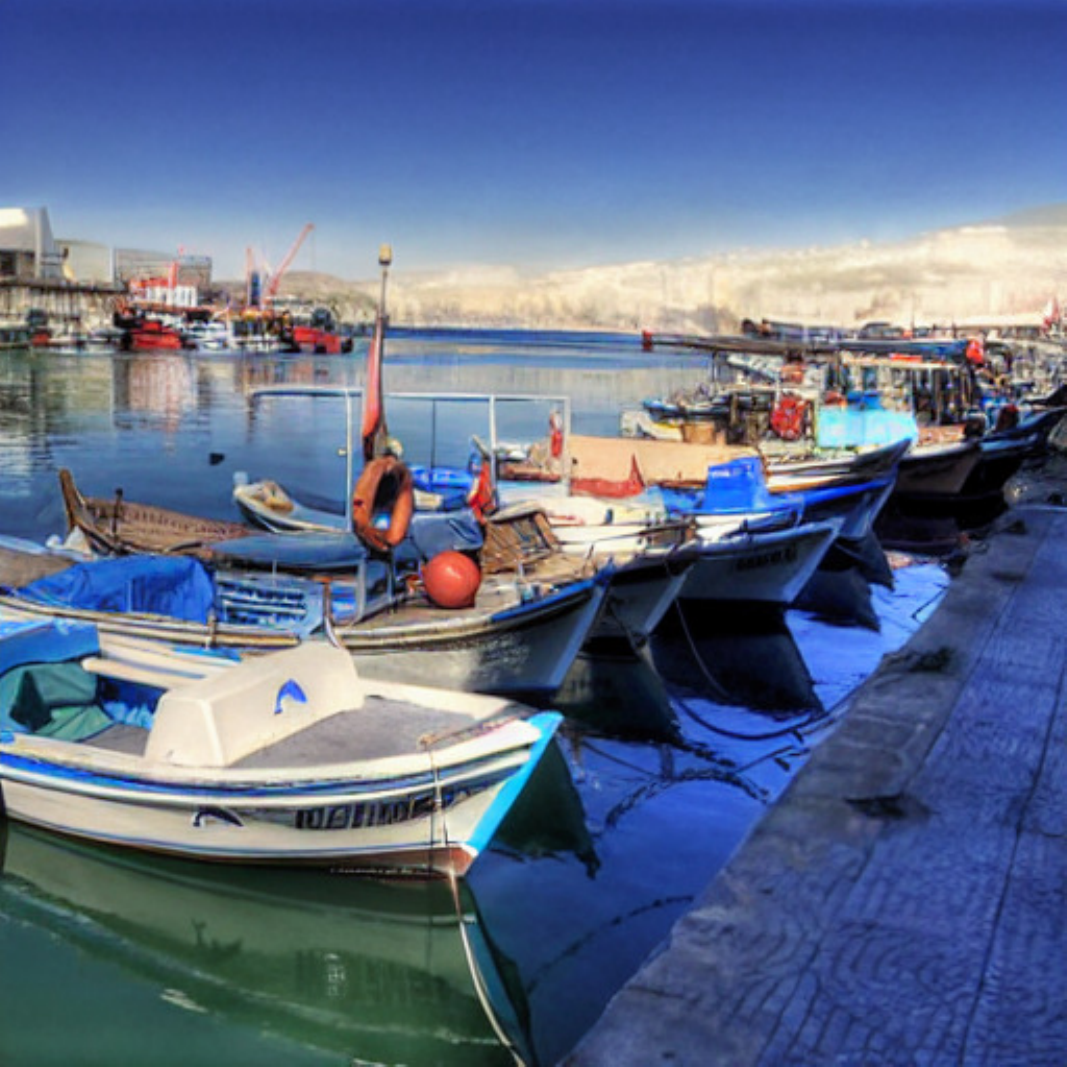}}&
\raisebox{-.5\height}{
\includegraphics[width=0.4\linewidth]{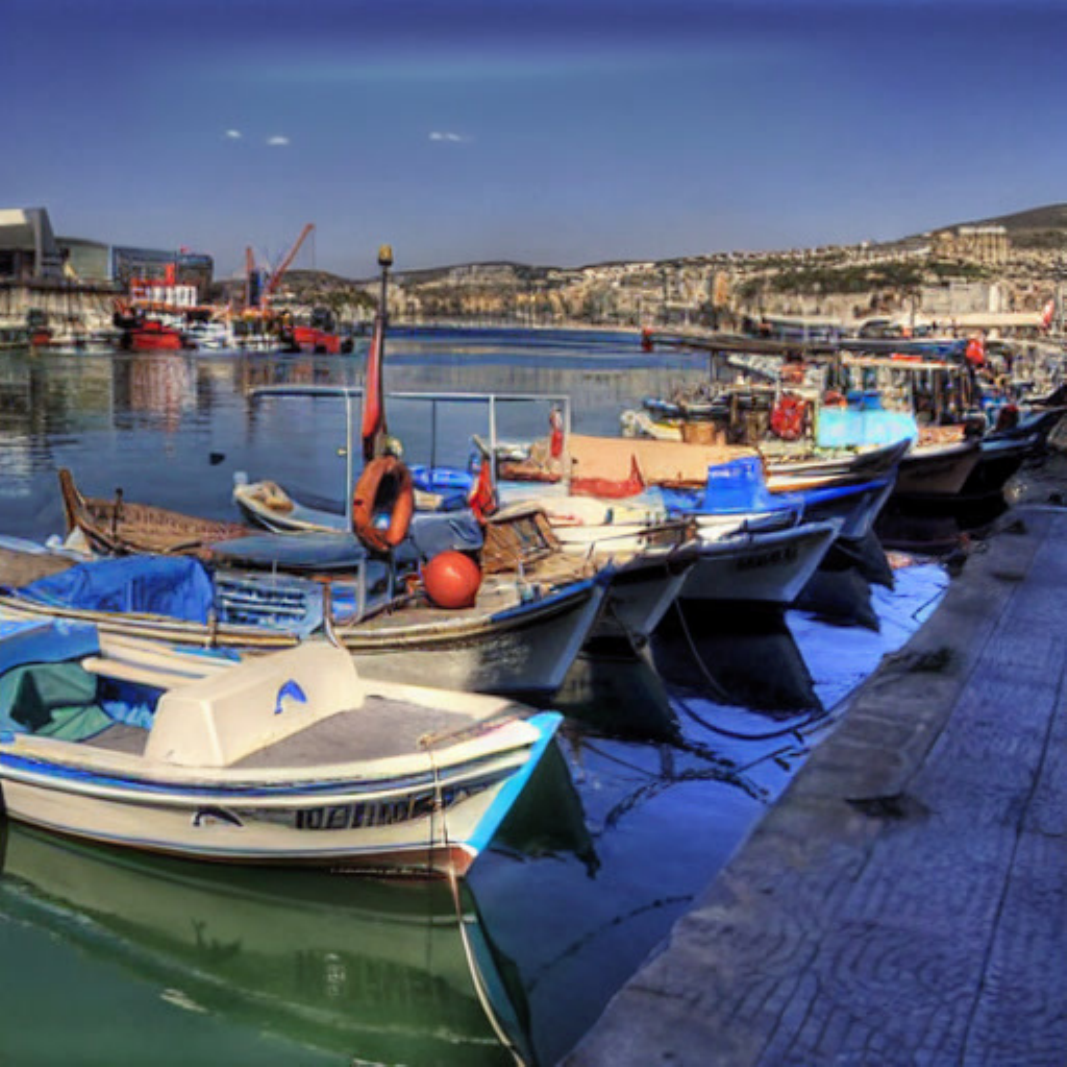}}&
\raisebox{-.5\height}{
\includegraphics[width=0.4\linewidth]{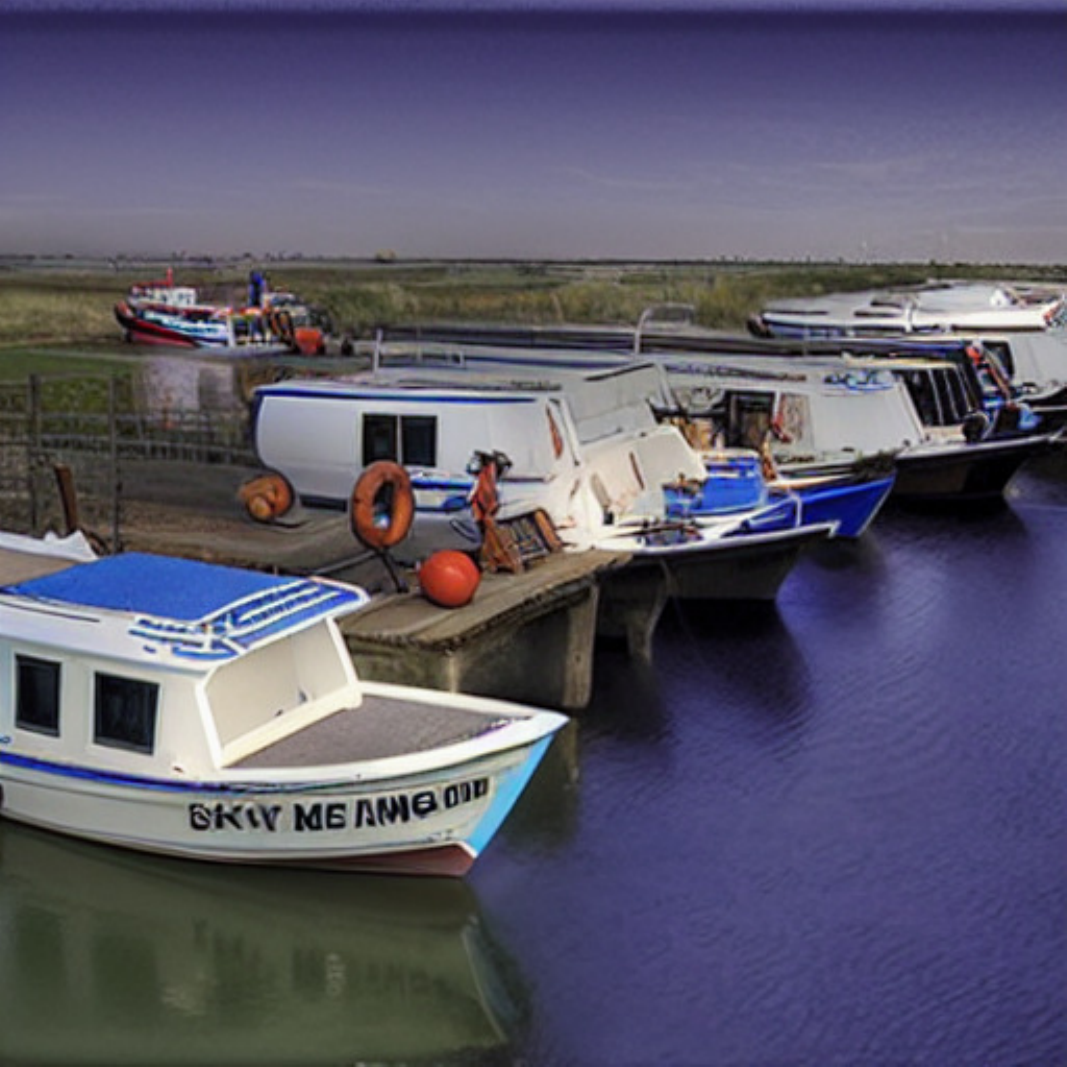}} \\

& \resizebox{!}{18px}{\begin{tabular}[x]{@{}c@{}}Original\end{tabular}} & 
\resizebox{!}{18px}{\begin{tabular}[x]{@{}c@{}} \textbf{\model} \end{tabular}} & 
\resizebox{!}{18px}{\begin{tabular}[x]{@{}c@{}}InstructPix2Pix \end{tabular}} &
\resizebox{!}{18px}{\begin{tabular}[x]{@{}c@{}} MagicBrush \end{tabular}} & 
\resizebox{!}{18px}{\begin{tabular}[x]{@{}c@{}} P2P \end{tabular}} \\
\end{tabular}
}
    \caption{Qualitative comparison of our model to baselines on \model Test Set.}
    \label{fig:comp_our_test2} %
\end{figure*}

\cref{fig:comp_our_images_sup,fig:comp_our_images_sup2} shows qualitative comparisons with baselines on generated samples. In addition, in \cref{fig:comp_our_test,fig:comp_our_test2} we present qualitative comparisons of our model with baselines on \model test set.

\subsection{Qualitative Comparison of Sequential Edit Thresholding}
\label{sec:multi_turn_qualitative_resutls}
In Fig.~\ref{fig:multiturn}, a qualitative comparison is provided to demonstrate the effectiveness of the proposed technique in maintaining image quality during multi-turn editing scenarios.
Specifically, we vary the value of the hyperparameter $\alpha$, which controls the degree to which pixel values are used during the editing process. With $\alpha=0$, no thresholding is applied and the output image is simply the result of passing the input image through the model. Conversely, when $\alpha=1$, the input image is used as the output image without any editing.
We present results for several values of $\alpha$, including $0.5$, $0.25$, $0.1$, $0.05$, $0.025$, $0.01$, and the baseline value of $0$.
As can be observed from the figure, when no clipping is applied (i.e., $\alpha=0$), artifacts tend to accumulate and manifest as general noise in the output image. On the other hand, applying a threshold helps preserve the quality of the output image even when multiple edit turns are applied. However, using a large value of $\alpha$ can interfere with the editing process and result in poor edit quality.
Based on these observations, we opt to use a value of $\alpha=0.03$ in our experiments, as it strikes a balance between preserving image quality and allowing for effective editing.

\section{Implementation Details}
\label{sec:impl_details}
We use a scaled-down version of \cite{emu} which is conditioned on CLIP ViT-L~\cite{radford2021learning} and T5-XL~\cite{2020t5}, and generates images at a resolution of $512 \times 512$. We adapt it to obtain image inputs by concatenating to the input channels following \cite{brooks2023instructpix2pix}. We condition on the text and task embeddings both through cross-attention and by addition to the timestep embeddings. 
For training, we employ the Adam optimizer with a batch size of 512. We use a learning rate of 2e-5 with a cosine decay schedule and a linear warmup of 2,000 iterations. The training spans 48,000 steps. %

\begin{figure*}
\lstset{backgroundcolor=\color{white},stringstyle=\color{black},language=python}
\begin{lstlisting}[label=cond-var-black]
def get_content_instruction(new_prompt):
    optional_verbs = choice(["include", "place", "position", "set", "incorporate", "alongside", "give", "put", "insert", "together with", "with", "make", "integrate", "have", "append", "make", "add", "include"])
    
    # system message #
    system_message = 
    f"<<SYS>>
    You are an assistant that only speaks JSON. Do not write normal text. The assistant answer is JSON with the following string fields: 'edit', 'edited object','output'. Here is the latest conversation between Assistant and User.
    <</SYS>>"

    # introduction message #
    intro_message =
    f"[INST]User: Hi, My job to take a given caption ('input') and to output the following: an instruction for {optional_verbs} an object to the image ('edit'), the object to {optional_verbs} ('edited object'), and the caption with the object ('output'). Please help me do it. I will give you the 'input', and you will help. When you reply, use the following format: {"edit": '<instruction>', 'edited object': '<object>', 'output': '<caption>'}[/INST]
    Assistant: Sure, I'd be happy to help! Please provide the actual input caption you'd like me to read and I'll assist you with writing an instruction to {optional_verbs} an object to the image, writing the added object and writing the caption with the object."


    # shuffling #
    random.seed(torch.randint(1 << 32, ()).item())
    shuffle(few_shot_examples)
    few_shot_examples = few_shot_examples[:int(len(few_shot_examples) * 0.6)]
    prompt = system + intro_message + "".join(<@\textcolor{blue}{few\_shot\_examples}@>)
    
    # add the test prompt #
    prompt = prompt + f"[INST]User: {new_prompt}[/INST]"

    return prompt
\end{lstlisting}
\caption{An example of in-context learning for the task of Add.}
\label{fig:llama_learning}
\end{figure*}

\begin{figure*}
\lstset{backgroundcolor=\color{white},stringstyle=\color{black},language=python}
\begin{lstlisting}[mathescape=true,label=cond-var-black-add]
<@\textcolor{blue}{few\_shot\_examples}@> = [
[INST]User: <@\textcolor{red}{"Beautiful cat with mojito sitting in a cafe on the street"}@>[/INST]
Assistant: {<@\textcolor{red}{"edit": "include a hat", "edited object": "hat", "output": "Beautiful cat wearing a hat with mojito sitting in a cafe on the street"}@>}
[INST]User:  <@\textcolor{red}{"robot playing chess at home."}@>[/INST]
Assistant: {<@\textcolor{red}{"edit": "add a cheerful smiling face.", "edited object": "robot", "output": "robot playing chess at home with a cheerful smiling face."}@>}
[INST]User:  <@\textcolor{red}{"A cute creature sits at the beach."}@>[/INST]
Assistant: {<@\textcolor{red}{"edit": "set a dog besides the creature", "edited object": "dog", "output": "A cute creature and a dog sit at the beach."}@>}
[INST]User:  <@\textcolor{red}{"Superhero on the street in sunny day working on his tablet."}@>[/INST]
Assistant: {<@\textcolor{red}{"edit": "put a vintage tie on the superhero.", "edited object": "tie", "output": "Superhero with a vintage tie on the street in sunny day working on his tablet."}@>}
[INST]User:  <@\textcolor{red}{"Picture clouds, birds, the wind, foliage, rainbow, hill, art, pair, guy"}@>[/INST]
Assistant: {<@\textcolor{red}{"edit": "together with a dog on the left", "edited object": "dog", "output": "Picture clouds, birds, the wind, foliage, rainbow, hill, art, pair, guy, dog on the left"}@>}
[INST]User:  <@\textcolor{red}{"horse on a red Boat Near Mountains During Golden Hour"}@>[/INST]
Assistant: {<@\textcolor{red}{"edit": "give the horse sunglassess", "edited object": "sunglassess", "output": "horse with sunglassess on a red Boat Near Mountains During Golden Hour"}@>}
[INST]User:  <@\textcolor{red}{"An animal family on studio background."}@>[/INST]
Assistant: {<@\textcolor{red}{"edit": "make them hold a teddy bear.", "edited object": "teddy bear", "output": "An animal family holding a teddy bear on studio background."}@>}
[INST]User:  <@\textcolor{red}{"Baked Salmon With Bell Peppers"}@>[/INST]
Assistant: {<@\textcolor{red}{"edit": "insert kale pesto to the dish", "edited object": "kale pesto", "output": "Baked Salmon With Kale Pesto And Bell Peppers"}@>}
[INST]User:  <@\textcolor{red}{"An airplaine is flying in the sky in rainy day."}@>[/INST]
Assistant: {<@\textcolor{red}{"edit": "add flowers in the windows", "edited object": "flowers", "output": "An airplaine with flowers in the windows is flying in the sky in rainy day."}@>}
[INST]User:  <@\textcolor{red}{"photo of mountains and trees"}@>[/INST]
Assistant: {<@\textcolor{red}{"edit": "position a castle between the trees", "edited object": "castle", "output": "photo of mountains, trees and castle between the trees"}@>}
[INST]User: <@\textcolor{red}{"Little bunny in the park"}@>[/INST]
Assistant: {<@\textcolor{red}{"edit": "Make the bunny play with a kite.", "edited object": "kite", "output": "Little bunny playing with a kite in the park"}@>}
[INST]User: <@\textcolor{red}{"Attic Bedroom With Large Ceilings"}@>[/INST]
Assistant: {<@\textcolor{red}{"edit": "decorate the room with beautiful chandeliers", "edited object": "chandeliers", "output": "Attic Bedroom With Beautiful Chandeliers on Large Ceilings"}@>}
[INST]User: <@\textcolor{red}{"Wedding rings and yellow flower on a red background"}@>[/INST]
Assistant: {<@\textcolor{red}{"edit": "place a bird by the yellow flower", "edited object": "bird", "output": "Wedding rings, a bird, and yellow flower on a red background"}@>}
[INST]User: <@\textcolor{red}{"Tree Near the lake in the morning"}@>[/INST]
Assistant: {<@\textcolor{red}{"edit": "Give it autumn leaves on top", "edited object": "leaves", "output": "Tree with autumn leaves on top Near the lake in the morning"}@>}
[INST]User: <@\textcolor{red}{"robot and alien sitting on hanging bridge at daytime"}@>[/INST]
Assistant: {<@\textcolor{red}{"edit": "make them hold three books.", "edited object": "three books", "output": "robot and alien holding three books while sitting on hanging bridge at daytime"}@>}
[INST]User: <@\textcolor{red}{"Skogafoss waterfall in the south of Iceland"}@>[/INST]
Assistant: {<@\textcolor{red}{"edit": "Set a colorful rainbow in the backhground!", "edited object": "rainboe", "output": "Skogafoss waterfall with a colorful rainbow in the south of Iceland"}@>}
[INST]User: <@\textcolor{red}{"Polar Bear with rubber gloves pushing shopping carts"}@>[/INST]
Assistant: {<@\textcolor{red}{"edit": "Make it wear a coat", "edited object": "coat", "output": "Polar Bear with a coat pushing shopping carts"}@>}
]
\end{lstlisting}

\caption{Examples of prompts for Add task.}
\label{fig:llama_prompting}
\end{figure*}

\clearpage

\clearpage

\end{document}